# La representación de la variación contextual mediante definiciones terminológicas flexibles

Antonio San Martín Pizarro

## Abstract

Definitions are one of the most important components of any high-quality terminological resource as well as a privileged medium for knowledge representation since they offer a direct natural-language explanation of the content of a concept. The adequacy of the definitions thus largely determines the overall usefulness of the terminological resource for the user.

This study has been motivated by the observation that terminological definitions often do not meet the needs of users. In this doctoral thesis, we apply premises of cognitive linguistics to terminological definitions and present a proposal called the flexible terminological definition. This consists of a set of definitions of the same concept made up of a general definition, in this case one encompassing the entire environmental domain, along with additional definitions describing the concept from the perspective of the subdomains in which it is relevant.

The main objective of this thesis is to analyze the effects of contextual variation on specialized environmental concepts with a view to their representation in terminological definitions. To accomplish this objective, we conducted an empirical study consisting of the analysis of a set of contextually variable concepts and the creation of a flexible definition for two of them.

As a result of the first part of the empirical study, we divided domain-dependent contextual variation into three different phenomena: modulation, perspectivization, and subconceptualization. In the second part, we applied these notions to terminological definitions and presented guidelines on how to build flexible definitions, from the extraction of knowledge to the actual writing of the definition.

This thesis contributes to the improvement of the quality of terminological definitions because, with this approach, the user is presented with a definition tailored to the domain that they have chosen. Furthermore, flexible terminological definitions provide a knowledge representation that better resembles the human conceptual system than traditional terminological definitions. This thesis is written in Spanish, except for a long summary, the results, and the conclusions, which are written in English.

# LA REPRESENTACIÓN DE LA VARIACIÓN CONTEXTUAL MEDIANTE DEFINICIONES TERMINOLÓGICAS FLEXIBLES

Antonio San Martín Pizarro

**tesis doctoral**

PROGRAMA OFICIAL DE POSGRADO
EN ESTUDIOS AVANZADOS
DE TRADUCCIÓN E INTERPRETACIÓN

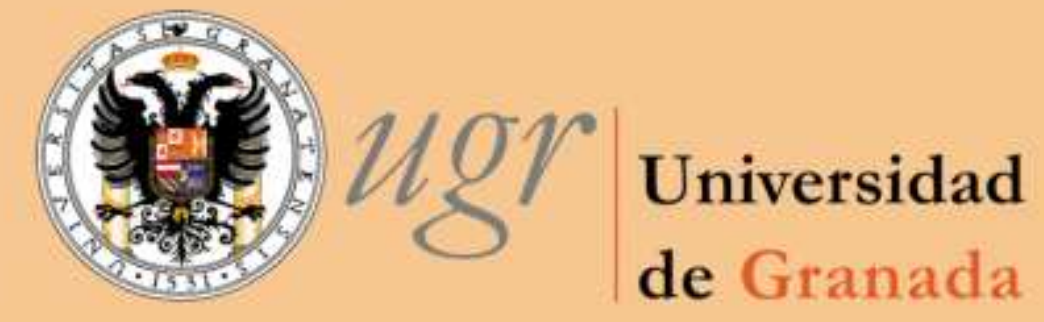

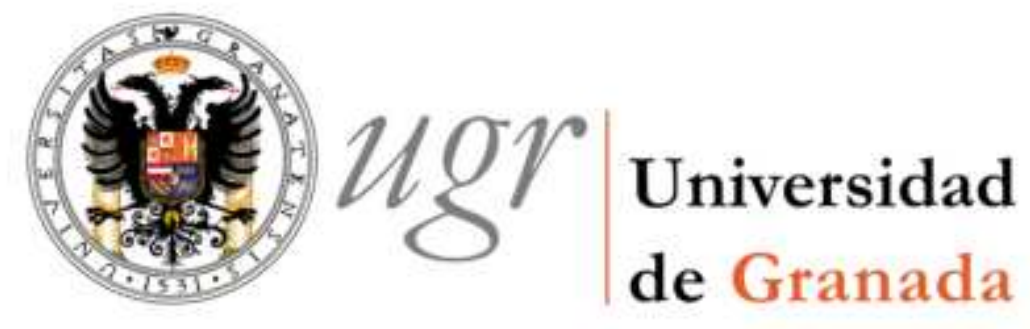

tesis doctoral

# LA REPRESENTACIÓN DE LA VARIACIÓN CONTEXTUAL MEDIANTE DEFINICIONES TERMINOLÓGICAS FLEXIBLES

Antonio San Martín Pizarro

Directoras: Pamela Faber Benítez
Pilar León Araúz

DEPARTAMENTO DE TRADUCCIÓN E INTERPRETACIÓN
PROGRAMA OFICIAL DE POSGRADO EN ESTUDIOS AVANZADOS DE TRADUCCIÓN E INTERPRETACIÓN

El doctorando, Antonio San Martín Pizarro, y las directoras de la tesis, Pamela Faber Benítez y Pilar León Araúz, garantizamos, al firmar esta tesis doctoral, que el trabajo ha sido realizado por el doctorando bajo la dirección de las directoras de la tesis y, hasta donde nuestro conocimiento alcanza, en la realización del trabajo, se han respetado los derechos de otros autores a ser citados, cuando se han utilizado sus resultados o publicaciones.

Granada, 9 de noviembre de 2015

| Fdo.: Pamela Faber Benítez | Fdo.: Pilar León Araúz | Fdo.: Antonio San Martín Pizarro |
|---|---|---|
| Directora de la tesis | Directora de la tesis | Doctorando |

*En la medida en que la palabra "conocimiento" tiene algún sentido, el mundo es cognoscible; pero también es interpretable. No hay un único significado, sino multitud de significados.*

Friedrich Nietzsche

# AGRADECIMIENTOS

Esta tesis doctoral no habría sido posible sin la excelente orientación de mis directoras, Pamela Faber y Pilar León. Deseo agradecer la confianza depositada en mí, el esfuerzo dedicado a la dirección de esta tesis y, sobre todo, la amistad, que siempre han puesto por encima de todo. A pesar de la distancia física en la última etapa de la tesis, han sabido guiarme hasta el final del camino. Me siento extraordinariamente afortunado de haber estado bajo su dirección.

También doy las gracias al resto de miembros del grupo de investigación LexiCon, con quienes he compartido tan buenos momentos y de quienes he aprendido tanto: Miriam Buendía, Beatriz Sánchez, Arianne Reimerink, Maribel Tercedor, Clara Inés López, Ana Belén Pelegrina, José Manuel Ureña, Pedro Magaña y Alejandro García.

Deseo agradecer también a Marie-Claude L'Homme, quien generosamente dirigió mi estancia de investigación en la Universidad de Montreal, siempre ha estado dispuesta a responder mis dudas y me ha transmitido su pasión por la investigación. Asimismo, doy las gracias a Benoît Robichaud, quien puso a mi disposición desinteresadamente sus conocimientos en programación para facilitarme el análisis de los datos de esta tesis. Asimismo, quiero agradecer a Patrick Drouin, François Lareau, Marjan Alipour, Mariana Botta, Jérémie Zhichao Jia, Florie Lambrey, Daphnée Azoulay, Gabriel Bernier-Colborne, Alicia Vico y Nizar Ghazzawi por hacer que mi paso por el Observatoire de linguistique Sens-Texte haya sido inolvidable.

Estoy igualmente agradecido a Daniel Williams, que resolvió mis dudas sobre química, y a Selja Seppälä, por su ayuda en la etapa final de la tesis doctoral.

Esta tesis tampoco habría sido posible sin el apoyo y cariño de mis amigos tanto de España (en particular Rocío Martínez, Esther Navarro, Eduardo Álvarez, Jesús Moreno, Beatriz Muñoz, Manuel Adán Bonald, Daniel

Pacheco, Benjamín Porto, Antonio Cossio, Juan Carlos Navarro, Luis Ortega y Agustín Otazo) como de Quebec (especialmente Anaïs Tatossian, Hicham González, Martin Harvey, Richard Cloutier, Rita Palacios, Jean Pichette, Michel Usereau y Manuel Meune).

Por último, dedico esta tesis doctoral a mis padres, Antonio y Teresa, y a mi hermano, Daniel. Gracias por el amor y apoyo incondicionales.

# ÍNDICE DE FIGURAS

# ÍNDICE DE TABLAS

# ABREVIATURAS

AGR: agronomía / agronomy

AIR: gestión de la calidad del aire / air quality management

ATM: ciencias atmosféricas / atmospheric sciences

AWL: academic word list

BCT: base de conocimiento terminológica

BIO: biología / biology

CDF: categoría derivada de un fin

CED: Collins English Dictionary

CEFLD: *Cobuild English for Learners Dictionary*

CEN: ingeniería química / chemical engineering

CHE: química / chemistry

CIV: ingeniería civil / civil engineering

DEC: *Dictionnaire explicatif et combinatoire du français contemporain*

DLE: *Diccionario de la lengua española*

DOEAC: *A Dictionary of Environment and Conservation*

ENE: ingeniería energética / energy engineering

ENV: medio ambiente general / general environment (subcorpus)

GE: general environmental

GEO: geología / geology

HYD: hidrología / hydrology

LCT: léxico científico transdisciplinar

LDOCE: *Longman Dictionary of Contemporary English*

LEC: lexicología explicativa y combinatoria

PHY: física / physics

SOI: ciencias del suelo / soil sciences

STL: scientific transdisciplinary lexicon

TBM: terminología basada en marcos

TCT: teoría comunicativa de la terminología

TFL: teoría funcional de la lexicografía

TGT: teoría general de la terminología

TSC: terminología sociocognitiva

TSSP: teoría de los sistemas de símbolos perceptuales

VGOS: vocabulaire général d'orientation scientifique

WAS: gestión de residuos / waste management

WAT: abastecimiento y tratamiento de aguas / water treatment and supply

WOS: way-of-seeing

# SUMMARY

Definitions are one of the most important components of any high-quality terminological resource as well as a privileged medium for knowledge representation since they offer a direct natural-language explanation of the content of a concept. The adequacy of the definitions thus largely determines the overall usefulness of the terminological resource for the user.

This study has been motivated by the observation that terminological definitions often do not meet the needs of users. This is partly due to certain preconceptions about the purpose of definitions as well as the nature of meaning itself. These preconceptions thus affect how terminological definitions are created.

Traditionally, defining a term has been seen as stating the necessary and sufficient characteristics that make up the meaning of the term. This approach, known as the Aristotelian definition (§3.2), presupposes the existence of a stable meaning, independent of the context in which the term is used. In addition, it is assumed that meaning (or semantic knowledge) is independent of world knowledge (or encyclopedic knowledge).

The key premises on which the traditional approach to definitions is based have been refuted in the field of cognitive linguistics (§2.1): i) it is not possible to determine the necessary and sufficient features of a concept because conceptual boundaries are fuzzy; ii) concepts have prototypical features not shared by all members of the category; iii) it is not possible to make a distinction between semantic and encyclopedic knowledge, nor between semantic and pragmatic knowledge.

From a cognitive point of view, encyclopedic knowledge plays a central role in the study of meaning because concepts always appear embedded in frames, which are structures based on encyclopedic knowledge which attribute sense to concepts (Fillmore 1982a). Moreover, meaning is not considered a stable entity. It is constructed in each usage event in accordance with the context (§2.1.2). As a consequence, meaning and context are inseparable.

In this doctoral thesis, we apply these premises of cognitive linguistics to terminological definitions and present a proposal called the *flexible terminological definition* (§3.6). This consists of a set of definitions of the same concept made up of a general definition (in this case, one encompassing the entire environmental domain) along with additional definitions describing the concept from the perspective of the subdomains in which it is relevant.

Our proposal assumes that by eliminating the artificial boundaries between semantic and pragmatic knowledge, the representation of contextual variation in the terminological definition will no longer be a mere possibility. Given the ubiquity of context and its effects on cognition and language, the representation of the traits activated by concepts in accordance with the context becomes a necessity if one aspires to fully meet the user's needs. This also entails the inclusion of prototypical characteristics in the definition, i.e. characteristics that are not always applicable to the concept, but which are relevant in a given context.

Similarly, encyclopedic knowledge in the terminological definition is no longer forbidden. It now forms an integral part of the definition. The role that the defined concept plays in the frames it activates should be, as far as possible, part of the definition.

Our proposal is specifically based on frame-based terminology (§2.2), in addition to the theories of grounded cognition (§2.1.1), frame semantics (§2.1.2.2), prototype theory (§2.1.3.2), and the theory theory (§2.1.3.3).

We took as a starting point the application of frame-based terminology to the representation of specialized knowledge and to terminological definitions in EcoLexicon (a terminological knowledge base on the environment created in accordance with frame-based terminology). In fact, our proposal for a flexible definition was inspired by the recontextualization of EcoLexicon (§3.6.1), as a result of which conceptual maps only show the relevant information for the subdomain of the environment chosen by the user. Recontextualization in EcoLexicon represents contextual variation and avoids information overload, thus increasing knowledge acquisition by the users.

EcoLexicon follows the principle proposed by Meyer, Bowker, and Eck (1992: 159) according to which, for a terminological knowledge base to be truly useful, it must reflect the same conceptual organization as in the human brain. Since terminological definitions are a kind of knowledge representation (Faber 2002), in this doctoral thesis, we assume that the creation of terminological definitions should also be based on the organization of the human conceptual system.

Since context is a determining factor in the construction of the meaning of lexical units (including terms), we assume that terminological definitions can, and should, reflect the effects of context, even though definitions have traditionally been treated as the expression of meaning void of any contextual effect.

The main objective of this thesis is to analyze the effects of contextual variation on specialized environmental concepts with a view to their representation in terminological definitions. Specifically, we focused on contextual variation based on thematic restrictions (§3.5.3.4), i.e. how the different areas of knowledge comprising the vast domain of the environment conceptualize differently the same concepts, and how this can be reflected in the definition. One of the main fundamentals of our proposal is the notion in cognitive linguistics that lexical units only have meaning in real use events (§2.1.2). Outside of any use event, a lexical unit does not have any meaning, only semantic potential.

A term's semantic potential is the raw material for its definition, not its object. The semantic potential is not the object because this would mean that defining a term would involve describing all the conceptual content that the term could activate. This is not viable since a term's semantic potential corresponds to a vast, immeasurable quantity of information that is never fully activated in real events.

Lexical units have not only semantic potential, but also associated conventional and contextual constraints. These constraints cause some conceptual content to be activated more often than others, giving rise to what Croft and Cruse (2004: 110) call *premeanings*. Pre-meanings are conceptual units that appear between the semantic potential and the meaning in the conceptualization process. The object of the definition is thus a subset of the semantic potential that corresponds to a pre-meaning.

The pre-meaning that becomes the object of a given definition depends on the contextual constraints applied to the definition. In all cases, this subset always corresponds to a portion of a single concept and the frames that it can activate. By contextual constraints, we mean any situational factors that affect meaning construction and, indirectly, the content of terminological definitions.

Given that the object of the definition (the pre-meaning) is an abstraction of the meanings that a lexical unit has under certain contextual constraints, we can state that the context associated with a pre-meaning is also a sort of abstraction from real contexts. As a consequence, we gave the name *pre-context* to the set of contextual restrictions that limit the semantic potential of a lexical unit in a relatively predictable way, giving rise to pre-meanings.

Context comprises the linguistic context, discursive context, sociocultural context, and spatial-temporal context. The pre-context for terminological definitions includes linguistic constraints, thematic constraints, cultural constraints, ideological constraints, and diachronic constraints.

Thematic constraints (i.e. discourse topic) allow for more accurate predictions about the way that the semantic potential of a given lexical unit is restricted than other contextual factors. Thematic constraints reduce the semantic potential of a lexical unit according to the topic at issue during a communicative act and the point of view taken. Our proposal of a flexible terminological definition relies on these types of constraint.

Domains, in terms of a knowledge field, allow for the systematic characterization of thematic constraints in terminological definitions. They can be understood as macroframes that guide knowledge organization and categorization in a given conceptual area. In this doctoral thesis, we have used a simplified version of the domain classification that was created specifically for EcoLexicon.

This work focuses on the phenomenon of contextual variation as opposed to lexical ambiguity (polysemy and homonymy). Contextual variation is the phenomenon that occurs when a concept does not always activate the same traits in use events and the relevance of these traits varies. For its part, lexical ambiguity is the phenomenon that occurs when a lexical unit is associated with more than one concept (Cruse 2011: 100).

To accomplish the objectives of this doctoral thesis, we conducted an empirical study (§5) consisting of the analysis of a set of contextually variable concepts and the creation of a flexible definition for two of them. Each of these two concepts presented different contextual profiles.

To select the concepts to be analyzed, a terminological extraction was performed on 14 corpora of different environmental subdomains, specifically compiled for this doctoral thesis (§4.1.2.1). The extraction was limited to simple nouns, and the results were compared so as to retain only those terms appearing (with a set frequency) in more than three domains. Polysemic terms were discarded manually.

To extract the knowledge needed for the conceptual analysis and the writing of the flexible terminological definitions, the methodology of frame-based terminology (with certain additions) was followed (§4.2.2). This methodology consists of a combined top-down and bottom-up approach. The top-down method includes mainly the analysis of definitions from other terminological resources, whereas the bottom-up approach comprises corpus analysis.

For more efficient knowledge extraction from the corpora, we employed hypernymic knowledge patterns (Meyer 2001: 290) coded as word-sketches for SketchEngine. This allows for the extraction of superordinate concept candidates for the choice of genus in definitions. Moreover, we created a word-sketch for the extraction of contextonyms, which are the lexical units that tend to co-occur with a given lexical unit in linguistic contexts (Ji, Ploux, and Wehrli 2003; Ji and Ploux 2003). In this work, the analysis of contextonyms was used to determine the semantic traits activated by a concept in a given domain.

As a result of the first part of our empirical study (the analysis of all the terms in our working list) (§5.1), we divided our notion of domain-dependent contextual variation into three different phenomena: i) modulation (similar to Cruse's modulation (§3.5.3.5.2)); ii) perspectivization (related to Cruse's ways-of-seeing (§3.5.3.5.3)); iii) subconceptualization (akin to Cruse's microsenses and local sub-senses (§3.5.3.5.4)). These phenomena are additive in that all concepts experience modulation, some concepts also undergo perspectivization, and finally, a small number of concepts are additionally subjected to subconceptualization.

Modulation (§5.2.1) is the type of contextual variation that only alters minor characteristics of a concept which are neither necessary nor prototypical. These alterations are not represented in a terminological definition. For its part, perspectivization (§5.2.2) results in the change in the level of prototypicality of certain traits for a concept in relation to the general environmental premeaning. Finally, subconceptualization (§5.2.3) is the type of contextual variation in which the extension of the concept in relation to the general environmental premeaning is modified.

In the second part of our empirical study (§5.4), we created two flexible terminological definitions, one for a concept with subconceptualizations (POLLUTANT) and another for a concept with perspectives (CHLORINE). In this section, we presented guidelines on how to build them, from the extraction of knowledge to the actual writing of the definition.

These guidelines ensure that the definition actually reflects how the defined concept is construed in different environmental domains, which might differ from the viewpoint adopted in the environment as a whole or other environmental subdomains.

This doctoral thesis contributes to the improvement of the quality of terminological definitions because, with our approach, the user is presented with a definition tailored to the domain that he/she has chosen, thus multiplying the probabilities that the definition will offer him/her the information he/she needs. Furthermore, flexible terminological definitions provide a knowledge representation that better resembles the human conceptual system than traditional terminological definitions. As a consequence, a flexible definition not only provides more relevant information, but it also accomplishes this in a way that potentially facilitates and enhances knowledge acquisition.

# 1 INTRODUCCIÓN

## 1.1 PRESENTACIÓN Y DELIMITACIÓN DEL ESTUDIO

El gran desarrollo de las tecnologías de la información permite la transmisión y representación del conocimiento de muy diversas maneras. Sin embargo, las definiciones siguen siendo uno de los componentes más importantes de cualquier recurso terminológico de calidad y un modo privilegiado de representar el conocimiento, pues ofrecen una explicación directa en lenguaje natural del contenido de un concepto. Aunque otras formas de representar el conocimiento pueden coexistir y complementar las definiciones en un recurso terminológico, la adecuación de las definiciones determinará en gran medida la utilidad global del recurso para el usuario.

La motivación de este estudio parte de la observación de que a menudo las definiciones terminológicas no satisfacen las necesidades de los usuarios.

Ello se debe, en parte, a ciertas preconcepciones sobre la función de las definiciones y sobre la naturaleza del significado que afectan a cómo se realizan las definiciones terminológicas. El verbo *definir* deriva del latín *definere,* que significa «poner límites». Precisamente, siempre se ha asumido que ese es el objetivo de las definiciones. Grosso modo, definir se ha tomado tradicionalmente por equivalente de enunciar las características suficientes y necesarias que componen el significado de un término y que lo distinguen de otros que pertenecen a la misma categoría. Este enfoque, conocido como la definición aristotélica (§3.2), presupone la existencia de un significado estable e independiente del modo en el que se empleen los términos en contexto. Además, se asume que el significado (o conocimiento semántico) es independiente del conocimiento del mundo (o conocimiento enciclopédico).

La consecuencia lógica de aplicar este enfoque es que todo concepto tendría una única definición ideal en cuanto a los rasgos semánticos representados. Tal definición especificaría los límites exactos del concepto estableciendo cuáles son las características suficientes y necesarias. Dichas características existirían objetivamente y simplemente sería necesario descubrirlas durante el proceso de elaboración de la definición.

Esta visión tradicional, aún con arraigo en la práctica terminológica actual y por herencia principalmente de la teoría general de la terminología (§2.2.1), hace que se cree una única definición para cada término o concepto con la que se espera satisfacer las necesidades de todos los usuarios. Además, en dicha definición, toda la información considerada «no definicional» se omite o, en el mejor de los casos, se incluye (a menudo de manera inconsistente) en otros campos del recurso terminológico, como la nota enciclopédica.

Las premisas fundamentales sobre las que se basa el enfoque tradicional sobre las definiciones han sido refutadas en el marco de lingüística cognitiva (§2.1). Por ejemplo, de acuerdo con la teoría de los prototipos (§2.1.3.2), la delimitación universal que se requiere para crear una definición aristotélica no es factible porque los límites de los conceptos son difusos. Por otra parte, hay ciertas características que están normalmente vinculadas a un concepto, pero no son suficientes ni necesarias dado que el sistema conceptual humano está organizado de tal manera que muestra «efectos de prototipo». Por ejemplo, el prototipo de PÁJARO tiene la característica «volador», pero ello no impide que un pingüino sea clasificado como un pájaro.

La lingüística cognitiva también defiende una visión enciclopédica del significado (§2.1.2.1), según la cual, no es posible establecer una clara división entre nuestro conocimiento sobre el significado de las palabras (conocimiento semántico) y nuestro conocimiento del mundo (conocimiento enciclopédico), puesto que, en realidad, son un contínuum. De hecho, el conocimiento enciclopédico adquiere un papel central en el estudio del significado, pues los conceptos siempre aparecen insertos en marcos, que son estructuras de conocimiento enciclopédico que les aportan sentido.

Por último, la lingüística cognitiva ha señalado que la distinción entre semántica y pragmática es también difusa. Tradicionalmente, las definiciones deben describir el contenido semántico de una palabra, es decir, lo que significa independientemente de cualquier factor pragmático. Sin embargo, el significado no es una entidad estable, sino que se construye en cada evento de uso a partir del contexto (§2.1.2). Por lo tanto, significado y contexto son inseparables. No obstante, esto no implica que no sea posible realizar abstracciones a partir del significado asociado a una palabra en diversos contextos de uso. De hecho, los prototipos son una suerte de abstracción. Sin embargo, limitar las definiciones a la especificación del contenido semántico estable a través de diversos contextos limita la utilidad de la definición.

En esta tesis doctoral, aplicamos estas premisas de la lingüística cognitiva a la definición terminológica y presentamos una propuesta que se denomina la *definición terminológica flexible* (§3.6). Consiste en un sistema de definiciones del mismo concepto compuesto por una definición general — en nuestro caso, que engloba el dominio del medio ambiente al

completo— junto con definiciones adicionales en las que se describe el concepto específicamente desde el punto de vista de los distintos subdominios en los que el concepto es relevante.

Nuestra propuesta parte de la base de que, al deshacerse las fronteras artificiales entre conocimiento semántico y pragmático, la representación de la variación contextual en la definición terminológica deja de ser una simple posibilidad. Dada la ubicuidad del contexto y sus efectos en la cognición y el lenguaje, representar qué rasgos activan los conceptos según el contexto se torna una necesidad si se aspira a satisfacer plenamente las necesidades de los usuarios. Asimismo, ello supone la inclusión en la definición de características prototípicas, es decir, que no siempre son aplicables al concepto, pero que son relevantes en un contexto dado.

De igual manera, el conocimiento enciclopédico en la definición terminológica deja de estar desterrado para formar parte íntegra de ella. La representación del papel que el concepto definido desempeña en los marcos que activa debe formar parte de la definición en la medida de lo posible.

Nuestra propuesta se entronca en la teoría de la terminología basada en marcos (§2.2) y toma como punto de partida la aplicación de dicha teoría a la representación del conocimiento especializado y la definición terminológica en EcoLexicon, base de conocimiento terminológica sobre el medio ambiente. En EcoLexicon, las definiciones en lenguaje natural son uno de los medios principales de representación del conocimiento junto con las redes conceptuales donde los conceptos están relacionados con otros conceptos a través de relaciones jerárquicas y no jerárquicas.

En particular, nuestra propuesta de definición flexible se inspira en la recontextualización de EcoLexicon (§3.6.1), a partir de la cual, los mapas conceptuales muestran únicamente la información relevante al subdominio del medio ambiente elegido por el usuario. Con ello, se representa la variación contextual y se evita la sobrecarga de información, aumentando la adquisición de conocimiento por parte de los usuarios.

EcoLexicon sigue el principio de Meyer, Bowker y Eck (1992: 159) según el cual para que una base de conocimiento terminológica sea realmente útil, esta debe reflejar la organización de los conceptos en la mente. Puesto que la definición terminológica es un tipo de representación conceptual (Faber 2002), en esta tesis doctoral, partimos de la base de que la construcción de definiciones terminológicas también debe fundamentarse en cómo se organiza el sistema conceptual humano.

Así pues, dado que el contexto es un factor determinante en la construcción del significado de cualquier unidad léxica, incluidas las terminológicas, asumimos que la definición terminológica puede y debe reflejar los efectos del contexto, a pesar de que tradicionalmente la definición se haya entendido como la expresión del significado despojado de los efectos del contexto.

## 1.2 OBJETIVOS

El objetivo principal de esta tesis doctoral es analizar los efectos de la variación contextual en conceptos especializados del medio ambiente con vistas a su representación en la definición terminológica. En particular, nos concentraremos en la variación contextual basada en restricciones temáticas. Esto es, en cómo las distintas áreas de conocimiento que forman el vasto dominio del medio ambiente conceptualizan de manera diferente los mismos conceptos y cómo ello puede reflejarse en la definición.

Con el fin de alcanzar dicho objetivo, se proponen los siguientes objetivos específicos:

- Determinar los componentes del contexto que afectan a la construcción del significado especializado y especificar aquellos que afectan a la representación conceptual en la definición terminológica.

- Caracterizar los fenómenos conceptuales a los que da lugar la variación contextual en el dominio del medio ambiente y establecer cómo gestionarlos mediante definiciones terminológicas flexibles.
- Desarrollar pautas de elaboración de definiciones terminológicas flexibles partiendo de la extracción de conceptos que varían contextualmente a partir de corpus hasta la redacción final de las definiciones, pasando por la extracción de conocimiento específica de cada subdominio y su representación en plantillas definicionales.

En definitiva, esta investigación pretende contribuir a la mejora de la calidad de las definiciones terminológicas. Para ello, se pretende avanzar en la comprensión de la variación contextual de conceptos especializados y aportar un marco metodológico para su representación en la definición terminológica basado en la lingüística cognitiva y más concretamente, en la terminología basada en marcos.

# 2 MARCO TEÓRICO

Una definición suele describirse a grandes rasgos como la enunciación del significado de una palabra, término o símbolo (P. Hanks 2006: 399). De ello se desprende que el modo en que se concibe la noción de significado debe ser la base de cualquier estudio en torno a la definición. Por este motivo, antes de concentrarnos en las cuestiones específicas que incumben a la definición en sí, clarificaremos las nociones de significado, concepto y categorización, entre otras. Lo haremos desde la lingüística cognitiva, la cual ofrece explicación a numerosos fenómenos relacionados con la definición y el significado en general que anteriormente habían sido obviados o relegados fuera del campo de la lingüística.

## 2.1 LA LINGÜÍSTICA COGNITIVA

La lingüística Cognitiva es un conjunto de teorías y enfoques relacionados acerca del estudio de la lengua que emergió en la década de los setenta y que ha sido especialmente activo a partir de la década de los ochenta (Croft y Cruse 2004: 1).

El adjetivo «cognitivo» en *lingüística cognitiva* ha resultado algo polémico. Los lingüistas generativistas argüían que ellos ya habían remarcado la naturaleza cognitiva de la habilidad lingüística del ser humano, mientras que los psicolingüistas también opinaban que la adopción de ese nombre despreciaba su trabajo sobre el lenguaje (Gibbs 1996: 29). Con todo, de acuerdo con Gibbs (1996: 49), la lingüística cognitiva es merecedora de su nombre por su compromiso con la búsqueda de la relación entre el conocimiento conceptual, la experiencia corporeizada y el lenguaje, así como por su objetivo de arrojar luz sobre el verdadero contenido de la cognición humana.

Aunque no se trata de una corriente homogénea, toda investigación enmarcada dentro de la lingüística cognitiva estará guiada por dos compromisos, según Lakoff (1990: 41): el compromiso de la generalización y el compromiso cognitivo. El compromiso de la generalización requiere que el lingüista cognitivo dé cuenta de todos los principios generales que rigen el lenguaje humano. Por su parte, el compromiso cognitivo se refiere a la necesidad de que tal generalización esté en consonancia con aquello que la investigación científica ha descubierto acerca de la mente humana. En caso de conflicto entre ambos compromisos, será el compromiso cognitivo el que prevalezca porque los lingüistas cognitivos requieren generalizaciones basadas en la realidad de la cognición (Lakoff 1990: 41).

Podría argumentarse que el compromiso de la generalización es un subproducto del compromiso cognitivo, dado que, como indican Evans y Green (2006: 36–40), la investigación sobre la cognición humana ya ha demostrado que muchos fenómenos tradicionalmente confinados a una determinada área del lenguaje son en realidad generales. Es el caso, por ejemplo, de la polisemia que puede encontrarse en otros subdominios lingüísticos distintos a la semántica como la morfología y la sintaxis. De ahí que los lingüistas cognitivos busquen principios generales aplicables a todas las subáreas de la lingüística sin considerar que hay subáreas más importantes que otras. De hecho, la división del lenguaje en subáreas se considera artificial, aunque en ocasiones puede resultar útil desde un punto de vista metodológico (Evans y Green 2006: 28).

Por otro lado, el compromiso cognitivo también implica una suerte de generalización. Mientras que por causa del compromiso de generalización se buscan principios generales a través de las subáreas de la lingüística, bajo el compromiso cognitivo, los lingüistas cognitivos generalizan al lenguaje aquello que se conoce de la cognición humana. Se considera que la habilidad lingüística es una habilidad cognitiva más regida por los mismos principios generales:

> [T]he organization and retrieval of linguistic knowledge is not significantly different from the organization and retrieval of other knowledge in the mind, and the cognitive abilities that we apply to speaking and understanding language are not significantly different from those applied to other cognitive tasks, such as visual perception, reasoning or motor activity (Croft y Cruse 2004: 2).

El hecho de que todo el trabajo desarrollado bajo este paradigma respete el compromiso de la generalización asegura que los descubrimientos realizados en lingüística cognitiva puedan aplicarse satisfactoriamente a la terminología. Esto se debe a que, sin perjuicio de las relaciones del campo de la terminología con otros dominios, los términos son esencialmente unidades lingüísticas (Cabré 2000c: 14). Como se verá más adelante, varias teorías recientes en terminología son de corte cognitivo (§2.2).

Por otro lado, en este trabajo, extendemos a la definición terminológica la idea de que, para que una base de conocimiento terminológica sea realmente útil, debe reflejar la organización de los conceptos en la mente (Meyer, Bowker y Eck 1992: 159). Por lo tanto, la definición terminológica también deberá dar cuenta de la organización de los conceptos en la mente y, como veremos más adelante, de cómo esos conceptos dan lugar al significado. Por consiguiente, tomar como fundamento teórico la lingüística cognitiva nos permite acercarnos a ese propósito gracias a su compromiso cognitivo.

La posición principal de la semántica cognitiva respecto al significado se basa en el principio de que, si la representación mental de cualquier conocimiento lingüístico y los procesos en que este se emplea son similares a los de cualquier otra estructura cognitiva, puede afirmarse que el conocimiento lingüístico es de naturaleza conceptual (Evans y Green 2006: 158). Como consecuencia de ello, la estructura semántica humana se considera un subconjunto de la estructura conceptual humana, la cual también abarca conocimiento conceptual no asociado directamente al lenguaje:

> Much thought is clearly nonverbal (consider the task of working a jigsaw puzzle), and many established concepts have no conventional linguistic symbolization (an example is the area above the upper lip and below the nose, where a moustache belongs) (Langacker 1987: 60).

El supuesto al que la lingüística cognitiva se opone consiste en que el conocimiento semántico es independiente y de naturaleza distinta a cualquier otro tipo de conocimiento conceptual. Es lo que ha venido a llamarse el *enfoque diccionarístico* (Evans y Green 2006: 158), puesto que extrapola la visión tradicional sobre la definición en los diccionarios a la representación léxica en la mente. De acuerdo con este enfoque, la información semántica en el lexicón mental se asemejaría al supuesto ideal de definiciones lexicográficas estáticas y bien delimitadas. Dado que las enciclopedias incluyen otra suerte de conocimiento en sus entradas (p. ej., información social, cultural, histórica, científica, etc.), el enfoque diccionarístico hacia el significado considera que todo conocimiento que no se suela representar en una definición lexicográfica pertenece al denominado «conocimiento enciclopédico» o «conocimiento del mundo». Así pues, solo el conocimiento definicional —el núcleo semántico de una palabra— sería el objeto de estudio de la semántica léxica según el enfoque diccionarístico (Evans y Green 2006: 208).

Un ejemplo destacado de este enfoque es el adoptado por los estructuralistas, que ha sido y sigue siendo muy influyente en lingüística. El estructuralismo define la lengua como un sistema independiente compuesto por signos lingüísticos, que son una combinación de significado (un concepto) y un significante (una forma) (Saussure 1916: 129). El significado, por tanto, se considera que surge por las relaciones existentes entre los diferentes signos que forman parte del sistema de la lengua y la referencia se considera solamente como un vínculo entre las palabras y las entidades del mundo real establecido por convención. Por lo tanto, el significado de una palabra (conocimiento semántico) sería diferente al conocimiento sobre la entidad del mundo real a que se refiere dicha palabra (conocimiento enciclopédico).

Asimismo, bajo el paradigma previo a la lingüística cognitiva, se considera que el contenido semántico de una palabra es independiente del contexto[1] de uso y que, por ende, existe un significado semántico por un lado y un significado pragmático por otro (Evans y Green 2006: 208–209). El significado semántico sería independiente del contexto, estable, determinado y estaría almacenado como tal en el lexicón mental. Este estaría en oposición al significado pragmático o inferido, que depende del contexto y que no sería parte del objeto de estudio de la semántica léxica. Esta dicotomía también tiene su reflejo en las definiciones en los diccionarios, pues tradicionalmente los lexicógrafos tratan de representar el significado independiente de los contextos en los que después esa palabra pueda utilizarse.

La lingüística cognitiva refuta la separación tradicional entre el conocimiento semántico y cualquier otro tipo de conocimiento partiendo de la idea arriba mencionada de que los procesos cognitivos que controlan el uso del lenguaje son los mismos que los de otras habilidades cognitivas (Croft y Cruse 2004: 2). De este modo, el conocimiento semántico está compuesto por conceptos que no son diferentes de otros conceptos almacenados en la mente humana (Evans y Green 2006: 159). Así, distinguir entre conocimiento semántico y enciclopédico no sería posible; el conocimiento semántico asociado a una unidad léxica es, pues, intrínsecamente enciclopédico. Esto explica por qué resulta a menudo tan difícil decidir qué elementos semánticos incluir al redactar una definición lexicográfica tradicional. Tratar de distinguir el contenido semántico del enciclopédico equivale a trazar una frontera inexistente[2]. Por ello, una simple comparación de las definiciones de cualquier palabra en varios diccionarios nos muestra que los lexicógrafos no siempre encuentran consenso respecto al contenido semántico de una unidad léxica.

---

[1] A lo largo de este trabajo, a menos que se indique lo contrario, utilizaremos el término *contexto* en su sentido más amplio, el cual, siguiendo la definición de Evans (Evans 2009a: 4), no solo abarca las palabras que acompañan a una palabra dada en un enunciado, sino que se hace referencia toda la situación comunicativa en la que se enmarca, lo cual incluye, entre otros factores, el conocimiento de fondo compartido entre el emisor y el receptor, la situación física y temporal en la que se produce la comunicación y la intención comunicativa del emisor. Trataremos esa cuestión en detalle en §3.5.3.

Finalmente, los lingüistas cognitivos defienden que realmente las unidades léxicas no portan significados en sí, sino que solamente inducen la generación mental del significado con arreglo a un contexto de uso real concreto. Así pues, la lengua guía la activación del conocimiento apropiado en una situación dada (Fauconnier 1994: xviii). El significado no está almacenado en la mente humana, el significado se crea en contexto. Por lo tanto, como corolario del destierro de la distinción entre conocimiento semántico y enciclopédico, la separación entre significado semántico y significado pragmático también queda invalidada. El significado es siempre pragmático, guiado por el contexto. Así, el significado lingüístico no equivale a los conceptos sino a concepciones que surgen en cada evento de uso mediante el proceso llamado *conceptualización* (Langacker 2008: 30).

Este enfoque semántico recibe el nombre de *tesis basada en el uso*. Según esta tesis, el ser humano aprehende qué contenido conceptual se asocia convencionalmente a una unidad léxica a partir de la abstracción de patrones de las concepciones que surgen en cada evento de uso al que se expone. Los patrones que surgen más habitualmente experimentan un proceso de arraigo (*entrenchment*) en la mente del usuario de la lengua (Langacker 2008: 16), lo cual conlleva que dichos patrones se establezcan como rutinas cognitivas. En otras palabras, si el usuario encuentra con frecuencia que un contenido conceptual forma parte habitualmente de las concepciones que emergen a partir de una unidad léxica concreta, el usuario asociará ese contenido a dicha unidad léxica y lo activará con más frecuencia en futuros eventos de uso en los que aparezca dicha unidad léxica.

[2] Dado que la frontera entre contenido semántico y enciclopédico es difusa, en este trabajo, defendemos que los límites de una definición terminológica vendrán marcados por factores ontológicos, funcionales y contextuales (§3.5) asociados a las necesidades del usuario que recibirá la definición.

### 2.1.1 La cognición fundamentada

En los últimos años, han tomado relevancia unas nuevas teorías de la cognición que han recibido el nombre de *cognición fundamentada* (*grounded cognition*). Las teorías fundamentadas de la cognición desafían las teorías tradicionales de la cognición proponiendo que las representaciones conceptuales que subyacen al conocimiento en la mente humana están fundamentadas en los sistemas sensoriales y motores, en lugar de estar representadas y ser procesadas de manera abstracta en estructuras amodales de información conceptual (Pezzulo et al. 2013: 1).

Aunque este estudio se trate de la definición terminológica, la cual no permite más que el empleo de las palabras para representar información conceptual, tener en cuenta las teorías de la cognición fundamentada puede optimizar la efectividad de las definiciones. Ello se debe a que definir, según estas teorías, supone una traducción lingüística de los símbolos modales asociados a un concepto:

> When people define a concept, they retrieve or construct a schematic image, focus attention on a subset of its perceptual symbols in a sequential manner, and describe the content of each focus with a linguistic description (Barsalou et al. 1993).

Las teorías tradicionales asumen que el conocimiento reside en un sistema de memoria semántica separada de los sistemas modales de la percepción (visión, audición, etc.), acción (movimiento, propiocepción, etc.) e introspección (estados mentales, emoción, etc.) (Barsalou 2008: 618). Por consiguiente, las representaciones provenientes de los sistemas modales se convertirían en símbolos amodales al almacenarse en la memoria semántica. Sin embargo, las teorías fundamentadas rechazan por lo general que sean símbolos amodales los que representen el conocimiento en una memoria semántica. De hecho, Barsalou (2008: 618) cuestiona que haya siquiera alguna clase de representaciones amodales en el cerebro.

Una de las nociones básicas sobre las que se cimientan las teorías fundamentadas es la de «simulación», que consiste en la recreación de estados perceptuales, motores e introspectivos adquiridos durante la experiencia con el mundo, el cuerpo y la mente (Barsalou 2008: 618):

> As an experience occurs (e.g., easing into a chair), the brain captures states across the modalities and integrates them with a multimodal representation stored in memory (e.g., how a chair looks and feels, the action of sitting, introspections of comfort and relaxation). Later, when knowledge is needed to represent a category (e.g., chair), multimodal representations captured during experiences with its instances are reactivated to simulate how the brain represented perception, action, and introspection associated with it (Barsalou 2008: 618–619)

Dado que la proliferación de teorías fundamentadas ha llevado a la confusión en el uso de los términos *fundamentado*, *corporeizado* y *situado*, Barsalou (2008) propone utilizar el término *cognición fundamentada* para abarcarlos todos. No obstante, Pezzulo et al. (2013: 4) aclaran que los efectos sobre la cognición y la representación conceptual de la fundamentación (propiedades físicas del mundo), la corporeización (limitaciones inherentes a las características físicas del cuerpo humano) y la situacionalidad (características específicas del entorno y el contexto, incluyéndose aspectos sociales y culturales) son aditivos.

En lingüística cognitiva se ha puesto el acento desde sus comienzos en la tesis de la corporeización. Según ella, todos los conceptos en la mente humana están determinados por la corporeización y no son, por ende, un simple reflejo de la realidad externa (Lakoff y Johnson 1999: 22). De este modo, se asume un enfoque denominado *realismo experiencial*, que recibe el nombre de *realismo* porque reconoce la existencia de un mundo real y el papel limitador de la realidad en el proceso de la conceptualización. Por otro lado, el término *experiencial* se emplea para remarcar que el sistema conceptual refleja la experiencia que el humano tiene de la realidad y no la realidad en sí, debido no solo a la mediatización de la corporeización sino también a la del entorno social y cultural (Lakoff 1987: xv; Lakoff y Johnson 1980: 180).

La relación de la naturaleza corporeizada de la cognición con el lenguaje se ha explorado de manera explícita en el marco de la teoría de los esquemas de imagen (Johnson 1987). Los esquemas de imagen son representaciones conceptuales esquemáticas de un patrón recurrente en diferentes experiencias sensoriales y perceptuales. Algunos esquemas de imagen propuestos por Johnson (1987: 126) son CAMINO, RECIPIENTE, CICLO, PARTE-TODO, VÍNCULO, EMPAREJAMIENTO, OBJETO, CONTACTO, PROCESO, ARRIBA-ABAJO, etc.

Una de las maneras en las que la corporeización influye en todo nuestro sistema conceptual es a través de proyecciones metafóricas (Johnson 1987: xv). La metáfora conceptual es el mecanismo por el cual un tipo de cosas es entendida y experimentada en términos de otro (Lakoff y Johnson 1980: 4). Esto incluye, entre otros tipos de proyección metafórica, la comprensión de conceptos abstractos en términos de esquemas de imagen. Por ejemplo, la metáfora MÁS ES ARRIBA (p. ej., «el número de estudiantes matriculados en esa asignatura se elevó el año pasado», «el precio de la luz ha subido en los últimos meses») se trata de un caso en el que una noción abstracta como CANTIDAD se entiende en términos de un esquema de imagen (ARRIBA-ABAJO) (Johnson 1987: xv).

Otra forma en que nuestro sistema conceptual depende en última instancia de la experiencia corpórea se explica en la teoría de los dominios de Langacker (1987). De acuerdo con esta teoría, para caracterizar cualquier unidad semántica, se requiere un contexto compuesto de experiencias mentales, espacios representacionales, conceptos y complejos conceptuales, lo que Langacker denomina un *dominio*. Aunque normalmente, la comprensión de un concepto presupone varios dominios o una matriz de dominios (Langacker 1987: 147).

El concepto de NUDILLO presupone, entre otros, el concepto de DEDO. Por ello, NUDILLO solo puede entender en relación con el dominio de DEDO. Al mismo tiempo, DEDO también presupone el dominio de MANO, MANO presupone BRAZO, etc. Con este ejemplo, Langacker (1987: 148) demuestra que los dominios forman jerarquías en las que los dominios de un nivel dependen de otros en el siguiente nivel y así sucesivamente. Sin embargo, llega un punto en que ciertos dominios no pueden entenderse en relación con otro dominio. Se trata de dominios primitivos que se derivan directamente de la experiencia corpórea y se llaman *dominios básicos*:

> If [FINGER] is the domain for [KNUCKLE], [HAND] for [FINGER], [ARM] for [HAND], and [BODY] for [ARM], what is the domain for [BODY]? The notion [BODY] (so far as shape is concerned) is a configuration in three-dimensional space, but it hardly seems appropriate or feasible to consider three-dimensional space as a concept definable relative to some other, more fundamental conception. It would appear more promising to regard the conception of space (either two- or three dimensional) as a basic field of representation grounded in genetically determined physical properties of the human organism and constituting an intrinsic part of our inborn cognitive apparatus[3]. (Langacker 1987: 148)

Todo dominio que depende de otro, mediante cualquier proceso conceptual, se denomina —en la terminología de Langacker— un *dominio abstracto* (Langacker 1987: 150). Aunque los dominios abstractos no sean el producto directo de la corporeización, todo nuestro sistema cognitivo está en última instancia fundamentado en la experiencia corpórea, ya que el nivel más bajo de la jerarquía de dominios está siempre ocupado por un dominio básico[4] (Langacker 1987: 149).

#### 2.1.1.1 La teoría de los sistemas de símbolos perceptuales

Barsalou (1993; 1999; 2003) propone una teoría del conocimiento basada en la cognición fundamentada llamada la teoría de los sistemas de símbolos perceptuales (TSSP). Su idea principal consiste en que el sistema conceptual humano está poblado por símbolos perceptuales consistentes en registros de la activación neural que surgen durante la percepción[5] (1999: 583). En otras palabras, un símbolo perceptual es un registro en el sistema conceptual de extractos de la percepción en la misma modalidad en la que se ha producido la percepción. Sin embargo, no se trata de una copia exacta y completa del estado cerebral que subyace a la percepción, sino que es esquemática y es la atención selectiva según el contexto la que aísla la información en la percepción y la almacena en la memoria a largo plazo (1999: 583).

---

[3] En la cita se ha respetado la ortotipografía original. Langacker escribe los dominios entre corchetes y en mayúsculas.

[4] Los dominios básicos de Langacker están íntimamente relacionados con los esquemas de imagen, pero no son lo mismo. Clausner y Croft (1999: 22) explican que algunos esquemas de imagen realmente corresponden a matrices de dominios básicos y, por tanto, en relación a la Teoría de los Dominios, serían considerados dominios abstractos. Uno de esos casos es el del esquema de imagen RECIPIENTE que se trata de un dominio abstracto porque presupone los dominios básicos de ESPACIO y OBJETO MATERIAL.

Asimismo, los símbolos perceptuales son dinámicos, es decir, las simulaciones a las que dan lugar no son siempre iguales, pues se ven afectadas tanto por otros símbolos perceptuales almacenados posteriormente como por factores contextuales:

> The subsequent storage of additional perceptual symbols in the same association area may alter connections in the original pattern, causing subsequent activations to differ. Different contexts may distort activations of the original pattern, as connections from contextual features bias activation toward some features in the pattern more than others (Barsalou 1999: 584).

El dinamismo de los símbolos conceptuales surge por el hecho de que los elementos que forman la realidad pueden tomar muy distintas formas, aparecer en distintas situaciones y servir distintos objetivos. Por ello, una única representación conceptual estática no resultaría óptima (Barsalou 2003: 553). Así pues, el contexto guía tanto el almacenamiento de los símbolos perceptuales en la memoria a largo plazo como las simulaciones a las que estos puedan dar lugar ulteriormente.

Por otro lado, Barsalou (1999: 584) destaca la naturaleza combinatoria de los símbolos perceptuales. Es decir, no son imágenes holísticas, sino que cada símbolo perceptual registra un aspecto concreto de la percepción. Por este motivo, los símbolos perceptuales no representan necesariamente entidades concretas, sino que según factores causales y contextuales un símbolo perceptual puede formar parte de la representación de diferentes referentes. Por ejemplo, un símbolo perceptual asociado a la percepción del color rojo puede estar asociado a distintos referentes en el mundo, como la sangre o la bandera de China.

---

[5] Barsalou utiliza una noción de *percepción* más amplia que la tradicional, pues incluye cualquier aspecto de la experiencia percibida, incluidas las propiocepción y la introspección (1999: 585).

##### *2.1.1.1.1 La organización del sistema conceptual*

El dinamismo del sistema conceptual no implica que los símbolos perceptuales sean independientes unos de otros. La TSSP postula que los símbolos perceptuales relacionados se organizan en torno a unas estructuras multimodales llamadas *marcos*. Los marcos en la TSSP comparten muchas características con otros constructos similares, como los dominios de Langacker (1987) o los esquemas de imágenes de Johnson (1987), que hemos visto anteriormente, así como los marcos de Minsky (1975) o Fillmore (1982a) (§2.1.2.2) y los guiones de Schank y Abelson (1977).

A partir de los marcos, el sistema cognitivo puede construir simulaciones (Barsalou 1999: 586). Un marco junto con las simulaciones a las que puede dar lugar se denomina un *simulador*. Dentro de la TSSP, un simulador correspondería a lo que habitualmente se considera un concepto mientras que las simulaciones equivaldrían a las distintas concepciones (Barsalou 1999: 587).

Barsalou (1991: 53) explica la organización conceptual en la mente humana mediante la metáfora del modelo del mundo, que es la idea que una persona tiene sobre el estado del mundo en este momento. La organización de dicho modelo del mundo está orientada hacia la consecución de objetivos (Barsalou 2003: 552). Para ello, es necesario que la persona sea capaz de integrar la información sobre cómo realizar acciones, qué secuencia se ha de seguir y qué elementos son necesarios con la información acerca de las características del entorno. En este sentido, hay dos motivaciones opuestas que configuran el sistema conceptual humano:

> At one extreme, people are intuitive taxonomists. Their goal is to discover the categorical structure of the world, develop taxonomic systems that represent this structure, and establish background theories that frame these taxonomies. At the other extreme, people are goal achievers who organise knowledge to support situated action. On this view, the primary organisation of the conceptual system supports executing actions effectively in the environment, with taxonomic hierarchies constituting a secondary-level (Barsalou 2003: 546).

El modelo del mundo permite la interfaz entre la acción y el entorno gracias a una estructura de marcos espacio-temporales que a su vez contienen marcos de entidades —ya sean concretas o abstractas— organizados en taxonomías (Barsalou 2003: 546). Los marcos contenidos en el modelo del mundo se dividen en cuatro tipos: los marcos que representan individuos, modelos, situaciones episódicas y situaciones genéricas (Barsalou et al. 1993)[6]. Por un lado, los marcos de individuo y las situaciones episódicas están relacionados con experiencias o conocimientos específicos, mientras que los modelos y las situaciones genéricas son generalizaciones a partir de marcos de individuo y situaciones episódicas.

Los marcos de individuo representan tanto entidades animadas como inanimadas existentes en el mundo (Barsalou et al. 1993), es decir, instancias reales como ELON MUSK o la TROPOSFERA. Cada vez que se activa ese marco y se procesa nueva información, esta se añade al mismo marco; no se crea uno nuevo. Por su parte, los marcos de modelo representan un tipo de individuo, es decir, una categoría (2.1.3). Mientras que los marcos de individuo tienen un único referente real en el mundo, los marcos de modelo son constructos mentales sin un referente real concreto:

> For example, people do not believe that a physical model of dog exists in the world that corresponds to their internal model of dog. Certainly, different individuals in the world may instantiate a model, but no direct physical counterpart to the model typically exists. Instead, people view their models for types as only existing mentally. (Barsalou et al. 1993)

[6] Aunque la TSSP no incluye en esta clasificación los marcos que se refieren, por ejemplo, a conceptos abstractos o a atributos, Barsalou (1992: 31) defiende que todos los tipos de concepto toman la forma de marco.

Como hemos indicado arriba, los marcos de individuo y modelo no solo forman taxonomías a partir de los procesos de categorización, sino que además están enmarcados dentro de eventos. Los eventos son conjuntos de marcos de situaciones, que, a su vez, están formadas por imágenes.

Barsalou et al. (1993) definen las imágenes como un conjunto de símbolos perceptuales que representan individuos y/o modelos en una configuración espacial estática percibida desde una perspectiva particular. Un ejemplo de imagen sería una vista frontal de un sismómetro encima de una mesa. Por su parte, las situaciones son una serie de imágenes que representan un conjunto relativamente constante de individuos y/o modelos que cambian de algún modo significativo de manera continua a lo largo del tiempo en una región del espacio relativamente constante. Un ejemplo de situación sería una serie de imágenes en las que una persona presiona un botón de un sismómetro y este produce un sismograma.

Asimismo, una situación puede ser episódica o genérica. Un marco de situación episódica (p. ej., EL ACCIDENTE NUCLEAR DE CHERNÓBIL) se crea a partir de un evento concreto ocurrido en el mundo y se almacena asociado a la ubicación en que tuvo lugar. Por su parte, un marco de situación genérica (p. ej., ACCIDENTE NUCLEAR) se crea como una abstracción de dos o más situaciones episódicas. Al igual que ocurre con los marcos de modelo, nuestro interés se centra en las situaciones genéricas, pues las situaciones episódicas no suelen ser objeto de definición.

Una cuestión importante que Barsalou et al. (1993) abordan es la de cuáles son los individuos y modelos que se incluyen dentro de una situación genérica. Por un lado, proponen un criterio cuantitativo según el cual dichos individuos y modelos deben ser comunes a varias de las situaciones episódicas que dan lugar a la situación genérica o estar incluidos de manera recurrente en distintas imágenes que forman las situaciones. Sin embargo, también se apela a la importancia de las creencias de fondo respecto a la situación, pues dichas creencias dictarán la relevancia de los distintos individuos/modelos dentro de la situación también pueden determinar su representación dentro del marco de la situación. Este segundo criterio coincide con los principios que rigen la categorización según de la teoría de la teoría (§2.1.3.3).

Como consecuencia del hecho de que el contexto determina cómo se simulan los conceptos, la representación de un individuo o un modelo se especializa dentro de una situación determinada. De acuerdo con Barsalou et al. (1993) es el principio de «un marco por entidad» el que determina que no se creen distintos marcos para cada vez que la entidad se activa en distintos contextos a pesar de los cambios que su simulación varíe:

> Although specialized models develop in different situations, they nevertheless reside in a single frame, because they all represent the same type of individual. For example, specialized models of car in the generic situations for buying a car, driving a car, and getting gas constitute different perspectives on the same type of individual, namely, cars (Barsalou et al. 1993).

Así pues, la información especializada sobre un modelo en las distintas situaciones en las que se activa se inserta en un único marco a modo de subconjuntos. Al activarse el modelo dentro de una situación, la información asociada a la situación se simulará, mientras que el resto del marco quedará inactivo (Barsalou et al. 1993).

Finalmente, la noción de evento consiste en una serie de dos o más situaciones relacionadas de modo coherente y que conducen a un resultado significativo (Barsalou et al. 1993). Por ejemplo, un evento podría incluir todas las situaciones que se encadenan desde la primera situación en la que una persona corta unas flores en un jardín hasta la situación en que pone un jarrón que contiene dichas flores encima de una mesa.

##### *2.1.1.1.2 Las propiedades de los marcos*

En vez de concebir que los símbolos perceptuales representan características dentro de los marcos (p. ej., el marco del modelo ENERGÍA FÓSIL incluiría la característica CONTAMINANTE), Barsalou (1992: 30) defiende que los marcos están compuestos por conjuntos de atributos y valores. Así, en lugar de listas de características, el marco de ENERGÍA FÓSIL tendría atributos como NIVEL DE CONTAMINACIÓN cuyo valor correspondiente sería ALTO. Tanto los atributos como los valores son marcos (es decir, conceptos) cuyo contenido se especializa por el hecho de estar activados dentro de otro marco.

Entre los atributos de un marco existen correlaciones, es decir, no son independientes. A las relaciones existentes entre los atributos, Barsalou (1992: 35) les da el nombre de *invariantes estructurales*, pues al relacionar dos atributos, se trata de una relación que se mantiene estable en las distintas concepciones del concepto. Por ejemplo, dentro del marco del modelo FUMIGADORA, los atributos correspondientes a las distintas partes de una fumigadora (TANQUE, MANGUERA, etc.) estarían unidos por invariantes estructurales que representarían la relación existente entre cada parte de una fumigadora (dónde está situada cada una respecto a otra, cómo interactúan entre ellas para hacer que la fumigadora funcione, etc.).

Asimismo, también pueden existir correlaciones entre los valores de los atributos. Dichas relaciones entre valores son lo que Barsalou (1992: 37) denomina *restricciones conceptuales*. Las restricciones conceptuales capturan las correlaciones que se dan en las distintas conceptualizaciones a las que puede dar lugar un marco (Barsalou 1993: 40). Por ejemplo, dentro del marco de modelo de PESTICIDA, si el atributo ORIGEN tiene SINTÉTICO como valor, automáticamente el atributo APTO PARA AGRICULTURA ECOLÓGICA adquiere el valor NO.

Finalmente, una de las propiedades que estructura los marcos y que hace que finalmente todos los marcos estén relacionados entre sí es la recursividad, que es la característica de los marcos de estar contenidos unos dentro de otros de manera indefinida (Barsalou y Hale 1993: 133). Este fenómeno surge por el hecho de que un marco comprende atributos y valores que a su vez son marcos:

> Frames can represent both the attributes of a frame and their values, as well as the structural invariants that integrate attributes and the

> constraints that link values. Human knowledge appears to be frames all the way down (Barsalou y Hale 1993: 130–131).

En relación con el conocimiento especializado, el fenómeno de la recursividad hará que la información conceptual que un especialista posea acerca de una noción de su dominio, comparada con la de un lego, esté compuesta por una estructura más amplia y compleja de marcos contenidos unos dentro de otros, dada la mayor profundidad de sus conocimientos:

> The knowledge of engines possessed by one group of conceptualizers, namely car mechanics, is highly complex, and this attribute includes many subordinate attributes each with corresponding values, which are themselves subordinate attributes with further values, and so on. In this way, a frame subsumes multiples frames which are embedded, capturing aspects of the larger units of which they are subparts (Evans 2009b: 204).

#### 2.1.1.2 La relevancia de la cognición fundamentada para la definición

Dado que los conceptos no se procesan aisladamente, sino enmarcados en situaciones o eventos de acuerdo con la interacción de la persona con el mundo, en la representación del conocimiento para los usuarios de un recurso terminológico resulta necesario dar cuenta del contexto en el que los conceptos se activan (Faber 2011: 14). De este modo, los usuarios legos pueden recrear la situación en que se activa un concepto de un modo situado, como lo haría un experto. Por ejemplo, el concepto EROSIÓN, ya sea conceptualizado como un proceso o el resultado de dicho proceso, no puede concebirse de manera aislada:

> It [erosion] is induced by an agent (wind, water, or ice) that affects a geographic entity (the Earth's surface) by causing something (solids) to move away. Moreover, any process takes place over a period of time, and can be divided into smaller segments. In this sense, erosion can happen at a specific season of the year, and may take place in a certain direction. All of this information about erosion should be available for potential activation when the user wishes to acquire knowledge about it (Faber 2011: 14).

Un experimento mediante resonancia magnética funcional llevado a cabo por Faber et al. (2014) demostró que los legos y los expertos presentan algunas diferencias respecto a la activación de áreas cerebrales al realizar tareas de asociación de términos especializados. Específicamente, dadas las regiones del cerebro activadas por los expertos, del estudio se desprende el papel fundamental que desempeña la contextualización y la situación en el procesamiento mental del conocimiento especializado.

Por lo tanto, con arreglo a la cognición fundamentada, para que una definición terminológica facilite la adquisición de conocimiento por parte de los usuarios, esta también debe representar o, al menos, servir como punto de acceso a las situaciones o eventos en los que el concepto que se define suele activarse, ya que son tanto o más importantes que la organización taxonómica.

Por otro lado, el carácter dinámico y composicional del sistema conceptual justifica el hecho de que haya distintos componentes semánticos que estén asociados a un concepto pero que no siempre se activen porque no resulten relevantes en el contexto.

Barsalou (2003: 545–546) explica que el dinamismo del sistema conceptual procede, entre otros motivos, del hecho de que la información que una persona necesita activar sobre una categoría varía según la situación y que, por ello, se crean distintas concepciones a partir de un concepto. En este sentido, la información que un usuario de una definición terminológica necesitará acerca de un concepto también variará según la situación. De ahí que en este trabajo se proponga la creación de distintas definiciones adaptadas a distintos contextos (§3.6).

Finalmente, dado que la organización más probable del sistema conceptual son los marcos, tanto intraconceptual como interconceptualmente, y que estos además permiten dar cuenta de los efectos del contexto sobre la conceptualización, la noción de marco y su carácter relacional se vuelven centrales con vistas a la elaboración de definiciones terminológicas, como veremos más adelante.

Mientras que en este apartado hemos tratado las características del sistema conceptual, a continuación, abordaremos la relación del sistema conceptual con el sistema lingüístico. En particular, nos centraremos en cómo ha aplicado la lingüística cognitiva la noción de marco para el análisis y la representación del significado.

## 2.1.2 La construcción del significado

### 2.1.2.1 La semántica enciclopédica

Como ya hemos indicado anteriormente, la lingüística cognitiva considera que el significado no lo portan las palabras en sí, sino que ellas simplemente proporcionan «minimal, but sufficient, clues for finding the domains and principles appropriate for building in a given situation» (Fauconnier 1994: xviii). Este proceso se denomina *construcción del significado* y se define como «the high-level, complex mental operations that apply within and across domains when we think, act, or communicate» (Fauconnier 1997: 1). En dichos dominios se incluyen aquellos descritos por Langacker (1987) así como los denominados *espacios mentales*, que son dominios temporales y parciales creados durante un evento comunicativo (Fauconnier 1997: 34). La teoría de los espacios mentales de Fauconnier (1994; 1997) explica los procesos que rigen la construcción del significado en enunciados, dentro de los cuales, las unidades léxicas desempeñan un papel fundamental, ya que conectan los distintos elementos que conforman los espacios mentales con el conocimiento necesario para su comprensión almacenado mentalmente a modo de dominios (Fauconnier 1994: xxiii).

Así pues, en lingüística cognitiva se distingue entre el estudio de la representación del conocimiento y el de la construcción del significado. El estudio de la representación del conocimiento intenta desentrañar qué tipos de conceptos pueblan el sistema cognitivo humano y las relaciones entre ellos. La noción clave de la representación del conocimiento es el concepto (Evans y Green 2006: 223). Por su parte, el estudio de la construcción del significado o conceptualización, analiza los procesos por los cuales el significado se crea en contexto. Como indicamos previamente, las representaciones mentales que se crean durante los eventos comunicativos se denominan *concepciones* en contraposición con la noción de *concepto* (Langacker 2008: 30).

De la distinción entre representación del conocimiento y construcción del significado se desprende que el conocimiento enciclopédico (los conceptos) no es equivalente al significado contextual (las concepciones). El significado de una unidad léxica es dinámico. Se construye en cada evento de uso por la activación de una parte concreta del conocimiento enciclopédico asociado a esa unidad léxica. Todo el contenido conceptual que una unidad léxica puede invocar es su potencial semántico (Evans 2009b)[7].

El potencial semántico de una unidad léxica se puede dividir en dos partes, a las que Langacker (1987) denomina *perfil* y *base*. El perfil es la subestructura dentro del potencial semántico que es el foco de atención y conforma el referente conceptual de la unidad léxica, es decir, el concepto o conceptos asociados convencionalmente a esa unidad léxica. Mientras que la base es el conocimiento de fondo que el perfil presupone, es decir, todos los marcos que esa unidad léxica podría invocar en eventos de uso concretos.

Los dos factores principales que intervienen en la construcción del significado, que hacen que no sea un proceso aleatorio, son la convención y el contexto:

> [W]e are not free to construe an utterance (or any of its parts) in whatever way our fancy dictates (otherwise communication would be impossible): there are pressures, of various types and strengths, to interpret in particular ways. Some constraints are very basic constraints which have particular relevance to the processing of utterances fall into two main types, conventional constraints and contextual constraints (Cruse 2011: 120).

[7] Similares a la noción de potencial semántico encontramos la de *potencial de significado* (Allwood 1999; Allwood 2003; P. Hanks 1988) y la de *significación* (Croft y Cruse 2004).

La convención interviene en dos aspectos fundamentales. Por un lado, una unidad léxica se asocia a un potencial semántico principalmente por convención. Una unidad léxica, al utilizarse en eventos de uso reales, da lugar a distintas concepciones que invocan diferentes marcos, lo cual va arraigando la asociación de esa unidad léxica con ese contenido conceptual que se activa al emplearse la unidad léxica. De este modo, se puede afirmar que el potencial semántico es una función de concepciones previas construidas a partir de una palabra empleada en situaciones específicas y, como tal, está continuamente actualizándose (Croft y Cruse 2004: 101). Además, cabe añadir que el potencial semántico también se ve influido por el carácter dinámico del sistema conceptual humano, que —como ya se señaló— también está continuamente actualizándose con nueva información. Es decir, el dinamismo del potencial semántico de una unidad léxica proviene tanto del proceso continuo de actualización de la asociación entre la unidad léxica y un contenido conceptual dado, como del continuo proceso de actualización que sufre dicho contenido conceptual.

Las restricciones convencionales también se aplican de la misma manera al lenguaje especializado, pues una unidad terminológica adquiere su potencial semántico mediante el uso que hacen los expertos de dicha unidad para transmitir una determinada porción de conocimiento.

Asimismo, cabe destacar que la convención puede hacer que exista una mayor tendencia a que se activen determinadas partes del potencial semántico asociado a una palabra, o —en términos de Langacker (2008)— a que haya determinados componentes de conocimiento que adquieran un carácter más central que otros:

> [K]nowledge components have varying degrees of centrality. […] For a given lexical meaning, certain specifications are so central that they are virtually always activated whenever the expression is used, while others are activated less consistently, and others are so peripheral that they are accessed only in special contexts (Langacker 2008: 39).

Se puede afirmar que la centralidad[8] afecta tanto al perfil como a la base, ya que los componentes del perfil presuponen los de la base, de modo que si una característica del concepto es central, los marcos que la presuponen resultarán también centrales. Por ejemplo, respecto al término *cloroplasto, se* activa con alta frecuencia (y, por tanto, es central) el contenido conceptual que indica que los cloroplastos son unos orgánulos propios de las células vegetales en las que tiene lugar la fotosíntesis. Como consecuencia de ello, el marco de la FOTOSÍNTESIS también será central con respecto al término *cloroplasto*.

Aunque una unidad léxica no tiene significado fuera de los eventos de uso, de acuerdo con Langacker (2008: 39), ello no quiere decir que el componente semántico asociado a una unidad léxica sea completamente libre, ya que, como se ha señalado, lo que una unidad léxica puede significar está limitado por su potencial semántico y por la centralidad de determinados componentes. Sin embargo, dicho límite tampoco es completamente fijo, puesto que la centralidad tiene grados y es sensible al contexto.

La sensibilidad al contexto de la centralidad se manifiesta de dos maneras distintas. Por un lado, puede haber contenido conceptual que tenga carácter central solo en determinados contextos (Croft y Cruse 2004: 102), por ejemplo, con respecto al término *arena*, la parte del concepto ARENA que se refiere a su función como material de construcción será central en un contexto de ingeniería civil, pero no en geología. Por otro lado, el propio contexto puede anular la activación de determinado contenido conceptual a pesar de su centralidad. Un ejemplo de ello sería el término *soja*: el uso de la soja como alimento humano es un componente conceptual central con alta tendencia de activación; sin embargo, en el contexto de la ingeniería energética, se activarán los componentes conceptuales que hacen referencia al uso del aceite de soja en la producción de biodiésel. Retomaremos la cuestión de las restricciones contextuales cuando tratemos la dimensión contextual de la selección de rasgos de la definición terminológica (§3.5.3).

[8] La teoría de los prototipos (§2.1.3.2) da cuenta detallada del hecho de que haya componentes conceptuales más centrales (o prototípicos) que otros.

En el siguiente apartado, nos centraremos en las distintas estructuras de conocimiento asociadas a una unidad léxica y que permiten su comprensión. Estas han recibido varios nombres, por ejemplo, como ya hemos visto, la base de Langacker (en contraposición con el perfil) en su teoría de los dominios (1987) o, como veremos, los marcos[9] de Fillmore (1982a).

Barsalou (1999: 591–592) distingue dos funciones del conocimiento enciclopédico respecto a la especificación del contenido de los conceptos. Por un lado, está la función de enmarcado, que se refiere al hecho de que los marcos aportan los conocimientos necesarios para entender un concepto dado. Por ejemplo, para entender el concepto de EVAPOTRANSPIRACIÓN es necesario tener nociones del ciclo hidrológico. Es decir, el concepto EVAPOTRANSPIRACIÓN está enmarcado forzosamente dentro del marco del CICLO HIDROLÓGICO, lo cual equivaldría a decir, en términos de Langacker (2008), que este marco tiene el mayor nivel de centralidad. Por otro lado, la otra forma en la que el conocimiento enciclopédico especifica el contenido de los conceptos es mediante lo que podemos denominar *contextualización*, que se refiere al hecho de que un concepto, al activarse en un contexto distinto, se inserta en distintos marcos y su contenido se modula en consecuencia.

Si bien estamos de acuerdo con Barsalou (1999) en que el conocimiento enciclopédico tiene esas dos funciones, consideramos que la frontera entre los marcos de fondo necesarios y los marcos de fondo que varían según el contexto es difusa, pues, la centralidad es una cuestión de grados y el contenido conceptual, dinámico. No obstante, dicha distinción puede resultar útil en el plano práctico.

[9] De ahora en adelante, utilizaremos el término *marco* de una manera más restringida que Barsalou et al. (1993) para hacer referencia al conocimiento de fondo necesario presupuesto por un concepto, en línea con Fillmore (1982a).

A continuación, desarrollaremos la teoría de la semántica de marcos que está centrada principalmente en la función de enmarcado de los marcos[10], aunque también es aplicable a la contextualización. Frente al desarrollo teórico realizado por Barsalou y colaboradores (Barsalou 1992; Barsalou y Hale 1993; Barsalou et al. 1993; Barsalou 1993) de los marcos, el cual no tenía como intención su uso como herramienta de análisis y representación semántica, la semántica de marcos destaca sobre todo por su vertiente práctica; en particular, por el proyecto FrameNet.

### 2.1.2.2 La semántica de marcos y FrameNet

La teoría que mejor representa el enfoque enciclopédico hacia el significado es la semántica de marcos (Fillmore 1976; Fillmore 1977; Fillmore 1982a; Fillmore 1985; Fillmore y Atkins 1992), que toma como punto de partida el hecho de que, para entender cualquier concepto, es necesario comprender toda la estructura de conocimiento en la que se enmarca (Fillmore 1982a: 111). Esta estructura es el marco semántico, que se define como una representación esquemática de las «estructuras conceptuales y patrones de creencias, prácticas, instituciones, imágenes, etc. que sirven de base para la interacción significativa en una determinada comunidad de habla». (Fillmore, Johnson yPetruck 2003: 235).

Según la semántica de marcos, las palabras no están relacionadas directamente entre sí, sino por medio de su asociación a un marco común y la forma en que el significado de cada palabra destaca elementos particulares de tales marcos (Fillmore y Atkins 1992: 77). No obstante, los marcos han considerarse como un prototipo (§2.1.3.2) más que como un conjunto real de supuestos de cómo es el mundo (Fillmore 1982a: 118).

En cuanto a la aplicación práctica de la semántica de marcos, Petruck (1996) expuso la necesidad de que esta abordara algunos asuntos prácticos como la determinación del contenido y los límites de los marcos y la forma en que los marcos interactúan. Para ello, surgió el proyecto FrameNet, a partir del cual, la semántica de marcos concibió un sentido más restringido de la noción de marco como herramienta de análisis léxico-semántico inspirada en la Gramática de Casos de Fillmore (1968).

[10] Mientras que para Barsalou y colaboradores los marcos pueden representar tanto los conceptos como las estructuras de conocimiento de fondo que enmarcan dicho concepto; la semántica de marcos (§2.1.2.2) restringe el uso de los marcos al ámbito del conocimiento de fondo.

FrameNet[11] es una base de datos léxica en línea para la lengua inglesa basada en la semántica de marcos. Se creó en 1997 en el *International Computer Science Institute* en Berkeley (California, Estados Unidos) (Baker, Fillmore y Lowe 1998: 86). Su objetivo principal es documentar la variedad de posibilidades combinatorias (valencias) semánticas y sintácticas de las palabras de la lengua inglesa en cada una de sus acepciones mediante la anotación asistida por ordenador de oraciones de ejemplo (Ruppenhofer et al. 2010: 5).

El componente principal de FrameNet son los marcos semánticos entendidos como una representación esquemática de una situación caracterizada mediante los roles semánticos que en ella participan, llamados *elementos del marco* (Fillmore y Petruck 2003: 359). Un ejemplo de marco es *Verification*[12], que tiene como definición en FrameNet: «An *Inspector* attains a degree of certainty in the *Unconfirmed_content*, generally by inspecting some evidence». La definición del marco en FrameNet presenta esquemáticamente un tipo de situación que subyace al contenido semántico de las palabras que evocan el marco haciendo uso de los roles semánticos (elementos del marco) (Fillmore et al. 2003: 305).

En el ejemplo de *Verification*, *Inspector* y *Unconfirmed_content* son los EM, que representan los tipos de entidad que pueden participar en los marcos y funcionan como roles semánticos de los predicados que evocan el marco (Fillmore, Johnson y Petruck 2003: 237). Por su parte, las palabras evocadoras del marco reciben el nombre de *unidades léxicas*. Una unidad léxica es una palabra tomada en una de sus acepciones (Fillmore et al. 2003: 297). Por lo general, las unidades léxicas en FrameNet son verbos, aunque también pueden ser sustantivos, adjetivos o adverbios, tanto predicativos como no predicativos. En el caso del marco *Verification*, las unidades léxicas que lo evocan son los verbos o sintagmas verbales *certify*, *confirm*, *identify*, *make sure*, *substantiate* y *verify*. También evocan el marco los sustantivos *confirmation* y *verification* y los adjetivos *unconfirmed* y *verifiable*.

[11] Disponible en: <https://framenet.icsi.berkeley.edu>.

[12] Todos los ejemplos en este apartado están tomados directamente de FrameNet.

A continuación, nos concentramos en dos cuestiones sobre FrameNet pertinentes a nuestro trabajo. En primer lugar, los marcos no eventivos y la representación en marcos de unidades léxicas no predicativas, y la representación del contenido semántico de las unidades léxicas.

#### *2.1.2.2.1 Los marcos no eventivos*

La estructura de FrameNet por estar enfocada hacia el estudio de la estructura argumental de predicados, no resulta adecuada para la caracterización de marcos que no representen eventos. A este respecto, Ruppenhofer et al. (2010: 5) argumentan que «[m]any common nouns, such as artifacts like *hat* or *tower*, typically serve as dependents rather than clearly evoking their own frames».

En esta afirmación, la noción de marco hay que entenderla desde el sentido restringido que se le da en FrameNet, pues como se ha expuesto anteriormente al tratar el enfoque enciclopédico de la lingüística cognitiva, todos los sustantivos evocan estructuras de conocimiento enciclopédico, aunque el diseño de los marcos en FrameNet no pueda caracterizar adecuadamente dichas estructuras. Los desarrolladores de FrameNet admiten que el diseño de su base de datos no es apto para la representación de entidades, por ello, prefieren derivar al usuario a WordNet para este tipo de información:

> [T]he FrameNet database is not readily usable as an ontology of things. In this area, we mostly defer to WordNet, which provides extensive coverage, including hierarchical relations of areas such as animals, plants, etc. (Ruppenhofer et al. 2010: 7)

Con todo, algunos artefactos y tipos naturales en FrameNet tienen su propio marco en los que los EM se refieren a sus subtipos, el material del que están hechos, el modo de producción, su función, etc., que como señalan Fillmore et al. (2003: 323) tiene ciertas correspondencias con los roles de *qualia* de Pustejovsky (Pustejovsky 1995). Por ejemplo, el marco *Weapon*, al que pertenecen unidades léxicas como *revolver*, *shotgun* o *sword*, tiene como EM nuclear *Weapon* (al ser un marco de entidad, el EM nuclear es la entidad en sí) y de EM periféricos *Creator*, *Descriptor*, *Material*, *Name*, *Part*, *Time_of_creation*, *Type*, *Use* y *Wielder*.

Este tipo de marcos se crean normalmente como complemento a un marco de evento y las entidades representadas en el marco de entidad suelen corresponderse con un EM del marco de evento. Por ejemplo, el marco *Weapon* está relacionado con el marco *Bearing_arms* mediante la relación *Is Used by*, y este último tiene un EM nuclear llamado *Weapon* cuyas instanciaciones son las unidades léxicas del marco *Weapon*.

Por lo general, la asignación a marcos de unidades léxicas nominales que no designan eventos es bastante inconsistente en FrameNet. Si se busca la unidad léxica *bachelor* en FrameNet, se encuentra que figura en el marco *Personal_relationship,* el cual tiene la siguiente definición: «The words in this frame have to do with people and the personal *Relationships* they are or can be a part of. Some of the words denote people engaged in a particular kind of *Relationship*, others denote the *Relationship*, yet others the events bringing about or ending the *Relationships*. Many of the words presuppose an understanding of states and events that must have occurred before another event takes place or before a person can be classified in a certain way».

Como ya presagia la definición del marco, su contenido resulta un conjunto desordenado de componentes, pues entre las unidades léxicas podemos encontrar palabras tanto predicativas como no predicativas que designan eventos, entidades o propiedades. Al consultar la entrada léxica de *bachelor* se encuentra la definición «a man who is not and has never been married», que no distingue el significado de *bachelor* de palabras relacionadas como *single*. En cuanto al resumen de las anotaciones, este muestra que *bachelor* solo se combina con los EM *Partner_1* y *Partner_2*, que están definidos en el marco como las personas participantes en una relación personal.

Así pues, podemos observar que el diseño de FrameNet no permite a los lexicógrafos representar el tipo de conocimiento enciclopédico al que aludía Fillmore en su famoso ejemplo de *bachelor* (Fillmore 1982b: 34) como el de la edad legal y la edad socialmente aceptable para contraer matrimonio que permiten entender por qué esa palabra se puede usar en algunos casos y en otros no.

##### *2.1.2.2.2 La representación del contenido semántico de las unidades léxicas*

La mayor parte del peso de la representación semántica en FrameNet recae en los marcos. Por consiguiente, ya se obtiene información semántica sobre una unidad léxica por su pertenencia a un marco. Si se desea obtener información concreta sobre una unidad léxica en particular, se puede consultar su entrada en FrameNet. En ella, se proporciona una definición, así como los esquemas de las realizaciones sintácticas y los patrones de valencia de los EM instanciados al emplearse la unidad léxica. Asimismo, se proporciona un enlace a los ejemplos anotados del corpus a partir de los cuales se han generado los esquemas de realizaciones sintácticas y los patrones de valencia.

Las definiciones de las unidades léxicas tienen un papel secundario, lo cual tiene su explicación en que el objetivo de FrameNet no es la representación del significado, si no, como ya indicamos anteriormente, documentar la variedad de posibilidades combinatorias semánticas y sintácticas de las palabras de la lengua inglesa. De hecho, en muchos casos, las definiciones proceden del *Concise Oxford Dictionary* (Ruppenhofer et al. 2010: 6).

Consideramos que la estructura de FrameNet podría reformarse para que los marcos y las definiciones se retroalimentaran mutuamente. Así, las definiciones podrían mostrar directamente cuál es la diferencia semántica entre las distintas unidades léxicas con referencia al marco que evocan, más allá de mostrar las realizaciones semánticas y sintácticas y ejemplos.

Con la exposición de la semántica de marcos y FrameNet hemos dado cuenta de las propuestas más importantes de la lingüística cognitiva en cuanto a la relación entre el significado y los marcos en el sistema conceptual. En el siguiente apartado, abordaremos el otro pilar de la organización del sistema conceptual, esto es, la estructuración taxonómica o categorización.

### 2.1.3 La categorización

La realidad se presenta como un número virtualmente infinito de estímulos. Por ello, una de las funciones más básicas de cualquier ser vivo es la de fraccionar la realidad en categorías de estímulos que no son idénticos pero que se tratan como si lo fueran (Rosch et al. 1976: 382). La categorización es uno de los procesos cognitivos más importantes del ser humano:

> Without the ability to categorize, we could not function at all, either in the physical world or in our social and intellectual lives. An understanding of how we categorize is central to any understanding of how we think and how we function, and therefore central to an understanding of what makes us human. (Lakoff 1987: 6)

De acuerdo con Cruse (2006: 29-30), una categoría se puede definir como una clase de entidades del mundo, entendiendo *entidad* en un sentido lato para incluir entidades físicas como un perro o una mesa, entidades abstractas como la paz o la derrota, así como propiedades como grande o verde, acciones como correr o estudiar, etc. Sloutsky (2003: 246) desde un punto de vista similar define categoría y categorización de este modo:

> Categories are equivalence classes of different (i.e. discriminable) entities and categorization is the ability to form such categories and treat discriminable entities as members of an equivalence class. (Sloutsky 2003: 246)

En este marco, los conceptos pueden entenderse como las representaciones mentales que almacenan conocimiento acerca de las categorías y que nos permiten asignar entidades a las categorías apropiadas (Cruse 2006: 30). Por ello, a menudo en el contexto de la categorización, los términos *concepto*, *categoría* y *categoría conceptual* suelen utilizarse intercambiablemente.

Existen dos tipos principales de categorización. Por un lado, la consideración de si un concepto (p. ej., GALLINA) es un subconjunto de un concepto más general (p. ej., AVE) y, por otro, si una entidad es una instancia de un concepto (p. ej., si una entidad en el mundo real sería categorizada como GALLINA). Dado que los experimentos han demostrado que no existen diferencias cognitivas relevantes entre estos dos tipos de categorización (Medin y Smith 1984: 116), no tendremos en cuenta esta distinción.

Para el ser humano, la habilidad de comprender el mundo a modo de categorías en vez de entidades individuales aporta grandes ventajas (Cruse 2006: 30):

1. Aprendizaje basado en la experiencia: Dado que las entidades y experiencias individuales rara vez se repiten de forma exacta, el almacenamiento de información sobre cada una de manera separada sería de poca utilidad. Sin embargo, si se agrupan entidades y experiencias similares en categorías, estas categorías podrán ser recurrentes y asociarse a una importante cantidad de conocimiento útil.
2. Comunicación: El lenguaje no podría funcionar si sus elementos no estuvieran asociados a categorías conceptuales compartidas entre los hablantes[13].

[13] La relación léxica que corresponde con la categorización se denomina *hiponimia* (M. L. Murphy 2003: 216-230). Por ejemplo, la unidad léxica *siamés* es hipónimo de *gato* o, dicho de otro modo, *gato* es hiperónimo de *siamés*.

3. Planificación: Los conceptos permiten la manipulación mental de las entidades del mundo y prever las consecuencias.
4. Economía: Lo que se aprende acerca de un miembro de una categoría puede generalizarse instantáneamente a otros miembros de la categoría. De manera inversa, también el descubrir que una entidad pertenece a una categoría concreta proporciona acceso inmediato a más información acerca de esa entidad.

Dado el papel fundamental de la categorización en la organización del sistema conceptual humano y el hecho de que definir consiste en describir un determinado contenido conceptual, no es de extrañar que ambos procesos estén íntimamente relacionados. De hecho, la teoría sobre la categorización imperante durante siglos es conocida, además de como *teoría clásica de la categorización*, por el nombre de *teoría definicional de la categorización*.

Probablemente, la manera más evidente en que se refleja esta relación entre categorización y definición es el hecho de que la mayor parte de definiciones en recursos lexicográficos y terminológicos en la actualidad hacen uso de la estructura de «género próximo y diferencias específicas» o aristotélica (§3.2), a pesar de que la teoría que fundamentaba originalmente este esquema definicional ha sido refutaba por teorías posteriores. Según este esquema el genus es una unidad léxica que representa el concepto superordinado más próximo respecto al definiéndum. Por lo tanto, el proceso de definir una unidad léxica o concepto comienza siempre por determinar su categorización.

El uso de la estructura aristotélica en la definición tiene la ventaja de resultar económica. Es decir, permite en poco espacio representar gran cantidad de información. Ello se debe a que al afirmar que un concepto pertenece a una categoría implica que el concepto comparte determinadas características con el resto de miembros de la categoría. Si no se recurre al uso de un género próximo, las definiciones resultan poco prácticas y de cuestionable utilidad. Por ejemplo, si tomamos la definición de *pedal* del

*Diccionario de la lengua española* (Real Academia Española 2001): «Palanca que acciona un mecanismo con el pie» y descomponemos su genus usando las definiciones contenidas en el propio diccionario hasta llegar a *objeto* obtendríamos una definición como la siguiente: «Objeto inanimado de una cantidad indeterminada de metal u otra materia, de forma generalmente prismática o cilíndrica y mucho más largo que grueso que suele componer un artefacto y que es inflexible, recto, angular o curvo, se apoya, puede girar sobre un punto, sirve para transmitir una fuerza y acciona un mecanismo con el pie».[14]

Ante cualquier trabajo definitorio es básico determinar qué fundamentos teóricos y metodológicos se adoptan respecto a la categorización, tanto en la manera en que se asume que esta se produce a nivel cognitivo como en el modo en que se considera adecuado gestionarla en la elaboración de definiciones.

La categorización, así como la cognición humana en su conjunto, es un objeto de estudio del que aún queda mucho por investigar y sobre el cual no existe un consenso en la comunidad científica. También conviene recordar que las teorías propuestas hasta el momento han sido presentadas para explicar la categorización del conocimiento general, por lo que conviene ser cautelosos al trasladarlas al ámbito del conocimiento especializado y al de la redacción de definiciones terminológicas.

#### 2.1.3.1 La teoría clásica

La teoría clásica recibe su nombre por dos motivos. Por un lado, se llama *clásica* porque tiene sus orígenes en la antigua Grecia y, por el otro, porque ha dominado la psicología, la filosofía y la lingüística (en especial, estructuralista y generativista) hasta bien entrado el siglo XX (Taylor 1995: 22).

[14] Este ejemplo sería una versión menos extrema de las definiciones de Wierzbicka, como, por ejemplo, las de *cup* y *mug* (1985: 33–36) que ocupan dos páginas cada una y fueron redactadas haciendo uso únicamente de primitivos semánticos (es decir, descomponiendo además del genus también todas las unidades léxicas que aparecen en la definición).

Como hemos indicado anteriormente, durante siglos, la teoría clásica de la categorización ha servido de fundamento teórico y metodológico a la práctica definitoria en diccionarios. Durante todo este tiempo, ha sido tomada como una verdad incuestionable en todos los ámbitos en los que ha sido adoptada, a pesar de no ser el resultado de estudios empíricos (Lakoff 1987: 6).

De acuerdo con la teoría clásica, todos los miembros de una categoría comparten unas características fundamentales que determinan su pertenencia a la categoría. Estas características forman la llamada *estructura definicional* de la categoría, que incluye las características necesarias y suficientes que debe poseer un concepto o entidad para ser considerado miembro de la categoría. Esa estructura recibe el calificativo de definicional porque permite demarcar claramente los límites del concepto, lo cual se consideraba que era también el fin de las definiciones lexicográficas (§3.2). En la base de la teoría se encuentra la concepción objetivista de que hay una única forma correcta de entender el mundo y el ser humano. Así pues, una definición que se ajuste a esta teoría representará esa realidad objetiva.

La teoría clásica es una teoría basada en la similitud porque lo que determina que dos elementos sean incluidos en la misma categoría es que compartan una serie de rasgos. Dentro de este marco, se considera que las categorías tienen límites precisos: si un elemento tiene las características necesarias y suficientes forma parte de la categoría; si no los tiene, no forma parte. Se trata de una cuestión de todo o nada, sin punto medio (Komatsu 1992: 502).

La teoría clásica de la categorización ha gozado de hegemonía durante tantísimo tiempo —además de porque la creencia de que los conceptos poseen características necesarias y suficientes está ampliamente extendida (McNamara y Sternberg 1983: 454)— gracias a su capacidad de explicar la inferencia analítica y la capacidad referencial de los conceptos.

Por un lado, la teoría clásica da explicación a la inferencia analítica ante un ejemplo como el enunciado «Todas las madres son mujeres». De acuerdo con la teoría clásica, el enunciado es una verdad analítica porque el concepto de MUJER está incluido dentro del de MADRE o, en otras palabras, porque pertenecer al sexo femenino es una condición necesaria para que una persona pueda ser categorizada como MADRE (Laurence y Margolis 1999: 12).

Por otro lado, la teoría clásica ofrece una explicación a la capacidad referencial de los conceptos a partir del razonamiento de que un concepto hará referencia a una entidad del mundo real siempre y cuando esta cumpla las condiciones necesarias y suficientes que forman parte de la estructura del concepto (Laurence y Margolis 1999: 14).

A pesar de su poder explicativo ante estos fenómenos, la teoría clásica ha sido refutada desde el punto de vista de su realidad cognitiva. Kintsch (1974: 230-233) demostró que las palabras consideradas definicionalmente más complejas no suponían mayor carga de procesamiento. Por ejemplo, *believe* implicaba la misma carga de procesamiento que *convince* (que es más compleja por tratarse del causativo de *believe*). Ello lleva a conjeturar que la estructura definicional que propone la teoría clásica no equivale a la organización conceptual de la mente humana.

Asimismo, con vistas a la aplicación práctica de la teoría clásica de la categorización a la definición terminológica, esta presenta numerosos problemas que hemos recogido en los siguientes subapartados.

##### *2.1.3.1.1 El problema de los rasgos necesarios y suficientes*

El problema de los rasgos necesarios y suficientes (Medin 1989: 1470; Laurence y Margolis 1999: 14-16) consiste en que, para la mayoría de conceptos, determinar de manera concluyente su estructura definicional es una tarea imposible. Basta con comparar las definiciones en distintos diccionarios del mismo concepto para darse cuenta de que los rasgos considerados definitorios no siempre concuerdan.

A menudo, se arguye a favor de la teoría clásica que, si bien una persona normal puede no saber las características necesarias y suficientes de un concepto, un experto sí las conoce. Sin embargo, ese sería únicamente el caso de nomenclaturas adoptadas internacionalmente y con una organización y principios claros. En este sentido, solo se podrían nombrar como ejemplo las nomenclaturas existentes para la zoología y la química (L'Homme 2014). Con todo, estas nomenclaturas están limitadas a un tipo concreto de entidades y no incluyen todos los conceptos asociados a dichos dominios. Por otro lado, una nomenclatura organiza el conocimiento desde un punto de vista concreto, lo cual no excluye que pudiera realizarse una categorización distinta o que los avances científicos fuercen su modificación. Así pues, se puede afirmar que el problema definicional se presenta también en los conceptos especializados, estén o no normalizados. Ello se puede observar en las definiciones extraídas de distintos diccionarios especializados de un mismo concepto. Por ejemplo, en el caso de CHLOROPLAST (*cloroplasto*), si bien parece que los distintos diccionarios están de acuerdo en determinados rasgos, en otros, hay variación:

ej. 1 CHLOROPLAST. a structure that is found in some cells of plants and contains chlorophyll. [DOEAC[15]]

ej. 2 CHLOROPLAST. the chlorophyll-containing, photosynthesizing organelle of plants.[16]

ej. 3 CHLOROPLAST. any of the chlorophyll-containing organelles (see plastid) that are found in large numbers in those plant and algal cells undergoing photosynthesis.[17]

El problema definicional es naturalmente uno de los principales escollos que presenta la teoría clásica con respecto a su aplicación a la definición. Trasladado al ámbito de las definiciones, este problema puede formularse como la imposibilidad de determinar concluyentemente la información que debe representarse en la definición como rasgos distintivos del definiéndum.

[15] *A Dictionary of Environment and Conservation* (Park y Allaby 2013).

[16] *A Dictionary of Genetics* (King, Mulligan y Stansfield 2014).

[17] *A Dictionary of Biology* (E. Martin y Hine 2014).

Uno de los ejemplos más conocidos que ilustran esta incapacidad de llegar a encontrar la estructura definicional de un concepto es JUEGO, que según Wittgenstein (1953: 66) presenta una estructura de «parecidos de familia». Es decir, cada uno de los conceptos miembros de la categoría JUEGO presenta una o varias características que comparte con varios de los otros miembros, pero no todos comparten un núcleo de características necesarias y suficientes:

> Consider, for example, the activities that we call "games". I mean board-games, card-games, ball-games, athletic games, and so on. […] [I]f you look at them, you won't see something that is common to all, but similarities, affinities, and a whole series of them at that. […] – Look, for example, at board-games, with their various affinities. Now pass to card-games; here you find many correspondences with the first group, but many common features drop out, and others appear. When we pass next to ball-games, much that is common is retained, but much is lost. – Are they all 'entertaining'? Compare chess with noughts and crosses. Or is there always winning and losing, or competition between players? Think of patience. In ball-games, there is winning and losing; but when a child throws his ball at the wall and catches it again, this feature has disappeared. Look at the parts played by skill and luck, and at the difference between skill in chess and skill in tennis. Think now of singing and dancing games; here we have the element of entertainment, but how many other characteristic features have disappeared! (Wittgenstein 1953: 66)

Asimismo, como corolario del problema definicional, se puede criticar de la teoría clásica su incapacidad para dar cuenta de las excepciones (Evans y Green 2006: 253). Por ejemplo, el concepto SEISMOMETER (*sismómetro*) aparece definido como «A device used to detect seismic waves originating from earthquakes». en el *Dictionary of Geology and Earth Sciences* (Allaby 2015). Esta definición no sería aplicable a un sismómetro que alguien utilice para detectar movimientos en la superficie terrestre originados por la actividad humana como una explosión, aunque sea el mismo aparato que utilizaría un geólogo. Tampoco podría aplicarse a un sismómetro que lleve mucho tiempo averiado y ya no se use. Sin embargo, contra lo que defiende la teoría clásica, ningún geólogo dejaría de categorizar esos sismómetros como tales, a pesar de que no presentan un rasgo que a priori debería ser necesario para ser categorizado como sismómetro.

#### 2.1.3.1.2 El problema de las categorías difusas

Si el problema definicional afectaba a las differentiae (rasgos distintivos) en una definición, el problema de las categorías difusas (Laurence y Margolis 1999: 23-24; Medin 1989: 1470) se refleja en las dificultades que presenta a menudo la selección del genus de una definición. De acuerdo con la teoría clásica, la determinación de si un elemento pertenece o no a una categoría se lleva a cabo simplemente comprobando si posee los rasgos necesarios y suficientes. Sin embargo, hay muchos casos en los que no es posible determinar de manera concluyente si un elemento pertenece o no a una categoría (Medin 1989: 1470). Por ejemplo, no existe consenso sobre si la energía nuclear debe ser categorizada como energía verde, ya que, si bien la generación de energía nuclear comporta una cantidad baja de emisión de gases de efecto invernadero, la gestión de los residuos radiactivos sigue siendo problemática.

Como indican Laurence y Margolis (1999: 24), el problema de las categorías difusas radica en el problema definicional, ya que, si no es posible determinar los rasgos definitorios de un concepto, tampoco es posible comprobar si un concepto los posee o no con vistas a su categorización.

#### 2.1.3.1.3 El problema de los efectos de tipicidad

El problema de los efectos de tipicidad (Laurence y Margolis 1999: 24-25; Medin 1989: 1470) surge ante la incapacidad de la teoría clásica de explicar por qué determinados miembros de una categoría, que cumplen los mismos requisitos para pertenecer a ella que los demás son considerados más representativos. Según la teoría clásica, todos los ejemplos de un concepto son igualmente representativos porque para pertenecer a la categoría han de poseer todos los rasgos definitorios requeridos (Medin 1989: 1470). Sin embargo, como demostró Rosch (1975a: 229), hay miembros de determinadas categorías que son considerados mejores ejemplos que otros, como es el caso de NARANJA y ARÁNDANO para la población anglohablante en Estados Unidos, pues el primero es considerado mejor ejemplo de la categoría FRUTA que el segundo.

#### 2.1.3.1.4 El problema de la multidimensionalidad

El desarrollo lógico de la teoría clásica lleva a concluir que los rasgos necesarios y suficientes de un concepto son universales y acontextuales. Sin embargo, ello choca con el fenómeno de la multidimensionalidad:

> Despite the classical theory's claim that there is only one correct way of classifying a given concept, it is commonly accepted that people can "see the same thing in different ways". We use the term *multidimensionality* to describe the phenomenon of classification that occurs when more than one characteristic can be used to distinguish between things, and hence those things can be classified in more than one way. (Bowker 1996: 784)

La multidimensionalidad tiene como consecuencia que un concepto puede pertenecer a distintas categorías al mismo tiempo así como que un mismo concepto puede tener una estructura definicional diferente según la dimensión desde la que se contemple. Ambos escenarios son habituales y se puede observar de manera evidente al comparar las definiciones terminológicas de un mismo concepto como, por ejemplo, ASBESTOS:

ej. 4 ASBESTOS**.** Any asbestiform mineral of the serpentine group (chrysotile, best adapted for spinning and the principal variety in commerce) or amphibole group (especially actinolite, anthophyllite, gedrite, commingtonite, grunerite, and tremolite).[18]

ej. 5 ASBESTOS**.** A natural material that is made up of tiny fibers used in insulations against fire and in tire brake liners.[19]

ej. 6 ASBESTOS**.** A mineral fiber that can pollute air or water and cause cancer or asbestosis when inhaled. The U.S. EPA has banned or severely restricted its use in manufacturing and construction and the ARB has imposed limits on the amount of asbestos in serpentine rock that is used for surfacing applications.[20]

ej. 7 ASBESTOS**.** A fire-resistant element that once was commonly used for insulation. It poses a lung hazard.[21]

---

[18] *TERMIUM Plus* (Translation Bureau / Bureau de la traduction [Canada] 2015)

[19] *Glossary of Cancer Terms* (The University of Texas MD Anderson Cancer Center 2015)

[20] *Glossary of Air Pollution Terms* (California Environmental Protection Agency 2015)

[21] *Canadian Financial, Real Estate and Mortgage Glossary* (The Mortgage Group 2015)

Como se puede ver en los ejemplos, el concepto ASBESTOS se ha categorizado de varias maneras: ASBESTIFORM MINERAL, NATURAL MATERIAL, MINERAL FIBER, FIRE-RESISTANT ELEMENT. Al mismo tiempo, cada definición ha destacado rasgos distintos del concepto según el dominio desde el que se ha definido el concepto. Un defensor de la teoría clásica podría argüir que las definiciones arriba mostradas son erróneas y que es necesario un análisis riguroso para obtener las características necesarias y suficientes de ASBESTOS. Sin embargo, al realizar el análisis nos toparíamos sin lugar a dudas con todos los problemas descritos en los apartados anteriores. Como consecuencia, a la teoría clásica no le quedaría más remedio que explicar el fenómeno de la multidimensionalidad como un tipo de homonimia o polisemia.

#### 2.1.3.2 La teoría de los prototipos

La teoría de los prototipos (Rosch 1975a; Rosch 1978; Rosch y Mervis 1975; Rosch et al. 1976) es una respuesta frente a los problemas que presenta la teoría clásica de la categorización. Al entroncarse dentro de la lingüística cognitiva está marcada por el rechazo al objetivismo de la teoría clásica y defiende el realismo experiencial. No obstante, comparte con la teoría clásica la idea de que lo que determina que dos elementos sean incluidos en la misma categoría es la noción de similitud, esto es, que presenten rasgos comunes.

La teoría de los prototipos se fundamenta sobre dos principios generales acerca de la formación de categorías (Rosch 1978): el principio de economía cognitiva y el principio de la estructura percibida del mundo.

##### *2.1.3.2.1 El principio de la economía cognitiva*

Según el principio de la economía cognitiva, los procesos de categorización responden a dos necesidades opuestas. Por un lado, a un ser vivo le interesa tener acceso al mayor número de propiedades posibles a partir de una sola propiedad, lo cual llevaría a la formación de un gran número de categorías con discriminaciones lo más precisas posibles entre categorías. Sin embargo, por el otro lado, uno de los objetivos principales de la categorización consiste en la reducción de las infinitas diferencias entre estímulos a proporciones abarcables desde el punto de vista cognitivo. (Rosch et al. 1976: 384). Gracias a la categorización, se consigue un equilibrio entre ambas.

Para que una categoría sea cognitivamente económica, es decir, que proporcione la mayor cantidad de información a un bajo coste cognitivo (Rosch et al. 1976: 428), debe compartir el máximo número de atributos intracategoriales manteniendo el mayor nivel de diferencia intercategorial (Evans y Green 2006: 261). Rosch calcula la economía cognitiva mediante la noción de validez de señal (*cue validity*), según la cual, un atributo adquiere relevancia para una determinada categoría cuanto más frecuentemente se asocie a miembros de esa categoría, mientras que un atributo pierde relevancia cuanto más frecuentemente se asocie a miembros de otras categorías (Evans y Green 2006: 261). Es decir, un rasgo es relevante para la categorización cuanto más frecuente y exclusivo sea de los miembros que componen una determinada categoría.

A partir de este principio, Rosch et al. (1976) encontraron la existencia de lo que llamaron el *nivel básico de categorización*, cuyo grado de abstracción da lugar a que la validez de la señal se maximice:

> Categories at higher levels of abstraction have lower cue validity than the basic because they have fewer attributes in common; categories subordinate to the basic have lower cue validity than the basic because they share most attributes with contrasting subordinate categories (Rosch et al. 1976: 428).

Por lo tanto, las categorías en la mente humana no estarían organizadas a modo de simples jerarquías taxonómicas, sino que la estructuración del conocimiento parte las categorías que están en la parte media de la jerarquía, las categorías de nivel básico (Lakoff 1987: 56).

Entre las categorías básicas estudiadas por Rosch y Mervis (1975) se incluyen CHAIR (*silla*) y CAR (*coche*). Estas categorías forman parte respectivamente de las categorías superiores FURNITURE (*mueble*) y VEHICLE (*vehículo*). Estas categorías son las llamadas *superordinadas* y se caracterizan porque sus miembros comparten solo unos pocos atributos y de tipo más abstracto que los del nivel básico. Por otro lado, dentro de la categoría básica CHAIR, se estudió KITCHEN CHAIR (*silla de cocina*) y, dentro de CAR, la categoría SPORTS CAR (*coche deportivo*). Estas se denominan *categorías subordinadas* y contienen poca información adicional con respecto a sus categorías básicas.

Las categorías superordinadas tienen una validez de señal inferior a las categorías básicas porque contienen menos atributos comunes dentro de la categoría (p. ej., la lista de atributos de FURNITURE es menor que la de CHAIR), mientras que las categorías subordinadas también tienen una validez de señal inferior porque comparten muchos atributos con otras categorías situadas en el mismo nivel de inclusividad (p. ej., SPORTS CAR y SEDAN CAR [*coche sedán*] comparten más atributos que CAR y TRUCK [*camión*] entre sí).

Una cuestión importante surgida en torno a las categorías de nivel básico es la de su carácter no universal. Estudios como el de Berlin et al. (1974) sobre la categorización de conceptos botánicos entre los tzetales contrastados con los resultados obtenidos por Rosch et al. (1976) con estudiantes universitarios estadounidenses, demostraron que el entorno cultural puede afectar a los niveles de categorización. Ante esta evidencia, Rosch et al. (1976: 393) conjeturaron sobre la posibilidad de que el nivel básico también se viese alterado como consecuencia de la adquisición de conocimiento experto:

> Would, for example, an ichthiologist, whether presented with an actual example of a category or with a fish name, have been able to list sufficient attributes specific to trout, bass, and salmon that the basic level for fish would have been placed at that level of abstraction? (Rosch et al 1976: 393)

Por ello, Tanaka y Taylor (1991) investigaron en tres experimentos distintos si había variaciones en el nivel básico dependiendo del nivel de conocimiento experto de los sujetos. Los experimentos arrojaron los siguientes resultados:

- De acuerdo con Rosch et al. (1976), los sujetos listan significativamente más atributos para la categoría de nivel básico que para la superordinada, mientras que el número de nuevos atributos añadidos para las categorías subordinadas es significativamente inferior al número añadido para las categorías básicas. En el primer experimento de Tanaka y Taylor (1991: 460-467), se demuestra que los expertos en su dominio de especialidad listan la misma cantidad de atributos nuevos para las categorías subordinadas como para las básicas.
- El segundo experimento de Tanaka y Taylor (1991: 467-470) demostró que mientras que los sujetos legos utilizan los nombres de las categorías básicas para identificar objetos, los expertos tienden a usar los nombres de las categorías subordinadas en sus campos de especialidad.
- Mientras que Rosch et al. (1976) demostraron que la asignación correcta de ejemplares a una categoría era más rápida para el nivel básico que para los otros dos. Tanaka y Taylor, en su tercer experimento (1991: 470-477), demostraron que los expertos categorizan el nivel básico y el subordinado a la misma velocidad en su propio dominio.

Tanaka y Taylor (1991: 478) llegaron a la conclusión de que el conocimiento experto no hace que las categorías subordinadas se conviertan en una suerte de segunda categoría básica, sino que simplemente se produce un aumento en la accesibilidad del nivel subordinado en los procesos de categorización. Así pues, las diferencias entre nivel básico y subordinado en cuanto a validez de señal se reducen por la adquisición de conocimiento, lo cual conlleva que el nivel subordinado alcance mayor relevancia en ámbitos especializados.

No obstante, la trascendencia de la postulación del nivel básico de categorización parece ser limitada. Si bien el descubrimiento arrojó luz sobre los procesos de categorización, existen dudas acerca de su aplicación más allá de entidades físicas. De hecho, Atran (1990) rechaza incluso que los niveles de categorización se puedan aplicar a artefactos, dado que el número de posibles categorías superordinadas es muchísimo mayor que para las entidades naturales.

#### *2.1.3.2.2 El principio de la estructura percibida del mundo*

Este principio defiende que los humanos perciben el mundo como una estructura de carácter altamente correlacional. Es decir, que los humanos perciben que determinados atributos coocurren con otros (p. ej., que los animales que tienen alas suelen también tener plumaje en vez de pelaje) y esa correlación se utiliza para organizar y formar categorías (Evans y Green 2006: 255). Rosch (1978) hace hincapié en que la categorización está basada en cómo se percibe el mundo y no en cómo es el mundo realmente. En eso, se opone a la teoría clásica, que se basa en que existe una realidad única y objetiva en la que se fundamenta la categorización:

> What attributes will be perceived given the ability to perceive them is undoubtedly determined by many factors having to do with the functional needs of the knower interacting with the physical and social environment. (Rosch 1978).

Por consiguiente, la teoría de los prototipos se alinea con el enfoque del realismo experiencial e introduce como factores en la categorización la propia corporeización (§2.1.1), así como el contexto en el que se produce la categorización. De ello se desprende, por lo tanto, que no existe una única categorización válida universal, sino que existen multiplicidad de ellas.

#### *2.1.3.2.3 La prototipicidad*

A partir de los dos principios expuestos arriba, Rosch y sus colaboradores desarrollaron la teoría de los prototipos, la cual recibe su nombre porque postula que las categorías en vez de tener una estructura como la descrita por la teoría clásica, se organizan en torno a prototipos. Las categorías según esta teoría poseen las siguientes características (Geeraerts 1989: 592-593): estructura prototípica, relaciones de parecidos de familia, estructura radial, grados de tipicidad y límites difusos. No obstante, cabe destacar que no todas las categorías presentan todas y cada unas de estas características de prototipicidad.

*Estructura prototípica*

Esta es la característica que da nombre a la teoría. Las categorías no pueden definirse por medio de un único conjunto de atributos necesarios y suficientes, sino mediante un prototipo. La noción de prototipo puede definirse como una representación compleja cuya estructura codifica un análisis estadístico de las propiedades que los miembros de una categoría suelen poseer (Laurence y Margolis 1999: 27). A pesar de que a menudo se ha empleado el término *prototipo* para hacer referencia al miembro más prototípico de una categoría, realmente no tiene por qué existir un miembro que equivalga al prototipo (Rosch 1978: 36).

*Relaciones de parecidos de familia y estructura radial*

Los miembros de una categoría no poseen forzosamente todas las propiedades del prototipo (Rosch et al. 1976: 433). Ello se explica porque entre los miembros de una categoría prototípica puede haber lo que Wittgenstein (1953) denominó *parecidos de familia*. Desarrollando la idea de Wittgenstein, Rosch y Mervis (1975: 574-575) describieron las relaciones de parecido de familia de la siguiente manera:

> A family resemblance relationship takes the form AB, BC, CD, DE. That is, each item has at least one, and probably several, elements in common with one or more items, but no, or few elements are common to all items. (Rosch y Mervis 1975: 574-575)

Por otro lado, un fenómeno relacionado al de los parecidos de familia que también se explica dentro de la teoría de los prototipos es el de las categorías radiales descritas por Lakoff (1987). Se trata de categorías que comprenden un elemento central junto con variantes convencionalizadas que no pueden predecirse mediante reglas generales (Lakoff 1987: 84). Por ejemplo, el prototipo de MADRE está compuesto por elementos procedentes de distintos subtipos de madre como la mujer que da a luz, la mujer que transmite su material genético, la mujer que cría a un niño, la mujer que está casada con el padre del niño, la mujer que tiene la custodia de un niño, etc. Una mujer que cumpla cualquiera de esas condiciones será considerada madre, pero cuantas más de ellas cumpla, más prototípica será.

*Grados de tipicidad*

Hay categorías que exhiben grados de tipicidad. Es decir, no todos los miembros de una categoría son considerados igualmente representativos de esta (Rosch 1978: 36). Rosch (1975a) realizó un experimento con estudiantes universitarios en Estados Unidos a los que pidió que puntuaran subjetivamente el nivel de representatividad de determinados miembros de unas categorías dadas. Como resultado, se obtuvo, por ejemplo, que SILLA era un buen ejemplo de MUEBLE, a diferencia de ESPEJO, o que una MUÑECA es un JUGUETE más típico que un GLOBO.

Rosch y Mervis (1975) demostraron que cuanto más prototípico es un miembro de una categoría, más atributos comparte con otros miembros de su categoría. Para ello, pidieron a un grupo de sujetos que listara los atributos de conceptos pertenecientes a distintas categorías. Cuanto más se repetía un atributo dentro de los miembros de una categoría, más puntuaban. Finalmente, la puntuación de cada concepto era la suma de la puntuación de sus atributos.

Los grados de tipicidad también se observan en los conceptos especializados. Por ejemplo, en el dominio de la oceanografía, dentro de la categoría TSUNAMI muestra más grados de tipicidad TSUNAMI MARINO que TSUNAMI LACUSTRE. Aunque ambos son miembros de la categoría, TSUNAMI MARINO resulta más representativo por el hecho de que la mayoría de tsunamis ocurren en el mar.

*Límites difusos*

Existen categorías que, además de presentar grados de tipicidad, tienen límites difusos (Rosch 1978: 35), lo cual conlleva que existan casos dudosos de pertenencia a una categoría. Un ejemplo no especializado es la categoría DEPORTE: conceptos como FÚTBOL o BALONCESTO son miembros claros de la categoría; sin embargo, asignar el concepto AJEDREZ a la categoría de DEPORTE resulta discutible. Las categorías de límites difusos también se encuentran en los campos de especialidad. Por ejemplo, la categoría de METAL: mientras que el HIERRO se considera claramente un tipo de METAL, el POLONIO es un caso dudoso.

Los grados de tipicidad no deben confundirse con el hecho de que haya categorías con límites difusos. Existen categorías que muestran grados de tipicidad y al mismo tiempo poseen límites precisos (es decir, que no hay duda respecto a qué miembros pertenecen o no a la categoría). Un ejemplo tomado de las matemáticas es la categoría de NÚMERO PRIMO. Mientras que no hay dudas respecto a qué ejemplares forman o no parte de la categoría, Armstrong, Gleitman, y Gleitman (1983: 277) demostraron que el número 3 es más representativo de la categoría que, entre otros, el número 501.

#### *2.1.3.2.4 Ventajas y problemas de la teoría de los prototipos aplicada a la definición*

Muchas de las críticas expuestas hacia la teoría de los prototipos parten de la asunción de que la teoría de los prototipos constituye una explicación acerca de la estructuración conceptual humana. Sin embargo, Rosch (1978) advierte de que los prototipos como representación de una categoría no existen y que lo único real son los efectos de prototipicidad:

> The pervasiveness of prototypes in real-world categories and of prototypicality as a variable indicates that prototypes must have some place in psychological theories of representation, processing, and learning. However, prototypes themselves do not constitute any particular model of processes, representations, or learning (Rosch 1978: 40).

Así pues, a pesar de no constituir una teoría sobre la representación conceptual, la noción de prototipo permite dar cuenta de fenómenos semánticos como el carácter flexible y dinámico del significado (Geeraerts 1989: 590-591). Asimismo, su aplicabilidad al conocimiento especializado ha sido demostrada por varios autores (Zawada y Swanepoel 1994; Weissenhofer 1995; León Araúz 2009; Temmerman 2000).

Sustituir el enfoque clásico por uno basado en prototipos supone grandes ventajas en la elaboración de definiciones terminológicas. De hecho, se puede decir que ofrece respaldo teórico a determinadas prácticas que ya se venían aplicando en la redacción de definiciones terminológicas (Seppälä 2012: 117).

La función de la definición deja de ser la de delimitar de manera precisa los límites de un concepto representando una estructura rara vez real de características necesarias y suficientes. Al perseguirse la descripción del prototipo, los rasgos que se representan en la definición son los que describen el prototipo, que no tienen por qué ser compartidos por todos los miembros de la categoría. Además, como señala Seppälä (2012: 118), frente a los atributos binarios de la teoría clásica, la teoría de los prototipos permite que los atributos asociados a un concepto tengan distintos niveles y matices.

Asimismo, como consecuencia del abandono de las características necesarias y suficientes y en línea con los principios de la lingüística cognitiva, ya no es menester distinguir entre conocimiento definicional y enciclopédico.

Finalmente, el prototipo de una categoría no es universal, sino que puede haber distintos prototipos según el contexto. De este modo, los rasgos relevantes para describir un concepto varían dependiendo de las dimensiones desde las que se pueda activar el concepto. Así pues, un concepto no tendría una única definición válida, sino que podrá tener varias.

Sin embargo, a pesar de sus ventajas, la teoría de los prototipos también presenta problemas en su aplicación a la definición terminológica.

*El problema de la inducción a error*

A partir de una definición basada en prototipos, el receptor de la definición podría categorizar erróneamente un concepto porque este posea mucha de las características del prototipo descrito. Al mismo tiempo, también existe el riesgo de que excluya de una categoría erróneamente un concepto porque considere que no comparta suficientes características con el prototipo.

Sin embargo, este problema no es exclusivo de las definiciones basadas en prototipos, ya que las definiciones clásicas también lo padecen. Dado que la estructura definicional de rasgos necesarios y suficientes no es real para la mayoría de categorías, una definición de este tipo tampoco puede asegurar que el receptor vaya a realizar categorizaciones correctas a partir de la información contenida en la definición. De hecho, dado que una definición basada en prototipos no limita el tipo ni cantidad de rasgos que se representan en la definición y permite adaptarse al contexto, es más probable que una definición prototípica induzca a menos errores.

*El problema de la herencia de propiedades*

La teoría clásica postulaba que si un concepto es subordinado de otro, este hereda las propiedades del primero y ello permitía una explicación sencilla y computacionalmente viable de la inferencia analítica (salvando el resto de problemas de la teoría). Por su parte, la teoría de los prototipos da solución a muchos de los problemas de la teoría clásica, pero, como contrapartida, la herencia de propiedades basada en prototipos no resulta sencilla. Dado que no todos los miembros poseen los rasgos típicamente asociados a la categoría, la herencia de rasgos característicos debe ser selectiva.

Si bien la herencia de propiedades selectiva puede resultar problemática computacionalmente, la teoría de los prototipos ha demostrado que para una persona no supone contradictorio que el concepto de PINGÜINO esté subordinado al de PÁJARO, a pesar de que el pájaro prototípico vuele y el pingüino no.

Con vistas a la definición, para representar este tipo de anulación de la herencia de rasgos, se pueden representar rasgos negativos en la definición. Así, por ejemplo, si la definición de PINGÜINO tiene PÁJARO como genus, uno de sus rasgos podría ser «no vuela».

Esto comporta que cada concepto ha de trabajarse más individualmente que en el enfoque clásico en el que se hereda información que a priori no necesita ser revisada.

*El problema de la composicionalidad*

Una crítica habitual a la teoría de los prototipos es el problema de la composicionalidad (Laurence y Margolis 1999: 35): los prototipos de conceptos complejos no son normalmente la función de los prototipos de los conceptos que lo componen. Este problema lo ilustran Fodor y Lepore (1996: 263) con el ejemplo de PET FISH (*pez mascota*):

> [K]nowing that PET and FISH have the prototypes that they do does not permit one to predict that the prototypical pet fish is more like a goldfish that like a trout or a herring, on the one hand, or a dog or a cat, on the other. (Fodor y Lepore 1996: 263)

Desde el punto de vista de la definición, es importante tener en cuenta que efectivamente el prototipo de un concepto compuesto puede no ser el resultado de unir la definición de sus componentes. Por lo tanto, se hace necesario analizar individualmente también los conceptos compuestos con vistas a su definición.

*El problema de las fuentes de tipicidad*

Barsalou (1985) demostró que las categorías conceptuales no solo presentan grados de tipicidad en torno a una representación prototípica que incluye los rasgos más compartidos por los miembros de la categoría. Según el tipo de categoría, los grados de tipicidad también pueden provenir de la formación de un ideal, la frecuencia de instanciación y la familiaridad.

Los ideales normalmente se forman cuando una categoría tiene una función asociada. Consisten en una representación de las características que los miembros de la categoría deben poseer para cumplir óptimamente la función en cuestión (Barsalou 1985: 630). Por lo general, los ideales suelen ser valores extremos que solamente cumplen algunos miembros de la categoría o incluso ninguno. Barsalou (1985: 640) demostró que los grados de tipicidad de la categoría ARMA están fuertemente influidos por un ideal basado en la máxima efectividad, a saber, en el cumplimiento de la función de herir o matar. Asimismo, también expuso la influencia de un ideal, en este caso basado en cuánto gustan a las personas determinados miembros, en categorías de entidades naturales como FRUTA y PÁJARO.

Por su parte, la familiaridad se define como la estimación subjetiva por parte de una persona respecto a la frecuencia con la que han experimentado una categoría en cualquier tipo de contexto (Barsalou 1985: 631). En los experimentos llevados a cabo por Malt y Smith (1982), se encontró que la familiaridad podría tener tan solo una ligera influencia en la estructura de tipicidad de las categorías.

Sin embargo, la frecuencia de instanciación, es decir, la estimación subjetiva por parte de una persona respecto a la frecuencia con la que han experimentado un concepto como miembro de una categoría específica, es mucho más importante que la familiaridad para la configuración de los grados de tipicidad (Barsalou 1985: 631). De hecho, los mismos conceptos contrastados frente a distintas categorías presentan diferentes grados de tipicidad (Smith, Shoben y Rips 1974). Por ejemplo, mientras que MOSCA es un miembro típico de la categoría INSECTO, respecto a la categoría ANIMAL es menos típico. Es decir, por un lado, cuando se utiliza el concepto INSECTO es más frecuente su instanciación como MOSCA que, por ejemplo, como LUCIÉRNAGA o TERMITA. Por el otro lado, PERRO o GATO son miembros más típicos de la categoría ANIMAL que MOSCA.

Finalmente, cabe destacar que Lakoff (1987: 85-90) señala otros factores que pueden afectar a la estructura de tipicidad de una categoría. Los más relevantes son los siguientes:

- Estereotipos sociales: Se trata de expectativas culturales respecto a las instancias de una categoría (Lakoff 1987: 85). Un ejemplo de estereotipo social sería la descripción de los andaluces como personas alegres y graciosas. Los estereotipos sociales a menudo no se ajustan a la realidad y son controvertidos.
- Parangones y ejemplos relevantes: Lakoff (1987: 87-89) arguye que el entendimiento de determinadas categorías se basa en instancias reales que representan un ideal (o su contrario) o simplemente un ejemplo conocido o memorable. Lakoff nombra el experimento de Michelson y Morley como parangón de experimento en el dominio de la Física, a partir del cual muchas personas entienden lo que es un gran experimento científico. En cuanto al ejemplo relevante, se podría afirmar que, para muchas personas, el concepto de accidente nuclear está fuertemente influido por los ejemplos de Chernóbil y Fukushima. Los parangones y los ejemplos relevantes, si bien facilitan la adquisición de conocimiento simplificando la realidad, pueden originar que una persona realice generalizaciones erróneas a partir de casos individuales.
- Generador: Los generadores son miembros centrales de una categoría a partir de los cuales, en conjunción con una serie de reglas, se generan los demás miembros de la categorías (Lakoff 1987: 88). Por ejemplo, los números de una sola cifra generan el resto de números, por eso aquellos son miembros más típicos de la categoría de los números naturales.
- Submodelo: Los submodelos equivalen a los puntos de referencia cognitiva de Rosch (1975b). Un ejemplo de submodelo, de nuevo en la categoría de los números naturales, son los múltiplos de diez que se utilizan como puntos de referencia en distintas operaciones cognitivas y, por tanto, tienen un lugar más cercano al centro de la categoría.

La información procedente de estereotipos sociales, parangones y ejemplos relevantes no suele formar parte de las definiciones lexicográficas, y especialmente tampoco de las terminológicas. A pesar de que integran el conjunto de rasgos que las personas asocian a ciertos conceptos, a menudo simplifican la realidad en exceso o la deforman. Por ello, puede ser útil tenerlos en cuenta como fuente de subjetividad en la extracción de conocimiento, pero su representación en la definición de un concepto no tendría lugar[22].

En cuanto los generadores y los submodelos, además del hecho de que son fenómenos minoritarios, la constatación de su existencia no es a priori especialmente relevante para la definición terminológica.

En definitiva, es evidente que los grados de tipicidad no proceden únicamente de los parecidos de familia. Además, como demostró Barsalou (1985: 645) la estructura de tipicidad de una categoría puede estar determinada simultáneamente por distintos factores según el contexto. Por lo tanto, las definiciones terminológicas basadas en la teoría de los prototipos habrán de dar cuenta de que hay categorías que cuyo miembro central no es un prototipo, sino de otra naturaleza y que ello dependerá del contexto.

*El problema de la similitud*

Como hemos visto, la similitud entre dos entidades se mide según la cantidad de rasgos comunes y el peso específico que se da a cada rasgo. De acuerdo con la noción de validez de señal, al categorizar, el peso específico de cada rasgo viene determinado por su frecuencia de aparición dentro de la categoría y no aparición fuera de ella. Sin embargo, el problema surge porque la teoría de los prototipos no explica qué rasgos, propiedades o atributos hay que tener en cuenta en un análisis de similitud y cuáles no (G.L. Murphy y Medin 1985: 292). Como ejemplifica G.L. Murphy (2002: 174), la lista de atributos de un concepto es virtualmente infinita:

> [R]obins have two legs, but they also have a leg; they also have fewer than three legs, and fewer than four legs, and so on. Robins have a red breast; or, one could say that they have a red chest and red belly. Which, if any of these, should be features? Why is it that red breast gets to be the feature, rather than red chest plus red belly? Why isn't having fewer than four legs a feature? And I saw a robin on my lawn this morning. Should this fact ("found on my lawn this morning") become a feature of robins? Or should the property "robins can be found on lawns" or "robins can be seen in the morning" or "robins can be found in the morning in the United States" . . . or . . . "in the midwest" or "in Illinois?" (G. L. Murphy 2002: 174)

[22] Los estereotipos sociales, parangones y ejemplos relevantes no deben confundirse con la connotación de las unidades léxicas. Mientras que la connotación es un fenómeno lingüístico (p. ej., *politicucho* es una palabra con connotación negativa para hacer referencia al concepto POLÍTICO), mientras que el un estereotipo es un fenómeno conceptual (el estereotipo de POLÍTICO tiene asociado los rasgos de corrupto y deshonesto).

El problema persiste al compararse dos conceptos, ya que si se hace una lista de los atributos que dos conceptos cualesquiera tienen en común, esta será virtualmente infinita:

> Suppose that one is to list the attributes that *plums* and *lawnmowers* have in common in order to judge their similarity. It is easy to see that the list could be infinite: Both weigh less than 10,000 kg (and less than 10,001 kg,…), both did not exist 10,000,000 years ago (and 10,000,001 years ago,…), both cannot hear well, both can be dropped, both take up space, and so on. Likewise, the list of differences could be infinite. (G.L. Murphy y Medin 1985: 292)

A partir de estos ejemplos, queda patente que la noción de similitud por sí sola no puede explicar por qué se forman unas categorías y otras no. Este problema es quizá el más importante en relación con la redacción de definiciones, pues la selección de rasgos que se van a representar en una definición queda sin justificar.

En el próximo apartado, tratamos la teoría de la teoría que corrige algunas de las deficiencias de la teoría de los prototipos y ofrece una guía sobre la selección de rasgos en las definiciones que se complementa con la noción de similitud.

### 2.1.3.3 La teoría de la teoría

La idea principal en la que está fundamentada la teoría de la teoría es que la organización mental de los conceptos está basada en el conocimiento y guiada por teorías mentales acerca del mundo (Medin 1989: 1474). La noción de teoría ha de entenderse en este contexto como los conocimientos de fondo o las explicaciones que posee una persona respecto al mundo que le rodea, es decir, se trata de un complejo conjunto de relaciones entre conceptos (G.L. Murphy y Medin 1985: 290). Dichas teorías pueden ser contradictorias y erróneas y los seres humanos las poseemos tanto para el conocimiento de sentido común como para el conocimiento especializado.

La teoría de la teoría (G.L. Murphy y Medin 1985; G.L. Murphy 1993; G.L. Murphy 2000) surge como reacción ante las teorías probabilísticas, como la teoría de los prototipos o la teoría de los ejemplares (Smith y Medin 1981). En especial, la crítica se centra en el uso que estas teorías hacen de la noción de similitud. La teoría de la teoría no rechaza dicha noción, simplemente aduce que la similitud entre los miembros de una categoría no es lo que hace que dicha categoría sea coherente, sino al revés, dado que la categoría es coherente, sus miembros parecen similares (G.L. Murphy y Medin 1985: 291). Dicha coherencia la aportan las teorías mentales, lo que supone la expansión de los límites de la representación conceptual, que ya no se limita a una lista de propiedades. En otras palabras, para explicar la coherencia conceptual, los procesos que operan sobre un concepto y las relaciones que este tiene con otros conceptos deben tenerse en cuenta además de la información directamente almacenado en él (G.L. Murphy y Medin 1985: 297). De acuerdo con la teoría de la teoría, los conceptos son representaciones cuya estructura consiste en sus relaciones con otros conceptos especificada por una teoría mental (Laurence y Margolis 1999).

La consecuencia principal de este enfoque para la manera en que se entiende la categorización es que esta deja de concebirse como un proceso en el que simplemente se compara un elemento con una lista de propiedades (Laurence y Margolis 1999: 43). Nuestro conocimiento del mundo ejerce presión para que los nuevos conceptos que se adquieran sean consistentes con dicho conocimiento previo (G. L. Murphy 2002: 60). En otras palabras, se busca una relación explicativa con las teorías en la que se enmarcan los nuevos conceptos (Medin 1989). Así pues, son esas teorías las determinan cuáles son las propiedades que se van a utilizar en el análisis de similitud.

Un caso ilustrativo es cuando un individuo hace una lista de los atributos de un concepto. Al hacerlo, no expresa todo lo que sabe del concepto, sino que nombra los atributos que son relevantes según su conocimiento de fondo y la situación (G.L. Murphy y Medin 1985: 299), ya que, como vimos anteriormente, dicha lista podría ser infinita. Un ejemplo es el atributo INFLAMABLE, que si bien lo poseen muchas entidades, solo sería relevante en la representación de algunas:

> [M]ost people realize, upon reflection, that the attribute, "flammable," applies to wood, money, certain plastics, and (sadly) even animals. Yet, it probably would be found only in the conceptual representation (and the listings) for the first of these categories, presumably because of the known role of wood in human activities. Some attributes are prominent in our concepts because of their importance in our other knowledge about the world, and others are excluded because of their irrelevance to our theories. The concept *money* is central to our theories of economic and social interaction, in which the attribute of flammability plays no role. Thus, it is apparently not part of our representation of money even though it may easily be inferred as true of most money. (G.L. Murphy y Medin 1985: 299-300)

Igualmente, al adquirir un concepto y categorizarlo, los rasgos que se le asociarán dependerán también del conocimiento de fondo que se aplique. Por ejemplo, si un ornitólogo se encuentra con un ave enjaulada de una especie que no conoce, prestará probablemente atención al color de su plumaje, la forma de su pico y su tamaño, y lo incorporará al concepto. Sin embargo, no incorporará al concepto la hora del día en que vio el ave o en qué parte de la jaula se encontraba (G.L. Murphy 2002: 147). Su conocimiento previo hará que se elija unos rasgos sobre otros para configurar su concepto sobre esta nueva especie. Sin embargo, en otro contexto o en el caso de otros conceptos, los rasgos variarán.

Este razonamiento es aplicable a la comparación de dos elementos con fines categorizadores. De la infinita lista de atributos compartidos, las teorías mentales que se apliquen, acotarán a qué atributos se va a limitar el análisis de similitud.

### *2.1.3.3.1 Las categorías derivadas de un fin*

A diferencia de la teoría de los prototipos, la teoría de la teoría es capaz de explicar la coherencia de una categoría en ausencia de similitud aparente entre conceptos. Ello se observa claramente en las llamadas *categorías derivadas de un fin* (CDF) de Barsalou (1983; 1985; 1987). Estas categorías se construyen espontáneamente para alcanzar un fin relevante en una determinada situación, aunque, si se utilizan frecuentemente pueden terminar arraigándose en la memoria (Barsalou 2010: 86)[23]. Ejemplos de CDF son ACTIVIDADES TURÍSTICAS QUE REALIZAR EN BARCELONA, COSAS QUE LLEVAR A UNA ACAMPADA, etc.

La estructura prototípica de estas categorías suele formarse en torno a un ideal. Dado que las CDF son creadas para cumplir un fin, los grados de tipicidad vienen marcados el ideal que maximiza el cumplimiento del fin (Barsalou 1985: 633). Un conocido ejemplo de CDF es la de COSAS QUE COMER DURANTE UNA DIETA. Cuando una persona crea esta categoría hace uso de su conocimiento previo para encontrar los miembros de la categoría, como, por ejemplo, el hecho de que una dieta se hace con el objetivo de perder peso, que la pérdida de peso se suele obtener reduciendo la cantidad de calorías ingeridas, etc. Por ello, el ideal en torno al que se crea la categoría es el de «cero calorías» y los miembros más prototípicos de la categoría serán aquellos que más se acerquen al ideal.

Este tipo de categorías también se dan en los dominios especializados. En agricultura, una CDF podría ser PLAGUICIDAS CONTRA LA ERINOSIS DEL OLIVO, creada mentalmente, por ejemplo, por un ingeniero agrónomo al encontrarse un caso de erinosis en unos olivares. Para conformar esta categoría es necesario saber que la erinosis es una plaga provocada por un tipo de ácaro y tener conocimiento sobre los distintos tipos de acaricidas disponibles. En la mayoría de casos, el ideal será el acaricida más efectivo contra la erinosis en el olivo. No obstante, dependiendo del contexto, el ideal varía y podría ser, por ejemplo, el acaricida con la mejor relación coste-eficacia, el acaricida más barato o el acaricida más fácil de obtener. En todos esos casos, el conocimiento desempeña un papel primordial.

[23] Las CDF que no están arraigadas en la memoria reciben el nombre de *categorías ad hoc*. La noción de CDF incluye tanto aquellas como las que sí están arraigadas (Barsalou 1985: 632).

Las CDF se caracterizan por violar la estructura correlacional por el hecho de que a menudo sus miembros pertenecen a distintas categorías taxonómicas (Barsalou 1985: 632). Por ejemplo, la categoría COSAS QUE LLEVARSE A UNA ACAMPADA incluye miembros muy distintos entre sí como COMIDA, ROPA, UTENSILIOS, etc. La violación de la estructura correlacional queda patente en el hecho de que sin contexto, sujetos en estudios experimentales encontraron difícil identificar las categorías al presentársele sus miembros (Barsalou 1983: 223). Así pues, lo que hace que una CDF sea coherente son las estructuras de conocimiento en la que se enmarca la categoría que se deriva del fin por el que la categoría se ha creado.

##### *2.1.3.3.2 Ventajas y problemas de la teoría de la teoría aplicada a la definición*

Como ya hemos apuntado anteriormente, la principal ventaja de la teoría de la teoría es que complementa la teoría de los prototipos aportando un factor explicativo a la selección de rasgos en el seno de los análisis de similitud. Esta cuestión es de gran importancia para la creación de definiciones terminológicas, pues de la selección de los rasgos que se representan en ellas depende en gran medida su utilidad.

La aplicación de los principios de la teoría de la teoría a la definición terminológica supone que el terminólogo, en la selección del genus y las differentiae, debe estar guiado por los conocimientos previos del receptor y por los conocimientos de fondo necesarios para entender el concepto que se define dentro del contexto en el que se enmarque la definición.

Por otro lado, la teoría de la teoría, al reconocer la importancia del conocimiento previo en la categorización y la adquisición de conceptos consigue dar respuesta a los problemas de la composicionalidad y de las fuentes de tipicidad de la teoría de los prototipos.

En lo que respecta a la composicionalidad y retomando el ejemplo de PET FISH, cabe destacar que mientras que el prototipo de PET FISH podría explicarse como el prototipo de la intersección entre PET y FISH, hay muchos otros conceptos compuestos que no se pueden explicar de esa manera. De hecho, G.L. Murphy y Medin (1985: 306) defienden que sin tener en cuenta el conocimiento de fondo no es posible predecir cómo se combinan dos conceptos para formar uno compuesto:

> An *expert repair* is a repair done by an expert, but an *engine repair* is probably not a repair done by an engine. So, no single relation (like set intersection) can describe all or even most compound concepts. Furthermore, the construction of complex concepts is not a simple operation on the features of the two concepts, such as feature overlap or projection. Although some of the features of *finger* get carried over onto *finger cup*, considerable knowledge is needed to specify which features are affected and how they are combined with the features of *cup*. Whenever people form complex concepts or understand compound nouns, they must be using their background knowledge of the way the world works in order to create the correct concept. (G. L. Murphy y Medin 1985: 306)

Por ejemplo, en el dominio de la ingeniería civil, encontramos los conceptos de POZO DE HINCA y POZO DE BOMBEO. En el primer caso, HINCA añade al concepto de POZO la información relativa al modo de construcción, mientras que en el segundo caso, BOMBEO se refiere a la función asignada al pozo. En ambos casos, sin el recurso a las teorías, no es posible predecir la interacción entre los conceptos que componen el concepto complejo. La consecuencia principal de este hecho es que las definiciones de conceptos complejos no se pueden obtener automáticamente a partir de las definiciones de los conceptos simples, pues requieren un análisis individual del conocimiento de fondo en que se enmarcan los conceptos.

En lo que respecta a las fuentes de tipicidad, ya hemos visto anteriormente que la representación central de una categoría puede estar determinada por una estructura de parecidos de familia, un ideal, la frecuencia de instanciación o la familiaridad. De nuevo, es precisamente el conocimiento de fondo y el contexto de activación los que determinan cuáles de estos factores serán los determinantes en cada caso.

Finalmente, cabe destacar que la teoría de la teoría no consigue solventar los problemas de la inducción a error y de la herencia de propiedades de la teoría de los prototipos, aunque las repercusiones de estos problemas para la definición terminológica son algo más limitadas en este caso.

En las definiciones creadas aplicando la teoría de la teoría sigue existiendo la posibilidad de que se induzca al receptor a error al no realizar este una categorización correcta a causa de inferencias erróneas. Sin embargo, si el terminólogo se ha guiado por los conocimientos previos del receptor y por los conocimientos de fondo necesarios según el contexto, las probabilidades de inducción a error se reducen. Además, hacer que la definición sea más coherente con los conocimientos previos del receptor incrementa la adquisición de conocimiento, pues G. L. Murphy y Wisniewski (1989) demostraron que una persona aprende más fácilmente las categorías que le resultan coherentes que las incoherentes.

Asimismo, una desventaja concreta de la teoría de la teoría es que si bien esta demuestra la importancia del conocimiento en la categorización y demuestra sus efectos, no aporta herramientas para su aplicación práctica, por ejemplo, en la definición terminológica.

#### 2.1.3.4 Modelo aplicable a la definición terminológica basado en prototipos y teorías

Como ya hemos apuntado anteriormente, la aplicación de la teoría de los prototipos a la definición terminológica aporta numerosas ventajas frente a la teoría clásica. Sus implicaciones son tan importantes que coincidimos con Seppälä (2012: 116) en afirmar que la teoría de los prototipos no solo redefine la noción de concepto sino la de definición.

Al aplicar la teoría de los prototipos, definir un concepto consistirá en situarlo dentro de una categoría apropiada de acuerdo con el contexto y representar los rasgos que la mayoría de los miembros de esa categoría comparten. Es posible indicar si algunos de los rasgos representados en la definición son compartidos solo por parte de los miembros (ej. 8), tiene algún grado (ej. 9) o le son aplicables restricciones de tipo temporal (ej. 10) o geográfico (ej. 11):

ej. 8 MACROBENTHOS. The larger organisms of the benthos, generally longer than 0.5 millimetres.[DOEAC]

ej. 9 DINOCAP. A dinitro fungicide and acaricide (2,6-dinitro-octylphenyl crotonates) used on fruit, vegetable, and ornamental crops. It is slightly to moderately toxic to humans but very toxic to fish. [DOEAC]

ej. 10 DEW POND. An artificial pond created by farmers to provide water for livestock on hills or other places with no natural water source. It originally consisted of a shallow excavation lined with impermeable clay. Nowadays a plastic membrane is normally used. […].[24]

ej. 11 MARRAM GRASS. Two species of Ammophila, which are grasses with an extensive, tightly knit underground root system which can survive in the relatively salty, nutrient-poor free-draining sand that is typical of beach dunes. They are native to Europe and North Africa (but noxious weeds in California) and are used to stabilize dunes. [DOEAC]

Introducir este tipo de información en una definición basada en prototipos puede ser útil en algunos casos. No obstante, no siempre será necesario porque el hecho de que la definición esté basada en un prototipo, ya implica que no todos los rasgos estarán siempre presentes en todos los miembros o instancias de la categoría. Además, dada la forma en que categorizan los seres humanos, el receptor de una definición está abierto a la existencia de conceptos que no comparten todos los rasgos del prototipo de la categoría a la que pertenecen (como, p. ej., PINGÜINO respecto a PÁJARO) o de instancias reales que, por diversos motivos, presenten excepciones (un perro que ha perdido una pata, un fruto cuyo tiempo de maduración ha sido alterado mediante modificación genética o un

[24] *A Dictionary of Weather* (Dunlop 2008).

instrumento al que se le cambia una característica para resultar más eficiente, etc.).

La aplicación de la teoría de la teoría a la definición terminológica permite aplicar criterios para la selección de los rasgos que representar en la definición. Como ya indicamos en el apartado anterior, la teoría de la teoría demuestra la importancia del conocimiento en la categorización y sus efectos, pero no aporta herramientas para su aplicación práctica en la definición terminológica. Para ello, vamos a recurrir al empleo de marcos y plantillas de categoría para la elaboración de definiciones terminológicas de acuerdo con la terminología basada en marcos (§2.2).

## 2.2 LA TERMINOLOGÍA BASADA EN MARCOS

### 2.2.1 Antecedentes

La terminología surge como campo de conocimiento a partir de la tesis doctoral de Eugen Wüster (1931). Más adelante, el propio Wüster y sus seguidores, conocidos como la Escuela de Viena, fueron desarrollando unos principios teóricos que quedaron plasmados en distintas publicaciones y, especialmente, en la obra póstuma de Wüster de 1979, en la que se compendia su teoría terminológica. Sus sucesores continuaron con dicha teoría, la cual empezó a conocerse como la teoría general de la terminología (TGT). Para la Escuela de Viena, la actividad terminológica es de carácter prescriptivo y está orientada a la normalización. De acuerdo con la TGT, la terminología se centra en la delimitación rígida de los conceptos y la asociación a cada uno de un único término evitando la sinonimia y la polisemia, con el objetivo de asegurar la univocidad de la comunicación profesional, especialmente en el contexto de la comunicación internacional.

Aunque, durante muchos años, la TGT ha sido el único conjunto de principios teóricos existentes para la terminología, en las últimas décadas numerosos especialistas en este campo han desechado las propuestas teóricas de Wüster proponiendo nuevos principios y enfoques.

A continuación, revisaremos el desarrollo que ha experimentado la terminología a partir de las primeras teorías que desafiaron la hegemonía de la TGT hasta llegar a la más reciente de las teorías terminológicas: la terminología basada en marcos, en la que se basa este trabajo.

Las primeras teorías surgidas en respuesta a la TGT fueron la socioterminología (§2.2.1.1) y la teoría comunicativa de la terminología (§2.2.1.2). Ambas presentan una visión de la terminología más realista, ya que son de carácter descriptivo y se centran en cómo los términos se emplean en contextos comunicativos reales. Describen las unidades terminológicas en el discurso y analizan las condiciones sociológicas y discursivas que originan los distintos tipos de textos (Faber 2009: 113).

#### 2.2.1.1 La socioterminología

La socioterminología (Boulanger 1991, 1995; Gambier 1987; Gaudin 1990, 1993, 2003) es una aproximación sociolingüística a la terminología que nació en la década de 1980 conjuntamente en Quebec y en Francia, aunque su desarrollo teórico explícito, se llevó a cabo durante la siguiente década (Gaudin 2003: 12). La socioterminología defiende que la terminología ha de dedicarse al estudio de los términos insertos en su contexto social particular:

> Cette discipline [la terminologie] a en charge l'étude de termes, c'est-à-dire de vocables servant à véhiculer des significations socialement réglées et insérées dans des pratiques institutionnelles ou des corps de connaissances (Gaudin 2003: 11).

Al dirigir la atención hacia el uso real de la lengua, se concibe el trabajo terminológico descriptivo y, por ende, se integra en él la variación terminológica —combatida por la TGT—. Además, según la socioterminología, en el trabajo terminológico descriptivo ha de incluirse también el estudio de la diacronía, pues, como afirma Gaudin (2003: 38), «ce qui fonde la spécificité des termes, leur caractérisation par des valeurs exactes, oblige à considérer leur historicité car ces valeurs exactes attachées aux termes sont provisoires et révisables».

La concepción del trabajo terminológico como descriptivo es una de las innovaciones más importantes introducidas por la socioterminología, ya que abrió la puerta a descubrir la realidad de la comunicación especializada en contraste con el universo cerrado de los enfoques tradicionales, donde los conceptos están perfectamente delimitados de forma artificial y asociados unívocamente a un solo término. De ahí que se integre en el trabajo terminológico la variación denominativa, pero también la conceptual, la cual ponen de relieve al proponer el estudio diacrónico de los términos.

Asimismo, también es destacable que la socioterminología defiende que el conocimiento especializado no está ordenado en compartimentos estancos como defendía la TGT, sino que está organizado en forma de nodos de conexiones, tal y como expuso Gambier en su teoría nodal:

> Une science, une technique réfère à d'autres sciences, d'autres techniques — elles mêmes branchées sur d'autres. […] Il n'y a pas de 'domaine' sans 'domaines' connexes : un 'domaine' revient alors à un nœud de connexions — d'autant plus ouvert, instable, que le savoir est nouveau, en cours de constitution, sans définition consensuelle […] (Gambier 1991: 37 *apud* Boulanger 1995: 198).

La teoría nodal resulta del estudio descriptivo de la comunicación especializada que pone de manifiesto que el conocimiento está en continua evolución y que un dominio no es más que una interconexión de dominios.

Aunque la socioterminología no aspira a constituir una teoría independiente, su importancia reside en el hecho de que abrió la puerta a otras teorías descriptivas de la terminología, que también tienen en cuenta factores sociales y comunicativos y que basan sus principios teóricos en la forma en que los términos se utilizan en el discurso especializado (Faber 2009: 114).

### 2.2.1.2 La teoría comunicativa de la terminología

La teoría comunicativa de la terminología (TCT) (Cabré y Feliu 2001; Cabré 1999, 2000a, 2000b; Lorente 2001) es una propuesta teórica de base lingüístico-comunicativa que surge a mediados de la década de 1990 y parte de la revisión de los principios teóricos de la TGT (Cabré 2001b: 22).

El objeto de estudio de la TCT son las unidades terminológicas propiamente dichas, las cuales forman parte del lenguaje natural y de la gramática de cada lengua. De acuerdo con Cabré (Cabré 1999: 102), la competencia general y la especializada del sujeto se encuentran integradas e incluyen unidades léxicas que, fuera de contexto, no son ni palabras ni términos. Los términos son módulos de rasgos asociados a las unidades léxicas, que se describen como unidades denominativo-conceptuales, dotadas de capacidad de referencia y que pueden ejercer funciones distintas. Así pues, las unidades terminológicas solo son potencialmente términos o no términos y pueden pertenecer a ámbitos distintos. El carácter de término lo activan en función de su uso en un contexto y una situación adecuados (Cabré 1999: 123). La TCT sostiene que el saber es un contínuum y que la segmentación de materias es una cuestión funcional:

> El valor de un término se establece por el lugar que ocupa en la estructuración conceptual de una materia de acuerdo con los criterios establecidos en un trabajo. Cada ámbito puede ser estructurado desde diferentes perspectivas y en diferentes concepciones, así como cada objeto temático puede ser abordado desde ámbitos y perspectivas distintos. Un concepto puede participar en más de una estructura con el mismo o diferente valor. Los términos *no pertenecen a un ámbito* sino que *son usados en un ámbito* con un valor singularmente específico (Cabré 1999: 124).

Cabré (1999: 100) sostiene que para poder dar cuenta de la terminología en toda su complejidad, se requiere un conjunto de teorías no contradictorias que trate los distintos aspectos de las unidades terminológicas. Por lo tanto, se concibe la terminología como un campo interdisciplinar que recibe aportaciones de tres teorías (Cabré 1999: 100-101):

- una teoría del conocimiento que explique cómo se produce la conceptualización de la realidad, los tipos de conceptualización, la relación entre conceptos y la relación de estos con sus posibles denominaciones por medio de las unidades de conocimiento y, dentro de ellas, de las unidades de conocimiento especializado (Cabré 2001a: 23);
- una teoría de la comunicación que describa explícitamente los tipos de situaciones que pueden producirse, que dé cuenta de la correlación entre tipo de situación y tipo de comunicación, y que explique las características, posibilidades y límites de los diferentes sistemas de expresión de un concepto y de sus unidades;
- una teoría del lenguaje que dé cuenta de las unidades de significación especializada dentro del lenguaje natural teniendo en cuenta que participan de todas sus características, pero singularizando su carácter terminológico y explicando cómo se activa este carácter en la comunicación. Las use incluyen las ut propiamente dichas (Cabré 2001a: 23).

A su vez, la teoría del lenguaje tiene que dar cuenta de tres vertientes de las unidades terminológicas (Cabré 1999: 101):

- de sus características gramaticales, semánticas y textuales, incluyendo todos sus aspectos gramaticales, semántico-cognitivos y pragmáticos;
- del uso que hacen los especialistas de las unidades terminológicas, por medio de una teoría pragmática y sociológica;
- de la adquisición del lenguaje en general y, en particular, de la terminología, mediante una teoría psicológica.

Cabré (1999: 103) también añade al conjunto de teorías una teoría de los signos, ya que el conocimiento especializado también puede representarse mediante sistemas simbólicos no lingüísticos. No obstante, Cabré (2002) solamente denomina *unidades terminológicas* o *términos* a las unidades de carácter lingüístico que se dan en el seno de la lengua natural.

Para explicar mejor el hecho de estudiar las unidades terminológicas desde distintos puntos de vista, Cabré (2003: 195–196) utiliza la «teoría de las puertas», según la cual, la unidad terminológica es como una casa con varias puertas de entrada. Todas las puertas dan acceso a todas las habitaciones de la casa, pero la elección de una puerta determinada condiciona cómo se llega al interior de la casa. El orden interno de las habitaciones no se altera, lo que cambia es cómo uno decide llegar. Por ello, Cabré se refiere a la unidad terminológica como un poliedro que se puede observar desde tres perspectivas: la cognitiva (el concepto), la lingüística (el término) y la comunicativa (la situación) (Cabré 2003: 187).

La TCT divide la terminología en dos vertientes interrelacionadas: la teórica y la práctica. La terminología teórica se dedica a la descripción formal, semántica y funcional de las unidades que pueden adquirir valor terminológico, explicar su activación y las relaciones con otros tipos de signos para hacer progresar el conocimiento sobre la comunicación especializada y las unidades que se usan en ella. Por otro lado, la terminología aplicada se encarga de la recopilación y el análisis de las unidades con valor terminológico. No obstante, aunque la TCT defiende el carácter descriptivo de la terminología, también reconoce la existencia de situaciones en las que el trabajo prescriptivo es necesario, como en el caso de las lenguas minoritarias o minorizadas.

Una de las principales aportaciones de la TCT es el estudio de la variación terminológica. Para la TCT, los términos son unidades de forma y contenido en las que el contenido es simultáneo a la forma (Cabré 1999: 123). Un contenido —es decir, un concepto— puede ser expresado por otras denominaciones del sistema lingüístico relacionadas semánticamente con la primera o mediante otros sistemas simbólicos. De este modo, la TCT destierra el principio de biunivocidad de la TGT y propone el estudio de la sinonimia y la polisemia. Cabré expone que todo proceso de comunicación comporta variación y, partiendo de ahí, formula el «principio de variación»:

> Este principio es universal para las unidades terminológicas, si bien admite diferentes grados según las condiciones de cada tipo de situación comunicativa. El grado máximo de variación de la terminología lo cumplirían los términos de las áreas más banalizadas del saber y los que se utilizarían en el discurso de registro comunicativo de divulgación de la ciencia y de la técnica; el grado mínimo de la variación sería propio de la terminología normalizada por comisiones de expertos; el grado intermedio, la terminología usada en la comunicación natural entre especialistas (Cabré 1999: 85).

Así pues, la variación terminológica no es solo un fenómeno evidente cuando se estudia el uso de los términos en el discurso. Esta se encuentra en la propia esencia de la comunicación especializada como parte integrante de la comunicación en su conjunto y está íntimamente relacionada con el nivel de especialidad y el contexto comunicativo en que se activen los conceptos.

La TCT supone la primera teoría viable capaz de sustituir a la TGT y ha dado lugar a numerosas investigaciones en el campo de la terminología (Faber 2009: 114). Muchos de sus principios siguen siendo asumidos como válidos por la mayoría de estudiosos de la terminología. La TCT, sobre todo, sentó las bases para que a la terminología se integrasen los conocimientos y avances de otras disciplinas además de la lingüística, que es la dimensión desde la que la TCT reconoce acceder al estudio de los términos.

#### 2.2.1.3 La terminología sociocognitiva

La terminología sociocognitiva (TSC) de Temmerman (2000, 2001) es junto con la terminología basada en marcos, la principal teoría terminológica basada en la lingüística cognitiva (§2.1). Una de sus nociones fundamentales es la de *unidad de comprensión,* que sustituye al término *concepto*, pues si bien se utiliza en lingüística cognitiva, en el campo de la terminología se ha venido asociado a los conceptos con límites precisos propugnados por la TGT (Temmerman 2000: 223–224). Temmerman estudió el dominio de la biología y llegó a la conclusión de que normalmente la unidad de comprensión es una categoría con una estructura prototípica como propuso Rosch (1978) para los conceptos no especializados (§2.1.3.2) y, rara vez, un concepto de límites precisos.

La TSC integra la noción de dinamismo al considerar que una categoría es como una «porción de conocimiento» que tiene un núcleo y una estructura, pero que existe en un proceso de continua reformulación y, por ende, se encuentra en continua transición. El dinamismo según Temmerman (2000: 227) se debe a varios factores activos que influyen simultáneamente en el sistema conceptual:

- la necesidad de cada vez más y mejor comprensión;
- la interacción entre usuarios de distintas lenguas;
- la estructura prototípica que puede verse tanto como el resultado como una de las causas de la evolución del significado.

Asimismo, la TSC defiende un enfoque semasiológico frente al onomasiológico de la TGT y propone que para estudiar y describir las categorías es necesario analizar la información obtenida del uso real de la lengua a partir de la combinación de cuatro perspectivas (Temmerman 2001: 79):

- la perspectiva nominalista, por la cual la unidad de comprensión es el sentido de la palabra;
- la perspectiva mentalista, por la cual la unidad de comprensión es una idea que existe en la mente de las personas;
- la perspectiva realista, por la cual la unidad de comprensión es una forma externa que existe en el universo;
- la perspectiva espaciotemporal, por la cual la unidad de comprensión evoluciona en el espacio y el tiempo.

Temmerman (2000: 227) demuestra, a partir del análisis de textos reales, que la sinonimia y la polisemia son funcionales en el proceso de la comprensión. A menudo, la (cuasi-)sinonimia puede explicarse por la existencia de diferentes perspectivas desde las que hacer referencia a una categoría. Por su parte, la polisemia puede darse por:

- un cambio tecnológico o social;
- un cambio en cómo se comprende una categoría;
- el hecho de que la estructura prototípica de las categorías propicia la evolución del significado y que los elementos de la lengua se influyan y restrinjan mutuamente.

La TSC también incluye la noción de marco, pero lo hace inspirada por los modelos cognitivos idealizados de Lakoff (1987). Partiendo de esta base, Temmerman (2000: 225) explica que las unidades de comprensión tienen estructuras tanto intracategoriales como intercategoriales

En cuanto a la estructura intracategorial, la TSC distingue diferentes módulos de información (rasgos, núcleo de la definición, información histórica o información procedimental) que variarán de relevancia según el tipo de categoría. En lo relativo a la estructura intercategorial se tiene en cuenta la perspectiva, el dominio y la intención dentro del modelo cognitivo idealizado.

Respecto a la estructura prototípica, Temmerman sostiene que para comprender un término es necesario profundizar en su estructura prototípica porque los límites de las categorías son difusos y habitualmente presentan una estructura de semejanza de familia y grados de pertenencia a la categoría (§2.1.3.2).

A partir de estos principios teóricos, Temmerman (2000) propone nuevos métodos de análisis terminológico tomados de la semántica cognitiva: el análisis de la estructura prototípica, el análisis del modelo cognitivo idealizado y el análisis diacrónico. De hecho, una de los principales puntos fuertes de la TSC reside especialmente en los nuevos enfoques que propone para llevar a cabo análisis conceptuales, tomados de la lingüística cognitiva, lo cuales revisaremos en el apartado dedicado a su metodología definicional (§3.4.1).

## 2.2.2 Bases teóricas y prácticas

La terminología basada en marcos (TBM) (Faber et al. 2006; Faber, León Araúz y Prieto Velasco 2009; León Araúz 2009; Faber 2012; Faber 2014) es el enfoque teórico principal en el que se fundamenta este trabajo. Surge en la primera década del siglo actual como una nueva aproximación cognitiva a la terminología. Por un lado, la TBM comparte premisas con la TCT (§2.2.1.2) y la TSC (§2.2.1.3), como que la distinción entre términos y palabras no es fructífera ni viable y que la mejor manera de estudiar las unidades de conocimiento especializado es mediante el análisis de su comportamiento en los textos (Faber, León Araúz y Prieto Velasco 2009: 4). Por otro lado, la TBM toma principios de diversas teorías lingüísticas y psicológicas como el modelo de la gramática léxica (Martín Mingorance 1990; Faber y Mairal Usón 1999), la semántica de marcos (§2.1.2.2), el lexicón generativo (Pustejovsky 1995) y las teorías de la cognición fundamentada (§2.1.1).

De acuerdo con Faber (2011: 10), los términos son unidades de conocimiento especializado que designan la conceptualización que hacemos de objetos, propiedades, estados y procesos en un dominio especializado y, por ese motivo, cualquier teoría de terminología debería aspirar a la adecuación psicológica y neurológica. Asimismo, en línea con (Meyer, Bowker y Eck 1992) y como corolario de esta visión de la noción de término, todo recurso terminológico y, en general, todo trabajo terminológico aplicado debe estar basado en el funcionamiento y organización de la cognición humana:

> [K]nowledge of conceptualization processes as well as the organization of semantic information in the brain should underlie any theoretical assumptions concerning the access, retrieval, and acquisition of specialized knowledge as well as the design of specialized knowledge resources (Faber 2011: 10).

De este modo, la TBM adopta el compromiso cognitivo, que como ya se indicó (§2.1) supone que las investigaciones llevadas a cabo bajo este paradigma deben concordar con lo que las ciencias cognitivas conocen hasta el momento acerca del sistema conceptual humano. Como consecuencia de ello, la noción de marco aparece como central, ya que, como demostró Barsalou (1992), esa es la organización más probable del sistema conceptual junto con las jerarquías taxonómicas. Además, los marcos como herramienta terminográfica permiten representar los procesos dinámicos que tienen lugar en campos de conocimiento especializados de un modo similar a la representación conceptual humana (Faber et al. 2007: 40).

Aunque inicialmente la noción de marco de la TBM provenía de la semántica de marcos, el desarrollo posterior de la TBM la ha alejado de dicha teoría para dar paso a una noción de marco mucho más amplia que la de la semántica de marcos, como la que ofrecen Evans y Green (2006: 211):

> [A] frame […] represents a schematization of experience (a knowledge structure), which is represented at the conceptual level and held in long-term memory and which relates element and entities associated with a particular culturally embedded scene, situation or event from human experience.

Desde un punto de vista terminográfico, la TBM define el marco como una representación que integra varias formas de combinar generalizaciones semánticas sobre una categoría o un grupo de categorías (Faber 2014: 15), convirtiendo así el marco en una herramienta de representación del significado mucho más versátil que la que propone la semántica de marcos. Como veremos más adelante, en la TBM, los marcos aparecen en todos los niveles de representación de un recurso terminológico, desde la macroestructura hasta la microestructura (León Araúz 2009). No obstante, es importante remarcar que los marcos de la TBM, aunque están basados en el lenguaje, pues es la principal puerta de acceso al sistema conceptual, no pretenden representar la estructura lingüística, sino la conceptual (León Araúz, Faber y Montero Martínez 2012: 117).

Desde sus inicios, la TBM ha tenido como uno de sus ejes centrales el estudio y representación de la naturaleza multidimensional del conocimiento especializado (Faber et al. 2006: 190) y, en ese sentido el marco permite dar cuenta del fenómeno de la multidimensionalidad al representar tanto la información de fondo necesaria para la comprensión de un concepto como la información que se active como consecuencia del contexto y que guía la conceptualización situada (Barsalou 1999: 591-592).

#### 2.2.2.1 Los eventos y los subeventos

El término *evento* se emplea de tres maneras distintas dentro de la TBM:

- proceso que tiene lugar en el mundo real y que es objeto de estudio por parte de un dominio de conocimiento especializado (en este trabajo, utilizaremos el término *proceso* para hacer referencia a este tipo de evento);
- marco que funciona de plantilla para la descripción de los procesos que ocurren dentro de un dominio de especialidad (p. ej., el evento medioambiental o el evento médico);
- marco que describe un proceso dentro de un dominio de especialidad (también recibe el nombre de *subevento*, p. ej., el subevento de SEDIMENTACIÓN o el subevento de EROSIÓN).

El marco más característico de la TBM es el evento, en la segunda acepción arriba presentada. Se fundamenta en la premisa de que la descripción de los dominios de especialidad se basa en los procesos que ocurren en ellos (Grinev y Klepalchenko 1999). La TBM defiende que la manera en que se representan los conceptos afecta a la configuración de la información en las entradas terminológicas individuales y los contenidos de cada campo de datos, especialmente en lo relativo a la definición de cada concepto (Faber et al. 2006: 191). Por ello, con el fin de representar el conocimiento especializado de manera dinámica, la TBM organiza las categorías más genéricas de un dominio en un evento prototípico:

> [W]e propose a frame-based organization of specialized fields in which a dynamic, process-oriented event frame provides the conceptual underpinnings for the location of sub-hierarchies of concepts within a specialized domain event (Faber et al. 2006: 189).

Por ende, la TBM defiende que todo área de conocimiento tendrá un evento que proporcionará una estructura para la organización de los conceptos dentro de dicho dominio (Faber y López Rodríguez 2012: 23) y que, aplicado a un recurso terminológico, es uno de los factores que permite que las entradas sean coherentes tanto interna como externamente (Faber et al. 2007).

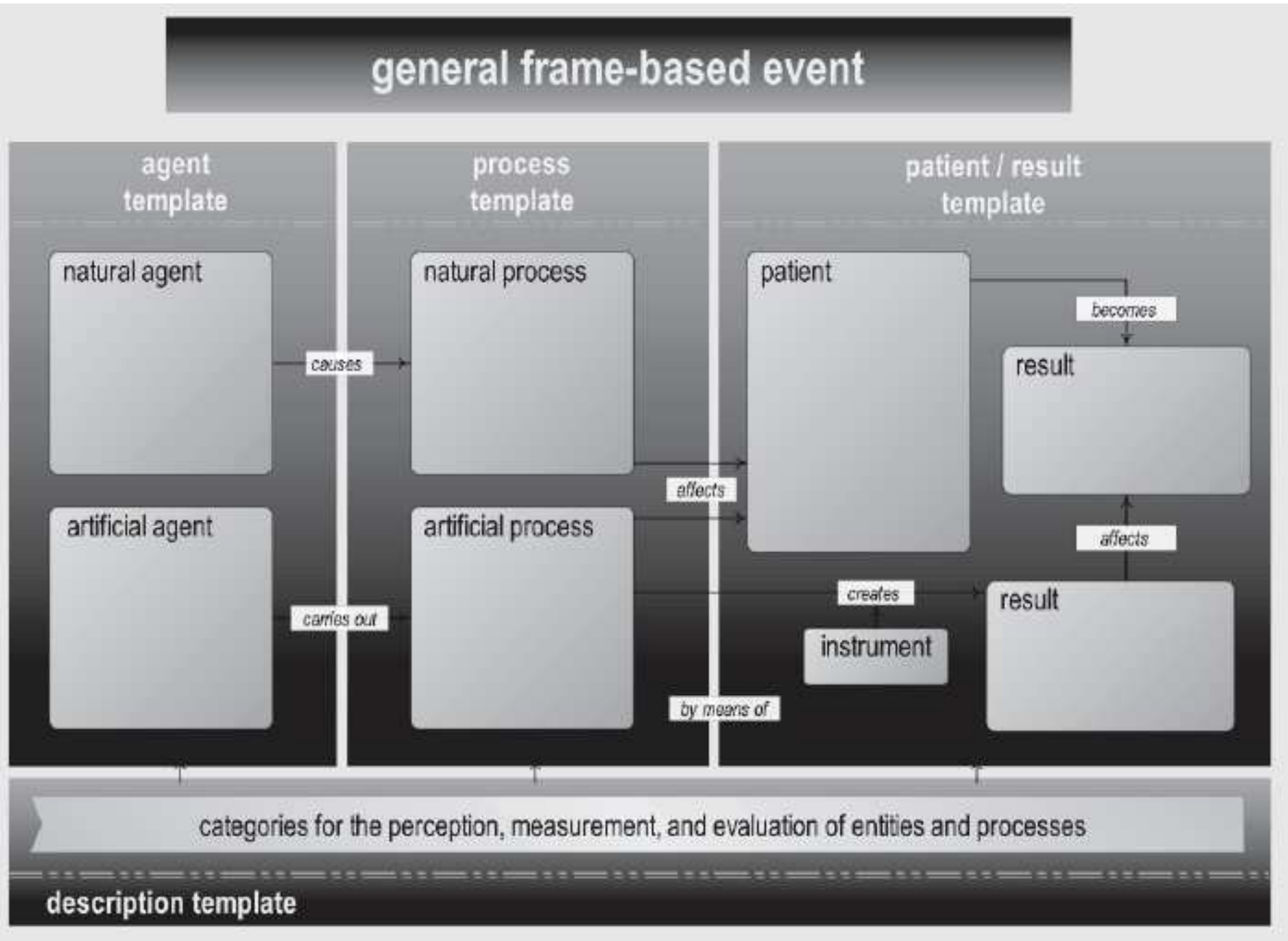

**Figura 1. Evento general de la TBM (León Araúz, Faber y Montero Martínez 2012: 119)**

Como modelo para la creación del evento de un dominio de especialidad la TBM ofrece el evento general (León Araúz, Faber y Montero Martínez 2012: 119) (Figura 1). El evento está dividido cuatro macrocategorías: AGENTE, PROCESO, PACIENTE/RESULTADO y DESCRIPCIÓN. Dentro de algunas de dichas plantillas se encuentran roles semánticos (AGENTE / PACIENTE / INSTRUMENTO / RESULTADO). A partir de dichas macrocategorías se pueden generar plantillas para los estados y eventos prototípicos que caracterizan un dominio de especialidad así como las entidades que participan en ellos (León Araúz, Faber y Montero Martínez 2012: 97). Así pues, en la TBM, las categorías están diseñadas de acuerdo con el papel que desempeñan dentro del evento y los subeventos y no únicamente a partir de su estructura interna como ENTIDAD, ACCIÓN o PROPIEDADES (León Araúz 2009: 147).

En la creación del evento de un dominio concreto se pueden añadir otros roles semánticos que sean necesarios, aunque de una manera limitada. La TBM se encuentra en un punto intermedio entre la creación de roles semánticos específicos para cada marco como en FrameNet y la generalización total en macrorroles (León Araúz, Faber y Montero Martínez 2012: 123).

Por otro lado, como se puede observar, las relaciones conceptuales primarias que unen a los distintos roles semánticos en el evento son relaciones no jerárquicas como *causa* o *afecta*, lo cual resuelve las limitaciones de las taxonomías jerárquicas y estáticas (Faber, León Araúz y Prieto Velasco 2009: 8).

Así pues, dado que cada dominio especializado tiene su propio evento prototípico; en la elaboración de un recurso terminológico que siga los principios de la TBM habrá de reconstruirse dicho evento a partir del cual se representarán los conceptos de todo el recurso y sus relaciones. El mejor ejemplo es el del evento medioambiental creado para la base de conocimiento terminológica (BCT) EcoLexicon (§2.2.2.5) (Figura 2).

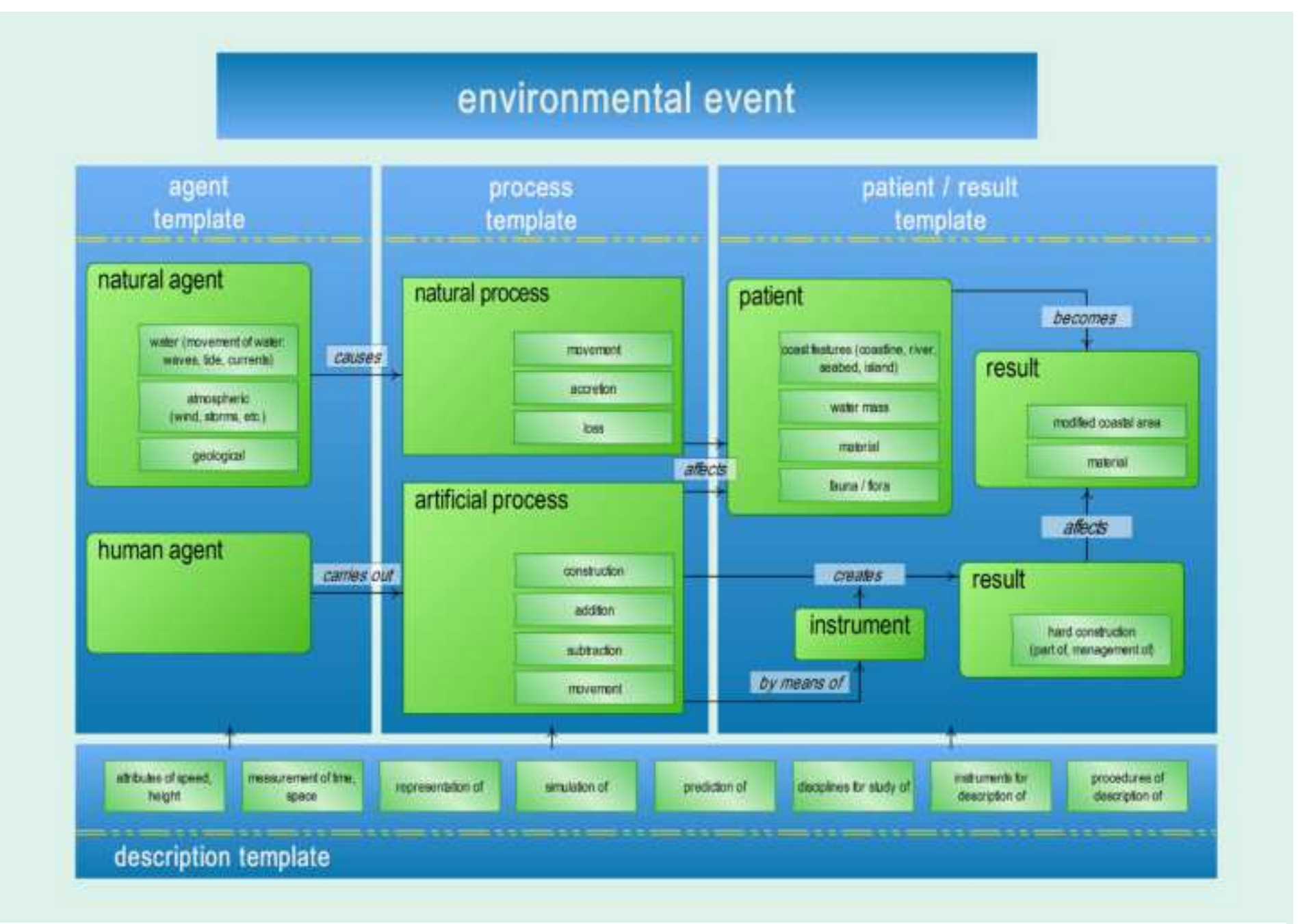

**Figura 2. Evento medioambiental (Faber, León Araúz y Prieto Velasco 2009: 8)**

<table>
<tr>
<td><b>Agent (A)</b><br>Natural agent (A.1)<br>Water Agent (A.1.1)<br>Atmospheric Agent (A.1.2)<br>Part of Atmospheric Agent (A.1.2.1)<br>Geological Agent (A.1.3)<br>Chemical Agent (A.1.4)<br>Physical Agent (A.1.5)<br>Biological Agent (A.1.6)<br>Artificial Agent (A.2)<br>Human (A.2.1)<br>Instrument (A.2.2)<br>System (A.2.3)<br>Entity / Institution (A.2.4)<br>Structure / Construction (A.2.5)<br>Substance (A.2.6)</td>
<td rowspan="3"><b>Patient / result (C)</b><br>Patient (C.1)<br>Natural Patient (C.1.1)<br>Coast feature (C.1.1.1)<br>Part of Coast Feature (C.1.1.1.1)<br>Water Mass (C.1.1.2)<br>Part of Water Mass (C.1.1.2.1)<br>Material (C.1.1.3)<br>Fauna/flora (C.1.1.4)<br>Geographic location (C.1.1.5)<br>Geographic feature (C.1.1.6)<br>Part of geographic feature (C.1.1.6.1)<br>Artificial Patient (C.1.2)<br>Construction (C.1.2.1)<br>Material (C.1.2.2)<br>Substance (C.1.2.3)<br>Result of natural process (C.2)<br>Effect (C.2.1)<br>Modified coastal area (C.2.2)<br>Material (C.2.3)<br>Geographic feature (C.2.4)<br>Modified geographic feature (C.2.5)<br>Substance (C.2.6)<br>Result of artificial process (C.3)<br>Construction (C.3.1)<br>Part of construction (C.3.1.1)<br>Material (C.3.3)<br>Modified coastal area (C.3.4)<br>Effect (C.3.5)<br>Modified geographic feature (C.3.6)<br>Substance (C.3.7)</td>
</tr>
<tr>
<td><b>Process (B)</b><br>Natural processes (B.1)<br>Movement (B.1.1)<br>Loss (B.1.2)<br>Addition (B.1.3)<br>Transformation (B.1.4)<br>Artificial processes (B.2)<br>Construction (B.2.1)<br>Addition (B.2.2)<br>Subtraction (B.2.3)<br>Movement (B.2.4)<br>Transformation (B.2.5)<br>Instrument (B.3)</td>
</tr>
<tr>
<td><b>Description (D)</b><br>Attributes / measurement of (D.1)<br>Representation of (D.2)<br>Spatial representation (D.2.1)<br>Temporal representation (D.2.2)<br>Simulation / prediction of (D.3)<br>Disciplines for study of (D.4)<br>Instruments / procedures of desc. of (D.5)<br>Part of instrument (D.5.1)</td>
</tr>
</table>

**Tabla 1. Categorías conceptuales del evento medioambiental**

El evento medioambiental conceptualizado como un proceso dinámico iniciado por un agente (ya sea natural o humano); después, el proceso afecta a un paciente y produce un resultado. La relación entre agentes naturales y procesos naturales es paralela a la existente entre agentes humanos y procesos artificiales (Faber et al. 2006: 194–195). Asimismo, se desarrolló un inventario de las principales categorías conceptuales que pueden aparecer dentro las macrocategorías del evento medioambiental (Tabla 1).

Subsumidos dentro del evento, se encuentran los subeventos que siguen el patrón del evento y describen procesos concretos que tienen lugar en el dominio. Algunos de los subeventos descritos en EcoLexicon incluyen, entre otros, CONSTRUCCIÓN, INUNDACIÓN, EROSIÓN, SEDIMENTACIÓN, DESERTIZACIÓN y DEFENSA COSTERA (León Araúz, Faber y Montero Martínez 2012: 119). A su vez, los subeventos pueden contener más subeventos, como ocurre, por ejemplo, con el subevento de la EROSIÓN DEL SUELO:

> [S]oil erosion in one area is a complex event that includes many sub-events, such as weather change, rainfall, over-exploitation of agriculture land, deforestation, etc. These sub-events are related since they influence each other over time. Participants in this event are soil, water, and wind. The final result is that soil is eroded. Each subevent is also observable as sets of similar more specific changes. (León Araúz, Faber, y Montero Martínez 2012: 119).

Los roles semánticos y las relaciones conceptuales que unen las categorías varían según el subevento y, además, dada la naturaleza multidimensional de los procesos de categorización humana, los conceptos se categorizan de manera distinta según el contexto, el mismo concepto puede asumir roles diferentes en subeventos relacionados (León Araúz, Faber y Montero Martínez 2012: 120). Por ejemplo, una categoría conceptual puede ser AGENTE en un subevento y PACIENTE en otro.

#### 2.2.2.2 Las relaciones conceptuales

La representación conceptual basada en marcos que defiende la TBM no solo supone la inclusión de los conceptos dentro de los marcos en los que participan, sino que también se representan las relaciones existentes entre los conceptos. De acuerdo con Faber, León Araúz y Prieto Velasco (2009: 7), dichas relaciones no pueden limitarse a las tradicionales, a saber, las genérico-específicas y las meronímicas, sin negar por ello el papel fundamental que estas desempeñan (Faber y León Araúz 2010: 84). La inclusión en una BCT de relaciones no jerárquicas, incluidas relaciones específicas de un dominio, aporta mayor coherencia y dinamicidad a la representación conceptual (Faber, León Araúz y Prieto Velasco 2009: 7).

Las relaciones conceptuales están presentes en todos los niveles de especificidad de una BCT basada en la TBM: desde el evento, en el que las relaciones especifican cómo interactúan las macrocategorías entre sí, hasta la descripción individual de cada concepto que, como veremos, consiste en su relación con otros conceptos y determinado por los subeventos en los que el concepto aparece inserto.

Las relaciones entre conceptos en la TBM se expresan mediante proposiciones formadas por dos conceptos y la relación que los une: «CONCEPTO *relación* CONCEPTO». Este formato no difiere mucho del propuesto por Barsalou (1992), pues básicamente también consistía en dos conceptos, que según el caso podía ser un atributo y un valor, o dos atributos, o dos valores, unidos por una relación. Así pues, al igual que los marcos de Barsalou, los de la TBM también tienen la característica de la recursividad, ya que en ellos se encuentran conceptos relacionados entre sí, pero que a su vez son marcos en sí mismos (que como veremos se configuran en la TBM como plantillas categoriales) en los que el concepto en cuestión se relaciona con otros conceptos, que a su vez se relacionan con otros, y así sucesivamente.

Las relaciones conceptuales en la TBM se estudiaron en el contexto de la BCT EcoLexicon (§2.2.2.5) mediante análisis de corpus y los resultados demostraron que la capacidad de una relación dada de unir dos conceptos depende de la naturaleza de los conceptos, es decir, si son una ENTIDAD FÍSICA, una ENTIDAD MENTAL, un PROCESO, un ESTADO o una PROPIEDAD (León Araúz y Faber 2010). Teniendo esto en cuenta, se estableció no solo una lista de las relaciones conceptuales que se usarían en EcoLexicon, sino que también se definieron sus criterios de uso (León Araúz 2009: 301-308; León Araúz y Faber 2010: 13; Faber y León Araúz 2010: 84-88; León Araúz, Faber y Montero Martínez 2012: 130-140):

- *Tipo-de*: Se trata de la tradicional relación genérico-específica. Representa el resultado de la categorización y, dentro de la BCT, estructura las jerarquías taxonómicas. La proposición «OXÍGENO *tipo-de* ELEMENTO QUÍMICO» supone que ELEMENTO QUÍMICO es un concepto superordinado de OXÍGENO, o en otras palabras, que OXÍGENO pertenece a la categoría de ELEMENTO QUÍMICO. Es posible que un concepto tenga varios superordinados, dándose así lugar a la multidimensionalidad y, por tanto, a la coexistencia de distintas jerarquías de conceptos.
- *Parte-de*: Esta es la relación meronímica básica, ya que en EcoLexicon se describen otras cinco relaciones meronímicas. *Parte-de* relaciona una entidad física con sus partes, por ejemplo, «ESTRIBO *parte-de* PRESA», o una disciplina científica con sus subdisciplinas como «GLACIOLOGÍA *parte-de* HIDROLOGÍA».
- *Fase-de*: Es una relación meronímica que se aplica para unir un proceso con sus fases, por ejemplo, «DIAGÉNESIS fase-de LITOGÉNESIS».
- *Compuesto-de*: Se trata de una relación meronímica que une tanto objetos artificiales como naturales con el material del que están hechos, por ejemplo, «PEDRAPLÉN *compuesto-de* PIEDRA».
- *Delimitado-por*: Relación meronímica específica del dominio del medio ambiente. Une objetos físicos que se delimitan mutuamente como «MESOSFERA *delimitado-por* ESTRATOSFERA».
- *Ubicado-en*: Esta relación meronímica se utiliza cuando la ubicación de un objeto físico es esencial en su descripción, es decir, que sin esa característica el concepto pierde su identidad. Un ejemplo de esta relación es «VIADUCTO *ubicado-en* DEPRESIÓN GEOGRÁFICA».

- *Tiene-lugar-en*: Esta relación meronímica une un proceso con su dimensión temporal («PLEAMAR SOSTICIAL MÁXIMA *tiene-lugar-en* SOLSTICIO») o espacial («TIFÓN *tiene-lugar-en* OCÉANO PACÍFICO»).
- *Resultado-de*: Relación no jerárquica que une una entidad o un proceso que es resultado de un proceso, por ejemplo, «NIEVE resultado-de PROCESO DE BERGERON».
- *Causa*: Relación no jerárquica que une una entidad con un proceso que esta provoca como «VIENTO *causa* EROSIÓN EÓLICA».
- *Afecta_a*: Esta relación no jerárquica une entidades o procesos que causan algún tipo de cambio en otra entidad o proceso sin producir un resultado final, por ejemplo, «EROSIÓN ANTRÓPICA *afecta-a* PLAYA». Aunque se han propuesto subtipos de esta relación (como *retarda* o *erosiona*), no se han aplicado por el momento en EcoLexicon.
- *Tiene-función*: Se trata de una relación no jerárquica que une una entidad o un proceso con su función («REVEGETACIÓN *tiene-función* PROTECCIÓN DEL SUELO»). No solo está limitado a las entidades y procesos artificiales, ya que el ser humano puede utilizar entidades y procesos naturales para sus propios fines («GUANO *tiene-función* FERTILIZACIÓN»). Esta relación, al igual que la de *afecta-a*, puede generar subtipos específicos del dominio del medio ambiente. En este caso, EcoLexicon sí ha incluido cuatro de ellas en su inventario: *se-hace-con* para unir exclusivamente un instrumento con el proceso que lleva a cabo («CONDENSACIÓN *se-hace-con* CÁMARA DE NEBLINA») o la entidad que crea («ECOGRAMA *se-hace-con* ECOSONDA»); *estudia* para unir una disciplina con su objeto de estudio («ZOOLOGÍA *estudia* FAUNA»); *mide* para instrumentos con una función de medición («SALINÓMETRO *mide* SALINIDAD»); y *representa* para gráficos y mapas («MAREOGRAMA *representa* NIVEL DEL MAR»).

- *Atributo-de*: Esta relación une una propiedad con una entidad o un proceso del que es característico, por ejemplo, «IMPERMEABLE atributo-de ACUÍFERO CONFINADO».

Todas las relaciones, excepto *delimitado-por* (por ser simétrica), tienen su inversa, por ejemplo: *tipo-de/tiene-tipo, parte-de/tiene-parte, se-hace-con/sirve-para*, etc. En la Tabla 2, se resumen las restricciones relacionales de los conceptos en EcoLexicon respecto a su naturaleza y su rol semántico.

#### 2.2.2.3 Las plantillas definicionales

Además de como eventos y subeventos, los marcos en la TBM se utilizan en la caracterización de categorías a modo de plantillas definicionales (o también llamadas *plantillas categoriales*). Por su parte, los eventos y los subeventos pueden considerarse representaciones intercategoriales, es decir, estructuras conceptuales a gran escala que conectan categorías mediante relaciones semánticas (Faber 2014: 15).

La distinción entre intracategorial e intercategorial es difusa, ya que finalmente tanto las relaciones intercategoriales como las intracategoriales son relaciones entre conceptos y ambas tienen como consecuencia la recursividad de los marcos unos dentro de otros. Sin embargo, a nivel metodológico es útil establecer esta diferencia, que podría caracterizarse como una diferencia de foco. Mientras que en las repreresentaciones intercategoriales (eventos y subeventos), el foco está en describir una estructura más amplia de conocimiento en el que las categorías se relacionan entre sí y aportan a la estructura una porción especializada de su contenido (la relevante a ese marco y adaptada a este), las representaciones intracategoriales (plantillas definicionales) ponen el foco en una categoría concreta, en la relación de esa categoría con otras categorías con las que interactúa dentro de uno o varios eventos o subeventos, según los objetivos de la representación.

| | **AGENTE** | | **PACIENTE** | |
|---|---|---|---|---|
| | **Poder relacional** | | | |
| **Tipo de concepto** | **Relación** | **Tipo de concepto** | **Relación** | **Tipo de concepto** |
| entidad física | *tipo_de*<br>*tiene_tipo*<br>*delimitado_por*<br>*tiene_parte*<br>*parte_de*<br>*compuesto_de*<br>*material_de*<br>*ubicado_en* | entidad física | *tipo_de*<br>*tiene_tipo*<br>*delimitado_por*<br>*tiene_parte*<br>*parte_de*<br>*compuesto_de*<br>*material_de*<br>*ubicado_en*<br>*ubicación_de* | entidad física |
| | *tiene_función* | proceso | | |
| | *sirve_para* | proceso/entidad | *se_hace_con* | |
| | *mide* | proceso/entidad | *se_mide_con* | |
| | *afecta_a* | proceso/entidad | *afectado_por* | proceso/entidad |
| | *causa* | proceso | | |

**Tabla 2. Restricciones conceptuales con arreglo a la naturaleza conceptual y al rol semántico (León Araúz y Faber 2010: 15)**

Los marcos intracategoriales reciben el nombre de *plantillas definicionales* en la TBM porque son la base la redacción de las definiciones terminológicas en cualquier BCT basada en la TBM. El formato y la metodología empleados en la configuración de las plantillas definicionales en la TBM está inspirado en los marcos de Martin (1998) y en la estructura de *qualia* del lexicón generativo. Trataremos con detenimiento la cuestión de las plantillas definicionales en el apartado dedicado a la metodología definicional de la TBM (§3.4.2).

#### 2.2.2.4 La extracción del conocimiento

En lo que respecta a la extracción del conocimiento, la metodología de la TBM se basa en la derivación del sistema conceptual del dominio mediante una aproximación tanto *top-down* como *bottom-up*. El método *top-down* consiste en analizar, filtrar, combinar y reestructurar información contenida en diccionarios especializados y otro material de referencia, la cual después es validada por expertos. El método *bottom-up* consiste en extraer la información de un corpus de textos en varias lenguas y que estén relacionados con el dominio (Faber, León Araúz y Prieto Velasco 2009: 6).

La metodología de la TBM en lo que respecta a la extracción del conocimiento para su posterior representación en un recurso terminológico, se detallará tanto en el apartado dedicado a la metodología definicional de la TBM (§3.4.2) como en el apartado de métodos de este trabajo (§4.2).

#### 2.2.2.5 EcoLexicon

EcoLexicon[25] es una BCT sobre el medio ambiente desarrollada por el grupo de investigación LexiCon de la Universidad de Granada desde el año 2003. Esta tesis doctoral se enmarca dentro de EcoLexicon y sus resultados se han generado de acuerdo con la viabilidad dentro de EcoLexicon y con vistas a su eventual aplicación en este.

EcoLexicon actualmente cuenta con 3.599 conceptos y 19.995 términos en inglés, español, alemán, griego moderno, francés, ruso y neerlandés. La BCT está principalmente almacenada en forma de base de datos relacional, aunque está en proceso de convertirse en una ontología (León Araúz y Magaña 2010; León Araúz, Faber y Magaña 2011). Los fundamentos teóricos y metodológicos de EcoLexicon son la TBM.

EcoLexicon está dirigido a una variedad de usuarios distintos como traductores, redactores técnicos, estudiantes y cualquier otro usuario que desee adquirir conocimientos especializados sobre medio ambiente y su terminología (Reimerink y Faber 2009: 629). En §3.5.2, profundizaremos en los usuarios de EcoLexicon y sus necesidades desde el punto de vista de la teoría funcional de la lexicografía.

En la Figura 3, se puede observar la interfaz de EcoLexicon. Las consultas a la BCT se pueden realizar por concepto (solo en español o en inglés) o por término (en cualquiera de las lenguas incluidas en EcoLexicon) a partir del

[25] Disponible en: <http://ecolexicon.ugr.es>.

campo de búsqueda en la barra superior. Si el término o concepto que se ha buscado se encuentra en la BCT, la información se muestra en la columna de la izquierda (módulo de definición, módulo de términos, módulo de recursos, módulo de categorías conceptuales y módulo de fraseología) y en el área central (módulo de mapa conceptual):

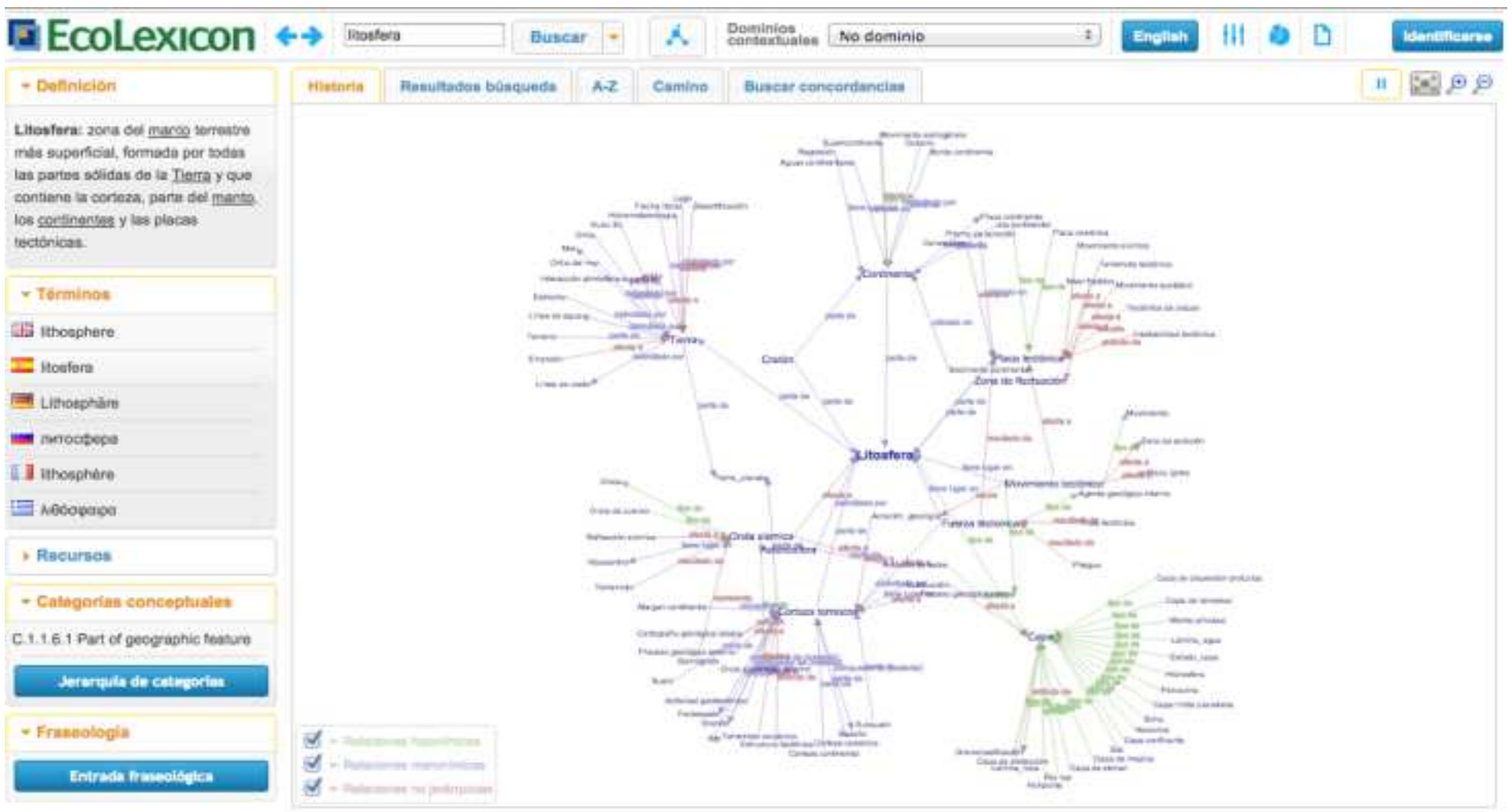

**Figura 3. Interfaz de EcoLexicon**

- Módulo de definición: Consiste en una definición terminológica del concepto (en inglés o en español, según la búsqueda). Aquellos conceptos que se mencionan en la definición y que están contenidos en la BCT aparecen subrayados y permiten al usuario acceder la información de dicho concepto. Trataremos en mayor profundidad la definición terminológica en EcoLexicon en §3.4.2.
- Módulo de términos: Consiste en una lista de términos que designan el concepto en las distintas lenguas de EcoLexicon. Si se pulsa sobre alguno de los términos, aparece una nueva ventana que proporciona información morfosintáctica acerca de la categoría gramatical del término (nombre, adjetivo, verbo o adverbio), el tipo de término (término principal, sinónimo, variante geográfica, acrónimo, etc.) y el género (masculino, femenino o neutro), así como contextos lingüísticos de uso (Reimerink, García De Quesada y Montero Martínez 2012) y fraseología (Buendía 2013).

- Módulo de recursos: Una lista de recursos multimedia asociados al concepto (imágenes, documentos, vídeos, etc.). Al hacer clic sobre alguno de los enlaces de esta sección, aparece una nueva ventana con información sobre el recurso en cuestión y el correspondiente hipervínculo hacia la página web donde se encuentra alojado. La selección de las imágenes en EcoLexicon está sujeta a una metodología que optimiza la adquisición de conocimiento especializado por parte del usuario (Prieto Velasco y Faber 2012).
- Módulo de categorías conceptuales: Cada concepto en EcoLexicon está asociado a una o varias categorías conceptuales que lo vinculan con el evento medioambiental. En este módulo, el usuario puede consultar a qué categoría pertenece el concepto dentro del evento medioambiental, así como consultar la jerarquía completa y los conceptos subsumidos en cada categoría. La lista completa está reproducida en la Tabla 1.
- Módulo de fraseología: Este módulo consiste en la clasificación de los verbos con arreglo a su significado nuclear —basado en los dominios léxicos de Faber y Mairal (1999)— y subdividido en dimensiones semánticas. Se proporciona información de uso sobre los verbos y sus restricciones semánticas. Como ya se ha señalado, a la información fraseológica también se puede acceder a través de los términos. La metodología de descripción fraseológica de EcoLexicon se encuentra descrita detalladamente en Buendía (2013).
- Módulo de mapa conceptual: Consiste en una representación dinámica de la red conceptual en la que se encuentra el concepto en cuestión vinculado a otros conceptos a través de relaciones conceptuales. Las relaciones genérico-específicas aparecen en color verde, las relaciones meronímicas en color azul y las relaciones no jerárquicas en color rojo. A partir de su última actualización, EcoLexicon permite las siguientes opciones relativas al mapa conceptual (Faber y Buendía 2014: 602):

- Cambiar la escala del mapa.
- Elegir qué tipos de relaciones se muestran y cuáles se ocultan.
- Ajustar el lugar de los nodos o eliminarlos.
- Obtener la URL de la representación de un concepto.
- Obtener información adicional sobre un concepto a partir de Google, Google Imágenes y Wolfram Alpha.

Asimismo, se han incluido dos nuevos modos de representación de los mapas: el árbol jerárquico (botón en la barra superior), que muestra la estructura jerárquica en la que está inserta un concepto; y el mapa del camino de relaciones conceptuales que une dos conceptos, disponible a partir de una pestaña en el área central.

En la barra superior, el usuario, además de poder acceder a los ajustes avanzados, las estadísticas sobre EcoLexicon, la opción de impresión y su cuenta personal, tiene la posibilidad de elegir un dominio contextual a partir de una lista desplegable. Esta función permite restringir la información que se muestra a solamente un campo de conocimiento. Hasta ahora, esta función solo afectaba a la representación en el mapa conceptual, pero a raíz del presente trabajo, se pretende que los dominios contextuales también determinen el contenido de la definición terminológica. En §3.6.1, retomaremos la cuestión de los dominios contextuales en EcoLexicon.

### 2.2.3 Otras aplicaciones de los marcos en terminología

Los marcos como herramienta de representación léxico-semántica y conceptual dentro del campo de la terminología han experimentado desde principios del siglo actual. En nuestra opinión, este hecho se debe por un lado al éxito de FrameNet en el ámbito de la Lexicografía, lo cual queda patente en que gran parte de las propuestas del uso de marco en terminología se inspiran en ella, y, por otro lado, en el giro cognitivo experimentado por la terminología como disciplina (Faber 2009a), que comporta el uso de nuevos métodos de representación del conocimiento especializado basados en los descubrimientos provenientes de las ciencias cognitivas, lo cual lleva al empleo de estructuras de gestión del conocimiento como los marcos.

Por ejemplo, los marcos se han aplicado, entre otros, a la terminología jurídica en portugués (Alves, Chishman y Quaresma 2007), a los sistemas de clasificación en Medicina (Wermuth 2009) y a la anotación de corpus médicos para el estudio de verbos especializados (Wandji, L'Homme y Grabar 2013; Wandji, L'Homme y Grabar 2014).

A pesar de que la mayoría de estudios terminológicos que hacen uso de los marcos lo hacen inspirándose en FrameNet, recientemente, han aparecido tres estudios que aplican explícitamente los principios de la TBM. En primer lugar, Hitcheva (2014) emplea la metodología de la TBM para la representación de colocaciones verbales en el dominio del procesamiento del acero. Asimismo, Peruzzo (2014), crea un evento sobre víctimas de delitos en el dominio del derecho europeo para aplicarlo a la extracción terminológica. Finalmente, Diez-Arroyo (2015) aplica la TBM para estudiar los neologismos en el lenguaje especializado de la moda.

En los siguientes apartados, haremos un breve repaso por algunas de las propuestas más destacadas inspiradas en la semántica de marcos y que han dado lugar a recursos terminológicos, a saber: el Kicktionary, el Dicionário da Copa do Mundo, BioFrameNet, DiCoEnviro y JuriDiCo.

#### 2.2.3.1 Kicktionary

El Kicktionary[26] (Schmidt 2006; Schmidt 2009) consiste en un recurso terminológico multilingüe (en inglés, alemán y francés) basado en FrameNet que abarca el dominio del fútbol. A diferencia de FrameNet, el

[26] Disponible en: <http://www.kicktionary.de/>.

Kicktionary se creó para que constituyera un recurso dirigido a usuarios humanos con la intención de servir de asistencia en la comprensión, traducción y creación de textos sobre fútbol, sin la intención de que fuera utilizado para el procesamiento del lenguaje natural.

Metodológicamente, la mayor diferencia con FrameNet estriba en que se siguió un enfoque *bottom-up,* esto es, se comenzó por la caracterización de las unidades léxicas para después agruparlas (Schmidt 2006: 76). La estructura del Kicktionary está basada en la diferencia entre escena y marco. Mientras que los marcos son estructuras lingüísticas y, por tanto, tienen unidades léxicas asociadas, las escenas son exclusivamente conceptuales y no tienen unidades léxicas asociadas:

> [A] *frame* is a structural entity used to group linguistic expressions which share a common *perspective* on a given conceptual *scene*. Where as a scene is defined in terms of pieces of abstract (and possibly non-linguistic) knowledge, the notion of a frame is concerned with the properties of concrete linguistic means of expressing this kind of knowledge. (Schmidt 2009: 102–103)

Dado el carácter multilingüe del Kicktionary, Schmidt consideró que la combinación de escenas y marcos permitía gestionar mejor las diferencias entre las lenguas incluidas en el recurso (2009: 125). Mientras que tuvieron que crearse marcos específicos de una o dos lenguas (Schmidt 2009: 109), todas las escenas contienen marcos que a su vez tienen unidades léxicas en las tres lenguas.

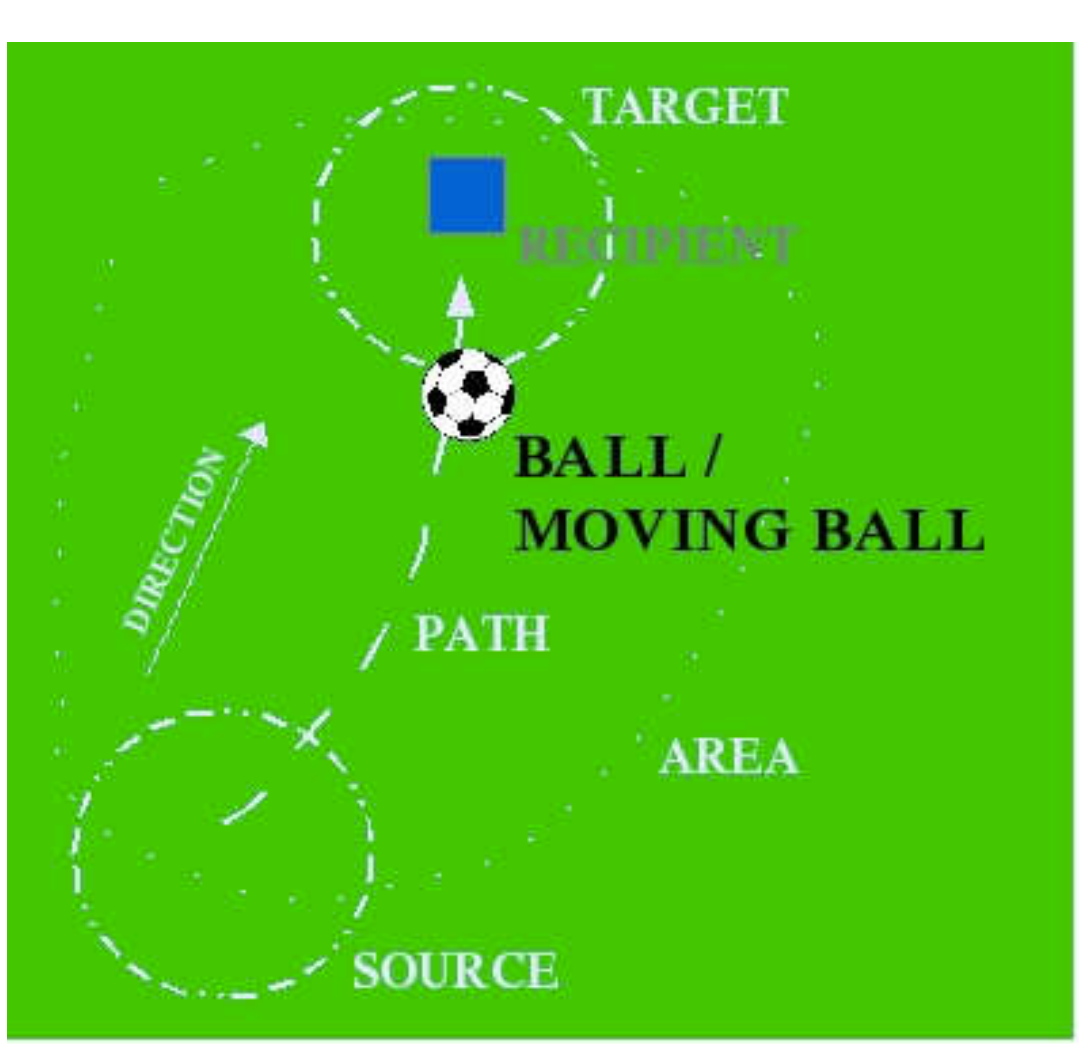

**Figura 4. Ilustración incluida en la descripción de la escena *Motion* (Fuente: Kicktionary)**

La única relación entre marcos en el Kicktionary es la existente entre las escenas y los marcos subsumidos en ellas, de las cuales proporcionan una perspectiva concreta. Un ejemplo de escena es *Motion*, que describe el movimiento del balón o de un jugador dentro del campo. Dicha escena incluye siete marcos organizados en tres grupos: *the ball moves*: *Ball_Move*, *Ball_Land* y *Ball_Escape*; *a bouncing ball*: *Ball_Bounce*; y *a player moves*: *Player_Move*, *Player_Move_With_Ball* y *Goal_Keeper_Advance*.

Las escenas se describen igual que las definiciones de los marcos en FrameNet, aunque a diferencia de este, también se incluyen imágenes que ilustran el evento que representa la escena. En la Figura 4 podemos ver la imagen que ilustra el grupo de marcos *the ball moves* dentro de la escena *Motion*.

Por su parte, los marcos se describen dentro de la definición de la escena, por lo que la entrada de un marco en Kicktionary solo incluye la lista de unidades léxicas (normalmente en las tres lenguas), los elementos del marco y un resumen de cuáles elementos del marco aparecen en la estructura argumental de cada unidad léxica. Por su parte, las entradas para las unidades léxicas incluyen ejemplos anotados y un resumen de posibles instanciaciones de cada elemento del marco en la estructura argumental de esa unidad léxica.

En el Kicktionary, a diferencia de FrameNet, las unidades léxicas están relacionadas entre sí mediante distintos tipos de relaciones léxicas (Schmidt 2009: 112–116). Entre unidades léxicas de la misma lengua puede haber relaciones de sinonimia, lo que da lugar a *synsets* al estilo de WordNet (Fellbaum 1998); la hiperonimia y la meronomia, que generan jerarquías de sustantivos y adjetivos; y la troponimia, que da lugar a jerarquías verbales. Asimismo, entre unidades léxicas de distintas lenguas se establece la relación de equivalencia de traducción.

Una deficiencia del Kitcktionary es la ausencia de definiciones para las unidades léxicas, a pesar de que parece que, en cierta etapa del desarrollo del recurso, sí se confeccionaron algunas definiciones, como se puede observar en Schmidt (2006). Así, por ejemplo, del sustantivo *upend* en inglés, aprenderemos que se encuentra en la escena *Foul* y el marco *Foul*, que tiene varios equivalentes en francés y alemán, que es un tropónimo de *foul*, pero tendríamos que recurrir a otro recurso para saber qué significa.

Asimismo, el Kicktionary admite tener el mismo problema que ya señalamos en FrameNet (§2.1.2.2) en lo que respecta a la descripción mediante marcos de unidades léxicas no predicativas:

> Nouns whose main function is to denote persons and objects (like goal-keeper, substitute, byline, penalty area) rather than to describe processes or activities (like most LUs exemplified in the previous sections) constitute a class of words that are especially difficult to characterize. In this case the concept of scenes and frames loses a lot of its intuitiveness. (Schmidt 2009: 110)

Para paliar en cierta medida esa deficiencia, en el Kicktionary se crearon las escenas estáticas *Actors* (que describe las personas que intervienen en un partido de fútbol) y *Field* (que describe los objetos que forman parte de o que aparecen en un campo de fútbol durante un partido), que a su vez contienen marcos específicos.

Otro problema que señala Schmidt (2009: 110–112) es la ausencia de fronteras claras entre las distintas escenas y marcos, lo cual se manifestó también en la asignación de unidades léxicas a los marcos. El hecho de que en el Kicktionary, a pesar del hecho de que las actividades que forman parte de un partido de fútbol están regidas por un conjunto de reglas convencionalizadas y que los actores y elementos que forman parte de dichas actividades están claramente delimitados (Schmidt 2009: 103), demuestra, por un lado, los límites difusos de cualquier tipo de estructura conceptual y la enorme influencia que, al final, tienen las decisiones metodológicas que se tomen en la representación de dichas estructuras en un recurso terminológico.

#### 2.2.3.2 Dicionário da Copa do Mundo

El Dicionário da Copa do Mundo[27] (Salomão, Torrent y Sampaio 2013; Torrent, Salomão, Campos et al. 2014; Torrent, Salomão, da Silva Matos et al. 2014) consiste en un diccionario en línea basado en la metodología de FrameNet concebido con la intención de ser de utilidad para no lingüistas (en particular, turistas, periodistas y trabajadores relacionados con la organización de la Copa del Mundo de Fútbol de 2014). Se trata de un diccionario multilingüe de inglés, portugués brasileño y español, y está centrado en los dominios del fútbol, el turismo y la propia Copa del Mundo (Torrent, Salomão, Campos et al. 2014: 10). En todo el recurso se utilizan los mismos marcos para las tres lenguas, porque según Torrent et al. (2014: 79) los dominios estudiados no varían suficientemente entre culturas como para crear marcos diferenciados. Sin embargo, ello contrasta, con que, como ya se señaló, en el Kicktionary, sí fue necesario crear marcos propios de una o dos lenguas.

[27] Disponible en: <http://www.dicionariodacopa.com.br/>.

La novedad del Dicionário da Copa do Mundo —desde el punto de vista de la terminología y en comparación con FrameNet— es que la metodología empleada incluye además de un enfoque *bottom-up*, también uno *top-down* (Torrent, Salomão, da Silva Matos et al. 2014: 81–82). En primer lugar, siguiendo un enfoque *bottom-up*, se extrajeron de unos corpus específicos de los dominios oraciones con unidades léxicas candidatas y se anotaron empleando etiquetas provisionales para los elementos del marco. A partir de la anotación, se definió una estructura de marcos preliminar. A continuación, el enfoque *top-down* consistió en alterar dicha estructura de marcos con arreglo a la organización del conocimiento que se propone en la literatura especializada de los dominios en cuestión.

Cabe destacar que si bien el Dicionário da Copa do Mundo no incluye contextos anotados ni información sintáctica, las entradas de cada una de sus unidades léxicas sí contienen una definición. Por lo demás, el resto de características destacables que presenta el Dicionário da Copa do Mundo están relacionadas con la adaptación de la estructura de la base de datos y del sistema de anotación de contextos lingüísticos para su uso en tres lenguas distintas, a pesar de que no están disponibles para el público.

### 2.2.3.3 BioFrameNet

BioFrameNet[28] (Dolbey, Ellsworth y Scheffczyk 2006; Dolbey 2009) es una extensión de FrameNet para el dominio de la biología molecular centrado particularmente en los procesos de transporte intracelular de proteínas. Se trata de un proyecto actualmente discontinuado que fue el objeto de la tesis doctoral de su creador (Dolbey 2009).

Dado que se trata de una extensión, la estructura de BioFrameNet y su metodología son similares a la de FrameNet. BioFrameNet está formado solamente por dos marcos: *Protein_transport* y *Cause_protein_transport.* El marco *Protein_transport,* que está enlazado con FrameNet mediante la relación *Inheritance* al marco *Motion* (Dolbey 2009: 75), cuenta con 4 elementos del marco y 32 unidades léxicas. Por su parte, *Cause_protein_transport* es el causativo de *Protein_transport* y, por tanto, les une la relación entre marcos: *Is Causative of*. *Cause_protein_transport* tiene 18 unidades léxicas y tiene un EM más que *Protein_transport* por ser su causativo.

El corpus utilizado para llevar a cabo el proyecto es una colección de textos anotados por el Laboratorio de Bioinformática de Lawrence Hunter de la Universidad de Colorado en Denver (Estados Unidos). Los textos anotados son descripciones de productos génicos obtenidos de la base de datos Gene[29], del Centro Nacional de Información Biotecnológica de Estados Unidos (NCBI). Dicha anotación estaba basada en una jerarquía de clases creada exclusivamente para clasificar fenómenos biológicos con el objetivo de introducir la información en una base de conocimiento. Esto es, no se tomaron en consideración cuestiones lingüísticas durante la anotación, a diferencia de BioFrameNet (Dolbey 2009: 8). El posterior uso de ese corpus anotado llevó al autor a destacar que dicha clasificación basada en conocimientos biológicos no tenía una correspondencia perfecta en la estructura representada en la lengua:

[28] BioFrameNet no está disponible en línea.

[29] Disponible en: <http://www.ncbi.nlm.nih.gov/gene/>.

> [T]he differences in biology which motivated division of the knowledge base in to the classes listed above are not directly reflected in the grammar and frame semantics of the language used to characterize these differences (Dolbey 2009: 9).

Así pues, mientras que el Laboratorio de Bioinformática de Lawrence Hunter había definido cinco tipos de mecanismos de transporte de proteínas de acuerdo con criterios biológicos (*protein transport*, *gated nuclear transport, transmembrane transport, vesicular transport* y *endocytosis*), en BioFrameNet, tras el análisis lingüístico y siguiendo las pautas para la creación de marcos en FrameNet, se decidió crear solamente dos marcos distintos: *Protein_transport* y *Cause_protein_transport* (Dolbey 2009: 72), siendo simplemente el segundo la variante causativa del primero.

Como queda patente en el proyecto BioFrameNet, la organización de marcos desde un punto de vista conceptual, como es el caso de la TBM, dará como resultados una división del conocimiento en marcos distintos de los que daría una configuración en marcos basada en un análisis sintáctico-semántico como el caso de FrameNet y sus derivados.

#### 2.2.3.4 DiCoEnviro

DiCoEnviro[30] (L'Homme 2012a; L'Homme 2012b; L'Homme 2015) es un recurso terminológico que contiene términos del medio ambiente en inglés, francés, español y portugués desarrollado por el Observatoire de linguistique Sens-Texte de la Universidad de Montreal (Canadá). Se ha compilado adaptando los principios teóricos y metodológicos de la lexicología explicativa y combinatoria (LEC) (Mel'čuk, Clas y Polguère 1995), al igual que su recurso hermano DiCoInfo[31] (L'Homme 2008), que contiene términos de informática.

DiCoEnviro parte de la premisa de que los verbos y sus derivados deben incluirse en los diccionarios especializados, ya que el análisis de los verbos especializados es un buen punto de partida para descubrir la estructura léxica de un campo de conocimiento. (L'Homme 2003: 405). De este modo, en DiCoEnviro se palían algunas de las deficiencias habituales en los recursos terminológicos (L'Homme 2008):

- La ausencia de términos que pertenecen a categorías gramaticales distintas del sustantivo, lo cual supone que términos con el mismo significado, pero de diferente categoría gramatical, se omitan, aunque a veces aparezcan como parte de compuestos.
- La falta de representación de determinadas diferencias semánticas como, por ejemplo, la polisemia regular que se da entre sustantivos deverbales que codifican una actividad o el resultado de esta.
- La omisión de la representación del comportamiento sintáctico de los términos y de su combinatoria.

[30] Disponible en: <http://olst.ling.umontreal.ca/dicoenviro>.

[31] Disponible en: <http://olst.ling.umontreal.ca/dicoinfo>.

Por lo tanto, en DiCoEnviro las entradas están dedicadas a términos de distintas categorías gramaticales: verbos, sustantivos, adjetivos, adverbios, así como locuciones. Por otro lado, cada entrada corresponde a una acepción distinta de cada término distinguiéndose así los casos de polisemia tanto regular como irregular (L'Homme 2008).

Una de las principales diferencias de DiCoEnviro con otros recursos terminológicos es que se proporciona la estructura argumental (o actancial, en la terminología de la LEC) de todos los términos predicativos. No obstante, a diferencia de la LEC, la estructura argumental en DiCoEnviro no se expresa mediante variables (X, Y, Z, etc.), sino mediante roles semánticos (L'Homme 2012b: 382), que, además, van acompañados del término o los términos que típicamente ocupan dicho argumento y una lista con otros términos que pueden instanciar los argumentos (L'Homme 2012a: 238).

Los roles semánticos utilizados en DiCoEnviro se encuentran, al igual que EcoLexicon, en un punto intermedio entre los macrorroles y los roles semánticos específicos para cada caso. Sin embargo, la lista de roles de DiCoEnviro es mucho más amplia que la de EcoLexicon e incluye, entre otros: AGENTE, PACIENTE, CAUSA, MANERA, RESULTADO, TIEMPO, etc., cuyo modo de empleo está claramente especificado en una guía para los terminólogos que trabajan en DiCoEnviro (L'Homme, Le Serrec y Laneville 2009).

Una entrada típica de DicoEnviro, como la de la Figura 5, incluye también información gramatical, sinónimos, contextos anotados con una metodología adaptada de la de FrameNet (L'Homme y Pimentel 2012) y relaciones léxicas de tipo paradigmático y sintagmático basadas en las funciones léxicas de la LEC.

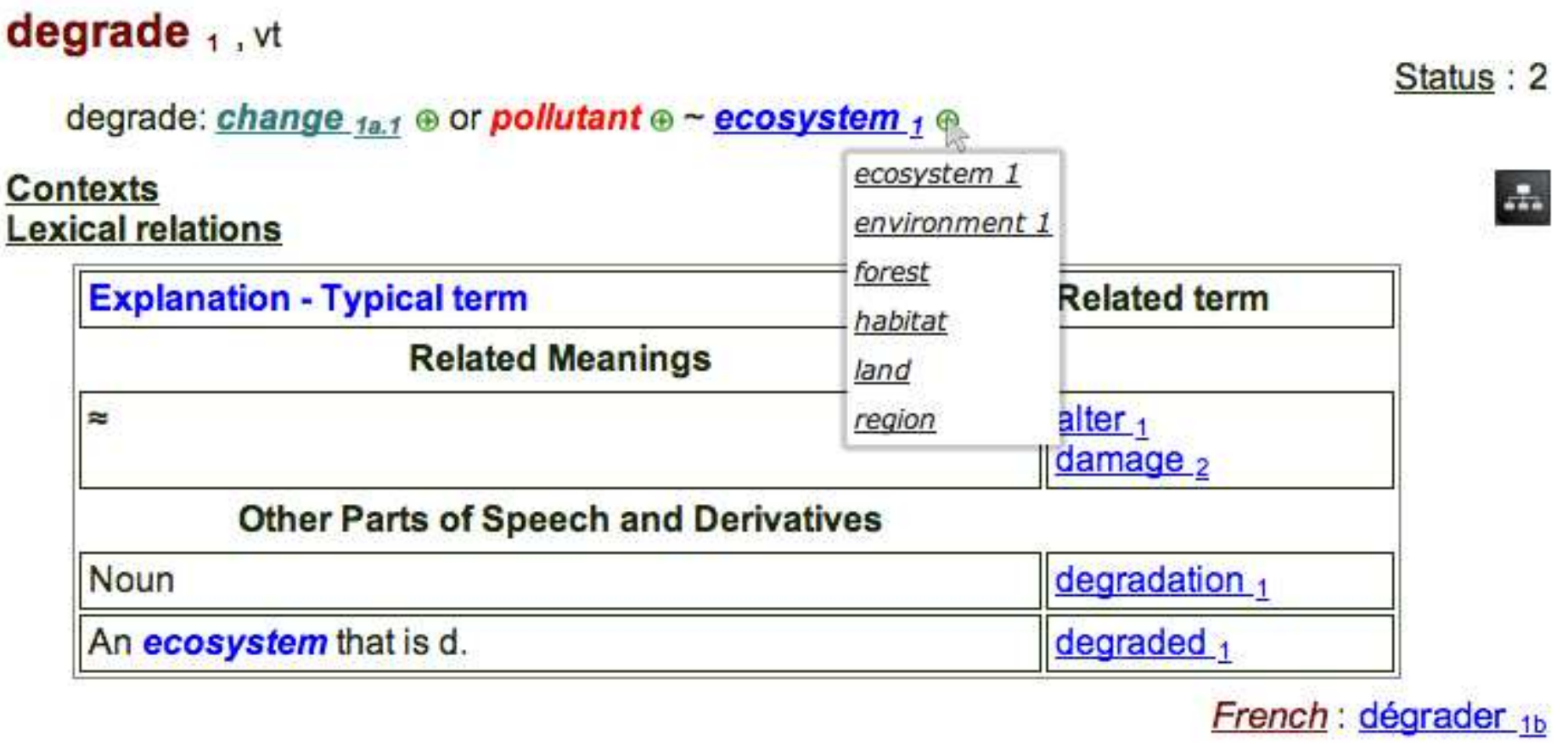

**Figura 5. Entrada en DiCoEnviro del término *degrade* (Fuente: DiCoEnviro)**

En cuanto a las definiciones, pocas entradas de DiCoEnviro las contienen en comparación con DiCoInfo, que se encuentra en un estado de desarrollo más avanzado en ese respecto. Hasta hace poco la redacción de definiciones en estos dos recursos no contaba con unas reglas precisas. Sin embargo, en San Martín y L'Homme (2014) se esbozaron unas primeras pautas para la confección de definiciones adaptando las reglas definicionales de la LEC (§3.3.9) a las características particulares de estos recursos.

Recientemente, se ha comenzado a dotar al recurso de una estructura de marcos basada en FrameNet para la versión inglesa y francesa de DicoEnviro (L'Homme, Robichaud y Subirats 2014; L'Homme y Robichaud 2014). El enfoque que se sigue consiste en descubrir los marcos de FrameNet evocados por los términos contenidos en DicoEnviro. Para ello, se realiza una comparación con los marcos de FrameNet a partir de la estructura argumental, las anotaciones de contextos y otra información léxico-semántica contenida en las entradas de los términos.

A diferencia de BioFrameNet, DiCoEnviro no es una extensión de FrameNet en sentido estricto. Mientras que BioFrameNet emplea la misma estructura y metodología que FrameNet y enlaza el recurso mediante una relación entre uno de sus marcos y otro de FrameNet, DiCoEnviro detecta aquellos marcos relevantes de FrameNet y los importa a su recurso con adaptaciones de distinto calibre según el caso, como por ejemplo, la sustitución de los elementos del marco por los roles semánticos de DiCoEnviro.

El enfoque empleado en DiCoEnviro toma como punto de partida los términos ya descritos en el recurso para llegar a los marcos; a diferencia de FrameNet, que comienza con la descripción del marco (L'Homme, Robichaud y Subirats 2014). Además, otra de las principales diferencias entre DiCoEnviro y FrameNet es que mientras DiCoEnviro describe los argumentos de cada término, FrameNet asocia los elementos del marco a un marco y no a una unidad léxica concreta (L'Homme, Robichaud y Subirats 2014: 1368). Teniendo eso en cuenta, se determinó que un término en DiCoEnviro evoca probablemente un marco existente en FrameNet si se podía establecer una relación entre los elementos del marco y los argumentos del término. No obstante, los autores tuvieron que establecer normas de equivalencias entre roles semánticos, ya que el inventario de FrameNet es muchísimo más específico que el de DiCoEnviro y, por otro lado, se tuvieron que gestionar determinadas diferencias entre el número de elementos del marco de FrameNet y el de argumentos de los términos en DicoEnviro (L'Homme y Robichaud 2014: 191–192).

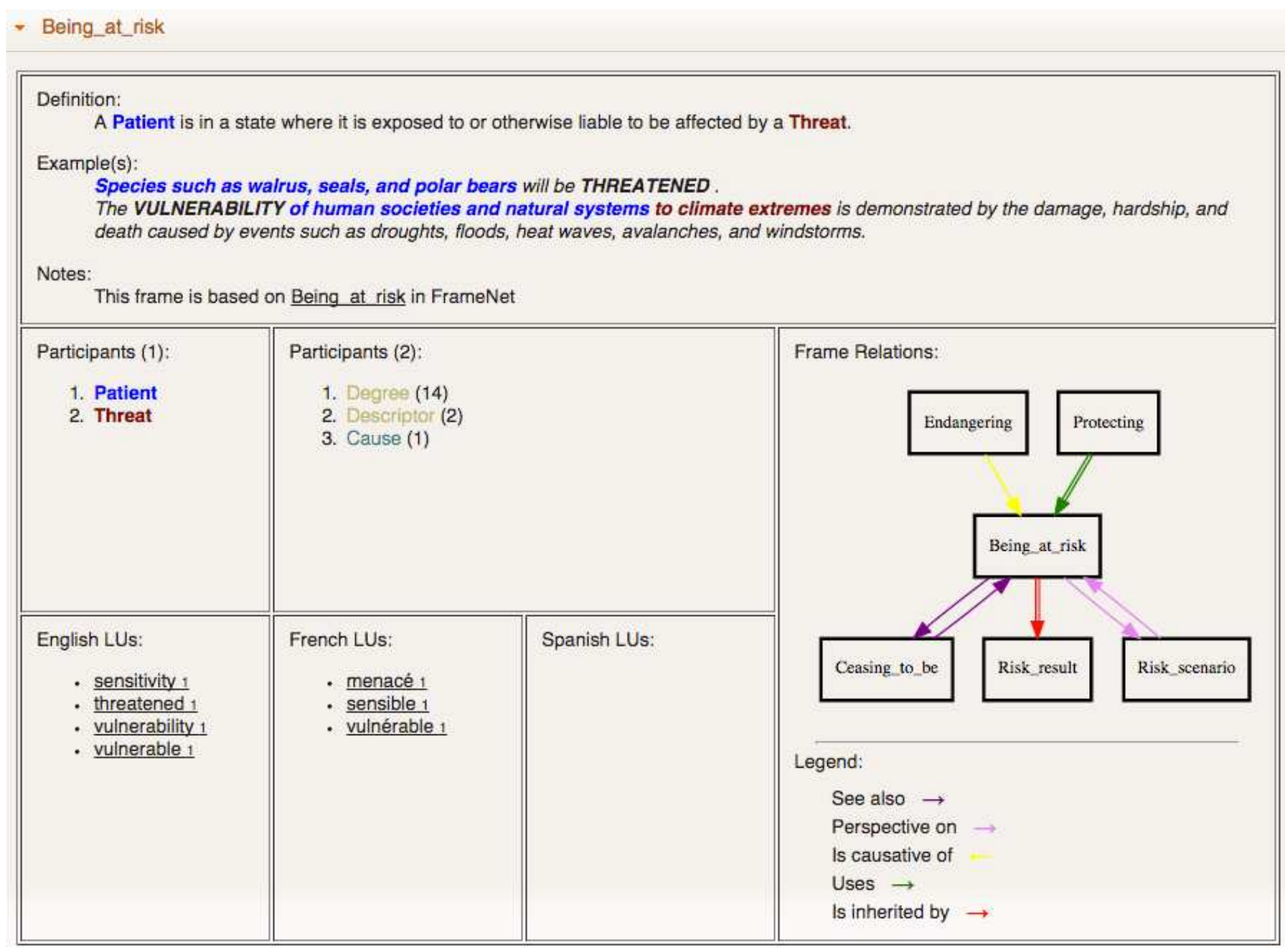

**Figura 6. Marco *Being_at_risk* en DiCoEnviro (Fuente: DiCoEnviro)**

Por otro lado, L'Homme y Robichaud (2014: 192–193) destacan que los marcos de DiCoEnviro —incluso aquellos tomados de FrameNet sin apenas cambios— difieren en cuanto que los marcos medioambientales son mucho más restringidos que los de FrameNet y que los términos que evocan los marcos podrían considerarse microsentidos (§3.5.3.5.4). Para ilustrarlo ponen el ejemplo del marco *Being_at_risk* (Figura 6), que importaron de FrameNet:

> [T]he Being_at_risk frame in the environment applies only to things such as species, ecosystems, plants, etc. In addition, the number of terms that evoke a frame is often much lower than those recorded in FrameNet. For instance, the terms evoking the Being_at_risk frame in the environment data are the following: sensitivity, threatened, vulnerability, vulnerable (whereas in FrameNet, the list comprises: danger.n, insecure.a, risk.n, safe.a, safety.n, secure.a, security.n, unsafe.a, vulnerability.n, vulnerable.a). (L'Homme y Robichaud 2014: 193)

En los casos en los que no se puede establecer una equivalencia con los marcos de FrameNet, DiCoEnviro crea marcos nuevos propios del dominio del medio ambiente (L'Homme y Robichaud 2014: 193), como, por ejemplo, *Cause_change_into_reusable_material* (Figura 7), que contiene los términos en inglés *compost* y *composting* que no están contenidos en FrameNet. No obstante, en francés, además de *compostage* y *composter*, también se incluye *transformer*, cuyo equivalente habitual en inglés (*transform*) sí está contenido en FrameNet evocando el marco *Cause_change*, lo cual se debe a que, como se indicó antes, el significado de *transformer* incluido en el marco de DiCoEnviro es una especialización o microsentido del significado no especializado del término.

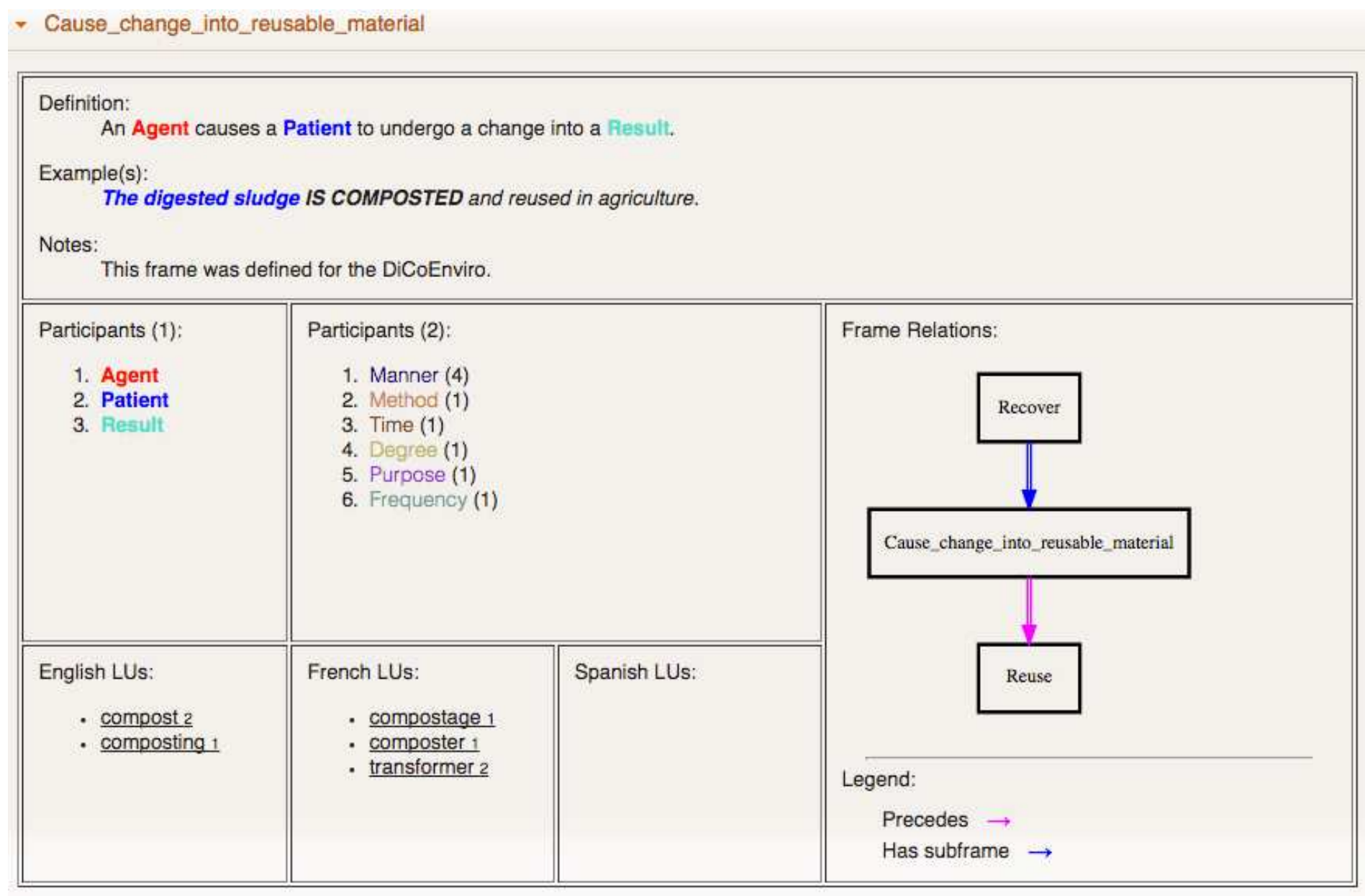

**Figura 7. Marco *Cause_change_into_reusable_material* en DiCoEnviro (Fuente: DiCoEnviro)**

A pesar del hecho de que la adición de marcos a DiCoEnviro está todavía en una fase inicial, sus resultados demuestran que una parte de los marcos ya descritos en FrameNet puede reutilizarse en un dominio especializado como el medio ambiente. Asimismo, es destacable el hecho de que los creadores hayan combinado principios de la LEC, de corte estructuralista, con la metodología de FrameNet, lo cual en un futuro podría permitir la combinación de las funciones léxicas con los marcos, llevando mucho más allá que el Kicktionary el uso de relaciones léxico-semánticas en conjunción con marcos.

#### 2.2.3.5 JuriDiCo

JuriDiCo[32] (Pimentel 2012a; Pimentel 2012b; Pimentel 2013; Pimentel 2014) es también un recurso terminológico concebido en el seno del Observatorie linguistique Sens-Texte de la Universidad de Montreal. Está dedicado al dominio del derecho e incluye términos en inglés, portugués y francés y, al igual que DiCoEnviro, está basado en la metodología de DiCoInfo (L'Homme 2008).

Por otro lado, también se diferencia de DiCoEnviro en la manera en que se han aplicado los marcos. Mientras que DiCoEnviro parte de los marcos ya descritos en FrameNet, JuriDiCo es independiente de FrameNet, aunque se nutre de su metodología. Sin embargo, al igual que DiCoEnviro se utiliza un enfoque *bottom-up*: se comienza por la caracterización de las unidades léxicas para después describir los marcos que estas evocan (Pimentel 2013: 245). Al igual que en FrameNet, un marco en JuriDiCo puede agrupar tanto sinónimos y cuasi-sinónimos como antónimos y términos relacionados.

El objetivo principal de JuriDiCo es emplear los marcos como herramienta interlingüística para la determinación de equivalentes de traducción de verbos especializados que aparecen en textos jurídicos (Pimentel 2013: 244). En este sentido, a diferencia de los diccionarios especializados tradicionales, JuriDiCo distingue entre equivalencia total o parcial. Para que una equivalencia se considere total es necesario que los términos evoquen el marco de la misma forma y que tengan el mismo número y tipo de argumentos (Pimentel 2013: 251).

Las entradas de términos en JuriDiCo son similares a las de DiCoEnviro con la excepción de que en JuriDiCo no se incluye el módulo de relaciones léxicas basadas en las funciones léxicas de la LEC. Sin embargo, todas las unidades léxicas van acompañadas de la información del marco que evocan (Figura 8) y siempre incluyen una definición al estilo de la LEC (§3.3.9).

[32] Disponible en: <http://olst.ling.umontreal.ca/juridico>.

Como se puede observar en la Figura 8, la entrada de un marco incluye una definición de este, que puede coincidir o no con la definición de alguno de las unidades léxicas que lo evocan (Pimentel 2012a: 298), la definición de los elementos del marco obligatorios, un listado de los elementos del marco opcionales y las unidades léxicas que lo evocan. A diferencia de FrameNet, la entrada de los marcos en JuriDiCo no incluyen relaciones entre marcos, pues no se han aplicado en el recurso.

JuriDiCo demuestra la utilidad de los marcos como herramienta interlingüística incluso en un dominio como el derecho en el que se presentan enormes diferencias de una lengua a otra dada la existencia de distintos sistemas jurídicos.

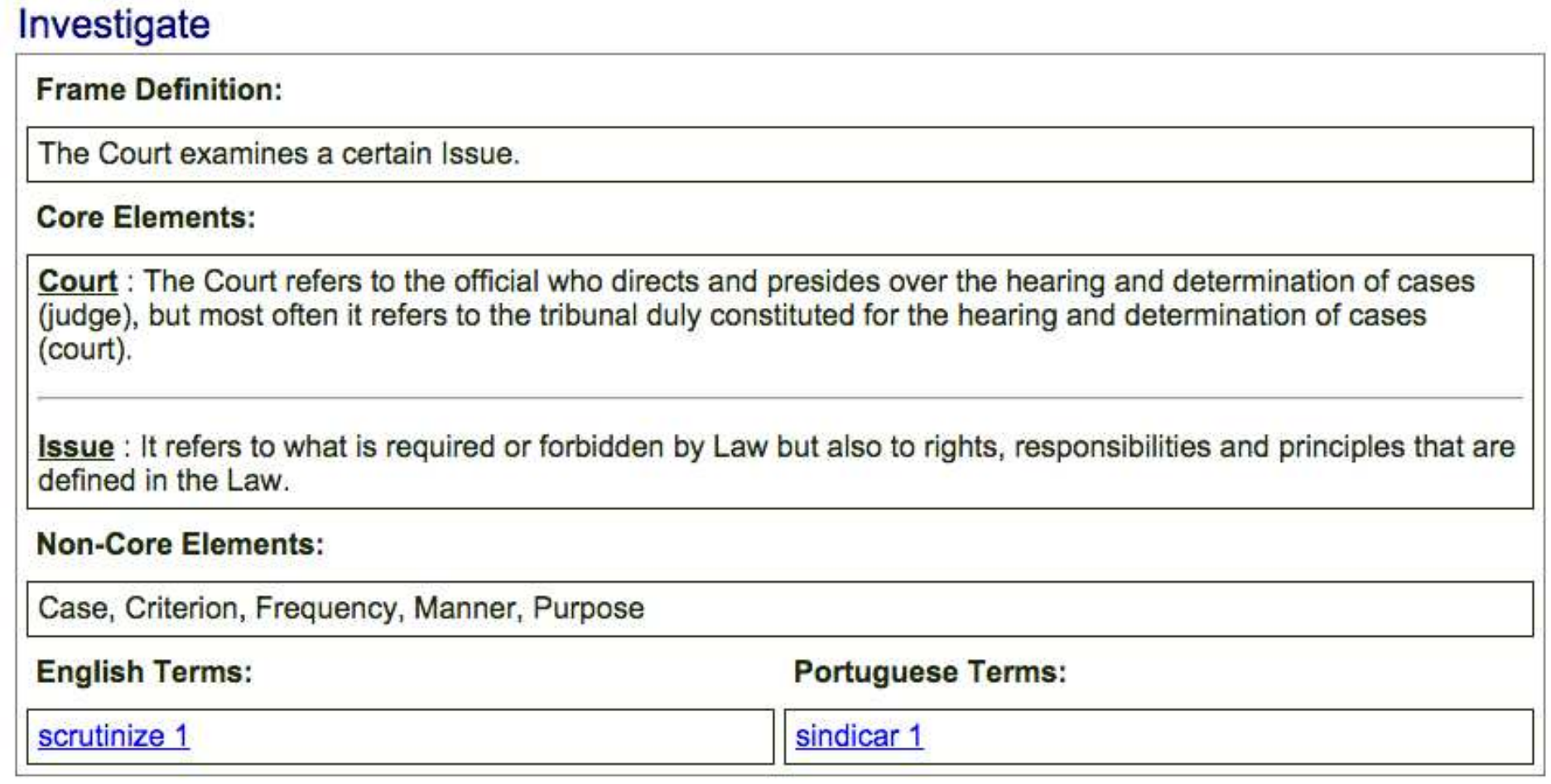

**Figura 8. Marco *Investigate* en JuriDiCo (Fuente: JuriDiCo)**

# 3 LA DEFINICIÓN

La definición es objeto de estudio principalmente dentro del ámbito de la lingüística y de la filosofía. Sin embargo, al ser un medio de transmisión del conocimiento, la definición se emplea en numerosas esferas de la vida humana:

> [T]he practice of defining is present in many domains of human activity and can be observed from such simple everyday situations as friendly and informal conversations in a pub, through definitions in textbooks to those found in very advanced fields of science. Language users define frequently, consciously or unconsciously, and for different purposes. (Fabiszewski-Jaworski 2012: 13)

Dada la multitud de situaciones distintas en las que se puede producir, la definición es una noción compleja que puede entenderse desde al menos cuatro puntos de vista distintos (Seppälä 2012: 5):

- *Actividad de definir*: actividad mental, verbalizada o no, realizada por un emisor, consistente en formar una *representación mental* definitoria acerca de X y comunicarla eventualmente a un destinatario en un *acto* definitorio.
- *Definición como representación mental*: representación mental compuesta por un conjunto de conocimientos o creencias acerca de X que constituyen el contenido definitorio.
- *Definición como artefacto representacional*: artefacto representacional lingüístico que toma la forma de un enunciado expresado por el emisor para representar el contenido definitorio resultado de la *actividad* definitoria con vistas a comunicarla al destinatario.
- *Acto de definir*: Acto comunicativo realizado directa o indirectamente por un emisor que consiste en responder mediante un enunciado definitorio (*artefacto representacional*) a la pregunta genérica «¿qué es X?» formulada por un destinatario.

Así pues, la definición puede concebirse, por un lado, como la actividad de definir que tiene como resultado la generación de una representación mental de un contenido definitorio y, por otro, como el acto de definir que consiste en verbalizar dicha representación mental mediante un artefacto representacional en forma de enunciado.

En el ámbito de la terminología, lo habitual es utilizar el término *definición* para hacer referencia principalmente a un enunciado que representa un contenido definitorio. Como se ha visto, ese contenido definitorio responde a la pregunta genérica «¿Qué es X?» hecha por un destinatario. En referencia a una definición terminológica, se entiende que dicha X es o bien un término (con sus diferentes acepciones) o bien un concepto.

Mientras que definir un concepto consiste en describir determinado contenido conceptual, ¿en qué consiste definir un término? Como explica (Seppälä 2012: 9), definir un término puede entenderse como definir el signo, definir el referente del término, definir el significado del término o definir el concepto asociado al término. A continuación, exponemos nuestro punto de vista al respecto.

En primer lugar, en este trabajo, descartamos que el objeto de la definición deba ser el signo, pues consideramos que la función de la definición no es la de describir el signo (p. ej., sus características morfosintácticas), ya que, para ello, se deben utilizar otros elementos dentro de un recurso como los campos destinados a la información gramatical, morfológica, etc., como indica Rey-Debove (1967: 145).

No obstante, las definiciones que tienen como objeto el signo existen y reciben el nombre de *definiciones metalingüísticas* y son, en la mayoría de casos, tachadas de incorrectas (Vézina et al. 2009: 25). Un ejemplo de este tipo de definición son aquellas encabezadas por «dícese de», «aplícase a», etc. (ej. 12) o definiciones encabezadas por una caracterización gramatical o de otro tipo del signo (ej. 13).

ej. 12 BARÓMETRO. Dícese del instrumento utilizado para medir la presión atmosférica.

ej. 13 CLINÓMETRO. Sustantivo masculino que designa un instrumento utilizado para medir ángulos de inclinación.

Asimismo, también descartamos que el objeto de la definición sea el referente de la unidad terminológica. La gran mayoría de términos no tienen como referente una única entidad en el mundo, sino que puede haber muchos ejemplares del referente. Por ejemplo, el término *altímetro* tiene como referente todas aquellas entidades en el mundo que se consideren miembros de la categoría ALTÍMETRO. Incluso suponiendo que la categoría ALTÍMETRO tuviera límites precisos, describir el referente de *altímetro* implicaría hacer una descripción de todos y cada uno de los altímetros que existen, han existido y existirán, lo cual es virtualmente imposible.

Incluso en el caso de términos con un único referente (*sol*, *luna*, etc.), los seres humanos también creamos conceptos para esos referentes (SOL, LUNA, etc.) que se integran en nuestro sistema conceptual (lo que Barsalou et al. [1993] denominan *marcos de individuo* [§2.1.1.1.1]) y, como cualquier otro concepto, se enriquecen con las nuevas experiencias y los conocimientos que se adquieren sobre ellos. Como ya se señaló, los conceptos no son un simple reflejo de la realidad externa (Lakoff y Johnson 1999: 22); están basados en ella y están influidos por nuestra corporeización, el entorno social y cultural (Lakoff 1987: xv; Lakoff y Johnson 1980: 180). Los conceptos con un único referente en el mundo tienen las mismas características que cualquier otro tipo de concepto, como la multidimensionalidad o la ausencia de límites precisos. Por lo tanto, dado que no es posible describir un referente sin la mediatización de la cognición, la definición no puede tener por objeto el referente en sí.

Al descartar el signo y el referente, la opción restante es el significado o el concepto. En el caso de la lexicografía, siguiendo un punto de vista estructuralista, se ha supuesto que el objeto de la definición lexicográfica es el significado, entendido como un conocimiento de tipo lingüístico estable e independiente del conocimiento enciclopédico. Por su parte, en terminología, desde un enfoque wüsteriano, el significado de un término se considera equivalente al concepto al que está asociado y, por ende, tradicionalmente el objeto de la definición terminológica es el concepto, el cual tiene límites precisos y características suficientes y necesarias.

En este trabajo, defendemos que las unidades léxicas solo tienen significado en eventos de uso concretos (§2.1.2). Fuera de los eventos de uso, es decir, de manera abstracta, una unidad léxica no tiene significado: tiene potencial semántico. Por ello, no es posible concebir que definir un término sea definir su significado.

Así pues, la opción restante es que definir un término implique definir un concepto. Sin embargo, es necesario realizar algunas precisiones. Como hemos indicado, las unidades léxicas tienen potencial semántico lo cual no es exactamente lo mismo que estar asociado a un concepto. Dentro del potencial semántico de un término podemos distinguir el perfil, que es el concepto (o conceptos, en caso de homonimia o polisemia) al que está asociada de manera convencional la unidad léxica[33], y la base, que son los marcos que presupone el perfil (Figura 9).

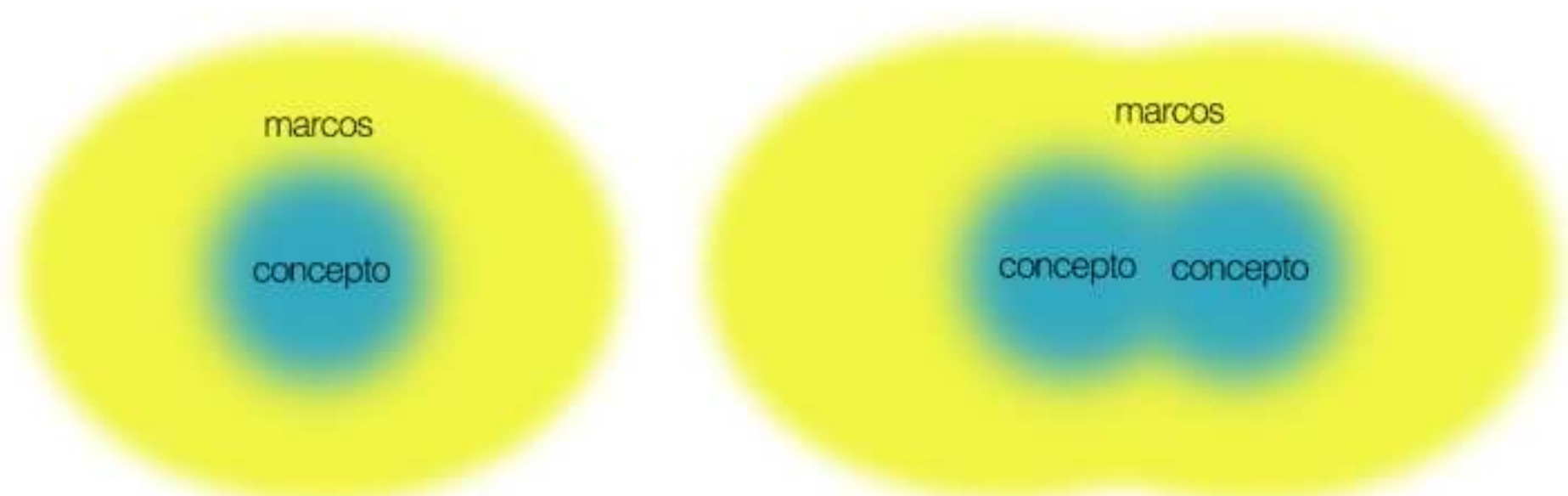

**Figura 9. Representación esquemática del potencial de significado de una unidad léxica asociada a un único concepto (a la derecha) y de una unidad léxica polisémica asociada a dos conceptos (a la izquierda)**

El potencial semántico del término es la materia prima de la definición, pero no el objeto, ya que asumir que el potencial semántico es el objeto implicaría que la definición sería una descripción de todo el contenido conceptual que una unidad léxica puede activar. Este enfoque no es viable porque el potencial semántico de una unidad léxica es una cantidad inabarcable de información, que, además, nunca se activa de manera completa en eventos de uso reales.

Como vimos en §2.1.2.1, las unidades léxicas, además de poseer potencial semántico, también tienen asociadas restricciones de tipo convencional y contextual. Dichas restricciones crean tendencias de activación de determinado contenido conceptual, formando lo que Croft y Cruse (2004: 110) denominan *presignificados*. Los presignificados son unidades conceptuales que, en el proceso de conceptualización, se encuentran a caballo entre el potencial semántico y el significado en un evento de uso concreto:

> Pre-meanings [are] units that appear somewhat further upstream in the construal process than full-fledged interpretations. [...] [A] pre-meaning is still subject to further construal. It is, however, more elaborated than 'raw' purport. (Croft y Cruse 2004: 110)

[33] Como se verá en §3.5.3.5.1, en caso de polisemia, una unidad léxica puede estar asociada a más de un concepto.

Los presignificados pueden aparecer en distintos niveles de conceptualización. Por un lado, pueden presentarse más cercanos al potencial semántico, como, por ejemplo, los presignificados correspondientes a los distintos conceptos a los que está asociada una unidad léxica polisémica u homónima. Por otro lado, pueden corresponderse con unidades subconceptuales, con la posibilidad de que los distintos presignificados estén anidados (Croft y Cruse 2004: 110).

Por ejemplo, en el caso del término *planta*, un primer nivel de presignificación se correspondería con la distinción entre los varios conceptos a los que está asociado, por ejemplo, PLANTA(*ser vivo*) y PLANTA(*instalación industrial*). La convención ha hecho que dichos conceptos estén asociados a esa unidad léxica y cada uno de ellos tiene una tendencia de activación mayor según el contexto. Además, las características de dichos conceptos tendrán distintos niveles de centralidad, es decir, que tendrán un prototipo asociado que activará con preferencia determinados marcos. Como hemos visto, los prototipos son también sensibles al contexto, por lo que dentro de los distintos conceptos, las restricciones contextuales harán que haya presignificados subsumidos. Por ejemplo, el siguiente nivel de presignificación de *planta*, a partir de PLANTA(*ser vivo*), podría corresponderse con cómo se conceptualiza ese concepto en el dominio de la agricultura.

El objeto de la definición es, pues, un subconjunto del potencial semántico que se corresponde con algún tipo de presignificado (Figura 10). El presignificado que será objeto de la definición dependerá de las restricciones contextuales que se impongan a la definición (§3.5.3.2). En todo caso, dicho subconjunto corresponderá con parte de un único concepto y de los marcos que este pueda activar. Si el potencial semántico es polisémico, es decir, que incluye más de un concepto, será necesario crear al menos una definición para cada concepto.

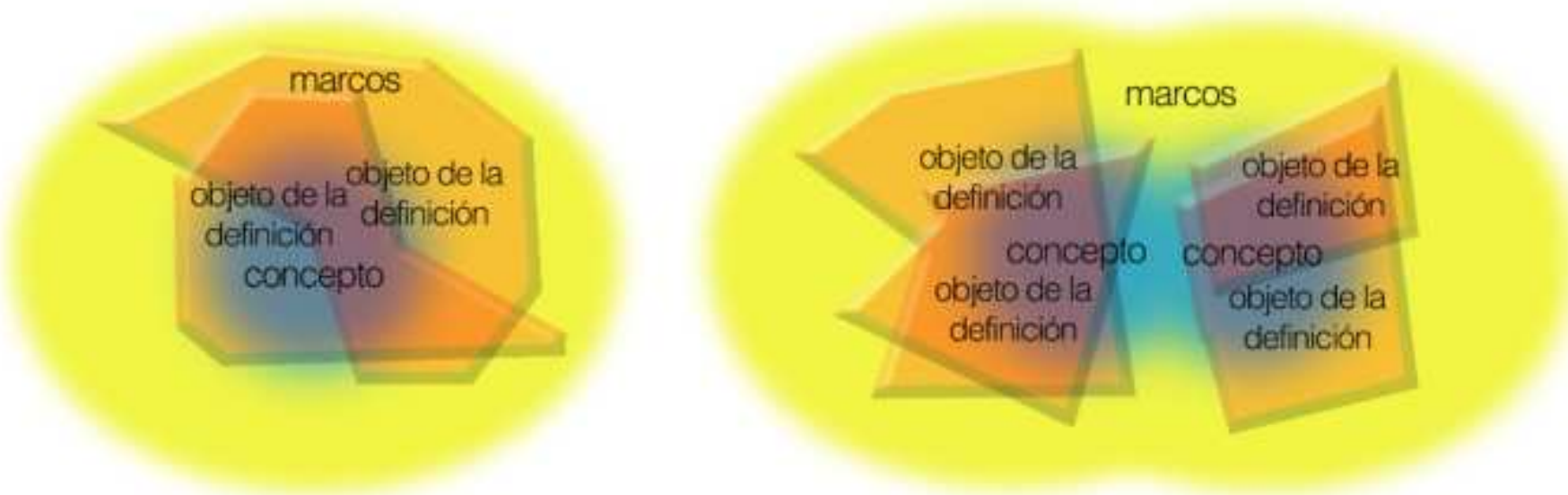

**Figura 10. Representación esquemática del potencial de significado y un ejemplo de un subconjunto que podría ser objeto de la definición de una unidad léxica asociada a un único concepto (a la derecha) y de una unidad léxica polisémica asociada a dos conceptos**

El potencial semántico de una unidad léxica (al igual que cualquier otro contenido conceptual) se encuentra almacenado en la mente de los hablantes. En el caso de las unidades terminológicas, el potencial semántico reside en la mente de los expertos. Sin embargo, dada la multidimensionalidad del conocimiento especializado, expertos de distintos campos del saber tendrán almacenado un potencial semántico distinto (con mayor o menor solapamiento) para el mismo término. Por ello, en línea con la noción de *división del trabajo lingüístico* (Putnam 1975), podríamos afirmar que el potencial semántico de los términos se encuentra repartido en la mente de los expertos de todos los campos del saber y se refleja en los eventos de uso que producen de dichos términos. Como se verá en el apartado relativo a la dimensión contextual de la selección de rasgos (§3.5.3), el terminólogo, para la elaboración de definiciones, reconstruye presignificados de un término correspondiente a una o varias áreas del saber.

Por lo tanto, desde nuestro punto de vista, definir un término supone ofrecer al menos una definición por cada concepto contenido en el potencial semántico del término[34] que describa la parte de dicho concepto y los marcos que este activa que el terminólogo considere relevantes. Por su parte, definir un concepto implica describir un subconjunto considerado relevante de sus rasgos junto con el conocimiento de fondo que necesariamente acompaña a ese subconjunto de contenido conceptual.

[34] Los conceptos pueden recibir múltiples definiciones. En este trabajo consideramos, pues, que definir implica proporcionar una o más definiciones de las múltiples posibles.

Como veremos en §3.5., la relevancia de los rasgos depende de factores de tipo ontológico, funcional y contextual.

## 3.1 LA DEFINICIÓN TERMINOLÓGICA

Tradicionalmente, desde un punto de vista estructuralista y wüsteriano, se había establecido una diferencia entre la definición lexicográfica, terminológica[35] y enciclopédica. La definición lexicográfica (o definición lingüística en términos de Cabré [1993: 209]) define el significado de las unidades léxicas no especializadas como parte de la estructura de la lengua y las explica en el contexto de otras unidades léxicas (Sager y Ndi-Kimbi: 1995). El significado de las palabras se considera puramente lingüístico y al considerarse que el objeto de la definición es el significado, ello supone que una definición lexicográfica no debe hacer referencia a ningún tipo de conocimiento extralingüístico.

Por su parte, la definición terminológica, que, desde el punto de vista tradicional, define el concepto que designa una unidad léxica especializada o término, tiene por objetivo la identificación de rasgos necesarios y suficientes del concepto dentro de los límites de un dominio concreto (Sager y Ndi-Kimbi 1995: 83). Por lo tanto, la definición terminológica se considera de naturaleza no lingüística.

Finalmente, la definición enciclopédica (o definición ontológica en términos de Cabré [1993: 209]) al igual que la terminológica, se concibe como extralingüística, pero al contrario de esta, no se limita a los rasgos necesarios y suficientes del concepto que describe, sino que proporciona una descripción global y exhaustiva de todos los aspectos del concepto, habitualmente con una intención didáctica (Sager y Ndi-Kimbi 1995: 83).

[35] De Bessé (1990: 253–254) y Béjoint (1993: 19) proponen una distinción entre definición terminográfica y terminológica. De acuerdo con estos autores, la definición terminográfica tiene la función de explicar un término, mientras que la definición terminológica sirve para fijar el significado de un término preexistente de manera prescriptiva o fijar el significado de un término de nueva creación. Esta distinción ya no se aplica comúnmente en Terminología y, en consecuencia, en este trabajo, consideramos *definición terminológica* y *definición terminográfica* como sinónimos.

Aplicar un enfoque cognitivista a la definición implica deshacer estas distinciones entre estos tipos de definición tal y como se han establecido tradicionalmente. Como vimos en §2.1.2.1, no es posible distinguir entre conocimiento lingüístico (semántico) y conocimiento extralingüístico (enciclopédico), ya que forman un contínuum.

Por lo tanto, la distinción entre definición terminológica y lexicográfica queda reducida a que la primera define un subconjunto del potencial semántico de los términos y la segunda, de las palabras de la lengua general. No obstante, dicha distinción traslada el problema a la dicotomía entre palabra y término, que también es difusa. De hecho, la TBM, como ya indicamos, considera que distinguir entre palabra y término no es solo inviable sino que, además, no resulta fructífero (Faber, León Araúz y Prieto Velasco 2009: 4). En este sentido, compartimos la recomendación de L'Homme (2005: 1113) que describe el término como una unidad léxica cuyo significado está relacionado con un dominio especializado e indica que para abordar la noción de término es esencial tener en cuenta su aplicación:

> La définition de « terme », contrairement à celle qui est donnée pour d'autres unités linguistiques, est donc relative. Elle dépend de la délimitation qu'on a faite d'un domaine spécialisé et les objectifs visés par une description terminologique (L'Homme 2005: 1125).

De este modo, en lo que a nuestros propósitos se refiere, más importante que determinar si una unidad léxica es un término es el hecho de decidir si una unidad léxica debe ser objeto de descripción dentro de un recurso terminológico, lo cual dependerá de los criterios establecidos para la inclusión de términos. Como indica Cabré (1999: 123), el carácter de término se activa en función del contexto. Así pues, dado que describir una unidad léxica y la información conceptual asociada a esta dentro de un recurso terminológico implica que la descripción se hará desde el punto de vista de uno o varios dominios especializados, dicha unidad adquirirá, en mayor o menor medida, un valor especializado.

Finalmente, en lo que respecta a la distinción entre definición terminológica y enciclopédica, defendemos que, dado que la identificación de rasgos suficientes y necesarios es virtualmente imposible, todas las definiciones serán enciclopédicas en lo que respecta al tipo de contenido representado. En cuanto a la supuesta globalidad y exhaustividad de la definición enciclopédica frente a la definición terminológica, una definición terminológica puede ser todo lo global y exhaustiva que el terminólogo considere adecuado según las características del recurso en el que se va a insertar.

De acuerdo con Seppälä (2012: 5), se puede caracterizar la definición terminológica en relación con cinco elementos: el emisor, el destinatario, la situación comunicativa, la modalidad de expresión y el objeto.

El emisor de la definición terminológica suele ser un terminólogo o un especialista del dominio, o ambos en colaboración, mientras que el destinatario es el público meta de la definición, el cual dependerá del recurso terminológico en el que se vaya a insertar la definición (Seppälä 2012: 7).

En lo que respecta a la situación comunicativa y la modalidad de expresión, la definición terminológica se produce en una situación comunicativa *in absentia*, es decir, el acto definitorio tiene lugar sin que el emisor esté presente (por ejemplo, cuando el destinatario lee la definición), y la modalidad de expresión es el enunciado lingüístico en contraposición, por ejemplo, con señalar con el dedo una entidad o situación, o mostrar una imagen[36] (Seppälä 2012: 8).

A continuación, presentamos la definición aristotélica, que es el formato definicional clásico y sigue siendo el más utilizado en recursos lexicográficos y terminológicos, en particular en la definición de sustantivos. Como se verá, en este trabajo, utilizaremos el modelo aristotélico, pero desde un punto de vista cognitivista.

[36] Algunos autores consideran que señalar con un dedo o mostrar una imagen es un tipo de definición llamada *definición ostensiva* (Marciszewski 1993).

## 3.2 LA DEFINICIÓN ARISTOTÉLICA

De manera habitual, cualquier tipo de definición consta de dos partes: el definiéndum (unidad léxica o concepto que se está definiendo) y el defíniens: enunciado cuya finalidad es definir el definiéndum. Habitualmente, recibe simplemente el nombre de *definición*.

El defíniens de la definición aristotélica (también llamada *definición tradicional, definición hiperonímica, definición analítica, definición intensional, definición por género próximo y diferencia específica, definición por genus y differentiae, definición inclusiva*, etc.) consiste en presentar el genus (hiperónimo o superordinado de la unidad léxica/concepto que se define) y una o varias differentiae (atributos que distinguen a la unidad de sus cohipónimos). Esta tradición procede de los neoplatónicos, en particular de Porfirio, que escribió una introducción (*Isagoge*) en griego a la lógica de Aristóteles. La *Isagoge* de Porfirio la tradujo al latín Boecio, a través de cuya traducción nos ha llegado la tradición de definir mediante genus y differentiae. Sin embargo, en la *Isagoge*, se exponían en total cinco conceptos de Aristóteles relacionados con la definición (P. Hanks 2006: 399):

- *Genus* (género): grupo de *species* que tienen atributos en común.
- *Species* (especie): grupo de cosas similares en esencia.
- *Differentia* (diferencia): parte de la esencia particular de una *species* en concreto y por tanto que la distingue de las demás.

- *Proprium* (propiedad): atributo compartido por todos los miembros de la especie que no forma parte de la esencia ni es necesario para diferenciarla.
- *Accidens* (accidente): atributo presente solamente en algunos miembros de la especie y que no forma parte de su esencia.

Por ejemplo, un ser humano (*species*) puede definirse como un animal (*genus*) racional (*differentia*). Asimismo, podemos afirmar que un ser humano tiene la capacidad de respirar, lo cual sería un *proprium* porque todos los seres humanos respiran; pero no es un atributo que forme parte de la esencia ni sirve para diferenciarlo de otras *species*. Por último, podemos decir que hay seres humanos que son rubios, lo cual sería un *accidens* porque es un atributo que solo se da en algunos seres humanos (Smith 2007).

Como expone P. Hanks (1987: 119), el estilo de las entradas en los diccionarios empezó a formalizarse a principios del siglo XVIII, pues anteriormente las definiciones eran una mezcla de equivalentes entre palabras y descripciones informales. A la sazón, Leibniz formuló la noción de que dos expresiones son sinónimas si una puede sustituirse por la otra *salva veritate* (quedando intacta la verdad). Esta noción tuvo gran repercusión en la lexicografía, pues comenzó a seguirse lo que ha venido a llamarse el *principio de sustitución*, que consiste en la elaboración de definiciones que pueden sustituirse en cualquier contexto por la palabra que se define. El objetivo primordial de este principio es el de verificar si las características representadas en la definición son suficientes y necesarias. Como ya vimos en §2.1.3.1.1, la determinación de características suficientes y necesarias no es factible, por lo que, en este trabajo, rechazamos el uso del principio de sustitución.

Defendemos el uso de una versión modificada de la definición aristotélica. El formato sigue siendo el mismo: «definiéndum = genus + differentiae», puesto que el uso del genus, es decir, de un concepto superordinado permite representar el proceso de categorización —que, como se vio, es un proceso básico de la cognición humana— y, por consiguiente, la herencia de propiedades. Sin embargo, la noción de differentiae la redefinimos y deja de referirse a rasgos suficientes y necesarios, para ser rasgos relevantes de esa categoría según distintos factores (§3.5).

A continuación, revisaremos algunos tipos de definición diferentes de la definición aristotélica (aunque, en algunos casos, variantes de esta) que habitualmente se utilizan —solos o combinados en una misma definición— en recursos lexicográficos y terminológicos.

## 3.3 OTROS TIPOS DE DEFINICIÓN

### 3.3.1 La definición extensional

La definición extensional —también llamada *definición denotativa* (Robinson, 1950; Sager, 1990) y definición enumerativa (Shelov, 2003)— consiste en la enumeración de todos los hipónimos de un concepto. Un ejemplo de definición extensional sería:

> ej. 14 EVENTO EXTREMO. Los eventos extremos son los terremotos, los maremotos, los deslizamientos de tierras, los huracanes, los tornados, las inundaciones y las erupciones.

Este tipo de definición, considerada junto con la aristotélica (§3.2) y la partitiva (§3.3.2), uno de los tres tipos principales de definición según la TGT (Temmerman 2000: 8), no está exenta de problemas. En primer lugar, carece de genus y, por ende, el definiéndum no se categoriza, lo cual dificulta que el usuario pueda situar el concepto dentro de sus conocimientos previos. Por otro lado, esta definición reposa sobre la suposición de que el usuario va a conocer los conceptos que se exponen en la definición. Pero, incluso aunque el usuario, en el ej. 14, conozca todos los tipos de EVENTO EXTREMO que se presentan, tendrá que hacer sus propias suposiciones, las cuales pueden estar erradas, para colegir cuáles son las características comunes a todos los cohipónimos.

Las relaciones genérico-específicas (además de la codificada por el genus) pueden resultar útiles en definiciones que reflejen además otros tipos de relaciones conceptuales, pero no en definiciones basadas exclusivamente en esas relaciones como la definición extensional.

### 3.3.2 La definición partitiva

La definición partitiva define un concepto mediante la enumeración de las partes que lo componen (ej. 15) o indicando de qué forma parte (Vézina et al. 2009: 38). Un ejemplo de definición partitiva sería:

> ej. 15 ATMÓSFERA. La atmósfera está formada por cinco capas principales: troposfera, estratosfera, mesosfera, termosfera y exosfera.

La definición partitiva presenta problemas similares a la definición extensional, como la carencia de genus y una dependencia excesiva de los conocimientos previos del usuario. En el ej. 15, podemos suponer que si el usuario conoce con precisión qué son la TROPOSFERA, la ESTRATOSFERA, la MESOSFERA, la TERMOSFERA y la EXOSFERA, entonces, sabrá sin duda qué es la ATMÓSFERA y no necesitará una definición. Al mismo tiempo, si el usuario desconoce dichos conceptos, esta definición será totalmente ineficaz.

Así pues, de manera similar a lo expresado sobre la definición extensional, las relaciones partitivas pueden resultar útiles en definiciones que reflejen además otros tipos de relaciones conceptuales, pero no en definiciones basadas exclusivamente en esta relación.

### 3.3.3 La definición sinonímica

La definición sinonímica (Bosque 1982; Ilson 1987; Robinson 1950; Sager 1990) consiste en definir ofreciendo uno o varios sinónimos. Este tipo de definición asume que el usuario conocerá el sinónimo, lo cual en recursos lexicográficos es «arriesgado» (Harris y Hutton 2009: 111) y se podría afirmar que en recursos terminológicos lo es todavía más, pues es aún menos probable que el usuario conozca el sinónimo en cuestión. En todo caso, en un recurso terminológico orientado al concepto como EcoLexicon, las definiciones sinonímicas no tienen lugar, puesto que los términos considerados sinónimos se asocian al mismo concepto y tienen, por ende, una definición común. Un ejemplo de definición sinonímica es:

ej. 16 COMIDA. Alimento.

También es habitual en algunos recursos lexicográficos ofrecer como definición varios sinónimos parciales, lo cual es aún menos riguroso porque se deja que el usuario adivine qué parte del contenido semántico asociado a cada sinónimo parcial debe elegir (Harris y Hutton 2009: 111).

### 3.3.4 La definición antonímica

La definición antonímica es una noción de Bosque (1982) que abarca la definición por exclusión y la definición por inclusión negativa de Rey-Debove (1967). El primer tipo consiste en definir una palabra negando su opuesto mediante un hiperónimo de significado negativo (ej. 17) y el segundo, mediante una negación sintáctica (ej. 18) (Rey-Debove 1967: 153-155).

ej. 17 ASIMETRÍA. Falta de simetría. [DLE[37]]

ej. 18 INCONFUNDIBLE. No confundible. [DLE]

Dentro de la definición antonímica, se puede concebir un tercer tipo: la definición mesonómica, consistente en una definición por exclusión doble (Borsodi 1966 *apud* Bosque 1982: 110). En una definición mesonómica, el definiéndum se identifica por medio de la exclusión de otras dos unidades:

ej. 19 INDIFERENCIA. Estado de ánimo en que no se siente inclinación ni repugnancia hacia una persona, objeto o negocio determinado. [DLE]

[37] *Diccionario de la lengua española* (Real Academia Española 2001)

La definición antonímica en terminología presenta los mismos problemas que la definición sinonómica en cuanto que confía en exceso en los conocimientos previos del usuario. Además, en el caso de la definición por excluyente negativo el genus es incorrecto, ya que, tomando el ej. 17, no se puede afirmar que *lentitud* sea un tipo de *falta*. No obstante, es posible incluir relaciones de oposición en una definición, pero no debe ser lo que determine el genus.

### 3.3.5 La definición serial

La definición serial (Bosque 1982: 109–110) es un tipo de definición en la que el definiéndum se sitúa en un punto de una determinada escala. Puede ser aristotélica o no. Bosque establece una tipología dentro de este tipo de definición recurriendo a las tres clases de estructuras seriales que distinguió Fillmore (1978 *apud* Bosque 1982: 109-110): ciclos, cadenas y redes.

Los ciclos representan series lineales cerradas como «mañana-tarde-noche» o «primavera-verano-otoño-invierno», es decir, la mañana precede a la tarde, la tarde precede a la noche, la noche precede a la mañana y así sucesivamente. Una definición serial de este tipo podría ser:

ej. 20 MARTES. Segundo día de la semana, posterior al lunes y anterior al miércoles.

Por su parte, las cadenas son series lineales no cíclicas, como «suspenso-aprobado-notable-sobresaliente». Por ejemplo:

ej. 21 NOTABLE. Calificación usada en los establecimientos de enseñanza, inferior al sobresaliente y superior al aprobado. [DLE]

Finalmente, las redes son estructuras que muestran relaciones cruzadas, como los sistemas de parentesco. Por ejemplo:

ej. 22 PRIMO. Respecto de una persona, hijo o hija de su tío o tía. [DLE]

Representar las relaciones entre conceptos que forman series puede ser útil en definiciones terminológicas, aunque es posible que en muchos casos deba añadirse más información, por lo que una definición exclusivamente serial es de limitada utilidad.

### 3.3.6 La definición morfosemántica

La definición morfosemántica (De Bessé 1990; Vézina, Darras, Bédard y Lapointe-Giguère 2009: 38) —también denominada *definición por paráfrasis* (Sager 1990)— consiste en describir una unidad léxica explicitando su morfología mediante una estructura semánticamente equivalente.

ej. 23 CONTAMINACIÓN. Acción y efecto de contaminar. [DLE]

Este tipo de definición obliga al usuario a consultar la definición de la otra unidad léxica a la que se le asocia en la definición. Ello se puede evitar sustituyendo dicha unidad léxica por su definición (en este caso, por la definición de *contaminar* en el mismo diccionario):

ej. 24 CONTAMINACIÓN. Acción y efecto de alterar nocivamente la pureza o las condiciones normales de una cosa o un medio por agentes químicos o físicos.

No obstante, para redactar una definición morfosemántica válida no existe una fórmula única para cada relación morfosemántica, sino que hay que realizar un análisis de la unidad léxica que se va a definir y de la relación semántica que la une a la unidad léxica de partida (San Martín y L'Homme 2014).

### 3.3.7 La definición implicativa

La definición implicativa (Robinson 1950: 106–108; Sager 1990: 43) —también llamada *definición contextual* (Harris y Hutton 2009: 118)— consiste en definir una unidad léxica mediante su presentación en un contexto explicativo. Se trata, en principio, de una definición no aristotélica (ej. 25), ya que el definiéndum no se categoriza, aunque puede haber casos en los que el genus esté presente (ej. 26). En el caso de que no haya genus, la definición corre el riesgo de no ser rigurosa ni suficientemente informativa. Sin embargo, si se presenta la unidad léxica en un contexto en el que se ofrece el genus, se puede considerar una variante aceptable de la definición aristotélica como las definiciones de oración completa (§3.3.10) ej. 25 OLEODUCTO. Un oleoducto transporta petróleo y sus derivados.

ej. 26 RECICLAR. Cuando una persona recicla algo, lo procesa para que puedan volverse a utilizar.

### 3.3.8 La definición relacional

La definición relacional (Rey-Debove 1967: 155–157) —también llamada *definición formularia* (Ilson 1987: 62)— hace referencia a la relación existente entre un definiéndum calificativo y las unidades léxicas que este puede calificar. Rey-Debove distingue dos tipos: las relativas (el defíniens es una oración de relativo) (ej. 27) y las preposicionales (el defíniens es un sintagma preposicional) (ej. 28).

ej. 27 RECICLABLE. Que se puede reciclar. [DLE]

ej. 28 DEPRISA. Con celeridad, presteza o prontitud. [DLE]

Este tipo de definición solo se emplea para adjetivos y adverbios y su uso está muy extendido como alternativa a la definición aristotélica para unidades léxicas pertenecientes a estas dos categorías gramaticales.

### 3.3.9 La definición en la lexicología explicativa y combinatoria

La lexicología explicativa y combinatoria (LEC) (Mel'čuk, Clas y Polguère 1995) forma parte de la teoría sentido-texto (Mel'čuk 1981; Mel'čuk 1998), cuyo principal ejemplo práctico es el *Dictionnaire explicatif et combinatoire du français contemporain* (DEC) (Mel'čuk 1984). En terminología, los recursos DiCoInfo y DiCoEnviro (§2.2.3.4) siguen las pautas definicionales de la LEC. Al tratarse de una teoría de corte estructuralista, se concibe el significado de las palabras como independiente del mundo y eso se refleja en su concepción de la definición.

Para la LEC, la definición ha de limitarse a expresar la información denotativa, ya que la información connotativa se representa en otras partes del DEC. No obstante, ante la dificultad de distinguir ambas informaciones (lo que estriba de la frontera difusa entre semántica y pragmática que vimos en §2.1.2.1), Iordanskaja y Mel'čuk (1984: 36) admiten carecer de criterios generales al respecto.

Mientras que en la teoría sentido-texto las representaciones semánticas se presentan bajo la forma de redes en las que cada elemento semántico está enlazado mediante relaciones de predicado-argumento (Mel'čuk, Clas y Polguère 1995: 73), en el DEC se prefiere representar la información semántica mediante definiciones, de modo que puedan interpretarse fácilmente por parte del usuario.

La forma y el contenido de las definiciones según el LEC están determinadas por seis reglas (Mel'čuk, Clas y Polguère 1995):

1. La regla de la forma proposicional
2. La regla de la descomposición
3. La regla del bloque máximo
4. La regla de la estandarización
5. La regla de la sustituibilidad mutua
6. La regla del orden según la pertinencia semántica

*La regla de la forma proposicional*

De acuerdo con esta regla, si la unidad léxica que se va a definir es predicativa, esta debe presentarse en su forma proposicional. En otras palabras, su estructura argumental debe hacerse explícita en el definiéndum mediante el uso de variables que representen los argumentos (Mel'čuk, Clas y Polguère 1995: 78). Por ejemplo, el definiéndum del verbo *dar* se presentaría de esta manera: «X da Y a Z».

En el defíniens, no solo se deben emplear las variables, sino que además se constriñe su naturaleza semántica cuando esta no resulta evidente para el usuario. Por ejemplo, en la definición de una de las acepciones de *éclipser* en el DEC, tanto la variable X como la variable Y son caracterizadas como «cuerpos celestes»:

ej. 29 ÉCLIPSER // X éclipse Y = Corps céleste X cause qu'un corps céleste Y disparaît (partiellement) de la vue d'un observateur, en se plaçant entre Y et l'observateur ou en projetant une ombre sur Y. (Elnitsky y Mantha 1992: 195)

La definición en la LEC es o bien de tipo aristotélico (en lo que respecta a que se puede localizar un genus y unas differentiae) como en el ej. 29 o de tipo relacional (§3.3.8) para adjetivos y adverbios, como en el ej. 30:

ej. 30 RISQUÉ // [X] risqué pour Y = [Fait X] qui représente un risqueI pour Y. (Milićević 1999: 306)

*La regla de la descomposición*

Para evitar la circularidad en el DEC, la regla de la descomposición exige que todas las unidades léxicas que se empleen para definir otra palabra deben ser más simples semánticamente que el definiéndum. Una unidad se considera más simple que otra cuando aquella no es necesaria para la definición de esta (Mel'čuk, Clas y Polguère 1995: 80). Por ejemplo, si la definición de la palabra *pulmón* contiene la palabra *respirar*, pero en la de *respirar* no aparece *pulmón*, entonces *respirar* es más simple que *pulmón*.

No obstante, en las definiciones terminológicas el problema de la circularidad es menor que en el de las definiciones en diccionarios de la lengua general. En este sentido, como explica Béjoint (1997: 23), un diccionario de la lengua general es de carácter cerrado, porque todas las palabras en él contenidas han de definirse en el propio diccionario, mientras que un recurso terminológico es abierto porque hay palabras (pertenecientes a la lengua general) que aparecen en los defíniens que no se definen.

*La regla del bloque máximo*

De acuerdo con esta regla, si el contenido semántico de dos unidades léxicas presentes en una definición puede transmitirse con una sola unidad, debe hacerse así (Mel'čuk, Clas y Polguère 1995: 84). Por ejemplo, según esta regla en una definición terminológica no debería emplearse «energía generada a partir del viento» porque, en su lugar, es posible utilizar «energía eólica» que condensa esa misma información.

Con esta regla, se evitan los problemas de las definiciones compuestas solo de primitivos semánticos que resultan excesivamente largas (como las que propone Wierzbicka [1996]). De acuerdo con Mel'čuk et al. (1995: 84), con la regla del bloque máximo, no hay necesidad de elaborar previamente una lista de primitivos semánticos y, además, las definiciones son de una longitud óptima.

La regla del bloque máximo tiene, por tanto, por objetivo evitar la redundancia en las definiciones y reducir su longitud; sin embargo, es posible que un pequeño nivel de redundancia o el uso de más de una palabra (aunque emplear solo una sea posible) pueda ser útil para la adquisición de conocimientos por parte del usuario. Así pues, en nuestra opinión, esta regla debería estar supeditada a la optimización de la comprensión por parte del receptor de la definición. De hecho, esta regla no forma parte de las de DiCoInfo y DiCoEnviro (San Martín y L'Homme 2014).

*La regla de la estandarización*

Esta regla tiene el objetivo de evitar el uso de palabras ambiguas o sinónimos en las definiciones. Para ello, el número de acepción de cada unidad léxica debe explicitarse para que el usuario sepa a qué acepción exactamente se está haciendo referencia. Por ejemplo, *mangerI.1.a* en la definición de *appétit* :

ej. 31 APPÉTIT // appétit de X pour Y = désir de X de mangerI.1.a Y. (Mel'čuk y Robitaille 1999: 94)

La regla de estandarización también implica el uso de la misma unidad léxica para los significados que puede expresarse mediante distintas unidades léxicas. Es decir, es necesario elegir solo una unidad léxica entre un grupo de sinónimos y emplearla siempre esa sola en las definiciones. Por ejemplo, el lexicógrafo habría de elegir entre usar *persona* o *ser humano* en todas las definiciones, no debe alternarlas. Lo mismo ocurre con las expresiones que representan las relaciones existentes entre los elementos de una definición. Por ejemplo, Mel'čuk et al (1995: 88-89) proponen como ejemplo las siguientes estandarizaciones:

| | | | |
|---|---|---|---|
| *montre* | dispositif pour savoir l'heure… | → | dispositif destiné à montrer… |
| *marteau* | outil de percussion… | → | outil destiné à frapper… |
| *couteau* | instrument servant à couper… | → | instrument destiné à couper… |
| *cuillère* | ustensile qui sert à porter… | → | ustensile destiné à porter… |

**Tabla 3. Estandarizaciones definicionales propuestas por Mel'čuk et al** (1995: 88-89).

Mel'čuk et al. (1995: 90) defienden incluso que puede ser necesario no respetar de manera estricta las coocurrencias léxicas o el comportamiento sintáctico de algunas unidades léxicas para poder limitar al máximo el número de expresiones que se emplean en las definiciones.

En nuestra opinión, el empleo de un lenguaje controlado para las definiciones puede aportar coherencia a un recurso e incluso facilitar la tarea del lexicógrafo o terminólogo. Sin embargo, no compartimos la necesidad de no respetar de manera estricta las coocurrencias léxicas o el comportamiento sintáctico de algunas unidades léxicas, pues eso podría dificultar la comprensión. La definición debe ser un texto aceptable desde el punto de vista estilístico y, en el caso de las definiciones terminológicas, se deben respetar, en todo momento, las convenciones del lenguaje especializado en el que se enmarca habitualmente el definiéndum.

En lo que respecta al uso de los números de acepciones, las definiciones en EcoLexicon enlazan de manera automatizada mediante un hipervínculo a buena parte de los conceptos activados en la definición. Está previsto que en un futuro dichos hipervínculos puedan servir para desambiguar a qué concepto polisémico se hace referencia en una definición dada.

*La regla de la sustituibilidad*

Una definición adecuada debe poder sustituir su definiéndum en cualquier contexto. Como se vio en §3.2, se trata de una de las reglas más tradicionales en la elaboración de definiciones aristotélicas y tiene el objeto de comprobar si los componentes de la definición son necesarios y suficientes.

*La regla del orden según la pertinencia semántica*

Esta regla dicta que es necesario ordenar los componentes semánticos de la definición de acuerdo con su pertinencia semántica. Ello implica que aquellos componentes más pertinentes han de situarse más cerca del genus. Consideramos adecuado seguir esta regla siempre y cuando no se vea comprometida la redacción clara de la definición.

### 3.3.10 Las definiciones de oración completa

P. Hanks (1987) critica el uso de los paréntesis en los diccionarios porque hacen que la definición se vuelva demasiado complicada y, en vez de lograr mayor precisión —objetivo primordial del uso de paréntesis en las definiciones—, crea dificultades en la interpretación por parte del usuario. Algunos ejemplos que expone P. Hanks (1987: 116) son:

ej. 32 PACK. To form (snow, ice, etc) into a hard compact mass or (of snow, ice, etc) to become compacted. [CED[38]]

ej. 33 PACK. To place or arrange (articles) in (a container), such as clothes in a suitcase. [CED]

ej. 34 FUSE. To (cause (metal) to) melt in great heat. [LDOCE[39]]

Ante este problema de los diccionarios monolingües y otras deficiencias observadas, Cobuild[40] desarrolló un inventario de estrategias que especialmente conllevaban el uso de un estilo simple y directo, empleando palabras comunes para la elaboración de lo que se ha venido a llamar *definiciones de oración completa* (Sinclair 2004: 5).

[38] CED: *Collins English Dictionary* (AA. VV. 1986)

[39] LDOCE: *Longman Dictionary of Contemporary English* (AA. VV. 1978)

Las definiciones de oración completa de Cobuild tienen dos partes. Con la primera parte ya se alejan de la tradición lexicográfica, pues se presenta la palabra dentro de una estructura típica (P. Hanks 1987: 117). Por ejemplo, como indica Rundell (2006: 325), los datos extraídos de los corpus de Cobuild, la palabra inglesa *temerity* suele aparecer con el patrón: «*have the temerity+TO-infinitive*», por lo tanto, el encabezamiento de la definición de *temerity* comienza con ese patrón (ej. 35).

ej. 35 If you say that a person has the *temerity* to do something… [CEFLD[41]]

En cambio, la segunda parte de la definición se asemeja a una definición tradicional y es la que identifica el significado:

ej. 36 … you are annoyed about something they have done which you think showed a lack of respect. [CEFLD]

No obstante, esta segunda parte expone las características y usos típicos de la palabra que se define y no debe entenderse como un conjunto de condiciones necesarias o suficientes. En los diccionarios *Cobuild* se muestran los usos típicos de una palabra y no todos sus usos posibles, que pueden llegar a ser innumerables en muchos casos (Rundell 2006: 331).

Además, si se comparan las definiciones de oración completa con las definiciones tradicionales, se puede comprobar que parte de la información que suele ofrecerse en la definición tradicional (ej. 37) aparece en el encabezamiento de la definición de oración completa (ej. 38), de ahí que puedan prescindir de los paréntesis:

ej. 37 FUSE. To (cause to) stop owing to a fuse. (LDOCE *apud* P. Hanks 1987: 123)

---

[40] Cobuild es un centro de investigación inicialmente dirigido por John Sinclair cuyos resultados se han plasmado en varios diccionarios de inglés.

[41] CEFLD: *Cobuild English for Learners Dictionary* (AA. VV. 2015)

ej. 38 When a light or some other piece of electrical apparatus *fuses* or when you fuse it, it stops working because of a fault, especially because too much electricity is being used. [CED *apud* P. Hanks 1987: 125]

Como se ha podido observar, para las unidades léxicas predicativas, las definiciones de oración completa permiten expresar la naturaleza de los argumentos, aunque sin implicar que las instanciaciones de los argumentos en la definición sean los únicos posibles:

> There is a lot more linguistic information in an FSD [full-sentence definition] than in a traditional type of entry, and it does not all need to be explicit; all sorts of typical features of the cotext can be reproduced, suggested, alluded to, hinted at, without the necessity of a bald statement. At the present time this lack of precise accountability is a positive advantage. A traditional entry that attempted the same detail would have to commit itself at times when the evidence was diverse, or omit a potentially important semantic observation. So Cobuild can say with impunity "If the police arrest you…" even though there exists the legal possibility of a citizen's arrest — Cobuild does not imply that no-one else can possibly arrest you; the traditional definitions are mealy-mouthed or vague in their attempts to wriggle out of commitment here (Sinclair, 2004: 5–6).

Por lo tanto, las definiciones de oración completa suponen una forma alternativa a las de la LEC (§3.3.9), en las que se explicitan los argumentos de las unidades léxicas predicativas. Si bien las definiciones de oración completa permiten representar de una manera más clara el significado de unidades léxicas predicativas, en el caso de unidades léxicas no predicativas las ventajas son más reducidas.

## 3.4 METODOLOGÍAS DEFINICIONALES COGNITIVAS EN TERMINOLOGÍA

### 3.4.1 La metodología definicional de la terminología sociocognitiva

La idea principal que defiende Temmerman (2000) respecto a la elaboración de definiciones es que es necesario reemplazar la definición tradicional por plantillas que puedan servir en la descripción de unidades de comprensión flexibles y difusas.

Como vimos en §2.2.1.3, Temmerman (2000) sustituye la noción de *concepto* por lo que denomina *unidad de comprensión*. Una unidad de comprensión es normalmente una categoría con una estructura prototípica y, rara vez, un concepto de límites precisos (Temmerman 2000: 223–224). Partiendo de esta base, Temmerman cree que solo los conceptos «tradicionales» permiten una definición que describa las características necesarias y suficientes del concepto. En lugar de ello, la autora presenta los siguientes principios (Temmerman 2000: 74):

- La comprensión es un proceso que está en evolución continua. Explicitar los periodos de este proceso puede ser esencial para la comprensión actual de algunos términos.
- Muchas unidades de comprensión se pueden dividir en varias facetas o aspectos de su estructura intracategorial.
- Toda unidad de comprensión puede observarse desde distintos puntos de vista, lo cual viene determinado por su estructura intercategorial.
- La intención del emisor influye en los elementos que se incluyen en la explicitación del significado de una categoría.

Tomando estos principios como punto de partida, Temmerman investigó la mejor manera de definir tres tipos de unidad de comprensión: INTRON (entidad), BLOTTING (actividad) y BIOTECHNOLOGY (categoría colectiva).

En primer lugar, determinó la viabilidad de la definición intensional y extensional de estas tres unidades de comprensión, esto es, si se pueden definir mediante características necesarias y suficientes:

- Para INTRON, una definición intensional es posible, ya que tiene un hiperónimo claro (SEQUENCE IN EUKARYOTIC SPLIT GENES) y se puede elaborar una lista de las características diferenciadoras con respecto a su único cohipónimo (EXON). En cuanto a la definición extensional, esta sería posible, pero resultaría demasiado larga.

- Para BLOTTING, podrían utilizarse varios conceptos como hiperónimo. Sin embargo, con cualquiera que se seleccione, la delimitación de las características diferenciadoras con respecto a otros conceptos con el fin de construir una definición intensional sería una tarea imposible. Ello se debe a la cantidad de conceptos cohipónimos de BLOTTING y las diferentes perspectivas desde las que se puede categorizar la unidad de comprensión, lo cual revelaría infinidad de relaciones que no se pueden presentar en una estructura lógica simple. Por otro lado, la definición extensional para BLOTTING sí es viable.
- Para BIOTECHNOLOGY, resulta casi imposible encontrar un concepto superordinado. Además, por su naturaleza interdisciplinar, cubre un gran espectro de actividades y resultados, por lo que describir su intensión no es posible al no poderse definir como parte de un estructura lógica u ontológica. Por otro lado, describir la extensión tampoco es posible por el solapamiento entre los distintos enfoques desde los que se puede activar el concepto BIOTECHNOLOGY.

A continuación, estudió la estructura prototípica de las unidades de comprensión:

- INTRON no tiene una estructura prototípica; los especialistas están de acuerdo en una delimitación clara del concepto. Sin embargo, se han producido nuevos descubrimientos sobre los intrones, lo cual da indicios de que podría empezar a desarrollarse una estructura prototípica en torno a INTRON.
- BLOTTING es una categoría que sí muestra las características de una estructura prototípica. Intensionalmente presenta semejanza de familia, lo cual ha dado lugar a distintas lexicalizaciones; mientras que extensionalmente, hay varias clases de BLOTTING, las cuales presentan distintos niveles de prototipicidad.

- BIOTECHNOLOGY presenta semejanza de familia respecto a los distintos contextos en los que se puede activar este concepto en situaciones reales. Es un conjunto de sentidos que se solapan en torno al núcleo de la definición que Temmerman (2000: 93) delimitó como «biotechnology is the application of biological techniques in order to achieve commercial results». En cuanto a la extensión, esta categoría tiene límites difusos debido a la multiplicidad de perspectivas desde las que se puede utilizar.

Finalmente, estudió la estructura intracategorial e intercategorial de las dos unidades de comprensión que mostraron estructura prototípica. Dichas estructuras cambian según el modelo cognitivo idealizado (marco) en el que se conciba la unidad de comprensión así como la perspectiva adoptada. Para obtener información acerca de la estructura intracategorial e intercategorial, Temmerman (2000: 97) defiende que es necesario recurrir al uso real de las categorías. La autora no considera que las unidades de comprensión existan como unidades independientes en el mundo objetivo, sino que lo que existe son los textos en los que los autores dan testimonio de cómo comprenden ellos las categorías dentro de un modelo cognitivo idealizado, que a su vez puede ser distinto del modelo cognitivo idealizado de otro autor.

BLOTTING es un proceso que consiste en varios pasos, por lo que para comprender la categoría es necesario conocer los pasos que lo estructuran intracategorialmente. Mientras que para comprender BIOTECHNOLOGY intracategorialmente es esencial la información histórica, sus aplicaciones y los tipos de empresa que se dedican a ello, intercategorialmente, será importante la perspectiva desde la que se conciba: otra disciplina, su carácter interdisciplinar, concienciación sobre la influencia positiva o negativa de la biotecnología, crecimiento económico, relación de la biotecnología con el derecho o con la educación.

A partir de los resultados de estos análisis, la autora llega a la conclusión de que la información esencial para comprender estas categorías es enciclopédica y que esa información tiene distintos grados de esencialidad y no se puede estructurar ni lógica ni partitivamente. Por ello, Temmerman (2000: 122) propone la siguiente plantilla para la descripción de unidades de comprensión:

| CATEGORY/TERM:<br>TYPE OF CATEGORY:<br>a) entity<br>b) activity<br>c) collective category<br>d) … etc.<br>CORE DEFINITION:<br>…………………………………………………………<br>INTRACATEGORIAL INFORMATION:<br>a) is a part of<br>b) consists of parts<br>c) is a type of<br>d) has the following types<br>e) aims<br>f) use<br>g) application<br>h) … etc.<br>INTERCATEGORIAL INFORMATION :<br>a) perspective<br>b) domains<br>c) intentions<br>HISTORICAL INFORMATION:<br>………………………………………………………… |
|---|

**Tabla 4. Plantilla para la descripción de unidades de comprensión (Temmerman 2000: 122)**

La plantilla de Temmerman (Tabla 4) no es más que una propuesta en la que la autora expone una lista de posibles tipos de información que cabe representar en una definición terminológica. Como se verá en la próxima sección, la TBM utiliza también plantillas definicionales, en las cuales a su vez se basan las que empleamos en los resultados de este trabajo (§5). A diferencia de la propuesta de Temmerman, las plantillas de la TBM tienen una aplicación práctica directa y, de hecho, se emplean actualmente en la redacción de definiciones en EcoLexicon.

### 3.4.2 La metodología definicional de la terminología basada en marcos

Para la TBM, las definiciones son representaciones del conocimiento y se conciben como la traducción a lenguaje natural de la estructura conceptual de un dominio (Faber 2002). La metodología definicional de la TBM está basada en dos pilares. El primero de ellos es la extracción del conocimiento mediante un enfoque tanto *top-down* como *bottom-up*, lo cual detallaremos en el capítulo dedicado a la metodología de este trabajo (§4.2.2). El segundo pilar son las plantillas definicionales, las cuales están basadas en los marcos de W. Martin (1994; 1998)[42], el cual, a su vez, se inspiró principalmente en Minsky (1975; 1977) y, en menor medida, en Fillmore (1985). W. Martin recibió las mismas influencias que Barsalou et al. (1993), por ello, la representación que ambos hacen de los marcos es similar. La diferencia radica en que Barsalou utilizó los marcos desde un punto de vista teórico para representar constructos conceptuales de distintas dimensiones, mientras que Martin los emplea de manera más restringida para la representación de la información conceptual de una categoría con vistas a la redacción de definiciones tanto lexicográficas (1994) como terminológicas (1998).

Las plantillas están formadas por casillas (*slots*) que, al aplicarse a un concepto, reciben sus respectivas especificaciones (*fillers*), al igual que los atributos y valores que proponía Barsalou (1992). Un ejemplo de plantilla definicional propuesta por W. Martin (1994: 248) es la de ANIMAL:

| **animal** | |
|---|---|
| zoocentric slots: | |
| - subtype | |

[42] W. Martin (1994; 1998) utiliza el término *marco* para hacer referencia a lo que en este trabajo denominamos *plantilla definicional*.

| - sex | |
|---|---|
| - size | |
| - shape | |
| - skin | |
| - design | |
| - colour | |
| - age | |
| - has-qual | |
| - has-part | |
| - type-action | |
| - typ-ability | |
| - move | |
| - sound | |
| - lives-off | |
| - birth | |
| - habitat-geo | |
| - habitat-eco | |
| anthropocentric slots: | |
| - has stereotypical-qual | |
| - function-man | |
| - human activity related to this animal | |
| - a part of this animal related to human activity | |
| - similar-to | |

**Tabla 5. Plantilla definicional para *animal* (W. Martin 1994: 247–248)**

La plantilla para ANIMAL se aplicaría a todos los conceptos que formen parte de dicha categoría. Sin embargo, W. Martin (2001: 68) advierte de que la lista de casillas no refleja características suficientes y necesarias, sino que son de carácter enciclopédico, así que, además de características necesarias, explica —basándose en Cruse (1986: 16)— pueden ser características esperables y características posibles[43]. Además, a la hora de aplicar la plantilla a los conceptos que pertenecen a la categoría, no

[43] El resto de estatus posibles de una característica —que no aparecerían en una plantilla definicional— son *no esperable* y *excluida* (Cruse 1986: 16).

siempre se rellenarán todas las casillas, sino que elegirán aquellas que sean relevantes (W. Martin 1994: 247). En la plantilla de la Tabla 6 se puede ver cómo Martin aplica la plantilla de ANIMAL a RABBIT:

| **rabbit** | |
|---|---|
| zoocentric slots: | |
| - subtype | rodent |
| - has-qual | quick |
| - habitat-eco | rabbit hole / rabbit hutch |
| anthropocentric slots: | |
| - has stereotypical-qual | shy / impatient |
| - function-man | |
| - human activity related to this animal | hunting / breeding |
| - a part of this animal related to human activity | meat / fur |
| - similar-to | hare <different-from:<br>shorter ears<br>shorter hind legs> |

**Tabla 6. Plantilla definicional para *rabbit*** (W. Martin 1994: 250)

En la plantilla de Tabla 6, podemos observar que no todas las casillas de la plantilla de ANIMAL se han rellenado. Por un lado, esto se debe a que se han seleccionado solo aquellas que son relevantes para este concepto. No obstante, por otro lado, tampoco se han rellenado aquellas casillas cuya especificación se hereda de RODENT, como, por ejemplo, que es un mamífero o que tiene dientes largos (W. Martin 1994: 250). A partir de la información contenida en la plantilla ya rellena, se elabora la definición.

Cabe reseñar la existencia algunas propuestas similares a las de Martin. Entre ellas podemos destacar la de Wegner (1985), que propone plantillas solamente para conceptos de alto nivel como persona, grupo, objeto, acción, etc.; la de Strehlow (1997), que defiende el uso de este tipo de plantillas para mostrar la información definicional de un concepto; y la de Atkins y Rundell (2008), que proponen plantillas para entradas lexicográficas completas. Las plantillas de estos dos últimos autores incluyen definiciones para determinadas categorías en las que tan solo hay que rellenar los huecos con la información correspondiente, por ejemplo, para unidades léxicas de la categoría *animal*: «a [size] [wild / domesticated] [carnivorous / herbivorous] mammal, Latin name XXX, having fur / hide [colour, markings], found in [habitat]. Also called XXX» (Atkins y Rundell 2008: 125). Asimismo, como hemos visto anteriormente, Temmerman (2000) también propone un plantilla para la confección de definiciones terminológicas, aunque su propuesta incluye una plantilla única que habrá de adaptarse según los casos.

El grupo de investigación LexiCon empezó a emplear las plantillas definicionales en su proyecto OncoTerm, dedicado al dominio de la Oncología. Para la redacción de definiciones se crearon diversas plantillas, como, por ejemplo, la de TREATMENT:

<table>
<tr><th>Categoría conceptual</th><th>Relación conceptual</th></tr>
<tr><td rowspan="4">TREATMENT</td><td><i>is-a</i></td></tr>
<tr><td><i>uses-instrument</i></td></tr>
<tr><td><i>has-function</i></td></tr>
<tr><td><i>has-location</i></td></tr>
</table>

**Tabla 7. Plantilla definicional para la categoría TREATMENT (Faber 2002)**

Si esta plantilla se proyecta en un concepto perteneciente a esta categoría, como RADIATION THERAPY, se generan los valores correspondientes a las relaciones conceptuales de la plantilla básica de TREATMENT:

| **RADIATION THERAPY** | |
|---|---|
| *is-a* | TREATMENT |
| *uses-instrument* | HIGH-ENERGY RAYS |
| *has-function* | ELIMINATION OF CANCER CELLS |
| *has-location* | BODY-PART |

**Tabla 8. Plantilla definicional para el concepto RADIATION THERAPY (Faber 2002)**

La definición resultante de esta plantilla es la siguiente:

ej. 39 RADIATION THERAPY. Treatment [*is-a*] involving the use of high-energy rays [*uses-instrument*] to damage cancer cells and stop them from growing and dividing [*has-function*].

La relación conceptual correspondiente a *has-location* no se activa en este nivel de abstracción, pero sí en sus conceptos subordinados como INTRAPERITONEAL RADIATION THERAPY, cuya plantilla está basada en la de RADIATION THERAPY habiendo, pues, herencia de propiedades:

| **INTRAPERITONEAL RADIATION THERAPY** | |
|---|---|
| *is-a* | INTERNAL RADIATION THERAPY |
| *uses-instrument* | HIGH-ENERGY RAYS |
| *has-function* | ELIMINATION OF CANCER CELLS |
| *has-location* | ABDOMEN / PELVIS |

**Tabla 9. Plantilla definicional para el concepto INTRAPERITONEAL RADIATION THERAPY (Faber 2002)**

La definición resultante de esta plantilla es la siguiente:

ej. 40 INTRAPERITONEAL RADIATION THERAPY. Internal radiation therapy [*is-a*] in which radioactive material is placed directly in the pelvic and abdominal cavities [*has-location*].

En este caso, las proposiciones conceptuales correspondientes a las relaciones *uses-instrument* y *has-function* no se activan en esta definición porque el concepto superordinado INTERNAL RADIATION THERAPY ya transmite esa información.

En EcoLexicon, las plantillas definicionales toman la siguiente forma: las casillas son las distintas relaciones conceptuales que puede activar el definiéndum y al cual se le aplica la plantilla, y las especificaciones son el concepto que está unido al definiéndum mediante la relación conceptual en cuestión. De este modo, la plantilla expresa proposiciones conceptuales en las que el definiéndum es el primer concepto. A continuación, se reproduce la plantilla definicional para la categoría HARD COASTAL DEFENSE STRUCTURE (la cual se aplica a todos los conceptos que forman parte de esa categoría en EcoLexicon) y la plantilla cumplimentada para GROIN, uno de los conceptos subordinados de HARD COASTAL DEFENSE STRUCTURE.

| **HARD COASTAL DEFENSE STRUCTURE** | |
|---|---|
| *type-of* | CONSTRUCTION |
| *located-at* | SHORELINE |
| *made-of* | MATERIAL |
| *has-function* | COASTAL DEFENSE |

**Tabla 10. Plantilla definicional para la categoría HARD COASTAL DEFENSE STRUCTURE** (León Araúz, Faber y Montero Martínez 2012: 156)

| **GROIN** | |
|---|---|
| Hard coastal defense structure made of concrete, wood, steel and/or rock perpendicular to the shoreline built to protect a shore area, retard littoral drift, reduce longshore transport and prevent beach erosion. | |
| *type-of* | HARD COASTAL DEFENSE STRUCTURE |
| *located-at* | PERPENDICULAR TO SHORELINE |
| *made-of* | CONCRETE<br>WOOD<br>METAL<br>ROCK |
| *has-function* | SHORE PROTECTION<br>LITTORAL DRIFT RETARDATION<br>LONGSHORE TRANSPORT REDUCTION<br>BEACH EROSION PREVENTION |

**Tabla 11. Plantilla cumplimentada y definición de GROIN a partir de la plantilla definicional de HARD COASTAL DEFENSE STRUCTURE (León Araúz et al. 2012: 156)**

Como se puede observar, la plantilla de HARD COASTAL DEFENSE STRUCTURE explicita las relaciones que se activarán en las definiciones de sus subordinados (*located-at*, *made-of* y *has-function*) y el tipo de concepto que ocupará la especificación (SHORELINE, MATERIAL y COASTAL DEFENSE). La casilla de *type-of* está reservada al concepto superordinado cuya plantilla se ha aplicado y es la que determina el genus de la definición.

Las plantillas definicionales en EcoLexicon están íntimamente ligadas a las redes conceptuales que cada concepto desarrolla y que se representan mediante relaciones jerárquicas y no jerárquicas con otros conceptos dentro de la BCT. Mientras que las redes conceptuales representan el potencial semántico de un concepto, la plantilla definicional restringe dicho potencial de significado según la relevancia de dichas relaciones de acuerdo con factores como el contexto o el perfil del usuario. Así pues, las plantillas explicitan la estructura prototípica de la descripción conceptual de una categoría (León Araúz, Faber y Montero Martínez 2012: 150).

Por otro lado, León Araúz (2009) aplica los roles de la estructura de *qualia*, concepto perteneciente a la teoría del lexicón generativo de Pustejovsky (Pustejovsky 1995), a las plantillas definicionales. Los roles de *qualia* captan el modo en que los seres humanos entienden los objetos y las relaciones en el mundo, al tiempo que explican el comportamiento de los elementos léxicos (Pustejovsky 2006: 104). Tal y como señalan León Araúz et al. (2012: 148), cada rol de *qualia* representa un segmento del significado de una unidad léxica y, por ello, la estructura de *qualia* ofrece una forma sistemática de representar dimensiones conceptuales. Los cuatro roles son los siguientes (Pustejovsky 1995):

- Rol formal: categoría básica que distingue el objeto dentro de un dominio más amplio.
- Rol constitutivo: relación entre el objeto y las partes que lo componen.
- Rol télico: finalidad y función del objeto.
- Rol agentivo: factores involucrados en el origen del objeto.

El rol formal se refiere a los mecanismos de categorización y, por ende, en EcoLexicon se corresponde con la relación conceptual *tipo-de*. Por su parte, el rol constitutivo hace referencia a las relaciones meronímicas como, por ejemplo, *parte-de* o *compuesto-de*. En lo que respecta al rol télico, la relación conceptual principal asociada a este rol es la de *tiene-función*. Por último, el rol agentivo concierne a las relaciones como *resultado-de* o *afectado-por*.

Por otro lado, en conjunción con los roles de qualia, la teoría del lexicón generativo además ofrece una tipología básica de conceptos relacionada con los roles de *qualia* (Pustejovsky et al. 2006). Primeramente, los conceptos se dividen en entidades, eventos y propiedades y, a partir de ahí, se subdividen en tipos naturales, tipos funcionales y tipos complejos (Pustejovsky et al. 2006: 1702):

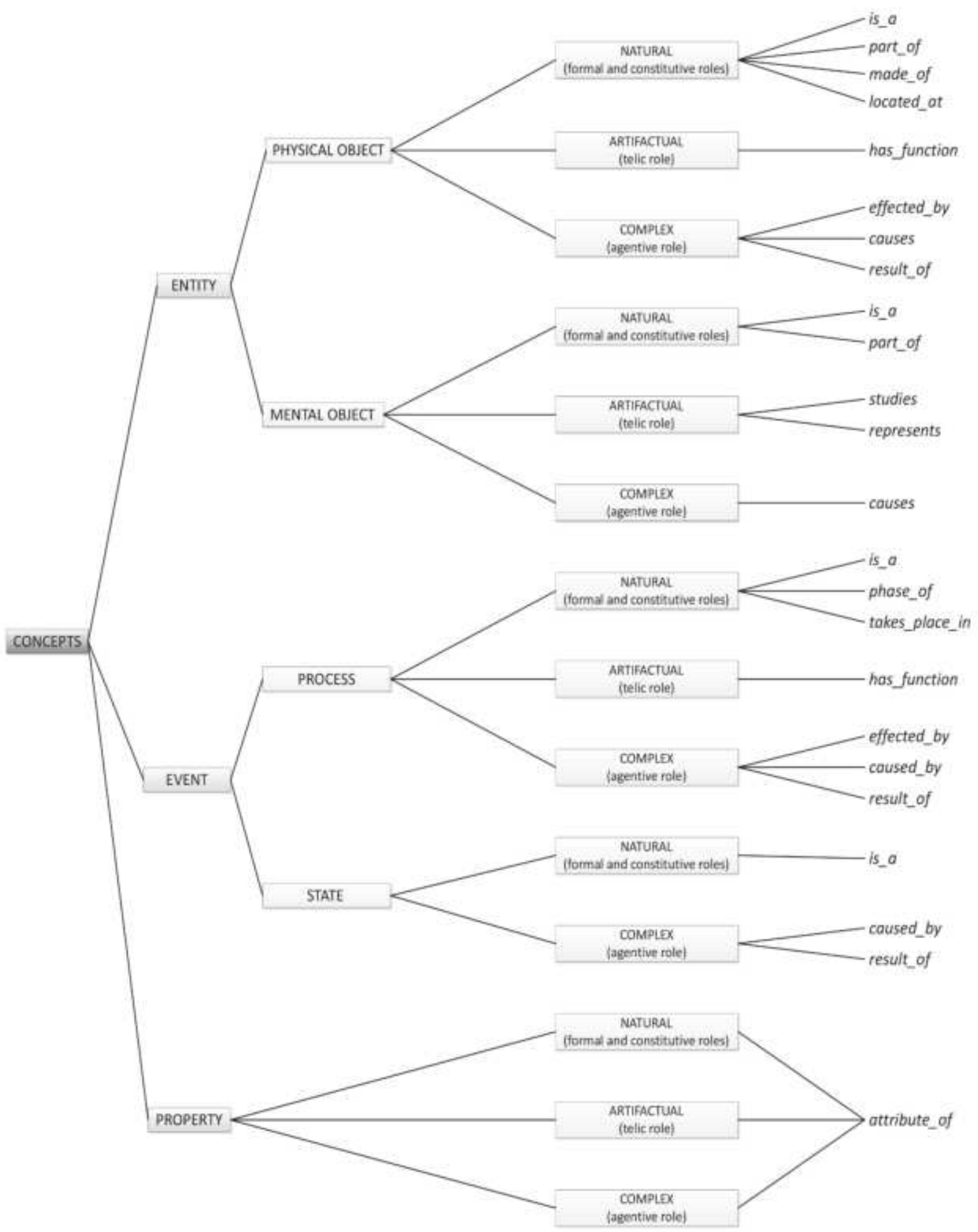

**Figura 11. Combinación de la tipología de conceptos y las relaciones conceptuales con los roles de qualia** (León Araúz, Faber y Montero Martínez 2012: 149)

- Tipos naturales: conceptos de clase natural que tan solo hacen referencia a los roles formal y constitutivo.
- Tipos artefactuales: conceptos que hacen referencia a su finalidad, función u origen.
- Tipos complejos: conceptos que integran una referencia a una relación entre ambos tipos.

León Araúz, Faber y Montero (2012: 149) —basándose en una versión inicial por Reimerink, León Araúz y Faber (2010: 19)— relacionan la tipología conceptual arriba presentada con los roles de *qualia* y las relaciones conceptuales de EcoLexicon (Figura 11).

De esta combinación, pueden extraerse cuáles son las relaciones conceptuales que deben ocupar un lugar preferencial en la descripción de conceptos en EcoLexicon y, por ende, en su definición. Por ello, León Araúz (2009) propone la integración de los roles de *qualia* en las plantillas definicionales de EcoLexicon. En las Tablas Tabla 12 y Tabla 13, podemos ver un ejemplo de la aplicación de los roles de qualia a las plantillas definicionales.

Las principales ventajas del uso de plantillas definicionales en un recurso terminológico son, por un lado, que estas dotan de consistencia y coherencia a las definiciones terminológicas, puesto que las definiciones de conceptos que pertenecen a la misma categoría recibirán definiciones con una estructura similar. Por otro lado, las plantillas definicionales pueden servir de guía al terminólogo en el proceso de extracción del conocimiento para su posterior representación en la definición terminológica.

| **SOFT COASTAL DEFENSE ACTION** | |
|---|---|
| formal role | *type-of* |
| constitutive role | *has-part* |
| agentive role | *result-of / effected-by* |
| telic role | *has-function* |

**Tabla 12. Plantilla definicional de la categoría SOFT COASTAL DEFENSE ACTION basada en los roles de qualia** (León Araúz, Faber y Montero Martínez 2012: 158)

| **BEACH NOURISHMENT** | | |
|---|---|---|
| Soft coastal defense action consisting of replenishing a beach with new dredged materials in order to protect the coastline form erosion, floods and storms. | | |
| formal role | *type-of* | SOFT COASTAL DEFENSE ACTION |
| constitutive role | *has-part* | REPLENISHING<br>DREDGING |
| agentive role | *effected-by*<br>*affects*<br>*result-of* | DREDGER<br>DREDGED MATERIALS<br>REPLENISHED BEACH |
| telic role | *has-function* | PROTECT THE COASTLINE FROM EROSION, FLOODS AND STORMS |

**Tabla 13. Plantilla definicional de la categoría SOFT COASTAL DEFENSE ACTION basada en los roles de qualia aplicada a BEACH NOURSISHMENT** (León Araúz, Faber y Montero Martínez 2012: 158)

## 3.5 LA SELECCIÓN DEL CONTENIDO DE LA DEFINICIÓN

La selección del contenido es una de las cuestiones esenciales respecto a la construcción de definiciones; sin embargo, la ausencia de pautas tanto en terminología como en lexicografía es muy destacable. Como señala Seppälä (2015: 24), mientras que existen algunas reglas sobre aspectos formales como la puntuación o el uso de las mayúsculas, no existen más que indicaciones generales sobre la selección de la información que se debe representar en una definición. Por ejemplo, estas son algunas de las reglas que exponen Cabré (1993: 210-213) y Vézina et al. (2009: 16-33):

- La definición debe tener forma afirmativa y expresar preferiblemente lo que es el concepto más que lo que no es.
- La definición solo debe contener términos conocidos por el usuario o que también estén definidos en el mismo recurso.
- La definición debe estar redactada de modo que se evite la circularidad.
- El genus debe ser de la misma categoría gramatical que el definiéndum.
- El genus no debe estar precedido por sintagmas como «especie de», «tipo de» o «clase de».[44]
- El definiéndum no debe estar contenido en la definición excepto en el caso de que se trate de un término homónimo o polisémico, o en el caso de que el definiéndum sea un término compuesto y el redactor de la definición se asegure de no caer en la tautología.

Como se puede observar, estas reglas suponen unos requisitos mínimos para la redacción de definiciones aristotélicas. Sin embargo, no aseguran que la definición resultante sea satisfactoria con respecto a su contenido. Como vimos en el apartado dedicado a la categorización (§2.1.3), la determinación de los rasgos de un concepto es una cuestión compleja. Con respecto a una definición terminológica dada, Seppälä (2009: 47-48) distingue cuatro tipos de rasgos que un terminólogo puede extraer de sus fuentes de información en relación con la definición:

- Rasgos latentes: Son aquellos rasgos del concepto que no se representarían en la definición porque son irrelevantes en el campo de conocimiento en el que se enmarca el concepto.
- Rasgos destacados: Son aquellos rasgos del concepto que podrían incluirse en una entrada de un diccionario o una ontología porque son de interés para el campo de conocimiento en el que se enmarca el concepto.
- Rasgos potencialmente relevantes: Son un subconjunto de los rasgos destacados que pueden considerarse relevantes para la definición del concepto teniendo en cuenta el campo de conocimiento en el que se enmarca.
- Rasgos relevantes: Son un subconjunto de los rasgos potencialmente relevantes que, finalmente, se consideran relevantes para la definición del concepto y, por tanto, se representan en ella.

[44] Esta pauta tiene la finalidad de evitar la definición metalingüística.

Así pues, la selección del contenido de la definición terminológica equivale a elegir aquellos rasgos que son relevantes. La relevancia de los rasgos depende de diversos factores, los cuales Seppälä (2012: 141-166; 2015) organiza en torno a tres dimensiones:

- Dimensión extensional:
    - Factor ontológico: Según este factor, la selección de rasgos se ve influenciada por el tipo de referente del término que se define.
    - Factor tipo-instancia: Según este factor, la selección de rasgos se ve influenciada por el hecho de que el referente sea un tipo o una instancia.
    - Factor extensional: Según este factor, la selección de rasgos se ve influenciada por si la extensión se compone de elementos homogéneos o heterogéneos.

- Dimensión contextual:
  - o Factor sistémico: Según este factor, la selección de rasgos se ve influenciada por el sistema conceptual.
  - o Factor individual: Según este factor, la selección de rasgos se ve influenciada por los conocimientos previos del usuario.
- Dimensión comunicativa:
  - o Factor funcional: Según este factor, la selección de rasgos se ve influenciada por las necesidades del usuario que consulta la definición.
  - o Factor comunicativo: Según este factor, la selección de rasgos se ve influenciada por la situación comunicativa.

Con la intención de adaptar dicha clasificación a las necesidades de esta investigación, hemos redefinido las dimensiones de la siguiente forma:

- Dimensión ontológica: Esta dimensión incluye todos los factores que determinan la selección de rasgos según el tipo ontológico del concepto que se define.
- Dimensión funcional: Esta dimensión incluye todos los factores que determinan la selección de rasgos según los usuarios y el recurso en el que se inserta la definición.
- Dimensión contextual: Esta dimensión incluye todos los factores que determinan la selección de rasgos según los elementos contextuales activados en la definición.

### 3.5.1 La dimensión ontológica

La dimensión ontológica incluye los factores que explican la relevancia de los rasgos en función de la categoría ontológica a la que pertenezca el concepto que se define. Las categorías ontológicas son las categorías aplicables a cualquier dominio de la realidad (Spear 2006: 28), como podría ser ENTIDAD, EVENTO y PROPIEDAD. Esta dimensión equivale a la descrita por Seppälä (2012: 142-151) bajo el nombre de *dimensión extensional*.

Sin embargo, cabe destacar que existe considerable desacuerdo respecto a cuáles son esas categorías generales que son relevantes en cualquier dominio y sus características (Hoehndorf 2010). De hecho, los filósofos llevan más de 25 siglos discutiendo cuáles son las categorías ontológicas (Sowa 1995: 669). Por ello, no es de extrañar que existan varios tipos de ontologías de alto nivel como BFO (Spear 2006), SUMO (Niles y Pease 2001), DOLCE (Gangemi et al. 2002), etc., que organizan dichas categorías de manera distinta. Esta ausencia de una única clasificación de categorías ontológicas independientes de la lengua y del dominio se explica por el efecto de la corporeización y de otros factores que mediatizan la percepción del ser humano de la realidad. Por lo tanto, ya que no hay un conjunto único de categorías ontológicas, la selección de uno de ellos será una cuestión de conveniencia, como expone Westerhoff (2005: 218).

Seppälä (2012; 2015) emprendió la formalización de un modelo de selección de rasgos en definiciones basado en categorías ontológicas. Para su estudio, Seppälä utilizó la versión 1.0 de BFO a la que añadió algunas categorías más (2015: 37). En total, elaboró treinta y siete modelos relacionales, uno para cada categoría.

El formato de los modelos relacionales es similar a las plantillas de W. Martin (1994; 1998) y, por ende, a las plantillas utilizadas en EcoLexicon. Asimismo, al igual que en EcoLexicon (§3.4.2), las plantillas están organizadas de manera jerárquica, lo que implica que se produce herencia de configuraciones relacionales (lo que en EcoLexicon se denominan *proposiciones*):

| **OBJECT** | |
|---|---|
| Relational configurations characterizing the entity type OBJECT | |
| *has_part* | OBJECT |
| *participates_in* | PROCESS |
| Relational configurations inherited from the entity type INDEPENDENT CONTINUANT | |
| *bearer_of* | QUALITY |
| *bearer_of* | REALIZABLE ENTITY |
| *located_at* | TEMPORAL REGION |
| *located_in* | SITE |
| *participates_in* | PROCESSUAL ENTITY |

**Tabla 14. Modelo relacional para la categoría** OBJECT (Seppälä 2015: 38)

Así pues, Seppälä (2012; 2015) verificó si los modelos relacionales de BFO pueden predecir qué tipo de rasgos son relevantes en la definición según el tipo de categoría ontológica. Para ello, la autora anotó un corpus de 240 definiciones terminológicas en francés de 15 dominios distintos. La anotación consistió en señalar dentro de cada definición qué porciones corresponden con el genus y cuáles con las differentiae. Asimismo, cada differentia se etiquetó según la configuración relacional correspondiente en los modelos relacionales de BFO, incluida una etiqueta para los casos dudosos. Asimismo, cada definición se asignó a una de las categorías de BFO para poder comparar las categorías BFO con los resultados del análisis. De las 37 categorías ontológicas incluidas en el estudio, aparecieron representadas 16 en el corpus.

Una vez anotadas todas las definiciones, Seppälä realizó un análisis para determinar cuáles son las proposiciones más habituales para definir las categorías en el corpus y las comparó con los modelos relacionales. El resultado total fue que cerca del 75% de las configuraciones relacionales expresadas en las definiciones analizadas estaban presentes en los modelos de BFO correspondientes (Seppälä 2015: 42). El resultado, por tanto, indica que existe una correlación importante entre la categoría ontológica a la que pertenece el concepto que se define y el tipo de rasgo que resulta relevante.

A partir de sus resultados, Seppälä (2015: 42) defiende que es posible crear modelos basados en una ontología de nivel superior —refinados con los resultados obtenidos a través del análisis de definiciones existentes— que sirva de guía para la selección de los rasgos que se deben activar en una definición de acuerdo con la categoría ontológica del concepto que se define.

Cabe destacar que dentro de esta misma dimensión de la selección de rasgos, Seppälä (2012; 2015) incluye otros dos factores más. Por un lado, la autora habla de lo que ella denomina el *factor tipo-instancia*. Según este factor, es posible que la relevancia de determinados rasgos se vea afectada por el hecho de que el referente sea un tipo o una instancia; es decir, si el referente es un grupo de referentes o un referente único (Seppälä 2012: 148). A este respecto, Seppälä (2012: 147) hipotetiza que en el caso de que el referente del definiéndum sea una instancia es probable que la localización espacial y temporal tengan mayor relevancia que si se trata de un tipo.

Por otro lado, bajo el nombre de *factor extensional*, Seppälä (2012: 150) también afirma que la relevancia de determinados rasgos puede que dependa de la homogeneidad de la extensión del definiéndum, entendiendo principalmente homogeneidad como similitud física. De este modo, cabría esperar que cuanto más homogénea sea la categoría, más rasgos relevantes de tipo perceptual habrá que serán frecuentemente típicos. Por el contrario, en el caso de categorías heterogéneas, los rasgos relevantes tenderán a ser necesarios y de tipo causal o funcional.

En conclusión, el estudio de Seppälä (2012; 2015) demuestra que la categoría ontológica es un factor importante en la selección de rasgos de la definición. Sin embargo, los modelos relacionales creados por la autora están limitados a categorías de alto nivel, lo cual limita su aplicación a dominios especializados concretos, pues los factores funcionales y contextuales ejercen una fuerte influencia en la selección de rasgos. La autora es consciente de estas limitaciones y recuerda que los modelos relacionales que propone deben complementarse con otro tipo de marcos descriptivos conceptuales o léxico-semánticos (Seppälä 2015: 46).

### 3.5.2 La dimensión funcional

La teoría funcional de la lexicografía (TFL) (Bergenholtz y Tarp 1995; Bergenholtz y Tarp 2003; Tarp 2008) está basada en el supuesto de que cualquier recurso lexicográfico[45] debe crearse con el objetivo de satisfacer ciertos tipos de necesidades sociales (Tarp 2008: 43). Dicho supuesto está inspirado en la siguiente recomendación de Householder: «Dictionaries should be designed with a special set of users in mind and for their specific needs» (1967: 279). Esta recomendación se puede extender a las definiciones y afirmar que estas deben diseñarse con un conjunto específico de usuarios en mente y con vistas a sus necesidades concretas. De este modo, dado que compartimos con la teoría funcional de la lexicografía la importancia que otorgamos a esta cuestión, describiremos la dimensión funcional de la selección de rasgos en las definiciones terminológicas en el marco de esta teoría. La dimensión funcional tal y como la describimos en este trabajo incluye tanto los factores que Seppälä ha incluido en la dimensión comunicativa como el factor individual que forma parte de su caracterización de la dimensión contextual.

Así pues, de acuerdo con la TFL, antes de la creación de cualquier recurso, es importante analizar las situaciones de los usuarios potenciales, sus cualificaciones y sus necesidades, así como de qué manera el recurso puede asistir al usuario a este respecto (Tarp 2008: 43).

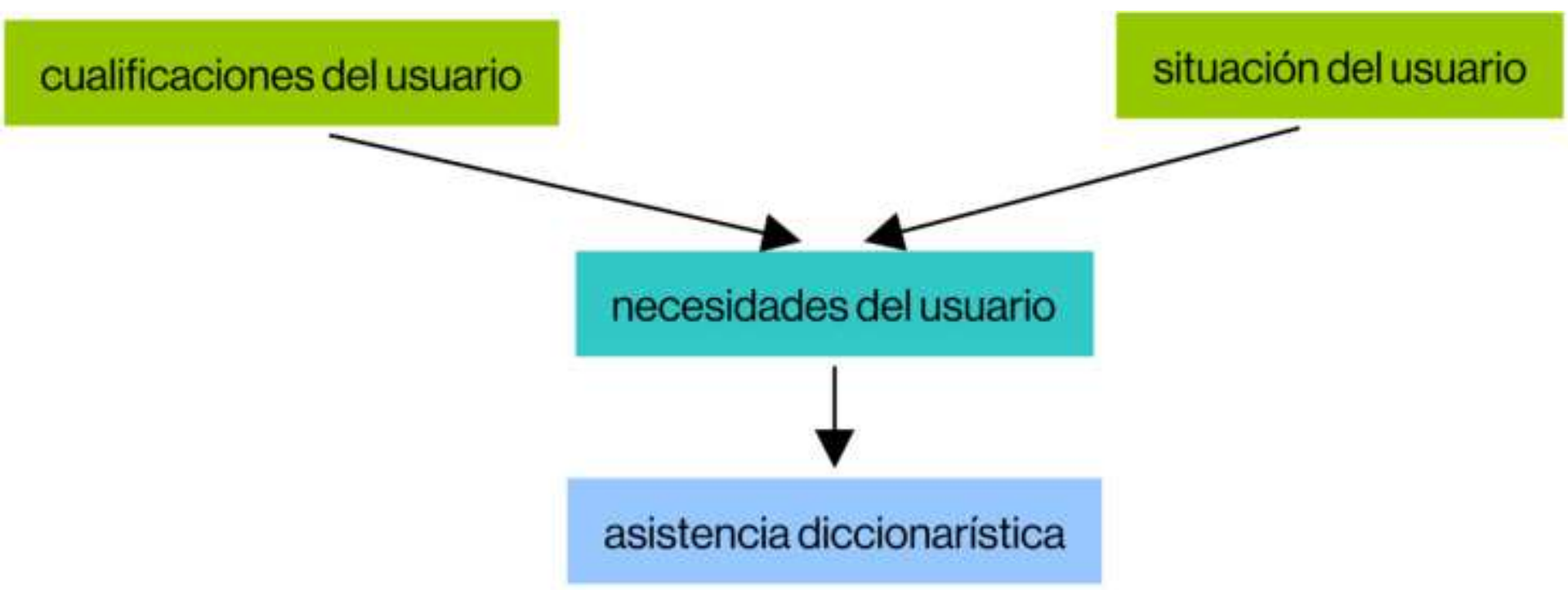

**Figura 12. La relación entre las cualificaciones del usuario, la situación del usuario, las necesidades y la asistencia diccionarística en la TFL**

[45] De acuerdo con la TFL, los recursos terminográficos son un tipo de recurso lexicográfico. Por ello, para seguir la terminología utilizada por los autores de la teoría, en esta sección utilizaremos el término *lexicográfico* y sus derivados para referirnos tanto a la lexicografía como a la terminografía.

Las situaciones de los usuarios, sus cualificaciones y sus necesidades son de naturaleza extralexicográfica, es decir, no están necesariamente relacionados con la lexicografía. Como puede observarse en la Figura 12, las cualificaciones de los usuarios potenciales y la situación extralexicográfica determinan las necesidades de los usuarios potenciales. Si estas necesidades pueden satisfacerse mediante la consulta de un recurso lexicográfico, entonces se consideran relevantes desde un punto de vista lexicográfico (Tarp 2008: 43-44). El cuarto elemento de la teoría es la asistencia diccionarística, que consiste en la información lexicográfica a partir de la cual los usuarios pueden extraer la información que necesitan (Tarp 2008: 58).

La noción central de la teoría es la de «función lexicográfica» que está basada en los cuatro elementos arriba mencionados. Una función lexicográfica se define como la satisfacción de los tipos específicos de necesidad relevantes desde un punto de vista lexicográfico que pueden surgir en un tipo específico de usuario potencial en un tipo específico de situación extralexicográfica (Tarp 2008: 81). En todo recurso lexicográfico, tanto el contenido como la forma deben concebirse tomando en consideración las funciones lexicográficas (Bergenholtz y Tarp 2003: 177).

Esta teoría es relevante para este estudio porque, cuando los lexicógrafos no tienen en cuenta las funciones lexicográficas a la hora de escribir definiciones, los recursos en los que estas se insertan no pueden satisfacer las necesidades de sus usuarios de forma óptima (Kwary 2011: 56). Por tanto, antes de escribir definiciones, es importante determinar las funciones lexicográficas del recurso en el que estas van a insertarse, de qué modo pueden asistir en el cumplimiento de dichas funciones y la relación de las definiciones con los otros elementos que conforman el recurso en lo que respecta al desempeño de esas funciones.

La dimensión funcional de la definición queda fuera del alcance de este trabajo. No obstante, es menester determinar las características funcionales de las que partiremos a la hora de tratar la dimensión contextual, puesto que al igual que las otras dos dimensiones, afecta directamente a la selección de rasgos. Tomaremos, pues, como punto de partida las características funcionales de las definiciones que forman parte de EcoLexicon. Con el fin de describirlas, a continuación revisamos los cuatro elementos que componen la TFL en relación con EcoLexicon y sus definiciones.

### 3.5.2.1 Las cualificaciones de los usuarios

Además de por la situación del usuario, que se tratará más adelante, las necesidades de los usuarios potenciales estarán en parte determinadas por sus conocimientos lingüísticos y enciclopédicos, ya que un usuario decidirá consultar un recurso lexicográfico para obtener o verificar algún tipo de conocimiento. Asimismo, en el diseño y desarrollo de un recurso lexicográfico, también es necesario tener en consideración el conocimiento que el usuario tiene sobre cómo emplear un recurso lexicográfico. Por este motivo, Tarp (2008: 55) divide las cualificaciones del usuario en lingüísticas y enciclopédicas por un lado, y en lexicográficas por el otro.

#### *3.5.2.1.1 Las cualificaciones lingüísticas y enciclopédicas*

Para definir las cualificaciones lingüísticas y enciclopédicas de los usuarios potenciales, Tarp (2008: 55) propone la siguiente lista de preguntas:

- ¿Cuál es la lengua materna de los usuarios?
- ¿Qué dominio tienen de su lengua materna?
- ¿Qué dominio tienen de una lengua extranjera específica?
- ¿Qué dominio tienen de una lengua de especialidad específica en su lengua materna?
- ¿Qué dominio tienen de una lengua de especialidad específica en una lengua extranjera?
- ¿Qué experiencia tienen en traducción?
- ¿Cuán grande es su cultura general?
- ¿Cuán grande es su conocimiento de la cultura de un área concreta donde se habla una lengua extranjera?

- ¿Cuántos conocimientos poseen acerca de un tema o una ciencia específicas?

En el caso de EcoLexicon, la lengua materna de los usuarios potenciales no es relevante. EcoLexicon está dirigido a usuarios que simplemente dominan al menos el inglés o el español, ya que la interfaz del usuario y ciertos módulos solo se ofrecen en estas dos lenguas. Asimismo, si entienden inglés o español, los usuarios que dominen el griego, alemán, ruso, francés o neerlandés pueden también beneficiarse de EcoLexicon, ya que se proporcionan términos equivalentes en estas lenguas.

Respecto al dominio de una determinada lengua de especialidad y un tema, la distinción entre legos, semiexpertos y expertos se torna relevante. Dado que nació en el seno del Departamento de Traducción e Interpretación de la Universidad de Granada, EcoLexicon se concibió como una BCT dirigida principalmente a traductores. Por tanto, EcoLexicon está dirigido principalmente a legos o a semiexpertos, ya que los traductores pueden ser categorizados como uno u otro dependiendo de su especialización y su experiencia.

Al definir los usuarios potenciales de EcoLexicon como legos y semiexpertos, los redactores científicos y técnicos y los estudiantes de cualquier disciplina perteneciente a los estudios medioambientales pueden incluirse también en el grupo de usuarios potenciales. Otro grupo potencial de un recurso terminológico medioambiental dirigido a legos y semiexpertos sería el de empleados del sector público y privado que trabajan a diario con información medioambiental, asesores en ciencia y tecnología para políticos, periodistas e incluso los propios políticos y líderes de opinión (Bergenholtz y Kaufmann 1997: 101).

Los expertos en cualquiera de las subdisciplinas de los estudios medioambientales tienen un estatus doble con respecto a EcoLexicon. Por un lado, dado que el medio ambiente es un vasto dominio multidisciplinar, es muy improbable que una persona pueda ser considerada como experto en todos y cada uno de los subdominios. Por ello, incluso un experto en un subdominio sería considerado semiexperto para una gran parte de la cobertura de EcoLexicon. Por el otro lado, aunque no es probable que las necesidades de conocimiento especializado de un experto las pueda satisfacer un recurso terminológico, el experto puede que no domine el inglés o español medioambiental y, por lo tanto, EcoLexicon puede asistirle a este respecto (Bergenholtz y Kaufmann 1997: 102).

En cuanto al conocimiento de cultura general, se espera que todos los usuarios potenciales de EcoLexicon sean usuarios con cierta formación general, así que dicha información no se proporciona. Sin embargo, no se asume ningún conocimiento cultural específico de un área geográfica, así que ese tipo de información se explicita en EcoLexicon cuando es necesario.

En lo que concierne a las definiciones, la lista de preguntas para determinar las cualificaciones lingüísticas y enciclopédicas propuestas por Tarp (2008: 55) puede adaptarse y reducirse a las siguientes cinco preguntas:

- ¿Qué dominio tienen de la lengua en la que estarán escritas las definiciones?[46]
- ¿Qué dominio tienen del lenguaje de especialidad específico en el que estarán escritas las definiciones?
- ¿Cuán grande es su cultura general?
- ¿Cuán grande es su conocimiento de la cultura en el área concreta relevante donde se habla una lengua extranjera?
- ¿Cuántos conocimientos poseen acerca del tema o ciencia específicas en el que se enmarcarán las definiciones?

La primera pregunta no incluye referencia al estatus de la lengua como materna o extranjera porque la única cuestión importante es el dominio que el usuario tenga de esta. Por otro lado, las definiciones en EcoLexicon no se presentan en ningún tipo de inglés o español simplificado; por lo

[46] Normalmente, la lengua del definiéndum y la del défíniens será la misma. Sin embargo, si este no es el caso, es la lengua del défíniens la que resultaría relevante.

tanto, los usuarios potenciales han de tener un buen dominio de inglés o español.

En cuanto a las otras preguntas, las respuestas son las mismas que para EcoLexicon como recurso en conjunto. En otras palabras, los usuarios potenciales de las definiciones en EcoLexicon son legos formados o semiexpertos y no se espera que tengan conocimientos culturales previos específicos de un área geográfica. Los expertos también se benefician de las definiciones de conceptos de su propio dominio pero, como veremos más adelante, solo para apoyar sus consultas lingüísticas.

#### *3.5.2.1.2 Las cualificaciones lexicográficas*

Las cualificaciones lexicográficas de los usuarios potenciales están basadas en el conocimiento sobre cómo utilizar un recurso lexicográfico. Para caracterizarlas, Tarp (2008: 56) propone las siguientes preguntas:

- ¿Cuánto saben los usuarios de Lexicografía?
- ¿Qué experiencia general tienen en el uso de diccionarios?
- ¿Qué experiencia específica tienen en el uso de un diccionario específico?

Estas cuestiones son de especial relevancia para el diseño de recursos lexicográficos y su organización. Cualquier deficiencia que posean los usuarios potenciales a este respecto puede ser compensada mediante la adición de información sobre cómo utilizar el recurso.

Las definiciones son uno de los componentes más básicos de cualquier entrada lexicográfica y han sobrevivido durante siglos de práctica lexicográfica a lo largo una variedad de culturas lexicográficas (Lew 2010: 292). En la educación obligatoria, normalmente, los estudiantes aprenden cómo consultar un diccionario. Por ende, una definición estándar (un definiéndum seguido de su defíniens) no requeriría ninguna cualificación especial por parte del usuario. No obstante, las cualificaciones lexicográficas de los usuarios potenciales se tornan relevantes si la definición se desvía de algún modo de esta norma, como es el caso de la propuesta presentada en este trabajo. En este caso, se debe proporcionar a los usuarios la información necesaria para que saquen el mayor provecho de las definiciones mediante, por ejemplo, una guía de uso o un vídeo tutorial.

### 3.5.2.2 Las situaciones de los usuarios

Originalmente, la TFL solo concebía dos situaciones posibles: la cognitiva y la comunicativa (Tarp 2010: 195). Más adelante en el desarrollo de la teoría, las situaciones operativa e interpretativa se incluyeron a la tipología[47]. Así pues, de acuerdo con Bergenholtz y Bothma (2011: 61-62), los cuatro tipos de situaciones en las que los usuarios pueden requerir la asistencia de un recurso lexicográfico son la cognitiva, la comunicativa, la operativa y la interpretativa.

#### *3.5.2.2.1 Las situaciones cognitivas*

Ambos tipos principales de situaciones del usuario (cognitiva y comunicativa) crean necesidades de adquisición de conocimiento. La distinción entre ellos radica en si hay una intención de usar el conocimiento obtenido en un evento comunicativo presente o planificado (situación comunicativa) o el usuario quiere simplemente almacenar ese conocimiento en su cerebro (situación cognitiva) (Bergenholtz y Bothma 2011: 61). Las necesidades cognitivas no deben confundirse con las necesidades de conocimiento enciclopédico, ni las necesidades comunicativas con necesidades de conocimiento lingüístico (Tarp 2008: 87). Por ejemplo, querer conocer el significado de *efflorescence* o el plural de *altostratum* podrían ser consideradas necesidades cognitivas o comunicativas según las situaciones en que surjan tales necesidades.

[47] Recientemente, se ha propuesto añadir a la lista las situaciones evaluativas. Una situación evaluativa se da cuando una persona ha de evaluar los conocimientos de otra, como, por ejemplo, en un examen (Rodríguez Gallardo 2013: 87). Sin embargo, hemos decidido excluirla porque aún no se considera parte integrante de la TFL.

En resumen, la principal diferencia entre una situación cognitiva y una comunicativa es que en la situación cognitiva no hay una intención comunicativa implicada:

> In cognitive situations the potential user has a need for knowledge of some kind. The purpose is not to use it in the concrete situation, e.g. during text reception or in a specific situation to act or react in that situation, but simply to get the knowledge. The user wants to know or to have the sought-after knowledge and stores it in the brain for later use. (Bergenholtz y Bothma 2011: 61)

No obstante, cualquier conocimiento adquirido en una situación cognitiva, puede ser usado más adelante en una situación comunicativa. La TFL admite este hecho, aunque lo considera irrelevante:

> The fact that information obtained by a user to fulfil his/her needs in a specific situation can also be used in another situation, is not relevant for the current concrete situation. (Bergenholtz y Bothma 2011: 62)

Siguiendo las definiciones dadas por Bergenholtz y Bothma (2011), un usuario potencial en una situación cognitiva sería, por ejemplo, un estudiante de geología que quiere saber lo que es la tomografía de resistividad eléctrica (situación cognitiva esporádica) o un legislador que desea entender la erosión costera y sus implicaciones en el desarrollo costero (situación cognitiva sistemática). Por otro lado, como se ha expuesto antes, de acuerdo con esta teoría, las situaciones cognitivas también pueden dar lugar a una necesidad de información lingüística, por ejemplo, un hidrólogo español que está aprendiendo inglés por motivos profesionales puede querer aprender la terminología en inglés relacionada con su dominio. En este caso, no hay un evento comunicativo concreto planificado, pero la razón por la que cualquier persona querría obtener algún tipo de información lingüística sería para usarla en un evento comunicativo futuro, ya sea determinado o indeterminado.

Tarp (2008: 45) describe ocho situaciones como principales ejemplos de situaciones cognitivas:

1. al leer: el deseo súbito de saber algo más acerca de una cuestión dada;
2. al escribir: la necesidad de saber más sobre un tema dado para finalizar un texto;
3. durante una conversación o discusión sobre algún tema específico;
4. durante procesos en el subconsciente: el deseo súbito de examinar algo;
5. durante la consulta de un diccionario: el deseo de saber más sobre un tema específico;
6. en relación con tareas de traducción e interpretación especializadas: la preparación para dichas tareas incluye aprender sobre el área de conocimiento en cuestión;
7. en relación con un programa de estudios: adquiriendo gradualmente conocimiento sobre un área de conocimiento específica;
8. en relación con un curso: la necesidad de saber más sobre un tema específico, por ejemplo.

Las ocho situaciones arriba expuestas demuestran que, a menudo, la distinción entre situación cognitiva y comunicativa puede ser difícil de establecer. En particular, cabe cuestionarse si los ejemplos 2, 3 y 6 no serían más bien situaciones comunicativas.

Aunque distinguir entre situaciones cognitivas y comunicativas no sea una tarea fácil, en nuestra opinión, sí resulta útil determinar las posibles situaciones extralexicográficas que pueden conducir a la consulta de un recurso lexicográfico sin necesidad de distinguir entre cognitivas y comunicativas.

EcoLexicon ha sido principalmente diseñado para satisfacer las necesidades de los usuarios potenciales en situaciones cognitivas que requieren la adquisición de información medioambiental o información lingüística relacionada con el inglés o el español especializado medioambiental (López Rodríguez, Buendía y García Aragón 2012: 62). De manera secundaria, EcoLexicon también asiste en situaciones cognitivas relacionadas con la necesidad de conocimiento relacionada con el griego, alemán, ruso, francés o neerlandés especializado medioambiental.

Por su parte, las definiciones en EcoLexicon asisten principalmente en la adquisición de información medioambiental en situaciones cognitivas. No obstante, la asistencia que ofrecen las definiciones en cuanto a la adquisición de información lingüística sobre el lenguaje especializado del medio ambiente en situaciones cognitivas también es importante, porque la terminología no puede divorciarse del significado que representa.

#### *3.5.2.2.2 Las situaciones comunicativas*

En situaciones comunicativas, el usuario potencial tiene una necesidad de información que le puede ser de asistencia en un evento comunicativo presente o planificado. Los siguientes son las situaciones comunicativas más relevantes para la Lexicografía (Tarp 2008: 53):

- producción textual en lengua materna;
- recepción textual en lengua materna;
- producción textual en lengua extranjera;
- recepción textual en lengua extranjera;
- traducción de lengua materna a lengua extranjera;
- traducción de lengua extranjera a lengua materna;
- traducción de lengua extranjera a otra lengua extranjera;
- revisión o calificación de un texto producido en lengua materna;
- revisión o calificación de un texto producido en lengua extranjera;
- revisión o calificación de un texto traducido de lengua materna a lengua extranjera;

- revisión o calificación de un texto traducido de lengua extranjera a lengua materna;
- revisión o calificación de un texto traducido de lengua extranjera a otra lengua extranjera.

En cuanto a EcoLexicon, se ha desarrollado principalmente para asistir a usuarios en las siguientes situaciones comunicativas (López Rodríguez, Buendía y García Aragón 2012: 62):

- la recepción de textos medioambientales en español;
- la producción de textos medioambientales en español;
- la recepción de textos medioambientales en inglés;
- la producción de textos medioambientales en inglés;
- la traducción de textos medioambientales en español hacia el inglés;
- la traducción de textos medioambientales en inglés hacia el español.

De forma secundaria, EcoLexicon también asiste en la recepción, producción y traducción de textos medioambientales en alemán, ruso, griego, francés y neerlandés.

En situaciones comunicativas, las definiciones en EcoLexicon, al igual que en el caso de situaciones cognitivas, asisten principalmente en la adquisición de información medioambiental y, en menor medida, a la adquisición de información lingüística sobre el lenguaje especializado del medio ambiente.

#### *3.5.2.2.3 Las situaciones operativas e interpretativas*

En situaciones operativas, el usuario potencial necesita pautas o instrucciones relativas a operaciones físicas o mentales (Bergenholtz y Bothma 2011: 62). Un usuario potencial en una situación operativa sería, por ejemplo, alguien que quiere saber cómo se utiliza un barómetro o cómo se determina la clorinidad de una muestra de agua del mar. Las situaciones operativas pueden ser vistas como un tipo de situación cognitiva, ya que el usuario quiere adquirir conocimiento independientemente de un evento comunicativo.

En las situaciones interpretativas, el usuario potencial necesita interpretar un signo no lingüístico de algún tipo (Bergenholtz y Bothma 2011: 62). Un usuario potencial en una situación interpretativa sería alguien que quiere entender una señal de tráfico o las banderas de color que indican si es seguro el baño o no en la playa. Se podría argüir que las situaciones interpretativas son un tipo especial de situación comunicativa, ya que, durante la interpretación de una señal no verbal, hay comunicación.

Las situaciones interpretativas y operativas se han estudiado muy escasamente en relación con la Lexicografía y muy pocas herramientas lexicográficas se han creado para ser usada en ese tipo de situaciones (Bothma y Tarp 2013: 91). En situaciones operativas, un usuario potencial puede obtener asistencia de un manual o una guía de uso (Tarp 2007: 177) y en situaciones interpretativas, por ejemplo, en material informativo procedente de autoridades y diferentes organizaciones (Bergenholtz y Bothma 2011: 62).

Dado que las herramientas lexicográficas no tienden a cubrir necesidades surgidas en situaciones operativas o interpretativas, las definiciones no están diseñadas para asistir en ninguna de las necesidades relacionadas. Aunque poco práctico, no sería imposible. Por otro lado, una herramienta lexicográfica que pretenda proporcionar asistencia en situaciones operativas podría incluir, en sus definiciones de conceptos con un componente funcional, instrucciones de cómo utilizar instrumentos o cómo llevar a cabo actividades, aunque esto probablemente excedería la longitud habitual de una definición. Por otro lado, el definiéndum de las definiciones en un diccionario interpretativo podría ser una imagen. Sin embargo, para asistir en estas dos situaciones, podrían crearse campos específicos para incluir esa información, sin que ello sustituyera la definición.

### 3.5.2.3 Las necesidades de los usuarios

Las necesidades de los usuarios equivalen a la información que requieren estos para resolver sus problemas específicos (Tarp 2004: 28). En relación con los recursos lexicográficos, las necesidades de los usuarios relevantes son aquellas que pueden satisfacerse por medio de la consulta de un diccionario. Por ejemplo, los problemas concernientes a la estructura argumental o la puntuación de un texto no son problemas que normalmente pueda resolver un diccionario (Tarp 2008: 56).

Las necesidades de los usuarios se han clasificado en las relativas a la función y las relativas al uso (Tarp 2008: 57). Las necesidades relativas a la función (o primarias) son aquellas necesidades de información requerida para resolver problemas u obtener conocimientos. Las necesidades relativas al uso (o secundarias) son aquellas necesidades relacionadas con el uso de los diccionarios o en relación con un recurso lexicográfico concreto. Dado que las necesidades relacionadas con el uso incluyen las necesidades de información que ayuda a los usuarios a encontrar y confirmar la información que precisan (Tarp 2008: 57), ciertos tipos de necesidades que estarían relacionadas con la función serían necesidades relacionadas con el uso dependiendo de la situación. Por ejemplo, como veremos más adelante, una necesidad de significado puede ser tanto una necesidad relacionada con la función como una necesidad relacionada con el uso.

Tarp (2008: 57) ofrece la siguiente lista de necesidades relativas a la función que pueden ser o cognitivas o comunicativas:

- información sobre lengua materna;
- información sobre lengua extranjera;
- información sobre lenguaje especializado en lengua materna;
- información sobre lenguaje especializado en lengua extranjera;

- información comparativa sobre lengua materna y lengua extranjera;
- información comparativa sobre lenguaje especializado en lengua materna y lengua extranjera;
- información de cultura general;
- información sobre cultura en un área lingüística específica;
- información sobre un tema o una ciencia específicas;
- información comparativa sobre un tema en cultura nacional y extranjera.

De esta lista, las definiciones pueden prestar asistencia en las siguientes necesidades:

- información sobre lenguaje especializado en lengua materna;
- información sobre lenguaje especializado en lengua extranjera;
- información de cultura general;
- información sobre cultura en un área lingüística específica;
- información sobre un tema o una ciencia específicas.

A este nivel de generalización, cualquiera de estas necesidades puede aparecer tanto en situaciones cognitivas como comunicativas. Por ende, en lo relativo a las definiciones y las necesidades del usuario, las cualificaciones lingüísticas y enciclopédicas de los usuarios potenciales son más importantes que la situación del usuario.

Las definiciones en EcoLexicon se centran en proporcionar información sobre un tema o ciencia específicos (más concretamente, el medio ambiente y sus subdominios). La información sobre lenguajes de especialidad no es directamente parte de la asistencia que proporcionan las definiciones en EcoLexicon porque las definiciones describen conceptos y no términos, aunque los términos están enlazados a los conceptos que designan. La información cultural se proporciona solo cuando se considera necesario debido a la naturaleza del concepto que se esté definiendo.

Por su parte, las necesidades relacionadas con el uso, que son aquellas que surgen en el momento en el que el usuario está obteniendo asistencia de un recurso lexicográfico (Tarp 2008: 57). Se las considera de tipo secundario porque se originan a partir de las de tipo primario. Las necesidades relacionadas con el uso pueden incluir las información o la instrucción necesarias para consultar un recurso lexicográfico, para lo cual, una definición no proporciona asistencia. Sin embargo, una necesidad secundaria también puede ser de tipo lingüístico o enciclopédico cuando se trata de confirmar o encontrar la información asociada a la necesidad primaria. Por ejemplo, una definición puede satisfacer una necesidad secundaria cuando aporta el significado de un término que aparece en la definición que el usuario consultó originalmente.

#### 3.5.2.4 La asistencia diccionarística

El cuarto elemento de la teoría es la asistencia diccionarística, que consiste en la información lexicográfica a partir de la cual los usuarios pueden extraer la información que necesitan (Tarp 2008: 58). Así pues, en el caso de las definiciones, la asistencia diccionarística será el contenido conceptual representado en la definición al que los usuarios pueden acceder para satisfacer sus necesidades.

#### 3.5.2.5 Las funciones lexicográficas y la necesidad de significado

Una vez que se han determinado los usuarios potenciales y sus necesidades lexicográficamente relevantes (en diferentes situaciones que puedan llevar a una consulta lexicográfica), es posible establecer las funciones lexicográficas del recurso (Bergenholtz y Tarp 2003: 176).

Tomando en consideración que los recursos lexicográficos pueden tener funciones con diferente nivel de importancia (Nielsen 2011: 212), las funciones principales de EcoLexicon (aunque no las únicas) serían:

- asistir a legos, semiexpertos y expertos que dominan el inglés y el español en la traducción de textos medioambientales de español a inglés;
- asistir a legos, semiexpertos y expertos que dominan el inglés y el español en la traducción de textos medioambientales de inglés a español;
- asistir a legos y semiexpertos que dominan el inglés en la recepción y producción de textos medioambientales en inglés;
- asistir a legos y semiexpertos que dominan el español en la recepción y producción de textos medioambientales en español;
- asistir a expertos que dominan el inglés como lengua extranjera en la recepción y producción de textos medioambientales en inglés;
- asistir a expertos que dominan el español como lengua extranjera en la recepción y producción de textos medioambientales en español;
- asistir a legos y semiexpertos que dominan el inglés o el español en la adquisición de conocimiento medioambiental especializado.

En EcoLexicon, las definiciones cumplen las mismas funciones que las listadas arribas para toda la BCT, pues puede haber una necesidad de consultar un significado en cualquier situación que implique recepción textual, producción textual y traducción, así como, naturalmente, en la adquisición de conocimiento especializado. Sin embargo, la asistencia diferirá según la situación.

De acuerdo con Tarp (2008: 70), la necesidad de significado en la recepción textual procede de la necesidad de entender lo que significa una palabra y es, por lo tanto, una necesidad relativa a la función. Sin embargo, en la producción textual sería una necesidad potencial relativa al uso, ya que el usuario puede querer consultar el significado de una palabra para asegurarse de que está consultando la entrada adecuada (Tarp 2008: 72).

En cuanto a la traducción, lo anteriormente tratado acerca de la recepción y la producción textuales es también aplicable a la recepción y producción textuales durante la traducción. Por su parte, en lo que respecta a la fase de transferencia, hay una necesidad potencial de significado sobre las diferencias semánticas de los posibles equivalentes de traducción (Tarp 2004: 32).

#### 3.5.2.6 La importancia de la dimensión funcional para la definición terminológica

Partiendo de lo expuesto sobre la TFL arriba, proponemos la división de la dimensión funcional de la selección de rasgos de la definición terminológica en dos vertientes interrelacionadas. La primera a la que llamaremos *factores relativos al usuario* y la segunda, *factores relativos al recurso*.

##### *3.5.2.6.1 Los factores relativos al usuario*

Los factores relativos al usuario son aquellos que afectan a la selección de rasgos de acuerdo con las funciones asignadas a la definición. Como hemos visto anteriormente, las funciones de la definición se determinan con arreglo a las necesidades del usuario que se quieren satisfacer, las cuales dependen, a su vez, de las cualificaciones de los usuarios y de la situación en la que surgen dichas necesidades.

En esta línea, dentro del marco del proyecto OncoTerm del grupo de investigación LexiCon, se elaboró una propuesta de definición variable que se adecuara al perfil del usuario (Jiménez Hurtado y Seibel 2004a; Seibel y Jiménez Hurtado 2004; Jiménez Hurtado y Seibel 2004b; Seibel 2002; Jiménez Hurtado y Seibel 2005). La propuesta de estas autoras hace mucho más hincapié en la adaptación del tipo de lenguaje utilizado (incluyendo la variación denominativa), que en los cambios en la selección del contenido de la definición, aunque también los contemplan:

> [L]a definición solo se puede considerar *variable* si se adapta al conocimiento previo o compartido entre emisor y receptor, esto es, si se entiende el receptor como un usuario con un perfil bien determinado,

> de modo que tanto la estructura de la definición como su contenido semántico se basen en las estructuras de percepción cognitiva del mencionado receptor. (Seibel y Jiménez Hurtado 2004: 108)

| **PELVISCOPIA** | |
|---|---|
| *tipo-de* | Procedimiento diagnóstico invasivo |
| *lugar* | realizado en una parte del cuerpo |
| *instrumento* | por medio de un endoscopio flexible o rígido |
| *función* | para la detección de un tumor. |
| *proceso* | Se pueden obtener muestras de tejido para una biopsia. |

**Tabla 15. Definición de PELVISCOPIA dirigida a un profesional de la salud (Jiménez Hurtado y Seibel 2004b: 125)**

| **PELVISCOPIA** | |
|---|---|
| *tipo-de* | Método de exploración |
| *lugar* | que se realiza en una parte del cuerpo |
| *instrumento* | con un aparato (endoscopio) flexible o rígido |
| *función* | para ver si hay un cáncer. |
| *proceso* | Se pueden tomar muestras para analizarlas en el microscopio. |

**Tabla 16. Definición de PELVISCOPIA dirigida a un paciente** (Jiménez Hurtado y Seibel 2004b: 124)

Para sistematizar la elaboración de definiciones variables de acuerdo con el usuario, las autoras proponen la creación de un lenguaje controlado[48] basado en lenguaje real extraído de corpus textuales discriminando los distintos tipos textuales (Jiménez Hurtado y Seibel 2005). Para adaptar la definición al usuario, en todas las definiciones se sigue el mismo esquema definicional, pero simplemente se cambia el tipo de lenguaje con el que se define, tanto recurriendo a estructuras sintácticas distintas para representar las relaciones conceptuales como a la variación denominativa, tal y como ocurre en textos reales que tienen distintos receptores. En las Tablas Tabla 15 y Tabla 16, se puede observar la diferencia entre la definición para un profesional de la salud del concepto PELVISCOPIA y la definición dirigida a un paciente.

[48] De acuerdo con Huijsen (1998: 23), un lenguaje controlado es «an explicitly defined restriction of a natural language that specifies constraints on lexicon, grammar, and style. The overall aim of this restriction is the reduction in ambiguity, redundancy, size, and complexity».

Por su parte, Nielsen (2011), desde la TFL, también propone crear definiciones múltiples en recursos lexicográficos para satisfacer las necesidades de los distintos tipos de usuario. No obstante, Nielsen va aún más allá y propone también definiciones múltiples que satisfagan funciones concretas:

> Experts, semi-experts and laypersons need different types of definition and different dictionary functions are best supported by meaning explanations that focus on different elements. [...] [O]nline dictionaries can cope with different definitions of the same concept depending on user type, i.e. user-related definitions. This capability may be extended to function related definitions (Nielsen 2011: 215).

Con respecto a los conocimientos previos del usuario (lo que en la terminología de la TFL se denomina *cualificaciones de los usuarios*), Seppälä (2012: 156-157) plantea la hipótesis, la cual compartimos, de que cuanto menores sean los conocimientos que el usuario tiene del sistema conceptual en el que se inserta el concepto que se define, mayor será la cantidad de rasgos relevantes. Al mismo tiempo, cuanto mayores sean sus conocimientos, menores serán los rasgos relevantes.

#### *3.5.2.6.2 Los factores relativos al recurso*

Los factores relativos al recurso en el que se inserta la definición se agrupan en dos clases. El primer grupo incluye los factores que intervienen en la selección de rasgos que surgen de la relación entre las características del recurso en el que se inserta la definición y las funciones asignadas a la definición dentro de dicho recurso. En este primer grupo, pueden englobarse todos los factores relativos al formato del recurso, al lugar que ocupa la definición dentro del recurso y cómo puede el usuario interactuar con la definición.

Si el recurso es en papel, la selección de rasgos puede verse afectada, por ejemplo, por la limitación de espacio. Asimismo, si se trata de un recurso electrónico, puede influir la posibilidad de incluir hiperenlaces o textos flotantes (*tooltips*) en la definición o que el terminólogo tenga permitido crear más de una definición para el mismo concepto.

El segundo grupo de factores que pueden afectar la selección de rasgos son los que se refieren a la relación entre la definición y otros elementos que transmiten información conceptual en el mismo recurso. De acuerdo con Lew (2010), los medios a partir de los cuales se puede transmitir significado en un recurso léxico-terminológico se pueden dividir entre aquellos de base verbal y aquellos de base no verbal[49]. Entre los medios de base verbal[50], se encuentra la definición, el equivalente en otra lengua y el ejemplo[51]. Mientras que entre los no verbales se encuentran las grabaciones de sonidos no lingüísticos, ilustraciones, fotografías, animaciones, vídeos y gráficos[52].

Por lo general, cuando un recurso contiene definiciones, estas suelen ser el medio primordial de transmisión de contenido conceptual y el resto de elementos que puedan cumplir esa función suelen ser auxiliares y estar subordinados a esta. Por ejemplo, si un recurso incluye definiciones e imágenes, la imagen suele apoyar la definición y no al contrario. En cualquier caso, el terminólogo puede decidir construir una definición independiente o hacer que esta necesite ir acompañada de algún otro elemento para satisfacer las necesidades del usuario. En el primer caso, que es el más habitual, la definición no se vería afectada por los otros elementos del recurso, mientras que en el segundo, la selección de rasgos sí

[49] Originalmente, Lew (2010) divide los medios entre verbales y no verbales, pero hemos considerado más adecuado denominarlos *de base verbal* y *de base no verbal*; porque en aquellos de base no verbal en ocasiones se utilizan también medios lingüísticos.

[50] A los medios de base verbal presentados por Lew (2010), podríamos añadir, entre otros, la nota enciclopédica, los sinónimos, los antónimos, los hiperónimos, etc.

[51] Especialmente, en el ámbito de la Terminología, destacan los estudios sobre «knowledge-rich contexts» (Meyer 2001) o contextos de uso que transmiten información conceptual que puede ser potencialmente relevante para un usuario.

[52] De acuerdo con la clasificación de Lew (2010), dentro de los gráficos se incluirían las representaciones gráficas de jerarquías o redes conceptuales.

se verá influida por la relación de la definición con los otros elementos relevantes.

En EcoLexicon, las definiciones han de ser coherentes con el resto de los módulos de EcoLexicon. Especialmente, el genus de las definiciones debe corresponderse con un concepto codificado como superordinado del definiéndum. Asimismo, siempre que resulte posible, la información que se representa en la definición debe estar también codificada en las proposiciones conceptuales que se muestran en los mapas conceptuales.

### 3.5.3 La dimensión contextual

La noción de contexto desempeña un papel crucial en diversas disciplinas como la psicología cognitiva, la inteligencia artificial, el procesamiento del lenguaje natural, el análisis del discurso, la semántica, la pragmática, la sociolingüística, etc. y cada una de ellas emplea la noción de formas diferentes. Relevantes a la terminología, podemos diferencias tres sentidos principales:

1. el contexto como las palabras que rodean una unidad lingüística en cuestión, también llamado *co-texto* (Lyons 1995: 271);
2. el contexto como ejemplo de uso que se añade en la entrada de un recurso lexicográfico o terminológico para mostrar una unidad lingüística usada un co-texto real (L'Homme 2004: 41);
3. el contexto, en su sentido más general, entendido como cualquier factor que afecta la interpretación real de un signo o una expresión (Kecskes 2014: 128), lo cual incluye también el co-texto.

Al hablar de la dimensión contextual de la definición terminológica, aludimos al contexto en la tercera acepción o, más específicamente, a los efectos que este tiene sobre la construcción del significado y, de manera indirecta, sobre el contenido de la definición terminológica.

Como ya se ha visto, en la definición no se describe el potencial semántico, pero tampoco se describen significados. Se describe un estadio intermedio que es una abstracción de los significados que tendría la unidad léxica en un contexto dado. Como ya se ha indicado, esa abstracción de significados que es el objeto de la definición recibe el nombre de *presignificado*. Sin contexto, no hay objeto de la definición, por lo cual, el contexto es una noción clave para la definición. De hecho, cuando se le presenta a una persona una unidad léxica sin un contexto explícito, esta crea un contexto prototípico a partir de eventos de uso pasados (Coulson 2001: 25).

Durante mucho tiempo en lingüística se había omitido el estudio del contexto porque se consideraba que era demasiado caótico e idiosincrático como para ser caracterizado de manera sistemática (Ervin-Tripp 1996: 35). De hecho, el contexto puede llegar a entenderse de una manera tan amplia que lo abarque absolutamente todo:

> Aside from the surrounding deictic coordinates, aside from the immediate linguistic co-text and accompanying gestural expressions at closer view, the following determinants can influence the attribution of sense: the entire frame of interaction, the individual biographies of the participants, the physical environment, the social embedding, the cultural and historical background, and—in addition to all these— facts and dates no matter how far removed in dimensions of time and space. Roughly speaking, 'context' can be the whole world in relation to an utterance act. (Pinkal 1985: 36) (Traducción de Quasthoff [1994: 733])

En la misma línea, W. Hanks (1996) describe el contexto como una sombra escurridiza por la dificultad de caracterizarlo dada su amplitud e inherente complejidad:

> What is context? Everything and nothing. Like a shadow, it flees from those who try to flee from it, evading the levels and categories of theory, and pursues those who try to flee from it, insinuating itself as the unnoticed ground upon which even the most explicit statements depend. If you are persuaded by the phenomenological concept of incompleteness, then context is inexhaustible. The more you try to specify it, the more blank spots you project, all in need of filling it. Ultimately, context is nothing less than the human world in which language use takes place and in relation to which language structure is organized. (W. Hanks 1996: 140)

Dada la inabarcable amplitud que puede adquirir la noción de contexto, Van Dijk (2008: 19) sugiere que el contexto debe definirse de manera más o menos amplia —esto es, con distinto nivel de generalidad y de granularidad— dependiendo del objetivo. Nuestro objetivo es el de caracterizar el contexto con respecto a la definición terminológica. Dado que el presignificado no es más que una abstracción de los significados de una unidad léxica a partir de determinadas restricciones contextuales, podemos afirmar que el contexto asociado a un presignificado también es una suerte de abstracción a partir de contextos reales. Por ello, llamaremos *precontexto* a las restricciones contextuales que se aplican a los presignificados en oposición a los contextos de uso concretos que determinan los significados. Antes de proceder a caracterizar los elementos que componen el precontexto de la definición terminológica, primero realizaremos la distinción entre contexto conceptual y contexto situacional.

#### 3.5.3.1 El contexto conceptual, el contexto situacional y el precontexto

En este trabajo, concebimos que la noción de contexto tiene dos vertientes principales. La primera es la noción de contexto como fenómeno mental (Ungerer y Schmid 1996: 47), el cual podemos llamar *contexto conceptual*, en la cual el contexto es una representación conceptual en forma de marcos que se utilizan para la interpretación de la realidad y, por extensión, de los enunciados lingüísticos. La segunda vertiente es la que a menudo se denomina la *situación* o *contexto situacional* que podría definirse como el estado de las cosas en el mundo real (Ungerer y Schmid 1996: 48). El contexto conceptual se genera a partir de aquellos elementos que cada interlocutor considera relevantes del contexto situacional. Así pues, el contexto situacional está compuesto por hechos objetivos mientras que el contexto conceptual está formado por interpretaciones intersubjetivas (Auer 2009: 94). El contexto situacional no afecta directamente el lenguaje y la cognición, solo lo hace mediante el contexto conceptual (Van Dijk 2008: 16).

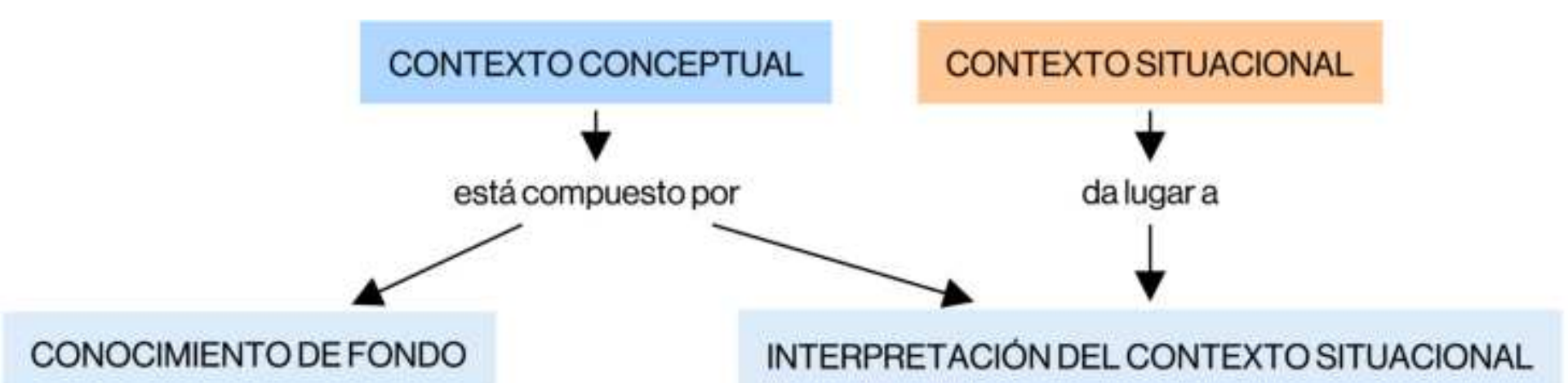

**Figura 13. Contexto conceptual y contexto situacional**

A su vez, como parte del contexto conceptual, podemos distinguir dos componentes interrelacionados: por un lado, la interpretación del contexto situacional per se y, por otro, el conocimiento de fondo que se activa para poder interpretar tanto la situación como el contenido de la comunicación (Figura 13). Por ejemplo, imaginemos un hidrólogo que lee la oración «La fracturación hidráulica contamina acuíferos» en un artículo de investigación. La parte de la interpretación del contexto situacional comprendería la representación mental acerca de los elementos de la situación que ayudan al hidrólogo a interpretar el texto (quién es el autor, el resto del texto que acompaña a la oración dentro del artículo, las características del lenguaje usado por el autor, el tema de que trata el artículo, el punto de vista adoptado por el autor, etc.). En lo que respecta al conocimiento de fondo, se trata de los marcos —incluidos los espacios mentales (Fauconnier 1994; 1997) (§2.1.2.1)— que activan las unidades léxicas contenidas en la oración a partir de su contextualización y que son necesarios para su comprensión. Los interlocutores asumen que buena parte de dicho conocimiento de fondo es compartido (Leech 1983: 13) y, como ya hemos visto, se puede considerar que, no solo da lugar al significado junto con la interpretación del contexto situacional, sino que forma parte integrante del significado.

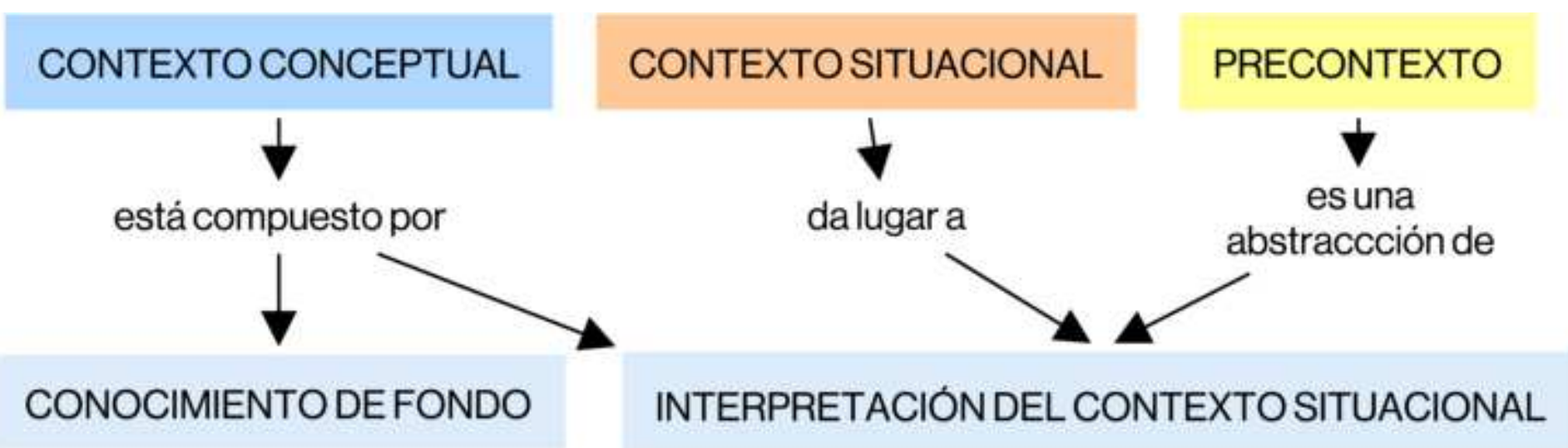

**Figura 14. Contexto conceptual, contexto situacional y precontexto**

El precontexto de la definición terminológica es una abstracción de la interpretación del contexto situacional (Figura 14). En este trabajo, puesto que nuestro enfoque es aplicado, entendemos el precontexto como un conjunto de restricciones contextuales que permiten limitar el potencial semántico de una unidad léxica de manera más o menos predecible dando lugar al presignificado.

Así pues, a continuación revisamos qué restricciones contextuales pueden formar parte del precontexto aplicado a la definición terminológica. Para ello, tomaremos como punto de partida los elementos más relevantes que forman parte del contexto situacional.

### 3.5.3.2 Los factores contextuales

La labor de caracterizar los factores contextuales que afectan a la interpretación de las unidades léxicas en eventos de uso reales es compleja, ya que, además de lo inabarcable que puede llegar a ser la tarea, los distintos factores están interrelacionados y es difícil delimitarlos de manera precisa. Son numerosos los autores que en las últimas décadas, de manera directa o indirecta, han estudiado el efecto del contexto en el significado (Clark 1996; Sperber y Wilson 1996; Coulson 2001; Carston 2002; Croft y Cruse 2004; Récanati 2004; Langacker 2008; Evans 2009; Asher 2011; Kecskes 2014; *inter alia*). Hemos clasificado los elementos más importantes reseñados por estos y otros autores en las categorías que describimos a continuación.

#### *3.5.3.2.1 El contexto lingüístico*

El contexto lingüístico o co-texto consiste, como ya se indicó, en las palabras que acompañan a la unidad léxica en cuestión en un evento de uso. Su importancia reside en que las unidades léxicas ven especificada su concepción al combinarse con otras palabras. Por ejemplo, si el término *pesticida* se combina con *tóxico* en un evento de uso concreto, el potencial semántico de *pesticida* se reduce y, a falta del resto de contexto, cobran importancia las características relativas a la toxicidad del pesticida y se activan marcos como el de la CONTAMINACIÓN O SALUD HUMANA.

Las palabras que están inmediatamente antes y después de la unidad léxica en cuestión reciben el nombre de *contexto local* (Dash 2008: 23). Un estudio de Choeuka y Lusignan (1985) hecho en lengua francesa descubrió que las personas son capaces de distinguir con una fiabilidad del 90% los sentidos de palabras polisémicas con una ventana de dos palabras de contexto delante y detrás de la unidad léxica en cuestión. Sin embargo, el contexto local no es suficiente para caracterizar el significado de una unidad léxica, aunque su influencia es destacable.

Otros contextos más amplios son el contexto enunciativo (Evans 2009b: 230), que se refiere al resto de palabras que componen el enunciado donde se encuentra la unidad léxica en cuestión. Este entorno, mayor que el del contexto local, permite reconocer las relaciones sintácticas con el resto de componentes (Dash 2008: 26), lo cual provee de mayor información sobre la modulación del significado de la unidad léxica. Asimismo, es posible considerar como contexto elementos lingüísticos de mayor tamaño; de hecho, no hay un límite determinado respecto a lo que abarca el contexto lingüístico en lo que se refiere a la longitud (Langacker 2008: 465).

Hay dos tipos principales de restricciones contextuales de tipo lingüístico que se pueden aplicar a la definición terminológica. La primera es muy común en los recursos terminológicos y se da cuando el definiéndum es una unidad terminológica compuesta, que puede concebirse como una manera de contextualizar lingüísticamente las unidades simples que lo componen. El otro tipo de restricción contextual, que se vio al tratar la regla de la forma proposicional de la metodología definicional de la LEC (§3.3.9), consiste en mostrar el definiéndum en su estructura argumental.

##### *3.5.3.2.2 El contexto discursivo*

Croft y Cruse (2004: 102) exponen que el tipo de discurso en el que se activa la unidad léxica afecta la construcción del significado. Para describir las características del discurso, unas de las nociones más empleadas son las

de campo, modo y tenor (Gregory y Carroll 1978), que Cabré (1999: 46) adapta al lenguaje especializado y convierte en cinco categorías[53]:

- El canal que se usa para transmitir la información: oral, escrito y todas las modalidades mixtas e híbridas.
- El propósito comunicativo (o tenor funcional): informar, evaluar, influir, argumentar, etc.
- El grado de formalidad entre los interlocutores: formal o informal.
- El tema de que se trata la comunicación.
- El nivel de abstracción con el que se transmite la información: de especializado a divulgativo.

Estos factores discursivos se utilizan en el estudio de la variación denominativa, aunque cada uno de ellos tienen distinto nivel de influencia (Freixa 2006: 56-57). Lo mismo ocurre con la variación conceptual en eventos de uso reales, pues cada uno de los factores tiene un nivel de influencia diferente.

Aunque sería necesario llevar a cabo estudios que lo demostraran, es de esperar que, si bien el canal, el propósito comunicativo y el grado de formalidad del discurso en determinados eventos de uso pueden afectar a la interpretación de una unidad terminológica, cada uno de ellos aislados no parece que puedan permitir predecir la interpretación; por ejemplo, no es probable que sea posible determinar de qué manera regular tiende a cambiar la conceptualización del término *solifluxión* cuando se utiliza en una comunicación oral frente a una escrita, cuando el emisor tiene un propósito comunicativo como argumentar o evaluar, o cuando el nivel de formalidad es más o menos alto. Por lo tanto, el precontexto de una definición terminológica no contendrá restricciones asociadas a estos factores.

[53] Cabré (1999: 46) originalmente propone cuatro categorías, pues incluye en la misma categoría el nivel de formalidad y el nivel de abstracción;.

No obstante, la situación es diferente respecto al tema y al nivel de especialización. Por un lado, dependiendo del tema del discurso —también denominado *contexto temático* (Miller y Leacock 2000; Dash 2008)— una unidad léxica tenderá a activar distinto contenido conceptual sin que se considere polisemia (retomaremos la cuestión de la polisemia y la variación contextual en §3.5.3.5.1). Por ejemplo, en el caso del término *agua* y su concepto asociado AGUA (en su sentido principal como sustancia cuya fórmula química es $H_2O$), varía su conceptualización si se activa en un evento comunicativo que tenga como tema la geología, pues tenderá a conceptualizarse como un agente en el proceso de erosión, mientras que si el tema es el tratamiento de aguas residuales, el concepto AGUA, siendo el mismo concepto, se activará como el paciente de distintos procesos de tratamiento (León Araúz y Faber 2010: 16). Por lo tanto, el contexto temático sí permite realizar una abstracción de los cambios en la conceptualización de una unidad terminológica.

El contexto temático cobra una relevancia particular en el ámbito de la terminología, dado que los términos son inseparables del área temática o dominio en el que se activen. Como expone Cabré (1999: 135), «[l]os términos son siempre temáticamente específicos, de forma que no hay término sin ámbito que lo acoja». Así pues, este factor sí se incluye en el precontexto de la definición terminológica. De hecho, en §3.5.3.4 lo presentaremos en mayor profundidad, pues se trata del foco principal de esta investigación.

En cuanto al nivel de abstracción (o nivel de especialización), este está íntimamente ligado al tema, pues puede definirse como la profundidad con la que este se trata. El nivel de abstracción dependerá principalmente de los conocimientos que tiene el receptor sobre el tema (Cabré 2000c: 29), dado que el emisor adaptará su mensaje para que el receptor pueda interpretarlo. Ello no solo conlleva el uso de variantes terminológicas más fáciles de comprender para el receptor, sino que también varía el contenido conceptual que se activa, que suele ser más impreciso (Fernández Silva 2011: 71).

Cabe señalar que aquí estamos haciendo referencia a los conocimientos previos de un receptor en un evento de uso concreto, que no es el mismo receptor que el de la definición terminológica. En el apartado dedicado a la dimensión funcional de la definición (§3.5.2), tratamos la cuestión de la adaptación a los conocimientos previos del usuario, que es esencial para la definición terminológica. Sin embargo, los conocimientos previos del receptor en un evento de uso —lo cual, como hemos señalado, determina el nivel de especialización del discurso— y el modo en que esto causa variación en el significado de una unidad léxica concreta se verá anulado en el marco de la definición terminológica por la necesidad de adaptación al receptor de la definición. Por ejemplo, aunque el término *solifluxión* se utilice en discursos de nivel especializado alto dirigido, por tanto, a especialistas, si la definición terminológica va dirigida a un público lego el contenido de la definición se adaptará a este público independientemente del destinatario habitual del discurso en el que suela activarse ese término. Así pues, el nivel de especialización no forma parte del precontexto de la definición terminológica.

#### *3.5.3.2.3 El contexto sociocultural*

El contexto sociocultural incluye la actividad social en la que se desarrolla la comunicación, las características de los participantes en la comunicación (clase social, sexo, edad, cultura, ideología, etc.) y relación entre los participantes (incluidas las relaciones de poder) (Croft y Cruse 2004: 103; Van Dijk 2008: 172-173; Auer 2009: 93). Este contexto se solapa en gran medida con el contexto discursivo, ya que la configuración del discurso se ve influida por los participantes y la situación social.

Por consiguiente, dejando de lado aquellos elementos que ya han sido cubiertos en el contexto discursivo y aquellos que es probable que no tengan gran influencia en la construcción del significado de las unidades terminológicas, los dos componentes del contexto sociocultural más relevantes para la definición terminológica son la cultura y la ideología.

Estas dos nociones son altamente complejas y controvertidas (Blommaert 2006: 510; Van Dijk 2009: 156) y conviene no confundirlas:

> It is important not to confuse (cultural) communities and (ideological) groups. The same community may have different ideological groups, which may be ideologically different, but share many of the cultural dimensions (language, norms, values, etc.) of their community. Thus, pacifists, feminists or socialists are not "cultural" communities, but ideological groups within cultural communities – so that Japanese feminists may be quite different feminists from those in the USA, for instance. (Van Dijk 2009: 158)

La dimensión cultural e ideológica de la terminología ha recibido la atención de diversos autores (Gambier 1991; Boulanger 1991; Wußler 1997; Lara 1999; Diki-Kidiri 2000; Gaudin 2003; Temmerman 2006; Faber 2009; Faber y León Araúz 2014, *inter alia*) que han destacado la importancia de los factores culturales e ideológicos no solo en la variación denominativa, sino también en la conceptual.

En lo que respecta a la dimensión cultural, Faber y León Araúz (2014) abogan por la necesidad de incluir en los recursos terminológicos el componente cultural en la representación conceptual y, entre los numerosos ejemplos que exponen, ofrecen el del concepto EROSION, que dependiendo del contexto cultural verá alterado su significado:

> Perception of EROSION can also be culture-bound since ice-produced erosion (and its related concepts) will be more salient or prototypical in language-cultures in Arctic regions. All of this context-modulated information should be available for potential activation when the user wishes to acquire knowledge about it. (Faber y León Araúz 2014: 140)

En cuanto a la ideología, a pesar de que varios autores han destacado su importancia en la terminología, no se han realizado estudios que hayan profundizado en la cuestión (Fernández Silva 2011: 68). No obstante, los efectos de la ideología en la conceptualización son evidentes. Un caso representativo de cómo la ideología subyacente a un evento de uso puede cambiar el significado de una unidad terminológica es el término *calentamiento global* que no activará el rasgo «provocado por la actividad humana» (o incluso activará el rasgo «no provocado por la actividad humana») si se emplea en un contexto de negacionismo del cambio climático.

El contexto cultural y el ideológico están muy relacionados con el contexto temático, que hemos visto en el apartado anterior. La frontera, de hecho, puede ser difusa, ya que dentro de una misma área temática pueden coexistir distintos puntos de vista que pueden considerarse o no debidos a diferencias culturales e ideológicas:

> En una mateixa disciplina poden conviure escoles de pensament diferents (i a vegades obertament enfrontades), corrents organitzats, treballs de recerca que es duen en paral·lel i que no són necessàriament coincidents, o punts de vista diferents sobre un mateix objecte. (Tebé 2005: 17)

En definitiva, tanto cultura como ideología son componentes que pueden formar parte del precontexto de la definición terminológica ya que son dos factores contextuales cuyo efecto sobre la conceptualización puede ser predicho y representado en una definición terminológica.

#### *3.5.3.2.4 El contexto espaciotemporal*

Como su propio nombre indica, este contexto tiene una vertiente espacial y temporal. La vertiente espacial hace referencia al lugar en el que se desarrolla la comunicación e incluye particularmente aquello que los participantes pueden percibir en su entorno inmediato (Croft y Cruse 2004: 103). En eventos de uso reales, especialmente en la comunicación oral, es muy importante porque permite la interpretación de elementos deícticos (Auer 2009: 92). Sin embargo, en lo que respecta a nuestros fines, no es probable que sea posible realizar una abstracción de los contextos espaciales posibles y encontrar patrones que determinen cómo varía el significado de las unidades léxicas según el entorno físico en el que se produce el evento de uso en el que se activan. Por lo tanto, la vertiente espacial no forma parte del precontexto para la definición terminológica.

En cuanto al contexto temporal, este se refiere a cuándo se produce la comunicación. Mientras que en eventos de uso reales el día de la semana o la hora pueden ser muy relevantes en la interpretación de enunciados así como en la interpretación de elementos deícticos, el aspecto del contexto temporal que puede resultar importante para la definición terminológica es la diacronía. Como ya se ha visto, varias corrientes teóricas terminológicas como la socioterminología (§2.2.1.1) y la TSC (§2.2.1.3) han incluido explícitamente el estudio de los términos en su evolución histórica. Dado el dinamismo del conocimiento especializado, el potencial semántico de las unidades terminológicas se va alterando con el paso del tiempo. Por tanto, el precontexto de la definición terminológica puede incluir restricciones contextuales diacrónicas.

### 3.5.3.3 El precontexto en la definición terminológica

Tras el análisis realizado en el apartado anterior de los principales factores contextuales en la construcción del significado, hemos concluido que el precontexto de la definición terminológica puede presentar las siguientes restricciones (Figura 15):

- Restricciones lingüísticas
- Restricciones temáticas
- Restricciones culturales
- Restricciones ideológicas
- Restricciones diacrónicas

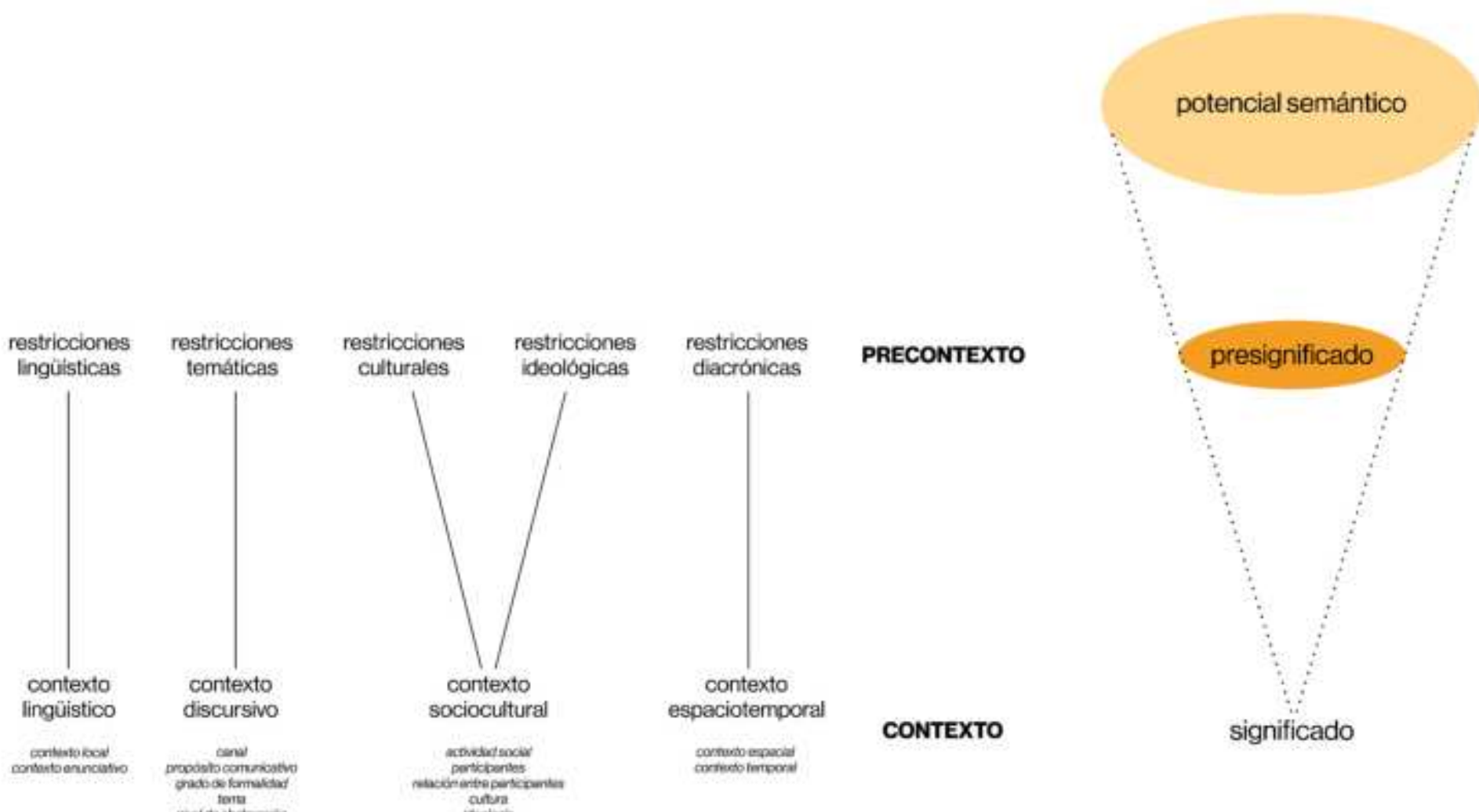

**Figura 15. Contexto, significado, precontexto, presignificado y potencial semántico**

A continuación, pasamos a tratar en mayor detalle las restricciones temáticas en la definición terminológica, pues se trata del foco de nuestra investigación.

#### 3.5.3.4 Las restricciones temáticas

Las restricciones temáticas reducen el potencial semántico de una unidad léxica con arreglo al asunto que se trata en el acto comunicativo y al punto de vista adoptado. Como hemos indicado antes, el tema del discurso permite predecir en mayor medida que otros factores contextuales de qué manera se va a restringir el potencial semántico de una unidad léxica dada. No obstante, la noción de tema es demasiado amplia e inespecífica, ya que el tema de un discurso puede caracterizarse con distintos niveles de generalidad y desde distintas perspectivas. Por poner un ejemplo, de un mismo discurso se podría decir que el tema es «la energía», «la energía eólica», «la energía eólica marina», «la energía eólica marina en la costa de Andalucía», etc. Asimismo, el mismo tema puede tratarse desde distintos puntos vista, así, la caracterización del tema en el ejemplo anterior podría incluir una matización sobre la perspectiva adoptada «desde el punto de vista del impacto medioambiental», «desde el punto de vista de la tecnología empleada», «desde el punto de vista legislativo», etc. A su vez, la perspectiva adoptada puede expresarse con mayor o menor granularidad, por ejemplo, en vez de «desde el punto de vista legislativo», podría ser «desde el punto de vista del Real Decreto 1028/2007».

En este trabajo, para caracterizar de manera sistemática la abstracción que suponen las restricciones temáticas recurrimos a la noción de *dominio*. Dicha noción puede entenderse, por un lado, como equivalente de categoría conceptual o marco, o como un campo de conocimiento especializado (Faber y León Araúz 2014: 142). Es el segundo sentido el que utilizamos en este trabajo, el cual es sinónimo (o cuasi-sinónimo) de disciplina, área temática, área especializada, ámbito temático, área especializada, campo temático, campo especializado, etc.

Los dominios pueden entenderse como una suerte de macromarcos que subsumen a su vez otros marcos a partir de los cuales se organiza y se categoriza una parcela de conocimiento. Los conceptos que se incluyen en un dominio no pertenecen a él, sino que se activan en él con una conceptualización específica que variará si se activa en otro dominio (Cabré 1999: 124).

Los dominios son constructos humanos y, como tales, no son universales ni estables, sino que son dinámicos y de límites difusos (Tebé 2005: 17). Los dominios pueden clasificarse según diferentes visiones o necesidades, no hay una clasificación única válida:

> Les domaines sont délimités en fonction des visions des connaissances, des pratiques sociales et des besoins des utilisateurs. Il existe plusieurs façons de procéder au découpage des connaissances et des activités, qui correspondent à plusieurs points de vue. […] C'est en fonction d'une perspective, d'un point de vue particulier, d'un cadrage, que sera détermine le contour d'un espace conceptuel. (De Bessé 2000: 187)

Las clasificaciones por dominios más conocidas que abarcan todo el conocimiento humano son las que tienen un fin documental. Las más utilizadas son la clasificación de la Biblioteca del Congreso de los Estados Unidos y la clasificación decimal Dewey, aunque también goza de popularidad la clasificación decimal universal (Lazarinis 2014: 139). Cada una de ellas, debido a sus diferentes características, son más o menos aptas para distinto tipo de tareas. Por ejemplo, como explica Lazarinis (2014: 150), las bibliotecas públicas tienden a utilizar más la clasificación decimal Dewey y las bibliotecas académicas la clasificación de la Biblioteca del Congreso de los Estados Unidos, debido a que clasificación decimal Dewey tiene un mayor número de categorías de alto nivel

Por lo general, los recursos terminológicos multidisciplinares no utilizan las clasificaciones documentales y emplean las suyas propias, como es el caso de TERMIUM Plus, el Grand Dictionnaire Terminologique o Termcat. Una excepción es IATE, cuya clasificación por dominios está basada en EuroVoc, un tesauro multilingüe creado para fines documentales por la Oficina de Publicaciones de la Unión Europea.

En este trabajo utilizamos una versión simplificada de la clasificación de dominios creada expresamente para EcoLexicon, la cual, como se expone en §3.6.1, se desarrolló para gestionar la multidimensionalidad mediante restricciones contextuales basadas en dominios.

### 3.5.3.5 La división de sentidos

#### *3.5.3.5.1 La polisemia y la variación contextual*

El fenómeno sobre el que se centra este estudio es la variación contextual (también conocido como *vaguedad*). No obstante, para caracterizar la variación contextual es necesario previamente abordar el fenómeno de la polisemia y sus diferencias con la variación contextual.

El potencial semántico de una unidad léxica se compone de uno o varios conceptos junto con los marcos que estos pueden evocar. Cuando una unidad léxica está asociada a más de un concepto, se produce el fenómeno de la ambigüedad léxica (Cruse 2011: 100). Cada uno de los conceptos asociados a una unidad léxica dada recibe el nombre de *sentido*. Tradicionalmente, se conciben dos tipos de ambigüedad léxica: la polisemia y la homonimia. La polisemia es el fenómeno que ocurre cuando los conceptos relacionados con una unidad léxica están relacionados entre sí (esto es, que presentan solapamiento en mayor o menor grado). Por su parte, la homonimia es cuando dichos conceptos no están relacionados.

La ambigüedad léxica no suele provocar problemas a nivel comunicativo (Falkum y Vicente 2015: 1), ni siquiera en ámbitos especializados, en los que la polisemia es también muy común independientemente del nivel de especialidad (Bertels 2011). Sin embargo, este fenómeno causa muchas dificultades a nivel empírico (Falkum y Vicente 2015: 1), lo cual incluye la redacción de definiciones.

Tradicionalmente, los diccionarios definen las unidades léxicas homónimas en entradas distintas (Atkins y Rundell 2008: 281), es decir, se las trata como si fueran palabras distintas que se escriben de la misma manera. En lo que respecta a las unidades léxicas polisémicas, se ofrece una definición distinta dentro de la misma entrada, organizadas en acepciones.

No obstante, la distinción entre polisemia y homonimia no es siempre fácil de establecer. De acuerdo con Croft y Cruse (2004: 111), existen tradicionalmente dos criterios principales para diferenciar la polisemia de la homonimia: el diacrónico y el sincrónico. El criterio diacrónico se basa en la noción de que la homonimia se produce porque que dos unidades léxicas distintas que, por la evolución de la lengua u otros motivos, terminaron teniendo la misma ortografía. No siempre es posible determinar el origen de las palabras y, por ende, este criterio a veces no se puede aplicar. Por ello, se recurre al criterio sincrónico, según el cual los sentidos no relacionados se consideran homónimos, mientras que aquellos en los que se puede observar algún tipo de proximidad semántica reciben el tratamiento de polisémicos. No existe una frontera clara entre la homonimia y la polisemia; por ello, Cruse (2011: 115) arguye que la homonimia y la polisemia forman un contínuum.

De acuerdo con Atkins y Rundell (2008: 281), en muchos diccionarios ya se está abandonando el tratamiento diferenciado de la polisemia y la homonimia por considerarse poco útil. Por su parte, en EcoLexicon y otros recursos terminológicos orientados al concepto, el tratamiento de la polisemia y de la homonimia es el mismo: la unidad terminológica se asocia a más de un concepto sin distinguir entre los dos fenómenos.

La polisemia[54] se opone a la monosemia, la cual se produce cuando una unidad léxica tiene un único sentido (Falkum y Vicente 2015: 1) o, en otras palabras, cuando está asociada solamente a un único concepto. Sin embargo, es necesario tener en cuenta que, en todo evento de uso, se activa una parte distinta del potencial semántico, esto es, un subconjunto del perfil (el concepto) y un subconjunto de la base (los marcos). Es decir, la monosemia no implica que una unidad léxica signifique siempre lo mismo, sino que una unidad léxica dada siempre activa un mismo único concepto.

[54] Dado que consideramos que la distinción entre polisemia y homonimia no es clara en muchos casos, en adelante, utilizaremos el término *polisemia* para englobar tanto la polisemia como la homonimia —a menos que se indique lo contrario—. Preferimos esta denominación a la de *ambigüedad léxica*, pues algunos autores la utilizan como sinónimo de homonimia (por ejemplo, Tuggy [1993] y Blank [1999]).

Así pues, la variación contextual es el fenómeno que surge cuando un concepto en eventos de uso reales no activa siempre los mismos rasgos y la relevancia de estos varía. Como ya se ha visto, este fenómeno es inherente a la cognición y comunicación humanas. Todos los conceptos varían según el contexto de activación; sin embargo, no siempre con la misma intensidad. Por ello, ampliando el contínuum entre polisemia y homonimia, resulta más efectivo entender la dicotomía entre polisemia y monosemia también como un contínuum (Langacker 1987: 18). Por lo tanto, dicho contínuum (Figura 16) va desde el mayor grado de ambigüedad léxica, que sería la homonimia, hasta el mayor grado de monosemia, que equivaldría a un bajo nivel de variación contextual.

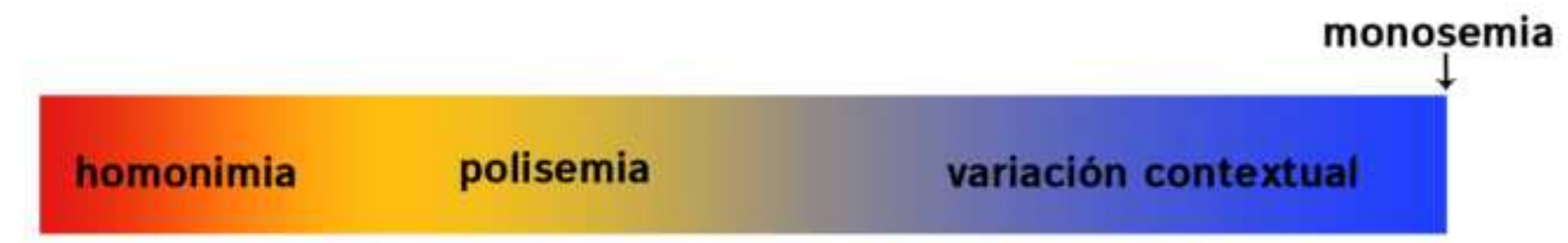

**Figura 16. Continuum desde la homonimia hasta la monosemia**

Aunque en teoría la distinción entre la polisemia y la variación contextual es clara, en la práctica resulta complicado distinguir entre monosemia con un alto grado de variación contextual y polisemia:

> [I]n normal-life situations, words only occur in concrete utterances and not in their more abstract and somewhat idealized dictionary definition. Thus, one has to define where contextual variation of one sense ends and where the semantic range of another sense starts – this is the distinction between vagueness and polysemy (Blank 1999: 15).

La frontera entre polisemia y variación contextual es difusa y su determinación es una cuestión metodológica. Heylen et al. (2015: 161) defienden que la decisión de agrupar o escindir sentidos dependerá finalmente de la función y los usuarios del recurso, lo cual conecta directamente con la dimensión funcional de la definición (§3.5.2). Es decir, como no existen límites precisos entre polisemia y variación contextual, el terminólogo propondrá la división de sentidos que considere más útil para el usuario.

Como ya se ha visto, en recursos como EcoLexicon, a partir del resultado de sus análisis, el terminólogo crea un concepto por cada sentido distinto en los casos de polisemia y, en los casos de variación contextual, la unidad terminológica la asociará a un único concepto. No obstante, gracias a la recontextualización (§3.6.1) y a la propuesta de definición terminológica flexible que se presenta en este trabajo (§3.6), en EcoLexicon se representa también de qué manera varían los conceptos según el dominio de activación. En esto se distingue de recursos terminológicos más tradicionales, en los que si se deciden agrupar determinados sentidos en un solo concepto, se deja de representar información que es potencialmente útil para los usuarios.

No obstante, en su tarea de distinción entre polisemia y variación contextual, el terminólogo puede ayudarse de determinadas pruebas. De acuerdo con Geeraerts (1993) existen tres tipos principales:

- Pruebas lógicas: De acuerdo con estas pruebas, entre las que destaca la propuesta por Quine (1960: 129), una unidad léxica es polisémica si puede ser simultáneamente verdadera y falsa respecto al mismo referente. (Geeraerts 1993: 229). Por ejemplo, se puede decir que la oración «Un dado es un cubo, pero no es un cubo», demuestra que la unidad léxica *cubo* es polisémica respecto a CUBO (forma geométrica) y CUBO (recipiente), ya que no es posible que un referente pertenezca a ambas categorías, como en el caso de un dado que pertenece a la primera categoría, pero no a la segunda.
- Pruebas de identidad semántica: Estas pruebas, como la de Lakoff (1970) y Zwicky y Sadock (1975: 21-31), implican la

aplicación de restricciones semánticas a oraciones que contienen dos ocurrencias relacionadas de la misma unidad léxica. Si la relación semántica entre ambas ocurrencias requiere su identidad semántica, los sentidos son polisémicos. Por ejemplo, puesto que el enunciado «Paula encontró la entrada y Jorge, también» requiere interpretar o bien que Paula y Jorge encontraron el lugar por donde se entra, o bien los billetes, ello implica que *entrada* es polisémica respecto a ENTRADA (lugar por donde se entra) y ENTRADA (billete). Si el emisor pretende que se interprete que cada uno de ellos encontró un tipo distinto de *entrada,* se está haciendo un juego de palabras.

- Pruebas definicionales: Según estas pruebas, una unidad léxica es polisémica si no es posible crear una definición mínimamente específica que cubra toda su extensión. Por ejemplo, la unidad léxica *manzana* es polisémica porque no es posible crear una definición que abarque MANZANA (fruta) y MANZANA (espacio delimitado por calles).

A pesar del atractivo de estas pruebas, Geeraerts (1993) demostró cómo para la misma unidad léxica cada uno de ellos daba un resultado distinto bajo las mismas condiciones y que incluso la misma prueba daba resultados divergentes en distintos contextos. Estas divergencias se deben al carácter difuso de los límites conceptuales y la variabilidad de las concepciones derivadas del potencial semántico en distintos eventos de uso. Es decir, la validez de estas pruebas reposa sobre una noción estática del significado y sobre la visión de que los conceptos están perfectamente delimitados y poseen rasgos suficientes y necesarios, la cual, como vimos, no se ajusta a lo que se conoce acerca de la cognición humana.

Por otro lado, cabe destacar que estas pruebas son más difíciles de aplicar en unidades terminológicas que en unidades léxicas generales porque, en muchos casos, dependen de que la persona que realice la prueba juzgue la validez de determinados enunciados o definiciones de acuerdo con su intuición lingüística. En caso de análisis transdisciplinares, para poder aplicar en condiciones similares estas pruebas a términos —esto es, pudiendo apelar al sentido de la intuición— sería necesario que el terminólogo tuviese unos conocimientos profundos sobre la terminología de varios dominios simultáneamente, lo cual impide recurrir a expertos, pues encontrar especialistas que lo sean simultáneamente de todos los dominios involucrados es altamente improbable[55].

Además de las pruebas reseñadas por Geeraerts (1993), existen también pruebas de tipo léxico-semántico aplicables a la terminología, de las cuales L'Homme (2004) destaca cinco[56]:

- Sustitución por un sinónimo: De acuerdo con esta prueba, si dos sentidos de una unidad léxica tienen sinónimos distintos, es probable que sean polisémicos. Por ejemplo, mientras que en «Defienden la energía verde», *verde* tiene como sinónimo *renovable;* en «Votaron al partido verde», el sinónimo sería *ecologista.* Ello es un indicio de que *verde* está asociado al menos a dos conceptos.
- Oposición diferencial: Esta prueba es similar a la de sustitución por un sinónimo, pero empleando, en este caso, un antónimo o término opuesto. Por ejemplo, en «Esa compañía emplea energía limpia», el antónimo de *limpio* es *contaminante;* sin embargo, en «Los operarios dejaron la playa limpia», su antónimo sería *sucio.* Ello es un indicio de que *limpio* está asociado al menos a dos conceptos.
- Derivación morfológica diferencial: De acuerdo con esta prueba, si una unidad léxica tiene derivados morfológicos que solo se aplican en determinados casos, es un indicio de polisemia. Por ejemplo, mientras que en «La contaminación es demasiado alta en la costa», un derivado morfológico de *costa* sería *costero;* en el enunciado «La multinacional petrolera tuvo que pagar las costas del juicio», *costa* no tendría como derivado *costero*. Ello indica que *costa* en cada caso podría tener sentidos distintos.

[55] Ello sin contar que en ocasiones los expertos no son capaces de proveer la información que el terminólogo requiere: «[D]espite the fact that experts may be very knowledgeable in their particular field, they are not experts in metacognition. In other words, they may know a great deal about their domain, but are not aware of how they know what they know, or how this knowledge is structured». (Faber y San Martín 2011: 50)

[56] Hemos omitido de la lista de pruebas recopilada por L'Homme (2004) el de la coocurrencia compatible por ser equivalente a las pruebas de identidad semántica vistas anteriormente.

- Presencia de relaciones paradigmáticas diferenciales: En esta prueba se incluye cualquier otra relación paradigmática no cubierta en las pruebas anteriores, como la cuasi-sinonimia, la meronimia, la hiponimia, etc.
- Coocurrencia diferencial (Mel'čuk, Clas y Polguère 1995: 66-67): De acuerdo con esta prueba, la coocurrencia con distintas estructuras sintácticas, argumentos, unidades léxicas, etc. es un indicio de la existencia de polisemia.

Cruse (1982: 68) considera que estas pruebas léxico-semánticas (las cuales él denomina *pruebas indirectas*) tienen un valor cumulativo y que cuanto más indicios de polisemia o variación se consigan a través de ellas, más posibilidades hay de que el diagnóstico sea correcto. Además, en el mismo artículo, presenta contraejemplos que apoyan su defensa de la necesidad de no considerar los resultados individuales de estas pruebas como resultado definitivo. Por ejemplo, en el caso de la prueba de sustitución por sinónimo, si tomamos el enunciado «Esa persona es mi padre», la unidad léxica *persona* podría sustituirse por el sinónimo *hombre,* mientras que en «*Esa persona es mi madre*» el sinónimo sería *mujer*. La prueba, por tanto, indicaría la existencia de polisemia, cuando realmente se trata de un caso de variación contextual. Por otro lado, a menudo no se pueden aplicar todas las pruebas porque, por ejemplo, no siempre es posible encontrar sinónimos u otros términos relacionados paradigmáticamente.

La falta de fiabilidad total de las pruebas que se proponen para distinguir entre polisemia y variación contextual proviene del hecho de que el potencial semántico de una unidad léxica es el conjunto de los posibles significados que puede tener dicha unidad, los cuales se solapan y tienen límites difusos.

No obstante, la activación conceptual de las distintas partes del potencial semántico no es aleatoria, sino que, debido a las restricciones convencionales, ciertos contenidos, a menudo agrupados, tienen mayor tendencia a ser activados:

> When we retrieve a word from the mental lexicon, it does not come with a full set of ready-made sense divisions. What we get is a purport, together with a set of conventional constraints. However, in particular cases there may be powerful stable constraints favoring the construal of certain sense units. If the permanent constraints are pushing very strongly in one direction, a correspondingly strong countervailing pressure will be necessary to go against them; if the permanent constraints are weak, whether a boundary is construed or not will depend on other, mainly contextual, factors (Croft y Cruse 2004: 109).

Para dar cuenta de esas fuertes restricciones convencionales, supeditadas a las restricciones contextuales, Cruse (2011) emplea la noción de *límites de sentido* y *límites de subsentido*, los cuales han de entenderse como una tendencia de activación de determinadas áreas de contenido conceptual que funcionan a modo de presignificados dentro del potencial semántico de una unidad léxica. Mientras que los límites de sentido dividen el potencial semántico en distintos conceptos y dan lugar al fenómeno de polisemia; los límites de subsentido se deben a la variación contextual dentro de un único mismo concepto. Naturalmente, la existencia de límites de sentido no impide que a su vez aparezcan límites de subsentido.

Cruse (2011: 101-110) caracteriza ambos límites mediante la noción de autonomía, inspirada en las numerosas pruebas existentes para distinguir la polisemia de la variación contextual. La autonomía explica que determinadas porciones del potencial semántico de una unidad léxica funcionen de manera independiente al resto del contenido semántico asociado a la unidad (Cruse 2006: 18). Cruse (2011: 101-103) describe tres tipos principales de autonomía:

- Autonomía atencional: Se dice que dos concepciones muestran autonomía atencional una respecto a la otra si son mutuamente antagónicas; es decir, cuando no pueden activarse de manera simultánea (Cruse 2011: 101). Por ejemplo, en el enunciado «El gato de Luis estaba en el jardín», no es posible interpretar al mismo tiempo *gato* como animal y *gato* como herramienta; por ello, se puede afirmar que hay autonomía atencional. Los ejemplos que aporta Cruse son reminiscentes de las pruebas lógicas y de identidad semántica. De acuerdo con Cruse, la polisemia, a diferencia de la variación contextual, se caracteriza por mostrar autonomía atencional radical (Cruse 2011: 104).

- Autonomía relacional: La autonomía relacional se da cuando dos concepciones distintas tienen relaciones paradigmáticas independientes: sinonimia, antonimia, meronimia, hiponimia, etc. Asimismo, Cruse (2011: 102) afirma que si las unidades léxicas relacionadas con cada uno de los supuestos sentidos están alejadas semánticamente, ello se puede interpretar como señal de mayor autonomía relacional. Por ejemplo, como dos de los sinónimos de *cuadro* (*cuadrado*, por un lado, y *pintura*, por otro) están alejados semánticamente, ello indica mayor autonomía relacional. Las pruebas léxico-semánticas basadas en relaciones paradigmáticas vistas anteriormente pueden servir para determinar la autonomía relacional.
- Autonomía composicional: La autonomía composicional se refiere al hecho de que uno de los elementos que participan en un proceso composicional (por ejemplo, la modificación de un nombre por un adjetivo) o la interacción de un verbo con sus argumentos, interactuará solamente con una parte del potencial semántico de su compañero. Por ejemplo, en «órgano desafinado» el adjetivo solamente puede modificar ÓRGANO como instrumento musical, mientras que en «órgano genital» solo puedo hacerlo a ÓRGANO como parte del cuerpo. El objetivo de la prueba léxico-semántica de coocurrencia diferencial se corresponde en gran parte con la detección de este tipo de autonomía.

Es importante remarcar que la autonomía no es una cuestión absoluta, sino que se presenta en diversos grados. A mayor autonomía, mayores posibilidades hay de que haya polisemia en vez de variación contextual. No obstante, en la variación contextual, hay unidades que presentan niveles destacables de autonomía, de ahí que sea difícil trazar la frontera con la polisemia.

Dado que la autonomía se presenta en distintos grados y formas, es posible caracterizar diferentes clases de variación contextual. A continuación, revisamos los tipos principales descritos por Cruse (1995; 2000; 2001; 2011)[57]. desde el que presenta menos autonomía al que más: la modulación contextual, los modos de ver y los microsentidos.

#### *3.5.3.5.2 La modulación contextual*

La modulación contextual es el nivel más bajo de variación contextual. Cruse (1986) expone dos parejas de fenómenos a los que da lugar la modulación contextual: la promoción y la democión, por un lado, y el resaltado (*highlighting*) y la relegación (*backgrounding*), por el otro.

La promoción y la democión (Cruse 1986: 52) se refieren al cambio de estatus que un rasgo puede experimentar a causa del contexto con respecto a su prototipo. Cruse (1986: 16) clasifica los características conceptuales en orden descendente según su grado de necesidad como necesarias, esperables canónicas[58], esperables no canónicas, posibles, no esperables y excluidas. De modo que la promoción supone subir en la escala (p. ej., que un rasgo en un contexto pase de ser posible a esperable) y la democión, lo contrario (p. ej., que un rasgo en un contexto pase de ser esperable a excluido).

---

[57] Cruse (2000; 2008) propone un cuarto tipo de fenómeno: las facetas. Sin embargo, en este trabajo, las concebimos como un tipo de polisemia. El tratamiento que reciben en EcoLexicon las facetas es el mismo que cualquier otro caso de polisemia. Además, son numerosos los autores (Frisson 2015; Pustejovsky 2005; Srinivasan y Snedeker 2011, *inter alia*) que consideran que *book*, el ejemplo prototípico de unidad léxica con facetas presentado por Cruse, es un caso de polisemia. La particularidad de las facetas es que es posible unificar en un contexto ambos sentidos (o facetas), a diferencia de otros casos de polisemia regular. Asimismo, cada una de las facetas corresponde a un tipo ontológico distinto, por ejemplo, la unidad léxica *libro,* que se puede dividir en dos facetas: TOMO, que es una entidad física, y TEXTO, que es una entidad abstracta.

[58] Una característica esperable canónica es aquella cuya ausencia es considerada como un defecto (Cruse 1986: 19). Por ejemplo, que un perro tenga cuatro patas es una característica esperable canónica.

Un ejemplo de promoción es el caso de *tsunami* en el enunciado «Existe riesgo de tsunami en el lago Tahoe». Dado que los tsunamis prototípicos son marinos, el rasgo relativo a que un tsunami tiene lugar en un lago pasa de ser posible (o incluso no esperable, dado el alto nivel de prototipicidad del tsunami marino), a asumir el estatus de necesario en este contexto. Al mismo tiempo, se produce la democión de esperable a excluido del rasgo relativo a que un tsunami tiene lugar en el mar.

En lo que respecta al resaltado y la relegación (Cruse 1986: 53), estos dos fenómenos surgen cuando el contexto altera la relevancia de los rasgos. El resaltado aumenta la relevancia de un rasgo, mientras que la relegación lo disminuye. Por ejemplo, en el enunciado «No tenemos suficiente dinero para instalar energía solar» se resalta el rasgo relativo a que la energía solar es cara, mientras que relega a un segundo plano que la energía solar es beneficiosa para el medio ambiente.

Estos cuatro fenómenos no son exclusivos de la modulación contextual, ya que se dan en los otros tipos de variación contextual que veremos a continuación. La diferencia entre la modulación y los demás tipos de variación es que las distintas interpretaciones a las que da lugar este fenómeno no muestran signos de autonomía (Cruse 2001: 38).

#### *3.5.3.5.3 Los modos de ver*

Los modos de ver (*ways-of-seeing [WOS]*) son un tipo de variación contextual cuyas conceptualizaciones resultantes presentan mayor nivel de autonomía que la simple modulación contextual (Cruse 2011: 111). Los modos de ver no suponen ningún cambio referencial, simplemente se adopta una perspectiva diferente:

> A simple way of explaining these would be by analogy with looking at everyday object from in front, the sides, from behind, from on top, etc. All these different views are perceptually distinct, but the mind unifies them into a single conceptual unity. (Cruse 2011: 111)

Para caracterizar los modos de ver, Croft y Cruse (2004: 137) se basan en los roles de *qualia* de Pustejovksy (1995):

- El modo de ver meronímico (*part-whole WOS*) (basado en el rol constitutivo): se observa la entidad en conjunto con sus partes. El ejemplo que Cruse (2011: 111) pone es el del veterinario, para el cual, lo más relevante del caballo es el correcto funcionamiento de su cuerpo con todas sus partes.
- El modo de ver taxonómico (*kind WOS*) (basado en el rol formal): se observa la entidad como un tipo entre otros tipos. Cruse (2011: 112) presenta el ejemplo del zoólogo taxonomista que verá principalmente al caballo según cómo este difiere de otras especies, cuáles son sus subespecies, etc.
- El modo de ver funcional (*functional WOS*) (basado en el rol télico): se observa la entidad en términos de su función. Cruse (2011: 112) lo ejemplifica con un jinete profesional, para el cual la característica principal de los caballos es el poder usarse en carreras.
- El modo de ver biohistórico (*life-history WOS*) (basado en el rol agentivo): se observa una entidad en términos de su historia vital, especialmente cómo se originó. De acuerdo con Cruse (2011: 112), este sería el caso de un constructor que verá una casa desde la perspectiva de su construcción.

De acuerdo con Cruse (2000: 47), podría haber más tipos de modos de ver, aunque no se trataría de una lista ilimitada. Asimismo, Croft y Cruse (2004: 138) reconocen que existen casos en los que es difícil asignar una conceptualización de manera inequívoca a uno de los modos de ver. Por otro lado, puede haber diferencias contextuales dentro de los modos de ver: por ejemplo, es posible atribuir más relevancia a una función que a otra dentro del modo de ver funcional o que se apliquen distintos criterios de clasificación dentro del modo de ver taxonómico.

Los modos de ver muestran, por un lado, un cierto nivel de autonomía relacional (Croft y Cruse 2004: 137). Por ejemplo, si HOTEL es visto como un tipo de inmueble, sus cohipónimos serán CASA, FÁBRICA, etc. Si es visto como un tipo de alojamiento, sus cohipónimos serán HOSTAL, ALBERGUE, etc.

Por el otro lado, muestran un mayor nivel de autonomía composicional. Por ejemplo, «hotel caro» puede significar «hotel caro respecto a lo que cuesta comprarlo» (modo de ver taxonómico), «hotel caro respecto a lo que cuesta alojarse en él» (modo de ver funcional) u «hotel caro respecto a lo que cuesta construirlo» (modo de ver biohistórico).

A partir de la descripción de los modos de ver por de Cruse, se desprende que se trata de un fenómeno de resaltado y relegación de rasgos. Sin embargo, no es descartable que los modos de ver puedan ir acompañados de promoción y democión de rasgos, aunque no fuera forzosamente atribuible al modo de ver sino a una modulación contextual colateral.

Como indica Tercedor (2011: 186) la conexión entre los modos de ver y la multidimensionalidad en terminología es obvia. Si bien dentro de un mismo dominio pueden observarse los conceptos simultáneamente desde varios modos de ver, es de esperar que los distintos dominios tengan tendencia a activar de manera privilegiada unos modos de ver frente a otros. Por ejemplo, los dominios técnicos como la ingeniería civil o la agricultura tienen probablemente mayor propensión a adoptar el modo de ver funcional que las ciencias como la química o la física.

#### *3.5.3.5.4 Los microsentidos*

Los microsentidos son áreas conceptuales dentro del potencial semántico de una unidad léxica que muestran ciertos grados de autonomía. Sin embargo, su autonomía no es suficientemente alta como para ser considerados sentidos polisémicos (Cruse 2001). Los microsentidos son, por tanto, subconjuntos dentro de un mismo concepto.

Los microsentidos tienen una organización jerárquica, con un sentido superordinado general y varios microsentidos subordinados que suponen especificaciones del sentido general en determinados contextos. Una de las características principales de los microsentidos es lo que Cruse (2011: 109) denomina *especificifidad por defecto*, la cual implica que, a menos que el contexto fuerce la conceptualización del sentido superordinado, la interpretación de la unidad léxica corresponderá con la de alguno de los microsentidos.

Los microsentidos dependen fuertemente de dominios específicos, en los que suelen funcionar como categorías de nivel básico y suelen ser prototípicos (§2.1.3.2), mientras que el sentido general, el cual es normalmente esquemático, presenta las características de una categoría de tipo superordinado sin una imagen visual clara, sin unos patrones de comportamiento definidos y no se corresponde con un prototipo (Cruse 1995: 40). Además, el sentido general no suele tener un lugar fijo en ningún sistema conceptual concreto (Croft y Cruse 2004: 127). Un ejemplo de unidad léxica con microsentidos es *knife* en inglés (Cruse 2001: 40):

ej. 41 This surgeon is very skilled at using the knife.

ej. 42 I will need a knife to eat this sirloin steak.

ej. 43 They sell all kinds of knife.

En el ej. 41, *knife* se activa contexto sobre cirugía y, por eso, surge el microsentido asociado al tipo de *knife* que usan los cirujanos. Por su parte, el ej. 42, activa el microsentido de *knife* como cubierto. Por último el ej. 43 muestra como el contexto puede forzar una interpretación no específica. De acuerdo con Kleiber (2006), una posible explicación del fenómeno de los microsentidos es su relación con las distintas funcionalidades que pueden adquirir los referentes del concepto en cuestión.

Aunque los microsentidos no son conceptos independientes y, por tanto, no se trata de un caso de polisemia, presentan mayor nivel de autonomía que las modulaciones contextuales y que los modos de ver.

Por un lado, los microsentidos muestran cierto grado de autonomía atencional; es decir, los distintos microsentidos son generalmente incompatibles entre sí. Para probarlo, Cruse (2001: 42-43) propone los siguientes ejemplos en los que aplica la prueba lógica (ej. 44) y la prueba de identidad semántica (ej. 45):

ej. 44 Mother: (at table; Johnny is playing with his meat with his fingers) Use your knife to cut your meat, Johnny.
Johnny: (who has a pen-knife in his pocket, but no knife of the proper sort) I haven't got one.

ej. 45 John needs a knife; so does Bill.

En el ej. 44, vemos cómo Johnny interpreta que lo que tienen en su bolsillo no es un *knife* según el microsentido al que se refiere su madre. De acuerdo con, Kleiber 2006 el ej. 44 no es una prueba de antagonismo y presenta como contraejemplo el caso de una persona A en el momento de subirse al tren en Estrasburgo con dirección a París a la que una persona B le pregunta si tiene un billete. La persona A responde que no porque el billete que tiene es el de vuelta, esto es, de París a Estrasburgo. Naturalmente, en este ejemplo de uso de *billete* no se trata de un microsentido.

Por otro lado, volviendo al ej. 44, si aplicamos el tipo de prueba lógica que se vio en §3.5.3.5.1, no es posible decir que lo que tiene Johnny en su bolsillo «is a knife, but not a knife», por lo que se puede afirmar que los microsentidos muestran autonomía atencional, pero en menor grado que en un caso de polisemia.

En cuanto al ej. 45, la prueba de identidad semántica no resulta conclusiva, ya que, aunque la tendencia es a interpretar que John necesita el mismo tipo de *knife*, es posible imaginar contextos en los que no se aplique dicha restricción. Así pues, la prueba lógica y la de identidad semántica muestran que los microsentidos presentan autonomía atencional, pero menor que en un caso de polisemia.

Los microsentidos también presentan autonomía relacional. Cruse (2001: 42) muestra cómo cada uno de ellos tiene distintos hiperónimos y cohipónimos:

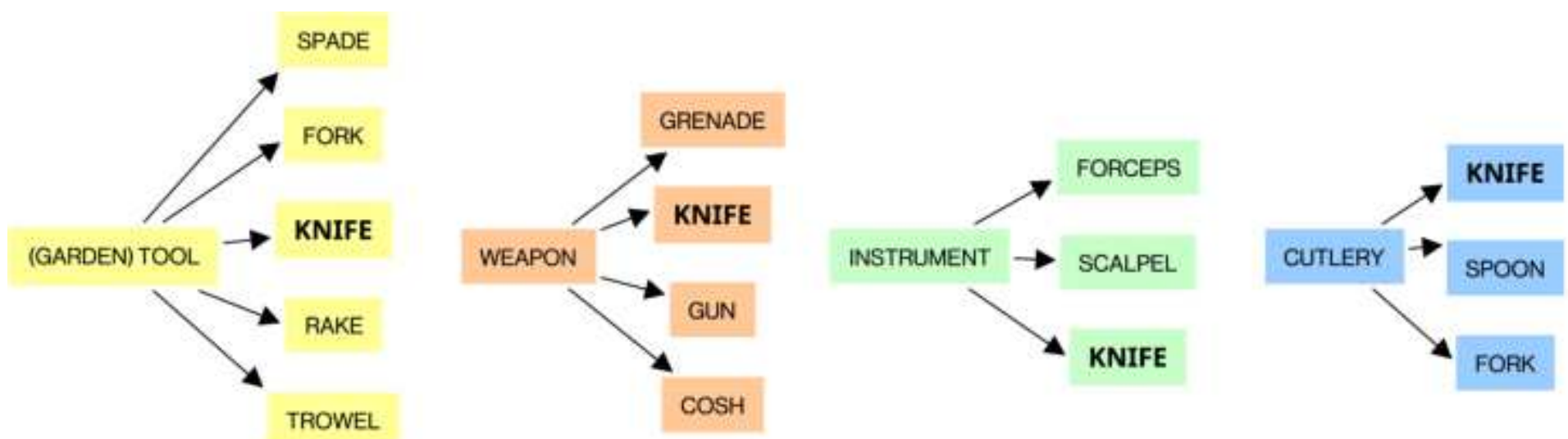

**Figura 17. Hiperónimos y cohipónimos de *knife*** (Cruse 2001: 42)

Asimismo, los microsentidos muestran cierta autonomía composicional. Por ejemplo, si la unidad léxica *card* en inglés se utiliza como objeto directo del verbo *swipe,* se activan solamente aquellos microsentidos de *card* con una banda magnética como las tarjetas bancarias y, a su vez, se descartarían microsentidos como, por ejemplo, el que en español se traduciría por *postal*.

A veces, aunque no siempre, hay una forma lexicalizada para el microsentido por la adición de algún modificador (Croft y Cruse 2004: 135). Por ejemplo, el *knife* que usa el cirujano puede recibir la denominación de *surgical knife*. Sin embargo, como indica Cruse (2011: 109) en el contexto de su dominio dicha especificación resulta extraña, ya que un cirujano no le pedirá a una enfermera que le pase un *surgical knife* sino simplemente le pedirá un *knife*.

Por otro lado, cabe destacar que, mientras que en los modos de ver tan solo había un cambio de perspectiva (lo cual daba lugar al resaltado y a la relegación de rasgos), en los microsentidos se producen cambios referenciales, por lo que se producirá principalmente promoción y democión de rasgos.

Cabe destacar también una variante de los microsentidos llamada *subsentidos locales* (Cruse 2011: 109-110). El ejemplo que Cruse plantea es el de una madre que acude al colegio de su hijo para una reunión de profesores y padres. Como no encuentra la sala donde se celebra tal reunión, se acerca a un hombre que está en la entrada y le pregunta: «¿Es usted profesor?». En este contexto, la interpretación más probable es que la madre le esté preguntando si es profesor en ese colegio. De modo que se está utilizando la unidad léxica *profesor* para hacer referencia a un tipo concreto de profesor.

De acuerdo con Cruse (2011: 110), los subsentidos locales se diferencian de los microsentidos en que la concepción supeordinada en este caso es rica y robusta, puede funcionar como unidad de nivel básico y participa en campos léxicos concretos.

Así pues, si bien la modulación contextual, los modos de ver y los microsentidos son fenómenos reales como bien demuestra Cruse con sus ejemplos, es de esperar que en la realidad la frontera entre ellos sea difusa y que una misma unidad léxica pueda presentar rasgos de los tres tipos de variación contextual al mismo tiempo.

En §5, tomaremos estos tres fenómenos como punto de partida en el estudio de la representación de la variación contextual mediante restricciones temáticas (por dominio) en la definición terminológica. Para la aplicación de dichas restricciones, proponemos un tipo de definición terminológica que hemos denominado *definición terminológica flexible* y que exponemos a continuación.

## 3.6 LA DEFINICIÓN TERMINOLÓGICA FLEXIBLE

Las definiciones, ya sean en recursos lexicográficos o terminológicos, tienen tendencia a ser demasiado generales, pues intentan abarcar todo tipo de contextos:

> A common problem with conventional definitions is that they are underspecified - that is, in trying to account for all possible instantiations of a word, they often resort to minimalist formulations that can be slotted into any conceivable context. (Rundell 2006: 330)

Para los conceptos con un alto nivel de variación contextual, una única definición que abarque todo el dominio medioambiental no resulta suficientemente informativa, como es el caso de las definiciones de OZONE en diferentes recursos:

ej. 46 A chemical that is made of three oxygen atoms joined together, and found in the Earth's atmosphere. There are two kinds of ozone: good ozone, and bad ozone. Good ozone is found high in the Earth's atmosphere, and prevents the sun's harmful rays from reaching the Earth. Bad ozone is found low to the ground, and can be harmful to animals and humans because it damages our lungs, sometimes making it difficult to breathe.[59]

ej. 47 An unstable gas, $O_3$. It is formed naturally in the atmosphere and also by an electric discharge in oxygen. It has a bluish colour and distinctive odour and is used in the purification of air and water.[60]

ej. 48 An allotropic form of oxygen containing three atoms in the molecule. It is a bluish gas, very active chemically, and a powerful oxidizing agent. Ozone is formed when oxygen or air is subjected to a silent electric discharge. It occurs in ordinary air in very small amounts only.[61]

Las definiciones en los ejemplos anteriores no resultan útiles para un usuario que tenga que enfrentarse al concepto de OZONE en diferentes subdominios del medio ambiente. Por ejemplo, en abastecimiento y tratamiento de agua, es importante saber que el ozono se utiliza como desinfectante dado su poder oxidante y viricida. Además, como se puede observar, no hay consenso respecto al concepto supeordinado de OZONE, pues cada una de las definiciones utiliza un genus distinto: CHEMICAL, GAS o ALLOTROPIC FORM OF OXYGEN.

Por ello, en este trabajo, se propone la creación de definiciones terminológicas flexibles. Una definición flexible es un sistema de definiciones del mismo concepto compuesto por una definición general (en nuestro caso, que abarca todo el dominio del medio ambiente) junto con definiciones contextualizadas en las que el concepto se sitúa en los distintos subdominios en los que el concepto es relevante. En cierto modo, nuestra propuesta de definiciones flexibles recoge la recomendación de Lorente (2001: 105) sobre cómo se debe representar la multidimensionalidad en la definición terminológica: «[P]ara superar las limitaciones del texto, para describir un concepto se deberían construir definiciones múltiples, parciales y complementarias, atendiendo a la parcialidad de perspectivas».

[59] *EPA Terminology Services* (United States Environmental Protection Agency 2015)

[60] *A Dictionary of Chemistry* (Daintith 2014)

[61] *GEMET* (European Environment Agency 2015)

Nuestra propuesta de definición terminológica flexible reposa sobre las nociones de multidimensionalidad y recontextualización por dominios tal y como se aplican en EcoLexicon. Por ello, antes de ofrecer los detalles acerca de la definición terminológica flexible, nos detendremos sobre esas dos cuestiones.

## 3.6.1 La multidimensionalidad y la recontextualización en EcoLexicon

Comunmente la multidimensionalidad se define como el fenómeno que surge cuando se clasifican los conceptos desde distintas perspectivas en un mismo sistema conceptual (Bowker 1997: 133). De acuerdo con Rogers (2004), se puede hablar de dos tipos de multidimensionalidad. El primero es el producido por la formalización de distintas relaciones genérico-específicas al mismo nivel con arreglo a distintas características diferenciadoras. En la Figura 18 se puede ver un ejemplo simple de sistema conceptual que es multidimensional porque los conceptos se categorizan según distintas características diferenciadoras simultáneamente.

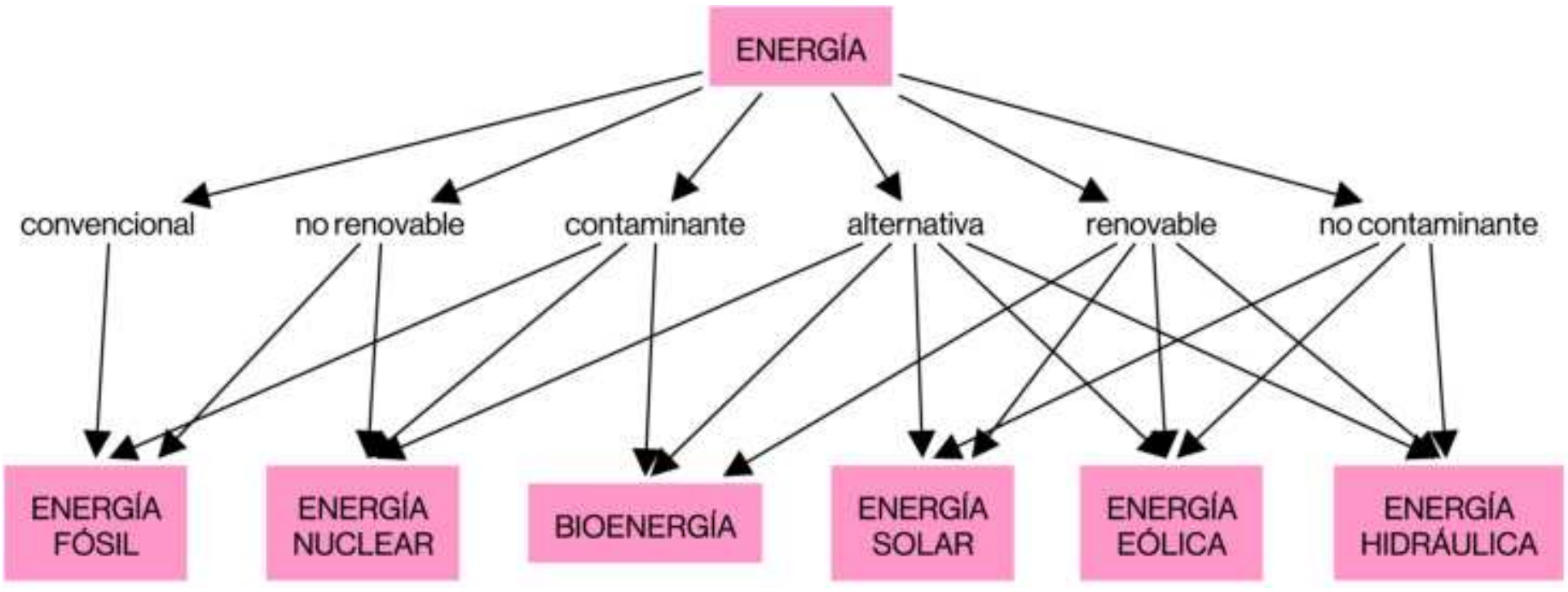

**Figura 18. Sistema conceptual multidimensional por el uso de distintas características diferenciadoras**

El segundo tipo de multidimensionalidad se refiere a la combinación de distintos tipos de relaciones en un mismo sistema conceptual, como relaciones genérico-específicas con partitivas o relaciones genérico-específicas con temporales. Por ejemplo, el sistema conceptual de la Figura 19 es multidimensional porque se emplean tanto relaciones genérico-específicas como partitivas al mismo tiempo.

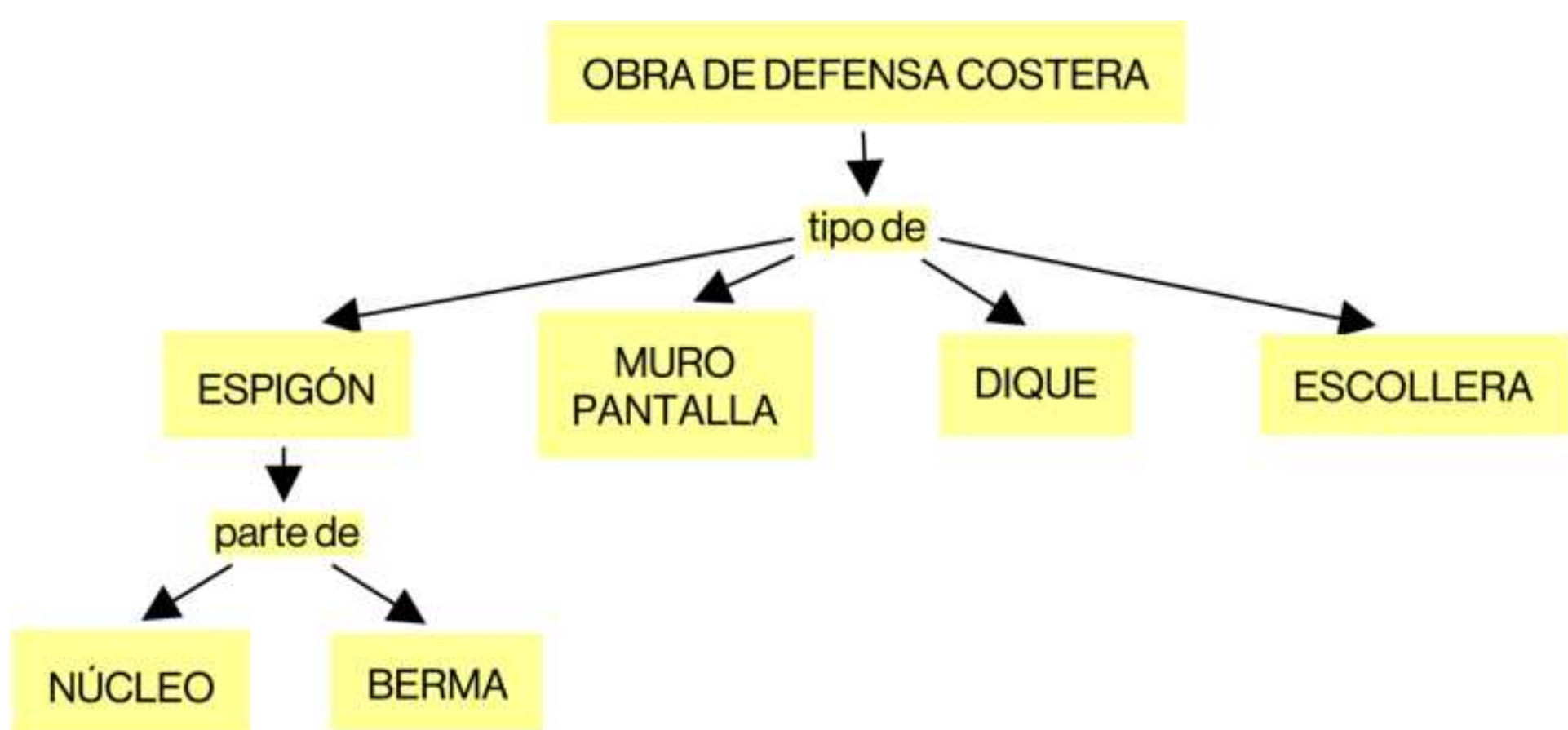

**Figura 19. Sistema conceptual multidimensional con relaciones genérico-específicas y partitivas**

No obstante, la multidimensionalidad puede entenderse en un sentido más lato del que exponen Bowker (1997) y Rogers (2004). La existencia de las distintas dimensiones de un concepto está determinada por diversos factores como su categoría conceptual, su rol conceptual, el dominio en que se activa, etc. y ello hace que las relaciones conceptuales (tanto jerárquicas como no jerárquicas) que establece el concepto varíen de una dimensión a otra. En las bases de conocimiento terminológicas, la multidimensionalidad permite enriquecer representaciones estáticas mediante la inclusión de distintos puntos de vista en un sistema conceptual; sin embargo, esto puede ocasionar una gran sobrecarga de información que impida la adquisición del conocimiento por parte del usuario (León Araúz y Faber 2010: 15). EcoLexicon padece este problema de sobrecarga de información en la representación gráfica de determinados conceptos porque el medio ambiente es un campo de estudio interdisciplinar muy amplio, lo cual ocasiona que un mismo concepto pueda activarse en distintos dominios contextuales, como la geología, la ingeniería o la meteorología. Esto ocurre especialmente en el caso de los conceptos versátiles, los cuales tienen una profundidad jerárquica reducida y, por tanto, se conceptualizan en variedad de contextos dentro de un dominio especializado como el medio ambiente (León Araúz 2009: 196).

La conceptualización en varios dominios distintos supone que, aunque el concepto sigue siendo el mismo, en distintos contextos, la relevancia de las distintas proposiciones conceptuales que se establecen varía o incluso se establecen proposiciones distintas. León Araúz (2009: 194) lo ejemplifica con el concepto SEDIMENTO el cual prototípicamente se categoriza como el RESULTADO de un PROCESO NATURAL, PACIENTE en el proceso de EROSIÓN y AGENTE del proceso de COLMATACIÓN. Sin embargo, si se restringe el dominio al de los PROCESOS COSTEROS, SEDIMENTO se convierte prototípicamente en el material que compone la LÍNEA DE COSTA o en el dominio de la DEFENSA DE COSTAS se convierte en el material usado para la REGENERACIÓN DE PLAYAS. De este modo, proposiciones que en otros contextos eran prototípicas pierden de manera parcial o total su relevancia en un nuevo contexto.

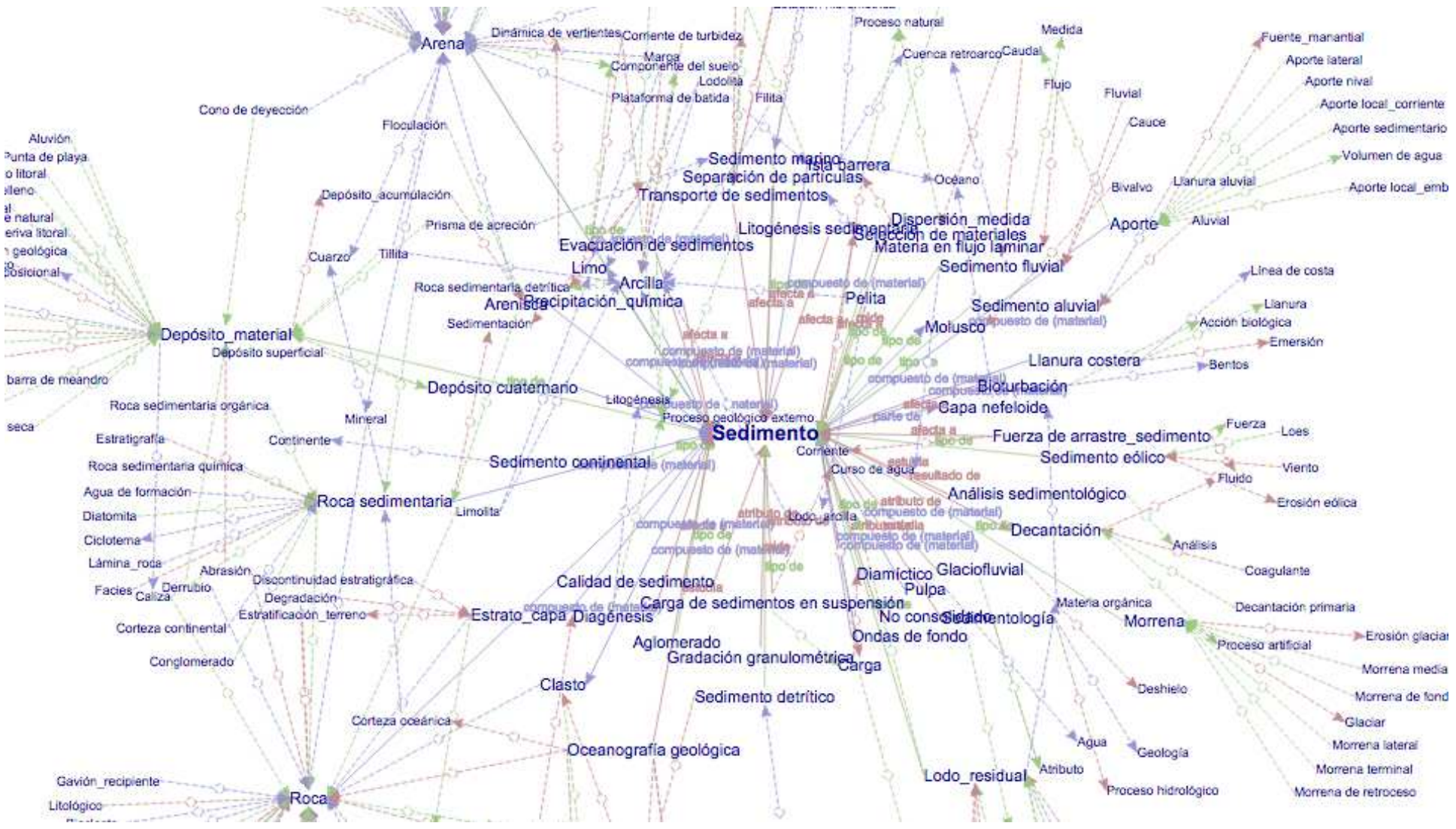

**Figura 20. Red conceptual de SEDIMENTO en EcoLexicon sin restricciones contextuales**

SEDIMENTO está conectado en su primer nivel con otros 50 conceptos en EcoLexicon. Si se representan gráficamente todas las proposiciones conceptuales en las que participa SEDIMENTO sin aplicar ningún tipo de contextualización, ocurre, por tanto, una sobrecarga de información (Figura 20).

A partir de esta representación gráfica, la adquisición de conocimiento por parte del usuario no es solo dificultosa por la grandísima cantidad de información que se muestra, sino también por la disparidad de los conceptos con los que se relaciona SEDIMENTO, como DECANTACIÓN, MOLUSCO o GLACIOFLUVIAL. Ante esta situación, León Araúz (2009) propuso la aplicación de restricciones contextuales mediante la asociación de las proposiciones conceptuales a los contextos en los que se activan, es decir, recontextualizarlas. De este modo, en la representación gráfica de las proposiciones conceptuales que establece un concepto, tan solo se mostrarán aquellas proposiciones relevantes a un contexto determinado dando así, además, cuenta de la variación contextual. En la Figura 21 puede verse la red conceptual de SEDIMENTO cuando se representa en el dominio contextual del abastecimiento y tratamiento de aguas.

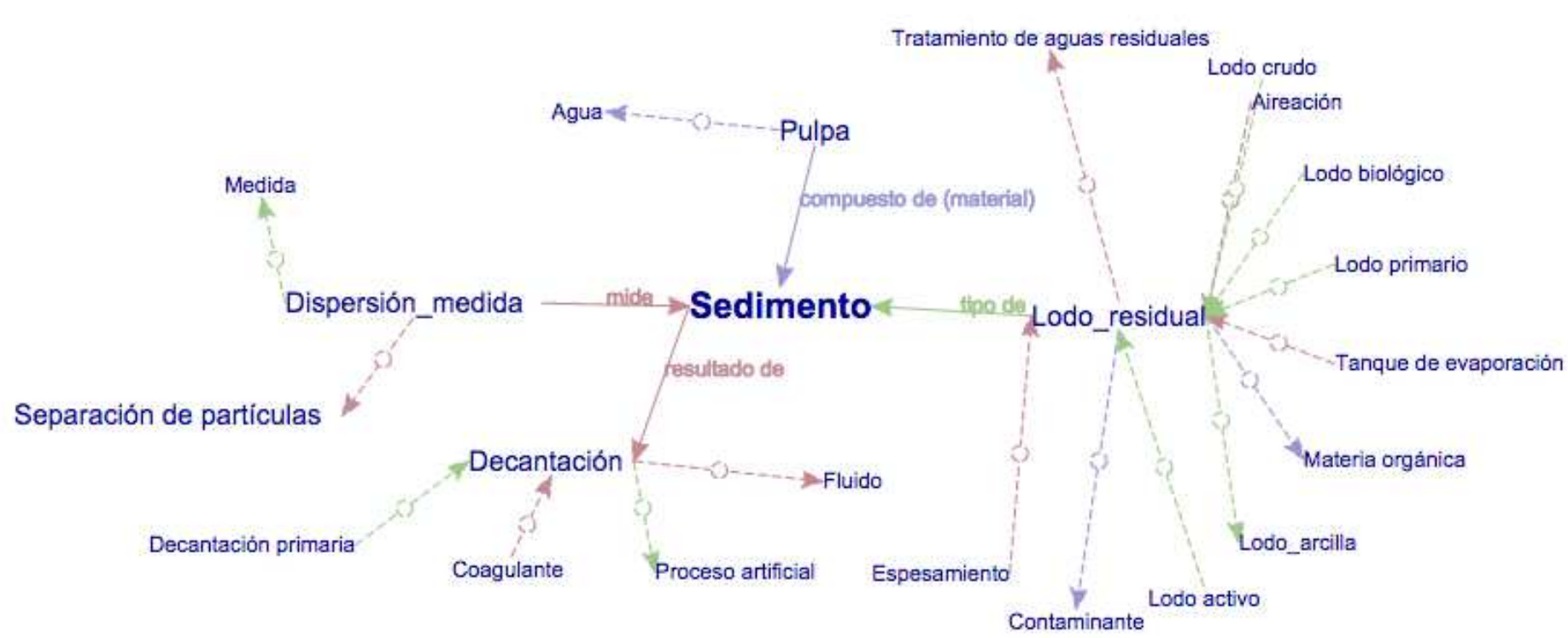

**Figura 21. Red conceptual de SEDIMENTO en EcoLexicon restringida al dominio de abastecimiento y tratamiento de aguas**

En EcoLexicon, para recontextualizar las proposiciones conceptuales, estas se adscriben a uno o varios de los siguientes dominios contextuales, que funcionan como restricciones temáticas (§3.5.3.4):

**1. PROTECCIÓN MEDIOAMBIENTAL**

- 1.1. Derecho medioambiental
- 1.2. Educación medioambiental
  - 1.2.1. Turismo sostenible
- 1.3. Gestión de recursos naturales

**2. CIENCIAS**

- 2.1. Geografía
- 2.2. Biología
  - 2.2.1. Oceanografía biológica
  - 2.2.2. Botánica
  - 2.2.3. Zoología
  - 2.2.4. Microbiología
  - 2.2.5. Biología Molecular
  - 2.2.6. Bioquímica
- 2.3. Física
  - 2.3.1. Geofísica
  - 2.3.2. Oceanografía física
- 2.4. Geología
  - 2.4.1. Hidrogeología
  - 2.4.2. Geofísica
  - 2.4.3. Geoquímica
  - 2.4.4. Oceanografía geológica
  - 2.4.5. Geomorfología
- 2.5. Hidrología
  - 2.5.1. Hidrogeología
  - 2.5.2. Hidrometeorología
- 2.6. Química
  - 2.6.1. Geoquímica
  - 2.6.2. Bioquímica
  - 2.6.3. Oceanografía química
- 2.7. Ciencias atmosféricas
  - 2.7.1. Meteorología
    - 2.7.1.1. Oceanografía meteorológica
    - 2.7.1.2. Hidrometeorología
  - 2.7.2. Climatología
- 2.8. Ecología
  - 2.8.1. Ecología humana
- 2.9. Ciencias del suelo
- 2.10. Oceanografía
  - 2.10.1. Oceanografía biológica
  - 2.10.2. Oceanografía física
  - 2.10.3. Oceanografía meteorológica
  - 2.10.4. Oceanografía geológica
  - 2.10.5. Oceanografía química

**3. INGENIERÍA**

- 3.1. Ingeniería naval
- 3.2. Ingeniería civil
  - 3.2.1. Ingeniería de transporte e infraestr.
  - 3.2.2. Ingeniería hidráulica
  - 3.2.3. Ingeniería costera
  - 3.2.4. Ingeniería de minas
  - 3.2.5. Ingeniería medioambiental
    - 3.2.5.1. Gestión de residuos
    - 3.2.5.2. Abastecimiento y trat. de aguas
    - 3.2.5.3. Gestión de la calidad del aire
    - 3.2.5.4. Gestión de la calidad del suelo
- 3.3. Ingeniería agronómica
- 3.4. Ingeniería química
- 3.5. Ingeniería energética
  - 3.5.1. Energías renovables

**Tabla 17. Dominios contextuales en EcoLexicon**

De este modo, en relación con un dominio contextual concreto, las proposiciones de un concepto pertenecen a dos conjuntos distintos (León Araúz y San Martín 2012: 579)[62]:

[62] Esta clasificación de las proposiciones en activa e inactiva y otras subdivisiones que se presentarán más adelante es una adaptación y ampliación de la clasificación de Seppälä (2009: 47-48) de rasgos latentes, destacados, potencialmente relevantes y relevantes vistos en §3.5.

- Proposiciones activas: Son aquellas que resultan relevantes respecto al concepto en ese dominio contextual.
- Proposiciones inactivas: Aquellas que o bien son irrelevantes (proposiciones latentes) o aquellas que nunca se activarían en ese dominio (proposiciones incompatibles).

La distinción entre proposición inactiva latente e incompatible puede explicarse mediante las nociones de resaltado y relegación, y de promoción y democión de Cruse (1986), que vimos en §3.5.3.5.4. Una proposición conceptual es irrelevante cuando el contexto ha provocado la relegación de esa característica, que en otros contextos medioambientales sí es relevante. Mientras que una proposición inactiva incompatible ha padecido democión: ha pasado de ser necesaria, esperable o posible a ser no esperable o excluida en ese dominio contextual. Por lo general, la democión conlleva también la relegación, aunque puede haber casos en que precisamente la democión de una característica la convierta en relevante.

### 3.6.2 Las proposiciones definicionales

En relación con la recontextualización de EcoLexicon, se puede decir que una definición flexible está compuesta por una definición general medioambiental que deriva de la representación del concepto en cuestión para todo el dominio del medio ambiente en EcoLexicon y varias definiciones para subdominios relevantes que derivan de la recontextualización de las proposiciones conceptuales.

No obstante, en una definición no se representan todas las proposiciones conceptuales activas del concepto, sino que hay que realizar una selección, lo cual divide las proposiciones activas en definicionales o no definicionales (todas las proposiciones inactivas son no definicionales).

Las definiciones contextualizadas han de poder funcionar independientemente. Esto es, deben transmitir toda la información relevante del presignificado en el subdominio en cuestión sin depender del resto de definiciones del conjunto para ello. Por su parte, la definición general medioambiental codifica los rasgos más relevantes en todo el dominio del medio ambiente, por lo cual incluye aquellos rasgos que son prototípicos en la mayoría de subdominios.

### 3.6.3 Las jerarquías contextualizadas y la categorización multidimensional

Una de las principales dificultades que plantea las definiciones flexibles es que, contrariamente a lo que pudiera pensarse, incluso las relaciones genérico-específicas están sujetas a la variación contextual (León Araúz y San Martín 2012: 581), tal y como se mostró con el ejemplo de OZONE y sus múltiples genus. Comprensiblemente, esto puede afectar a la herencia de propiedades en una jerarquía.

Dado que es necesario especificar una jerarquía coherente antes de realizar una definición para asegurar la herencia de propiedades correcta, en el caso de las definiciones flexibles, cada dominio contextual requiere su propia jerarquía. Como se verá en §4.2, las principales fuentes de información que utilizamos para determinar cómo clasificar un concepto en cada subdominio son las definiciones en otros recursos y el análisis de corpus.

Sin embargo, todos los candidatos a superordinado extraídos con estos dos métodos se pueden utilizar solamente como una guía. Los conceptos pueden clasificarse de varias maneras, incluso en el mismo dominio contextual. De hecho, muchas de las categorías que se pueden extraer con estos dos métodos se podrían considerar categorías derivadas de un fin (Barsalou 1983; Barsalou 1985) (§2.1.3.3.1), construidas para un propósito específico en una determinada situación y carente convencionalización, en lugar de categorías bien establecidas.

Así pues, una vez extraído todo el conocimiento necesario sobre el concepto y las distintas formas de categorizarlo, las principales directrices para la estructuración de jerarquías contextualizadas en EcoLexicon son la coherencia (para una herencia de propiedades correcta una vez que EcoLexicon se convierta en una ontología) y la activación del marco conceptual subyacente más prototípico (§3.6.4).

Dado que conviven varias jerarquías al mismo tiempo en EcoLexicon, un concepto puede tener tres o más conceptos superordinados distintos que se pueden clasificar del siguiente modo:

- Superordinado general medioambiental: Es el concepto superordinado de un concepto dado en la jerarquía general medioambiental y funciona como genus en la definición medioambiental general. Es aplicable al concepto en todos los dominios contextuales, aunque en algunos dominios puede no resultar prototípico. Por ejemplo, el concepto DICHLORODIPHENYLTRICHLOROETHANE tiene como superordinado general medioambiental CHEMICAL COMPOUND, es aplicable en cualquier dominio, sin embargo, en agricultura no es el más relevante.
- Superordinado contextual preferencial: Se corresponde con el concepto superordinado empleado en cada jerarquía contextualizada para cada dominio. El superordinado preferencial contextual es el que se utiliza como genus para la definición contextualizada. Por ejemplo, en agricultura, DICHLORODIPHENYLTRICHLOROETHANE tiene como superordinado contextual preferencial el concepto PESTICIDE, pues destaca su función principal dentro de ese dominio.
- Superordinados contextuales no preferenciales: En este grupo se incluyen aquellos superordinados que no son el preferencial del concepto en un dominio dado, pero que no es incompatible. Por ejemplo, en agricultura, DICHLORODIPHENYLTRICHLOROETHANE tiene como superordinado contextual no preferencial al concepto PERSISTENT ORGANIC POLLUTANT.

Todas las relaciones genérico-específicas conllevan herencia de propiedades. En consecuencia, en EcoLexicon, incluso los conceptos contextualizados heredan proposiciones a partir de más de un concepto superordinado. Por ejemplo, DICHLORODIPHENYLTRICHLOROETHANE hereda proposiciones de sus tres superordinados: CHEMICAL COMPOUND, PESTICIDE y PERSISTENT ORGANIC POLLUTANT. El hecho de que un concepto tenga varios superordinados da lugar a la existencia de distinto tipo de proposiciones definicionales:

- Proposiciones definicionales directas: Se corresponden con lo que tradicionalmente se conoce como differentiae. Son proposiciones que el concepto establece directamente y no son heredadas. Un ejemplo sería la proposición «DICHLORODIPHENYLTRICHLORO-ETHANE *made-of* CHLORINE» para el concepto DICHLORODIPHENYLTRICHLOROETHANE en ingeniería química, que la establece el concepto directamente, no la hereda de sus superordinados.

- Proposiciones definicionales indirectas: Son aquellas heredadas por el concepto a través de sus conceptos superordinados. Hay dos tipos:

    - Proposiciones definicionales indirectas implícitas: Son aquellas heredadas del concepto superordinado que actúa de genus. Estas proposiciones generalmente no se representan en la definición, pues puede resultar redundante. Por ejemplo, la proposición «DICHLORODIPHENYLTRICHLORO-ETHANE *affects(=kills)*[63] PEST» se hereda de PESTICIDE y no se representa en la definición del concepto en agricultura porque sería redundante, ya que el genus es PESTICIDE. Sin embargo, sí se representan si alguno de los conceptos que forman la proposición se activa de manera más específica.

[63] Por lo general, la relación *affects* no es lo suficientemente expresiva. En aquellos casos en los que sea pertinente, como el presente, precisaremos el sentido de la relación.

- Proposiciones definicionales indirectas especificadas: Son aquellas heredadas del concepto superordinado que actúa de genus y, que como se activa de manera más específica, sí se representa en la definición. Por ejemplo, FUNGICIDE es un subtipo de PESTICIDE. En este caso, la relación «PESTICIDE *affects(=kills)* PEST» se hereda y se representa en la definición porque se especifica como «FUGINCIDE *affects(=kills)* FUNGUS».

- Proposiciones definicionales indirectas explícitas: Son aquellas que se heredan de los conceptos superordinados contextuales no preferenciales y, además, resultan relevantes en el dominio en cuestión. Por ejemplo, «DICHLORODIPHENYLTRICHLOROETHANE *affects(=pollutes)* ENVIRONMENT» se hereda de PERSISTENT ORGANIC POLLUTANT y, al ser relevante en el dominio de la agricultura (debido a su naturaleza contaminante, muchos países han restringido o prohibido su uso), se representaría en la definición contextualizada en ese dominio.

### 3.6.4 Los marcos conceptuales subyacentes

Fillmore (2003) propuso la integración de la noción de marco en la definición de cualquier palabra mediante lo que denominó *definiciones de dos pisos*. Dicha definición contendría dos partes o pisos: la definición *per se* y una descripción del marco en el que se inserta el definiéndum.

Como se vio en §2.1.3.3, la noción de teoría es imprescindible en la selección de los rasgos con los que describir un concepto, pues explica en gran medida por qué los seres humanos realizamos determinadas categorizaciones frente a otras posibles. En este trabajo, las teorías se aplican mediante los dominios contextuales y los marcos. Los dominios contextuales pueden entenderse a efectos prácticos como grandes conjuntos de marcos que explican las relaciones que establecen los conceptos entre sí y, por ende, la relevancia de estos.

De este modo, coincidimos con Fillmore (2003) en la necesidad de representar los marcos que subyacen a los conceptos en las definiciones. No obstante, en lugar de presentarlos por separado, los integramos en la definición. Esto es, consideramos que la definición terminológica debe describir (de acuerdo con las restricciones contextuales pertinentes) los rasgos relevantes del concepto en cuestión así como el marco o marcos que dan sentido a dichos rasgos[64].

Para ello, el terminólogo durante el proceso de documentación y análisis de corpus para la realización de la definición terminológica deberá detectar en qué marco o marcos se inserta el concepto que se está definiendo. Con vistas la definición, dado que EcoLexicon no tiene un sistema de marcos granular como el de FrameNet, la detección del marco consiste en determinar qué otros conceptos suelen coactivarse con el que se está definiendo[65] y qué relaciones los une. En el caso de que dichas relaciones no estén aún contenidas en EcoLexicon, conviene introducir en la base de datos las correspondientes proposiciones conceptuales, pues podrán utilizarse posteriormente para construir la definición.

En las plantillas definicionales que se utilizan actualmente en EcoLexicon solo se incluyen proposiciones en las que participa el definiéndum. Sin embargo, en esta tesis, proponemos integrar los marcos dentro de la definición mediante la representación de proposiciones en las que participan los otros conceptos activados en la definición o conceptos cuya denominación es un derivado morfológico del definiéndum. Denominamos dichas proposiciones *proposiciones enmarcantes (framing propositions)* porque aportan información sobre el marco o marcos en los que se inserta el definiéndum. Por ejemplo, en la plantilla definicional de CHLORINE para el dominio de la *GESTIÓN DE LA CALIDAD DEL AIRE*, la proposición «CHLOROFLUOROCARBON *affected-by* ULTRAVIOLET RADIATION» explica una etapa esencial del marco OZONE DEPLETION, activado por CHLORINE en ese dominio. En esta proposición enmarcante, no participa CHLORINE directamente, pero lo hace un concepto (CHLOROFLUOROCARBON) que previamente dentro de la definición se ha asociado a CHLORINE mediante una proposición directa («CHLOROFLUOROCARBON *made-of* CHLORINE»).

[64] Nuestra decisión de integrar los marcos dentro de la definición en lugar de describirlos aparte se debe principalmente a una cuestión de índole práctica. Consideramos que aportar las descripciones de marcos junto con las definiciones implica contar con un sistema de marcos estructurado y la creación de tal sistema queda fuera del alcance de esta tesis.

[65] Para ello, es de especial utilidad práctica la detección de contextónimos que describimos en §4.2.2.3.2.

EcoLexicon, con su herramienta de mapa conceptual, permite consultar las proposiciones que establecen los conceptos asociados a un primer concepto. Ello proporciona al terminólogo una representación visual de todas las proposiciones enmarcantes. Aquellas proposiciones enmarcantes que finalmente forman parte de la definición, reciben el nombre de *proposiciones enmarcantes definicionales*. La Figura 22 representa visualmente todos los tipos de proposiciones que hemos expuesto en este apartado.

Por otro lado, el marco que activa prototípicamente el concepto en un dominio contextual debe guiar en la medida de lo posible la selección del superordinado contextual preferencial. A este respecto, podemos dividir los tipos de genus inspirándonos en los roles de *qualia* de Pustejovsky (1995):

- Genus télico: El genus es un concepto que tiene como característica necesaria principal una función, un propósito o un uso que se le da al referente. Por ejemplo, FERTILIZANTE, MEDIDOR, COAGULANTE, DEPURADORA, etc.

- Genus agentivo: El genus es un concepto cuya característica necesaria principal indica el origen del referente. Por ejemplo, PRODUCTO, SUBPRODUCTO, RESIDUO, RESULTADO, etc.

- Genus constitutivo: El genus es un concepto cuya característica necesaria principal indica que el referente es parte de otra entidad o proceso. Por ejemplo, PARTE, COMPONENTE, MATERIAL, FASE, etc.

- Genus formal: El genus es un concepto cuya característica necesaria principal hace referencia las propiedades ontológicas del concepto. Por ejemplo, ENTIDAD, PROCESO, ATRIBUTO, etc.

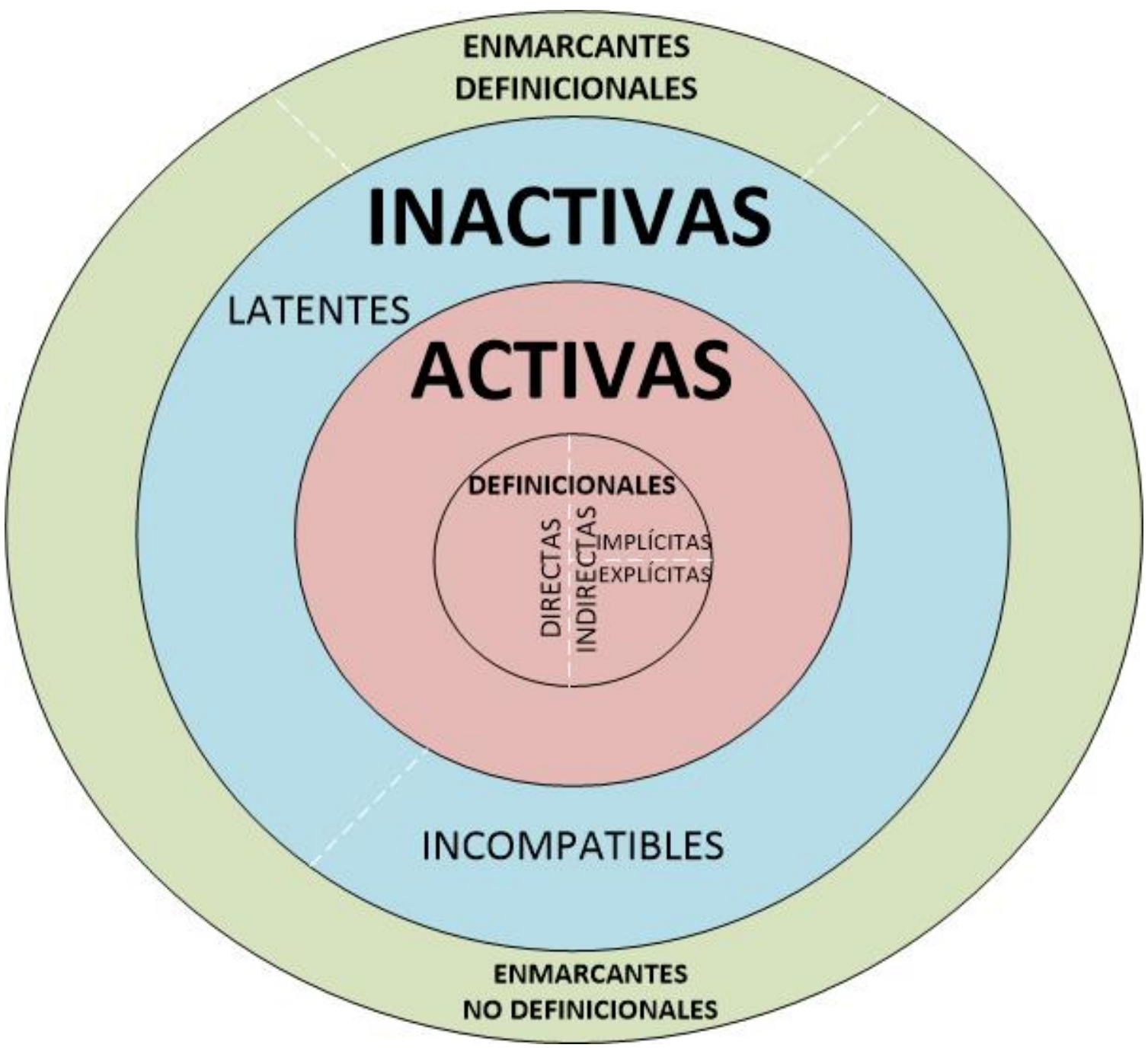

**Figura 22. Representación gráfica de las proposiciones activas, inactivas y enmarcantes**

A menudo, un concepto incluye características necesarias de varios tipos, por lo que, a menudo, esta clasificación que acabamos de presentar es algo difusa. Por ejemplo, un concepto télico como DEPURADORA incluye no solo como característica necesaria su función de depuración, sino que también incluye que se trata de un artefacto, lo cual, además de incluir el hecho de que los artefactos tienen una función, también implicaría un rasgo formal en cuanto que es una entidad física y un rasgo agentivo en cuanto que ha sido construido por el ser humano. Sin embargo, consideramos DEPURADORA y conceptos similares como télicos porque su función es el rasgo que define su papel en los principales marcos en los que participa.

Por consiguiente, es necesario elegir, siempre que resulte posible, un genus del tipo que mejor represente el papel del concepto en el principal marco que activa en un precontexto dado. Por ejemplo, en ingeniería civil, STEEL tendrá como superordinado contextual preferencial BUILDING MATERIAL (de tipo télico), en lugar de ALLOY, (de tipo formal-agentivo) porque el primero pone de relieve la función de STEEL en ese dominio.

### 3.6.5 Las proposiciones conceptuales en el mapa conceptual y en la definición

Dado que la manera en que se codifica la información conceptual en proposiciones en EcoLexicon para su representación en mapas conceptuales no es del todo satisfactoria con vistas a la creación de definiciones, consideramos conveniente ciertas pautas de codificación para aumentar su utilidad a este respecto.

Aunque EcoLexicon se encuentra en transición a convertirse en una ontología formal, por el momento, la base de conocimiento no aplica ningún tipo de herencia de propiedades ni inferencias. En lo que respecta al usuario, esta carencia se ve compensada por el hecho de que el usuario puede ver varios niveles de proposiciones conceptuales. Por ejemplo, si el usuario consulta el concepto PLUVIOGRAMA puede ver en el mapa todas las relaciones que establece ese concepto y a su vez las relaciones que establecen los conceptos relacionados con PLUVIOGRAMA e incluso las relaciones que establecen estos últimos con otros conceptos. Todo ello simultáneamente en el mismo mapa conceptual.

Así pues, para evitar en lo posible la sobrecarga de información (problema que, como ya hemos visto, es frecuente en EcoLexicon) y, de manera temporal hasta que EcoLexicon se convierta en una ontología, entendemos que conviene dar por supuesta la herencia de propiedades y ciertas inferencias.

En lo que respecta a la herencia de propiedades, un ejemplo sería METHANE. Si se categoriza METHANE como GREENHOUSE GAS y GREENHOUSE GAS establece la proposición «GREENHOUSE GAS *causes* GREENHOUSE EFFECT», no se debe codificar la relación «METHANE CAUSES GREENHOUSE EFFECT», pues ha de entenderse como heredada.

Consecuentemente, solo han de codificarse aquellas proposiciones que sean aplicables también a los subordinados de los conceptos en cuestión. Por ejemplo, el concepto OZONE-DEPLETING SUBSTANCE se refiere a sustancias que contienen cloro o bromo (o ambos), por ejemplo, METHYL BROMIDE (que contiene bromo, pero no cloro). Si se codifica «OZONE-DEPLETING SUBSTANCE *made of* CHLORINE» y «OZONE-DEPLETING SUBSTANCE *made of* BROMINE», ello daría lugar a que se interprete que METHYL BROMIDE contiene cloro, lo cual no es correcto. Por ello, en ese caso y similares, al no ser posible añadir disyunciones a las proposiciones en EcoLexicon, este tipo de proposiciones no se deben codificar en el supeordinado y se debe suponer que, en un futuro, se podrán inferir en una ontología a partir de los subordinados.

Una excepción es cuando el concepto subordinado hereda una proposición, pero la activa de manera más específica. En ese caso, sí ha de codificarse en EcoLexicon. Por ejemplo, la proposición «WATER DISINFECTION *effected-by* WATER DISINFECTANT» es heredada por chlorination porque es su subordinado. Sin embargo, CHLORINATION activa de manera más específica dicha proposición («CHLORINATION *effected-by* CHLORINE»), de modo que dicha especificación sí se debe codificar en EcoLexicon.

En cuanto a las inferencias, también es necesario dar por supuestas algunas de ellas, aunque el sistema no las aplique. De lo contrario, al igual que con la herencia de propiedades, se corre el riesgo de duplicar la información y sobrecargar la representación visual en el mapa conceptual. Un ejemplo es cuando un proceso tiene un agente que lo causa. A menudo, para expresar el paciente de dicho proceso es posible codificarlo tanto respecto al agente como al proceso. Por ejemplo, la relación entre FERTILIZER y FERTILIZATION solo hay una manera posible de expresarla mediante proposiciones de EcoLexicon «FERTILIZER causes FERTILIZATION». Sin embargo, para expresar que la fertilización afecta al suelo, es posible codificarlo como «FERTILIZER *affects* SOIL» o «FERTILIZATION *affects* SOIL». En ese caso, como medida de uniformización, proponemos codificar solamente la relación que establece el proceso y, en un futuro, configurar la ontología para que lo infiera a partir de «FERTILIZER causes FERTILIZATION».

Describir todas las posibles inferencias necesarias en EcoLexicon queda fuera del alcance de esta tesis doctoral. Sin embargo, en este trabajo, seguiremos el principio de evitar en todo lo posible la duplicidad de información heredable e inferible en lo que respecta a la codificación de proposiciones en EcoLexicon.

En cuanto a la relación de las proposiciones conceptuales codificadas en EcoLexicon y su relación con las definiciones, conviene remarcar que, si bien defendemos que las definiciones en EcoLexicon deben ser coherentes con la información codificada por las proposiciones, hay una diferencia esencial entre ellas. Mientras que en las proposiciones se ha de evitar la duplicidad de información, en las definiciones no, en el sentido de que dos definiciones pueden compartir parcialmente la misma información explícita (especialmente, aquellas cuyos definiéndums tienen denominaciones que son derivados morfológicos).

En lo que respecta a la herencia de propiedades, en la definición no se expresa la información heredada del concepto superordinado que ejerce de genus. Sin embargo, como ya hemos visto en §3.6.3, proponemos hacer una excepción con los conceptos superordinados que denominamos *no preferenciales*. Mientras que en las proposiciones, la herencia se da por supuesta y no se debe duplicar la información; en la definición, se representan aquellas proposiciones relevantes que se heredan del concepto superordinado no preferencial.

En lo que respecta a las inferencias, es necesario reconstruirlas manualmente para hacer la definición y representar aquellas que tras el análisis conceptual se determinen como relevantes. Por ejemplo, para la definición de OZONE-DEPLETING SUBSTANCE, la característica de que este tipo de sustancias contienen cloro o bromo se inferirá manualmente para representarla en la definición.

# 4 MATERIALES Y MÉTODOS

## 4.1 MATERIALES

### 4.1.1 Aplicaciones informáticas

#### 4.1.1.1 TermoStat Web 3.0

TermoStat[66] es un extractor de términos desarrollado por Drouin (2003) en el Observatoire de linguistique Sense-Texte de la Universidad de Montreal. Actualmente TermoStat trabaja con el inglés, el francés, el español, el italiano y el portugués. Su funcionamiento se basa en la oposición de un texto especializado proporcionado por el usuario con un corpus no especializado (corpus de referencia) con vistas a la identificación de términos (Drouin 2010a).

---

[66] TermoStat se ofrece como una aplicación web accesible en <http://termostat.ling.umontreal.ca/>.

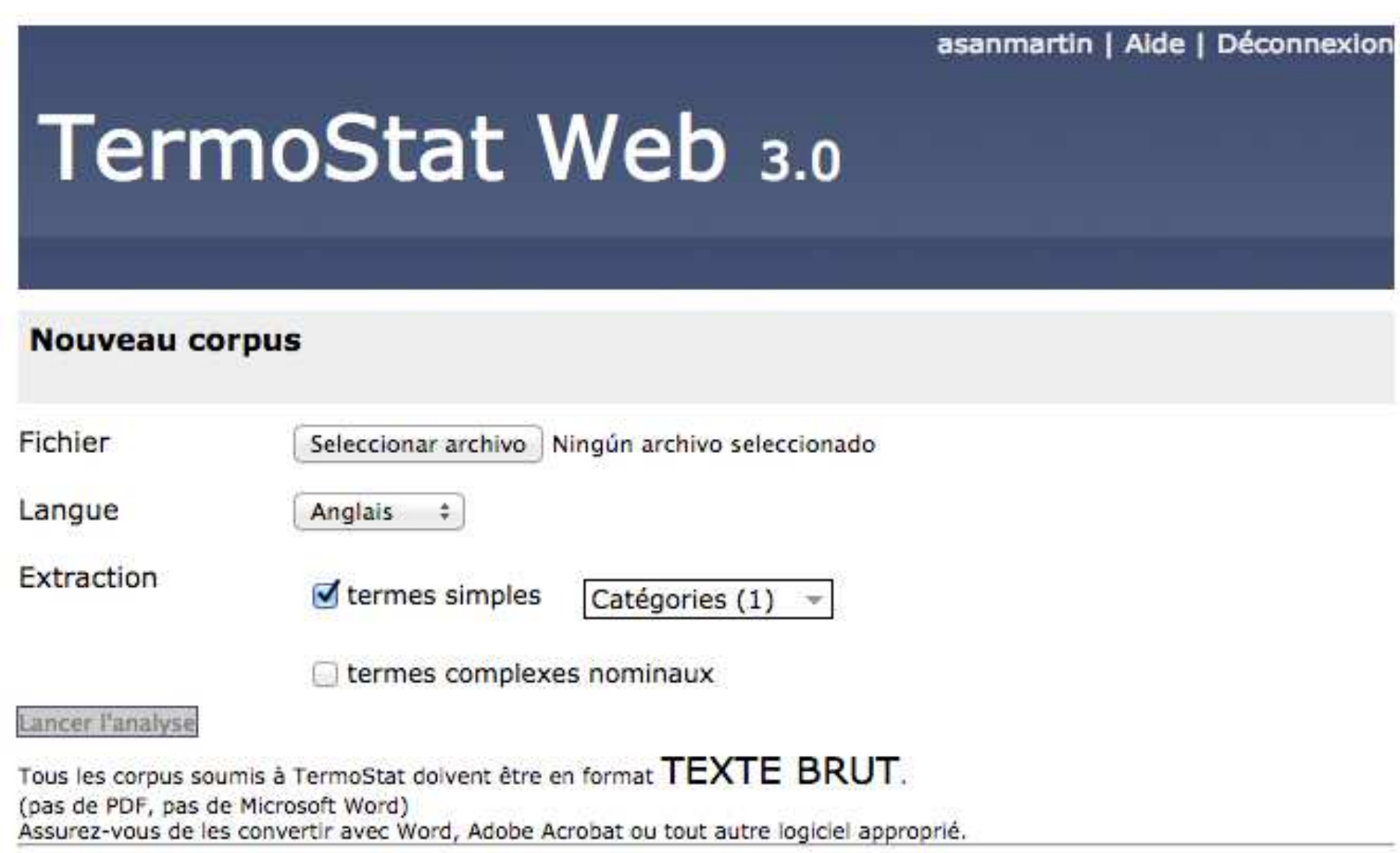

**Figura 23. Interfaz de carga de archivos en TermoStat Web 3.0**

El corpus de referencia en inglés de TermoStat, lengua en la que realizaremos la extracción, tiene aproximadamente 8.000.000 de palabras y en torno a 465.000 formas. Se trata de un corpus no técnico compuesto por artículos periodísticos de temas variados del periódico canadiense Montreal Gazette publicados en 1989 y textos provenientes del British National Corpus (Drouin 2010a).

Al cargar un texto para la extracción terminológica (que debe estar en formato de texto plano), el usuario tiene la opción de elegir si desea que los términos que se extraigan sean simples y/o compuestos. Mientras que los términos compuestos pueden ser nominales, en el caso de los términos simples, es posible limitar la extracción a sustantivos, verbos, adjetivos y/o adverbios. La Figura 23 muestra la interfaz de carga de textos en TermoStat Web 3.0.

Una vez cargado el texto, la aplicación le devuelve como resultado principal una lista de términos candidatos extraídos del texto. En dicha lista, los términos aparecen acompañados de la siguiente información (Figura 24):

- *Fréquence*: frecuencia de aparición en el texto provisto por el usuario

- *Score*: puntuación basada en su frecuencia en el corpus analizado y su frecuencia en el corpus de referencia (Drouin 2010a). Se considera que cuanto más alta la puntuación, mayor pertinencia tendrá el término. El usuario puede elegir qué tipo de fórmula se utiliza para obtener dicha puntuación: especificidad, $X^2$, logaritmo de verosimilitud o logaritmo de la razón de momios.

- *Variantes orthographiques*: variantes ortográficas y flexiones

- *Matrice*: categoría gramatical

La lista de términos candidatos junto con la información asociada es exportable. Asimismo, al pulsar en un término candidato se ofrecen las concordancias donde aparece el término en el texto del que se ha extraído el término.

Además de la lista de términos, el usuario puede hacer clic en otras pestañas con funciones adicionales:

- *Nuage*: Se muestra de manera visual los 100 términos con mayor puntuación a modo de nube de palabras. Cuanto mayor sea la puntuación, más grande aparece el término.

- *Statistiques*: Se ofrece el número de términos extraídos y estadísticas según la estructura gramatical de los términos.

- *Structuration*: Se presenta la lista de términos candidatos y cada uno aparece acompañado de aquellos términos compuestos que lo contienen. Al pulsar en el icono amarillo que acompaña a cada término, se accede a más información relacional y se ofrece la posibilidad de generar gráficos con dicha información.

- *Bigramme*: Se muestra la lista de los bigramas (verbo-sustantivo) más comunes en el texto proporcionado por el usuario. Incluye información sobre la frecuencia y la puntuación de asociación.

Corpus >> kwicsand — asanmartin | Aide | Déconnexion

Résultats

Liste des termes | Nuage | Statistiques | Structuration | Bigrammes

| Candidat de regroupement | Fréquence | Score (Spécificité) | Variantes orthographiques | Matrice |
|---|---|---|---|---|
| sand | 10011 | 450.05 | *sand*<br>*sands* | Nom |
| sediment | 1792 | 189.85 | *sediment*<br>*sediments* | Nom |
| beach | 1746 | 175.85 | *beach*<br>*beaches* | Nom |
| dune | 1223 | 156.98 | *dune*<br>*dunes* | Nom |
| erosion | 1068 | 144.05 | *erosion* | Nom |
| clay | 876 | 121.93 | *clay*<br>*clays* | Nom |
| silica | 614 | 109.75 | *silica* | Nom |
| gravel | 606 | 107.05 | *gravel*<br>*gravels* | Nom |
| deposit | 681 | 97.9 | *deposit*<br>*deposits* | Nom |
| silica sand | 454 | 96.42 | *silica sand*<br>*silica sands* | Nom Nom |
| grain | 519 | 87.77 | *grain*<br>*grains* | Nom |
| wave | 601 | 87.1 | *wave*<br>*waves* | Nom |

**Figura 24. Lista de términos candidatos en TermoStat Web 3.0**

#### 4.1.1.2 SketchEngine

SketchEngine (Kilgarriff et al. 2004; Kilgarriff et al. 2014) es una herramienta de análisis de corpus a la que se accede a través de su aplicación web. Una de sus funciones principales y la que inspira su nombre es la de los *word-sketches*. Si bien, SketchEngine incluye una gran multitud de funciones, nos concentraremos en aquellas de las que hemos hecho uso en este trabajo.

Aunque SketchEngine ofrece numerosos corpus precargados en distintas lenguas o la posibilidad de crear un corpus automáticamente con la herramienta WebBootCat; en este trabajo, hemos cargado nuestros propios corpus especializados (§4.1.2).

Una vez cargado el corpus en la aplicación, el usuario tiene la posibilidad de compilar el corpus para poder hacer uso de todas las funciones que SketchEngine ofrece. El compilado del corpus incluye el etiquetado morfosintáctico del corpus y la aplicación de una gramática de *sketches*. En nuestro caso, el etiquetador morfosintáctico elegido fue TreeTagger 2.5 (inglés) y la gramática de *sketches* fue la que ofrece SketchEngine por defecto para ese etiquetador, aunque le añadimos varios nuevos *sketches* que se describen en §4.2.2.3. Una vez compilado el corpus, es posible definir subcorpus, lo cual configuramos de modo que pudiéramos realizar consultas tanto de todo el corpus como por dominio contextual.

A continuación, repasamos la funciones *concordance*, *word-sketch* y *word list*, que fueron las que empleamos en este trabajo.

*Concordance*

La función *concordance* permite consultar directamente el corpus (o los subcorpus) para obtener líneas de concordancias. En la búsqueda simple (*simple query*), tan solo es necesario escribir una cadena de caracteres para obtener las concordancias. Asimismo, SketchEngine permite realizar búsquedas más avanzadas mediante el lenguaje CQL (*Corpus Query Language*), lo cual permite, entre otras cosas, emplear etiquetas morfosintácticas, lemas y expresiones regulares. Por ejemplo, la búsqueda «[tag="JJ" & lemma!="hazardous"] [lemma="chemical" & tag=”N.*”]» devuelve todas las concordancias en la que aparece el sustantivo *chemical* (lo cual incluye, por tanto, el término en singular y plural, y excluye el término usado como adjetivo) precedido de cualquier adjetivo que no sea *hazardous*.

Al mostrar las concordancias, la aplicación permite, entre otras opciones, aplicar diversos filtros para ordenar los resultados, eliminar aquellos que no sean relevantes, mostrarlos como KWIC (*KeyWord In Context*) u oración completa, o exportar las concordancias en formato txt o xml.

*Word-sketches*

Kilgariff et al. (2004: 105) definen un *word-sketch* como un resumen automático del comportamiento gramatical y colocacional de una palabra dispuesto en una sola página y basado en corpus. Cada *word-sketch* incluye columnas con listas con las palabras que coocurren con la palabra introducida en la búsqueda en un patrón predeterminado en la gramática de *sketch* elegida para compilar el corpus. Algunos patrones que vienen incluidos en la gramática de sketch por defecto en inglés aplicables a sustantivos son, por ejemplo:

- *object_of:* verbos con los que el sustantivo funciona como objeto en el corpus*;*
- *subject_of:* verbos con los que el sustantivo funciona como sujeto en el corpus*;*
- *modifier:* palabra que funciona como modificador del sustantivo en el corpus*;*
- *modifies:* palabra para la que el sustantivo funciona como modificador*;*
- *and/or:* otro sustantivo que coocurre con el sustantivo en una enumeración*;*

En la Figura 25, se reproduce un ejemplo de *word-sketches* tal y como se muestran en la interfaz de SketchEngine.

*Word List*

La función *Word List* permite obtener listas de palabras de un corpus junto con su frecuencia de acuerdo con una gran diversidad de parámetros configurables como el uso de expresiones regulares, lista blanca o negra, comparación con otros corpus, etc.

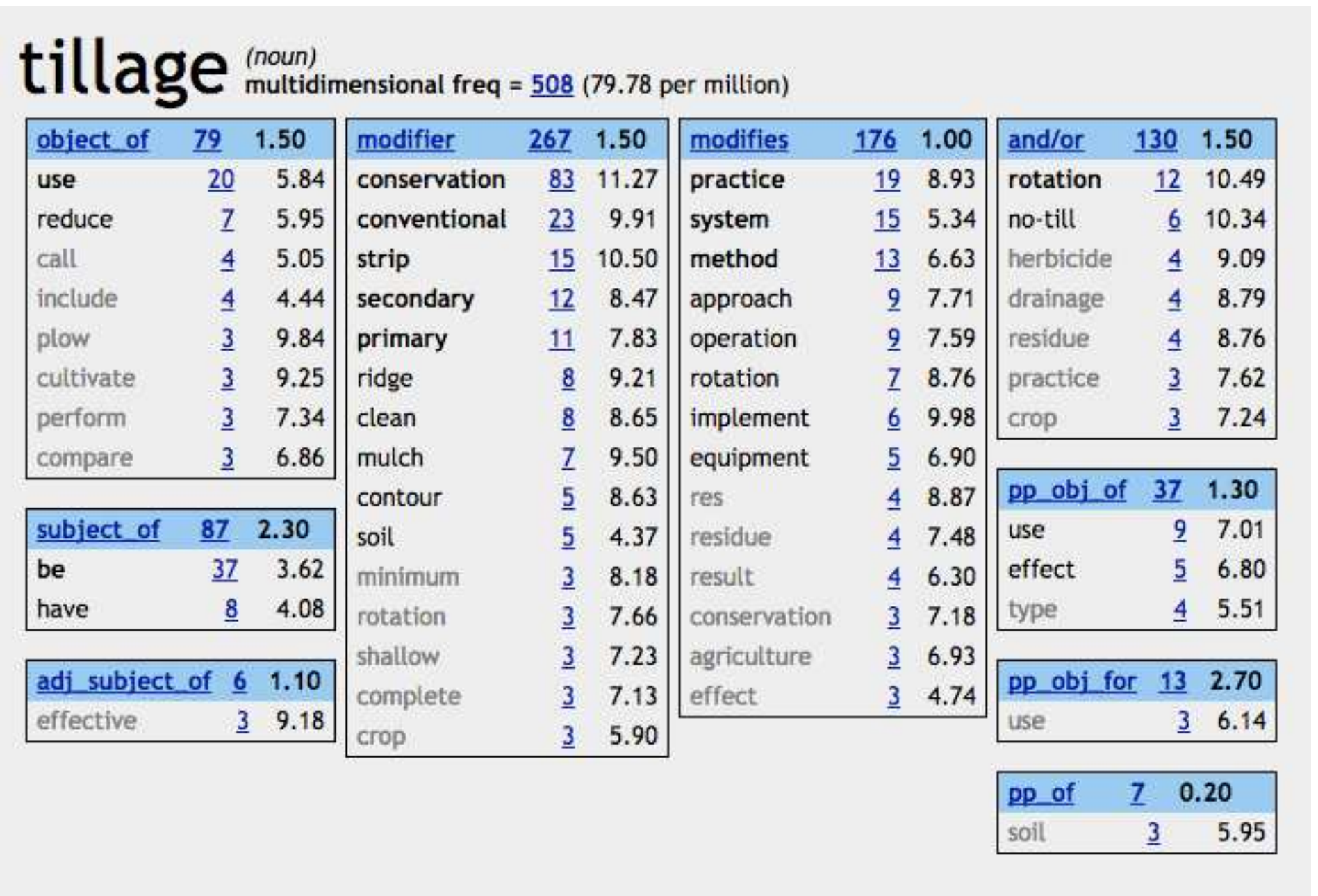

**tillage** (noun) multidimensional freq = 508 (79.78 per million)

| object_of | 79 | 1.50 |
|---|---|---|
| use | 20 | 5.84 |
| reduce | 7 | 5.95 |
| call | 4 | 5.05 |
| include | 4 | 4.44 |
| plow | 3 | 9.84 |
| cultivate | 3 | 9.25 |
| perform | 3 | 7.34 |
| compare | 3 | 6.86 |

| subject_of | 87 | 2.30 |
|---|---|---|
| be | 37 | 3.62 |
| have | 8 | 4.08 |

| adj_subject_of | 6 | 1.10 |
|---|---|---|
| effective | 3 | 9.18 |

| modifier | 267 | 1.50 |
|---|---|---|
| conservation | 83 | 11.27 |
| conventional | 23 | 9.91 |
| strip | 15 | 10.50 |
| secondary | 12 | 8.47 |
| primary | 11 | 7.83 |
| ridge | 8 | 9.21 |
| clean | 8 | 8.65 |
| mulch | 7 | 9.50 |
| contour | 5 | 8.63 |
| soil | 5 | 4.37 |
| minimum | 3 | 8.18 |
| rotation | 3 | 7.66 |
| shallow | 3 | 7.23 |
| complete | 3 | 7.13 |
| crop | 3 | 5.90 |

| modifies | 176 | 1.00 |
|---|---|---|
| practice | 19 | 8.93 |
| system | 15 | 5.34 |
| method | 13 | 6.63 |
| approach | 9 | 7.71 |
| operation | 9 | 7.59 |
| rotation | 7 | 8.76 |
| implement | 6 | 9.98 |
| equipment | 5 | 6.90 |
| res | 4 | 8.87 |
| residue | 4 | 7.48 |
| result | 4 | 6.30 |
| conservation | 3 | 7.18 |
| agriculture | 3 | 6.93 |
| effect | 3 | 4.74 |

| and/or | 130 | 1.50 |
|---|---|---|
| rotation | 12 | 10.49 |
| no-till | 6 | 10.34 |
| herbicide | 4 | 9.09 |
| drainage | 4 | 8.79 |
| residue | 4 | 8.76 |
| practice | 3 | 7.62 |
| crop | 3 | 7.24 |

| pp_obj_of | 37 | 1.30 |
|---|---|---|
| use | 9 | 7.01 |
| effect | 5 | 6.80 |
| type | 4 | 5.51 |

| pp_obj_for | 13 | 2.70 |
|---|---|---|
| use | 3 | 6.14 |

| pp_of | 7 | 0.20 |
|---|---|---|
| soil | 3 | 5.95 |

**Figura 25. *Word-sketch* del lema *tillage* (sustantivo) en el corpus MULTI**

## 4.1.2 Descripción de los corpus

En este estudio se utilizaron dos corpus distintos. El primero de ellos es uno elaborado específicamente para este trabajo (que denominamos MULTI, como apócope de *multidimensional*) y el segundo es el corpus medioambiental monolingüe de inglés elaborado en el marco del proyecto PANACEA.

### 4.1.2.1 Corpus MULTI

El corpus MULTI se elaboró ad hoc para este trabajo y consta de 15 subcorpus: uno que abarca todo el dominio del medio ambiente (ENV) y 14 de subdominios del medio ambiente:

| | **Dominio** | **Palabras** |
|---|---|---|
| AGR | agronomía<br>agronomy | 322.979 |
| AIR | gestión de la calidad del aire<br>air quality management | 319.449 |

| | Dominio | Palabras |
|---|---|---|
| ATM | ciencias atmosféricas<br>atmospheric sciences | 324.559 |
| BIO | biología<br>biology | 321.167 |
| CEN | ingeniería química<br>chemical engineering | 321.672 |
| CHE | química<br>chemistry | 321.569 |
| CIV | ingeniería civil<br>civil engineering | 322.023 |
| ENE | ingeniería energética<br>energy engineering | 323.963 |
| ENV | medio ambiente general<br>general environment | 929.171 |
| GEO | geología<br>geology | 323.523 |
| HYD | hidrología<br>hydrology | 320.599 |
| PHY | física<br>physics | 327.402 |
| SOI | ciencias del suelo<br>soil sciences | 324.248 |
| WAS | gestión de residuos<br>waste management | 322.234 |
| WAT | abastecimiento y tratamiento de aguas<br>water treatment and supply | 322.390 |
| TOTAL | | 5.446.948 |

**Tabla 18. Número de palabras en cada subcorpus y en total**

El subcorpus ENV está compuesto al 100% por manuales semiespecializados de nivel universitario sobre el medio ambiente en general y de una enciclopedia especializada del medio ambiente (ver Anexo 1). Por su parte, todos los subcorpus de dominios están formados en un 62,5% de manuales semiespecializados de nivel universitario del dominio respectivo (aprox. 200.000 palabras) (ver Anexo 1) y un 37,5% de artículos de la Wikipedia (aprox. 120.000 palabras). Tan solo hay una excepción y es el subcorpus de ciencias del suelo que en vez de un 62,5% de manuales semiespecializados, se incluyeron artículos de una enciclopedia sobre ciencias del suelo con textos de características similares a los de un manual, ya que no se encontró ningún manual en formato electrónico de las características deseadas.

La inclusión de manuales semiespecializados y artículos de enciclopedia en el corpus está motivada por el hecho de que, al no estar dirigidos a un público completamente experto, existe mayor probabilidad de que los autores hagan explícitas las características más relevantes de los conceptos que activan en sus textos.

Por su parte, la Wikipedia es una enciclopedia multilingüe en línea colaborativa de libre acceso creada en el año 2001 (Jemielniak 2014: 11). En comparación con la Enciclopedia Británica, sus artículos en inglés son 10 veces más largos (Medelyan et al. 2009: 717) y un controvertido estudio publicado en la revista *Nature* destacó que su fiabilidad es similar (Giles 2005). Tal y como resumen Mesgari et al. (2015: 10), hay numerosos estudios respecto a su fiabilidad en diversos dominios, cuyo resultado es que la Wikipedia es, por lo general, una fuente fiable de información en diversos dominios, excepto en ciencias de la salud donde los resultados son mixtos.

Como exponen Medelyan et al. (2009), la Wikipedia se utiliza ampliamente en procesamiento del lenguaje natural, recuperación y extracción de información y construcción de ontologías. Además, los trabajos de Vivaldi y Rodríguez (2010a; 2010b; 2011; 2012) demuestran los buenos resultados del empleo de la Wikipedia como fuente para la extracción terminológica.

Para este trabajo, se eligieron manualmente los artículos para cada dominio a partir de las categorías y subcategorías correspondientes en la Wikipedia. El subcorpus ENV no incluye artículos de la Wikipedia porque la cantidad de artículos con un enfoque general sobre el medio ambiente es muy reducida.

#### 4.1.2.2 Corpus PANACEA medioambiental

El corpus monolingüe de inglés medioambiental PANACEA[67] lo obtuvieron de manera automatizada investigadores de varios países europeos en el marco del proyecto PANACEA (*Platform for Automatic, Normalized Annotation and Cost-Effective Acquisition of Language Resources for Human Language Technologies*)[68], llevado a cabo como parte del Séptimo Programa Marco de investigación de la Comisión Europea.

[67] Disponible en: <http://catalog.elra.info/product_info.php?products_id=1184>.

Se trata de un corpus exclusivamente en inglés, adquirido automáticamente de Internet con textos clasificados como pertenecientes al dominio del medio ambiente. Contiene en total 50.541.538 palabras repartidas en 28.071 documentos obtenidos a partir de 3.121 sitios web.

Dadas las características del corpus, parte de los documentos que componen el corpus son de tipo divulgativo o de escasa fiabilidad. Sin embargo, su enorme tamaño compensa esas deficiencias para algunas tareas. En este trabajo, este corpus solo se empleó en la extracción de hiperónimos para los conceptos que se definieron en §5.4 porque tras una extracción automática de concordancias, se realizó un filtrado manual.

### 4.1.3 El léxico científico transdisciplinar

Como ya se indicó en §3.1, la distinción entre palabra y término es una cuestión controvertida. En este sentido, Drouin (2010b) defiende que a caballo entre las unidades léxicas generales y las unidades terminológicas existen unas unidades pertenecientes a lo que él denomina el *léxico científico transdisciplinar* (LCT) (Drouin 2007; Drouin 2010b):

> [O]n considère que le LST transcende les domaines de spécialité et présente un noyau lexical commun significatif entre les disciplines. Le lexique scientifique transdisciplinaire n'est pas saillant dans les textes scientifiques dans la mesure où, contrairement à la terminologie, il se rencontre également dans la langue commune. Par contre, il est au cœur même de l'argumentation et de la structuration du discours et de la pensée scientifique. (Drouin 2007: 45).

El LST de Drouin tiene como antecedentes principales el VGOS (*Vocabulaire général d'orientation scientifique*) de Phal (1971) en francés y

[68] Para más información sobre PANACEA, se puede visitar la página web del proyecto: <http://www.panacea-lr.eu/>.

la AWL (*Academic Word List*) de Coxhead (2000) en inglés. Sin embargo, del VGOS le distingue el hecho de incluir todas las disciplinas científicas, incluidas las ciencias humanas (Drouin 2007: 45). Por su parte, la diferencia de AWL es que el foco de este es la lengua académica (como su propio nombre indica) con vistas a su enseñanza del inglés (Drouin 2007: 46) Asimismo, un proyecto de características similares al del LCT, aunque solamente en francés, es el del *Lexique transdisciplinaire des écrits scientifiques* de Tutin (2007a; 2007b).

El LCT es un proyecto que se encuentra aún en desarrollo y se centra en el francés y el inglés. Para obtener la primera versión de la LCT en inglés, Drouin (2010b) se basó en dos factores: la distribución en un corpus transdisciplinar y la especificidad de las unidades léxicas en comparación con un corpus de referencia.

El corpus transdisciplinar estaba compuesto por artículos científicos y tesis doctorales pertenecientes a los dominios de antropología, química, informática, ingeniería, geografía, historia, derecho, física y psicología. En total el corpus contenía aproximadamente 4 millones de palabras. Por su parte, el corpus de referencia era una parte del British National Corpus.

Para que una unidad léxica fuera considerada parte del LCT, debía tener un alto nivel de especificidad en el corpus transdisciplinar y aparecer en todos los dominios. Las 10 unidades del LCT más frecuentes en inglés fueron: *model*, *analysis*, *function*, *phase*, *system*, *structure*, *method*, *state*, *design*, *interaction* (Drouin 2010b: 300). La lista actual completa en inglés está compuesta por 1517 unidades[69].

En este trabajo, haremos uso de la lista de unidades del LCT en inglés para descartarlas de nuestro análisis. Aunque sería necesario un análisis sistemático, las unidades LCT en general no experimentan un nivel suficiente de variación contextual como para que sea necesaria una definición flexible.

[69] Puede consultarse en: <http://olst.ling.umontreal.ca/lexitrans/nomenclature.php>.

El filtrado de las unidades LCT se hizo de manera automática, por lo que se descartaron con ello algunas unidades que, en nuestra opinión, sí podrían necesitar de una definición flexible en el medio ambiente como *water*, *energy* o *air*. Sin embargo, este tipo de casos son aislados.

## 4.2 MÉTODOS

### 4.2.1 Selección de conceptos de análisis

El primer paso de nuestro estudio empírico consistió en la obtención de una lista de conceptos que fuesen relevantes simultáneamente en varios dominios contextuales. Para ello, se realizó una extracción terminológica con TermoStat Web 3.0 para cada uno de los 14 subcorpus de dominio contextual del corpus MULTI. La extracción se limitó a términos simples nominales. Cada una de las 14 listas fueron exportadas en formato TXT mediante la función correspondiente en TermoStat Web 3.0. Las listas exportadas de TermoStat contienen, además de los términos candidatos, el número de ocurrencias en el corpus y su puntuación de especificidad, entre otros datos.

A continuación, se realizó una comparación automática de las listas mediante un *script* PHP[70]. La lista que devuelve el *script* organiza los resultados de la comparación en varias columnas: 1) el término candidato; 2) el número de dominios en el que aparece, 3) indicación de si el término está contenido en la lista del LCT; y 4) una columna para cada dominio contextual que contiene el número de ocurrencias en ese dominio.

El *script* permitía asimismo establecer un umbral mínimo de ocurrencias que un término debía tener en un dominio dado para ser incluido como presente en dicho dominio en la lista. Gracias a ello, se pudo extraer numerosas listas con distintos umbrales para establecer el más adecuado para nuestros fines (a saber, 64 ocurrencias).

---

[70] El *script* PHP lo desarrolló expresamente para este trabajo el investigador Benoît Robichaud, del Observatoire de linguistique Sens-Texte de la Universidad de Montreal.

La lista final de trabajo estaba formada por los términos que aparecían en tres o más subcorpus de dominios contextuales con una frecuencia superior a 64 ocurrencias. De manera automática se descartaron las unidades pertenecientes al LCT. Manualmente, se excluyeron de la lista las abreviaciones y acrónimos, y los términos polisémicos. Asimismo, se recalculó el número de ocurrencias de algunos términos con variantes ortográficas.

Todos los conceptos representados por los términos de la lista fueron analizados a partir del conocimiento extraído mediante las técnicas que se exponen en el siguiente apartado. Asimismo, haciendo uso de estas mismas técnicas, se procedió a desarrollar la definición flexible de dos conceptos representativos de los fenómenos observados en el análisis de los conceptos.

### 4.2.2 Extracción del conocimiento

Desde un enfoque descriptivo de la terminología, como es nuestro caso, la metodología empleada para extraer el conocimiento cobra una gran importancia. Dado que nuestra base teórica fundamental es la TBM y los resultados de este trabajo están orientados a su incorporación a EcoLexicon, se siguieron los principios metodológicos que emanan de la TBM y que se aplican actualmente a EcoLexicon. No obstante, añadimos algunos elementos nuevos a los que haremos referencia más adelante.

Como ya se indicó en §2.2.2.4, el rasgo principal de la metodología de extracción del conocimiento de la TBM es su combinación tanto de un enfoque *top-down* como *bottom-up* (Faber, León Araúz y Prieto Velasco 2009: 6). El enfoque *top-down* de la TBM incluye tres elementos principales:

- Extracción de conocimiento a partir de otros recursos terminológicos (bases de datos terminológicas, diccionarios especializados, glosarios, etc.)
- Extracción de conocimiento a partir de otro tipo de obras de referencia especializada (enciclopedias, manuales, libros de texto, etc.)
- Consulta a expertos en el dominio correspondiente[71].

Por su parte, el enfoque *bottom-up* incluye el análisis de corpus, que como veremos en §4.2.2.3, puede tomar distintas formas. Asimismo, cabe destacar que la misma fuente puede emplearse desde un punto de vista *top-down* o *bottom-up*. Por ejemplo, las obras de referencia especializadas pueden incluirse en un corpus y, en ese caso, el enfoque adoptado pasa a ser de tipo *bottom-up*.

#### 4.2.2.1 Análisis de definiciones de otros recursos

La extracción de conocimiento a partir de otros recursos terminológicos es el método *top-down* principal que se emplea en EcoLexicon. Su principal modalidad es la del análisis de definiciones terminológicas de otros recursos, aunque no es la única, ya que también es posible extraer conocimiento a partir de, por ejemplo, relaciones conceptuales ya codificadas en otra base de datos o cualquier otro elemento de un recurso que represente conocimiento conceptual.

La consulta de otros recursos es una práctica común en lexicografía, pues, como indica Čermák (2003: 19), los lexicógrafos siempre consultan otros diccionarios cuando trabajan en un proyecto lexicográfico. No existen motivos para pensar que la situación no sea similar entre los terminólogos. El auge de la explotación de la información contenida en las definiciones

[71] En este trabajo, no se recurrió a la consulta a expertos por el carácter altamente multidisciplinar de esta investigación, ya que se habría requerido contar con un elevado número de expertos, lo cual resultó inviable. Asimismo, los expertos no son siempre capaces de proporcionar la información que necesita el terminólogo (León Araúz 2009: 88) y cada experto posee una conceptualización propia de su dominio (Hameed, Sleeman y Preece 2002: 270) que no se ajusta necesariamente a la convención dentro de esa área de conocimiento.

(u otros elementos) de otros recursos para crear o enriquecer uno propio llegó a partir de la de década de 1980 con la aparición de los diccionarios en formato electrónico (O'Hara 2005: 35). Más de tres décadas de desarrollo y las múltiples aplicaciones que se le han dado han demostrado la enorme utilidad de recurrir al conocimiento ya representado en otros recursos.

Inspirada por el modelo lexemático-funcional (Martín Mingorance 1984; Martín Mingorance 1995; Faber y Mairal Usón 1999), la TBM concibe las definiciones terminológicas como mini-representaciones de conocimiento. Faber y Mairal Usón (1999: 88-91) defienden la utilidad de analizar definiciones de otros recursos de la unidad léxica que se pretende representar, pues, al segmentarlas y contrastar los constituyentes de manera sistemática, podemos obtener información semántica estructurada. Aunque los autores son conscientes de las limitaciones de este método —ya que las definiciones no son siempre satisfactorias— afirman que no se debe dejar de lado, porque si se busca información semántica, los diccionarios son el primer lugar al que hay que acudir:

> [D]ictionaries are an extremely valuable resource in any type of lexical research, and not using them because of certain understandable limitations would be a little like throwing the baby out with the bath water. They are undoubtedly the first place one must look into in order to find information about meaning. (Faber y Mairal Usón 1999: 91)

A pesar de que las definiciones no siempre están basadas en estudios de corpus, su sistematicidad es a menudo deficiente, padecen habitualmente de circularidad y algunos autores simplemente copian la definición de otros recursos (León Araúz 2009: 95; Faber et al. 2007: 41), los rasgos más repetidos tienen una gran probabilidad de ser definicionales (Faber 2002).

Además de la búsqueda individual en determinados recursos terminológicos, en este trabajo, se utilizaron cuatro sitios web que permiten consultar las definiciones de numerosos recursos especializados al mismo tiempo:

- MetaGlossary[72]: Al introducir un término en inglés, este sitio devuelve definiciones de cientos de glosarios en Internet. La página de resultados agrupa las definiciones según palabras claves, lo cual a menudo corresponde con sentidos polisémicos del término. Cada definición viene acompañada de la URL de la fuente, lo cual permite comprobar su fiabilidad.
- Oxford Reference[73]: Este portal permite consultar con una sola búsqueda cerca de 300 obras de referencia de la editorial Oxford University Press.
- EcoRessources[74]: Este sitio web ha sido desarrollado por el Observatoire de linguistique Sens-Texte de la Universidad de Montreal. Al buscar un término, muestra la definición contenida en diversos recursos terminológicos sobre el medio ambiente como EcoLexicon, DiCoEnviro o las bases de datos terminológicas de la EPA, la EIONET o la FAO.
- OneLook[75]: Este sitio web permite interrogar con una única búsqueda más de 1000 diccionarios o glosarios, muchos de ellos especializados. A diferencia de los tres anteriores, presenta el inconveniente de no mostrar en una única página las definiciones, sino que es necesario visitar cada uno de los enlaces que OneLook proporciona.

#### 4.2.2.2 Consulta de obras de referencia

La extracción de conocimiento a partir de otro tipo de obras de referencia especializada permite al terminólogo adquirir una visión de conjunto del área de conocimiento sobre la que esté trabajando. Cumple una función muy importante en una fase inicial de familiarización y en la resolución de dudas posteriormente. Los textos sobre dominios especializados codifican sistemas conceptuales parciales que el terminólogo puede utilizar como punto de partida. En este sentido, las obras de referencia como enciclopedias, manuales o libros de texto, dado que van dirigidos a un público lego o semiespecializado, suelen codificar de manera más completa los sistemas conceptuales, pues se asume que el receptor no tiene ese conocimiento previo que sí posee el experto. En este trabajo, se hizo un uso extensivo de variedad de obras de referencia a lo largo de todas las etapas.

[72] Disponible en: <http://www.metaglossary.com>.

[73] Disponible en: <http://www.oxfordreference.com>.

[74] Disponible en: <http://termeco.info/ecoressources>.

[75] Disponible en: <http://www.onelook.com>.

#### 4.2.2.3 El análisis de corpus

Todo trabajo terminológico se debe basar en gran medida en el análisis de corpus por dos motivos principales (Bourigault y Slodzian 1999: 30):

- Las aplicaciones del trabajo terminológico son casi siempre de tipo textual (traducción, redacción, etc.), así que para que el resultado sea aplicable a nuevos textos, la fuente debe provenir de otros textos.
- En los textos producidos o utilizados por la comunidad de expertos es donde se expresan y se hacen accesibles los conocimientos compartidos por esa comunidad.

Al igual que las definiciones de otros recursos, el análisis de corpus permite extraer información conceptual que posteriormente puede clasificarse y analizarse para caracterizar los conceptos dentro su marco de activación (León Araúz, Faber y Montero Martínez 2012: 106). Además, si los textos incluidos en el corpus son fiables (Buendía y Ureña 2009), la información extraída del corpus se puede utilizar para verificar la obtenida a través de otros medios (Faber, López Rodríguez y Tercedor 2001).

La manera más básica de utilizar un corpus es mediante la lectura manual de las líneas de concordancia del término sobre el que se esté trabajando. Sin embargo, esto consume muchísimo tiempo, por ello existen métodos para analizar y extraer la información de los corpus de manera más eficiente. En este trabajo, además de los *word-sketches* de SketchEngine, se utilizaron patrones de conocimiento y contextónimos.

#### *4.2.2.3.1 Los patrones de conocimiento*

Con el fin de extraer información definicional de un corpus, un enfoque común es la búsqueda de contextos ricos en conocimiento (Meyer 2001). Un contexto rico en conocimiento es un contexto que indica al menos un elemento de conocimiento del dominio que podría ser útil para el análisis conceptual (Meyer 2001: 281). Para encontrar contextos ricos en conocimiento en los corpus, recurre al uso de los patrones de conocimiento, que consisten en patrones lingüísticos y paralingüísticos que transmiten una relación semántica específica (Meyer 2001: 290). En el marco de EcoLexicon, se han utilizado los patrones de conocimiento para analizar corpus tanto de manera manual (Tercedor y López Rodríguez 2008: 171-178; Faber, León Araúz y Reimerink 2011: 376) como (semi)automatizada (León Araúz y Faber 2012; León Araúz 2014; León Araúz y Reimerink 2010).

Los patrones de conocimiento más estudiados son los que transmiten una relación hiperonímica (Auger y Barrière 2008: 4). Algunos ejemplos de tales patrones en inglés son *comprise(s), consist(s), define(s), denote(s), designate(s), is/are, is/are called, is/are defined as, is/are known as* (Pearson 1998: 140). Los patrones de conocimiento hiperonímicos identifican lo que se ha denominado *contextos ricos en conocimiento definitorios* (Meyer 2001: 283), *expositivos definitorios* (Pearson 1998: 135) o, simplemente, *contextos definitorios* (Malaisé, Zweigenbaum y Bachimont 2005; Sierra 2009; Alarcón Martínez 2009). Los contextos definitorios poseen los mismos componentes que una definición tradicional: definiéndum, genus y, ocasionalmente, differentiae.

Por otro lado, Meyer (2001: 283) denomina *contextos ricos en conocimiento explicativos* a cualquier otro tipo de contexto rico de conocimiento. Este tipo de contextos pueden ser de tipo causal, funcional, meronímico, etc. Para extraer cada tipo es necesario emplear patrones de conocimiento específicos. Por ejemplo, *cause(s)*, *produce(s)* y *generate(s)* son patrones de conocimiento causales (León Araúz y Faber 2012).

En este trabajo, tan solo hicimos uso de patrones de conocimiento de tipo hiperonímico. Se emplearon para la extracción de candidatos a genus. Dicha extracción se realizó de manera automática con SketchEngine mediante la adición de nuevos tipos de *word-sketches* basados en patrones de conocimiento a la gramática que incluye SketchEngine por defecto en inglés.

A continuación, se reproducen los patrones empleados[76] tal y como se codificaron para su uso en este trabajo, acompañados de una explicación de cada elemento y de ejemplos extraídos del corpus MULTI.

<table>
<tr><td colspan="2">2:"N.*" [word="and"]? [tag!="V.*"]{0,5} [lemma="be"] "RB.*|DT.*"{0,4} "JJ.*"? [lemma="type|kind|example|group|class"] [word="of"] [tag!="V.*"]{0,3} 1:"N.*" within <s/></td></tr>
<tr><td>2:"N.*"</td><td>El hipónimo es un sustantivo.</td></tr>
<tr><td>[word="and"]?</td><td>La palabra <i>and</i> opcional.</td></tr>
<tr><td>[tag !="V.*"]{0,5}</td><td>Entre 0 y 5 palabras que no sean verbos.</td></tr>
<tr><td>[lemma="be"]</td><td>Una palabra cuyo lema sea <i>be</i>.</td></tr>
<tr><td>"RB.*|DT.*"{0,4}</td><td>Entre 0 y 4 adverbios o determinantes.</td></tr>
<tr><td>"JJ.*"?</td><td>Un adjetivo opcional.</td></tr>
<tr><td>[lemma="type|kind|example|group|class"]</td><td>Una palabra cuyo lema sea <i>type</i>, <i>kind</i>, <i>example</i>, <i>group</i> o <i>class</i>.</td></tr>
<tr><td>[word="of"]</td><td>La palabra <i>of</i>.</td></tr>
<tr><td>[tag !="V.*"]{0,3}</td><td>Entre 0 y 3 palabras que no sean verbos.</td></tr>
<tr><td>1:"N.*"</td><td>El hiperónimo es un sustantivo.</td></tr>
<tr><td>within <s/></td><td>El patrón debe ocurrir en una única oración.</td></tr>
<tr><td colspan="2">Ejemplos en el corpus:<br>on the same plant. Corn is an example of a monoecious plant. The corn tassel is<br>several nutrients. Lime is a special type of fertilizer that can apply Ca or<br>from biogenic sources. Terpenes are a class of hydrocarbons that evaporate from<br>fortified with vitamins. Lipids are a group of organic compounds that are insoluble<br>photochemical smog. Smog is a kind of air pollution. Classic smog results</td></tr>
</table>

**Tabla 19. Primer patrón de conocimiento hiperonímico codificado para SketchEngine**

[76] Los diez patrones son una ampliación y refinamiento de los tres patrones incluidos en Buendía, Sánchez Cárdenas y León Araúz (2014).

<table>
<tr><td colspan="2">[lemma="type|kind|example|class"] [word="of"] [tag!="V.*|IN.*"]{0,5} 1:"N.*" []{0,3} [lemma="include|be"] "DT.*"? [tag!="V.*|IN.*"]{0,5} [word="and"]? 2:"N.*" within <s/></td></tr>
<tr><td>[lemma="type|kind|example|clas s"]</td><td>Una palabra cuyo lema sea <i>type</i>, <i>kind</i>, <i>example</i>,o <i>class</i>.</td></tr>
<tr><td>[word="of"]</td><td>La palabra <i>of.</i></td></tr>
<tr><td>[tag!="V.*|IN.*"]{0,5}</td><td>Entre 0 y 5 palabras que no sean verbos ni preposiciones.</td></tr>
<tr><td>1:"N.*"</td><td>El hiperónimo es un sustantivo.</td></tr>
<tr><td>[]{0,3}</td><td>Entre 0 y 3 palabras cualesquiera.</td></tr>
<tr><td>[lemma="include|be"]</td><td>Una palabra cuyo lema sea <i>include</i> o <i>be.</i></td></tr>
<tr><td>"DT.*"?</td><td>Un determinante opcional.</td></tr>
<tr><td>[tag!="V.*|IN.*"]{0,5}</td><td>Entre 0 y 5 palabras que no sean verbos ni preposiciones.</td></tr>
<tr><td>[word="and"]?</td><td>La palabra <i>and</i> opcional.</td></tr>
<tr><td>2:"N.*"</td><td>El hipónimo es un sustantivo.</td></tr>
<tr><td>within <s/></td><td>El patrón debe ocurrir en una única oración.</td></tr>
<tr><td colspan="2">Ejemplos en el corpus:<br>climacteric crops. Another class of phytohormones is the jasmonates, first isolated from<br>surface charge. Some examples of these substances are rubber, plastic, glass, and<br>Related, more specific types of toponym include hydronym for a body of water<br>The two most common types of folds are anticlines and synclines (FIGURE<br>highly demanded. Some examples of basic chemicals are: ethylene, benzene, chlorine</td></tr>
</table>

**Tabla 20. Segundo patrón de conocimiento hiperonímico codificado para SketchEngine**

<table>
<tr><td colspan="2">1:"N.*" [tag!="V.*"]{0,5} [word="such"] [word="as"] "DT.*"? [tag!="V.*"]{0,5} [word="and"]? 2:"N.*" within <s/></td></tr>
<tr><td>1:"N.*"</td><td>El hiperónimo es un sustantivo.</td></tr>
<tr><td>[tag!="V.*"]{0,5}</td><td>Entre 0 y 5 palabras que no sean verbos.</td></tr>
<tr><td>[word="such"]</td><td>La palabra <i>such</i>.</td></tr>
<tr><td>[word="as"]</td><td>La palabra <i>as</i>.</td></tr>
<tr><td>"DT.*"?</td><td>Un determinante opcional.</td></tr>
<tr><td>[tag!="V.*"]{0,5}</td><td>Entre 0 y 5 palabras que no sean verbos.</td></tr>
<tr><td>[word="and"]?</td><td>La palabra <i>and</i> opcional.</td></tr>
<tr><td>2:"N.*"</td><td>El hipónimo es un sustantivo.</td></tr>
<tr><td>within <s/></td><td>El patrón debe ocurrir en una única oración.</td></tr>
<tr><td colspan="2">Ejemplos en el corpus:<br>ozone and toxic heavy metals such as lead , cadmium, and copper<br>phosphates; due to human activities such as agriculture and discharge from<br>describe the various types of pesticides, such as insecticides and herbicides. Any<br>linked to a variety of cancers , such as leukemia and cancers of the<br>which gobble up the poisons such as petroleum and other hydrocarbons</td></tr>
</table>

**Tabla 21. Tercer patrón de conocimiento hiperonímico codificado para SketchEngine**

<table>
<tr><td colspan="2">1:"N.*" [tag!="V.*"]{0,5} [word="including"] [tag!="V.*"]{0,5} [word="and"]? 2:"N.*" within <s/></td></tr>
<tr><td>1:"N.*"</td><td>El hiperónimo es un sustantivo.</td></tr>
<tr><td>[tag!="V.*"]{0,5}</td><td>Entre 0 y 5 palabras que no sean verbos.</td></tr>
<tr><td>[word="including"]</td><td>La palabra *including*.</td></tr>
<tr><td>[tag !="V.*"]{0,5}</td><td>Entre 0 y 5 palabras que no sean verbos.</td></tr>
<tr><td>[word="and"]?</td><td>La palabra *and* opcional.</td></tr>
<tr><td>2:"N.*"</td><td>El hipónimo es un sustantivo.</td></tr>
<tr><td>within <s/></td><td>El patrón debe ocurrir en una única oración.</td></tr>
<tr><td colspan="2">Ejemplos en el corpus:<br>from the milk of other mammals , including sheep , goats, buffalo, and<br>been found in other foods , including barley, cassava, corn, rice<br>steam heated using fossil fuel (including coal, gas and oil) or nuclear<br>southwest has desert plants , including yucci and cacti. The cultivated<br>mounts of many other elements , including titanium and manganese, and</td></tr>
</table>

**Tabla 22. Cuarto patrón de conocimiento hiperonímico codificado para SketchEngine**

<table>
<tr><td colspan="2">2:"N.*" [tag !="V.*"]{0,5} [word="and"] [word="other"] [tag !="V.*"]{0,5} 1:[tag="N.*" & lemma!="type|sort|kind|example|group|part"] within <s/></td></tr>
<tr><td>2:"N.*"</td><td>El hipónimo es un sustantivo.</td></tr>
<tr><td>[tag!="V.*"]{0,5}</td><td>Entre 0 y 5 palabras que no sean verbos.</td></tr>
<tr><td>[word="and"]</td><td>La palabra *and*.</td></tr>
<tr><td>[word="other"]</td><td>La palabra *other*.</td></tr>
<tr><td>[tag!="V.*"]{0,5}</td><td>Entre 0 y 5 palabras que no sean verbos.</td></tr>
<tr><td>1:[tag="N.*" & lemma!="type|sort|kind|example|group|part"]</td><td>El hiperónimo es un sustantivo cuyo lema no sea *type*, *sort*, *kind*, *example*, *group* ni *part*.</td></tr>
<tr><td>within <s/></td><td>El patrón debe ocurrir en una única oración.</td></tr>
<tr><td colspan="2">Ejemplos en el corpus:<br>iron, manganese and sulfur, and other chemical pollutants such as fertilisers<br>acids, and sugars) to fermentation and other metabolic processes leading to the formation<br>such as the telescope, sextant and other devices that use telescopes<br>normally grow. Also, insects and other animals might eat the plants<br>of the definition, chemists and other scientists use the term "chemical</td></tr>
</table>

**Tabla 23. Quinto patrón de conocimiento hiperonímico codificado para SketchEngine**

<table>
<tr><td colspan="2">2:"N.*" [word="and"]? [tag!="V.*"]{0,5} "MD"? [lemma="be"] [word!="not"]? [word="defined|classified|categori.ed"] [word="as"] "DT.*"? [lemma="type|kind|example|class"]? [word="of"]? [tag!="V.*"]{0,2} 1:[tag="N.*" & lemma!="type|sort|kind|example|group|class"] within <s/></td></tr>
<tr><td>2:"N.*"</td><td>El hipónimo es un sustantivo.</td></tr>
<tr><td>[word="and"]?</td><td>La palabra <i>and</i> opcional.</td></tr>
<tr><td>[tag!="V.*"]{0,5}</td><td>Entre 0 y 5 palabras que no sean verbos.</td></tr>
<tr><td>"MD"?</td><td>Un verbo modal opcional.</td></tr>
<tr><td>[lemma="be"]</td><td>El lema <i>be</i>.</td></tr>
<tr><td>[word!="not"]?</td><td>Una palabra opcional que no sea <i>not</i>.</td></tr>
<tr><td>[word="defined|classified|categori.ed"]</td><td>La palabra <i>defined</i>, <i>classified</i> o una palabra que comience por <i>categori</i> acabe en <i>ed</i> y en medio solo tenga un carácter.</td></tr>
<tr><td>[word="as"]</td><td>La palabra <i>as</i>.</td></tr>
<tr><td>"DT.*"?</td><td>Un determinante opcional.</td></tr>
<tr><td>[lemma="type|kind|example|class"]?</td><td>Una palabra opcional cuyo lema sea <i>type</i>, <i>kind</i>, <i>example</i> o <i>class</i>.</td></tr>
<tr><td>[word="of"]?</td><td>La palabra <i>of</i> opcional.</td></tr>
<tr><td>[tag!="V.*"]{0,2}</td><td>Entre 0 y 2 palabras que no sean verbos.</td></tr>
<tr><td>1:[tag="N.*" & lemma!="type|sort|kind|example|group|class"]</td><td>El hiperónimo es un sustantivo cuyo lema no sea <i>type</i>, <i>sort</i>, <i>kind</i>, <i>example</i>, <i>group</i> ni <i>class</i>.</td></tr>
<tr><td>within <s/></td><td>El patrón debe ocurrir en una única oración.</td></tr>
<tr><td colspan="2">Ejemplos en el corpus:<br>whereas vitamins and minerals are classified as micronutrients that are required in<br>to whether eye irritation should be categorized as a significant health effect since no other<br>such as carbon, nitrogen , and oxygen, are classified as non-metals. Non-metals lack the<br>Species Act. Critical habitat is defined as any place where a threatened<br>natural coordinates. Shear is defined as the rate of change of 1 The</td></tr>
</table>

**Tabla 24. Sexto patrón de conocimiento hiperonímico codificado para SketchEngine**

<table>
<tr><td colspan="2">2:"N.*" [word="and"]? [tag!="V.*"]{0,5} "MD"? [lemma="be"] [word!="not"]{0,1} [word="considered"] [word="to"]? [word="be"]? "DT.*"? [tag!="V.*|IN.*" | lemma!="part"]? [lemma="type|kind|example|class"]? [word="of"]? [tag!="V.*|IN.*"]{0,3} 1:"N.*" within <s/></td></tr>
<tr><td>2:"N.*"</td><td>El hipónimo es un sustantivo.</td></tr>
<tr><td>[word="and"]?</td><td>La palabra <i>and</i> opcional.</td></tr>
<tr><td>[tag!="V.*"]{0,5}</td><td>Entre 0 y 5 palabras que no sean verbos.</td></tr>
<tr><td>"MD"?</td><td>Un verbo modal opcional.</td></tr>
</table>

<table>
<tr><td>[lemma="be"]</td><td>El lema *be*.</td></tr>
<tr><td>[word!="not"]{0,1}</td><td>Entre 0 y 1 palabras que no sean *not*.</td></tr>
<tr><td>[word="considered"]</td><td>La palabra *considered*.</td></tr>
<tr><td>[word="to"]?</td><td>La palabra *to* opcional.</td></tr>
<tr><td>[word="be"]?</td><td>La palabra *be* opcional.</td></tr>
<tr><td>[tag="DT.*"]?</td><td>Un determinante opcional.</td></tr>
<tr><td>[tag!="V.*|IN.*" | lemma!="part"]?</td><td>Una palabra que no sea verbo, ni preposición ni cuyo lema sea *part*.</td></tr>
<tr><td>[lemma="type|kind|example|class"]?</td><td>Una palabra opcional cuyo lema sea *type*, *kind*, *example* o *class*.</td></tr>
<tr><td>[word="of"]?</td><td>La palabra *of* opcional.</td></tr>
<tr><td>[tag!="V.*|IN.*"]{0,3}</td><td>Entre 0 y 3 palabras que no sean ni verbos ni preposiciones.</td></tr>
<tr><td>1:"N.*"</td><td>El hiperónimo es un sustantivo.</td></tr>
<tr><td>within <s/></td><td>El patrón debe ocurrir en una única oración.</td></tr>
<tr><td colspan="2">Ejemplos en el corpus:<br>oxidizing. Photochemical smog is therefore considered to be a problem of modern<br>irritation and headaches. Carbon dioxide is also considered a greenhouse gas. Water<br>Non-rechargeable batteries are considered hazardous waste and should only be kilometers<br>Most groundwater is considered a nonrenewable resource because it has taken recharging<br>The electric car is considered a zero emissions vehicle. Even if power plant</td></tr>
</table>

**Tabla 25. Séptimo patrón de conocimiento hiperonímico codificado para SketchEngine**

<table>
<tr><td colspan="2">2:"N.*" [lemma="be"] "DT.*"? 1:[tag="N.*" & lemma!="type|sort|kind|example|group|part"] within <s/></td></tr>
<tr><td>2:"N.*"</td><td>El hipónimo es un sustantivo.</td></tr>
<tr><td>[lemma="be"]</td><td>El lema *be*.</td></tr>
<tr><td>"DT.*"?</td><td>Un determinante opcional.</td></tr>
<tr><td>1:[tag="N.*" & lemma!="type|sort|kind|example|group|part"]</td><td>El hiperónimo es un sustantivo cuyo lema no es *type*, *sort*, *kind*, *example*, *group* ni *part*.</td></tr>
<tr><td>within <s/></td><td>El patrón debe ocurrir en una única oración.</td></tr>
<tr><td colspan="2">Ejemplos en el corpus:<br>and the Family Farm. Agrarianism is the belief that farming is the<br>experience nitrification, as ammonia is a nutrient for bacterial growth<br>part of a chromosome. Mutations are changes in the DNA sequence<br>communities. For instance, foxes are omnivores whose diet includes<br>energy. In physics, energy is a property of objects which can<br>two solid surfaces. Lubrication is a technique employed to reduce</td></tr>
</table>

**Tabla 26. Octavo patrón de conocimiento hiperonímico codificado para SketchEngine**

<table>
<tr><td colspan="2">[lemma="define"] "DT.*"? [word="and"]? [tag!="V.*"]{0,3} 2:"N.*" [word="as"] "DT.*"? [lemma="type|kind|example|class"]? [word="of"]? [tag!="V.*"]{0,2} 1:[tag="N.*" & lemma!="type|sort|kind|example|group"] within <s/></td></tr>
<tr><td>[lemma="define"]</td><td>Una palabra cuyo lema sea <i>define</i>.</td></tr>
<tr><td>"DT.*"?</td><td>Un determinante opcional.</td></tr>
<tr><td>[word="and"]?</td><td>La palabra <i>and</i> opcional.</td></tr>
<tr><td>[tag!="V.*"]{0,3}</td><td>Entre 0 y 3 palabras que no sean verbos.</td></tr>
<tr><td>2:"N.*"</td><td>El hipónimo es un sustantivo.</td></tr>
<tr><td>[word="as"]</td><td>La palabra <i>as</i>.</td></tr>
<tr><td>"DT.*"?</td><td>Un determinante opcional.</td></tr>
<tr><td>[lemma="type|kind|example|class"]?</td><td>Una palabra opcional cuyo lema sea <i>type</i>, <i>kind</i>, <i>example</i> o <i>class</i>.</td></tr>
<tr><td>[word="of"]?</td><td>La palabra <i>of</i> opcional.</td></tr>
<tr><td>[tag !="V.*"]{0,2}</td><td>Entre 0 y 2 palabras que no sean verbos.</td></tr>
<tr><td>1:[tag="N.*"& lemma!="type|sort|kind|example|group"]</td><td>El hiperónimo es un sustantivo cuyo lema no es <i>type</i>, <i>sort</i>, <i>kind</i>, <i>example</i> ni <i>group</i>.</td></tr>
<tr><td>within <s/></td><td>El patrón debe ocurrir en una única oración.</td></tr>
<tr><td colspan="2">Ejemplos en el corpus:<br>EU) Floods Directive defines a flood as a covering by water of land not<br>Ernst Mayr in 1968 defined beta taxonomy as the classification of ranks higher than<br>Hydraulic Engineering defines hydrostatics as the study of fluids at rest.<br>Organization (FAO) has defined pesticide as: any substance or mixture of substances<br>an apple. Finally, we define pressure as force applied per unit area</td></tr>
</table>

**Tabla 27. Noveno patrón de conocimiento hiperonímico codificado para SketchEngine**

<table>
<tr><td colspan="2">2:"N.*" [lemma="refer"] [word="to"] "DT.*"? [lemma="type|kind|example|class"]? [word="of"]? [word="and"]? [tag!="V.*"]{0,3} 1:[tag="N.*" & lemma!="type|sort|kind|example|group|part"] within <s/></td></tr>
<tr><td>2:"N.*"</td><td>El hipónimo es un sustantivo.</td></tr>
<tr><td>[lemma="refer"]</td><td>El lema <i>refer</i>.</td></tr>
<tr><td>[word="to"]</td><td>La palabra <i>to</i>.</td></tr>
<tr><td>"DT.*"?</td><td>Un determinante opcional.</td></tr>
<tr><td>[lemma="type|kind|example|class"]?</td><td>Una palabra opcional cuyo lema sea <i>type</i>, <i>kind</i>, <i>example</i> o <i>class</i>.</td></tr>
<tr><td>[word="of"]?</td><td>La palabra <i>of</i> opcional.</td></tr>
<tr><td>[word="and"]?</td><td>La palabra <i>and</i> opcional.</td></tr>
<tr><td>[tag!="V.*"]{0,3}</td><td>Entre 0 y 3 palabras que no sean verbos.</td></tr>
<tr><td>1:[tag="N.*" &</td><td>El hiperónimo es un sustantivo</td></tr>
</table>

<table>
<tr><td>lemma!="type|sort|kind|example|group|part"]</td><td>cuyo lema no es <i>type</i>, <i>sort</i>, <i>kind</i>, <i>example</i>, <i>group</i> ni <i>part</i>.</td></tr>
<tr><td>within <s/></td><td>El patrón debe ocurrir en una única oración.</td></tr>
<tr><td colspan="2">Ejemplos en el corpus:<br>cause degradation. Porosity refers to the air-holding capacity of the soil. See also<br>reducing pH.[12] Soil fertility refers to the ability of a soil related to<br>original materials.[2] Solidification refers to the physical changes in the contaminated<br>at the global scale. Desertification refers to the process of becoming desert<br>Transformation. Decomposition refers to the chemical and biochemical reactions</td></tr>
</table>

**Tabla 28. Décimo patrón de conocimiento hiperonímico codificado para SketchEngine**

#### *4.2.2.3.2 La extracción de contextónimos*

Los contextónimos (Ji, Ploux y Wehrli 2003; Ji y Ploux 2003) de una unidad léxica son aquellas otras unidades léxicas con las que tiende a coocurrir en contextos lingüísticos. El contexto lingüístico puede abarcar desde una ventana de pocas palabras a varios párrafos. La relación de contextonimia no es ni transitiva (p. ej, que *silla* sea contextónimo de *mesa* y que *respaldo* sea contextónimo de *silla*, no implica que *respaldo* sea contextónimo de *mesa*) ni simétrica (p. ej., que *fuego* sea contextónimo de *mechero* no implica que *mechero* lo sea de *fuego).* Además, los contextónimos, a diferencia de sinónimos y antónimos, no son necesariamente de la misma categoría gramatical (Ji, Ploux y Wehrli 2003: 195).

La noción de contextónimo parte de la base de que las palabras relacionadas contextualmente con otra palabra señalan de manera significativa el valor semántico de dicha palabra en un contexto dado (Ji, Ploux y Wehrli 2003: 194). De hecho, como se vio en §3.5.3.2.1, el contexto lingüístico es un factor elemental en el proceso por el cual el potencial semántico de una unidad léxica da lugar a un significado concreto.

En este trabajo, se utilizó la noción de contextónimo para determinar qué rasgos semánticos activan un concepto dado según el dominio contextual. Para extraer las listas de contextónimos de un término, utilizamos SketchEngine. Creamos el *word-sketch* reproducido en la Tabla 29, a partir del cual un contextónimo de una palabra clave es cualquier verbo, sustantivo o adjetivo (excepto los lemas *be, have, such, other, much, many,* *more, do, make, another, most*[77]) que se encuentre antes o después de la palabra clave con entre cero y 44 palabras entre ellos.

La ventana de 44 palabras se determinó a partir del estudio que se describe en San Martín (2014). En dicho estudio, se realizó una comparación entre una lista de referencia compuesta por una lista de términos extraídos con TermoStat Web 3.0 a partir de un corpus de 133 definiciones en inglés de MAGMA y listas de términos extraídas a partir de cinco corpus compuestos de líneas de concordancia del término *magma* de longitud variable.

```
2:[tag="V.*|N.*|J.*" &
!lemma="be|have|such|other|much|many|more|do|make|another|most"]
[]{0,44} 1:[tag="N.*"|tag="V.*"|tag="J.*"]
1:[tag="N.*"|tag="V.*"|tag="J.*"] []{0,44} 2:[tag="V.*|N.*|J.*" &
!lemma="be|have|such|other|much|many|more|do|make|another|most"]
```

**Tabla 29. *Word-sketch* creado para extraer contextónimos**

Para reducir la interferencia por la variación terminológica, los términos de la lista de referencia se agruparon en 23 categorías según la proposición conceptual expresada en relación con MAGMA.

La longitud de las concordancias de los corpus era respectivamente: 1) 100 caracteres antes y después de *magma*; 2) 250 caracteres antes y después de *magma;* 3) 500 caracteres antes y después de *magma;* 4) 750 caracteres antes y después de *magma*. El quinto corpus contenía solo oraciones en las que aparecía *magma*. La comparación con la lista de referencia se hizo con los 50 y los 100 términos más frecuentes de cada lista.

El resultado del estudio indicó que la lista que presentaba mayor precisión y exhaustividad en comparación con la lista de referencia era la de los 100 términos más frecuentes extraídos a partir del corpus de concordancias de 250 caracteres antes y después de *magma.*

[77] Esos lemas se excluyeron tras observar en pruebas iniciales que ocupaban habitualmente los primeros lugares de las listas de contextónimos y que nunca transmitían una relación concreta ni significativa con respecto a la palabra clave en las concordancias correspondientes.

Dado que una lista de términos extraída de un corpus de concordancias es en gran medida equivalente a una lista de contextónimos para un término especializado, se asumió la ventana de 250 caracteres como parámetro para configurar la extracción de contextónimos. Para calcular el equivalente en palabras de 250 caracteres[78], se dividió 250 por el número de caracteres de media de las palabras contenidas en el corpus MULTI (5,52), lo cual dio como resultado (redondeado) 45 palabras[79].

[78] La necesidad de utilizar un número de palabras en vez de un número caracteres se debe a una limitación de SketchEngine.

[79] En el *word-sketch* se estableció la ventana de 44 palabras en vez de 45 porque los 250 caracteres de las concordancias en el estudio de San Martín (2014) incluyen los caracteres del contextónimo.

# 5 RESULTS

## 5.1 SELECTION OF CONCEPTS REQUIRING A FLEXIBLE DEFINITION

### 5.1.1 Extraction of terms used in several domains

Each of the 14 domain subcorpora was loaded to TermoStat Web 3.0 in order to obtain a list candidate terms. Only monolexical nouns were extracted. The result was the following:

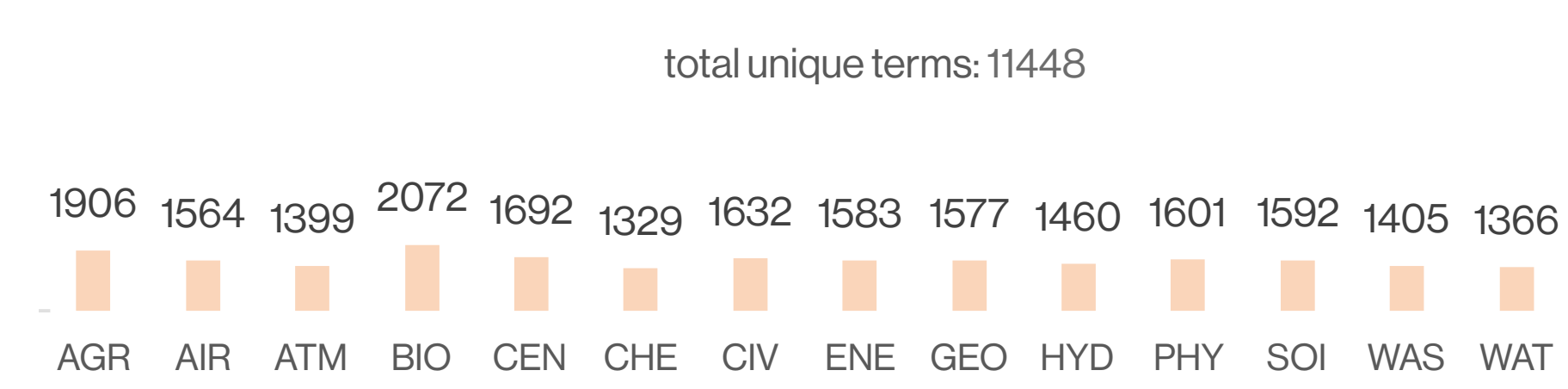

**Figure 26. Number of candidate terms in total and per contextual domain**

All the candidate terms extracted by TermoStat had a minimum frequency of two occurrences. To ensure representativeness, a term was only considered to be associated with a domain if it had at least 64 occurrences in the corresponding subcorpus (one occurrence per 5,000 words). The 64-occurrence threshold was selected after analyzing the resulting lists with different thresholds. In most cases, a term with fewer than 64 occurrences in a subcorpus was not sufficiently representative. However, we are aware that this threshold eliminated from our working list some relevant associations between terms and domains. Therefore, in a real application of this procedure, the terminologist will likely need to vary the threshold to tailor it to the needs of his/her project. Figure 27 shows the number of remaining terms in each domain and the total number of unique terms after applying the 64-occurrence threshold. Figure 28 shows in how many domains terms appear.

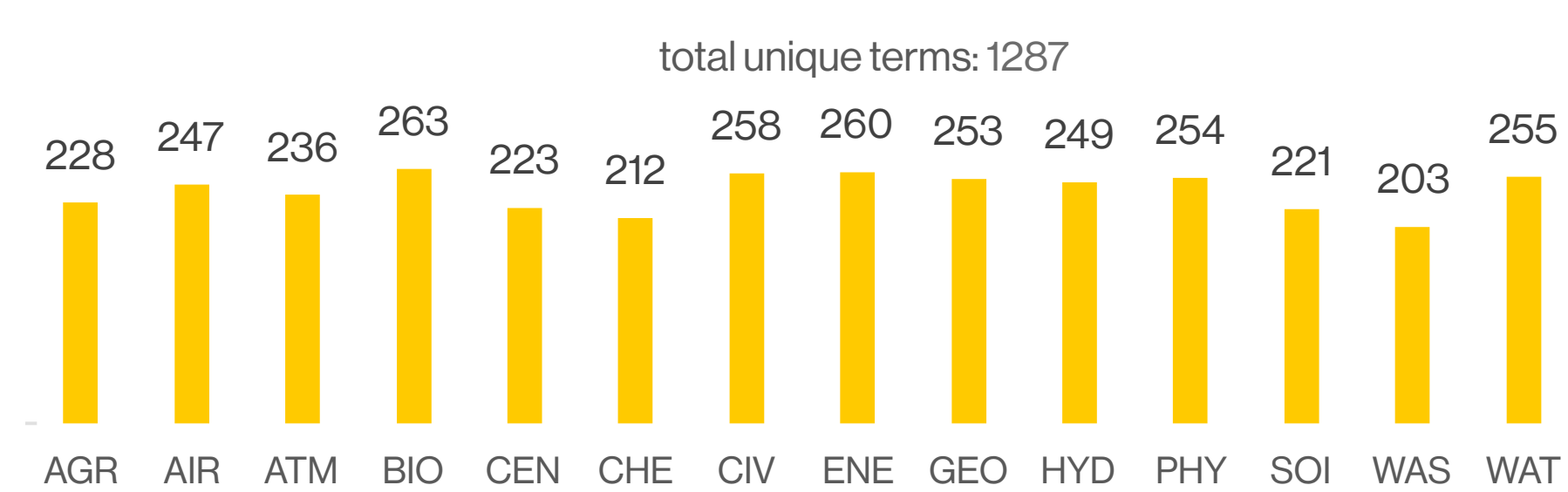

**Figure 27. Number of candidate terms in total and per contextual domain after applying the 64-occurrence threshold**

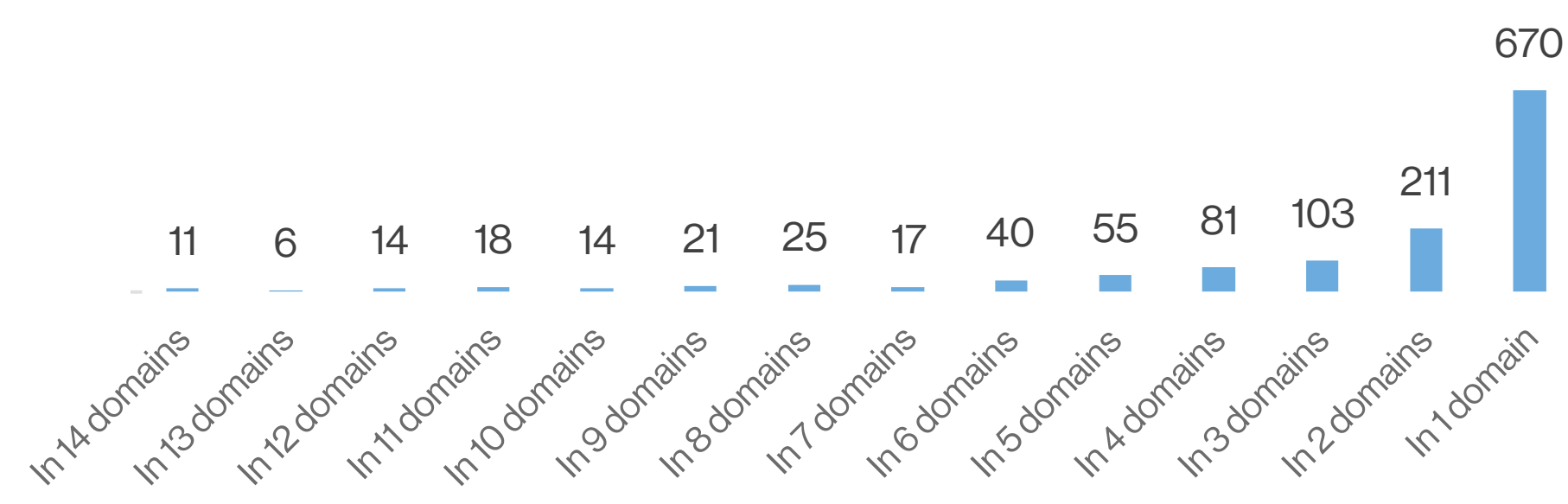

**Figure 28. Number of domains in which the remaining terms appear**

The associated domains and frequency numbers of several terms (such as *groundwater/ground-water*, *vapor/vapour*, and *sulfur/sulphur*) were recalculated to account for spelling variations. From the 64-occurrence-threshold list of terms, only those present in at least three domains were retained, i.e., 405 terms. Then, all abbreviations and acronyms terms were eliminated after manual verification in the corpus. As a result, there were 380 terms distributed as follows:

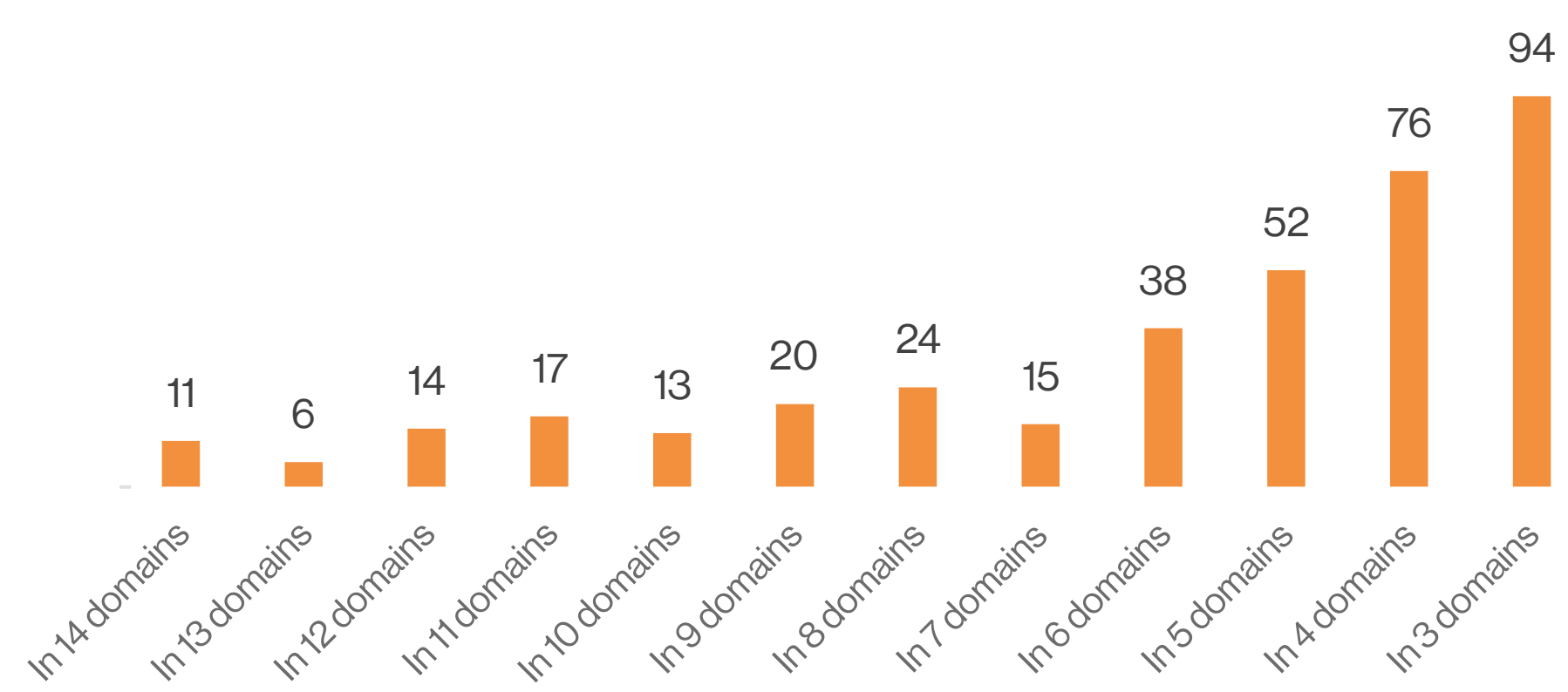

**Figure 29. Number of domains in which the remaining terms appear after the removal of abbreviations and acronyms**

### 5.1.2 Filtering out scientific transdisciplinary lexicon units

The 380 terms were compared to the scientific transdisciplinary lexicon (STL) list in order to retain only those not present in the list STL. The STL list used was an expansion of the English list available on the website of the STL project[80]. We expanded it following the same guidelines as its developer[81]. Consequently, after automatic filtering, we classified as an STL unit each term that was not originally detected as such whether it was a morphological derivative of an STL unit already in the English list or whether it had an equivalent in the French list. The final result is shown in Table 30.

[80] Available at: <http://olst.ling.umontreal.ca/lexitrans/>.

[81] In personal communication, Patrick Drouin, main developer of the STL, informed me that the English STL was currently being enriched with morphological derivates and translations from the French STL list.

| | **STL units (= discarded)** | **Non-STL units ( = retained)** | **% STL units** |
|---|---|---|---|
| In 14 domains | 1 | 10 | 90,91 |
| In 13 domains | 0 | 6 | 100 |
| In 12 domains | 2 | 12 | 85,71 |
| In 11 domains | 1 | 16 | 94,12 |
| In 10 domains | 2 | 11 | 84,62 |
| In 9 domains | 2 | 18 | 90 |
| In 8 domains | 2 | 22 | 91,67 |
| In 7 domains | 3 | 12 | 80 |
| In 6 domains | 10 | 27 | 72,97 |
| In 5 domains | 17 | 36 | 67,92 |
| In 4 domains | 76 | 27 | 49 |
| In 3 domains | 94 | 46 | 48 |
| TOTAL | 113 | 267 | 70,26 |

**Table 30. Number of STL units, non-STL units, and percentage**

Not surprisingly, since our method of finding contextually variable terms in different domains is similar to the one used by Drouin (2007; 2010b) to draw up the STL list, more than 70% of the terms coincide. However, since Drouin's extraction included many other domains in addition to the environment, the elimination of STL units ensured that the terms were specific to the environmental domain.

### 5.1.3 The final working list

In order to obtain the final list, only the terms whose referent is an entity (abstract or concrete) were retained. In other words, we excluded concepts referring to attributes, states, and processes. The following table shows the final working list of terms with their frequency in each domain (when the frequency is lower than 64, a hyphen is inserted):

| | **AGR** | **AIR** | **ATM** | **BIO** | **CEN** | **CHE** | **CIV** | **ENE** | **GEO** | **HYD** | **PHY** | **SOI** | **WAS** | **WAT** |
|---|---|---|---|---|---|---|---|---|---|---|---|---|---|---|
| acid | 209 | 177 | - | 247 | 190 | 931 | - | - | - | - | - | 127 | 104 | 115 |
| animal | 277 | - | - | 721 | - | - | - | - | 122 | - | - | - | - | - |
| aquifer | - | - | - | - | - | - | 159 | - | - | 378 | - | - | - | 86 |
| atom | - | 74 | - | - | 76 | 1321 | - | - | 132 | - | 189 | - | - | - |
| bacterium | 124 | - | - | 236 | - | - | - | - | - | - | - | 92 | 103 | 164 |

| | AGR | AIR | ATM | BIO | CEN | CHE | CIV | ENE | GEO | HYD | PHY | SOI | WAS | WAT |
|---|---|---|---|---|---|---|---|---|---|---|---|---|---|---|
| basin | - | - | - | - | - | - | - | - | 200 | 996 | - | - | - | 187 |
| bed | - | - | - | - | 87 | - | - | - | 176 | 76 | - | - | 65 | 257 |
| biomass | - | - | - | - | - | - | - | - | - | - | - | 87 | 150 | 88 |
| calcium | - | - | - | - | - | 71 | - | - | 74 | - | - | 103 | - | 91 |
| carbon | 252 | 435 | 104 | 101 | 111 | 380 | - | 134 | 113 | - | - | 475 | 206 | 124 |
| cation | - | - | - | - | - | 122 | - | 84 | - | - | 80 | 574 | 85 | - |
| chlorine | - | 75 | - | - | - | 78 | - | - | - | - | - | - | - | 165 |
| cloud | - | 116 | 890 | - | - | - | - | - | 75 | 67 | - | - | - | - |
| coal | - | 221 | - | - | - | - | - | 106 | 185 | - | - | - | 84 | - |
| contaminant | - | 92 | - | - | 80 | - | - | - | - | - | - | - | 155 | 132 |
| crop | 2236 | - | - | - | - | - | - | - | - | 94 | - | 465 | - | - |
| crystal | - | - | 151 | - | 113 | 124 | - | - | 191 | - | 66 | - | - | - |
| dioxide | 71 | 291 | - | - | - | 166 | - | - | - | - | - | - | 76 | - |
| drought | 98 | - | 85 | - | - | - | - | - | - | 86 | - | - | - | - |
| ecosystem | 110 | 72 | - | 178 | - | - | - | - | - | 64 | - | 168 | - | - |
| electron | - | - | - | 199 | - | 750 | - | - | 92 | - | 315 | - | - | - |
| enzyme | 65 | - | - | 134 | - | 107 | - | - | - | - | - | - | - | - |
| feed | 79 | - | - | - | 275 | - | - | - | - | - | - | - | - | 91 |
| fertilizer | 226 | - | - | - | - | - | - | - | - | - | - | 152 | 64 | - |
| forest | 124 | 222 | - | 87 | - | - | - | - | - | 111 | - | 194 | - | - |
| fuel | 76 | 304 | - | - | - | 79 | - | 424 | - | - | - | - | 375 | - |
| fungus | 116 | - | - | 314 | - | - | - | - | - | - | - | 117 | - | - |
| glass | - | - | - | - | - | - | 66 | 129 | - | - | 82 | - | 247 | - |
| glucose | 74 | - | - | 102 | - | 80 | - | - | - | - | - | - | - | - |
| grain | 678 | - | - | - | - | - | - | - | 165 | - | - | 103 | - | - |
| groundwater | - | - | - | - | - | - | - | - | 149 | 387 | - | - | - | 144 |
| heat | - | 103 | 473 | - | 403 | 292 | - | 1108 | 154 | 133 | 418 | - | 150 | 71 |
| hydrogen | - | 71 | - | - | - | 539 | - | 424 | - | - | 64 | 68 | 152 | - |
| ice | - | - | 592 | - | - | - | - | - | 433 | 227 | - | - | - | - |
| ion | - | - | - | 152 | - | 732 | - | - | 136 | - | 146 | 164 | - | 193 |
| iron | - | - | - | - | - | 126 | 89 | - | 112 | - | - | 123 | - | 92 |
| lake | - | - | - | 70 | - | - | - | - | 187 | 259 | - | - | - | - |
| metal | - | 93 | - | - | 78 | 541 | 102 | 103 | - | - | 77 | 118 | 360 | 79 |
| meter | - | - | - | - | - | - | - | - | 207 | - | 84 | - | - | 206 |
| methane | - | 89 | - | - | - | - | - | - | - | - | - | - | 93 | 64 |
| mineral | 79 | - | - | - | - | - | 246 | - | 914 | - | - | 523 | - | - |
| molecule | - | 101 | 139 | 448 | 200 | 933 | - | - | - | - | 118 | - | - | - |
| nitrate | - | 83 | - | - | - | - | - | - | - | - | - | 104 | - | 95 |
| nitrogen | 289 | 158 | - | - | - | 144 | - | - | - | - | - | 270 | - | 110 |

| | AGR | AIR | ATM | BIO | CEN | CHE | CIV | ENE | GEO | HYD | PHY | SOI | WAS | WAT |
|---|---|---|---|---|---|---|---|---|---|---|---|---|---|---|
| nutrient | 235 | - | - | 84 | - | - | - | - | - | - | - | 314 | - | - |
| ocean | - | 71 | 349 | 102 | - | - | - | - | 390 | 99 | - | - | - | - |
| oil | 179 | - | - | - | 105 | - | - | 430 | - | - | - | - | 166 | - |
| organism | 171 | - | - | 1035 | - | 73 | - | - | 154 | - | - | 109 | - | 82 |
| oxide | - | 199 | - | - | - | 245 | - | - | - | - | - | 194 | - | - |
| oxygen | 85 | 70 | 125 | 114 | - | 518 | - | - | 94 | 125 | - | 130 | 125 | 265 |
| ozone | - | 516 | 138 | - | - | - | - | - | - | - | - | - | - | 65 |
| particle | 80 | 681 | 463 | - | 158 | 158 | 166 | - | 246 | 84 | 829 | 301 | 157 | 198 |
| pathogen | 94 | - | - | 93 | - | - | - | - | - | - | - | - | - | 116 |
| pesticide | 127 | - | - | - | 236 | - | - | - | - | - | - | - | 75 | - |
| pipe | - | - | - | - | 110 | - | 78 | 200 | - | - | - | - | - | 350 |
| plane | - | - | - | - | - | - | 137 | - | 76 | - | 205 | - | - | - |
| planet | - | - | 79 | - | - | - | - | - | 219 | - | 76 | - | - | - |
| plant | 2637 | 359 | 77 | 1163 | 390 | - | 87 | 416 | 182 | 181 | - | 881 | 453 | 812 |
| plate | - | 67 | - | - | 69 | - | 99 | - | 544 | - | 83 | - | - | - |
| pollutant | - | 1031 | - | - | - | - | - | - | - | - | - | - | 108 | 137 |
| potassium | - | - | - | - | - | 91 | - | - | 64 | - | - | 65 | - | - |
| pump | - | - | - | - | 76 | - | - | 194 | - | - | - | - | - | 188 |
| reactor | - | - | - | - | 271 | - | - | 90 | - | - | - | - | 90 | 125 |
| reservoir | - | - | - | - | - | - | - | 145 | 66 | 360 | - | - | - | 80 |
| river | - | - | - | - | - | - | 141 | - | 230 | 860 | - | - | - | 124 |
| rock | 76 | - | - | - | - | - | 185 | 70 | 2203 | 83 | - | 186 | - | - |
| salt | - | - | - | 88 | 71 | 115 | - | - | - | 70 | - | 180 | - | 87 |
| sand | - | - | - | - | - | - | 191 | - | 250 | - | - | 224 | - | 228 |
| sea | - | 66 | 215 | 114 | - | - | 88 | - | 225 | 176 | - | - | - | - |
| sediment | - | - | - | - | - | - | 92 | - | 518 | 108 | - | - | - | - |
| sodium | - | - | - | - | - | 176 | - | - | - | - | - | 107 | - | 97 |
| steel | - | - | - | - | - | - | 222 | 65 | - | - | - | - | 89 | - |
| sulfate | - | 99 | - | - | - | 68 | - | - | - | - | - | 183 | - | - |
| sulfur | 66 | 213 | - | - | - | 91 | - | - | - | - | - | - | - | - |
| summer | 98 | - | 112 | - | - | - | - | - | - | 81 | - | - | - | - |
| tank | - | - | - | - | 135 | - | - | 132 | - | - | - | - | 114 | 619 |
| tube | - | - | - | - | 87 | - | - | 130 | - | - | 86 | - | - | - |
| vapor | - | 151 | 304 | - | 229 | 209 | - | 104 | - | 147 | - | - | - | - |
| vegetation | 79 | - | - | - | - | - | - | - | 66 | 164 | - | 165 | - | - |
| wind | 114 | 124 | 1116 | - | - | - | 142 | 217 | 208 | 72 | - | - | - | - |
| winter | 213 | - | 106 | - | - | - | - | - | - | 67 | - | - | - | - |

**Table 31. Final working list of terms**

### 5.1.4 Polysemous terms

As indicated in §3.5.3.5.1, the distinction between polysemy and contextual variation is fuzzy. However, within the framework of a resource that uses flexible definitions, as defended in this thesis, the impact of sense splitting or lumping is reduced, because not only is polysemy represented but also contextual variation.

It is important to note that the concepts associated with a polysemous term experience contextual variation as well. Nevertheless, since our focus is on contextual variation, with the intention of avoiding interference, polysemous terms were discarded[82].

In order to classify a term as polysemous, we used our corpus as a point of reference. In other words, only those terms that appeared as polysemous in our corpus were treated as such, even if their polysemous status was well identified in dictionaries or other terminographical sources. For instance, the term *atom* was included in the analysis because, in spite of being polysemous (ATOM_1: *a unit of matter consisting of a single nucleus surrounded by a number of electrons equal to the number of protons in the nucleus* / ATOM_2: *a number or symbol of a measure algebra other than zero*[83]), only occurrences of ATOM_1 were found in our corpus.

The instances of non-linear polysemy[84] were identified by means of the tests in §3.5.3.5.1. In those cases in which the tests were not conclusive, they were considered polysemous.

What Cruse called *facets,* i.e., terms with two related senses that belong to different ontological categories (but that can be reunited in certain contexts) were considered polysemous. The reason is that it is a phenomenon not related to domain-specific contextual variation, and, therefore, it is beyond the scope of this work. Furthermore, EcoLexicon treats them as polysemous terms. For example, *pollution* was considered polysemous because it has a PROCESS facet (*the process of polluting*) and an ENTITY facet (*substances that pollute*). Nonetheless, we believe that further study is needed to decide whether this is the best option for EcoLexicon and how to streamline the representation of regular polysemy.

[82] Since our corpus was POS-tagged, terms belonging to more than one part of speech were not considered polysemous for our purposes.

[83] The source of both definitions is TERMIUM Plus (Translation Bureau / Bureau de la traduction [Canada] 2015).

[84] Non-linear polysemy is the type of polysemy that occurs when there is no relation of inclusion between polysemous words (Cruse 2011: 115).

Autohyperonimic and automeronymic senses were considered polysemous if the different senses were not linked to different domains. For example, the term *oil* can refer to a kind of unctuous liquid substance or to petroleum, which is a subtype of the first sense. The term *oil* refers to any of the two senses in any domain.

The list of polysemous terms with a definition for each sense is in Annex 2.

## 5.2 ADAPTING THE TYPES OF CONTEXTUAL VARIATION

After the analysis of all the terms in our working list, it became evident that the division of contextual variation in microsenses (including local sub-senses) (§3.5.3.5.4), WOS (§3.5.3.5.3), and modulation (§3.5.3.5.2) was not adequate for our purposes.

Our notion of domain-dependent contextual variation includes three phenomena: (i) modulation (similar to Cruse's modulation); (ii) perspectivization (related to Cruse's WOS); (iii) subconceptualization (akin to Cruse's microsenses and local sub-senses). These phenomena are additive in that all concepts experience modulation; some concepts also undergo perspectivization; and finally a small number of concepts additionally are subjected to subconceptualization.

### 5.2.1 Modulation

Our notion of modulation is similar to Cruse's. It is the simplest kind of contextual variation. In this work, we regarded contextual modulation as a type of contextual variation that solely alters minor non-necessary and non-prototypical characteristics of a concept. As a consequence, these alterations are not represented in a definition. If a concept only undergoes modulation in a given domain, then no flexible definition is created for it. For instance, BREAKWATER, which is a type of coastal defense structure, is only modulated in *AGRONOMY*. BREAKWATER is so seldom activated in that domain that no different necessary or prototypical characteristics are conventionally attached to the concept in that context with respect to the general environmental premeaning.

### 5.2.2 Perspectivization

For a premeaning to be considered a perspective, it needs to change the level of prototypicality of certain traits of a concept in a consistent way in relation to the general environmental premeaning. For example, SULFUR is conceptualized from two distinct perspectives in *AGRONOMY* and *AIR QUALITY MANAGEMENT*. In *AGRONOMY*, SULFUR is an important nutrient that plants need for different functions, such as chlorophyll production. In *AIR QUALITY MANAGEMENT*, SULFUR (as part of sulfur dioxide) is a major air pollutant that can have severe effects on the environment and human health.

Perspectivization is related to Cruse's WOS, but differs from it in two ways. The first is that Cruse's WOS are opposed to microsenses. In other words, if a premeaning is considered to be a microsense, then it is not a WOS and vice versa. However, in our approach, perspectives and subconceptualizations are not in opposition to each other. In fact, all subconceptualizations also include a perspective.

The second way that our notion of perspective differs from Cruse's WOS is that we use Pustejovsky's (1995) qualia roles differently. Cruse characterizes the WOS adopted by a domain with qualia roles. Nonetheless, we defend that the prevailing qualia role for a concept should be determined at frame level and not at domain level. The reason for this is that, in a given domain, a concept may participate in several frames at the same time. As a consequence, different qualia roles could be assigned in the same domain depending on the activated frame. For example, the concept FUNGUS in *AGRONOMY* can be seen from a telic-agentive perspective if FUNGUS is conceptualized in the frame of MUSHROOM CULTURE. FUNGUS can also be construed from a different telic perspective if it is categorized as a PEST.

As explained in §3.6.4, genus choice in flexible definitions is guided by qualia roles. In the case of a concept that activates several frames such as FUNGUS, the most frequent frame according to corpus analysis is used to select the genus. However, if possible, the role of the concept in the other relevant frames is to be represented in the definition for a given domain as well. For instance, FUNGUS in *AGRONOMY* would take PEST as a genus, but the fact that it can be cultivated should also be featured in the definition.

### 5.2.3 Subconceptualization

Our third type of contextual variation is subconceptualization. It includes both Cruse's microsenses and local sub-senses. Nonetheless, it is even more general, since there were cases where a concept manifested contextual variation akin to these two kinds but did not meet all the requirements.

For our purposes, we define a subconceptualization as a premeaning in which the extension of the concept in relation to the reference point (in our case, the general environmental premeaning) is modified. For instance, the concept CRYSTAL gives rise to a subconceptualization in *ATMOSPHERIC SCIENCES* because the extension of the concept is restricted to CRYSTALS made of ICE in that domain.

Nevertheless, it should be highlighted that subconceptualization is not a clear-cut phenomenon. Since a subconceptualization relies on the notion of conceptual extension, it is inherently fuzzy. As shown in §2.1.3.2, conceptual limits are fuzzy and show prototypical effects.

The upper limit of a subconceptualization is polysemy: when the subconceptualization shows too much autonomy. The lower limit is perspectivization, which is when a context-variable trait is not strong enough to be considered extension-changing.

In both cases, we believe that the decision whether to regard a premeaning as a separate concept, a subconceptualization or a perspective must be made depending on user needs and the characteristics of the resource where the concept or concepts are to be represented.

None of our analyzed concepts met all the requirements to be considered a microsense strictly in the way that Cruse described them. Nevertheless, some concepts did manifest one of its main characteristics, namely, that the relation between the general environmental premeaning and the domain premeaning is hierarchical. This is the reason why we have named this phenomenon subconceptualization.

The main difference between perspectivation and subconceptualization regarding terminological definitions is that a contextualized definition for a subconceptualization represents as necessary a trait that in the general environmental definition does not have necessary status. This is not possible in a contextualized definition that is only a perspective. For instance, for an entity to be considered a CONTAMINANT in *WATER TREATMENT AND SUPPLY*, it needs to contaminate water, which becomes a necessary characteristic. Therefore, CONTAMINANT is subconceptualized in that domain. However, HYDROGEN in *ENERGY ENGINEERING* is not a subconceptualization, but simply a perspective. Although it is conceptualized as used for energy storage, this is not a necessary characteristic. In other words, an entity will still be considered HYDROGEN in that domain even if it is not used for that purpose.

The complexity of the phenomenon of subconceptualization and the different degrees in which it manifested itself was evident during our analysis. The following sections comment on certain factors that affected the identification of subconceptualizations and the implications for their practical application to the terminological definition.

#### 5.2.3.1 The entrenchment of the general environmental premeaning

Our point of reference is what we have called the general environmental premeaning. A subdomain premeaning is compared to the general environmental premeaning in order to determine whether there is a subconceptualization. However, the entrenchment of the general environmental premeaning is in many cases doubtful, which, at the same time, is one of the reasons why flexible definitions are necessary.

The general environmental premeaning is a summary of the most relevant characteristics of a given concept across its domain-specific conceptualizations. Most of the time, in the general environmental domain, there is not a clear frame associated with the concept, but rather several of them in different subdomains. Moreover, different premeanings may have different levels of prototypicality when taken into account together in the whole environmental domain. In conclusion, since our point of reference normally lacks entrenchment, this also hinders the clear identification of subconceptualizations.

#### 5.2.3.2 The entrenchment of subconceptualizations

The phenomenon of subconceptualizations occurs when, in a certain domain, a concept is construed in such a way that the resulting premeaning is a subtype of the concept. In other words, the extension of the concept is modified (and, more often than not, restricted). Nonetheless, there are cases where the subconceptualization seems to be weaker than in other cases, and this hinders the identification of subconceptualizations.

For instance, POLLUTANT has its extension limited in the domain of *AIR QUALITY MANAGEMENT* to those agents that pollute the air. In the general environmental premeaning, POLLUTANT includes any agent that pollutes any kind of environment. However, in the *AIR QUALITY MANAGEMENT* subcorpus, the complex terms *air pollutant* and *atmospheric pollutant* were fairly frequent (164 occurrences out of a total of 1,092 occurrences of *pollutant* in the *AIR QUALITY MANAGEMENT* subcorpus). This need for specification is indicative of weaker entrenchment.

In contrast, the subconceptualization of DROUGHT in *AGRONOMY* shows greater entrenchment than the one of POLLUTANT in *AIR QUALITY MANAGEMENT*. In our *AGRONOMY* subcorpus, the term *drought* was never preceded by a modifier that indicated the subconceptualization, only adjectives that qualified drought mainly in terms of its severity or length:

| AGR | The Midwestern U.S. is experiencing a **severe drought**. Farmers are being limited on how much they can irrigate |
|---|---|
| AGR | on the sandstone cliffs around their cities. Two **lengthy droughts** around 1090 and 1130 CE, social unrest, and land |
| AGR | 1895–1995 Percent of time in severe and **extreme drought** 5–9.99% Less than 5% 10–14.9% 15–19.9% 20% or |
| AGR | and rainfall information to determine **long-term drought**. Severe or extreme drought in each year from 1895 to 1995, |
| AGR | temperatures and below-normal rainfall caused **widespread drought** during the 1930s from North Dakota to Texas. |

**Table 32. Concordances of the lemma *drought* (noun) preceded by a modifier in the *AGRONOMY* subcorpus**

### 5.2.3.3 The problem of the subconceptualization's extension

POLLUTANT is prototypically preconceptualized as an AIR POLLUTANT in the domain of *AIR QUALITY MANAGEMENT*. It is not a case of polysemy; it is simply that context narrows the extension of the concept by default (a subconceptualization). The problem is that in EcoLexicon, AIR POLLUTANT is an independent concept in itself. In other words, the subconceptualization's extension (POLLUTANT in *AIR QUALITY MANAGEMENT*) corresponds to another concept's extension (AIR POLLUTANT).

In this case, AIR POLLUTANT is a relevant concept in the domain of *AIR QUALITY MANAGEMENT*. The definition of POLLUTANT for the domain of *AIR QUALITY MANAGEMENT* will consist of a hyperlink to the definition of AIR POLLUTANT in that same domain:

| POLLUTANT (*AIR QUALITY MANAGEMENT*) |
| --- |
| → AIR POLLUTANT (*AIR QUALITY MANAGEMENT*) |

**Table 33. Definition of POLLUTANT in *AIR QUALITY MANAGEMENT***

The case of FUEL is different. The extension of the subconceptualization of FUEL in *AGRONOMY* corresponds to the extension of CHEMICAL FUEL (in contrast to NUCLEAR FUEL). However, CHEMICAL FUEL is not a relevant concept in the domain, since the term *chemical fuel* does not appear in the *AGRONOMY* subcorpus. In this case, the definition for FUEL describes the subconceptualization and the perspective, but no explicit reference is made to the concept CHEMICAL FUEL[85].

#### 5.2.3.4 The problem of the hierarchical organization

The representation of the subconceptualization structure formed by the general environmental premeaning and its subconceptualizations in flexible definitions are highly dependent on the domain hierarchy being used and its depth. For instance, taking Cruse's example of *ball,* the resulting subconceptualization (TENNIS BALL, GOLF BALL, etc.) could not be accounted for in a flexible definition if the domain Sports was not subdivided.

The mismatch between the subconceptualization structure and our domain hierarchy occurred frequently. For example, the concept SUMMER has two subconceptualizations: one in *ASTRONOMY*, and one in *ATMOSPHERIC SCIENCES*, particularly in *METEOROLOGY*. On the one hand, the astronomical subconceptualization limits the extension of SUMMER from the summer solstice to the autumnal equinox, thus covering the months of June (partially), July, August, and September (partially) in the Northern Hemisphere, and December (partially), January, February, and March (partially) in the Southern Hemisphere. On the other hand, in *ATMOSPHERIC SCIENCES*, SUMMER covers June, July, and August in the Northern Hemisphere, and December, January, February in the Southern Hemisphere.

[85] An extract of the contextualized definition of FUEL in *AGRONOMY* is shown in Table 50.

The problem arose when that subconceptualization structure was mapped onto those domains in which the term *summer* had more than 64 occurrences: *AGRONOMY*, *ATMOSPHERIC SCIENCES*, and *HYDROLOGY*. Firstly, *ASTRONOMY* is not part of our domain hierarchy. Secondly, while *ATMOSPHERIC SCIENCES* has its own subconceptualization, *AGRONOMY* and *HYDROLOGY*'s premeanings include both the astronomical and meteorological subconceptualizations along with a geographical factor (the perceived start and end of the summer depends on the meteorological conditions of a given place), similarly to the general environmental premeaning. Nevertheless, having a similar extension does not mean that the definitions will be the same, since, as previously mentioned, concepts also undergo perspectivization in each domain.

Cruse presented microsenses as being hierarchical and incompatible between domains. However, our notion of subconceptualization is more flexible because it permits several domains to share the same subconceptualization (though with different perspectives). Furthermore, subconceptualizations do not need to be incompatible. For instance, in the case of POLLUTANT, a HEAVY METAL is at the same time a kind of POLLUTANT in *AIR QUALITY MANAGEMENT* and *WATER TREATMENT AND SUPPLY*, even if they have different subconceptualizations.

## 5.3 CONCEPT ANALYSIS

### 5.3.1 Chemical elements

Our first set of analyzed concepts corresponds to those that are categorized as CHEMICAL ELEMENTS in the domain of *CHEMISTRY*. The terms corresponding to these concepts and their occurrence in each subdomain corpus are shown on Table 34.

| | AGR | AIR | ATM | BIO | CEN | CHE | CIV | ENE | GEO | HYD | PHY | SOI | WAS | WAT |
|---|---|---|---|---|---|---|---|---|---|---|---|---|---|---|
| calcium | - | - | - | - | - | 71 | - | - | 74 | - | - | 103 | - | 91 |
| carbon | 252 | 435 | 104 | 101 | 111 | 380 | - | 134 | 113 | - | - | 475 | 206 | 124 |

| | AGR | AIR | ATM | BIO | CEN | CHE | CIV | ENE | GEO | HYD | PHY | SOI | WAS | WAT |
|---|---|---|---|---|---|---|---|---|---|---|---|---|---|---|
| chlorine | - | 75 | - | - | - | 78 | - | - | - | - | - | - | - | 165 |
| hydrogen | - | 71 | - | - | - | 539 | - | 424 | - | - | 64 | 68 | 152 | - |
| iron | - | - | - | - | - | 126 | 89 | - | 112 | - | - | 123 | - | 92 |
| nitrogen | 289 | 158 | - | - | - | 144 | - | - | - | - | - | 270 | - | 110 |
| oxygen | 85 | 70 | 125 | 114 | - | 518 | - | - | 94 | 125 | - | 130 | 125 | 265 |
| potassium | - | - | - | - | - | 91 | - | - | 64 | - | - | 65 | - | - |
| sodium | - | - | - | - | - | 176 | - | - | - | - | - | 107 | - | 97 |
| sulfur | 66 | 213 | - | - | - | 91 | - | - | - | - | - | - | - | - |

**Table 34.Terms whose corresponding concepts can be categorized as CHEMICAL ELEMENTS**

Most of the known chemical elements occur naturally in the environment, as is the case of the ten elements analyzed in this work. CHEMICAL ELEMENTS are substances that cannot be decomposed into simpler substances. The elements can appear in different forms. More specifically, the bonding between the atoms of an element may vary resulting in allotropes (e.g., ozone is an allotrope of oxygen); the number of neutrons of an atom may change originating isotopes or the same element (e.g., chlorine-35 is the most common stable isotope of chlorine); or an atom or molecule of an element may gain or lose electrons becoming an ion (e.g., if a calcium atom loses electrons, it becomes a calcium ion). Furthermore, elements may either be pure or appear with other elements in compounds or mixtures, or in different states (mainly, solid, liquid, or gas).

However, whatever the form of an element, the number of protons in its atoms remains unaltered. If an element undergoes a change in the number of protons, it becomes another element and thus loses its identity. Moreover, no other characteristic is added by default to the premeanings of CHEMICAL CONCEPTS in the contextual domains analyzed. For instance, for something to be categorized as OXYGEN in *BIOLOGY*, *GEOLOGY*, or *HYDROLOGY*, it needs to have 8-proton atoms. This condition applies in all domains. Since the alteration of necessary characteristics is compulsory for a conceptualization to be a kind of subconceptualization, CHEMICAL ELEMENTS do not give rise to that phenomenon.

In order to verify if a premeaning is a subconceptualization in a given domain, one must ascertain whether the additional characteristics activated are necessary, given the set of contextual constraints. For example, ATMOSPHERIC OXYGEN is the most prototypical type of OXYGEN in *ATMOSPHERIC SCIENCES*. Nonetheless, when OXYGEN is activated in this domain, "OXYGEN *located-in* ATMOSPHERE" is not a necessary characteristic. In other words, for an entity to be considered OXYGEN in *ATMOSPHERIC SCIENCES,* it does not need to be located in the atmosphere, despite the fact that, in this domain, OXYGEN is conceptualized as located in the atmosphere more frequently than in other environmental subdomains. Therefore, the characteristic "OXYGEN *located-in* ATMOSPHERE" only has prototypical status in *ATMOSPHERIC SCIENCES,* and the concept OXYGEN only gives rise to a perspective in this domain, not a subconceptualization.

It should also be underlined that, since CHEMICAL ELEMENTS do not give rise to subconceptualizations, when a CHEMICAL ELEMENT is preconceptualized in an environmental subdomain, it does not stand for the most prototypical compound or substance in which it appears. For instance, when the concept CALCIUM is activated in *GEOLOGY*, it does not stand for CALCIUM CARBONATE (its most usual form in that contextual domain). Instead, the information that CALCIUM prototypically appears as a part of CALCIUM CARBONATE becomes active and should be represented in the corresponding definition.

Nevertheless, in the same way as the other CHEMICAL ELEMENTS, CARBON does not undergo subconceptualization. During the analysis of the term *carbon* in the contextual domain of *AIR QUALITY MANAGEMENT*, it was found that sometimes it does stand for the whole compound in which it prototypically appears (CARBON DIOXIDE) or even for all sorts of greenhouse gasses. However, this cannot be considered a subconceptualization, because when the concept CARBON is activated in *AIR QUALITY MANAGEMENT*, it does not stand for CARBON DIOXIDE by default. It could not be considered a case of automeronymic polysemy either because this usage is limited in our corpus to complex terms such as *carbon emission* or *carbon footprint*. Thus, it appears that it is simply a shortened form of *carbon dioxide emission* or *carbon dioxide footprint*. In our *AIR QUALITY MANAGEMENT* subcorpus, when *carbon* appeared without being part of a noun phrase, it always referred to CARBON and not CARBON DIOXIDE:

| | |
|---|---|
| AIR | chlorofluorocarbon (CFC) is an organic compound that contains only **carbon**, chlorine, and fluorine, produced as a volatile |
| AIR | article: Organofluorine chemistry As in simpler alkanes, **carbon** in the CFCs and the HCFCs is tetrahedral. Because the fluorine |
| AIR | Furthermore, many examples are known for higher numbers of **carbon** as well as related compounds containing bromine. Uses |
| AIR | unburned combustibles present in the samples. Exactly how **carbon** works as a charge carrier is not fully understood, but it is |
| AIR | Life is carbon based, and CO2 is the source of that **carbon**. Carbon dioxide is also a major greenhouse gas and, because |

**Table 35. Concordances of the lemma *carbon* (noun) in the *AIR QUALITY MANAGEMENT* subcorpus**

As shown, for CHEMICAL ELEMENTS, the forms in which the element prototypically appears in a given domain is a relevant characteristic. To determine them, in addition to reference material, the *modifier*, *modifies,* and *contextonym* word-sketches are very useful. The *modifier* and *modifies* word-sketches show the noun phrases in which the lemma appears. For instance, the *modified* word-sketch for *potassium* in the *GEOLOGY* subcorpus (Table 36) shows that potassium feldspar, a type of mineral, is one of the main forms in which potassium is studied in that domain. As can be observed in Table 37, CAST IRON, WROUGHT IRON, and PIG IRON are the most frequent forms of IRON in *CIVIL ENGINEERING*[86]. Finally, Table 38 shows how the *contextonym* word-sketch indicates that in *AGRONOMY*, SULFUR is prototypically conceptualized as part of SULFATES and SULFIDES.

| potassium (noun) / subcorpus: GEO / frequency: 56[87] (152.92 per million) | |
|---|---|
| **modifies** | |
| feldspar | 16 |
| calcium[88] | 3 |
| bicarbonate | 2 |

**Table 36. *Modifies* word-sketch of *potassium* (noun) in the *GEOLOGY* subcorpus**

[86] In the case of IRON in *CIVIL ENGINEERING*, its most prototypical form is CAST IRON (along with PIG IRON, WROUGHT IRON, and STEEL). Although CAST IRON could be regarded as a kind of IRON, it is a material with a very high proportion of iron in its composition, but, strictly speaking, not a type of IRON.

[87] Divergences in the count of occurrences between TermoStat and SketchEngine are attributable to the fact that they use different versions of TreeTagger.

[88] The detection of *calcium* as being modified by *potassium* is an error in the *modifies* word-sketch, since in all 3 cases, there is a comma between *calcium* and *potassium*.

| iron (noun) / corpus: CIV / frequency: 98 (260.30 per million) | |
|---|---|
| **modifier** | |
| cast | 12 |
| wrought | 9 |
| pig | 4 |

**Table 37. *Modifier* word-sketch of *iron* (noun) in the *CIVIL ENGINEERING* subcorpus**

| sulfur (noun) / corpus: AGR / frequency: 67 (175.50 per million) | |
|---|---|
| **contextonym** | |
| plant | 52 |
| compound | 27 |
| sulfate | 23 |
| sulfide | 21 |

**Table 38. Contextonyms of *sulfur* (noun) in the *AGRONOMY* subcorpus**

It is worth noting that CHEMICAL ELEMENTS show relational and compositional autonomy, as Cruse described for WOS. Relational autonomy in the domain-specific conceptualizations of CHEMICAL ELEMENTS takes the form of multidimensional categorization. In other words, they have different hyperonyms, co-hyponyms, and hyponyms, depending on the subdomain. For instance, in *AGRONOMY*, NITROGEN has NUTRIENT as one of its hyperonyms, and PHOSPHORUS as a cohyponym. Meanwhile, in *AIR QUALITY MANAGEMENT*, one of its hyperonyms is ATMOSPHERIC GAS and one of its co-hyponyms is OXYGEN. Concepts tend to have several hyperonyms, and this is one of the difficulties of crafting flexible definitions.

As for compositional autonomy, it is usual for the terms denoting CHEMICAL ELEMENTS to have different collocations, depending on the domain. In fact, the collocations that a term has in different domains, which can be obtained by means of the various word sketches, are another key to the characterization of the perspective of a concept in a given domain. One example is the adjective *usable* in conjunction with *nitrogen*. In our corpus, this adjective is only used with *nitrogen* in *SOIL SCIENCES*. It refers to the nitrogen that can be used by plants and thus indicates that nitrogen is an important nutrient for plants.

| | |
|---|---|
| SOI | an unfertilised field, this is the most important source of **usable nitrogen**. In a soil with 5% organic matter perhaps 2 to 5% of that |
| SOI | are converted by way of mineralisation. Some amount of **usable nitrogen** is fixed by lightning as nitric oxide (NO) and nitrogen |
| SOI | fraction of nitrogen is held this way. Nitrogen losses **Usable nitrogen** may be lost from soils when it is in the form of nitrate, |

**Table 39. Concordances of the lemmas *usable* (adjective) and *nitrogen* (noun) in the *SOIL SCIENCES* subcorpus**

An example of a verb that collocates with a chemical element only in a given domain is *liquefy*, which takes *hydrogen* as an object in the *ENERGY ENGINEERING* domain. This collocation shows that hydrogen needs to be liquefied in order to be used as an energy storage medium in *ENERGY ENGINEERING*:

| | |
|---|---|
| ENE | boil-off during storage. The energy required to **liquefy hydrogen** is about 3040% of the energy content of the gas, |
| ENE | during the liquefaction process. Thermodynamically, **liquefying hydrogen** involves three heat transfer stages: the rst |
| ENE | by a liquefaction cycle. The minimum theoretical work to **liquefy** normal **hydrogen** to 99.79% para form is 14,280 kJ |

**Table 40. Concordances of the lemmas *liquefy* (verb) and *hydrogen* (noun) in the *ENERGY ENGINEERING* subcorpus**

Finally, some CHEMICAL ELEMENTS in certain contextual domains pose a problem that we have called *hyperversality*. Hyperversality is a property of certain concepts in relation to a given contextual domain. A concept is considered hyperversatile when it participates in a wide range of frames in a contextual domain. For example, OXYGEN is a hyperversatile concept in most of its contextual domains given that it is the most abundant chemical element on Earth, and since it is highly reactive, it participates in many processes. Therefore, the premeaning of OXYGEN in any of its contextual domains is not linked to only one frame.

Hyperversatility is not absolute; there are degrees, and it is intimately linked to the size of the contextual domains. A concept that is hyperversatile in a given domain tends to be less hyperversatile with respect to its subdomains. For instance, OXYGEN in *BIOLOGY* is more hyperversatile than in *BOTANY*.

Hyperversatility is a problem because formulating a flexible definition for a hyperversatile concept entails a longer process of documentation. It is necessary to make sure that the definition summarizes the most important roles of that concept in the contextual domain. In the case of EcoLexicon, hyperlinks in the definitions and the conceptual map ensure that the user can obtain further information if needed.

## 5.3.2 Chemical compounds and other substances

In this set, we have included concepts that have many conceptual similarities with CHEMICAL ELEMENTS, since, to a varying degree, they have a specific chemical composition that acts as a necessary characteristic. This set of concepts corresponds to those that are categorized as CHEMICAL COMPOUNDS in the domain of *CHEMISTRY* (in the case of DIOXIDE, GLUCOSE, and METHANE) as well as other concepts that can be categorized as SUBSTANCES but are neither CHEMICAL ELEMENTS nor CHEMICAL COMPOUNDS. This is the case of OZONE (an allotrope of oxygen), ENZYME (a kind of protein), COAL, and STEEL, which can be categorized as SUBSTANCES[89]. The terms corresponding to the concepts in this set and their occurrence in each subdomain corpus are shown in Table 41.

| | AGR | AIR | ATM | BIO | CEN | CHE | CIV | ENE | GEO | HYD | PHY | SOI | WAS | WAT |
|---|---|---|---|---|---|---|---|---|---|---|---|---|---|---|
| coal | - | 221 | - | - | - | - | - | 106 | 185 | - | - | - | 84 | - |
| dioxide | 71 | 291 | - | - | - | 166 | - | - | - | - | - | - | 76 | - |
| enzyme | 65 | - | - | 134 | - | 107 | - | - | - | - | - | - | - | - |
| glucose | 74 | - | - | 102 | - | 80 | - | - | - | - | - | - | - | - |
| methane | - | 89 | - | - | - | - | - | - | - | - | - | - | 93 | 64 |
| ozone | - | 516 | 138 | - | - | - | - | - | - | - | - | - | - | 65 |
| steel | - | - | - | - | - | - | 222 | 65 | - | - | - | - | 89 | - |

**Table 41. Terms whose corresponding concepts can be categorized as CHEMICAL COMPOUNDS or SUBSTANCES**

GLUCOSE, METHANE, and OZONE have a chemical formula. The formula, $C_6H_{12}O_6$ for GLUCOSE, $CH_4$ for METHANE, and $O_3$ for OZONE, acts as a sufficient and necessary characteristic. Because of their chemical formulas, their contextual behavior resembles that of CHEMICAL ELEMENTS. As a result, they have perspectives but do not give rise to subconceptualizations.

For example, METHANE is conceptualized as a GAS emitted by the decomposition of organic waste in *WASTE MANAGEMENT* or as a by-product of wastewater treatment in *WATER TREATMENT AND SUPPLY*. In contrast, in *AIR QUALITY MANAGEMENT*, it is a GREENHOUSE GAS. In none of the three domains, are the necessary characteristics of METHANE altered. However, the frames in which the concept participates are not the same in every domain, and as a consequence, certain conceptual relations become highlighted and others backgrounded.

[89] We have excluded from this group those concepts that can be categorized as SUBSTANCES, but whose composition is too variable. Such concepts are included in §5.3.5.

In *AIR QUALITY MANAGEMENT*, the highlighted information for METHANE is that it contributes to the greenhouse effect and that it can be found in the atmosphere due to natural as well as anthropogenic sources. In that domain, its impact on the environment and how it can be managed or reduced are also highlighted. Nonetheless, most of this information is backgrounded in *WASTE MANAGEMENT*, where the fact that the decomposition of organic waste emits methane is much more relevant.

Although METHANE and OZONE are not hyperversatile in their contextual domains, this is not the case of GLUCOSE in *BIOLOGY*. GLUCOSE in *AGRONOMY* activates the frames of PHOTOSYNTHESIS and PLANT RESPIRATION. Therefore, its roles in those frames are explained in the definition. However, in *BIOLOGY*[90], GLUCOSE participates in many more frames, including PHOTOSYNTHESIS and PLANT RESPIRATION. As a consequence, the definition should be less detailed regarding the roles of GLUCOSE in each frame because, otherwise, the definition would be extremely long.

The composition of COAL and STEEL is variable. Nevertheless, this does not give rise to subconceptualizations. The reason is that, even though varying compositions and properties are possible, in the contextual domains there are no fixed compositions or properties different from the general environmental premeaming of those concepts. For instance, STEEL is an alloy of iron and carbon, and can sometimes contain other elements as well, such as manganese, nickel, chromium, molybdenum, or silicon. There are different types of steel depending on the proportion of its elements or the method of production (among other factors), which make the steel suitable for different uses. However, when STEEL is preconceptualized in one of its contextual domains, there are not any characteristics whose status changes to necessary in these contextual domains (as compared to the general environmental domain). In other words, the extension of the premeanings remains the same as in the general environmental domain. This entails that the conditions required for an entity to be categorized as IRON are the same across environmental domains.

[90] Most of our concepts are hyperversatile in *BIOLOGY* because it is a very large domain. In fact, in EcoLexicon, *BIOLOGY* is subdivided in *BIOLOGICAL OCEANOGRAPHY*, *BOTANY*, *ZOOLOGY*, *MICROBIOLOGY*, *MOLECULAR BIOLOGY*, and *BIOCHEMISTRY*.

Finally, DIOXIDE and ENZYME are also perspectived, but manifest a kind of hyperversatility that we have called superordinate hyperversatility. This hyperversatility is due to the fact that the concept has many subordinate concepts and each subordinate participates in different frames, which makes the superordinate concept hyperversatile. This causes them to behave as a superordinate-level concept in the domain in question. For instance, ENZYME is a conceptual category that encompasses proteins that act as catalysts in biochemical reactions. Given that each enzyme is specific to a kind of reaction or set of reactions, ENZYME does not participate in any specific frame per se and remains rather generic despite the fact that it is restricted to one contextual domain. This is the case in *AGRONOMY*, given that plants and the soil contain an enormous variety of enzymes with very different functions.

Superordinate hyperversatility can be generally detected by means of the *modifier* word-sketch if the term has modifiers that refer to subtypes. However, the terms for subtypes of ENZYME do not contain the term *enzyme*. In this case, they need to be specifically searched for. For instance, the two most common enzymes in the *AGRONOMY* subcorpus are *RuBisCO* and *reductase*. RuBisCo was detected because it appeared in the contextonym word-sketch for enzyme in the *AGRONOMY* subcorpus. Another method of extracting types of ENZYME is to generate a wordlist containing only the words ending in *–ase* (the suffix for enzymes):

| .*ase / subcorpus: AGR | |
|---|---|
| oxygenase | 6 |
| synthase | 6 |

| α-amylase | 6 |
|---|---|
| reductase | 6 |
| carboxylase | 4 |
| lyase | 3 |
| anhydrase | 3 |
| kinase | 3 |
| nitrogenase | 3 |
| amylase | 2 |
| synthetase | 2 |
| mono-oxygenase | 2 |

**Table 42. Most frequent enzymes in the *AGRONOMY* subcorpus (filtered list of words ending in -ase)**

The term *RuBisCO* is the abbreviation of *ribulose-bisphosphate carboxylase/oxygenase*. It belongs to the ENZYME category of LYASES and participates in the frame of the CALVIN CYCLE OF PHOTOSYNTHESIS. *Reductase* is a generic name for all the enzymes that catalyze a reduction reaction[91] and in the *AGRONOMY* subcorpus, its most common type is *APS reductase*, which takes part in the frame of SULFUR METABOLISM.

For its part, DIOXIDE is also a superordinate hyperversatile concept, but in contrast to *enzyme*, *dioxide* is always preceded by a modifier, e.g.: *carbon dioxide*, *sulfur dioxide*, *nitrogen dioxide*. In this case, the contextualized definition of DIOXIDE for each domain would include its necessary characteristics ("oxide that contains two atoms of oxygen in each molecule") and a reference to its main subtypes in the domain according to corpus evidence[92].

### 5.3.3 Artifactual concepts

In this group of concepts, we have included those whose referents are artifacts:

[91] Its corresponding concept (REDUCTASE) is thus also superordinate hyperversatile.

[92] Since DIOXIDE is a chemical compound, in order to determine the most relevant subtypes, it is necessary to generate a wordlist of the terms containing the string "O2", as well as the *modifier* word-sketch of the term *dioxide*.

| | AGR | AIR | ATM | BIO | CEN | CHE | CIV | ENE | GEO | HYD | PHY | SOI | WAS | WAT |
|---|---|---|---|---|---|---|---|---|---|---|---|---|---|---|
| pipe | - | - | - | - | 110 | - | 78 | 200 | - | - | - | - | - | 350 |
| pump | - | - | - | - | 76 | - | - | 194 | - | - | - | - | - | 188 |
| tank | - | - | - | - | 135 | - | - | 132 | - | - | - | - | 114 | 619 |
| tube | - | - | - | - | 87 | - | - | 130 | - | - | 86 | - | - | - |

**Table 43. Terms whose corresponding concepts can be categorized as ARTIFACTS**

All the examples offered by Cruse for microsenses were artifacts (*knife*, *ball*, *card*). However, none of the artifactual concepts analyzed in this work gave rise to subconceptualizations with respect to our domains[93]. These concepts only manifest perspectives and are superordinate hyperversatiles in their respective domains. For instance, the three first results of the *modifier* word-sketches for *tank* in its different contextual domains are the following:

| tank (noun) / corpus: WAT / frequency: 641 (1,700.37 per million) | |
|---|---|
| **modifier** | 326 |
| aeration | 64 |
| storage | 59 |
| sedimentation | 23 |

**Table 44. *Modifier* word-sketch for the term *tank* in the *WATER TREATMENT AND SUPPLY* subcorpus**

| tank (noun) / corpus: ENE / frequency: 134 (361.88 per million) | |
|---|---|
| **modifier** | 76 |
| mud | 27 |
| storage | 17 |
| cryogenic | 4 |

**Table 45. *Modifier* word-sketch for the term *tank* in the *ENERGY ENGINEERING* subcorpus**

| tank (noun) / corpus: CEN / frequency: 151 (404.47 per million) | |
|---|---|
| **modifier** | 67 |
| storage | 22 |
| frp | 4 |
| cargo | 4 |

**Table 46. *Modifier* word-sketch for the term *tank* in the *CHEMICAL ENGINEERING* subcorpus**

[93] It is probable, though, that they give rise to subconceptualizations in very restricted contexts.

| tank (noun) / corpus: WAS / frequency: 116 (313.56 per million) | |
|---|---|
| **modifier** | 84 |
| septic | 30 |
| storage | 12 |
| underground | 10 |

**Table 47. *Modifier* word-sketch for the term *tank* in the *WASTE MANAGEMENT* subcorpus**

As can be observed, each contextual domain has an array of representative types of TANK, although STORAGE TANK, a tank with the specific function of storing a fluid, appears in all of them. For instance, in *WATER TREATMENT AND SUPPLY*, there is also AERATION TANK (a tank where air is injected into water) and SEDIMENTATION TANK (a tank where suspended solids in water are allowed to settle so that they can be removed) (see Table 44). Tanks in *WATER TREATMENT AND SUPPLY* are mainly used to hold water, but they can also hold other substances needed to treat water or the by-products of the treatment. There is not a characteristic so prototypical in any of its domains that it modifies the extension of the concept. As a consequence, the necessary characteristics of TANK in *WATER TREATMENT AND SUPPLY* are not modified with respect to the general environmental conceptualization and TANK participates in a broad range of frames through its subordinate concepts.

## 5.3.4 Functional concepts

In this group, we have included those having a function either in natural or artificial processes as a necessary characteristic. All of them, except for FERTILIZER, give rise to subconceptualizations:

| | AGR | AIR | ATM | BIO | CEN | CHE | CIV | ENE | GEO | HYD | PHY | SOI | WAS | WAT |
|---|---|---|---|---|---|---|---|---|---|---|---|---|---|---|
| contaminant | - | 92 | - | - | 80 | - | - | - | - | - | - | - | 155 | 132 |
| fertilizer | 226 | - | - | - | - | - | - | - | - | - | - | 152 | 64 | - |
| fuel | 76 | 304 | - | - | - | 79 | - | 424 | - | - | - | - | 375 | - |
| nutrient | 235 | - | - | 84 | - | - | - | - | - | - | - | 314 | - | - |
| pathogen | 94 | - | - | 93 | - | - | - | - | - | - | - | - | - | 116 |
| pesticide | 127 | - | - | - | 236 | - | - | - | - | - | - | - | 75 | - |
| pollutant | - | 1031 | - | - | - | - | - | - | - | - | - | - | 108 | 137 |

**Table 48. Terms whose corresponding concepts are functional**

CONTAMINANT, NUTRIENT, PATHOGEN, and POLLUTANT generate subconceptualizations in some of their contextual domains. It follows that their extension is different in those domains with respect to the general environmental premeaning. Table 49 shows a summary of the domain-specific contextual variation of these concepts.

| | AGR | AIR | ATM | BIO | CEN | SOI | WAS | WAT |
|---|---|---|---|---|---|---|---|---|
| NUTRIENT | sub: plant nutrient | | | pers | | pers | | |
| PATHOGEN | sub: plant pathogen | | | pers | | | pers | pers |
| POLLUTANT | | sub.: air pollutant | sub.: air pollutant | | | | pers | sub.: water pollutant |
| CONTAMINANT | | sub.: air contaminant | | | pers | | pers | sub: water con. |
| sub: subconceptualization<br>pers: perspective | | | | | | | | |

**Table 49. Summary of the domain-specific contextual variation of NUTRIENT, PATHOGEN, POLLUTANT, and CONTAMINANT**

NUTRIENT in its general environmental conceptualization could be said to have the necessary characteristic of being a substance that provides nourishment to a living organism. This characteristic of NUTRIENT is restricted to the provision of nourishment to plants in *AGRONOMY*, thus modifying the extension of the category. However, in the case of *SOIL SCIENCES*, although *nutrient* usually refers to PLANT NUTRIENT, it is also applied to other types of NUTRIENT (such as ANIMAL NUTRIENT, or MICROORGANISM NUTRIENT). Therefore, given that the characteristic of the provision of nourishment to plants is not strong enough to become necessary, NUTRIENT has the same extension in *SOIL SCIENCES* as in the general environmental premeaning.

NUTRIENT does not have different superordinate concepts in *AGRONOMY* with respect to its other contextual domains and the general environmental conceptualization. NUTRIENT is a superordinate-level concept in all of the analyzed domains, and it follows that its only possible superordinate concept is SUBSTANCE. The situation is similar for CONTAMINANT, PATHOGEN, and POLLUTANT, which also give rise to subconceptualizations and do not show different superordinate concepts depending on the domain.

The subconceptualizations of these concepts have a lexicalized term associated, and their extension corresponds to a full-fledged concept in EcoLexicon. For instance, the subconceptualization of CONTAMINANT in *AIR QUALITY MANAGEMENT* corresponds to AIR CONTAMINANT. As a consequence, the definition of contaminant in *AIR QUALITY MANAGEMENT* will consist of a hyperlink to the definition of AIR CONTAMINANT contextualized in that domain.

The kind of subconceptualization that functional concepts show are associated with their nature of quasi-predicates (Mel'čuk and Polguère 2008; Polguère 2012). While the concepts referring to events and attributes are predicative because they convey a fact and, therefore, need arguments, the concepts denoting entities are not predicative because they can be conceived without reference to arguments: WIND, OXYGEN, PLANET. However, quasi-predicate concepts are entity concepts that need arguments:

> Comme le terme l'indique, un quasi-prédicat n'est pas un prédicat au sens strict, car il ne satisfait pas la propriété sémantique définitoire des prédicats : celle de dénoter un fait. Cependant, l'entité dénotée par un quasi-prédicat est particulière, en ce sens qu'elle est définie plutôt par son implication dans un fait que par ses propriétés intrinsèques ; à cause de cela, un quasi-prédicat contrôle des positions actancielles (Mel'čuk and Polguère 2008: 6).

In DicoEnviro, examples of quasi-predicates are encoded with their argument structure:

ej. 49 PASSENGER: PATIENT(*user*) is a ~ of INSTRUMENT(*vehicle*)

ej. 50 DEPOSIT: ~ of PATIENT(*substance*)

ej. 51 DRIVER: AGENT(*user*) is a ~ of PATIENT(*vehicle*)

In our opinion, the specification of the arguments of a quasi-predicate is in many cases subjective and context-sensitive. Consequently, the arguments of CONTAMINANT, NUTRIENT, PATHOGEN, and POLLUTANT could be the following:

- CONTAMINANT => CONTAMINANT of x(ENVIRONMENT)
- NUTRIENT => NUTRIENT for x(LIVING ORGANISM)
- PATHOGEN => PATHOGEN to x(LIVING ORGANISM)
- POLLUTANT => POLLUTANT of x(ENVIRONMENT)

Although PATHOGEN is an entity, it also carries an inner predicate that could be expressed as follows: "PATHOGEN *causes disease in* X". The X in that inner predicate corresponds to the argument of PATHOGEN (PATHOGEN TO X). This inner predicate is a necessary characteristic of the concept, and the context may specify it further. Depending on the value of X, the extension of the concept changes. If, in a given domain's premeaning, the value of X is different from the value of X in the general environmental premeaning, then there is a subconceptualization. For example, in *AGRONOMY*, the argument of PATHOGEN is by default PLANT in contrast to the general environmental premeaning, where it is a LIVING ORGANISM. In *BIOLOGY*, it is also a LIVING ORGANISM (therefore, there is no subconceptualization), but if we only consider *ZOOLOGY*, the argument would be ANIMAL and then it would be a subconceptualization.

*PESTICIDE and FUEL*

PESTICIDE and FUEL are more complex cases of domain-specific subconceptualizations. The main characteristic of PESTICIDE is that the extension of its subconceptualization in *AGRONOMY* is at the same time more general and more restrictive. As for FUEL, the extension of its subconceptualization in *AGRONOMY* corresponds to that of a concept that is not relevant in that domain, since its lexicalization is not found in the corresponding subcorpus.

PESTICIDE is preconceptualized in the general environmental domain as any AGENT used to repel, kill or control pests. However, in *AGRONOMY*, its subconceptualization is more specific with respect to certain characteristics, but also more extense concerning others. The premeaning of PESTICIDE in *AGRONOMY* specifies that it refers to an agent used to repel, kill or control pests in plants or plant products. Therefore, in *AGRONOMY*, the extension of PESTICIDE does not include those pesticides used, for instance, to protect human from pests. On the other hand, it prototypically includes other type of products such as defoliants or desiccants, which help crops grow or offers them protection.

The term *plant protection product* lexicalizes this subconceptualization, even though it does not include the term *pesticide*. This term is only a synonym for *pesticide* in *AGRONOMY*. Currently, EcoLexicon does not allow the inclusion of context-dependent synonyms, but this will be implemented in the future.

As for FUEL, its corresponding term appears (with more than 64 occurrences) in five domains in our corpus: *AGRONOMY*, *AIR QUALITY MANAGEMENT*, *CHEMISTRY*, *ENERGY ENGINEERING,* and *WASTE MANAGEMENT*. The general environmental premeaning of FUEL can be roughly glossed as a substance from which energy can be obtained by chemical or nuclear reaction. It is precisely the type of reaction that produces a subconceptualization in *AGRONOMY*, which is the only domain where NUCLEAR FUEL is not pertinent. In *AGRONOMY*, the extension of FUEL only includes those from which energy can be obtained through a chemical reaction (normally, combustion). They are relevant because agricultural machinery needs fuel to function, and also because certain crops can be used to produce biofuel. CHEMICAL FUEL is not a relevant concept in *AGRONOMY* because the term *chemical fuel* does not appear in the *AGRONOMY* subcorpus. As a consequence, the *AGRONOMY* contextualized definition of FUEL is not replaced by a hyperlink to the definition of CHEMICAL FUEL, as is the case of POLLUTANT in *AIR QUALITY MANAGEMENT* with AIR POLLUTANT. The definition of FUEL in *AGRONOMY* represents as necessary (instead of simply prototypical) the characteristic that gives rise to the context-specific change of extension (in this case, the type of reaction):

| FUEL (general environmental) |
|---|
| Substance that, as a result of a chemical or nuclear reaction, can provide energy… |
| FUEL (*AGRONOMY*) |
| Substance that, as a result of a chemical reaction, can provide energy… |

**Table 50. Extract of the contextualized definition of FUEL in the general environmental domain and *AGRONOMY***

If FUEL in *AGRONOMY* was a simple perspective without a subconceptualization, the trait "as a result of a chemical reaction" would be represented as prototypical, for instance, by adding an adverb such as *normally* or *usually*: "normally as a result of a chemical reaction".

Therefore, in cases like fuel, where the extension of the subconceptualization corresponds to a concept that is not relevant in that domain the contextualized definition represents as necessary a characteristic that is a prototypical or optional trait in the general environmental definition. This is also the case for subconceptualizations that do not correspond to another independent concept.

In the other domains, both chemical and nuclear fuels are relevant. For instance, in *WASTE MANAGEMENT*, FUEL activates the frame BIOFUEL PRODUCTION FROM WASTE, where only chemical fuels are relevant. It also invokes the frame NUCLEAR WASTE MANAGEMENT, where USED NUCLEAR FUEL is categorized as a RADIOACTIVE WASTE that needs to be safely disposed of.

If we had divided the *ENERGY ENGINEERING* domain, NUCLEAR FUEL would be a subconceptualization of FUEL in a hypothetical subdomain of *NUCLEAR ENERGY*. However, throughout *ENERGY ENGINEERING*, both CHEMICAL FUEL and NUCLEAR FUEL are relevant subtypes of FUEL. Therefore, the extension of the *ENERGY ENGINEERING* premeaning is similar.

*FERTILIZER*

FERTILIZER is the only concept in this group that did not give rise to subconceptualizations because its extensions are not altered in its respective domains. FERTILIZER can also be regarded as a quasi-predicate. However, in the three domains in which it appears (*AGRONOMY*, *SOIL SCIENCES*, and *WASTE MANAGEMENT*), its Y argument is always SOIL (A X(HUMAN BEING) uses a FERTILIZER to render Y(SOIL) more fertile). No other characteristic of the concept gives rise to subconceptualizations. Therefore, it only undergoes perspectivization in its domains.

## 5.3.5 Other concepts

In this section, we analyze the rest of the concepts on our working list:

| | AGR | AIR | ATM | BIO | CEN | CHE | CIV | ENE | GEO | HYD | PHY | SOI | WAS | WAT |
|---|---|---|---|---|---|---|---|---|---|---|---|---|---|---|
| animal | 277 | - | - | 721 | - | - | - | - | 122 | - | - | - | - | - |
| aquifer | - | - | - | - | - | - | 159 | - | - | 378 | - | - | - | 86 |
| atom | - | 74 | - | - | 76 | 1321 | - | - | 132 | - | 189 | - | - | - |
| bacterium | 124 | - | - | 236 | - | - | - | - | - | - | - | 92 | 103 | 164 |
| cation | - | - | - | - | - | 122 | - | 84 | - | - | 80 | 574 | 85 | - |
| crystal | - | - | 151 | - | 113 | 124 | - | - | 191 | - | 66 | - | - | - |
| drought | 98 | - | 85 | - | - | - | - | - | - | 86 | - | - | - | - |
| ecosystem | 110 | 72 | - | 178 | - | - | - | - | - | 64 | - | 168 | - | - |
| electron | - | - | - | 199 | - | 750 | - | - | 92 | - | 315 | - | - | - |
| fungus | 116 | - | - | 314 | - | - | - | - | - | - | - | 117 | - | - |
| groundwater | - | - | - | - | - | - | - | - | 149 | 387 | - | - | - | 144 |
| ice | - | - | 592 | - | - | - | - | - | 433 | 227 | - | - | - | - |
| ion | - | - | - | 152 | - | 732 | - | - | 136 | - | 146 | 164 | - | 193 |
| lake | - | - | - | 70 | - | - | - | - | 187 | 259 | - | - | - | - |
| molecule | - | 101 | 139 | 448 | 200 | 933 | - | - | - | - | 118 | - | - | - |
| planet | - | - | 79 | - | - | - | - | - | 219 | - | 76 | - | - | - |
| river | - | - | - | - | - | - | 141 | - | 230 | 860 | - | - | - | 124 |
| sediment | - | - | - | - | - | - | 92 | - | 518 | 108 | - | - | - | - |
| summer | 98 | - | 112 | - | - | - | - | - | - | 81 | - | - | - | - |
| vapor | - | 151 | 304 | - | 229 | 209 | - | 104 | - | 147 | - | - | - | - |
| vegetation | 79 | - | - | - | - | - | - | - | 66 | 164 | - | 165 | - | - |
| wind | 114 | 124 | 1116 | - | - | - | 142 | 217 | 208 | 72 | - | - | - | - |
| winter | 213 | - | 106 | - | - | - | - | - | - | 67 | - | - | - | - |

**Table 51. Remaining terms in the working list**

*Concepts with subconceptualizations*

There are five concepts that give rise to subconceptualizations in their domains: VAPOR, CRYSTAL, SUMMER, WINTER, and DROUGHT.

VAPOR and CRYSTAL both have subconceptualizations linked to their quasi-predicate arguments. VAPOR can be roughly defined as the gas of a substance that is at a state below its critical temperature. Its argument structure is VAPOR of (X). In all of its domains except for *HYDROLOGY* and *ATMOSPHERIC SCIENCES*, the value of X is SUBSTANCE by default. In *HYDROLOGY* and *ATMOSPHERIC SCIENCES*, it is WATER, thus generating a subconceptualization.

CRYSTAL is also a quasi-predicate whose argument expresses the substance that composes it: CRYSTAL of X. This gives rise to a subconceptualization in *ATMOSPHERIC SCIENCES*, for which the value of X is ICE. In the other domains, since it is preconceptualized as being composed of many different substances, the extension is the same as in the general environmental premeaning.

SUMMER, WINTER, and DROUGHT all refer to time periods. Their referents are also vague. Different domains determine differently the start and the end of those periods. This is the reason why they generate subconceptualizations.

The case of WINTER is similar to SUMMER, which was discussed in §5.2.3.4. As for DROUGHT, experts in different environmental domains have already remarked that it is a concept that has different definitions, depending on the domain (Wilhite and Glantz 1985; Mishra and Singh 2010). While DROUGHT can be roughly defined as a period of time with a deficiency of precipitation, only the domain determines how long and how severe the precipitation deficiency must be for it to be considered a DROUGHT. Droughts can have various effects, and each domain focuses on different ones to delimit the extension of the concept.

In our corpus, the term *drought* had over 64 occurrences in *AGRONOMY*, *ATMOSPHERIC SCIENCES*, and *HYDROLOGY*. This matched the classification of definitions proposed by Wilhite and Glantz (1985), which also includes a fourth type of DROUGHT, namely, socio-economic drought[94].

*ATMOSPHERIC SCIENCES* only takes into account the duration and severity of the precipitation deficiency in comparison to climate records. However, a DROUGHT in *AGRONOMY* is based on the effects that the deficiency of precipitation has on agricultural activities. For its part, *HYDROLOGY* restricts its premeaning of DROUGHT to the cases where the deficiency of precipitation affects the hydrological system, especially river basins.

The general environmental definition would encompass all the subconceptualizations of DROUGHT, including socio-economic drought. The domain definitions of DROUGHT would describe their extension and the perspective taken by the domain.

*Concepts with perspectives*

Among the concepts with perspectives, PLANET is an interesting example. Although it did not produce subconceptualizations, there were occurrences in the corpus where *planet* without any modifier referred to EARTH (this also happens in general language). Some concordances from our corpus are shown in Table 52.

| | |
|---|---|
| ATM | atmosphere, especially during the early evolution of **the planet** (See History of Earth for more details on this topic). |
| ATM | gases determines the amount of solar energy retained by **the planet**, leading to global warming or global cooling. The |
| PHY | of Earth. Equator, great circle in the plane of rotation of **the planet**. The inclination of this equatorial plane against the |
| GEO | are disregarded, Milankovitch theory predicts that **the planet** will continue to undergo glacial periods at least until the |
| GEO | is constantly being added at the oceanic ridges, **the planet** is not growing in size—its total surface area remains |

**Table 52. Concordances of the string "the planet" in the *ATMOSPHERIC SCIENCES*, *PHYSICS* and *GEOLOGY* subcorpora**

Since the Earth is the planet where we live, if the term is used with a definite article ("*the* planet") without any contextual cue, this activates the concept EARTH. However, this is not a case of autohyponymic polysemy. It is simply a way of denoting a referent by using its hyperonym, such as when there is a dog in the room, and someone says "the animal".

[94] The socio-economic view of DROUGHT defines it as having consequences on the socio-economic activities that depend on water supply (Wilhite and Glantz 1985: 115).

Most of the concepts in this last group are hyperversatile in most of their domains. Relevant examples include all the concepts that referred to living organisms (ANIMAL, BACTERIUM, FUNGUS, and VEGETATION). The main reason for this is that all of them are superordinate-level concepts. Therefore, they have many subordinate concepts that participate in many different frames.

Although a larger sample would be necessary, it seems that hyperversatility is a very common phenomenon. In fact, in our opinion, this is the reason that FrameNet (§2.1.2.2) lacks a consistent representation of certain types of lexical unit, such as those denoting artifacts.

## 5.4 FLEXIBLE DEFINITIONS

### 5.4.1 A concept with subconceptualizations: POLLUTANT

POLLUTANT, as reflected in the results of the terminological extraction, is relevant in the domains of *AIR QUALITY MANAGEMENT*, *WASTE MANAGEMENT* and *WATER TREATMENT AND SUPPLY* (Table 53). The following sections describe the process of extracting the knowledge associated with this concept, how contextual variation affects its domain-specific construals, and how it can then be then represented in a flexible terminological definition.

| | AGR | AIR | ATM | BIO | CEN | CHE | CIV | ENE | GEO | HYD | PHY | SOI | WAS | WAT |
|---|---|---|---|---|---|---|---|---|---|---|---|---|---|---|
| pollutant | - | 1031 | - | - | - | - | - | - | - | - | - | - | 108 | 137 |

**Table 53. Frequency of the term *pollutant* (noun) in the corresponding contextual domains according to TermoStat**

The analysis followed this order: (i) analysis of the definitions in other resources; (ii) analysis of contextonyms; (iii) analysis of hyperonyms candidates. The first two stages helped determine definitional propositions and the third (which used data from the first stage) focused on genus choice. Finally, our proposal of a flexible definition for POLLUTANT is presented along with an explanation of how it was created.

The first step was the analysis of definitions from other resources because this was the fastest of obtaining an overall understanding of the concept. However, as shall be seen, these definitions tended to fall short in their reflection of contextual variation.

We then continued with the analysis of contextonyms because it was the most efficient way to identify how each domain construed a given concept and the other concepts activated along with it. Although the study of the contextonyms of a term in different domains was a longer process than consulting definitions from other resources, it provided a better picture of contextual variation, and made it easier to select context-specific differentiae.

Finally, hyperonym candidates were analyzed, and the genus for each contextual domain was chosen. This was the final step before writing the definitions because the choice of genus had to be in accordance with the frame or frames activated by the concept in its respective contextual domains. The choice of genus was thus the product of an in-depth understanding of the concept. This could only be achieved after the previous stages had been completed.

#### 5.4.1.1 Definitions of POLLUTANT from other resources

A total of 50 specialized definitions of *pollutant* were collected from different terminological sources (see Annex 3 for the complete list of definitions). Only those definitions of the term *pollutant* without any modifier were included in the analysis. The definitions were classified according to their domain. No definitions for *pollutant* were found in *AIR QUALITY MANAGEMENT* resources. Two of the definitions belonged to *WASTE MANAGEMENT*, 17 to *WATER TREATMENT AND SUPPLY*, and 31 belonged to the general environmental domain. We considered that a definition belonged to the general environmental domain if it was related to another environmental domain other than *AIR QUALITY MANAGEMENT*, *WASTE MANAGEMENT*, and *WATER TREATMENT AND SUPPLY* or if it encompassed the whole environmental domain.

The following tables summarize the conceptual characteristics featured by the definitions analyzed according to the domain. In the tables, we used a wording similar to the one in the definitions to reflect the different viewpoints. As a consequence, there are certain overlaps.

| POLLUTANT (general environmental domain) | |
|---|---|
| *type of* | by-product, chemical, compound, contaminant, energy, gas, liquid, material, solid, substance, thing, waste |
| *result of* | emission, human activity, vehicle braking |
| *has undesirable effect on*<br>*causes harm to*<br>*adversely alters* | air quality, animals, biological processes in the ecosystem, environment, fish passage, food, habitat, health of individuals, humans, land, natural resource organisms, plants, resource, soil, water quality |
| *has attribute* | chemical, inorganic or organic, living or not living, monitored, present in excessive concentrations, regulated, resistant to biodegradation |
| *has type* | agricultural waste, biological materials, carbon dioxide, carcinogen, cellar dirt, chemical wastes, chemicals, dredged spoil, garbage, heat, incinerator residue, industrial waste, municipal waste, munitions, nitrogen, noise, other harmful substances, oxygen-demanding material, pathogen, pesticide, radioactive materials, rock, sand, sewage, sewage sludge, solid waste, sound, substances at higher temperatures than that of the receiving media, toxic metal, waste product, wrecked or discarded equipment |
| *exceeds* | environmental quality standard |
| *introduced into/discharged into* | gaseous environmental medium, liquid environmental medium, soil, solid environmental medium |

**Table 54. Characteristics of POLLUTANT represented in the definitions from other resources that belong to other environmental domains or the whole environmental domain**

| POLLUTANT (*WASTE MANAGEMENT*) | |
|---|---|
| *type of* | substance, contaminant |
| *degrades*<br>*impairs usefulness of* | air, soil, water |
| *adversely alters* | biological properties of environment, chemical properties of environment, physical properties of environment |

**Table 55. Characteristics of POLLUTANT represented in the definitions from other resources that belong to the *WASTE MANAGEMENT* domain**

| POLLUTANT (*WATER TREATMENT AND SUPPLY*) | |
|---|---|
| *type of* | chemical, component, contaminant, energy, impurity, material, mixture, organism, substance |
| *has attribute* | high concentration, chemical, physical, objectionable |
| *endangers* | atmosphere, ecosystem, life of organisms, living things, public health, soil, water |
| *introduced into* | water, land, air |
| *subject to* | effluent limitations |
| *affects* | biological properties of water, chemical properties of water, physical properties of water, purity of air, purity of water |
| *has type* | chemicals, disease-producing organisms, nutrients, oxygen-demanding materials, pesticides, sediment, silt, toxic metals |
| *impairs usefulness of* | air, soil, water |

**Table 56. Characteristics of POLLUTANT represented in the definitions from other resources that belong to the *WATER TREATMENT AND SUPPLY* domain**

As can be seen, there is consensus that a POLLUTANT is an entity with an undesirable effect on the environment or human beings. This effect is due to its being where it should not be or to its having an unacceptably high concentration in a certain medium or at a certain location. Some definitions indicate as well that a POLLUTANT is normally the result of human activity.

Certain definitions provide more detail on the damage that a POLLUTANT may cause depending on what it affects. Moreover, a small number of definitions underline that POLLUTANTS tend to be regulated by standards. It is usual that examples of the extension of the category are also included in the definitions.

According to the definitions, the extension of POLLUTANT includes not only SUBSTANCES (e.g., CARBON DIOXIDE, PESTICIDE, TOXIC METAL), but also LIVING BEINGS, HEAT, and SOUND. Additionally, some definitions include all sorts of WASTES as POLLUTANTS. However, this is not accurate because WASTES contain POLLUTANTS, but are not POLLUTANTS *per se*.

The domain-specific definitions do not help determine the specificities of the premeaning in contextual domains. For instance, in *WASTE MANAGEMENT*, they make no reference to the role of the concept in the domain. As for *WATER TREATMENT AND SUPPLY*, only three out of the 17 analyzed definitions limited the extension of POLLUTANT to WATER POLLUTANT.

#### 5.4.1.2 Contextonyms of *pollutant*

The analysis of contextonyms provides an insight into the role of a concept in a given domain. We analyzed the first 100 contextonyms of *pollutant* in each of its contextual domains as well as the general environmental contextonyms (the contextonyms of *pollutant* in the whole MULTI corpus). The full lists with the frequency of each contextonym in each contextual domain and the general environmental domain are in Annex 4.

In order to visualize the results more clearly and be able to compare between contextual domains, we represented the first 50 contextonyms in each contextual domain in a Venn diagram (Figure 30)[95]. As can be seen, among the terms shared by *AIR QUALITY MANAGEMENT*, *WASTE MANAGEMENT*, and *WATER TREATMENT AND SUPPLY* are *concentration*,

[95] In Figure 30, since the objective was to compare between the contextual domains, the general environmental contextonyms were not included. In Figure 31 we compared the general environmental contextonyms with the ones from the contextual domains.

*quality*, *reduce*, and *source*. This shows that, in all the contextual domains, (i) the concentration and source of pollutants are important parameters; (ii) pollutants affect the quality of the medium in which they are introduced; and (iii) efforts are made to reduce the amount of pollutants in the environment. The other shared contextonyms are a derivative that indicates the process in which POLLUTANT participates (*pollution*), and terms that do not convey a definite relation in their corresponding concordances (*include*, *system*, *use*). For instance, *system,* since it is a very versatile term, appears in the contextonym concordances as part of diverse complex terms such as *sewer system*, *aquatic system*, *permit system,* or *gastrointestinal system*. Thus, it is not possible to determine a clear relation

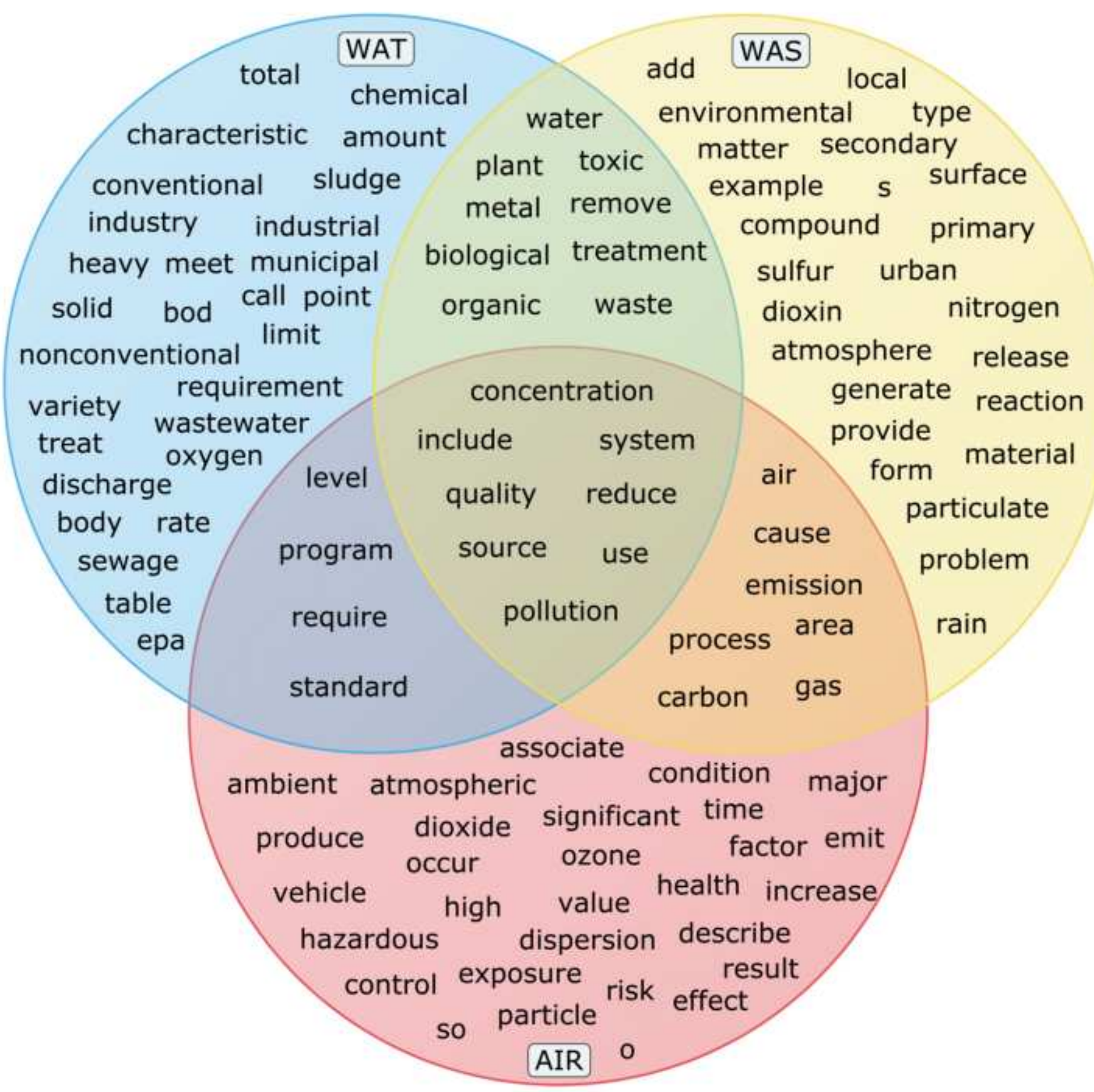

**Figure 30. Venn diagram of the first 50 contextonyms of *pollutant* in *AIR QUALITY MANAGEMENT*, *WASTE MANAGEMENT*, and *WATER TREATMENT AND SUPPLY***

between POLLUTANT and SYSTEM.

The following sections describe what the contextonyms extracted from each contextual domain and the general environmental domain show about how POLLUTANT is construed contextually.

##### 5.4.1.2.1 AIR QUALITY MANAGEMENT *contextonyms of* pollutant

The subconceptualization of POLLUTANT as an AIR POLLUTANT in *AIR QUALITY MANAGEMENT* becomes evident in view of its contextonym list. Terms like *air*, *atmospheric*, and *atmosphere* rank high on the list. Neither *water* nor *soil* appear. In the following paragraphs, we outline the information that according to the *pollutant* contextonyms in *AIR QUALITY MANAGEMENT* are relevant to its construal in this contextual domain.

Atmospheric pollution (also known as *air pollution*) is due to the emission of pollutants from very different sources into ambient air (for instance, industrial activities or the use of hydrocarbon fuels in vehicles). Many substances considered pollutants are already present in the atmosphere, but it is their high concentration in relation to natural levels and their adverse effects what causes them to be categorized as pollutants. Natural sources (such as volcanoes and the ocean) also contribute to air pollution. However, there are no terms related to natural pollution in the first 100 contextonyms, which indicates that human activities are the prototypical source of air pollution.

Although *AIR QUALITY MANAGEMENT* studies all the effects that air pollution has on the environment, as revealed in the contextonym list, one of the most important is the associated risk to human health of pollution exposure. This domain also focuses on the control and reduction of air pollution, where air pollution dispersion is a major factor.

Major air pollutants (most of which are in the contextonym list) include tropospheric ozone ($O_3$), particle pollution, carbon monoxide (CO), nitrogen oxides, sulfur dioxide ($SO_2$) and lead (Pb). Greenhouse gasses, such as carbon dioxide (CO2) or methane ($CH_4$) are also common air pollutants. Those air pollutants that cause severe health effects (such as cancer) or have environmental impact are categorized as hazardous air pollutants (such as benzene or mercury). The emission of pollutants is regulated by air quality standards.

In the light of the contextonym analysis, POLLUTANT is subconceptualized as AIR POLLUTANT and takes part in the AIR POLLUTION frame. This is a generic frame that has many subframes since there are a great variety of air pollutants, and their source and effects are very diverse. However, in the definition of AIR POLLUTANT, it is possible to provide a summary of the most important aspects of the general frame of air pollution, which would include the information outlined above.

#### *5.4.1.2.2* WASTE MANAGEMENT *contextonyms of* pollutant

*Air* and *water* rank high on the list of contextonyms. *Soil* and *land* are present as well in a lower position on the list. This indicates that the extension of POLLUTANT is not limited in this contextual domain as far as the patient of the process of pollution is concerned. Below, we present the information that, according to the 100 first *pollutant* contextonyms in *WASTE MANAGEMENT*, are relevant to its construal in this contextual domain.

Among the main objectives of *WASTE MANAGEMENT* is to avoid waste to become pollutants or reduce its effect. However, many waste management treatments cause pollution such as incineration or landfill use. Incineration, which is used to deal with organic waste materials, emits pollutants to the atmosphere such as nitrogen oxide, carbon dioxide, sulfur dioxide, dioxins and particulate matter (which includes heavy metals). The use of landfills also releases pollutants. On the one hand, landfills may emit gasses such as methane. On the other hand, pollutants may also leach from the landfill and pollute the soil and groundwater, being rain one of the primary causes. To a lesser extent, composting can also pollute the environment. It is worth noting that only *incineration* appears in the first 100 contextonyms. This was reflected in the corresponding contextualized definition.

The terms *primary* and *secondary* are also contextonyms of *pollutant* in this contextual domain. They point to the distinction between those air pollutants that are emitted directly from a source (primary pollutants) and those that result from some reaction after emission. The reason why these contextonyms are in the first-100-contextonym list in *WASTE MANAGEMENT* though not in *AIR QUALITY MANAGEMENT* is attributable to the size of the corpus. In the *WASTE MANAGEMENT* subcorpus, there were only 112 occurrences of *pollutant* and a portion of one of the books on *WASTE MANAGEMENT* included in the corpus made use of that categorization. Therefore, the terms *primary* and *secondary* were overrepresented in the *WASTE MANAGEMENT* contextonym list.

In view of the contextonyms of *pollutant* in *WASTE MANAGEMENT*, it became evident that POLLUTANT participates in many different frames in this contextual domain. Therefore, it can be considered hyperversatile. However, all of these frames tend to conceptualize POLLUTANTS as results from waste disposal or waste management processes. This raises the question of whether POLLUTANT in *WASTE MANAGEMENT* is a subconceptualization instead of a perspective. In other words, is the extension of POLLUTANT constrained to only those members of the category resulting from waste disposal or waste management processes?

As stated in §5.2.3, the distinction between subconceptualization and perspective is fuzzy, and this is an example. In our opinion, this is a case of perspectivization because other types of pollutants different from waste-related pollutants are also probably activated when the concept POLLUTANT is employed in this contextual domain. Consequently, the characteristic that POLLUTANT can be the result of waste-related processes has prototypical status (not necessary), and, therefore, does not limit the extension of the concept.

#### *5.4.1.2.3* WATER TREATMENT AND SUPPLY *contextonyms of* pollutant

The contextonyms of *pollutant* in *WATER TREATMENT AND SUPPLY* reveal that the extension of the concept in this domain is more limited than the one in the general environmental domain. POLLUTANT gives rise to a subconceptualization that coincides with the extension of the concept WATER POLLUTANT. In what follows, we outline the information that, based on the list of contextonyms in *WATER TREATMENT AND SUPPLY*, is relevant to the premeaning of POLLUTANT in this contextual domain.

Wastewater treatment plants have the function of making wastewater (for instance, sewage or industrial wastewater) suitable for discharge into the water cycle or for reuse. Among other things, this entails, the removal of pollutants.

Water pollution can be divided into two types depending on its source[96]. The first type is non-point-source pollution, which comes from many diffuse sources. This pollution affects water normally due to runoff that carries away pollutants. Point-source pollution has its origin from a single discrete, identifiable source that discharges pollutants into a water body such as industrial facilities or wastewater treatment plants. In most countries, discharge water needs to meet quality standards that limit the amount of pollutants. The disposal of sludge (the solid waste resulting from water treatment) is also required to comply with pollutant concentration limits.

Because of the origin of the texts, certain terms that point to the categorization of water pollutants in the United States appear in the contextonym list. In the USA, the Environmental Protection Agency (EPA) classifies water pollutants into three types: priority, conventional, and nonconventional:

- Priority pollutants are 126 chemical toxic pollutants for which the EPA has developed analytical test methods. Most of them are organic compounds. None of the priority pollutants are among the first 100 contextonyms of pollutant in this domain, but the terms *heavy* and *metal* do appear. Heavy metals (such as mercury and lead) are priority pollutants.

[96] This categorization is also applicable to air pollution, but, according to our corpus, it is more frequently used for water pollution.

- Conventional water pollutants are those that an American municipal wastewater plant can remove. As a consequence, this category includes common chemical and biological pollutants in municipal wastewater: biochemical-oxygen-demanding (BOD) substances, total suspended solids, pH-altering substances, fecal coliform bacteria, and oil and grease.

- Nonconventional pollutants include any other water pollutant not included in the other two categories, such as ammonia, nitrogen or phosphorus.

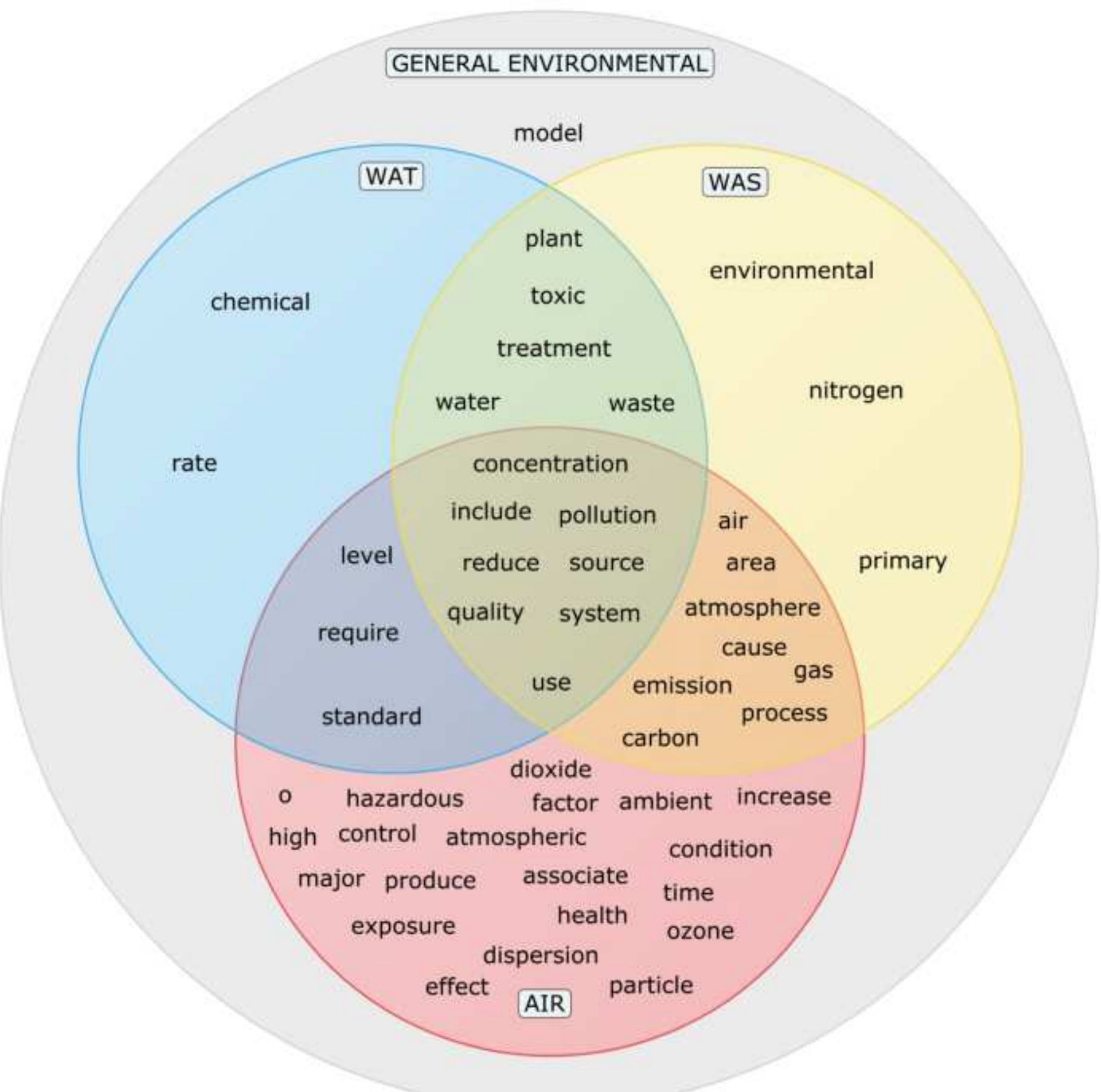

**Figure 31. Venn diagram of the first 50 general environmental contextonyms of POLLUTANT that are also present on the list of the first 50 contextonyms in *AIR QUALITY MANAGEMENT*, *WASTE MANAGEMENT*, and *WATER TREATMENT AND SUPPLY***

Since POLLUTANT is construed as a subconceptualization in *WATER TREATMENT AND SUPPLY*, the contextualized definition of POLLUTANT for this domain would forward to the contextualized definition of WATER POLLUTANT for the same contextual domain. In that definition, WATER POLLUTANT would be represented as a participant in a general frame that could receive the name of WATER POLLUTION. This frame would have many subframes, for instance, NON-POINT-SOURCE WATER POLLUTION and POINT-SOURCE WATER POLLUTION, which would in its turn have many other subframes.

##### *5.4.1.2.4 General environmental contextonyms of* pollutant

As part of the analysis of the 100 first general environmental contextonyms of *pollutant* (i.e., the contextonyms of *pollutant* in the MULTI corpus), we compared them with the contextonyms of POLLUTANT in its contextual domains. Our results show that 98 of the general environmental contextonyms were also contextonyms in the contextual domains. Only *method* and *fuel* were not contextonyms in the contextual domains. *Method* is a versatile term, and no clear relation between POLLUTANT and METHOD could be determined. *Fuel* refers to the emission of air pollutants by the combustion of fossil fuels, which is one of the main sources of air pollution (it is the 109$^{th}$ contextonym in *AIR QUALITY MANAGEMENT*).

To create the visual representation in Figure 31, we compared the 50 first contextonyms. As can be seen, the *AIR QUALITY MANAGEMENT* contextonyms are the most coincident with the general environmental ones, which indicates the prototypicality of AIR POLLUTION in comparison to other kinds of POLLUTION. The only contextonym not shared with its contextual domains is *model*[97], a term that did not convey a clear relation in the concordances because it made reference to very different kinds of model.

[97] Since *model* is the 77th *AIR QUALITY MANAGEMENT* contextonym, it was not among the terms that were not shared between the general environmental contextonyms and those of the contextual domains.

In conclusion, the general environmental contextonyms show that the general environmental premeaning encompasses parts of all the contextual domains, although the characteristics related to *AIR QUALITY MANAGEMENT* are slightly more relevant.

#### 5.4.1.3 Extraction of superordinate concepts and choice of genera for POLLUTANT

We extracted superordinate concept candidates from the definitions of other resources (the same definitions analyzed in §5.4.1.1), and from corpora by means of hypernymic knowledge patterns.

As explained in §3.6.3, the genera of definitions in EcoLexicon coincide with the superordinate concepts of the definiendum in the knowledge base. The general environmental definition corresponds to the general environmental superordinate, and each contextualized definition's genus is the corresponding contextual preferential superordinate concept. In order to determine the genus of each definition, we extracted superordinate candidates.

However, the superordinate candidates extracted from the definitions of other resources and corpora are only to be used as a guide, since the coherence of the conceptual hierarchies must take precedence. Moreover, if possible, the most prototypical frame invoked by the concept should be reflected in the genus.

This stage only affects the definition of POLLUTANT in the general environmental domain and *WASTE MANAGEMENT*, since both the definitions of AIR POLLUTANT and WATER POLLUTANT take POLLUTANT as a superordinate concept. For this reason, the genus of the definitions collected for these two domains were counted along with general environmental definitions (which include definitions from other domains or the whole environmental domain). The result from the extraction from the definitions of other resources organized by headword is represented in Table 57:

| | **GE** | **WAS** |
|---|---|---|
| **MATERIAL** | **3** | **0** |
| material | 2 | 0 |
| waste material | 1 | 0 |
| **CHEMICAL** | **4** | **0** |
| chemical | 3 | 0 |
| harmful chemical | 1 | 0 |
| **AGENT** | **1** | **0** |
| physical agent | 1 | 0 |
| **SUBSTANCE** | **28** | **1** |
| substance | 21 | 1 |
| emitted substance | 1 | 0 |
| chemical substance | 2 | 0 |
| harmful substance | 1 | 0 |
| inorganic or organic substance | 1 | 0 |
| undesirable substance | 2 | 0 |
| **CONTAMINANT** | **10** | **1** |
| contaminant | 9 | 1 |
| environmental contaminant | 1 | 0 |
| **COMPOUND** | **1** | **0** |
| compound | 1 | 0 |
| **GAS** | **1** | **0** |
| introduced gas | 1 | 0 |
| **LIQUID** | **1** | **0** |
| introduced liquid | 1 | 0 |
| **SOLID** | **1** | **0** |
| introduced solid | 1 | 0 |
| **THING** | **1** | **0** |
| living or not living thing | 1 | 0 |
| **BY-PRODUCT** | **1** | **0** |
| by-product of human activities | 1 | 0 |
| **COMPONENT** | **1** | **0** |
| chemical or physical component | 1 | 0 |
| **WASTE** | **1** | **0** |
| waste | 1 | 0 |
| **ORGANISM** | **1** | **0** |
| organism | 1 | 0 |
| **ENERGY** | **2** | **0** |

| | GE | WAS |
|---|---|---|
| energy | 2 | 0 |
| **IMPURITY** | **1** | **0** |
| impurity | 1 | 0 |
| **MIXTURE** | **1** | **0** |
| mixture of substances | 1 | 0 |

**Table 57. Superordinate concept candidates of POLLUTANT extracted from the definitions of other resources**

As can be seen, the superordinate SUBSTANCE is the most common in the environmental definitions. In *WASTE MANAGEMENT*, there are only two superordinate concept candidates: SUBSTANCE and CONTAMINANT.

This list of superordinate concepts was complemented with those extracted from the PANACEA corpus since no hypernymic knowledge-rich contexts for *pollutant* were extracted from the MULTI corpus. We only considered those cases where *pollutant* was not modifying or being modified by another lexical unit. We obtained a total of seven superordinate concepts candidates from the PANACEA corpus. Table 58 shows the concordances and Table 59 shows the results.

| | |
|---|---|
| PAN | terms, it is the "bloom" or... and littering. Pollutants A **pollutant** is a **waste material** that pollutes air, water or soil. Three |
| PAN | two major laboratories tested for 210 industrial compounds, **pollutants** , and other **chemicals** in the blood and urine of nine volunteers |
| PAN | organisms adapt. Pollutants may have entered the ecosystem. **Pollutants** are defined as any **agent** that causes harm. Pollutants can |
| PAN | affected by events far upstream, and **concentrate materials** such as **pollutants** and sediments. [14] Land run-off and industrial, |
| PAN | management practices, pest activity, beneficial organisms, fire, **pollutants** , storm damage, and other **factors** including climate change |
| PAN | can CO2 be considered a pollutant? A broader definition of **pollutant** is a **substance** that causes instability or discomfort to an |
| PAN | on delta flows and not enough on other **conditions** , such as **pollutants** and invasive species, that could be harming fish populations |

**Table 58. Hypernymic knowledge-rich contexts for the lemma *pollutant* (noun) in the PANACEA corpus**

| | GE |
|---|---|
| **FACTOR** | **1** |
| factor | 1 |
| **MATERIAL** | 2 |
| waste material | 1 |
| concentrate material | **1** |
| **CHEMICAL** | 1 |
| chemical | 1 |
| **AGENT** | 1 |
| agent | 1 |
| **SUBSTANCE** | **1** |

| substance | 1 |
|---|---|
| **CONDITION** | 1 |
| condition | 1 |

**Table 59. Superordinate concept candidates of POLLUTANT extracted from corpora**

Of the list of superordinate concepts extracted from the corpus, only two candidates were not present on the list extracted from the definitions: *factor*, and *condition*.

The combined results of the definitions of other resources and of hypernymic knowledge-rich contexts from corpora are shown in the following table (for the sake of clarity, only headwords are taken into account):

| | GE | AIR | WAS | WAT | TOTAL |
|---|---|---|---|---|---|
| **SUBSTANCE** | 20 | 0 | 1 | 9 | 30 |
| **CONTAMINANT** | 3 | 0 | 1 | 7 | 11 |
| **MATERIAL** | 4 | 0 | 0 | 1 | 5 |
| **CHEMICAL** | 4 | 0 | 0 | 1 | 5 |
| **AGENT** | 2 | 0 | 0 | 0 | 2 |
| **ENERGY** | 1 | 0 | 0 | 1 | 2 |
| **FACTOR** | 1 | 0 | 0 | 0 | 1 |
| **CONDITION** | 1 | 0 | 0 | 0 | 1 |
| **COMPOUND** | 1 | 0 | 0 | 0 | 1 |
| **GAS** | 1 | 0 | 0 | 0 | 1 |
| **LIQUID** | 1 | 0 | 0 | 0 | 1 |
| **SOLID** | 1 | 0 | 0 | 0 | 1 |
| **THING** | 1 | 0 | 0 | 0 | 1 |
| **BY-PRODUCT** | 1 | 0 | 0 | 0 | 1 |
| **COMPONENT** | 0 | 0 | 0 | 1 | 1 |
| **WASTE** | 1 | 0 | 0 | 0 | 1 |
| **ORGANISM** | 0 | 0 | 0 | 1 | 1 |
| **IMPURITY** | 0 | 0 | 0 | 1 | 1 |
| **MIXTURE** | 0 | 0 | 0 | 1 | 1 |

**Table 60. Superordinate concept candidates of POLLUTANT extracted from the definitions of other resources and from corpora**

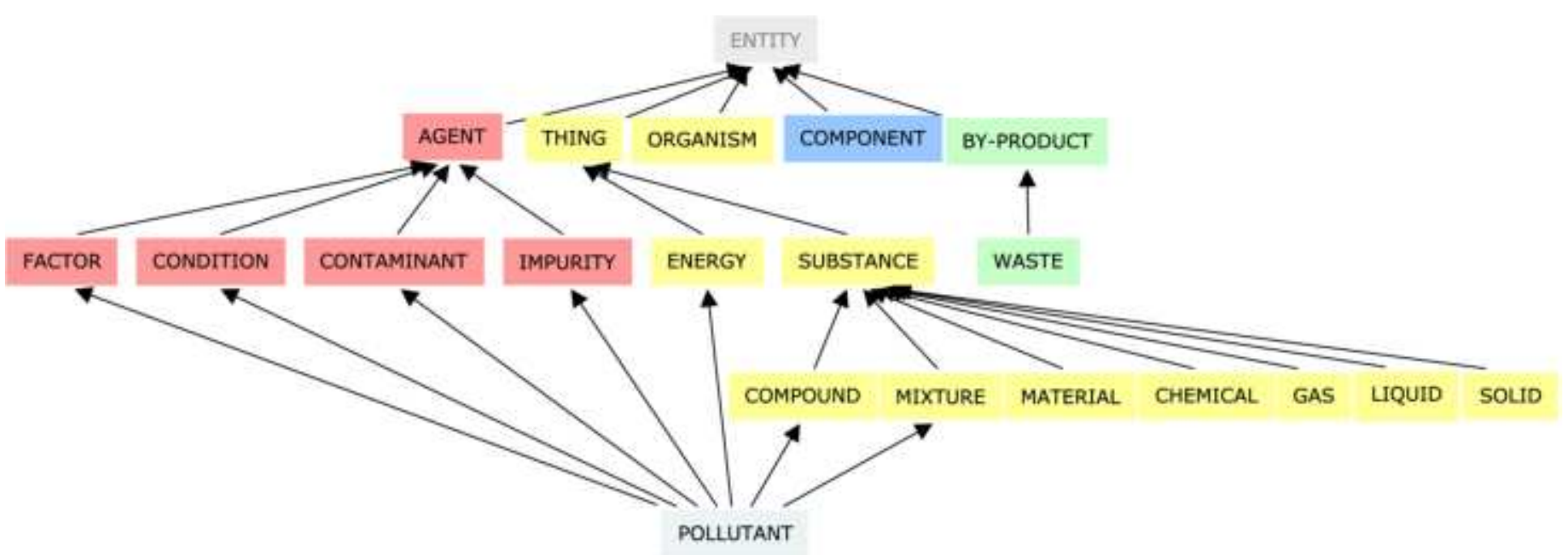

**Figure 32. Hierarchical organization of the superordinate concept candidates for POLLUTANT**

Based on these results, SUBSTANCE is the most frequent superordinate concept of POLLUTANT, followed by CONTAMINANT, MATERIAL, CHEMICAL, AGENT, and ENERGY. In the hierarchy shown in Figure 32, the telic concepts are in pale red; the constitutive concepts are in blue; the agentive concepts are in green; and the formal concepts are in yellow.

In the process of genus choice, it is important to take into account the role that the concept plays in that domain and choose a type of genus accordingly. Both in the general environmental domain and in the domain of *WASTE MANAGEMENT*, its main role is having a harmful effect, which is a telic trait. Therefore, a telic genus should be chosen in both cases: CONTAMINANT, AGENT, FACTOR, CONDITION, or IMPURITY.

FACTOR and CONDITION were discarded because they are goal-derived categories (§2.1.3.3.1). They were both extracted from corpora. In the first case, POLLUTANT was categorized as a FACTOR that affects forest health. In the second case, POLLUTANT was conceptualized as a CONDITION that may be responsible for the decline of fish population in the Sacramento-San Joaquin River Delta (California). As for IMPURITY, it is not suitable either because it excludes POLLUTANTS that are not SUBSTANCES such as HEAT, LIGHT, SOUND or MICROORGANISMS.

CONTAMINANT also poses the problem that its extension does not encompass the whole extension of POLLUTANT. CONTAMINANT includes SUBSTANCES and MICROORGANISMS that are in an environment in places where they are not found naturally or in a concentration higher than usual for that environment. In most cases, CONTAMINANT comprises the extension of POLLUTANT (POLLUTANT can be defined as a harmful CONTAMINANT in those cases[98]). However, LIGHT, SOUND, and HEAT are not part of the extension of CONTAMINANT, but are considered POLLUTANTS (although they are not prototypical members of the category). In other words, whereas most POLLUTANTS are types of CONTAMINANT, not all POLLUTANTS are CONTAMINANTS. Therefore, CONTAMINANT cannot be used as the genus of POLLUTANT.

AGENT is a generic concept that includes any kind of ENTITY that produces a result or is used to produce a result. Since it does encompass the whole extension of POLLUTANT, we designated AGENT as the general environmental superordinate concept and as the *WASTE MANAGEMENT* preferential superordinate concept for POLLUTANT.

#### 5.4.1.4 Definition of POLLUTANT

This section presents the flexible definition for POLLUTANT and discusses it. However, it is first necessary to explain certain modifications that we made to the definitional templates used in EcoLexicon (§3.4.2).

The first two rows contain the concept being defined (with an indication of its contextual domain), and the definition. The remaining rows are reserved for the definitional conceptual propositions. A proposition row has six columns:

1. Identification number: All propositions are identified by the letter P followed by a number: P1, P2, P3, and so on.
2. Type: The following abbreviations are used to indicate the type of proposition (§3.6.3):
    - Direct proposition (DI)
    - Indirect specified proposition (SP)

[98] Sometimes, the term *contaminant* is also used erroneously as a synonym of *pollutant*.

- Indirect explicit proposition (EX)
- Framing proposition (FR) (they are always placed at the end of the template)

3. Status: In the third column, we distinguish whether the proposition is absolute, prototypical, absolute negative, or prototypical negative. Moreover, although relations have their inverse in EcoLexicon, it is important to note that the inverse relation does not necessarily have the same status. The following symbols are used:
   - Absolute (@) A proposition is absolute when all the members of the conceptual category being defined meet that condition. For instance, the proposition "METHANE *made-of* HYDROGEN" is absolute for METHANE because all the members of the category METHANE are made of HYDROGEN. However, as stated above, it does not follow that the inverse "HYDROGEN *component-of* METHANE" is absolute as well because not all the members of the category HYDROGEN are components of METHANE. The absolute symbol can be combined with the negative symbol.
   - Prototypical (π) A proposition is prototypical when the members of the conceptual category being defined prototypically meet that condition. For instance, if the proposition "POLLUTANT *result-of* ARTIFICIAL PROCESS" is marked as prototypical, it means that a POLLUTANT is prototypically the result of an ARTIFICIAL PROCESS, although not all members of the category POLLUTANT meet that condition. This sort of prototypical proposition is represented on the definitional template but not codified in EcoLexicon (because it would give rise to false inheritance). In definitions, prototypical propositions are inherited, and they maintain the prototypical status in the subordinates concepts (unless overriden in the corresponding template). For instance, "POLLUTANT *result-of* ARTIFICIAL PROCESS has prototypical status, and it is inherited by AIR POLLUTANT as "AIR POLLUTANT *result-of* ARTIFICIAL PROCESS" also with prototypical status. It can be combined with the negative symbol.

- Negative (!): This symbol is used to make the linking relation negative. For example, if a proposition containing the relation "*affects*" is marked with the symbol !, the relation is to be interpreted as "does not affect". The negative symbol is always used in combination with either the absolute symbol or the prototypical symbol. For instance, on the one hand, if the proposition "CHEMICAL DECOMPOSITION *affects* CHEMICAL ELEMENT" is marked with "@!" (absolute negative), this would mean that CHEMICAL COMPOSITION never affects CHEMICAL ELEMENTS. On the other hand, if the proposition "CHEMICAL DECOMPOSITION *affects* CHEMICAL ELEMENT" is marked with "π!" (prototypical negative), this would mean that CHEMICAL COMPOSITION normally does not affect CHEMICAL ELEMENTS. Negative propositions are represented in the definitional template but are not codified in EcoLexicon because the knowledge base currently does not allow it.

4. First concept(s): This column is used for the first concept or concepts of the proposition. If more than one concept is included in the cell, they are linked by any of these operators[99]:

[99] The order of precedence is: x, & and |. When there are several operators of the same type, they operate from left to right. If the order of precedence needs to be changed, parentheses are used.

- & (conjunction): This operator links concepts in a proposition when both of them apply. It corresponds to what in natural language is expressed as "and". For instance: "METHANE *made-of* CARBON & HYDROGEN" means that METHANE contains both CARBON and HYDROGEN at the same time.
- | (inclusive disjunction): This operator corresponds to what in natural language is expressed as "and/or". For instance, "OZONE-DEPLETING SUBSTANCE *made-of* CHLORINE | BROMINE", means that an OZONE-DEPLETING SUBSTANCE contains CHLORINE, BROMINE, or both.
- x (exclusive disjunction): This operator expresses that only one of concepts is applicable separately but never at the same time. For instance, the prototypical proposition "POLLUTANT *result-of* INCINERATION x LANDFILL USE" means that a pollutant is usually the result of either incineration or landfill use. The same member of the category POLLUTANT cannot result from both processes simultaneously.

5. Relation: Only the relations in EcoLexicon were used. However, in the case of the *affects* relation, a specification in brackets was added when deemed necessary.
6. Second concept(s). This column is used for the second concept or concepts of the proposition. The may also make use of the operators presented above.

Finally, it is important to note that inferred propositions are included in the templates if they are represented in the definition. For the reasons explained on §3.6.5, they are not codified in EcoLexicon propositions.

##### 5.4.1.4.1 AIR QUALITY MANAGEMENT *contextualized definition of POLLUTANT*

Given that POLLUTANT in *AIR QUALITY MANAGEMENT* is subconceptualized as AIR POLLUTANT, the contextualized definition of POLLUTANT in this contextual domain consists of a hyperlink to the *AIR QUALITY MANAGEMENT* contextualized definition of that concept:

| POLLUTANT (*AIR QUALITY MANAGEMENT*) |
|---|
| → AIR POLLUTANT (*AIR QUALITY MANAGEMENT*) |

**Table 61. Filled-out definitional template of POLLUTANT (*AIR QUALITY MANAGEMENT* contextual domain)**

##### 5.4.1.4.2 WASTE MANAGEMENT *contextualized definition of POLLUTANT*

POLLUTANT in *WASTE MANAGEMENT* only has one contextual preferential superordinate concept (AGENT), which coincides with the general environmental one. Its hierarchical path is POLLUTANT>AGENT>ENTITY. ENTITY is considered a semantic primitive and it has no propositions associated with it. Therefore, POLLUTANT only inherits propositions from AGENT. Since AGENT does not show significant contextual variation, it only has one definition and its propositions are not contextualized:

| AGENT (all domains) | | | | | |
|---|---|---|---|---|---|
| Entity that has an effect on another entity or process, causes a process or is used for some purpose. | | | | | |
| P1 | DI | @ | AGENT | *type-of* | ENTITY |
| P2 | DI | π | AGENT | *affects* | ENTITY \| PROCESS |
| P3 | DI | π | AGENT | *causes* | PROCESS |
| P4 | DI | π | AGENT | *effects* | PROCESS |

**Table 62. Filled-out definitional template of AGENT (all domains)**

Table 63 shows the contextualized definition of POLLUTANT in *WASTE MANAGEMENT*.

| POLLUTANT (*WASTE MANAGEMENT*) |
|---|
| Agent that adversely affects the environment (e.g., air, water, and soil) or human health. Types of pollutant include substances (e.g., carbon dioxide, sulfur dioxide, or dioxins), pathogens, and certain forms of |

<table>
<tr><td colspan="6">energy (e.g., heat). A pollutant is at an unacceptably high concentration or is in a place where it is not naturally present. It is generally the result of human actions. For instance, incineration emits pollutants into the atmosphere, and landfills emit air pollutants and can cause the leaching of pollutants into the soil and groundwater. The release of pollutants into the environment normally needs to comply with regulations.</td></tr>
<tr><td>P1</td><td>DI</td><td>@</td><td>POLLUTANT</td><td><i>type-of</i></td><td>AGENT</td></tr>
<tr><td>P2</td><td>SP</td><td>@</td><td>POLLUTANT</td><td><i>affects (adversely)</i></td><td>ENVIRONMENT | AIR | WATER | SOIL | HUMAN HEALTH</td></tr>
<tr><td>P3</td><td>DI</td><td>@</td><td>CARBON DIOXIDE & SULFUR DIOXIDE & DIOXIN & PATHOGEN & HEAT</td><td><i>type-of</i></td><td>POLLUTANT</td></tr>
<tr><td>P4</td><td>DI</td><td>π</td><td>POLLUTANT</td><td><i>has-attribute</i></td><td>(HIGH) CONCENTRATION</td></tr>
<tr><td>P5</td><td>DI</td><td>π</td><td>POLLUTANT</td><td><i>result-of</i></td><td>ARTIFICIAL PROCESS</td></tr>
<tr><td>P6</td><td>DI</td><td>π</td><td>POLLUTANT</td><td><i>result-of</i></td><td>INCINERATION X LANDFILL USE</td></tr>
<tr><td>P7</td><td>DI</td><td>π</td><td>POLLUTANT</td><td>affected-by</td><td>REGULATION</td></tr>
<tr><td>P8</td><td>FR</td><td>@</td><td>AIR & WATER & SOIL</td><td><i>part-of</i></td><td>ENVIRONMENT</td></tr>
<tr><td>P9</td><td>FR</td><td>@</td><td>CARBON DIOXIDE & SULFUR DIOXIDE & DIOXIN</td><td><i>type-of</i></td><td>SUBSTANCE</td></tr>
<tr><td>P10</td><td>FR</td><td>@</td><td>HEAT</td><td><i>type-of</i></td><td>ENERGY</td></tr>
<tr><td>P11</td><td>FR</td><td>@</td><td>INCINERATION & LANDFILL USE</td><td><i>type-of</i></td><td>ARTIFICIAL PROCESS</td></tr>
<tr><td>P12</td><td>FR</td><td>@</td><td>LANDFILL LEACHING</td><td><i>result-of</i></td><td>LANDFILL USE</td></tr>
<tr><td>P13</td><td>FR</td><td>π</td><td>LANDFILL LEACHING</td><td><i>affects (pollutes)</i></td><td>SOIL | GROUNDWATER</td></tr>
</table>

**Table 63. Filled-out definitional template of POLLUTANT (*WASTE MANAGEMENT* contextual domain)**

P1 is the proposition that establishes the genus in all definitional templates. Other propositions might be inherited from it, like P2 in this case, which is inherited from “AGENT *affects* ENTITY | PROCESS”. Since POLLUTANT in *WASTE MANAGEMENT* has no non-preferential superordinate concepts, it only inherits propositions from its preferential genus.

Moreover, in P8, a framing proposition, it is specified that AIR, WATER and SOIL are part of the ENVIRONMENT because they are the most prototypical patients of the process of POLLUTION.

Many of the analyzed definitions of POLLUTANT used SUBSTANCE as a genus. However, since not all POLLUTANTS are SUBSTANCES, it cannot be the genus of the definition. Neither can it be said that SUBSTANCES are kinds of POLLUTANT. Instead, by means of P3, we provided the most prototypical examples of POLLUTANTS, according to the contextonyms of POLLUTANT in *WASTE MANAGEMENT*. Furthermore, in P9, we specified that some of those kinds of POLLUTANT are, at the same time, SUBSTANCES. P10 was also added for the same reason but in relation to HEAT and ENERGY. Both P9 and P10 were included in order to convey the type of entities that can be considered POLLUTANTS, since such specification is not inherited from the genus AGENT. Moreover, it is important to note that CARBON DIOXIDE and SULFUR DIOXIDE are specifically AIR POLLUTANTS, therefore, the proposition "CARBON DIOXIDE & SULFUR DIOXIDE *type-of* POLLUTANT" was inferred from "CARBON DIOXIDE & SULFUR DIOXIDE *type-of* AIR POLLUTANT" and "AIR POLLUTANT *type-of* POLLUTANT".

In P4, due to the lack of expressiveness of the *has-attribute* propositions in EcoLexicon, we further specified it in brackets. Furthermore, it is not possible to encode that POLLUTANTS may be in a place where they are not naturally present by employing EcoLexicon propositions.

P5 is marked as prototypical (π) because not all POLLUTANTS are the result of ARTIFICIAL PROCESSES, although most are. The same is true for P7 since not all POLLUTANTS are regulated by standards, and regulations are more or less strict depending on the country. As can be seen, if a proposition has prototypical status instead of absolute status, it is reflected in the wording of the definition by the addition of an adverb (*normally, generally, usually*, etc.) or the use of modal verbs (*can*, *could*, *may*, etc.).

In P6, INCINERATION and LANDFILL USE are the most relevant subordinates of ARTIFICIAL PROCESS that cause POLLUTION in *WASTE MANAGEMENT*. That relation is represented with P11. An exclusive disjunction was used to link

INCINERATION and LANDFILL USE IN P6 because the same instance of POLLUTANT cannot be simultaneously the result of both INCINERATION and LANDFILL USE.

For their part, P12 is a framing proposition needed to specify the way LANDFILL USE can lead to POLLUTION (by LANDIFLL LEACHING), and P13 specifies that LANDFILL LEACHING typically pollutes SOIL and GROUNDWATER.

Some of the above-mentioned propositions are not directly established by POLLUTANT, but are inferred from "POLLUTANT *causes* POLLUTION". We did not represent this proposition in the definition to avoid the use of terms morphologically related to the term denominating the definiendum. However, users can see that proposition on the conceptual map in EcoLexicon.

##### 5.4.1.4.3 WATER TREATMENT AND SUPPLY *contextualized definition of POLLUTANT*

POLLUTANT in *WATER TREATMENT AND SUPPLY* is subconceptualized as WATER POLLUTANT. As a consequence, the contextualized definition of POLLUTANT in this contextual domain consists of a hyperlink to the *WATER TREATMENT AND SUPPLY* contextualized definition of that concept:

| POLLUTANT (*WATER TREATMENT AND SUPPLY*) |
| --- |
| → WATER POLLUTANT (*WATER TREATMENT AND SUPPLY*) |

**Table 64. Filled-out definitional template of POLLUTANT (*WATER TREATMENT AND SUPPLY* contextual domain)**

##### *5.4.1.4.4 General environmental definition of POLLUTANT*

The general environmental definition (Table 65) greatly resembles the one in *WASTE MANAGEMENT* because the *WASTE MANAGEMENT* premeaning is a perspective of the general environmental premeaning, and they both share the same genus (AGENT) (Table 62).

<table>
<tr><td colspan="6">POLLUTANT (General environmental definition)</td></tr>
<tr><td colspan="6">Agent that adversely affects the environment (e.g., air, water, and soil) or human health. Types of pollutant include substances (e.g., carbon dioxide, nitrogen oxides, or sulfur dioxide), pathogens, and certain forms of energy (e.g., heat, light, or sound). A pollutant is at an unacceptably high concentration or is in a place where it is not naturally present. It is generally the result of human actions, for instance, industrial activity or vehicle exhaust. The release of pollutants into the environment normally needs to comply with regulations.</td></tr>
<tr><td>P1</td><td>DI</td><td>@</td><td>POLLUTANT</td><td><i>type-of</i></td><td>AGENT</td></tr>
<tr><td>P2</td><td>SP</td><td>@</td><td>POLLUTANT</td><td><i>affects (adversely)</i></td><td>ENVIRONMENT | AIR | WATER | SOIL | HUMAN HEALTH</td></tr>
<tr><td>P3</td><td>DI</td><td>@</td><td>CARBON DIOXIDE & NITROGEN OXIDE & SULFUR DIOXIDE & PATHOGEN & HEAT & LIGHT & SOUND</td><td><i>type-of</i></td><td>POLLUTANT</td></tr>
<tr><td>P4</td><td>DI</td><td>π</td><td>POLLUTANT</td><td><i>has-attribute</i></td><td>(HIGH) CONCENTRATION</td></tr>
<tr><td>P5</td><td>DI</td><td>π</td><td>POLLUTANT</td><td><i>result-of</i></td><td>ARTIFICIAL PROCESS</td></tr>
<tr><td>P6</td><td>DI</td><td>π</td><td>POLLUTANT</td><td><i>result-of</i></td><td>INDUSTRIAL ACTIVITY X VEHICLE EXHAUST</td></tr>
<tr><td>P7</td><td>DI</td><td>π</td><td>POLLUTANT</td><td><i>affected-by</i></td><td>REGULATION</td></tr>
<tr><td>P8</td><td>FR</td><td>@</td><td>AIR & WATER & SOIL</td><td><i>part-of</i></td><td>ENVIRONMENT</td></tr>
<tr><td>P9</td><td>FR</td><td>@</td><td>CARBON DIOXIDE & NITROGEN DIOXIDE & SULFUR DIOXIDE</td><td><i>type-of</i></td><td>SUBSTANCE</td></tr>
<tr><td>P10</td><td>FR</td><td>@</td><td>HEAT & LIGHT & SOUND</td><td><i>type-of</i></td><td>ENERGY</td></tr>
<tr><td>P11</td><td>FR</td><td>@</td><td>INDUSTRIAL ACTIVITY & VEHICLE EXHAUST</td><td><i>type-of</i></td><td>ARTIFICIAL PROCESS</td></tr>
</table>

**Table 65. Filled-out definitional template of POLLUTANT (general environmental domain)**

Several of the propositions of the general environmental definition are the same as in the *WASTE MANAGEMENT* definition: P2, P5, P7, and P8.

In P3, the most prototypical examples of POLLUTANTS, based on the general environmental contextonyms of POLLUTANT, are represented. LIGHT and SOUND were also added because they were important to show the extension of the concept. As with the *WASTE MANAGEMENT* definition of POLLUTANT, in P9 and P10, we added that those prototypical POLLUTANTS are, at the same time, SUBSTANCES or ENERGY. The propositions "CARBON DIOXIDE & NITROGEN OXIDE & SULFUR DIOXIDE *type-of* POLLUTANT" were inferred from "CARBON DIOXIDE & NITROGEN OXIDE & SULFUR DIOXIDE *type-of* AIR POLLUTANT" and "AIR POLLUTANT *type-of* POLLUTANT".

As can be observed, the examples of source of POLLUTANTS in P6 are not INCINERATION and LANDFILL USE, but rather INDUSTRIAL ACTIVITY and VEHICLE EXHAUST, following the analysis of the general environmental contextonyms. Finally, P11 specifies that INDUSTRIAL ACTIVITY and VEHICLE EXHAUST are types of ARTIFICIAL PROCESS.

##### *5.4.1.4.5 Flexible definition of POLLUTANT*

Table 66 shows the flexible definition of POLLUTANT. Users would access the definitions separately, depending on the contextual domain chosen in EcoLexicon.

| POLLUTANT | |
|---|---|
| *GENERAL ENVIRONMENTAL* | Agent that adversely affects the environment (e.g., air, water, and soil) or human health. Types of pollutant include substances (e.g., carbon dioxide, nitrogen oxides, or sulfur dioxide), pathogens, and certain forms of energy (e.g., heat, light, or sound). A pollutant is at an unacceptably high concentration or is in a place where it is not naturally present. It is generally the result of human actions, for instance, industrial activity or vehicle exhaust. The release of pollutants into the environment normally needs to comply with regulations. |

| *AIR QUALITY MANAGEMENT* | → AIR POLLUTANT (*AIR QUALITY MANAGEMENT*) |
|---|---|
| *WASTE MANAGEMENT* | Agent that adversely affects the environment (e.g., air, water, and soil) or human health. Types of pollutant include substances (e.g., carbon dioxide, sulfur dioxide, or dioxins), pathogens, and certain forms of energy (e.g., heat). A pollutant is at an unacceptably high concentration or is in a place where it is not naturally present. It is generally the result of human actions. For instance, incineration emits pollutants into the atmosphere, and landfills emit air pollutants and can cause the leaching of pollutants into the soil and groundwater. The release of pollutants into the environment normally needs to comply with regulations. |
| *WATER TREATMENT AND SUPPLY* | → WATER POLLUTANT (*WATER TREATMENT AND SUPPLY*) |

**Table 66. Flexible definition of POLLUTANT**

## 5.4.2 A concept with perspectives: CHLORINE

As shown in Table 67, CHLORINE appears with a frequency of over 64 occurrences in the domains of *AIR QUALITY MANAGEMENT*, *CHEMISTRY*, and *WATER TREATMENT AND SUPPLY*.

| | AGR | AIR | ATM | BIO | CEN | CHE | CIV | ENE | GEO | HYD | PHY | SOI | WAS | WAT |
|---|---|---|---|---|---|---|---|---|---|---|---|---|---|---|
| chlorine | - | 75 | - | - | - | 78 | - | - | - | - | - | - | - | 165 |

**Table 67. Frequency of the term *chlorine* (noun) in the corresponding contextual domains according to TermoStat**

For the creation of the flexible definition of CHLORINE, we followed the same steps as for POLLUTANT.

### 5.4.2.1 Definitions of CHLORINE from other resources

A total of 29 definitions of CHLORINE were extracted from different terminological resources. The complete list of definitions can be found in Annex 5. Two of those definitions belonged to the domain of *AIR QUALITY*

*MANAGEMENT*, two to *CHEMISTRY*, 12 to *WATER TREATMENT AND SUPPLY*, and 13 were environmental definitions (i.e., belonging to other environmental domains or encompassing the whole environmental domain).

The following tables summarize the characteristics represented in the environmental definitions (Table 68), *AIR QUALITY MANAGEMENT* definitions (Table 69), *CHEMISTRY* definitions (Table 70) and *WATER TREATMENT AND SUPPLY* definitions (Table 71).

| CHLORINE (general environmental) | |
|---|---|
| *type of* | element<br>gas |
| *has attribute* | gaseous<br>green/greenish yellow<br>halogen<br>heavy<br>highly toxic<br>necessary for plant growth<br>persistent in the environment<br>poisonous (to fish and invertebrates)<br>respiratory irritant<br>strong/pungent odor<br>very/highly reactive<br>widespread |
| *occurs naturally in* | biological tissues<br>earth crust<br>salt lakes<br>sea-water<br>underground deposits |
| *has usual form* | common salt (NaCl) |
| *used to produce* | chlorinated organic solvents (such as polychlorinated biphenyls (PCBs))<br>hypochlorite bleaches<br>insecticides<br>organochloride pesticides (such as DDT)<br>pharmaceuticals<br>plastics (polyvinyl chloride plastics (PVC), thermoplastics)<br>polycarbonates and polyurethanes (as intermediate) |
| *causes* | ozone hole/destruction of ozone |
| *used as* | biocide |
| | (water) disinfectant<br>sterilizing agent<br>oxidizing agent<br>bleach |
| *kills* | bacteria<br>algae |
| *has atomic number* | 17 |
| *has atomic weight* | 35.453 |
| *has valence* | 1- |
| *has isotopes* | chlorine-35, chlorine-33, chlorine-39 |
| *turns liquid at* | -34° C |
| *is obtained by* | salt electrolysis |

**Table 68. Characteristics of CHLORINE represented in the definitions from other resources that belong to other environmental domains or the whole environmental domain**

| CHLORINE (*AIR QUALITY MANAGEMENT*) | |
|---|---|
| *type of* | gas<br>halogen |
| *depletes* | ozone (when released in stratosphere) |
| *is part of* | CFCs<br>HCFCs<br>methyl chloroform |
| *has attribute* | greenish-yellow<br>strong odor |
| *used as* | water disinfectant |

**Table 69. Characteristics of CHLORINE represented in the definitions from other resources that belong to the domain of *AIR QUALITY MANAGEMENT***

| CHLORINE (*CHEMISTRY*) | |
|---|---|
| *type of* | element |
| *has attribute* | halogen<br>oxidizing<br>poisonous<br>very reactive |
| *has atomic number* | 17 |
| *has atomic weight* | 35.453 |
| *has density* | 3.214 g dm$^{-3}$; |
| *has melting point* | -100.98°C |
| *has boiling point* | –34.6°C |
| *has usual form* | carnallite<br>halite<br>sodium chloride |

| | sylvite |
|---|---|
| *is obtained by* | Downs process for making sodium<br>electrolysis of brine |
| *used for* | bleaching<br>chlorination of drinking water<br>manufacture of organic chemicals |
| *has oxidation states* | 1, 3, 5, 7 |

**Table 70. Characteristics of CHLORINE represented in the definitions from other resources that belong to the domain of *CHEMISTRY***

| CHLORINE (*WATER TREATMENT AND SUPPLY*) | |
|---|---|
| *type of* | chemical<br>disinfectant<br>element<br>gas |
| *is added to* | water |
| *kills/ protects from* | bacteria<br>disease-causing organisms<br>germs<br>small organisms |
| *used as* | oxidizing agent (for organic mater, manganese, iron, hydrogen sulfide)<br>water disinfectant |
| *has attribute* | gas<br>greenish-yellow<br>liquid<br>strong odor<br>widely used |
| *used in* | swimming pools<br>waste treatment plants |

**Table 71. Characteristics of CHLORINE represented in the definitions from other resources that belong to the domain of *WATER TREATMENT AND SUPPLY***

The environmental definitions represent a variety of characteristics: color, standard state, toxicity, odor, reactivity, atomic number, atomic weight, usual location, natural form, how humans use it, the consequences of its use, etc. Each definition focuses on certain of those characteristics.

As for the two definitions belonging to the domain of *AIR QUALITY MANAGEMENT*, one of them focuses on its role in ozone depletion, whereas the other one gives a general description of the concept, without reference to it role in the domain. This is the reason why in Table 69, the attribute of strong odor or its function as water disinfectant are listed, although they are not relevant in the domain.

The chemical definitions offer some characteristics that did not appear in other definitions such as density, boiling point, mode of production and oxidation states.

Finally, the *WATER TREATMENT AND SUPPLY* definitions for CHLORINE focus on its use in water treatment, such as what type of undesirable elements in water are affected by chlorine and in which kind of facilities chlorine is employed.

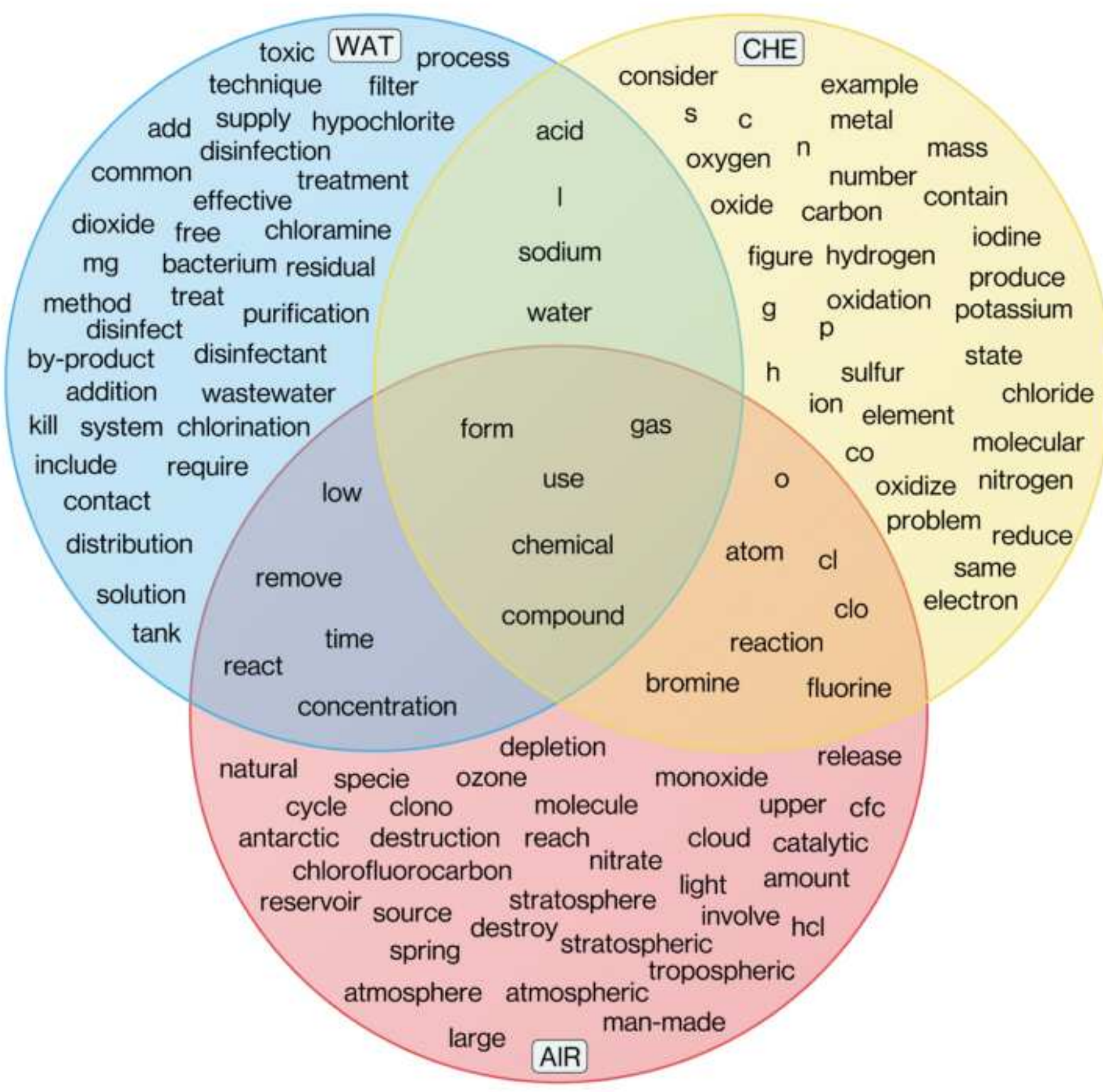

**Figure 33. Venn diagram of the first 50 contextonyms of *chlorine* in *AIR QUALITY MANAGEMENT*, *CHEMISTRY*, and *WATER TREATMENT AND SUPPLY***

### 5.4.2.2 Contextonyms of *chlorine*

In order to obtain a clear image of the contextual variation that the concept chlorine experiences in its contextual domains, we analyzed the first 100 contextonyms for the MULTI corpus and the corresponding subcorpora (*AIR QUALITY MANAGEMENT*, *CHEMISTRY,* and *WASTE MANAGEMENT*). The complete lists are in Annex 6.

Figure 33 is a visual representation in a Venn diagram of the first 50 contextonyms for the contextual domains of *AIR QUALITY MANAGEMENT*, *CHEMISTRY* and *WASTE MANAGEMENT*.

The following subsections outline the results of the analysis of the contextonyms (with the help of their corresponding concordances) in each contextual domain and of the general environmental contextonyms (extracted from the whole MULTI corpus).

#### *5.4.2.2.1* AIR QUALITY MANAGEMENT *contextonyms of* chlorine

The most important contextonyms of *chlorine* in *AIR QUALITY MANAGEMENT* show how CHLORINE is conceptualized as contributing to the depletion of stratospheric ozone. The process of ozone depletion is a series of subprocesses that start with the emission of ozone-depleting substances, which are man-made compounds containing chlorine or bromine, such as chlorofluorocarbons (CFCs), hydrochlorofluorocarbons (HCFCs). Some ozone-depleting substances contain fluorine as well, but this chemical element does not contribute significantly to ozone depletion. Although some natural processes release chlorine to the atmosphere, man-made substances are the main source of stratospheric chlorine, which has experienced an increase in concentration since the 1960s.

Ozone-depleting substances do not destroy ozone ($O_3$) directly. Firstly, they reach the stratosphere transported by air motions. Then, they are converted by ultraviolet light and chemical reactions into hydrochloric acid (HCl), and chlorine nitrate ($ClONO_2$), which are gasses that act as chlorine reservoirs species. Those two compounds are finally converted into chlorine monoxide (ClO) (especially in Antarctic polar stratospheric clouds) or free chlorine atoms, which react with ozone molecules in a catalytic cycle destroying large amounts of ozone before being removed from the stratosphere.

##### *5.4.2.2.2* WATER TREATMENT AND SUPPLY *contextonyms of* chlorine

The first 50 contextonyms for *chlorine* in *WATER TREATMENT AND SUPPLY* clearly show that CHLORINE is part of a frame that could be called WATER DISINFECTION. Disinfection with chlorine is called *chlorination*, and it is a treatment that consists of the addition of chlorine to water in order to deactivate or kill pathogens, such as bacteria or viruses. It is usually carried out in a chlorine-contact tank. The main forms in which it is used are chlorine gas ($Cl_2$) sodium hypochlorite, calcium hypochlorite, chlorine dioxide or chloramine.

Chlorination is considered to be part of the tertiary treatment of water. The primary includes mechanical methods (e.g., sedimentation), and the secondary treatment includes biological methods (e.g., the activated sludge process). Filtration is included in many different forms during primary and secondary treatment.

While chlorination is an effective method of disinfection, it has two main drawbacks. Chlorine reacts with organic compounds that may be present in the water to form harmful by-products: trihalomethanes and haloacetic acids. Moreover, since residual chlorine in treated water is toxic to aquatic life, it needs to be removed before discharging in aquatic ecosystems through a process called *dechlorination*. The presence of high levels of chlorine in water is always the result of chlorination. Therefore, the process of DECHLORINATION could be regarded as an optional phase in the CHLORINATION frame, in which CHLORINE is a participant.

##### *5.4.2.2.3* CHEMISTRY *contextonyms of* chlorine

CHEMICAL ELEMENTS tend to be hyperversatile in the domain of *CHEMISTRY* because they can participate in many different types of process. CHLORINE is not an exception and the fact that its contextonyms in *CHEMISTRY* are very varied, and point to many different processes indicates its hyperversatility in this domain.

The first three contextonyms correspond to its symbol (*Cl*), a hyperonym (*element*) and its standard state and hyperonym (*gas*). The analysis also shows that chlorine possesses a strong oxidizing power, which means that it has the tendency to react and cause oxidation (loss of electrons) on other elements and simultaneously undergo reduction. As well as the other halogen elements (i.e., fluorine, bromine, iodine, and astatine), it is diatomic in its elemental state, and when it reacts with a metal (e.g., sodium or potassium) it results in salts (such as sodium chloride), which are ionic compounds.

Some of its most important contextonyms are other chemical elements with which it tends to form compounds: *oxygen* (e.g., chlorine monoxide), *hydrogen* (e.g., hydrogen chloride (or hydrochloric acid in water)), *sulfur* (e.g., sulfur dichloride), *nitrogen* (e.g., chlorine nitrate), and *carbon* (e.g., carbon tetrachloride).

#### *5.4.2.2.4 General environmental contextonyms of* chlorine

As can be seen in Figure 34, the first 50 contextonyms in the general environmental domain coincide almost entirely with one of the three relevant contextual domains. The exceptions are the terms *product* and *PCB*. In the concordances, *product* mainly refers to commodities manufactured using chlorine. As for PCB, it is, in fact, a chlorine-containing product that, before being banned for its high toxicity, was widely employed in many different industries. In view of these two contextonyms, an example of some of the most common products made with chlorine was included in our general environmental definition of CHLORINE.

We also contrasted the first 100 general environmental contextonyms with those of the contextual domains. The general environmental list had the following exclusive contextonyms (including *PCB* and *product*): *ammonia*, *chlorinate*, *cost*, *high*, *material*, *ppm*, and *waste.*

*High* and *material* were too versatile to be able to determine a clear relation with CHLORINE. As for the remaining ones, they are all related to the contextual domain of *WATER TREATMENT AND SUPPLY*. According to the concordances, *ammonia* and *chlorine* are related because they form chloramine together (employed in chlorination). *Chlorinate* is the verb for chlorination. *Cost* is mostly used to make reference to the cost of the different chlorine-derived disinfectants. *Waste* appears on the list due to the alternative spelling of *wastewater* as two separate words. Finally, *ppm* stands for *parts per million*, which is a usual unit of measurement to

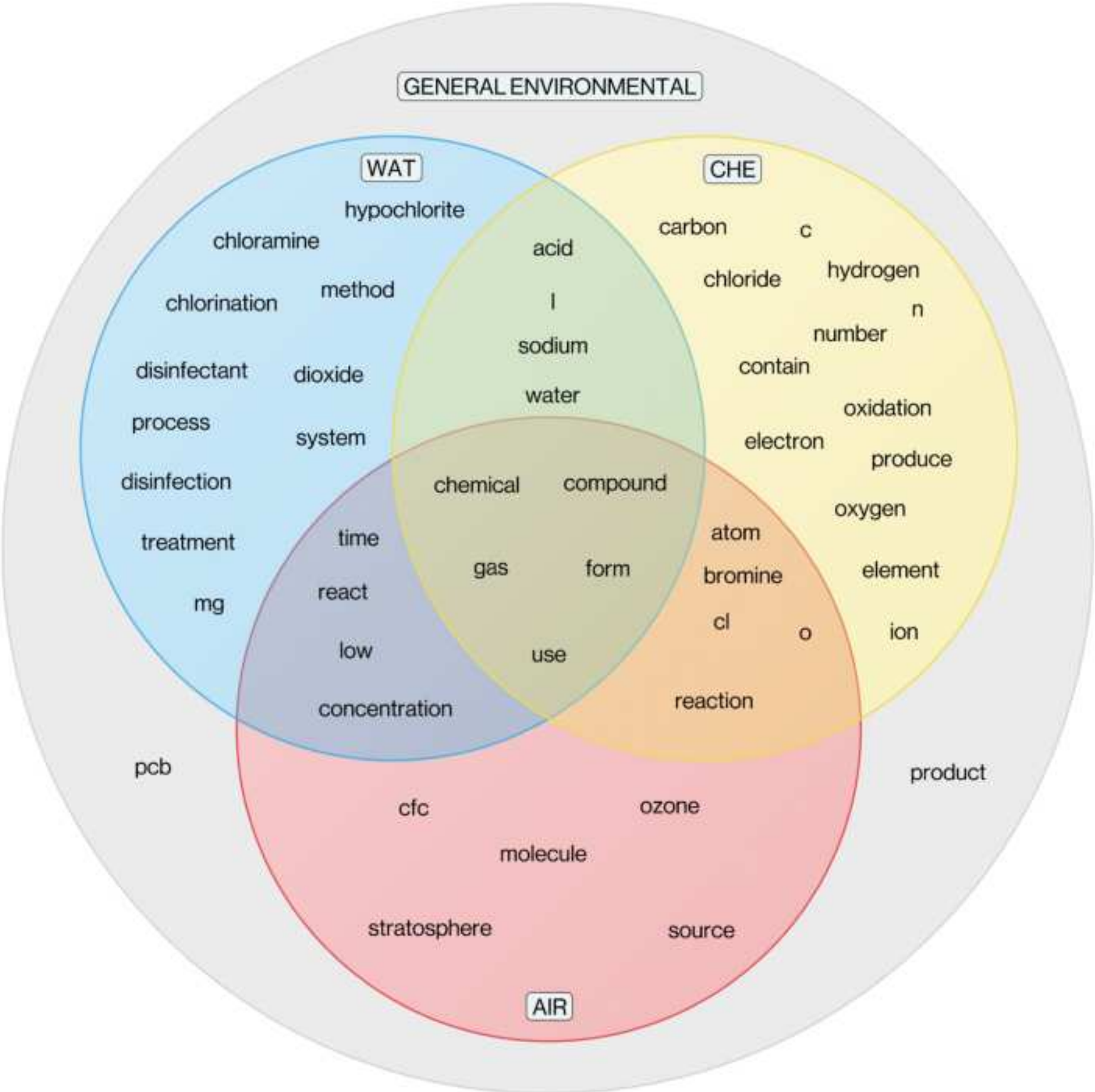

**Figure 34. Venn diagram of the first 50 general environmental contextonyms of CHLORINE that are also present on the list of the first 50 contextonyms in *AIR QUALITY MANAGEMENT*, *CHEMISTRY*, and *WATER TREATMENT AND SUPPLY***

quantify the concentration of chlorine in water.

#### 5.4.2.3 Extraction of superordinate concepts and choice of genera for CHLORINE

As with POLLUTANT, superordinate concept candidates were extracted from the definitions of other resources that we had already analyzed and from corpora. Table 72 shows the result of the extraction of genera from the definitions.

| | GE | AIR | CHE | WAT |
|---|---|---|---|---|
| **ELEMENT** | **10** | **0** | **2** | **1** |
| element | 4 | 0 | 0 | 1 |
| chemical element | 2 | 0 | 0 | 0 |
| halogen element | 2 | 0 | 2 | 0 |
| gaseous element | 1 | 0 | 0 | 0 |
| poisonous element | 1 | 0 | 0 | 0 |
| **GAS** | **3** | **1** | **0** | **3** |
| greenish yellow gas | 1 | 1 | 0 | 0 |
| gas | 0 | 0 | 0 | 2 |
| non-metallic gas | 1 | 0 | 0 | 0 |
| halogen gas | 1 | 0 | 0 | 0 |
| chemical gas | 0 | 0 | 0 | 1 |
| **HALOGEN** | **0** | **1** | **0** | **0** |
| halogen | 0 | 1 | 0 | 0 |
| **CHEMICAL** | **0** | **0** | **0** | **6** |
| chemical | 0 | 0 | 0 | 5 |
| liquid or gas chemical | 0 | 0 | 0 | 1 |
| **DISINFECTANT** | **0** | **0** | **0** | **2** |
| disinfectant | 0 | 0 | 0 | 1 |
| drinking water disinfectant | 0 | 0 | 0 | 1 |

**Table 72. Superordinate concept candidates of CHLORINE extracted from the definitions of other resources**

The general environmental definitions have a preference for the genus ELEMENT, followed by GAS. In *AIR QUALITY MANAGEMENT* only two genera were extracted: GAS and HALOGEN. In *CHEMISTRY*, both analyzed definitions had ELEMENT as superordinate concept. Finally, *WATER TREATMENT AND SUPPLY* favored CHEMICAL, followed by GAS and DISINFECTANT.

These results were complemented with superordinate concepts extracted from corpora. Six were obtained from the MULTI corpus: one from the *AIR QUALITY MANAGEMENT* subcorpus, one from the *CHEMISTRY* subcorpus, and three from the general environmental subcorpus:

| AIR | a tendency to be "poisoned" by **contaminants** such as **chlorine** (Cl), S, and Pb; plugging by PM; and damage by high-temperature |
|---|---|
| CHE | important for the production of **chemical elements**, such as **chlorine** [23] or aluminium. The reverse process in which electrons are |
| ENV | Other forms of chemical disinfectant are the **halogens** such as **chlorine**, bromine, iodine, and the powerful, unstable oxidant, ozone |
| ENV | change.Most of the hundreds of other **air pollutants**—such as **chlorine**, lead, hydrochloric acid, formaldehyde, radioactive substances |
| ENV | other **hazardous chemicals**, such as ammonia, sulfuric acid, and **chlorine**.Chemical safety programs have traditionally stressed |

**Table 73. Hypernymic knowledge-rich contexts for the lemma *chlorine* (noun) in the MULTI corpus**

We also obtained 17 superordinate concept candidates from the PANACEA corpus:

| PAN | plus the **chemical contaminants** used to treat sewage such as **chlorine**; runoff from surface water that can include chemical fertilisers |
|---|---|
| PAN | together with moderate levels of **disinfectants** such as **chlorine**. Treatment Most vegetable washing waters are extensively recycled |
| PAN | are one new approach. Because the use of **chemicals** such as **chlorine** or fluoride may result in the production of harmful by-products |
| PAN | or its compounds such as chloramine or chlorine dioxide. **Chlorine** is a **strong oxidant** that rapidly kills many harmful micro-organisms |
| PAN | inactivation usually involves the use of **disinfectants** such as **chlorine**, ozone, and chlorine dioxide, and a combination of chlorine |
| PAN | underground aquifers and is treated with **chemicals** — such as **chlorine** — to clean it up. According to the U.S. Environmental |
| PAN | untreated domestic sewage, and **chemical contaminants**, such as **chlorine**, from treated sewage; release of waste and contaminants |
| PAN | less-reactive metals, **highly reactive non-metals** (such as **chlorine**, fluorine, and oxygen), and some almost completely nonreactive |
| PAN | drop back to 1970s levels by 2060. But will the decline in **chlorine** and other **ozone-damaging substances** lead directly to the revival |
| PAN | A compound formed by the reaction of a **disinfectant** such as **chlorine** with organic material in the water supply. dissolved oxygen |
| PAN | groundwaters. These compounds MAY react with **halogens** (such as **chlorine**) to form trihalomethanes (try-HAL-o-METH-hanes) |
| PAN | ultraviolet light and other **agents** in the atmosphere such as **chlorine**; uptake in ocean surface waters; and chemical and biological |
| PAN | quartz, and silica, as well as **elements** such as sulfur, **chlorine**, and helium. Petroleum geology Main article: Petroleum geology |
| PAN | present together with moderate levels of **disinfectants** such as **chlorine**. Treatment Most vegetable washing waters are extensively |
| PAN | Sat, 11 Apr 2009 10:12:31 -0700 Though the concentration of **chlorine** and other **ozone-depleting substances** in the stratosphere will |
| PAN | compounds that often contain other **elements** such as hydrogen, **chlorine**, or bromine. Common fluorocarbons include |
| PAN | Freon) They all contain carbon and **halogens**, such as Cl (**chlorine**), F (fluorine), or Br (bromine), and, in the case of the HCFC |

**Table 74. Hypernymic knowledge-rich contexts for the lemma *chlorine* (noun) in the PANACEA corpus**

Table 75 summarizes the results of the extraction of superordinate concepts candidates from corpora:

| | **GE** | **AIR** | **CHE** |
|---|---|---|---|
| **ELEMENT** | **4** | **0** | **1** |
| element | 2 | 0 | 0 |
| chemical element | 2 | 0 | 1 |
| **HALOGEN** | **3** | **0** | **0** |
| halogen | 3 | 0 | 0 |
| **CHEMICAL** | **3** | **0** | **0** |
| chemical | 2 | 0 | 0 |
| hazardous chemical | 1 | 0 | 0 |
| **DISINFECTANT** | **4** | **0** | **0** |
| disinfectant | 4 | 0 | 0 |

| drinking water disinfectant | 0 | 0 | 0 |
|---|---|---|---|
| **CONTAMINANT** | **2** | **1** | **0** |
| contaminant | 0 | 1 | 0 |
| chemical contaminant | 2 | 0 | 0 |
| **POLLUTANT** | **1** | **0** | **0** |
| air pollutant | 1 | 0 | 0 |
| **OXIDANT** | 1 | 0 | 0 |
| strong oxidant | 1 | 0 | 0 |
| **NON-METAL** | 1 | 0 | 0 |
| highly-reactive non-metal | 1 | 0 | 0 |
| **SUBSTANCE** | **2** | **0** | **0** |
| ozone-damaging substance | 1 | 0 | 0 |
| ozone-depleting substance | 1 | 0 | 0 |
| **AGENT** | **1** | **0** | **0** |
| agent | 1 | 0 | 0 |

**Table 75. Superordinate concept candidates of CHLORINE extracted from corpora**

Among the general environmental contexts (the ENV subcorpus and PANACEA), the most common superordinate concepts were ELEMENT and DISINFECTANT, followed by HALOGEN, CHEMICAL, CONTAMINANT, and SUBSTANCE. The only superordinate extracted for *AIR QUALITY MANAGEMENT* was CONTAMINANT, and the only one for *CHEMISTRY* was ELEMENT.

Some of the candidates were a combination of simpler candidates, such as *chemical element*, *halogen gas*, *gaseous element*, *halogen element,* etc. However, for the sake of clarity, Table 76 shows the results merged from the definitions of other resources and from hypernymic knowledge-rich contexts from corpora and limited only to the head in case the of nominal phrases:

| | **GE** | **AIR** | **CHE** | **WAT** | **TOTAL** |
|---|---|---|---|---|---|
| ELEMENT | 14 | 0 | 3 | 1 | 32 |
| CHEMICAL | 3 | 0 | 0 | 6 | 9 |
| GAS | 3 | 1 | 0 | 3 | 7 |
| DISINFECTANT | 4 | 0 | 0 | 2 | 6 |
| HALOGEN | 3 | 1 | 0 | 0 | 4 |
| CONTAMINANT | 2 | 1 | 0 | 0 | 3 |
| SUBSTANCE | 2 | 0 | 0 | 0 | 2 |

| | GE | AIR | CHE | WAT | TOTAL |
|---|---|---|---|---|---|
| POLLUTANT | 1 | 0 | 0 | 0 | 1 |
| OXIDANT | 1 | 0 | 0 | 0 | 1 |
| NON-METAL | 1 | 0 | 0 | 0 | 1 |
| AGENT | 1 | 0 | 0 | 0 | 1 |

**Table 76. Superordinate concept candidates of CHLORINE extracted from the definitions of other resources and from corpora**

For the whole environmental domain, the most frequent superordinate concepts were in this order: ELEMENT, CHEMICAL, GAS, DISINFECTANT, and HALOGEN. For the contextual domains, the results are inconclusive because of the small sample. In *AIR QUALITY MANAGEMENT*, GAS, HALOGEN and CONTAMINANT appear with one occurrence each. In *CHEMISTRY*, ELEMENT has three occurrences. In *WATER TREATMENT AND SUPPLY*, the most frequent superordinate concept is CHEMICAL, followed by GAS and DISINFECTANT.

All the extracted superordinate concepts were structured in a hierarchy (Figure 35). Functional concepts appear in green, and formal concepts appear in orange.

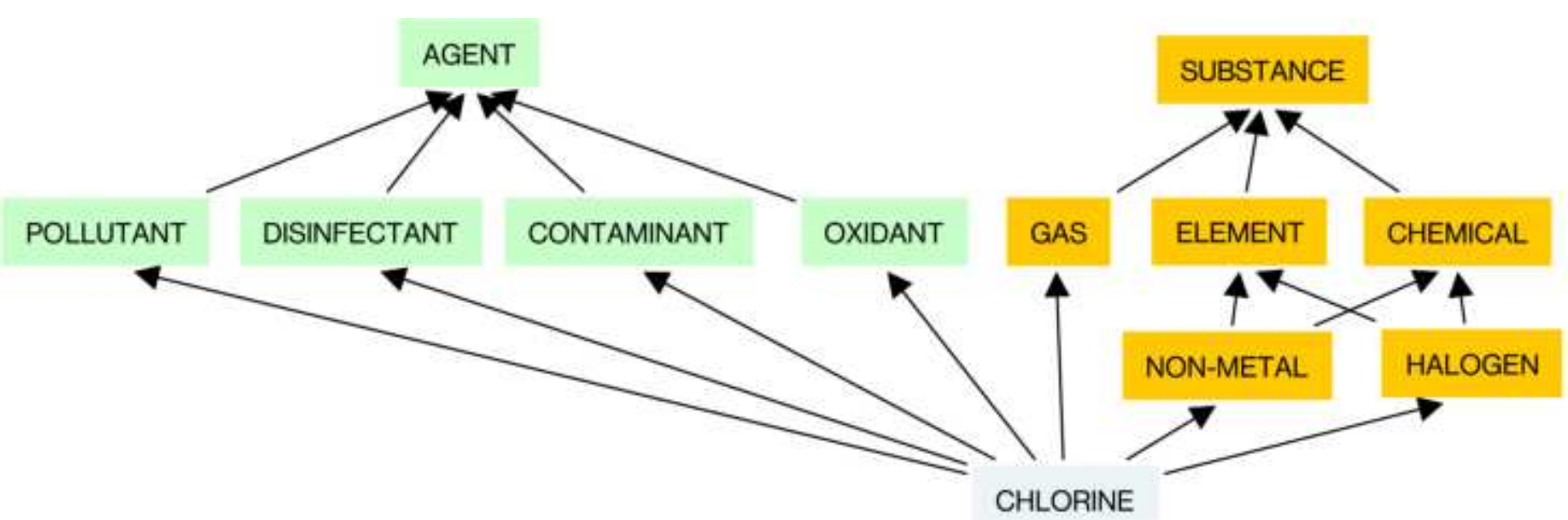

**Figure 35. CHLORINE's superordinate concepts candidates organized in a hierarchy**

In *AIR QUALITY MANAGEMENT* and *WATER TREATMENT AND SUPPLY*, a telic genus is needed since in both domains the concept CHLORINE is profiled in a frame where they serve a function. In *CHEMISTRY* and the general environmental domain, a formal genus is needed.

The general environmental domain takes a superordinate concept that is applicable to all other domains. In this case, CHEMICAL ELEMENT was the most suitable since in all contextual domains CHLORINE is also regarded as a CHEMICAL ELEMENT. The hierarchical path was CHEMICAL

ELEMENT>SUBSTANCE>ENTITY. As explained in §3.6.3, the general environmental superordinate concept is at the same time a non-preferential genus for all other contextual domains. Furthermore, the general environmental domain takes as non-preferential all the other preferential and non-preferential genera in contextual domains.

In the case of *AIR QUALITY MANAGEMENT*, we used AIR POLLUTANT as genus because, although the polluting nature of CHLORINE stems mainly from the fact that it depletes the stratospheric ozone, the category of OZONE-DEPLETING SUBSTANCE is limited to substances containing chlorine or bromine as per the *Montreal Protocol on Substances that Deplete the Ozone Layer* (United Nations Environment Programme 2000) that was approved 1989 but is still in force. CONTAMINANT was discarded because it is less specific than POLLUTANT. Thus, the preferential hierarchical path for CHLORINE in this contextual domain is AIR POLLUTANT>POLLUTANT>AGENT>ENTITY. HALOGEN ELEMENT and GAS are also included as non-preferential superordinate concepts.

WATER DISINFECTANT was the chosen genus for CHLORINE in *WATER TREATMENT AND SUPPLY*. OXIDANT was discarded because it does not reflect as clearly the role of CHLORINE in WATER TREATMENT as WATER DISINFECTANT does. However, the oxidizing nature of CHLORINE, which is what makes it a powerful WATER DISINFECTANT, was included as a characteristic of the concept in our *WATER TREATMENT AND SUPPLY* definition. The preferential hierarchical path for CHLORINE in this contextual domain was thus WATER DISINFECTANT>DISINFECTANT>AGENT.

Finally, we chose HALOGEN ELEMENT as the preferential superordinate genus of chlorine in *CHEMISTRY* since CHEMICAL ELEMENT was too generic for the domain. We chose HALOGEN ELEMENT because it ranked higher than NON-METAL ELEMENT in our superordinate concept extraction. Consequently, the preferential hierarchical path for CHLORINE in *CHEMISTRY* is HALOGEN ELEMENT>CHEMICAL ELEMENT>SUBSTANCE>ENTITY. NON-METAL ELEMENT and GAS are also included as non-preferential superordinate concepts.

All of the resulting hierarchical paths (general environmental and contextual subdomains) for CHLORINE are represented in Figure 36.

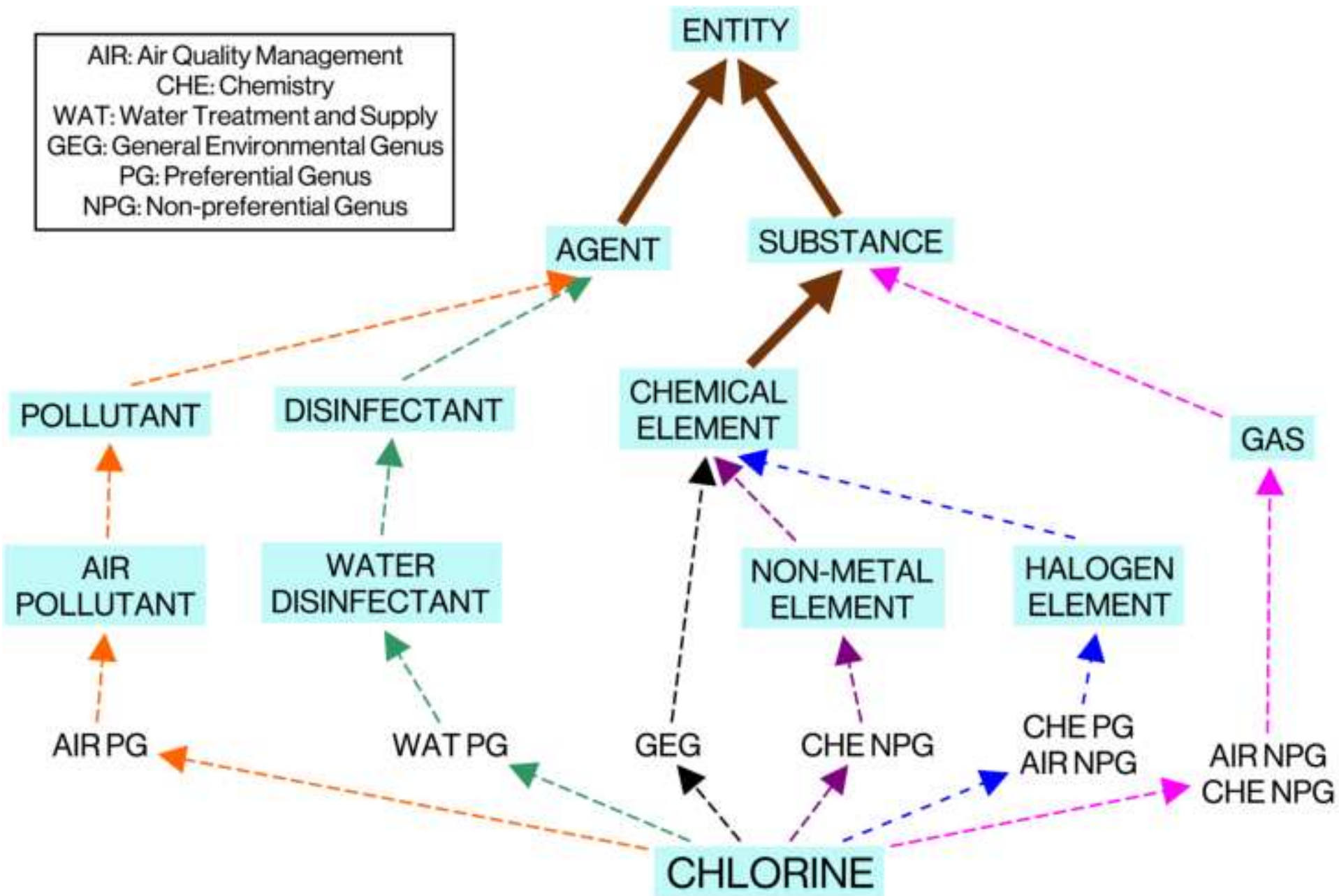

**Figure 36. Hierarchical paths of CHLORINE in all domains**

### 5.4.2.4 Definition of CHLORINE

#### *5.4.2.4.1* AIR QUALITY MANAGEMENT *contextualized definition of CHLORINE*

CHLORINE in *AIR QUALITY MANAGEMENT* has AIR POLLUTANT as preferential genus, CHEMICAL ELEMENT as general environmental genus (which thus acts as non-preferential genus), as well as HALOGEN ELEMENT and GAS as non-preferential genera. All the hierarchical paths are represented in Figure 37.

The top superordinate concept for all of the genera of CHLORINE is ENTITY, which, as has already been said it is considered a semantic primitive, and has no propositions attached. The second level is occupied by AGENT (Table 62) and SUBSTANCE (Table 77), all of which are the same for all domains. Thus, they have only one definition each, and their propositions are not contextualized.

| SUBSTANCE (all domains) | | | | | |
|---|---|---|---|---|---|
| Entity that has a definite composition. | | | | | |
| P1 | DI | @ | SUBSTANCE | *type-of* | ENTITY |
| P2 | DI | @ | SUBSTANCE | *made-of* | ENTITY |

**Table 77. Filled-out definitional template of SUBSTANCE (all domains)**

In the third level of preferential hierarchical path for *AIR QUALITY MANAGEMENT*, the subordinate of AGENT is POLLUTANT. As seen in §5.4.1.2.1, POLLUTANT in *AIR QUALITY MANAGEMENT* is subconceptualized as AIR POLLUTANT. AIR POLLUTANT in *AIR QUALITY MANAGEMENT* is a subordinate concept of POLLUTANT as conceptualized in the general environmental domain. Therefore, CHLORINE inherits from POLLUTANT as conceptualized in the general environmental domain (Table 65) and, then, as AIR POLLUTANT as conceptualized in *AIR QUALITY MANAGEMENT* (to which we will come back later).

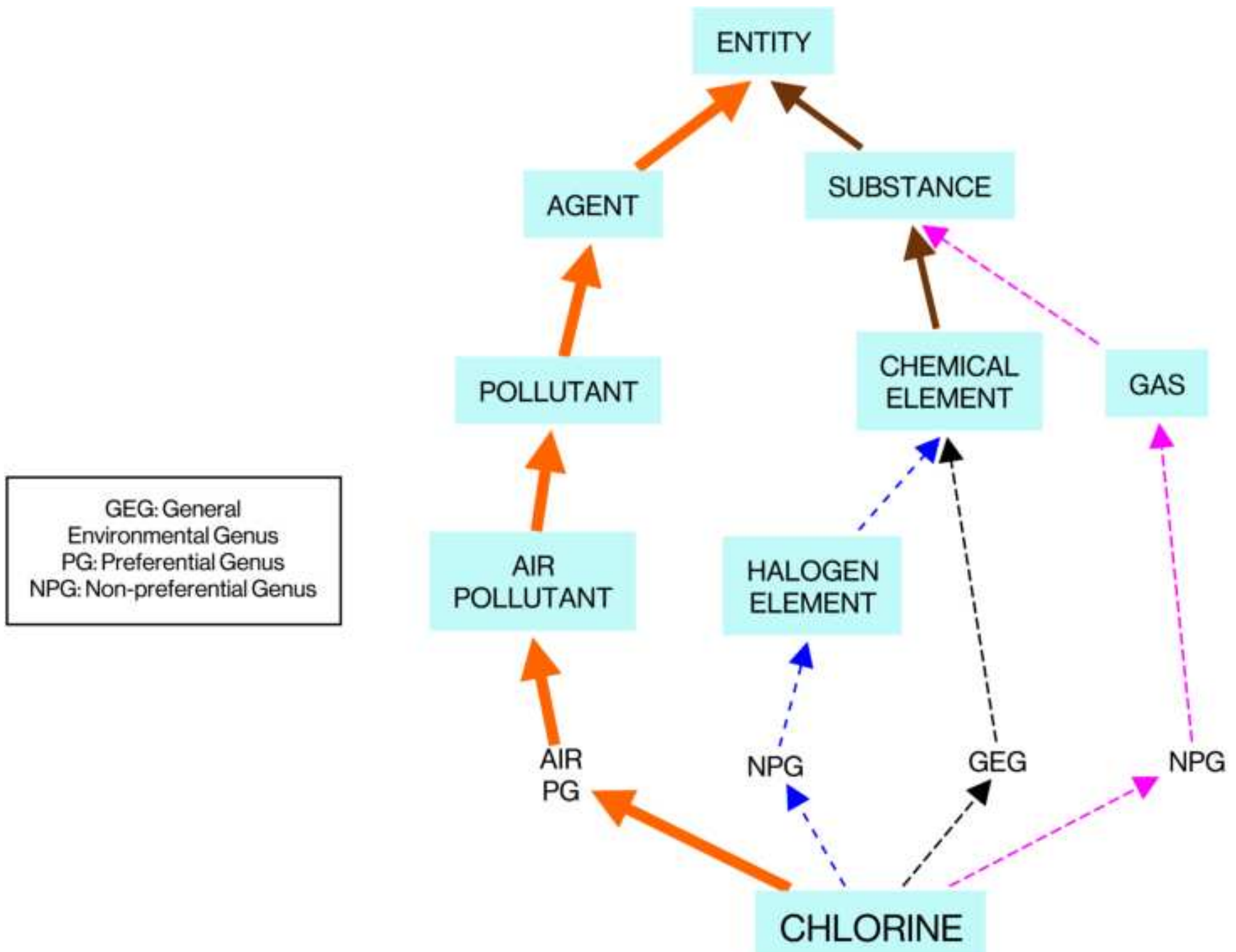

**Figure 37. Hierarchical paths of CHLORINE in *AIR QUALITY MANAGEMENT***

In the third level, there are also CHEMICAL ELEMENT (general environmental path) and GAS (non-preferential path). CHEMICAL ELEMENT

(Table 78) is not contextualized in *AIR QUALITY MANAGEMENT* and, therefore, CHLORINE inherits from its general environmental template. However, GAS is contextualized in this domain and, thus, CHLORINE inherits from a template specific for *AIR QUALITY MANAGEMENT* (Table 79).

| CHEMICAL ELEMENT (all domains) | | | | | |
|---|---|---|---|---|---|
| Substance consisting of the same type of atoms and which cannot be decomposed into chemically simpler substances. Examples of chemical elements are hydrogen, oxygen, and carbon. | | | | | |
| P1 | DI | @ | CHEMICAL ELEMENT | *type-of* | SUBSTANCE |
| P2 | SP | @ | CHEMICAL ELEMENT | *made-of* | ATOM |
| P3 | DI | !@ | CHEMICAL ELEMENT | *affected-by* | CHEMICAL DECOMPOSITION |
| P4 | DI | @ | HYDROGEN & OXYGEN & CARBON | *type-of* | CHEMICAL ELEMENT |

**Table 78. Filled-out definitional template of CHEMICAL ELEMENT (all domains)**

| GAS (*AIR QUALITY MANAGEMENT*) | | | | | |
|---|---|---|---|---|---|
| Substance whose particles are more separate than those of liquids at standard temperature and pressure. Gasses do not have constant volume or shape. The atmosphere is naturally made of different gasses (mainly nitrogen, oxygen, and argon), but also contains polluting gasses (such as carbon dioxide or sulfur dioxide) that are not naturally present or are in a higher concentration due to human activities. | | | | | |
| P1 | DI | @ | GAS | *type-of* | SUBSTANCE |
| P2 | SP | @ | GAS | *made-of* | GAS PARTICLE |
| P3 | DI | !@ | GAS | *has-attribute* | (CONSTANT) VOLUME |
| P4 | DI | !@ | GAS | *has-attribute* | (CONSTANT) SHAPE |
| P5 | DI | @ | ATMOSPHERE | *made-of* | GAS |
| P6 | DI | @ | CARBON DIOXIDE & SULFUR DIOXIDE | *type-of* | GAS |
| P7 | FR | @ | GAS PARTICLE | *has-attribute* | SEPARATENESS |
| P8 | FR | @ | ATMOSPHERIC GAS | *type-of* | GAS |
| P9 | FR | @ | NITROGEN & OXYGEN & ARGON | *type-of* | ATMOSPHERIC GAS |

| P10 | FR | @ | CARBON DIOXIDE & SULFUR DIOXIDE | *type of* | AIR POLLUTANT |
|---|---|---|---|---|---|
| P11 | FR | $\pi$ | AIR POLLUTANT | *has-location* | ATMOSPHERE |
| P12 | FR | $\pi$ | AIR POLLUTANT | *has-attribute* | (HIGHER THAN NATURAL) CONCENTRATION |
| P13 | FR | $\pi$ | AIR POLLUTANT | *result-of* | ARTIFICIAL PROCESS |

**Table 79. Filled-out definitional template of GAS (*AIR QUALITY MANAGEMENT* domain)**

Finally, on the fourth level, there is one non-preferential superordinate (HALOGEN ELEMENT) (Table 80) that is a subordinate of CHEMICAL ELEMENT, and the preferential superordinate AIR POLLUTANT (Table 81). Both of them have contextualized templates in this contextual domain.

| HALOGEN ELEMENT (*AIR QUALITY MANAGEMENT*) | | | | | |
|---|---|---|---|---|---|
| Chemical element that belongs to group 17 of the periodic table, which includes fluorine, chlorine, bromine, iodine, and astatine. Halogens tend to be very reactive and toxic. Some organic compounds, such as chlorofluorocarbons and methyl bromide, contain halogen elements and, when emitted to the atmosphere, contribute to ozone depletion. | | | | | |
| P1 | DI | @ | HALOGEN ELEMENT | *type-of* | CHEMICAL ELEMENT |
| P2 | DI | $\pi$ | HALOGEN ELEMENT | *has-attribute* | (HIGH) REACTIVITY |
| P3 | DI | $\pi$ | HALOGEN ELEMENT | *has-attribute* | (HIGH) TOXICITY |
| P4 | DI | @ | FLUORINE & CHLORINE & BROMINE & IODINE & ASTATINE | *type-of* | HALOGEN ELEMENT |
| P5 | FR | @ | CHLOROFLUOROCARBON & METHYL BROMIDE | *made-of* | HALOGEN ELEMENT |
| P6 | FR | @ | CHLOROFLUOROCARBON & METHYL BROMIDE | *type-of* | ORGANIC COMPOUND & OZONE-DEPLETING SUBSTANCE |

**Table 80. Filled-out definitional template of HALOGEN ELEMENT (*AIR QUALITY MANAGEMENT* domain)**

| AIR POLLUTANT (*AIR QUALITY MANAGEMENT*) | | | | | |
|---|---|---|---|---|---|
| Pollutant emitted to the atmosphere that degrades air quality and which can affect human health in case of exposure. Some of the main sources of air pollutants are industrial activity or vehicle exhaust. The most important air pollutants are tropospheric ozone, particulate matter, carbon monoxide, nitrogen oxides, sulfur dioxide, lead, and greenhouse gasses (such as carbon dioxide or methane). | | | | | |
| P1 | DI | @ | AIR POLLUTANT | *type-of* | POLLUTANT |
| P2 | SP | @ | AIR POLLUTANT | *affects (adversely)* | AIR \| HUMAN HEALTH |
| P3 | SP | $\pi$ | AIR POLLUTANT | *result-of* | INDUSTRIAL ACTIVITY & VEHICLE EXHAUST |
| P4 | DI | @ | TROPOSPHERIC OZONE & PARTICULATE MATTER & CARBON MONOXIDE & NITROGEN OXIDE & SULFUR DIOXIDE & LEAD & GREENHOUSE GAS | *type-of* | AIR POLLUTANT |
| P5 | FR | @ | CARBON DIOXIDE & METHANE | *type-of* | GREENHOUSE GAS |

**Table 81. Filled-out definitional template of AIR POLLUTANT (*AIR QUALITY MANAGEMENT* domain)**

Finally, we reproduce below the contextualized definition for CHLORINE in *AIR QUALITY MANAGEMENT*, which, as previously explained, has AIR POLLUTANT as its genus, but also GAS, HALOGEN ELEMENT, and CHEMICAL ELEMENT as its non-preferential superordinate concepts.

| CHLORINE (*AIR QUALITY MANAGEMENT*) |
|---|
| Air pollutant emitted mainly as chlorofluorocarbons, hydrochlorofluorocarbons, carbon tetrachloride, and methyl chloroform, which are ozone-depleting substances. When these substances reach the stratosphere, ultraviolet radiation and chemical reactions break them apart, and they are converted into hydrochloric acid and chlorine nitrate. Those two compounds are finally converted into chlorine monoxide or free chlorine atoms, which destroy |

| stratospheric ozone because of their high reactivity. Along with bromine, chlorine is the main contributor to ozone depletion. | | | | | |
|---|---|---|---|---|---|
| P1 | DI | @ | CHLORINE | *type-of* | AIR POLLUTANT |
| P2 | DI | @ | CHLOROFLUOROCARBON & HYDROCHLOROFLUOROCARBON & CARBON TETRACHLORIDE & METHYL CHLOROFORM | *made-of* | CHLORINE |
| P3 | DI | @ | HYDROCHLORIC ACID & CHLORINE NITRATE | *made-of* | CHLORINE |
| P4 | DI | @ | CHLORINE MONOXIDE | *made-of* | CHLORINE |
| P5 | EX | @ | CHLORINE | *has-attribute* | (HIGH) REACTIVITY |
| P6 | SP | @ | CHLORINE | *affects (destroys)* | STRATOSPHERIC OZONE |
| P7 | FR | @ | CHLOROFLUOROCARBON & HYDROCHLOROFLUOROCARBON & CARBON TETRACHLORIDE & METHYL CHLOROFORM | *type-of* | OZONE-DEPLETING SUBSTANCE |
| P8 | FR | $\pi$ | CHLOROFLUOROCARBON & HYDROCHLOROFLUOROCARBON & CARBON TETRACHLORIDE & METHYL CHLOROFORM | *affected-by (decomposed by)* | ULTRAVIOLET RADIATION & CHEMICAL REACTION |
| P9 | FR | @ | HYDROCHLORIC ACID & CHLORINE NITRATE | *type-of* | CHEMICAL COMPOUND |
| P10 | FR | @ | CHLORINE MONOXIDE | *affects (destroys)* | STRATOSPHERIC OZONE |
| P11 | FR | @ | BROMINE | *affects (destroys)* | STRATOSPHERIC OZONE |

**Table 82. Filled-out definitional template of** CHLORINE (*AIR QUALITY MANAGEMENT* **domain)**

It was not possible to accurately encode various of the definitional characteristics of CHLORINE in the template. The main reason for this is that the underlying frame of this definition (OZONE DEPLETION) comprises several phases in which many participants interact. As a consequence, there were certain relations that could only be expressed in the definition.

For instance, the part of the process consisting of (chlorine-containing) OZONE-DEPLETING SUBSTANCES reaching the STRATOSPHERE and then being converted into HYDROCHLORIC ACID and CHLORINE NITRATE by ULTRAVIOLET RADIATION and CHEMICAL REACTIONS could not be completely expressed. The only possible proposition related to it is P8, which is marked as prototypical instead of absolute because it does not express a characteristic always associated with those concepts. The decomposition of these substances by ultraviolet radiation and chemical reactions into hydrochloric acid and chlorine nitrate only happens naturally under specific circumstances (i.e., after they have reached the stratosphere when they have been transported by air motions and when certain conditions are met).

P5 is an explicit indirect proposition inherited from the non-preferential superordinate concept HALOGEN ELEMENT. If it had been the preferential genus, this proposition would have been redundant. However, since a definition can only have one genus, this proposition needs to be represented explicitly in the definition. As for GAS, the other non-preferential superordinate concept, no propositions inherited from it were made explicit because the gaseous state of chlorine is evident from the context.

P6 is the only specified indirect proposition. It is inherited from "AIR POLLUTANT *affects (adversely)* ENVIRONMENT" and ENVIRONMENT has been specified as STRATOSPHERIC OZONE, which is in a meronymic relation with it.

As can be seen, more than a third of the propositions are framing propositions. If the definition were limited to direct and indirect propositions, the user would only obtain a shallow understanding of the concept. However, by explaining the form in which CHLORINE is emitted to the ATMOSPHERE, the changes that it undergoes and how it ends up depleting STRATOSPHERIC OZONE, the user is able to integrate the concept in a larger knowledge structure and understand the relevance of CHLORINE in this domain.

#### 5.4.2.4.2 CHEMISTRY *contextualized definition of* CHLORINE

HALOGEN ELEMENT Table 83, which is a subordinate of the general environmental genus (CHEMICAL ELEMENT) (Table 78), is the preferential genus of CHLORINE in *CHEMISTRY*. At the same time, NON-METAL ELEMENT Table 84 and GAS (Table 85) are its non-preferential genera. All of them have specific definitional templates for *CHEMISTRY*. The hierarchical paths are represented in Figure 38.

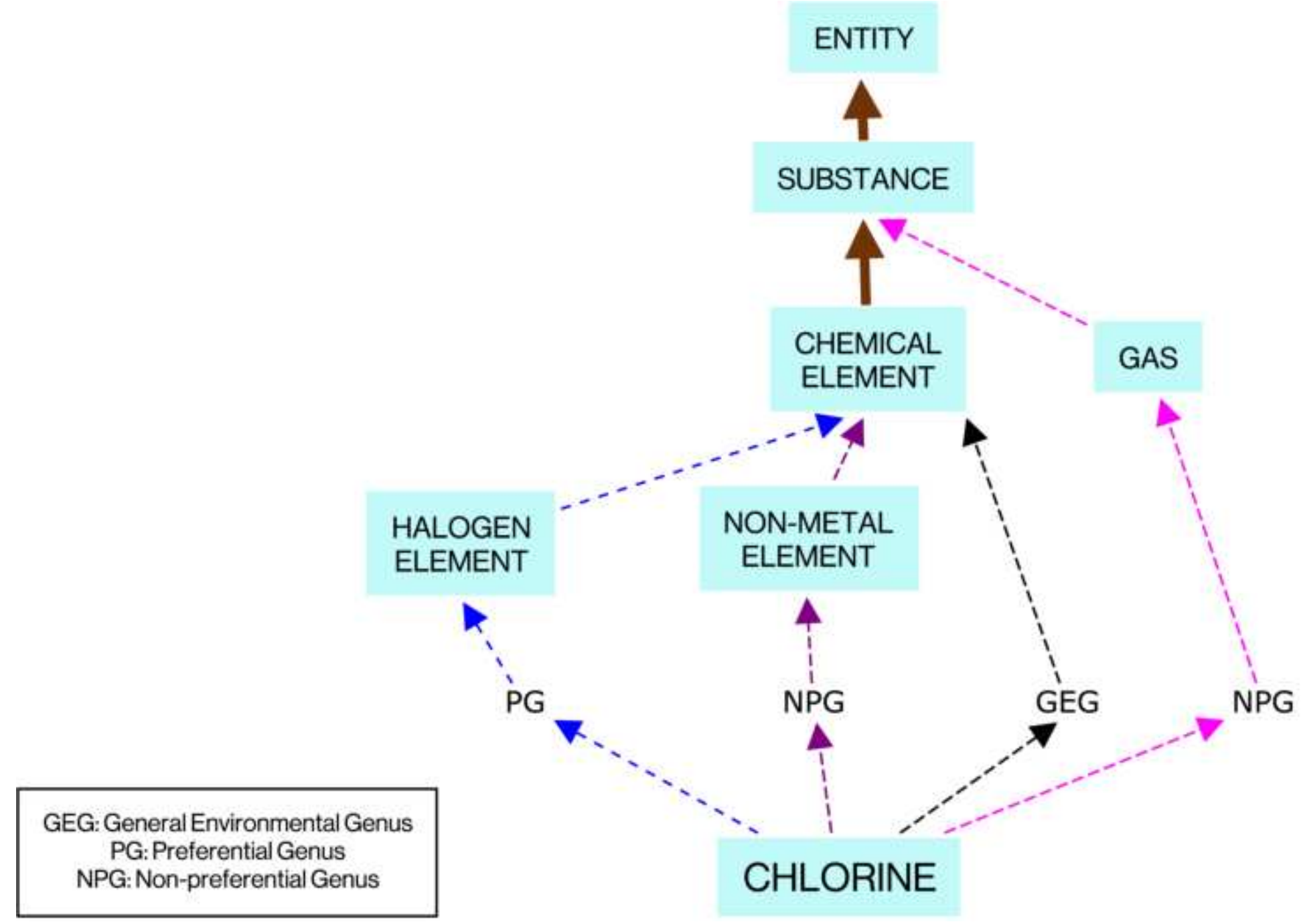

**Figure 38. Hierarchical paths of CHLORINE in *CHEMISTRY***

| HALOGEN ELEMENT (*CHEMISTRY*) | | | | | |
|---|---|---|---|---|---|
| Chemical element that belongs to group 17 of the periodic table, which includes fluorine, chlorine, bromine, iodine, and astatine. It is probable that unenseptium is also a halogen element. Halogens are diatomic in their elemental state, though because of their reactivity, they are never found as such in nature. Since they are electronegative, they can act as oxidizing agents. When they react with metals, they form salts. Halogen elements tend to be very toxic. | | | | | |
| P1 | DI | @ | HALOGEN ELEMENT | *type-of* | CHEMICAL ELEMENT |
| P2 | DI | @ | FLUORINE &<br>CHLORINE &<br>BROMINE &<br>IODINE & | *type-of* | HALOGEN ELEMENT |

| | | | ASTATINE | | |
|---|---|---|---|---|---|
| P3 | DI | $\pi$ | UNUNSEPTIUM | *type-of* | HALOGEN ELEMENT |
| P4 | DI | $\pi$ | HALOGEN | *has-attribute* | DIATOMIC |
| P5 | DI | @ | HALOGEN ELEMENT | *has-attribute* | REACTIVITY & ELECTRONEGATIVITY & OXIDIZING |
| P6 | DI | $\pi$ | HALOGEN ELEMENT | *affects (reacts with)* | METAL |
| P7 | DI | $\pi$ | SALT | *made-of* | HALOGEN ELEMENT |
| P8 | DI | $\pi$ | HALOGEN ELEMENT | *has-attribute* | TOXIC |
| P9 | DI | $\pi$ | HALOGEN ELEMENT | *has-attribute* | TOXICITY |

**Table 83. Filled-out definitional template of HALOGEN ELEMENT (*CHEMISTRY* domain)**

| NON-METAL ELEMENT (*CHEMISTRY*) | | | | | |
|---|---|---|---|---|---|
| Non-lustrous, brittle chemical element that are normally poor electrical and thermal conductors. Since most non-metal elements have high ionization energy and high electron affinity, they tend to be electronegative. Except for hydrogen, nonmetals are located in the upper right side of the periodic table, such as helium, fluorine or neon. | | | | | |
| P1 | DI | @ | NON-METAL ELEMENT | *type-of* | CHEMICAL ELEMENT |
| P2 | DI | @ | NON-METAL ELEMENT | *has-attribute* | LUSTER & BRITTLENESS |
| P3 | DI | !$\pi$ | NON-METAL ELEMENT | *has-attribute* | ELECTRICAL CONDUCTIVITY & THERMAL CONDUCTIVITY |
| P4 | DI | $\pi$ | NON-METAL ELEMENT | *has-attribute* | HIGH IONIZATION ENERGY & HIGH ELECTRON AFFINITY & ELECTRONEGATIVITY |
| P5 | DI | @ | HYDROGEN & HELIUM & FLUORINE & NEON | *type-of* | NON-METAL ELEMENT |

**Table 84. Filled-out definitional template of NON-METAL ELEMENT (*CHEMISTRY* domain)**

| GAS (*CHEMISTRY*) | | | | | |
|---|---|---|---|---|---|
| Substance whose particles are more separate than those of liquids or solids at standard temperature and pressure and do not experience any attractive or repulsive force with each other. Gasses do not have constant volume or shape. Gas particles are in continuous movement colliding with each other. | | | | | |
| P1 | DI | @ | GAS | *type-of* | SUBSTANCE |
| P2 | SP | @ | GAS | *made-of* | GAS PARTICLE |
| P3 | DI | !@ | GAS | *has-attribute* | (CONSTANT) VOLUME |
| P4 | DI | !@ | GAS | *has-attribute* | (CONSTANT) SHAPE |
| P5 | FR | @ | GAS PARTICLE | *has-attribute* | SEPARATENESS & (CONTINUOUS) MOVEMENT |
| P6 | FR | !@ | GAS PARTICLE | *affected-by* | ATTRACTIVE FORCE & REPULSIVE FORCE |
| P7 | FR | @ | GAS PARTICLE | *affected-by* | COLLISION |

**Table 85. Filled-out definitional template of GAS (*CHEMISTRY* domain)**

| CHLORINE (*CHEMISTRY*) | | | | | |
|---|---|---|---|---|---|
| Non-metallic halogen element with atomic number 17 that exists as a greenish-yellow gas at standard temperature and pressure. It is only found naturally in compounds, such as sodium chloride (common salt) in seawater and in halite (rock salt) or potassium chloride in sylvite. Chlorine has an atomic weight of 35.45 u with $^{35}$Cl and $^{37}$Cl as its stable isotopes. | | | | | |
| P1 | DI | @ | CHLORINE | *type-of* | HALOGEN ELEMENT |
| P2 | EX | @ | CHLORINE | *has-attribute* | NON-METALLIC |
| P3 | DI | $\pi$ | CHLORINE | *has-attribute* | GREENISH YELLOW |
| P4 | EX | $\pi$ | CHLORINE | *has-attribute* | GASEOUS |
| P5 | DI | @ | SODIUM CHLORIDE & POTASSIUM CHLORIDE | *made-of* | CHLORINE |
| P6 | DI | @ | CHLORINE-35 & CHLORINE-37 | *type-of* | CHLORINE |
| P7 | FR | @ | SODIUM CHLORIDE & POTASSIUM | *type-of* | CHEMICAL COMPOUND |
| | | | CHLORIDE | | |
| P8 | FR | @ | SEAWATER & HALITE | *made-of* | SODIUM CHLORIDE |
| P9 | FR | @ | SYLVITE | *made-of* | POTASSIUM CHLORIDE |

**Table 86. Filled-out definitional template of CHLORINE (*CHEMISTRY* domain)**

Many of the characteristics that were relevant according to the analysis of the definitions of other resources and contextonyms were not represented in the definition because they have already been described as part of its genus HALOGEN ELEMENT. Examples of this are as its reactivity, electronegativity or its oxidizing power.

CHLORINE also inherits from its non-preferential superordinate concepts NON-METAL ELEMENT and GAS. In this case, instead of inheriting selected propositions from them, we opted for adding two attributes that indicate that chlorine is also a GAS and a NON-METAL ELEMENT (P2, and P4).

P2 and P3 are prototypical because CHLORINE is not always gaseous and greenish-yellow in color. However, P4 is absolute because CHLORINE is always non-metallic.

The atomic number and atomic weight cannot currently be represented with EcoLexicon propositions. Moreover, since EcoLexicon is a not specialized in *CHEMISTRY*, we propose to encode the isotopes of an element as subordinate concepts, although a specific relation would be necessary for greater precision.

The framing propositions in this definition add further information about the two most common CHEMICAL COMPOUNDS where CHLORINE can be found in nature.

#### *5.4.2.4.3* WATER TREATMENT AND SUPPLY *definition of* CHLORINE

The *WATER TREATMENT AND SUPPLY* definition of CHLORINE has WATER DISINFECTANT as preferential genus, and CHEMICAL ELEMENT as non-

preferential since it is the general environmental superordinate concept. Both hierarchical paths are depicted in Figure 39.

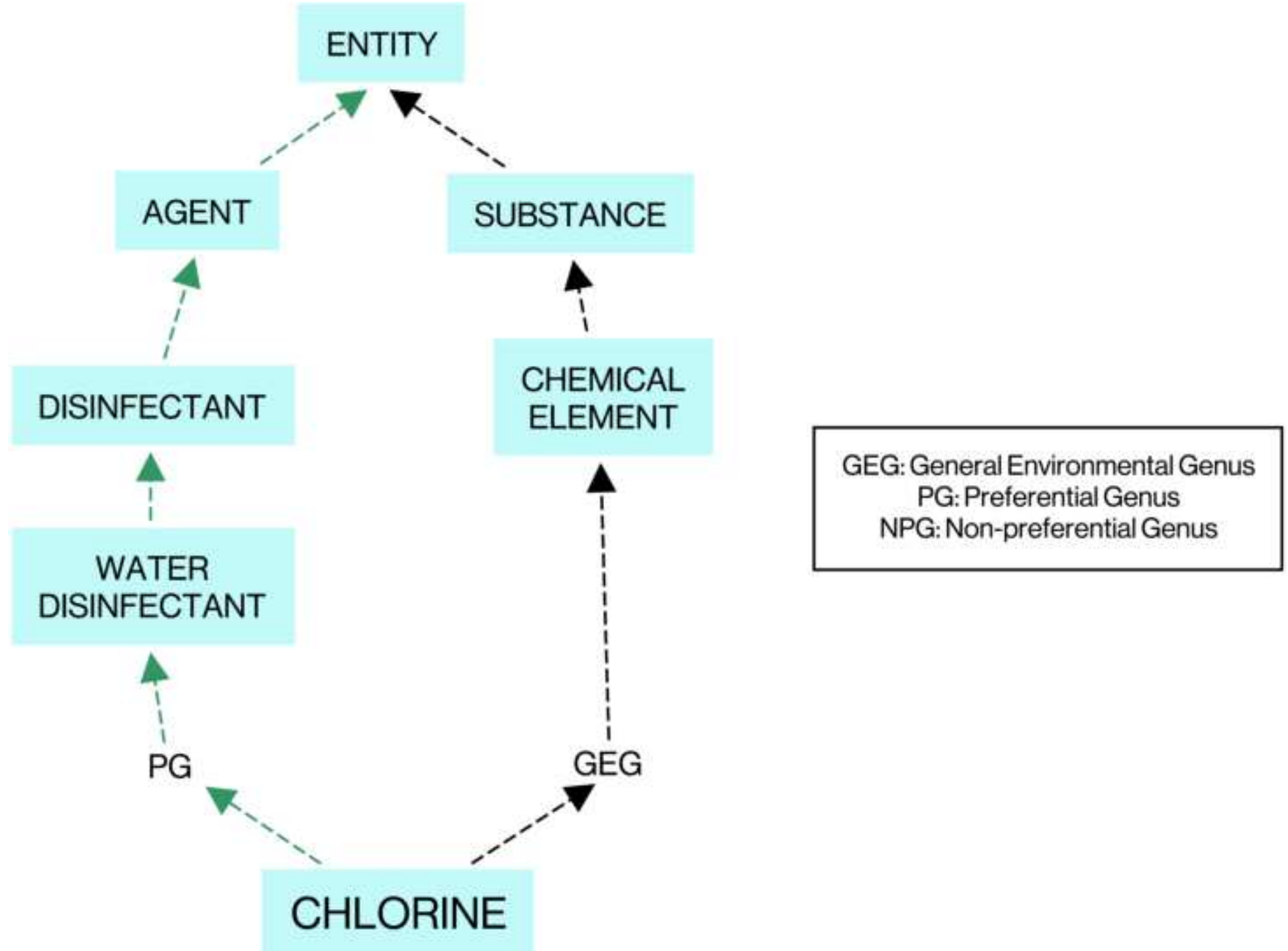

**Figure 39. Hierarchical paths of** CHLORINE **in** ***WATER TREATMENT AND SUPPLY***

Since ENTITY, SUBSTANCE (Table 77) and CHEMICAL ELEMENT (Table 78) are not contextualized in *WATER TREATMENT AND SUPPLY*, their general environmental definitional templates apply to this domain too. For its part, DISINFECTANT is subconceptualized as WATER DISINFECTANT in *WATER TREATMENT AND SUPPLY*. Consequently, CHLORINE in this domain inherits characteristics from DISINFECTANT in its general environmental template (Table 87) and, then, from WATER DISINFECTANT in its *WATER TREATMENT AND SUPPLY* conceptualization (Table 88). The resulting contextualized definition is shown in Table 89.

| DISINFECTANT (all domains) | | | | | |
|---|---|---|---|---|---|
| Agent that kills or inactivates pathogens in a non-living entity such as water, air or a surface. Examples of disinfectants include substances (e.g., sodium hypochlorite, ozone, phenols) and ultraviolet radiation. | | | | | |
| P1 | DI | @ | DISINFECTANT | *type-of* | AGENT |
| P2 | SP | @ | DISINFECTANT | *affects (kills or* | PATHOGEN |

| | | | | *inactivates)* | |
|---|---|---|---|---|---|
| P3 | SP | @ | DISINFECTANT | *affects* | NON-LIVING ENTITY \| WATER \| AIR \| SURFACE |
| P4 | DI | @ | SODIUM HYPOCHLORITE & OZONE & PHENOL & ULTRAVIOLET RADIATION | *type-of* | DISINFECTANT |
| P5 | DI | @ | SODIUM HYPOCHLORITE & OZONE & PHENOL | *type-of* | SUBSTANCE |
| P6 | DI | | WATER & AIR & SURFACE | *type-of* | NON-LIVING ENTITY |

**Table 87. Filled-out definitional template of** DISINFECTANT **(all domains)**

| WATER DISINFECTANT (*WATER TREATMENT AND SUPPLY*) | | | | | |
|---|---|---|---|---|---|
| Disinfectant that is added to water to kill or inactivate pathogens. The most common water disinfectants are chlorine-containing substances such as sodium hypochlorite (bleach) and chloramine. Ozone and ultraviolet radiation are also used as water disinfectants. | | | | | |
| P1 | DI | @ | WATER DISINFECTANT | *type-of* | DISINFECTANT |
| P2 | SP | @ | WATER DISINFECTANT | *affects (is added to)* | WATER |
| P3 | DI | @ | SODIUM HYPOCHLORITE & CHLORAMINE & OZONE & ULTRAVIOLET RADIATION | *type-of* | WATER DISINFECTANT |
| P6 | DI | @ | SODIUM HYPOCHLORITE & CHLORAMINE | *type-of* | SUBSTANCE |
| P7 | DI | @ | SODIUM HYPOCHLORITE & CHLORAMINE | *made-of* | CHLORINE |

**Table 88. Filled-out definitional template of** WATER DISINFECTANT (***WATER TREATMENT AND SUPPLY*** **domain)**

| CHLORINE (*WATER TREATMENT AND SUPPLY*) | | | | | |
|---|---|---|---|---|---|
| Water disinfectant that because of its oxidizing power, is added to water usually in a contact tank to kill or inactivate pathogens as part of wastewater treatment. Chlorine is used in pure form or as sodium hypochlorite (bleach), calcium hypochlorite, chlorine dioxide, or chloramine. Chlorine reacts with organic compounds that may be present in water to form harmful by-products: trihalomethanes, and haloacetic acids. Since chlorine is toxic to aquatic life, residual chlorine in treated wastewater needs to be removed (normally with sulfur dioxide) before being discharged into aquatic ecosystems. | | | | | |
| P1 | DI | @ | CHLORINE | *type-of* | WATER DISINFECTANT |
| P2 | DI | @ | CHLORINE | *has-attribute* | OXIDIZING |
| P3 | DI | @ | SODIUM HYPOCHLORITE & CALCIUM HYPOCHLORITE & CHLORINE DIOXIDE & CHLORAMINE | *made-of* | CHLORINE |
| P4 | DI | π | CHLORINE | *affects (reacts with)* | ORGANIC COMPOUND |
| P5 | DI | @ | CHLORINE | *has-attribute* | TOXICITY (TO AQUATIC LIFE) |
| P6 | DI | π | CHLORINE | *affected-by (removed by)* | SULFUR DIOXIDE |
| P7 | FR | @ | CHLORINATION | *takes-place-in* | CONTACT TANK |
| P8 | FR | π | CHLORINATION | *phase-of* | WASTEWATER TREATMENT |
| P9 | FR | π | TRIHALOMETHANE & HALOACETIC ACID | *result-of* | CHLORINATION |
| P10 | FR | @ | TRIHALOMETHANE & HALOACETIC ACID | *has-attribute* | HARMFUL |

**Table 89. Filled-out definitional template of** CHLORINE (***WATER TREATMENT AND SUPPLY*** **domain)**

The contextualized definition of CHLORINE in *WATER TREATMENT AND SUPPLY* inserts the concept in the frame of CHLORINATION, from which certain propositions are also inferred. However, as with POLLUTANT, the relation «CHLORINE effects CHLORINATION» is not represented in the definition because they are morphologically related, though the user is presented with this proposition in the conceptual map in EcoLexicon. The same is true for DECHLORINATION.

P2 is a direct proposition because HALOGEN ELEMENT (from which it could have been inherited) is not a non-preferential genus of CHLORINE in this contextual domain. Apart from the fact that in the analysis in §5.4.2.3 there was no occurrence of HALOGEN ELEMENT as a superordinate concept of CHLORINE in *WATER TREATMENT AND SUPPLY*, the term *halogen* only appears once in the *WATER TREATMENT AND SUPPLY* subcorpus. Therefore, there was no justification for making HALOGEN ELEMENT a non-preferential genus of CHLORINE in this contextual domain. As a consequence, all the propositions are either direct or framing.

P4 and P6 have prototypical status because they only happen under certain circumstances. P4 refers to the reaction of chlorine with certain organic compounds when it is added to water, while P6 is prototypical because DECHLORINATION is not always included as part of wastewater treatment, and sometimes the use of SULFUR DIOXIDE is replaced by other methods.

The framing propositions in P7, P8, P9, and P10 provide more details about CHLORINATION: where it is carried out, the larger event in which it is integrated, and its drawbacks.

##### *5.4.2.4.4 General environmental definition of* CHLORINE

Given that the general environmental definition (Table 90) encompasses all the contextual domains, many of its propositions are shared with those of contextualized definitions. The propositions that do not appear in the contextualized definitions are based on what the analysis of the definitions from other resources and the general environmental contextonyms revealed.

| CHLORINE (general environmental) |
|---|
| Non-metallic chemical element that belongs to the halogen family and exists as a greenish-yellow gas at standard temperature and pressure. Because of its high reactivity, it is only found naturally in compounds such as sodium chloride (common salt) in seawater, and halite (rock |

<table>
<tr><td colspan="6">salt). Since it has a strong oxidizing power, it is used as a water disinfectant and bleaching agent, particularly as sodium hypochlorite (bleach). Chlorine is a component of polychlorinated biphenyls (PCBs) and dichlorodiphenyltrichloroethane (DDT), two harmful artificial compounds, and polyvinyl chloride (PVC), a widely used plastic. As a consequence of the emission of chlorofluorocarbons (CFCs), hydrochlorofluorocarbons (HCFCs), and other man-made compounds, chlorine is released into the stratosphere and destroys ozone.</td></tr>
<tr><td>P1</td><td>DI</td><td>@</td><td>CHLORINE</td><td><i>type-of</i></td><td>CHEMICAL ELEMENT</td></tr>
<tr><td>P2</td><td>DI</td><td>@</td><td>CHLORINE</td><td><i>has-attribute</i></td><td>GASEOUS & NON-METALLIC & HALOGEN & GREENISH YELLOW & (HIGH) REACTIVITY & OXIDIZING</td></tr>
<tr><td>P3</td><td>DI</td><td>@</td><td>SODIUM CHLORIDE & SODIUM HYPOCHLORITE & POLYCHLORINATED BIPHENYL & DICHLORODIPHENYLTRICHLOROETHANE & POLYVINYL CHLORIDE & CHLOROFLUOROCARBON & HYDROCHLOROFLUOROCARBON</td><td><i>made-of</i></td><td>CHLORINE</td></tr>
<tr><td>P4</td><td>DI</td><td>@</td><td>CHLORINE & SODIUM HYPOCHLORITE</td><td><i>effects</i></td><td>WATER DISINFECTION & BLEACHING</td></tr>
<tr><td>P5</td><td>DI</td><td>@</td><td>CHLORINE</td><td><i>affects (destroys)</i></td><td>STRATOSPHERIC OZONE</td></tr>
<tr><td>P6</td><td>FR</td><td>@</td><td>SEAWATER & HALITE</td><td><i>made-of</i></td><td>SODIUM CHLORIDE</td></tr>
<tr><td>P7</td><td>FR</td><td>@</td><td>CHLOROFLUOROCARBON & HYDROCHLOROFLUOROCARBON & POLYCHLORINATED</td><td><i>type-of</i></td><td>ARTIFICIAL CHEMICAL COMPOUND</td></tr>
</table>

| | | | BIPHENYL & DICHLORODIPHENYLTRICHLOROETHANE & POLYVINYL CHLORIDE | | |
|---|---|---|---|---|---|
| P8 | FR | @ | POLYCHLORINATED BIPHENYL & DICHLORODIPHENYLTRICHLOROETHANE | *has-attribute* | HARMFUL |
| P9 | FR | @ | POLYVINYL CHLORIDE | *type-of* | PLASTIC |

**Table 90. Filled-out definitional template of CHLORINE (general environmental domain)**

P2 includes six attributes of CHLORINE that are also represented in its contextual domains. GASEOUS, NON-METALLIC, HALOGEN, and GREENISH YELLOW in *CHEMISTRY*; (HIGH) REACTIVITY in *AIR QUALITY MANAGEMENT*, and OXIDIZING in *WATER TREATMENT AND SUPPLY*.

P3 describes all the relevant chemical compounds that contain CHLORINE. On the one hand, most of this information is already present in the contextualized definitions. SODIUM CHLORIDE, as the main natural source of CHLORINE, comes from *CHEMISTRY* (this proposition is complemented with P6, like in *CHEMISTRY*). SODIUM HYPOCHLORITE is featured in *WATER TREATMENT AND SUPPLY* definition, although in this definition, its function as BLEACHING AGENT, in addition to WATER DISINFECTANT, is also conveyed in P4. The fact that CHLOROFLUOROCARBON and HYDROCHLOROFLUOROCARBON are made of CHLORINE originates from the *AIR QUALITY MANAGEMENT* contextual domain and P5 and P7 further specified that they are artificial compounds that end up releasing CHLORINE into the ATMOSPHERE and resulting in OZONE DEPLETION.

On the other hand, in contrast with the contextual domain definitions, P3 states that also POLYCHLORINATEDBIPHENYL, DICHLORODIPHENYLTRICHLOROETHANE, and POLYVINYL CHLORIDE contain CHLORINE. This proposition is complemented with the framing propositions P7, P8 and P9 that add that POLYCHLORINATED BIPHENYL and DICHLORODIPHENYLTRICHLOROETHANE are HARMFUL COMPOUNDS, and that POLYVINYL CHLORIDE is a PLASTIC.

#### 5.4.2.4.5 Flexible definition of CHLORINE

The resulting flexible definition of CHLORINE is reproduced in the following table:

| CHLORINE | |
|---|---|
| *GENERAL ENVIRONMENTAL* | Non-metallic chemical element that belongs to the halogen family and exists as a greenish-yellow gas at standard temperature and pressure. Because of its high reactivity, it is only found naturally in compounds such as sodium chloride (common salt) in seawater, and halite (rock salt). Since it has a strong oxidizing power, it is used as a water disinfectant and bleaching agent, particularly as sodium hypochlorite (bleach). Chlorine is a component of polychlorinated biphenyls (PCBs) and dichlorodiphenyltrichloroethane (DDT), two harmful artificial compounds, and polyvinyl chloride (PVC), a widely used plastic. As a consequence of the emission of chlorofluorocarbons (CFCs), hydrochlorofluorocarbons (HCFCs), and other man-made compounds, chlorine is released into the stratosphere and destroys ozone. |
| *AIR QUALITY MANAGEMENT* | Air pollutant emitted mainly as chlorofluorocarbons, hydrochlorofluorocarbons, carbon tetrachloride, and methyl chloroform, which are ozone-depleting substances. When these substances reach the stratosphere, ultraviolet radiation and chemical reactions break them apart, and they are converted into hydrochloric acid and chlorine nitrate. Those two compounds are finally converted into chlorine monoxide or free chlorine atoms, which destroy stratospheric ozone because of their high reactivity. Along with bromine, chlorine is the main contributor to ozone depletion. |
| *CHEMISTRY* | Non-metallic halogen element with atomic number 17 that exists as a greenish-yellow gas at standard temperature and pressure. It is only found naturally in compounds, such as sodium chloride (common salt) in seawater and in halite (rock salt) or potassium chloride in sylvite. Chlorine has an atomic weight of 35.45 u with $^{35}Cl$ and $^{37}Cl$ as its stable isotopes. |
| *WATER* | Water disinfectant that because of its oxidizing power, |
| *TREATMENT AND SUPPLY* | is added to water usually in a contact tank to kill or inactivate pathogens as part of wastewater treatment. Chlorine is used in pure form or as sodium hypochlorite (bleach), calcium hypochlorite, chlorine dioxide, or chloramine. Chlorine reacts with organic compounds that may be present in water to form harmful by-products: trihalomethanes, and haloacetic acids. Since chlorine is toxic to aquatic life, residual chlorine in treated wastewater needs to be removed (normally with sulfur dioxide) before being discharged into aquatic ecosystems. |

**Table 91. Flexible definition of CHLORINE**

# 6 CONCLUSIONS

This doctoral thesis analyzed the effects of contextual variation in environmental concepts with a focus on its representation in terminological definitions. Specifically, we sought to characterize the conceptual phenomena resulting from contextual variation and to develop guidelines on how to reflect them in terminological definitions from the extraction of knowledge to the actual writing of the definition.

By applying a cognitive linguistics approach, the following general conclusions on terminological definitions were derived from our research:

- Defining a term is not the same as describing its meaning because terms do not have meaning outside of real events. They have semantic potential.

- A term's semantic potential is the raw material for its definition, not its object. The semantic potential is not the object because this would mean that defining a term would involve describing all the conceptual content that the term can activate. This is not viable because a term's semantic potential corresponds to a vast quantity of information that cannot be measured and which is also never activated entirely in real events.

- The object of a definition is a subset of the term's semantic potential, which corresponds to a premeaning. A premeaning is an intermediate stage between semantic potential and meaning. It is an abstraction of the meaning that the term could have in a given context. Thus, the chosen premeaning depends on the contextual constraints applied to the definition.

- Defining a term entails offering a definition for each concept contained in the semantic potential of the term. In the definition, a relevant part of the concept (a premeaning) and the frames that it activates are described.

- Defining a concept entails the description of a relevant subset of its traits (a premeaning) along with the background knowledge that always accompanies this subset of conceptual content.

- In view of prototype effects in conceptual representation, the notion of differentiae in definitions needs to be redefined. Differentiae are not necessary and sufficient features, but rather relevant features.

- In line with Seppälä (2012), the factors that determine trait relevance in terminological definitions can be grouped in three dimensions: (i) ontological (based on the ontological type of the definiendum), functional (based on the users and resource), and contextual (based on contextual constraints).

One the specific objectives of this work was to determine the components of context that affect specialized meaning construction and also the ones that affect conceptual representation in terminological definitions. This led us to the following conclusions:

- Since the object of the terminological definition is an abstraction of the situated meanings of a term under certain contextual constraints (i.e., a premeaning), the context associated with a premeaning is also an abstraction of real contexts (i.e., a precontext).

- There are two main types of context: (i) the mental phenomenon (conceptual context); (ii) the situation (situational context). A precontext is an abstraction of a conceptual context, which is composed of the interpretation of the situational context and background knowledge.

- Context comprises linguistic context, discursive context, sociocultural context, and spatial-temporal context. The precontext for terminological definitions includes linguistic constraints, thematic constraints, cultural constraints, ideological constraints, and diachronic constraints.

- Domains, in their sense of knowledge field, allow the systematic characterization of thematic constraints in terminological definitions. They can be understood as macroframes that guide knowledge organization and categorization of in a given conceptual area.

We also sought to characterize the phenomena resulting from contextual variation in the environmental domain. Our conclusions were the following:

- Contextual variation can be characterized as including three different phenomena: modulation, perspectivization, and subconceptualization. These phenomena are additive in that all concepts experience modulation; some concepts also undergo perspectivization; and finally, a few concepts additionally are subjected to subconceptualization.

- Modulation is the type of contextual variation that only alters minor characteristics of a concept that are neither necessary nor

prototypical. These alterations are not represented in a terminological definition.

- Perspectivization is the type of contextual variation that results in the change of the level of prototypicality of certain traits of a concept consistently in relation to the general environmental premeaning.

- Subconceptualization is the type of contextual variation in which the extension of the concept in relation to the general environmental premeaning is modified.

Regarding the management of the previously mentioned phenomena and, more generally, the development of guidelines for the creation of flexible terminological definition, we obtained the following conclusions:

- By feeding a term extractor with domain-specific corpora and comparing the output, it is possible to obtain a list of contextually variable candidate terms. This list also contains polysemous terms.

- The extraction of domain-specific contextonyms from corpus has proven to be the most effective way of identifying the most relevant traits that characterize contextualized premeanings.

- Domain-specific definitions in most terminological resources do not always reflect the role of a concept in the domain. Therefore, they are less useful than contextonyms to identify how concepts are construed differently depending on the domain.

- The distinction between the three types of contextual variation described in this work is fuzzy. Consequently, determining the kind of contextual variation must be carried out according to the needs of the user and the characteristics of the resource in which the definitions are to be inserted.

- The main difference between perspectivation and subconceptualization regarding the application in flexible

terminological definitions is that a contextualized definition for a subconceptualization represents as necessary a trait that in the general environmental definition does not have necessary status. This is not possible in a contextualized definition of a premeaning that is only a perspective.

- A common problem with perspectivization is that many concepts participate in a broad range of frames, even if only one contextual domain is taken into account. This phenomenon is what we have called *hyperversatility*.

- A special type of hyperversatility is superordinate hyperversatility, which occurs when a concept has many subordinate concepts and each subordinate participate in different frames. As a consequence, the concept behaves as a superordinate-level concept in the domain in question.

- Both cases of hyperversatility seem to be common phenomena. They might be the reason why frame-based resources tend to avoid or inconsistently represent certain types of concept, such as artifacts.

- The creation of a contextualized definition for a hyperversatile or superordinate hyperversatile concept implies that a summary of the main roles of the concept in the corresponding contextual domain needs to be provided in the definition. This entails a longer process of documentation for the terminographer.

- The identification of subconceptualizations is hindered by the fact that conceptual boundaries are not clear-cut, and by the lack of entrenchment of the general environmental premeaning or of the subconceptualization itself. Furthermore, the hierarchical organization of subconceptualizations does not always match conventionalized domain hierarchies.

- A subconceptualization's extension can correspond to another concept's extension. In this case, the contextualized definition should refer to that other concept if it is a relevant concept in that domain.

- If a subconceptualization's extension corresponds to a concept that is not relevant in that domain or does not correspond to another concept, the contextualized definition will represent as necessary the characteristic that gives rise to the context-specific change of extension.

- The choice of the genus of a flexible definition should be guided by the coherence with the rest of the resource and with the frames that the concept activates in accordance with the applied contextual restrictions.

- Genus candidates for terminological definitions can be obtained by means of hypernymic knowledge-patterns and by extracting them from the definitions of other resources. Each contextual domain needs to have its own conceptual hierarchy that reflects how concepts are categorized in that domain.

- The genus of a definition should convey the role of the concept in the most relevant frame it activates. Superordinate concepts can be classified into qualia roles in order to facilitate the selection of genus.

- Frames can be integrated into definitions by means of framing propositions. Unlike conventional definitional propositions that only link the definiendum to other concept, framing propositions relate the concepts activated in the definition with other concepts.

- Since all generic-specific propositions entail property inheritance, concepts may inherit from several superordinate concepts at the same time in a definition.

Finally, our proposal of flexible terminological definition contributes to the improvement of the quality of terminological definitions for these three reasons:

- With our approach, the user is presented with a definition tailored for the domain that he/she has chosen, multiplying the probabilities that the definition will offer him/her the information he/she needs.

- As the analysis of the definitions from other resources showed, terminological definitions are sometimes not properly adapted to the domain from which they are supposed to be written. In our approach, the knowledge represented in each contextualized definition has been chosen based on the result of the analysis of contextonyms. This ensures that the definition actually reflects how the definiendum is construed in that domain, which might differ from the viewpoint adopted in the environment as a whole or in other environmental subdomains.

- The genus and differentiae selection in flexible definitions is guided by contextonym-based analysis and the aim of inserting the concept in the frames in which it participates. By representing concepts in context-specific hierarchies and frames, the flexible terminological definition provides a knowledge representation that better resembles the human conceptual system than traditional terminological definitions. As a consequence, a flexible definition not only provides more relevant information, but it also accomplishes this in a way that potentially facilitates and enhances knowledge acquisition.

As for future work, we plan to address certain limitations that were encountered during this research. First of all, the way in which conceptual propositions are encoded in EcoLexicon has proven not to be entirely suitable for the representation of the information in flexible definitions. To this respect, an ontology-based system for the management of definitional templates could provide more expressiveness and flexibility with the additional advantage that property inheritance would be automatic, and inferences could be implemented.

This ontology-based system for the management of definitional templates could also encode frames. Nonetheless, further study is necessary to determine the best way in which frames can be formalized in order to support the creation of flexible definitions. The problem of hyperversatility will certainly present a challenge in this respect.

Another limitation that will be addressed in future work is the fact that the creation of flexible terminological definitions is a laborious process that would benefit from being streamlined. More specifically, the contextonym-based analysis of contextual variation is time-consuming. The development of a black list filtering terms that are not informative (such as *example*, *call or require)* would speed up the analysis. However, this list would have to be carefully compiled because certain common English words that may apparently be too generic, such as *use* or *part,* are also indicative of important relations (for example, functional or meronymic in the case of *use* and *part*).

Additionally, contextonym analysis could be further improved if morphological derivations on the lists were grouped together, if complex terms were also included as contextonyms, or by developing a system that would automatically compare contextonyms in different domains. Furthermore, we are aware that the effectiveness of contextonym analysis using SketchEngine as presented in this work is currently limited to non-polysemous terms. A possible solution would involve the use of word-disambiguation techniques to create sense-specific contextonym lists.

The methodology for the extraction of domain-specific definitional knowledge could also be enhanced by the development of other knowledge-pattern-based sketch grammars to extract conceptual relations other than generic-specific ones, such as functional, causal, or meronymic.

Finally, we envision extending the scope of the study of contextual variation and its application to terminological definitions in two ways. On the one hand, since our analysis was limited to ENTITIES designated by simple terms, we propose to include ATTRIBUTES and PROCESSES, as well as concepts associated with complex terms. On the other hand, it would be relevant to analyze contextual variation in specialized domains other than the environment and to incorporate other languages, especially from a contrastive point of view.

We believe that this research has contributed to the understanding of contextual variation in terminology and has shown effective ways of representing it in terminological definitions. Nonetheless, having welcomed context into the realm of terminological definitions, countless possibilities open up, and new research avenues await to be explored.

# 7 BIBLIOGRAFÍA

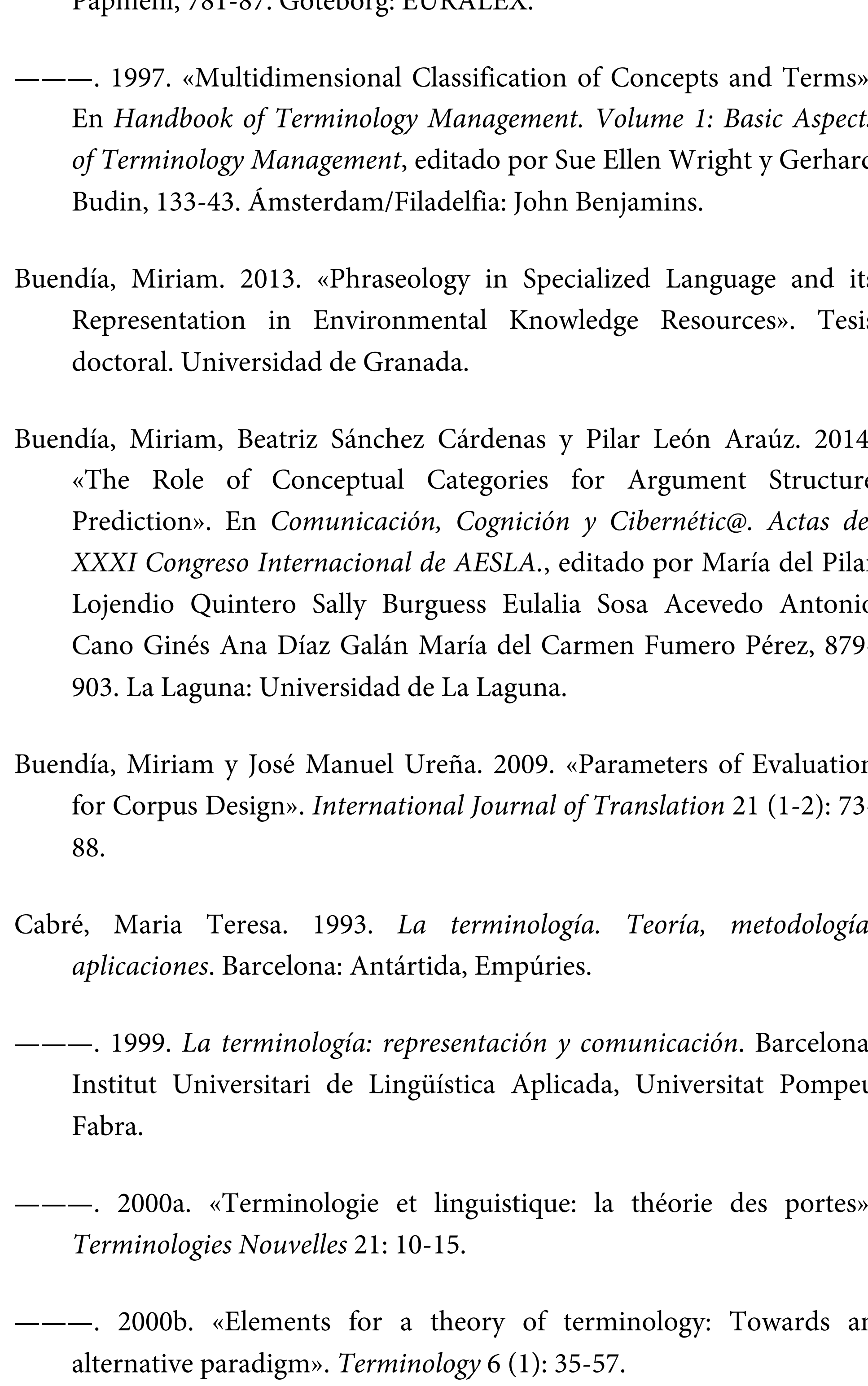

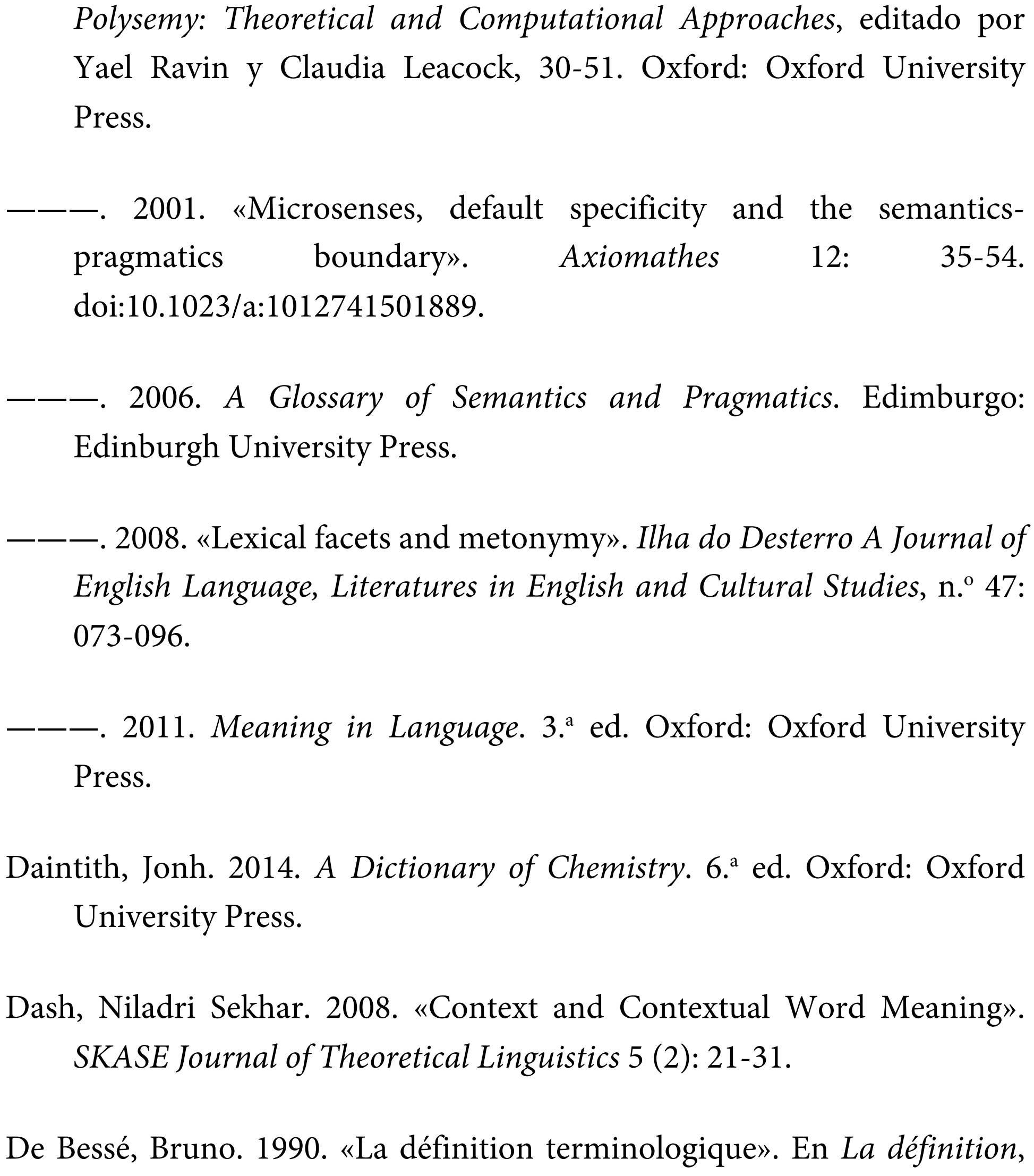

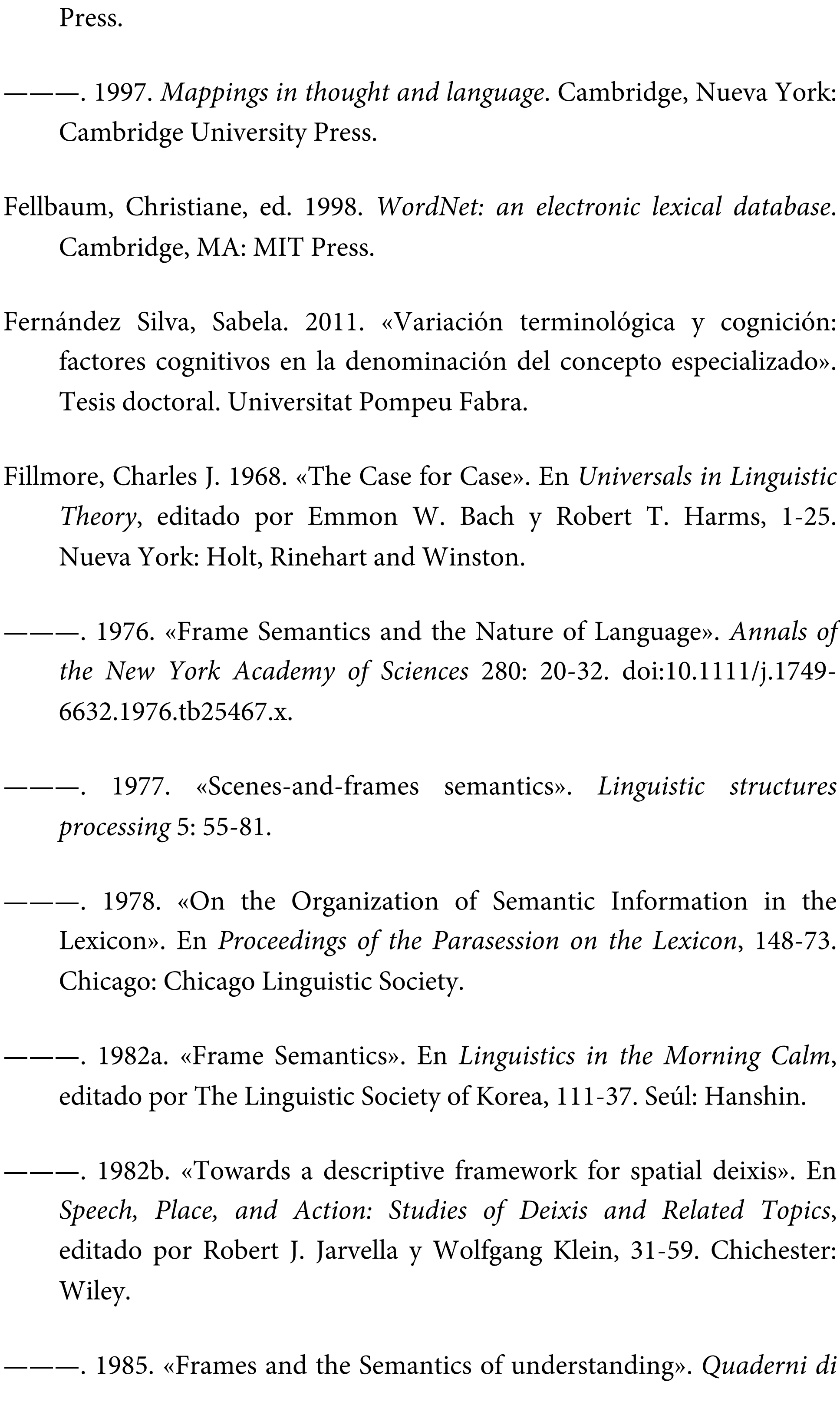

# ANEXO 1. OBRAS INCLUIDAS EN EL CORPUS MULTI

| Subcorpus | Referencias bibliográficas |
|---|---|
| AGR | Sheaffer, Craig y Kristine M. Moncada. 2012. *Introduction to Agronomy : Food, Crops, And Environment*. Clfton Park NY: Delmar Cengage Learning. |
| AIR | Godish, Thad. 2004. *Air quality*. Boca Raton: Lewis Publishers.<br>Schnelle, Karl B. y Charles A. Brown. 2002. *Air Pollution Control Technology Handbook*. Boca Raton: CRC Press. |
| ATM | Wallace, John y Peter Victor Hobbs. 2006. *Atmospheric Science: an Introductory Survey*. Ámsterdam, Boston: Elsevier Academic Press. |
| BIO | Reece, Jane B, Lisa A Urry, Michael L Cain, Steven Alexander Wasserman, Peter V Minorsky, Rob Jackson y Neil A Campbell. 2014. *Campbell Biology*. Boston: Pearson. |
| CEN | Green, Don W. y Robert H. Perry, eds. 2008. *Perry's Chemical Engineer's Handbook*. Nueva York, Chicago: McGraw-Hill. |
| CHE | Chang, Raymond y Kenneth A. Goldsby. 2013. *Chemistry*. Nueva York: McGraw-Hill. |
| CIV | Bhavikatti, S S. 2010. *Basic Civil Engineering*. Nueva Delhi: New Age International.<br>Chen, Wai-Fah y J Y Richard Liew, eds. 2003. *The Civil Engineering Handbook*. Boca Raton: CRC Press.<br>Darling, Peter, ed. 2011. *SME Mining Engineering Handbook*. Englewood: Society for Mining Metallurgy and Exploration.<br>Sorensen, Robert. 2006. *Basic Coastal Engineering*. Nueva York: Springer. |
| ENE | Thumann, Albert y Paul Mehta. 2012. *Handbook of Energy Engineering*. Lilburn, GA: Fairmont Press.<br>Georgiadis, Michael, Eustathios S Kikkinides y Efstratios N Pistikopoulos. 2008. *Energy Systems Engineering*. Weinheim: Wiley-VCH. |
| ENV | Raven, Peter, Linda R Berg y David M Hassenzahl. 2010. *Environment*. Hoboken: Wiley.<br>Kaushik, Anubha y C P Kaushik. 2010. *Basics of Environment and Ecology*. Nueva Delhi: New Age International.<br>Pfafflin, J y E N Ziegler. 2006. *Encyclopedia of Environmental Science* |

| | *and Engineering*. Nueva York: Taylor & Francis. |
|---|---|
| GEO | Lutgens, Frederick y Edward J Tarbuck. 2012. *Essentials of Geology*. Boston: Prentice Hall. |
| HYD | Davie, Tom. 2008. *Fundamentals of Hydrology*. Londres, Nueva York: Routledge.<br>Hingray, Benoît, A Musy y Cécile Picouet. 2015. *Hydrology : a Science for Engineers*. Boca Raton: CRC Press. |
| PHY | Benenson, Walter, Horst Stöcker, J Harris y Holger Lutz. 2006. *Handbook of Physics*. Nueva York: Springer. |
| SOI | Lal, Rattan, ed. 2006. *Encyclopedia of Soil Science*. Nueva York: Taylor & Francis. |
| WAS | Letcher, T.M. y Daniel A. Vallero, eds. 2011. *Waste: a Handbook for Management*. Burlington, MA: Academic Press.<br>Vaughn, Jacqueline. 2009. *Waste Management: a Reference Handbook*. Santa Barbara: ABC-CLIO. |
| WAT | Davis, Mackenzie. 2010. *Water and Wastewater Engineering Design Principles and Practice*. Nueva York : McGraw-Hill. |

# ANEXO 2. LISTA DE TÉRMINOS POLISÉMICOS CON DEFINICIONES

| **Término**[100] | **Definición 1** | **Fuente** | **Definición 2** | **Fuente** |
|---|---|---|---|---|
| acid | A molecular entity or chemical species capable of donating a hydron (proton) | IUPAC. 2014. «IUPAC Compendium of Chemical Terminology - the Gold Book». http://goldbook.iupac.org/index.html. | [A molecular entity or chemical species] capable of forming a covalent bond with an electron pair. | IUPAC. 2014. «IUPAC Compendium of Chemical Terminology - the Gold Book». http://goldbook.iupac.org/index.html. |
| basin | A depression, usually of considerable size, which may be erosional or structural in origin. | Allaby, Michael. 2010. *A Dictionary of Ecology*. Oxford: Oxford University Press. | Total land area drained by a river and its tributaries. | FAO. 2015. «FAO TERM portal». http://www.fao.org/faoterm/en/. |
| bed | The floor or bottom on which any body of water rests. | Translation Bureau / Bureau de la traduction [Canada]. 2015. «TERMIUM Plus». http://www.btb.termiumplus.gc.ca/. | A layer of deposited sediment within a sedimentary rock. | Park, Chris y Michael Allaby. 2013. *A Dictionary of Environment and Conservation*. 2nd ed. Oxford: Oxford University Press. |
| biomass | The total weight of all living organisms in an area. The term is sometimes used to refer specifically to organisms of one type. | Park, Chris y Michael Allaby. 2013. *A Dictionary of Environment and Conservation*. 2nd ed. Oxford: Oxford University Press. | Living material (such as wood and vegetation) that is grown or produced for use as fuel." | Park, Chris y Michael Allaby. 2013. *A Dictionary of Environment and Conservation*. 2nd ed. Oxford: Oxford University Press. |

[100] Aunque muchos de estos términos presentan varios sentidos, esta lista se ha limitado a los dos más comunes en el corpus.

| **Término**[100] | **Definición 1** | **Fuente** | **Definición 2** | **Fuente** |
| --- | --- | --- | --- | --- |
| cloud | A mass of water droplets and/or ice crystals in the air, formed by condensation of water vapour around nuclei such as dust, salt, and soil particles as a mass of air rises and cools. There are three main types of cloud: stratus, cumulus, and cirrus. | Park, Chris y Michael Allaby. 2013. A Dictionary of Environment and Conservation. 2nd ed. Oxford: Oxford University Press. | Any suspension of particulate matter, such as dust or smoke, dense enough to be seen. | Translation Bureau / Bureau de la traduction [Canada]. 2015. «TERMIUM Plus». http://www.btb.termiumplus.gc.ca/. |
| crop | A plant (such as cereals, vegetables, or fruit plants) that is cultivated and harvested for use by people or livestock. | Park, Chris y Michael Allaby. 2013. *A Dictionary of Environment and Conservation*. 2nd ed. Oxford: Oxford University Press. | The amount of a crop that is harvested from a given area of land (such as a field) in a single growing season, usually expressed as yield per unit area (for example, tonnes per hectare). | Park, Chris y Michael Allaby. 2013. *A Dictionary of Environment and Conservation*. 2nd ed. Oxford: Oxford University Press. |
| feed | Ground or processed grains that are fed to animals, including hay such as alfalfa or silage | Park, Chris y Michael Allaby. 2013. *A Dictionary of Environment and Conservation*. 2nd ed. Oxford: Oxford University Press. | Material supplied to a processing unit for treatment. | Translation Bureau / Bureau de la traduction [Canada]. 2015. «TERMIUM Plus». http://www.btb.termiumplus.gc.ca/. |
| forest | An ecosystem that is dominated by trees, the crowns of which touch to form a continuous canopy. | Park, Chris y Michael Allaby. 2013. *A Dictionary of Environment and Conservation*. 2nd ed. Oxford: Oxford University Press. | The trees that comprise such an ecosystem. | Park, Chris y Michael Allaby. 2013. *A Dictionary of Environment and Conservation*. 2nd ed. Oxford: Oxford University Press. |

| **Término**[100] | **Definición 1** | **Fuente** | **Definición 2** | **Fuente** |
|---|---|---|---|---|
| glass | An amorphous solid in which the atoms form a random network. Glasses do not have the rigidity of crystals and have a viscosity that increases as the temperature is lowered. At very low temperatures, when the viscosity is very large, glasses can become elastic and brittle. […] | Daintith, Jonh. 2014. *A Dictionary of Physics*. Oxford: Oxford University Press. | A hard, amorphous, inorganic, usually transparent, brittle substance made by fusing silicates, sometimes borates and phosphates, with certain basic oxides and then rapidly cooling to prevent crystallization. | Translation Bureau / Bureau de la traduction [Canada]. 2015. «TERMIUM Plus». http://www.btb.termiumplus.gc.ca/. |
| grain | A part of a metal, timber, or ceramic material representing an individual crystal. Usually grains are visible with an optical aid, for example, a microscope. | Gorse, Christopher, David Johnston y Martin Pritchard. 2013. *A Dictionary of Construction, Surveying and Civil Engineering*. Oxford: Oxford University Press. | The dry, starchy seed that is produced by cereal grasses. | Park, Chris y Michael Allaby. 2013. *A Dictionary of Environment and Conservation*. 2nd ed. Oxford: Oxford University Press. |
| heat | Energy caused by the movement of atoms and molecules that can be transferred from one place to another by conduction, convection, or radiation. | Gorse, Christopher, David Johnston y Martin Pritchard. 2013. *A Dictionary of Construction, Surveying and Civil Engineering*. Oxford: Oxford University Press. | The perception of warmth. | Gorse, Christopher, David Johnston y Martin Pritchard. 2013. *A Dictionary of Construction, Surveying and Civil Engineering*. Oxford: Oxford University Press. |
| metal | Any of a large class of chemical elements that, with certain exceptions, are ductile, malleable solids with a characteristic lustre, are good conductors of electricity and heat, and generally form simple cations. | Cammack, Richard, Teresa Atwood, Peter Campbell, Howard Parish, Anthony Smith, Frank Vella y John Stirling. 2008. *Oxford Dictionary of Biochemistry and Molecular Biology*. Oxford: Oxford University Press. | Any substance or material consisting entirely or predominantly of one or more metals. | Cammack, Richard, Teresa Atwood, Peter Campbell, Howard Parish, Anthony Smith, Frank Vella y John Stirling. 2008. *Oxford Dictionary of Biochemistry and Molecular Biology*. Oxford: Oxford University Press. |

| **Término**[100] | **Definición 1** | **Fuente** | **Definición 2** | **Fuente** |
|---|---|---|---|---|
| meter | An instrument or device for measuring (and recording) the quantity of something (either instantaneously or cumulatively). | Cammack, Richard, Teresa Atwood, Peter Campbell, Howard Parish, Anthony Smith, Frank Vella y John Stirling. 2008. *Oxford Dictionary of Biochemistry and Molecular Biology*. Oxford: Oxford University Press. | The fundamental unit of length in the metric system, which is equal to 3.28 ft, 39.37 in, or 100 cm. | Translation Bureau / Bureau de la traduction [Canada]. 2015. «TERMIUM Plus». http://www.btb.termiumplus.gc.ca/. |
| mineral | A solid, naturally occurring, usually inorganic substance with a characteristic chemical composition, typically with a crystalline structure, and possessing physical properties of hardness, lustre, colour, cleavage, relative density, and fracture. Rocks are composed of minerals. | Park, Chris y Michael Allaby. 2013. *A Dictionary of Environment and Conservation*. 2nd ed. Oxford: Oxford University Press. | Any organic substance that is obtained by mining (e.g. coal, oil, etc.). | Park, Chris y Michael Allaby. 2013. *A Dictionary of Environment and Conservation*. 2nd ed. Oxford: Oxford University Press. |
| nitrate | The $NO_3^-$ ion, derived from nitric acid. | Cammack, Richard, Teresa Atwood, Peter Campbell, Howard Parish, Anthony Smith, Frank Vella y John Stirling. 2008. *Oxford Dictionary of Biochemistry and Molecular Biology*. Oxford: Oxford University Press. | Any salt or ester of nitric acid. | Cammack, Richard, Teresa Atwood, Peter Campbell, Howard Parish, Anthony Smith, Frank Vella y John Stirling. 2008. *Oxford Dictionary of Biochemistry and Molecular Biology*. Oxford: Oxford University Press. |

| **Término**[100] | **Definición 1** | **Fuente** | **Definición 2** | **Fuente** |
|---|---|---|---|---|
| ocean | One of the major divisions of the vast expanse of salt water covering part of the earth | European Environment Agency. 2015. «Environmental Terminology and Discovery Service». http://glossary.eea.europa.eu/. | The vast body of salt water which covers almost three fourths of the earth's surface. | Translation Bureau / Bureau de la traduction [Canada]. 2015. «TERMIUM Plus». http://www.btb.termiumplus.gc.ca/. |
| oil | Any of numerous mineral, vegetable and synthetic substances, and animal or vegetable fats that are generally unctuous, slippery, combustible, viscous, liquid or liquefiable at room temperatures, soluble in various organic solvents, but not in water. | Translation Bureau / Bureau de la traduction [Canada]. 2015. «TERMIUM Plus». http://www.btb.termiumplus.gc.ca/. | A thick, black, sticky hydrocarbon substance that is used to produce fuel (petroleum) and materials (plastics). Oil is one of the world's principal fossil fuel resources, and known reserves are fast being depleted. […] | Park, Chris y Michael Allaby. 2013. *A Dictionary of Environment and Conservation*. 2nd ed. Oxford: Oxford University Press. |
| organism | An individual living system, such as an animal, plant, or microorganism, that is capable of reproduction, growth, and maintenance. | Martin, Elizabeth y Robert Hine. 2014. *A Dictionary of Biology*. 6th ed. Oxford: Oxford University Press. | A system or organization consisting of interdependent parts, compared to a living being. | Stevenson, Angus, ed. 2010. *Oxford Dictionary of English*. Oxford University Press. |

| **Término**[100] | **Definición 1** | **Fuente** | **Definición 2** | **Fuente** |
|---|---|---|---|---|
| oxide | Any binary compound of oxygen and some other element. | Cammack, Richard, Teresa Atwood, Peter Campbell, Howard Parish, Anthony Smith, Frank Vella y John Stirling. 2008. *Oxford Dictionary of Biochemistry and Molecular Biology*. Oxford: Oxford University Press. | Any compound in which an oxygen atom has been attached to a heteroatom of a precursor compound. This definition includes compounds of general formulae: RC≡NO (nitrile oxides; e.g. benzonitrile oxide), >C=N(O)R or R,R′,R″NO (N-oxides, amine oxides; e.g. trimethylamine oxide); and R,R′SO (S-oxides, sulfoxides; e.g. methionine S-oxide; dimethyl sulfoxide). | Cammack, Richard, Teresa Atwood, Peter Campbell, Howard Parish, Anthony Smith, Frank Vella y John Stirling. 2008. *Oxford Dictionary of Biochemistry and Molecular Biology*. Oxford: Oxford University Press. |
| particle | Small part or fragment of a material. | Gorse, Christopher, David Johnston y Martin Pritchard. 2013. *A Dictionary of Construction, Surveying and Civil Engineering*. Oxford: Oxford University Press. | One of the fundamental components of matter. | Daintith, Jonh. 2014. *A Dictionary of Physics*. Oxford: Oxford University Press. |
| plane | A flat surface defined by the condition that any two points in the plane are joined by a straight line that lies entirely in the surface. | Daintith, Jonh. 2014. *A Dictionary of Physics*. Oxford: Oxford University Press. | An aeroplane. | Stevenson, Angus, ed. 2010. *Oxford Dictionary of English*. Oxford University Press. |

| **Término[100]** | **Definición 1** | **Fuente** | **Definición 2** | **Fuente** |
|---|---|---|---|---|
| plant | Major equipment and machinery used in industrial processes. | Schaschke, Carl. 2014. *A Dictionary of Chemical Engineering*. Oxford: Oxford University Press. | Any organism of the kingdom Plantae. Plants are characterized by their ability to effect photosynthesis and the possession of rigid cell walls that contain cellulose. | Cammack, Richard, Teresa Atwood, Peter Campbell, Howard Parish, Anthony Smith, Frank Vella y John Stirling. 2008. *Oxford Dictionary of Biochemistry and Molecular Biology*. Oxford: Oxford University Press. |
| plate | A flat perforated horizontal sheet of metal used in distillation and absorption columns designed to provide an intimate contact between a rising vapour or gas and a liquid to allow vapour–liquid equilibrium to be achieved. | Schaschke, Carl. 2014. *A Dictionary of Chemical Engineering*. Oxford: Oxford University Press. | A segment of the lithosphere, which has little volcanic or seismic activity but is bounded by almost continuous belts (known as plate margins) of earthquakes and, in most cases, by volcanic activity and young subsea or subaerial mountain chains. | Allaby, Michael. 2015. *A Dictionary of Geology and Earth Sciences*. 4.a ed. Oxford: Oxford University Press. |
| reactor | A vessel for containing and controlling a biochemical, chemical or nuclear reaction process. | Atkins, Tony y Marcel Escudier. 2014. *A Dictionary of Mechanical Engineering*. Oxford: Oxford University Press. | Any device, such as an inductor or capacitor, that introduces reactance into an electrical circuit. | Daintith, Jonh. 2014. *A Dictionary of Physics*. Oxford: Oxford University Press. |
| reservoir | A substance that captures another and retains it temporarily. | Translation Bureau / Bureau de la traduction [Canada]. 2015. «TERMIUM Plus». http://www.btb.termiumplus.gc.ca/. | An artificial or natural storage place for water, such as a lake or pond, from which the water may be withdrawn as for irrigation, municipal water supply, or flood control. | European Environment Agency. 2015. «GEneral Multilingual Environmental Thesaurus (GEMET)». http://www.eionet.europa.eu/gemet. |

| **Término**[100] | **Definición 1** | **Fuente** | **Definición 2** | **Fuente** |
|---|---|---|---|---|
| rock | Rocky hill, mountain, or cliff; or a large boulder. | Translation Bureau / Bureau de la traduction [Canada]. 2015. «TERMIUM Plus». http://www.btb.termiumplus.gc.ca/. | A hard material consisting of a collection of different minerals, which forms the majority of the relatively thin outer curst of the surface of the earth. Rock falls into three broad groups based on its origin, namely: igneous,metamorphic, and sedimentary. | Gorse, Christopher, David Johnston y Martin Pritchard. 2013. *A Dictionary of Construction, Surveying and Civil Engineering*. Oxford: Oxford University Press. |
| salt | Any member of a class of compounds formed, together with water, by reaction of an acid with a base (most commonly an inorganic acid and metallic base). | Cammack, Richard, Teresa Atwood, Peter Campbell, Howard Parish, Anthony Smith, Frank Vella y John Stirling. 2008. *Oxford Dictionary of Biochemistry and Molecular Biology*. Oxford: Oxford University Press. | Sodium chloride. | Cammack, Richard, Teresa Atwood, Peter Campbell, Howard Parish, Anthony Smith, Frank Vella y John Stirling. 2008. *Oxford Dictionary of Biochemistry and Molecular Biology*. Oxford: Oxford University Press. |
| sand | In the commonly used Udden–Wentworth scale, particles between 62.5 and 2000μm. Other classifications exist. In pedology, sand is defined as mineral particles of diameter 2.0–0.02 mm in the international system, and as 2.0–0.5 mm diameter particles in the USDA (American) system. | Allaby, Michael. 2015. *A Dictionary of Geology and Earth Sciences*. 4.a ed. Oxford: Oxford University Press. | A class of soil texture. | Allaby, Michael. 2015. *A Dictionary of Geology and Earth Sciences*. 4.a ed. Oxford: Oxford University Press. |

| **Término**[100] | **Definición 1** | **Fuente** | **Definición 2** | **Fuente** |
|---|---|---|---|---|
| sea | In general, the marine section of the globe as opposed to that of the land. | European Environment Agency. 2015. «GEneral Multilingual Environmental Thesaurus (GEMET)». http://www.eionet.europa.eu/gemet. | The name given to a body of salt water smaller than an ocean and generally in proximity to a continent. | European Environment Agency. 2015. «GEneral Multilingual Environmental Thesaurus (GEMET)». http://www.eionet.europa.eu/gemet. |
| sulfate | The anion $SO_4^{2-}$. | Cammack, Richard, Teresa Atwood, Peter Campbell, Howard Parish, Anthony Smith, Frank Vella y John Stirling. 2008. *Oxford Dictionary of Biochemistry and Molecular Biology*. Oxford: Oxford University Press. | A salt or ester of sulfuric acid of which most of the salts except those of barium, lead, strontium, and calcium are fairly soluble in water. | Translation Bureau / Bureau de la traduction [Canada]. 2015. «TERMIUM Plus». http://www.btb.termiumplus.gc.ca/. |

# ANEXO 3. DEFINICIONES DE POLLUTANT

| Fuente | Dominio | Definición |
| --- | --- | --- |
| European Environment Agency. 2015. «GEneral Multilingual Environmental Thesaurus (GEMET)». http://www.eionet.europa.eu/gemet. | ENV | Any **substance**, usually a residue of human activity, which has an undesirable effect upon the environment. |
| United States Environmental Protection Agency. 2015. «Terminology Services». http://ofmpub.epa.gov/sor_internet/registry/termreg/searchandretrieve/termsandacronyms/search.do | ENV | An **emitted substance** that is regulated or monitored for its potential to cause harm to the health of individuals or to the environment. |
| United States Environmental Protection Agency. 2015. «Terminology Services». http://ofmpub.epa.gov/sor_internet/registry/termreg/searchandretrieve/termsandacronyms/search.do | ENV | A **contaminant** in a concentration or amount that adversely alters the physical, chemical, or biological properties of the environment. The term includes pathogens, toxic metals, carcinogens, oxygen-demanding materials, and all other harmful substances. With reference to nonpoint sources, the term is sometimes used to apply to contaminants released in low concentrations from many activities that collectively degrade water quality. As defined in the federal Clean Water Act, pollutant means dredged spoil; solid waste; incinerator residue; sewage; garbage; sewage sludge; munitions; chemical wastes; biological materials; radioactive materials; heat; wrecked or discarded equipment; rock; sand; cellar dirt; and industrial, municipal, and agricultural waste discharged into water. |
| United States Environmental Protection Agency. 2015. «Terminology Services». http://ofmpub.epa.gov/sor_internet/registry/termreg/searchandretrieve/termsandacronyms/search.do | ENV | A **compound** in the air that has an impact on air quality. This is a general term and does not imply a criteria air pollutant, a hazardous air pollutant, or a greenhouse gas. Mode-specific pollutants for mobile sources are also considered pollutants (e.g., BRK__PM10 for $PM10_{10}$ emitted from vehicle braking). |
| United States Environmental Protection Agency. 2015. «Terminology Services». http://ofmpub.epa.gov/sor_internet/registry/termreg/searchandretrieve/termsandacronyms/search.do | ENV | Any **introduced gas, liquid or solid** that makes a resource unfit for a specific purpose. |

| Fuente | Dominio | Definición |
|---|---|---|
| FAO. 2015. «FAO TERM portal». http://www.fao.org/faoterm/en/. | ENV | Any **substance** not intentionally added to food, which is present in such food as a result of production (including operations carried out in crop and animal husbandry), manufacture, processing, preparation, treatment, packing, packaging, transport or holding of such food or as a result of environmental contamination. The term includes chemical and biological substances not desirable in food but does not include insect fragments, rodent hairs and other extraneous matter. |
| FAO. 2015. «FAO TERM portal». http://www.fao.org/faoterm/en/. | ENV | A **contaminant** that in a certain concentration or amount will adversely alter the physical, chemical, or biological function of the environment; includes pathogens, heavy metals, carcinogens, oxygen-demanding materials, and all other harmful substances, including dredged spoil, solid waste, incinerator residue, sewage, garbage, sewage sludge, munitions, chemical wastes, biological materials, radioactive materials, and industrial, municipal and agricultural wastes discharged into the aquatic environment. |
| FAO. 2015. «FAO TERM portal». http://www.fao.org/faoterm/en/. | ENV | **Substance** that is present in concentrations that may harm organisms (humans, plants and animals) or exceed an environmental quality standard. The term is frequently used synonymously with contaminant. |
| Lenntech. «Water Glossary». http://www.lenntech.com/water-glossary.htm | WAT | A **contaminant** at a concentration high enough to endanger the life of organisms. |
| Translation Bureau / Bureau de la traduction [Canada]. 2015. «TERMIUM Plus». http://www.btb.termiumplus.gc.ca/. | ENV | **Any thing, living or not living, or any physical agent** (e.g., heat, sound) that in its excess makes any part of the environment undesirable; if water, undesirable for drinking, recreation, visual enjoyment, or as a habitat for the aquatic life normal to it; if air, undesirable for breathing, for the condition of buildings and monuments exposed to it, or for animal and plant life; if soil and land, undesirable for raising food and fiber, animals, or for recreation or aesthetic enjoyment. |
| Park, Chris y Michael Allaby. 2013. *A Dictionary of Environment and Conservation*. 2nd ed. Oxford: Oxford University Press. | ENV | A **substance** that pollutes or causes pollution. |
| Schaschke, Carl. 2014. *A Dictionary of Chemical Engineering*. Oxford: Oxford University Press. | ENV | A **chemical substance** that when released into the environment gives rise to harmful and damaging effects on living organisms. The substance can be either a toxic substance that is harmful to the environment by being resistant to biodegradation such as pesticides, or can already be present in the environment but is added in excessive amounts such as nitrogen into the soil that accumulates in lakes and rivers. |

| **Fuente** | **Dominio** | **Definición** |
| --- | --- | --- |
| Allaby, Michael. 2013. *A Dictionary of Plant Sciences*. Oxford: Oxford University Press. | ENV | A **substance** that enters the environment or becomes concentrated within it, and that will, or may, harmfully affect human life or that of desirable species. Pollutants are by-products of human activity (compare allelopathy) and the term embraces noise and the release of substances at temperatures markedly higher than those of the receiving media. |
| Martin, Elizabeth y Robert Hine. 2014. *A Dictionary of Biology*. 6th ed. Oxford: Oxford University Press. | ENV | Any **substance**, produced and released into the environment as a result of human activities, that has damaging effects on living organisms. Pollutants may be toxic substances (e.g. pesticides) or natural constituents of the atmosphere (e.g. carbon dioxide) that are present in excessive amounts. See pollution. |
| Allaby, Michael. 2010. *A Dictionary of Ecology*. Oxford: Oxford University Press. | ENV | A **by product of human activities** which enters or becomes concentrated in the environment, where it may cause injury to humans or desirable species. In addition to chemical substances, the term also embraces noise, vibration, and alterations to the ambient temperature. |
| Pinelands. «Water Glossary». http://www.state.nj.us/pinelands/infor/educational/curriculum/pinecur/pwg.htm | WAT | A **chemical** or **physical component** found in water, land or air that makes conditions less favorable for living things (Improper fertilizer application can act as a pollutant.) |
| Oasis Environmental. 2006. «Glossary». http://oasisenviro.co.uk/Glossary%20N%20to%20R.htm | ENV | A **chemical** or **substance** that causes harm in the environment |
| Oregon Museum of Science and Industry. 2005. «Experiencing Chemistry Glossary». https://web.archive.org/web/20051120153013/http://www.omsi.edu/visit/chemistry/glossary.cfm | ENV | A **chemical** that is unwanted in a particular environment |
| 1st Water Filters. 2007. «Water Filtration Glossary». https://web.archive.org/web/20070122224404/http://www.1st-water-filters.com/water-filtration-glossary.html | WAT | A **contaminant** at a concentration high enough to endanger the life of organisms. |
| Pure1. 2010. «The WQA Glossary of Terms». http://www.pure1.com/glossary | WAT | A **contaminant** existing at a concentration high enough to endanger the environment or the public health or to be otherwise objectionable. |

| Fuente | Dominio | Definición |
|---|---|---|
| Aquatic Testing Laboratory. «Terminology». http://www.aquatictestinglabs.com/term.html | WAT | A **contaminant** introduced into a receiving water which is subject to technology-based or water-quality based effluent limitations in the permit. |
| Indian and Northern Affairs Canada. «Glossary of Water Management Terms». http://www.aadnc-aandc.gc.ca/eng/1100100028092/1100100028094 | WAT | A **contaminant** that negatively impacts the physical, chemical, or biological properties of the environment |
| Department of Ecology (State of Washington). «Glossary of terms». http://www.ecy.wa.gov/programs/wq/plants/management/manual/appendixa.html | WAT | A **contaminant**, a substance that is not naturally present in water or occurs in unnatural amounts that can degrade the physical, chemical, or biological properties of the water. Pollutants can be chemicals, disease-producing organisms, silt, toxic metals, oxygen-demanding materials, to name a few. |
| San Francisco Estuary Partnership. 2012. «Glossary». http://sfep.sfei.org/wp-content/uploads/2012/12/9CCMPGlossary.pdf | WAT | A **harmful chemical** or **waste material** discharged into the environment. Persistent pollutants are those that do not degrade, causing potential long-term chronic toxicity to biotas. |
| FirstEnergy. 2015. «Environmental Glossary». https://www.firstenergycorp.com/content/fecorp/environmental/environmental_glossary.html | ENV | A **substance** in certain concentrations that is capable of degrading the environment's usefulness. Many such substances in diluted forms are not harmful to humans or to the environment. |
| PhysicalGeography.net. 2006. «Glossary of terms». http://www.physicalgeography.net/physgeoglos/p.html | ENV | A **substance** that has a harmful effect on the health, survival, or activities of humans or other living organisms. |
| Minnesota Pollution Control Agency. 2002. «Glossary». https://web.archive.org/web/20020202072241/http://www.pca.state.mn.us/gloss/glossary.cfm?alpha=P&header=1 | ENV | A **substance** that pollutes air, water or land. |
| Healthy Waterways. 2006. «Glossary». https://web.archive.org/web/20060821164846/http://www.healthywaterways.org/PAGE170746PMIJM4TZ.html | WAT | A **substance** which may naturally occur but be present at harmful levels (e.g. sediment or nutrients) or which may be unnatural in the environment (e.g. pesticides) |

| Fuente | Dominio | Definición |
|---|---|---|
| Save Our Seine. 2007. «Glossary». https://web.archive.org/web/20070623011458/http://www.saveourseine.com/glossary/index.html | WAT | A **substance** which negatively affects the purity of some component of the environment such as air or water |
| Veolia Environmental Services. 2008. «Dewatering Glossary». https://web.archive.org/web/20080402004951/http://www.onyxdewaters.com/dewatering_glossary.htm | WAT | A **substance**, **organism** or **energy form** present in amounts that impair or threaten an ecosystem to the extent that its current or future uses are prevented. |
| US Department of Defense. 2013. «Environmental Exposure Report Glossary». https://web.archive.org/web/20130219054330/http://www.gulflink.osd.mil/owf_ii/owf_ii_taba.htm | ENV | An **environmental contaminant**. |
| Ocean Institute. 2004. «Science Glossary». https://web.archive.org/web/20041029211837/http://www.ocean-institute.org/edu_programs/materials/P/Glo/S_Glos.htm | ENV | Any **chemical substance** from outside an ecosystem, whether natural or man-made, that impacts the biological processes in the ecosystem. |
| Wetland Curriculum Resource (Ontario). «Glossary». http://www.torontozoo.com/adoptapond/curriculum/m1-glossary.html | WAT | Any **impurity**, **contaminant**, or **harmful substance** found in the environment |
| Wordatlas. 2015. «Glossary». http://worldatlas.com/h2oterms.htm | ENV | Any **inorganic or organic substance** that contaminates air, water or soil |
| Society for Risk Analysis. 2006. «Risk Analysis Glossary». https://web.archive.org/web/20060117202135/http://www.sra.org/resources_glossary_p-r.php | ENV | Any **material** entering the environment that has undesired effects. |
| The Coastal Correspondent. «Glossary». http://www.marshmission.com/coastal_glossary.cfm | ENV | Any **material** that can damage living things or their environments |

| Fuente | Dominio | Definición |
|---|---|---|
| New Jersey Department of Environmental Protection. 2015. «Glossary of Technical Terms». http://www.nj.gov/dep/srp/guidance/fspm/glossary.htm | ENV | Any **substance** defined as such pursuant to the Water Pollution Control Act, N.J.S.A 58:10A- 1 et seq. |
| WaterBank. «Glossary of terms». http://waterbank.com/glossary.html | WAT | Any **substance** introduced into the environment that adversely affects the usefulness of a resource. |
| MSC Environment. 2004. «Household Waste Management». https://web.archive.org/web/20060716160447/http://ipp-environment.com/hhwm/glossary.html | WAS | Any **substance** of such character and in such quantities that upon reaching the environment (soil, water or air), is degrading in effect so as to impair the environment's usefulness or render it offensive. |
| Montana State University. 2010. «Water Quality Glossary». https://web.archive.org/web/20100609153201/http://www.animalrangeextension.montana.edu/LoL/Module-3/3-Glossary.htm | WAT | Any **substance** of such character and in such quantities that when it reaches a body of water, soil, or air, it is degrading in effect so as to impair the water, soil, or air's usefulness or render it offensive. |
| Suffolk County Government. 2006. «Glossary of Aquatic Terms». https://web.archive.org/web/20060222110516/http://www.co.suffolk.ny.us/webtemp3.cfm?dept=18&id=812 | WAT | Any **substance** or **mixture of substances** that defile or contaminate the soil, water or atmosphere. |
| Johnson County Wastewater. 2013. «Glossary». https://web.archive.org/web/20131106005350/http://www.jcw.org/edglossary.htm | WAT | Any **substance** suspended or dissolved in water that builds up in sufficient quantity to impair water quality. |

| Fuente | Dominio | Definición |
|---|---|---|
| Coastwide Laboratories. «Glossary of Terms and Abbreviations Related to Green Chemistry, Green Cleaning and Community Sustainability». http://www.coastwidelabs.com/Technical%20Articles/Green%20Cleaning%20Glossary.htm | ENV | Any **substance** that directly or indirectly creates an adverse human health or environmental effect when introduced into any environmental media. |
| Galveston Bay Estuary Program. 2007. «Glossary». https://web.archive.org/web/20070731041739/http://www.gbep.state.tx.us/glossary/glossary2.asp | WAT | Any **substance**, as certain chemicals or waste products, that renders the air, soil, water, or other natural resource harmful or unsuitable for a specific purpose. The term "Pollution" also includes such things as impairment to habitat and barriers to fish passage. |
| San Joaquin County Public Works Department. 2006. «Glossary of terms». https://web.archive.org/web/20060520180333/http://207.104.50.39/pubworks/storm%20water/GlossaryofTerms.htm | ENV | Generally are **substances** introduced into the environment that adversely affect the usefulness of a resource. |
| Biodiversity and Human Health. 2010. «Glossary of Ecology Terms». https://web.archive.org/web/20100601052655/http://www.wms.org/biod/library/glos_NS.html | ENV | A particular **chemical** or **form of energy** that can adversely affect the health, survival, or activities of humans or other living organisms. |
| Carolina Environmental Diversity Exploration. 2006. «Glossary». https://web.archive.org/web/20060820211529/http://www.learnpress.org/editions/cede_capefear/glossary | ENV | **Substance** (such as a waste material) that pollutes or contaminates the air, soil, or water and can damage the environment. |
| IUPAC. 2007. «Glossary of Terms Relatings to Pesticides». https://web.archive.org/web/20070924195412/http://www.iupac.org/reports/1996/6805holland/o-p1.html | ENV | **Undesirable substance** introduced into a solid, liquid or gaseous environmental medium totally or partially by human activities. See also contaminant. (after Duffus, 1993) |

| Fuente | Dominio | Definición |
|---|---|---|
| Coalition Opposed to PCB. 2008. «Glossary». https://web.archive.org/web/20080907001809/http://www.copa.org/library/glossary/p.htm | ENV | **Undesirable substances** discharged to soil or water, making the receiving body unfit for use. |
| California Association of Sanitation Agencies. 2006. «Glossary». https://web.archive.org/web/20060820211529/http://www.learnpress.org/editions/cede_capefear/glossary | ENV | **Waste** discharged into water, including dredged soil, solid waste, incinerator residue, sewage, garbage, sewage sludge, munitions, chemical wastes, biological materials, radioactive materials, discarded equipment, and rock. |
| Waste Business Journal. 2015. «Glossary of Waste Terms». http://www.wastebusinessjournal.com/glossary6.htm#P | WAS | A **contaminant** that adversely alters the physical, chemical, or biological properties of the environment. |

# ANEXO 4. CONTEXTÓNIMOS DE *POLLUTANT*

| GENERAL MEDIOAMBIENTAL | | GESTIÓN DE LA CALIDAD DEL AIRE | |
|---|---|---|---|
| Contextónimo | Frecuencia | Contextónimo | Frecuencia |
| air | 2096 | air | 661 |
| source | 927 | emission | 365 |
| emission | 887 | source | 319 |
| pollution | 788 | concentration | 306 |
| water | 787 | pollution | 274 |
| concentration | 683 | effect | 252 |
| effect | 588 | use | 250 |
| use | 554 | quality | 232 |
| quality | 551 | exposure | 226 |
| exposure | 462 | control | 210 |
| include | 434 | level | 210 |
| health | 415 | atmospheric | 198 |
| control | 396 | health | 197 |
| level | 396 | include | 184 |
| gas | 394 | associate | 167 |
| area | 377 | gas | 164 |
| standard | 355 | standard | 153 |
| atmosphere | 338 | area | 142 |
| system | 312 | atmosphere | 140 |
| high | 310 | major | 139 |
| chemical | 304 | ambient | 134 |
| plant | 290 | o | 133 |
| atmospheric | 289 | high | 125 |
| process | 284 | particle | 123 |
| produce | 274 | produce | 121 |
| cause | 273 | hazardous | 116 |
| carbon | 261 | require | 114 |
| waste | 261 | system | 114 |
| associate | 253 | emit | 109 |
| model | 253 | occur | 107 |
| primary | 248 | condition | 106 |
| time | 241 | dispersion | 102 |
| particle | 235 | factor | 102 |
| o | 233 | process | 102 |
| ozone | 232 | dioxide | 101 |
| reduce | 231 | result | 100 |

| GENERAL MEDIOAMBIENTAL | | GESTIÓN DE LA CALIDAD DEL AIRE | |
|---|---|---|---|
| toxic | 230 | time | 100 |
| major | 227 | significant | 97 |
| hazardous | 225 | ozone | 96 |
| treatment | 220 | cause | 94 |
| dioxide | 219 | increase | 94 |
| require | 215 | value | 93 |
| nitrogen | 207 | vehicle | 92 |
| condition | 204 | carbon | 92 |
| ambient | 203 | program | 91 |
| factor | 202 | describe | 88 |
| dispersion | 200 | risk | 87 |
| rate | 199 | s | 87 |
| increase | 196 | so | 85 |
| environmental | 195 | reduce | 84 |
| industrial | 192 | regulatory | 83 |
| oxide | 190 | co | 83 |
| environment | 190 | new | 83 |
| remove | 188 | environment | 83 |
| surface | 188 | primary | 82 |
| problem | 187 | substance | 82 |
| sulfur | 185 | rate | 82 |
| occur | 185 | plant | 82 |
| emit | 184 | human | 81 |
| organic | 184 | problem | 81 |
| substance | 183 | determine | 80 |
| human | 182 | low | 79 |
| risk | 180 | use | 79 |
| secondary | 178 | activity | 78 |
| reaction | 178 | criterion | 77 |
| soil | 178 | nitrogen | 77 |
| so | 170 | monitoring | 76 |
| use | 169 | pm | 76 |
| program | 168 | motor | 76 |
| epa | 166 | reaction | 76 |
| point | 166 | datum | 75 |
| table | 165 | concern | 74 |
| large | 165 | control | 73 |
| matter | 164 | volume | 72 |
| describe | 164 | chemical | 72 |
| co | 164 | environmental | 71 |

| GENERAL MEDIOAMBIENTAL | | GESTIÓN DE LA CALIDAD DEL AIRE | |
|---|---|---|---|
| urban | 163 | model | 71 |
| low | 163 | regulate | 70 |
| result | 161 | sulfur | 70 |
| study | 161 | case | 70 |
| vehicle | 160 | study | 70 |
| s | 159 | see | 70 |
| reduction | 156 | provide | 70 |
| value | 155 | requirement | 69 |
| provide | 155 | specific | 69 |
| example | 155 | affect | 67 |
| combustion | 153 | change | 67 |
| see | 153 | plume | 66 |
| plume | 152 | surface | 66 |
| determine | 152 | particulate | 65 |
| amount | 151 | result | 64 |
| fuel | 149 | example | 64 |
| river | 146 | deposition | 63 |
| particulate | 145 | transport | 63 |
| affect | 145 | h | 63 |
| method | 145 | toxic | 62 |
| limit | 142 | limit | 62 |
| compound | 142 | combustion | 62 |
| new | 142 | reduction | 62 |
| activity | 140 | release | 60 |

| GESTIÓN DE RESIDUOS | | ABASTECIMIENTO Y TRATAMIENTO DE AGUA | |
|---|---|---|---|
| Contextónimo | Frecuencia | Contextónimo | Frecuencia |
| air | 48 | water | 84 |
| waste | 39 | wastewater | 47 |
| water | 28 | treatment | 47 |
| compound | 26 | sewage | 36 |
| pollution | 26 | industrial | 36 |
| problem | 24 | use | 33 |
| source | 24 | discharge | 32 |
| material | 24 | system | 31 |
| include | 24 | pollution | 29 |
| use | 23 | concentration | 29 |

| GESTIÓN DE RESIDUOS | | ABASTECIMIENTO Y TRATAMIENTO DE AGUA | |
|---|---|---|---|
| organic | 18 | remove | 27 |
| reduce | 18 | quality | 27 |
| treatment | 17 | source | 27 |
| area | 17 | conventional | 26 |
| release | 16 | epa | 26 |
| release | 16 | waste | 26 |
| cause | 16 | include | 25 |
| system | 16 | plant | 25 |
| rain | 15 | discharge | 24 |
| primary | 15 | point | 24 |
| nitrogen | 15 | chemical | 24 |
| matter | 15 | treat | 23 |
| emission | 15 | table | 23 |
| reaction | 15 | organic | 23 |
| toxic | 14 | level | 22 |
| local | 14 | nonconventional | 20 |
| atmosphere | 14 | municipal | 20 |
| surface | 14 | industry | 20 |
| biological | 13 | biological | 20 |
| metal | 13 | amount | 20 |
| type | 13 | limit | 19 |
| process | 13 | meet | 18 |
| sulfur | 12 | toxic | 17 |
| secondary | 12 | standard | 17 |
| urban | 12 | program | 17 |
| remove | 12 | heavy | 16 |
| quality | 12 | sludge | 16 |
| environmental | 12 | body | 16 |
| carbon | 12 | variety | 15 |
| concentration | 12 | reduce | 15 |
| provide | 12 | require | 15 |
| example | 12 | rate | 15 |
| plant | 12 | bod | 14 |
| particulate | 11 | characteristic | 14 |
| form | 11 | requirement | 14 |
| s | 11 | metal | 14 |
| gas | 11 | oxygen | 14 |
| dioxin | 10 | solid | 14 |
| add | 10 | total | 14 |
| generate | 10 | call | 14 |

| GESTIÓN DE RESIDUOS | | ABASTECIMIENTO Y TRATAMIENTO DE AGUA | |
|---|---|---|---|
| scale | 10 | cause | 14 |
| result | 10 | use | 14 |
| common | 10 | high | 14 |
| amount | 10 | process | 14 |
| increase | 10 | npdes | 13 |
| produce | 10 | permit | 13 |
| large | 10 | domestic | 13 |
| chemical | 10 | pathogen | 13 |
| time | 10 | design | 13 |
| energy | 10 | generate | 13 |
| monoxide | 9 | vary | 13 |
| volatile | 9 | surface | 13 |
| hydrocarbon | 9 | priority | 12 |
| contaminant | 9 | loading | 12 |
| treat | 9 | runoff | 12 |
| oxide | 9 | establish | 12 |
| industrial | 9 | act | 12 |
| facility | 9 | disposal | 12 |
| wastewater | 9 | load | 12 |
| solid | 9 | apply | 12 |
| involve | 9 | river | 12 |
| allow | 9 | category | 11 |
| several | 9 | national | 11 |
| term | 9 | facility | 11 |
| form | 9 | base | 11 |
| contain | 9 | effect | 11 |
| environment | 9 | type | 11 |
| condition | 9 | pretreatment | 10 |
| land | 9 | cfr | 10 |
| heat | 9 | suspend | 10 |
| effect | 9 | sewer | 10 |
| soil | 9 | process | 10 |
| incineration | 8 | wide | 10 |
| numerous | 8 | demand | 10 |
| runoff | 8 | removal | 10 |
| storm | 8 | city | 10 |
| regulation | 8 | good | 10 |
| ozone | 8 | control | 10 |
| chemical | 8 | large | 10 |
| discharge | 8 | area | 10 |

| GESTIÓN DE RESIDUOS | | ABASTECIMIENTO Y TRATAMIENTO DE AGUA | |
|---|---|---|---|
| associate | 8 | potw | 9 |
| apply | 8 | discharger | 9 |
| depend | 8 | nonpoint | 9 |
| consider | 8 | effluent | 9 |
| new | 8 | list | 9 |
| result | 8 | agency | 9 |
| management | 8 | ph | 9 |
| control | 8 | practice | 9 |
| small | 8 | percent | 9 |
| o | 8 | country | 9 |

# ANEXO 5. DEFINICIONES DE CHLORINE

| **Fuente** | **Dominio** | **Definición** |
|---|---|---|
| European Environment Agency. 2015. «Environmental Terminology and Discovery Service». http://glossary.eea.europa.eu/. | ENV | A very reactive and highly toxic green, **gaseous element**, belonging to the halogen family of substances. It is one of the most widespread elements, as it occurs naturally in sea-water, salt lakes and underground deposits, but usually occurs in a safe form as common salt (NaCl). Commercially it is used in large quantities by the chemical industry both as an element to produce chlorinated organic solvents, like polychlorinated biphenyls (PCBs), and for the manufacture of polyvinyl chloride plastics, thermoplastic and hypochlorite bleaches. Chlorine was the basis for the organochlorine pesticides, like DDT and other agricultural chemicals that have killed wildlife. The reactivity of chlorine has proved disastrous for the ozone layer and has been the cause of the creation of the ozone hole, which was first detected in the Southern Hemisphere over Antarctica and then over the Northern Hemisphere. |
| FAO. 2015. «FAO TERM portal». http://www.fao.org/faoterm/en/. | ENV | The **chemical element** often used as a biocide in the cooling waters of power plants (mainly those using brackish and marine waters as coolant), and as disinfectant in aquaculture (especially in quarantine stations and hatcheries). |
| Translation Bureau / Bureau de la traduction [Canada]. 2015. «TERMIUM Plus». http://www.btb.termiumplus.gc.ca/. | ENV | Chlorine [is] an **element** that is found in biological tissues as the chloride ion. The body contains about 100g of chloride and the average diet contains 6-7 g, mainly as sodium chloride. Free chlorine is used as a sterilising agent e.g. in drinking water. |
| Park, Chris y Michael Allaby. 2013. *A Dictionary of Environment and Conservation*. 2nd ed. Oxford: Oxford University Press. | ENV | A highly reactive **halogen gas** that is added to drinking water to kill bacteria and algae, and used as a raw material for products such as plastics, pharmaceuticals, and pesticides. It is a very toxic biocide, poisonous to fish and invertebrates, persistent in the environment, and important in the destruction of ozone. |
| Allaby, Michael. 2010. *A Dictionary of Ecology*. Oxford: Oxford University Press. | ENV | An **element** that is necessary for normal plant growth. Its main role seems to be that of controlling turgor, but it may also be involved in the light reaction of photosynthesis. If plants are deficient in chloride ions, wilting occurs and the young leaves become blue-green and shiny. Later they become bronze-coloured and chlorotic (see chlorosis). |
| King, Robert C., Pamela K. Mulligan y William D. Stansfield. 2014. *A Dictionary of Genetics*. 8.a ed. Oxford: Oxford University Press. | ENV | An **element** universally found in small amounts in tissues. Atomic number 17; atomic weight 35.453; valence 1–; most abundant isotope 35Cl; radioisotopes 33Cl; half-life 37 minutes; 39Cl, half-life 55 minutes; radiation emitted—beta particles. |

| Fuente | Dominio | Definición |
|---|---|---|
| Euro Chlor. 2015. «Chlorine Glossary». http://www.eurochlor.org/facts-figures-glossary/chlorine-glossary.aspx | ENV | In its "normal" state, chlorine is a **greenish yellow gas**, but at -34°C it turns to a liquid. It is the eleventh most common element in the earth's crust and is widespread in nature. Chlorine is a key building block of modern chemistry and used in three principal ways: direct use (e.g. to disinfect water); as a raw material for chlorine-containing products (e.g. plastics, pharmaceuticals, pesticides) and as an intermediate to manufacture non-chlorinated products (eg polycarbonates and polyurethanes) |
| Cammack, Richard, Teresa Atwood, Peter Campbell, Howard Parish, Anthony Smith, Frank Vella y John Stirling. 2008. *Oxford Dictionary of Biochemistry and Molecular Biology*. Oxford: Oxford University Press. | CHE | A **halogen element** of group 17 of the IUPAC periodic table; atomic number 17; relative atomic mass 35.453. The 15th most abundant element of the Earth's crust, occurring in the combined state, mostly as inorganic chlorides. |
| Daintith, Jonh. 2014. *A Dictionary of Chemistry*. 6.a ed. Oxford: Oxford University Press. | CHE | A **halogen element**; a.n. 17; r.a.m. 35.453; d. 3.214 g dm−3; m.p. −100.98°C; b.p. −34.6°C. It is a poisonous greenish-yellow gas and occurs widely in nature as sodium chloride in seawater and as halite (NaCl), carnallite (KCl.MgCl2.6H2O), and sylvite (KCl). It is manufactured by the electrolysis of brine and also obtained in the Downs process for making sodium. It has many applications, including the chlorination of drinking water, bleaching, and the manufacture of a large number of organic chemicals.<br><br>It reacts directly with many elements and compounds and is a strong oxidizing agent. Chlorine compounds contain the element in the 1, 3, 5, and 7 oxidation states. It was discovered by Karl Scheele in 1774 and Humphry Davy confirmed it as an element in 1810. |
| Anglian Water. «Glossary». https://www.anglianwater.co.uk/_assets/media/Glossary.pdf | WAT | A **gas** which is added to domestic water supplies to kill bacteria |
| Atmospheric Research and Information Centre (UK). 2007. «Glossary». https://web.archive.org/web/20070715090732/http://www.ace.mmu.ac.uk/Resources/Fact_Sheets/Key_Stage_3/Global_Warming/glossary.html | ENV | An **element** that is poisonous in large amounts, and usually found as a greenish-yellow gas. |

| Fuente | Dominio | Definición |
|---|---|---|
| Australian Green Procurement 2005. «Glossary». https://web.archive.org/web/20060505015636/http://www.greenprocurement.org/database/glossary.jsf | ENV | **Poisonous element** with strong odour, and a powerful bleach and disinfectant. Known respiratory irritant. |
| Indian and Northern Affairs Canada. 2006. «Glossary». https://web.archive.org/web/20080315063646/http://ainc-inac.gc.ca/ncp/pub/hig/hig27_e.html | ENV | Chlorine is a naturally occurring, poisonous greenishyellow **non-metallic gas** used to purify water, for bleaching and in the manufacture of many organic chemicals. It occurs naturally only as a component of salt, e.g., in sea water. |
| Center for Global Environmental Education (Hamline University Graduate School of Education). 2000. «Glossary». https://web.archive.org/web/20071006063959/http://cgee.hamline.edu/waters2thesea/glossary2.htm | ENV | A **halogen element** that is isolated as a heavy greenish yellow gas of pungent odor and is used especially as a bleach, oxidizing agent, and disinfectant in water purification |
| ESP Water Products. 2015. «Glossary of Water Purification and Water Filtration Terms» https://www.espwaterproducts.com/water-dictionary/ | WAT | A **gas** widely used in the disinfection of water and as an oxidizing agent for organic matter, manganese, iron, and hydrogen sulfide. |
| Greenpeace Ozone Layer. 2012. «Glossary». https://web.archive.org/web/20120310035842/http://archive.greenpeace.org/ozone/chlorine/6chlor.html | AIR | A **halogen** that depletes ozone when released into the stratosphere. Of the chemical groups covered by this report, CFCs, HCFCs and methyl chloroform release chlorine into the stratosphere. |
| H2o University (Southern Nevada Water Authority). 2015. «Glossary». http://www.h2ouniversity.org/html/library_glossary_af.html | WAT | A **chemical** used to clean water. |
| Hydro Polymers. 2008. «Glossary». https://web.archive.org/web/20051001005223/http://www.hydropolymers.com/en/media_room/glossary/ | ENV | One of the most common **chemical elements** which can be derived from salt using by electrolysis. Used as a raw material for the production of PVC. Also used in the manufacture of a range of solvents, pharmaceutical preparations, insecticides, weed killers and other polymers. Used as a bleaching agent, and as a disinfectant e.g. in drinking water and in swimming pools |

| Fuente | Dominio | Definición |
| --- | --- | --- |
| Michigan Environmental Education Curriculum Drinking Water Treatment. «Glossary». http://techalive.mtu.edu/meec/module03/Glossary.htm | WAT | A **chemical** added to drinking water and wastewater to kill disease-causing organisms. |
| Natural Resources Defense Council. «Glossary of Environmental Terms». http://www.nrdc.org/reference/glossary/c.asp | ENV | A highly reactive **halogen element**, used most often in the form of a pungent gas to disinfect drinking water. |
| North Wales Water Authority. «Glossary». http://www.nwwater.com/index.cfm/nodeID/ab8731e8-2fb8-4f99-9c19-855a210d8282/fuseaction/showContent.page | WAT | A liquid or gas **chemical** that is used as a disinfectant in the drinking water treatment process. |
| NRV Regional Water Authority. «Glossary». http://www.h2o4u.org/education/glossary.shtml | WAT | An **element** ordinarily existing as a greenish-yellow gas about 2.5 times as heavy as air. |
| US Department of State. 2001. «The Ozone Layer. Glossary of important terms». https://web.archive.org/web/20010912011548/http://exchanges.state.gov/forum/journal/env8appendix.htm | AIR | A **greenish-yellow gas** that has a strong odor and is often used to disinfect water |
| Portland Water Bureau. 2013. «Glossary of Water Terms». https://web.archive.org/web/20130511045452/http://www.portlandoregon.gov/water/article/327617 | WAT | A **chemical** which destroys small organisms in water. |
| South Central Connecticut Regional Water Authority. 2002. «Glossary of Water Terms». https://web.archive.org/web/20070103083943/http://www.rwater.com/glossary/gloss_al.htm | WAT | A **chemical** used to kill bacteria in the water. |
| The Watershed. «Glossary». http://www.h2ohero.org/landing/glossary.htm | WAT | A widely used **disinfectant** in waste treatment plants, drinking water, and swimming pools. |
| Water FAQs. 2007. «Glossary of Water-related Terms». https://web.archive.org/web/20070302233536/http://www.waterfaqs.com/water-glossary.html | WAT | A **chemical gas** that is commonly added to tap water to insure bacteriological safety. |

| Fuente | Dominio | Definición |
|---|---|---|
| City of Tucson. 2015. «Water Quality Terms and Definitions». https://www.tucsonaz.gov/water/terms-and-definitions | WAT | Chlorine is the most widely used **drinking water disinfectant** in North America. Adding a small amount of chlorine to drinking water protects the water from bacteria and other microorganisms. Chlorine is added to drinking water as either a gas or after having been already dissolved in water. Chlorine is measured in the field using portable analytical instrumentation. In order to meet Federal and State microbiological drinking water regulations, it is Tucson Water policy to ensure that there is a detectable amount of chlorine in 95 percent of the samples taken from the water distribution system. To optimize microbiological protection and minimize taste and odor problems, Tucson Water works to maintain a chlorine level between 0.6 parts per million and 1.0 parts per million throughout the distribution system. When the level of chlorine is above 0.5 parts per million, many people can smell chlorine in the water. |
| UNICEF. 2010. «Glossary. Water, environment and sanitation». https://web.archive.org/web/20100610093225/http://www.unicef.org/voy/explore/wes/explore_1870.html | WAT | A **chemical** that is added to water to kill off some germs and make water safe for drinking. |

# ANEXO 6. CONTEXTÓNIMOS DE *CHLORINE*

| GENERAL MEDIOAMBIENTAL | | GESTIÓN DE LA CALIDAD DEL AIRE | |
|---|---|---|---|
| Contextónimo | Frecuencia | Contextónimo | Frecuencia |
| water | 492 | ozone | 44 |
| ozone | 221 | stratosphere | 38 |
| use | 215 | cfc | 34 |
| cl | 185 | atom | 33 |
| gas | 183 | atmosphere | 28 |
| atom | 181 | stratospheric | 27 |
| o | 175 | compound | 26 |
| reaction | 164 | source | 26 |
| disinfection | 163 | cl | 23 |
| sodium | 140 | reaction | 23 |
| compound | 138 | o | 22 |
| chemical | 130 | depletion | 20 |
| treatment | 110 | reach | 20 |
| pcb | 109 | bromine | 18 |
| molecule | 102 | low | 17 |
| acid | 102 | time | 16 |
| carbon | 102 | chemical | 15 |
| system | 101 | hcl | 14 |
| form | 96 | monoxide | 14 |
| stratosphere | 95 | remove | 14 |
| l | 93 | molecule | 14 |
| oxygen | 91 | natural | 14 |
| use | 89 | destruction | 13 |
| chloride | 87 | destroy | 13 |
| concentration | 87 | release | 13 |
| chloramine | 86 | atmospheric | 13 |
| dioxide | 86 | cycle | 13 |
| hydrogen | 84 | catalytic | 12 |
| ion | 83 | nitrate | 12 |
| contain | 83 | specie | 12 |
| low | 83 | concentration | 12 |
| source | 82 | man-made | 11 |
| hypochlorite | 81 | reservoir | 11 |
| chlorination | 81 | form | 11 |
| produce | 81 | gas | 11 |
| time | 81 | use | 11 |

| **GENERAL MEDIOAMBIENTAL** | | **GESTIÓN DE LA CALIDAD DEL AIRE** | |
|---|---|---|---|
| bromine | 80 | chlorofluorocarbon | 10 |
| electron | 78 | fluorine | 10 |
| method | 77 | clo | 10 |
| number | 77 | light | 10 |
| react | 74 | amount | 10 |
| element | 73 | clono | 9 |
| process | 73 | antarctic | 9 |
| cfc | 72 | spring | 9 |
| disinfectant | 71 | cloud | 9 |
| n | 70 | involve | 9 |
| oxidation | 69 | large | 9 |
| c | 69 | tropospheric | 8 |
| mg | 68 | react | 8 |
| product | 68 | upper | 8 |
| high | 68 | significant | 8 |
| table | 67 | hydrogen | 8 |
| solution | 66 | measure | 8 |
| organic | 66 | acid | 8 |
| level | 66 | carbon | 8 |
| reduce | 63 | emission | 8 |
| g | 62 | value | 8 |
| include | 62 | year | 8 |
| contact | 61 | level | 8 |
| supply | 61 | misconception | 7 |
| wastewater | 61 | hcfcs | 7 |
| cost | 61 | spray | 7 |
| remove | 60 | hole | 7 |
| h | 60 | catalyst | 7 |
| disinfect | 58 | observation | 7 |
| plant | 58 | break | 7 |
| ppm | 57 | second | 7 |
| clo | 56 | present | 7 |
| stratospheric | 55 | affect | 7 |
| example | 55 | allow | 7 |
| oxidize | 52 | know | 7 |
| chlorinate | 51 | contain | 7 |
| amount | 51 | condition | 7 |
| large | 51 | show | 7 |
| waste | 51 | occur | 7 |
| ammonia | 50 | effect | 7 |

| GENERAL MEDIOAMBIENTAL | | GESTIÓN DE LA CALIDAD DEL AIRE | |
|---|---|---|---|
| form | 50 | air | 7 |
| find | 50 | irrelevant | 6 |
| figure | 49 | long-lived | 6 |
| drinking | 48 | one | 6 |
| treat | 48 | sink | 6 |
| add | 48 | related | 6 |
| common | 48 | accept | 6 |
| occur | 48 | efficient | 6 |
| temperature | 48 | polar | 6 |
| air | 48 | true | 6 |
| material | 47 | methane | 6 |
| agent | 46 | demonstrate | 6 |
| release | 46 | chloride | 6 |
| atmosphere | 46 | decrease | 6 |
| require | 46 | contribute | 6 |
| toxic | 45 | chemical | 6 |
| effective | 45 | oxygen | 6 |
| metal | 45 | production | 6 |
| s | 45 | result | 6 |
| fluorine | 44 | cause | 6 |
| addition | 44 | number | 6 |
| show | 44 | produce | 6 |
| increase | 44 | example | 6 |
| pressure | 44 | temperature | 6 |

| QUÍMICA | | ABASTECIMIENTO Y TRATAMIENTO DE AGUA | |
|---|---|---|---|
| Contextónimo | Frecuencia | Contextónimo | Frecuencia |
| cl | 44 | water | 128 |
| element | 30 | use | 65 |
| gas | 29 | disinfection | 64 |
| o | 27 | use | 47 |
| oxidation | 23 | gas | 47 |
| oxygen | 22 | hypochlorite | 37 |
| g | 22 | treatment | 37 |
| fluorine | 21 | sodium | 36 |
| molecular | 21 | method | 36 |
| reaction | 21 | system | 34 |

| QUÍMICA | | ABASTECIMIENTO Y TRATAMIENTO DE AGUA | |
|---|---|---|---|
| sodium | 20 | dioxide | 31 |
| n | 20 | chloramine | 30 |
| bromine | 19 | chlorination | 29 |
| chloride | 19 | supply | 29 |
| electron | 17 | kill | 28 |
| hydrogen | 17 | disinfectant | 27 |
| atom | 17 | purification | 27 |
| h | 17 | treat | 27 |
| example | 17 | common | 27 |
| compound | 14 | filter | 26 |
| acid | 14 | chemical | 26 |
| oxidize | 13 | disinfect | 25 |
| oxide | 13 | solution | 25 |
| metal | 13 | free | 23 |
| form | 13 | mg | 23 |
| number | 13 | tank | 23 |
| s | 13 | addition | 23 |
| figure | 13 | form | 22 |
| ion | 12 | contact | 21 |
| co | 12 | acid | 21 |
| mass | 12 | l | 21 |
| reduce | 12 | require | 21 |
| chemical | 12 | time | 21 |
| potassium | 11 | residual | 20 |
| l | 11 | bacterium | 20 |
| same | 11 | add | 20 |
| carbon | 11 | technique | 20 |
| state | 11 | distribution | 20 |
| use | 11 | remove | 20 |
| sulfur | 10 | include | 20 |
| nitrogen | 10 | by-product | 19 |
| consider | 10 | react | 19 |
| p | 10 | wastewater | 19 |
| contain | 10 | low | 19 |
| problem | 10 | effective | 18 |
| produce | 10 | compound | 18 |
| c | 10 | form | 18 |
| water | 10 | process | 18 |
| clo | 9 | toxic | 17 |
| iodine | 9 | concentration | 17 |

| QUÍMICA | | ABASTECIMIENTO Y TRATAMIENTO DE AGUA | |
|---|---|---|---|
| calcium | 9 | ozone | 16 |
| edit | 9 | storage | 16 |
| industrial | 9 | level | 16 |
| follow | 9 | dissolve | 15 |
| different | 9 | strong | 15 |
| halogen | 8 | service | 15 |
| silicon | 8 | allow | 15 |
| br | 8 | united | 15 |
| agent | 8 | become | 15 |
| formula | 8 | large | 15 |
| calculate | 8 | residual | 14 |
| increase | 8 | pathogen | 14 |
| f | 8 | first | 14 |
| solution | 8 | base | 14 |
| find | 8 | states | 14 |
| table | 8 | organic | 14 |
| value | 8 | thms | 13 |
| show | 8 | drink | 13 |
| see | 8 | avoid | 13 |
| pressure | 8 | few | 13 |
| process | 8 | problem | 13 |
| electronegative | 7 | reduce | 13 |
| argon | 7 | plant | 13 |
| electronegativity | 7 | bleach | 12 |
| aluminium | 7 | harmful | 12 |
| magnesium | 7 | minute | 12 |
| prepare | 7 | membrane | 12 |
| phosphorus | 7 | salt | 12 |
| react | 7 | sand | 12 |
| say | 7 | need | 12 |
| mole | 7 | available | 12 |
| bond | 7 | develop | 12 |
| name | 7 | source | 12 |
| charge | 7 | haloacetic | 11 |
| similar | 7 | giardia | 11 |
| molecule | 7 | cryptosporidium | 11 |
| group | 7 | dechlorination | 11 |
| k | 7 | liquefy | 11 |
| important | 7 | medical | 11 |
| equation | 7 | compressed | 11 |

| QUÍMICA | | ABASTECIMIENTO Y TRATAMIENTO DE AGUA | |
|---|---|---|---|
| increase | 7 | iodine | 11 |
| temperature | 7 | british | 11 |
| oxoacids | 6 | filter | 11 |
| hclo | 6 | dose | 11 |
| bromide | 6 | lime | 11 |
| aq | 6 | drinking | 11 |
| anion | 6 | agent | 11 |
| gain | 6 | calcium | 11 |
| exercise | 6 | cl | 11 |
| atm | 6 | protection | 11 |